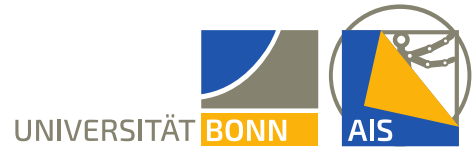

ANALYTIC BIPEDAL WALKING WITH FUSED ANGLES AND
CORRECTIVE ACTIONS IN THE TILT PHASE SPACE

PHILIPP ALLGEUER

Bonn, March 2020

*To the one who wraps me, and to my mother Karin,*
*without whose tireless efforts spanning two decades*
*I would not even be writing this.*

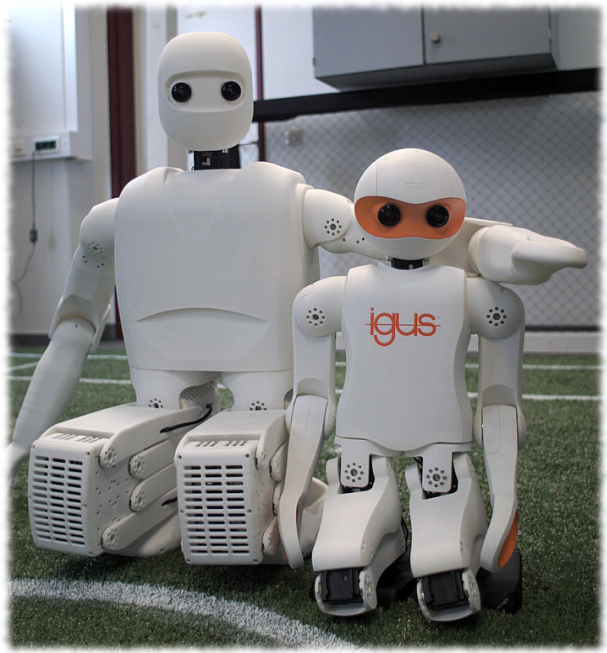

## ABSTRACT

As humanoid robots start to move from research labs to workplace environments and homes, the topic of how they should best and most reliably locomote in the face of unknown disturbances will be a topic of increasing importance. This thesis presents algorithms and methods for the feedback-stabilised walking of bipedal humanoid robotic platforms, along with the underlying theoretical and sensorimotor frameworks required to achieve it. Bipedal walking is inherently complex and difficult to control due to the high level of nonlinearity and significant number of degrees of freedom of the concerned robots, the limited observability and controllability of the corresponding states, and—especially for low-cost robots—the combination of imperfect actuation with less-than-ideal sensing. The methods presented in this thesis deal with these issues in a multitude of ways, ranging from the development of an actuator control and feed-forward compensation scheme, to the implementation of countless sensor calibration, autocalibration and processing schemes, to the inclusion of inherent filtering in almost all of the gait stabilisation feedback pipelines.

Two main gaits are developed and investigated in this work, the direct fused angle feedback gait, and the tilt phase controller. Both gaits follow the design philosophy of internally leveraging a semi-stable open-loop gait generator, and extending it through stabilising feedback via the means of so-called corrective actions. The idea of using corrective actions is to modify the generation of the open-loop joint waveforms in such a way that the balance of the robot is influenced and thereby ameliorated. Examples of such corrective actions include modifications of the arm swing and leg swing trajectories, the application of dynamic positional and rotational offsets to the hips and feet, and adjustments of the commanded step size and timing.

Underpinning both feedback gaits and their corresponding gait generators (the CPG and KGG) are significant advances in the field of 3D rotation theory. These advances in particular include the development of three novel rotation representations, the tilt angles, fused angles, and tilt phase space representations. All three of these representations are founded on a new innovative way of splitting 3D rotations into their respective yaw and tilt components.

All of the algorithms presented in this thesis were implemented as part of the Humanoid Open Platform ROS Software release, and tested on a multitude of real and simulated robots, including in particular the igus Humanoid Open Platform and NimbRo-OP2. The notable walking stability that was achieved critically contributed to Team NimbRo's yearly wins at the international RoboCup competition from 2016 onwards.

# ZUSAMMENFASSUNG

Humanoide Roboter verlassen zunehmend die Forschungslabore und beginnen das alltägliche Leben zu durchdringen, sowohl in Arbeitsumgebungen als auch in Privathaushalten. Angesichts dieser Entwicklung gewinnt die Frage, wie Roboter sich unter dem Einfluss unvorhersehbarer Störungen zuverlässig fortbewegen können, eine immer größere Bedeutung. Die vorliegende Arbeit stellt Methoden, Algorithmen, sowie die dazu erforderlichen theoretischen und sensomotorischen Grundlagen des stabilen Gangs auf zwei Beinen vor. Die Bipedie ist von Natur aus komplex und schwierig zu steuern, da zweibeinige Roboter ein hohes Maß an Nichtlinearität und eine hohe Anzahl an Freiheitsgraden aufweisen. Weitere Schwierigkeiten ergeben sich durch die eingeschränkte Beobachtbarkeit und Kontrollierbarkeit des Zustands, sowie die insbesondere bei kostengünstigen Robotern auftretende Kombination aus unpräziser Motorik und ungenauer Sensorik. Die vorgestellten Algorithmen lösen diese Problematik durch die Kombination verschiedener Techniken, insbesondere durch ein neuartiges Aktuator-Regelungsschema, zahlreiche Sensorkalibrierungs-, Autokalibrierungs- und Prozessschemata, sowie die Anwendung von Filtern in den Rückkopplungspfaden.

In dieser Arbeit werden zwei grundlegende Gangarten entwickelt und untersucht, der „direct fused angle feedback gait" und der „tilt phase controller". Beide Gangarten benutzen einen semistabilen Open-Loop-Ganggenerator, der zur Stabilisierung des Roboters mit Korrekturmaßnahmen erweitert wird. Diese Korrekturmaßnahmen modifizieren die erzeugten Gelenktrajektorien, und beeinflussen dadurch das Gleichgewicht des Roboters. Beispiele für diese Korrekturmaßnahmen stellen die gezielte Abänderung der Bewegungsbahnen der Arme und Beine, sowie die Adaptierung der Schrittgrößen und Zeitabfolgen dar.

Beide Gangarten, sowie die dazugehörigen Ganggeneratoren (CPG und KGG), bauen auf umfangreichen Neuentwicklungen auf dem Gebiet der 3D-Rotationstheorie auf, die ebenfalls in dieser Arbeit vorgestellt werden. Diese umfassen insbesondere die Entwicklung von drei neuartigen Rotationsdarstellungen, die „tilt angles", „fused angles" und „tilt phase space" Darstellungen. Alle drei basieren auf einem neuen, innovativen Ansatz zur Aufteilung von Rotationen im dreidimensionalen Raum in Gier- und Neigungskomponenten.

Alle in dieser Arbeit vorgestellten Algorithmen wurden in der Humanoid Open Platform ROS Software implementiert und auf mehreren humanoiden Robotern praktisch erprobt, unter anderem auch auf der igus Humanoid Open Platform. Die hohe Stabilität des damit erzielten Gangs trug maßgeblich zu den zahlreichen Siegen des Teams NimbRo bei den jährlichen RoboCup Wettbewerben ab 2016 bei.

## ACKNOWLEDGEMENTS

I would first of all like to thank Lynne Winkler for her unquestionable support in all aspects of my life, be it robots, RoboCup, climbing, Ninja Warrior competitions, or anything else. I would also like to thank Marcell Missura for inducting me into the world of bipedal walking, Hafez Farazi for the hard work we shared preparing for those many RoboCups, and Grzegorz Ficht for building the NimbRo-OP2 and NimbRo-OP2X robots, with the help of André Brandenburger, who also kept the robots in good shape for whenever I needed them. I would also like to thank Michael Schreiber for his general mechanical and mechatronic support, Max Schwarz for helping me whenever I had a particularly fiendish Linux problem or bug, and Prof. Dr. Sven Behnke for the opportunity to pursue my research interests and compete at the RoboCup competitions.

x

# CONTENTS

## LIST OF FIGURES

## LIST OF TABLES

## LIST OF ALGORITHMS

# LIST OF ACRONYMS

For more than just the page number of the *first* use of each acronym, please refer to the index on page 521, in which the page numbers of all definitions of the acronyms are listed.

## NOTATION

---

**General Conventions**

| | |
|---|---|
| $a, b, c$ | Scalars (regular italic symbols) |
| $\mathbf{a}, \mathbf{b}, \mathbf{c}$ | Vectors (bold upright symbols) |
| $A, B, C$ | Matrices (uppercase regular italic symbols) |
| $\mathbf{A}, \mathbf{B}, \mathbf{C}$ | Gait keypoints (uppercase bold upright symbols) |
| $\mathbb{A}, \mathbb{B}, \mathbb{C}$ | Domain sets (blackboard bold uppercase symbols) |
| $\mathcal{A}, \mathcal{B}, \mathcal{C}$ | General sets (calligraphic uppercase symbols) |
| $\alpha, \beta, \gamma$ | Angles (lowercase regular italic Greek letters) |
| {A}, {B}, {C} | Cartesian coordinates frames in 3D space |

**Rotation Representations**

| | |
|---|---|
| $R = [R_{ij}] \in \mathrm{SO}(3)$ | Rotation matrix representation |
| $q = (w, x, y, z)$ | Quaternion representation (unit norm) |
| $\quad = (q_0, \mathbf{q}) \in \mathbb{Q} \subset \mathbb{H}$ | where $q_0 = w$ and $\mathbf{q} = (x, y, z)$ |
| $T = (\psi, \gamma, \alpha) \in \mathbb{T}$ | Tilt angles representation |
| $F = (\psi, \theta, \phi, h) \in \mathbb{F}$ | Fused angles representation |
| $E = (\psi_E, \theta_E, \phi_E) \in \mathbb{E}$ | ZYX Euler angles representation |
| $\tilde{E} = (\psi_{\tilde{E}}, \phi_{\tilde{E}}, \theta_{\tilde{E}}) \in \tilde{\mathbb{E}}$ | ZXY Euler angles representation |
| $P = (p_x, p_y, p_z) \in \mathbb{P}^3$ | Relative tilt phase space representation |
| $\tilde{P} = (\tilde{p}_x, \tilde{p}_y, \tilde{p}_z) \in \tilde{\mathbb{P}}^3$ | Absolute tilt phase space representation |
| $A = (\hat{\mathbf{e}}, \theta_a) \in \mathbb{A}$ | Axis-angle representation |

| | | | |
|---|---|---|---|
| $\psi$ | Fused yaw | $\theta$ | Fused pitch |
| $\gamma$ | Tilt axis angle | $\phi$ | Fused roll |
| $\alpha$ | Tilt angle | $h$ | Hemisphere |
| $\psi_E$ | ZYX Euler yaw | $\psi_{\tilde{E}}$ | ZXY Euler yaw |
| $\theta_E$ | ZYX Euler pitch | $\phi_{\tilde{E}}$ | ZXY Euler roll |
| $\phi_E$ | ZYX Euler roll | $\theta_{\tilde{E}}$ | ZXY Euler pitch |
| $p_x$ | Relative tilt phase x | $\tilde{p}_x$ | Absolute tilt phase x |
| $p_y$ | Relative tilt phase y | $\tilde{p}_y$ | Absolute tilt phase y |
| $p_z$ | Relative tilt phase z | $\tilde{p}_z$ | Absolute tilt phase z |

| | |
|---|---|
| $\square^{-1}$ | Inverse rotation of $\square$, where $\square$ is $E, F, T, P, \ldots$ |
| $R^T$ | Matrix transpose, equivalent to $R^{-1}$ for rotation matrices |
| $q^*$ | Quaternion conjugate, equivalent to $q^{-1}$ for unit quaternions, i.e. quaternion rotations |
| $L_q(\mathbf{v})$ | The resulting vector of $\mathbf{v}$ rotated by quaternion $q$ |

$R_x(\cdot)$     The rotation matrix corresponding to a counterclockwise rotation by $\cdot$ radians about the x-axis. Similarly for other representations and axes, e.g. $q_y(\cdot)$, $F_z(\cdot)$, …

$R_{\mathbf{v}}(\cdot)$     The rotation matrix corresponding to a counterclockwise rotation by $\cdot$ radians about the vector $\mathbf{v}$. Similarly for other representations, e.g. $q_{\mathbf{v}}(\cdot)$, $T_{\mathbf{v}}(\cdot)$, …

$\Psi(\square)$     The fused yaw of the rotation $\square$

$\Theta(p, q)$     The smaller of the two angles between the 4D lines given by $\pm p$ and $\pm q$ (the 'quaternion lines' $p$ and $q$)

## Vectors, Rotations and Bases

$\mathbf{v} = (v_x, v_y, v_z)$     Unless otherwise specified, vector components are given by adding $x$, $y$ and $z$ subscripts

$\mathbf{\Omega} = (\omega_x, \omega_y, \omega_z)$     Angular velocity components are given by $\omega_*$

$^G\mathbf{v}$     A vector expressed in the coordinates of frame {G}

$\mathbf{x}_A$     The x-axis of frame {A} expressed in the coordinates of the global fixed frame {G}. Similarly for other combinations of frames and basis vectors, e.g. $\mathbf{y}_B$, $\mathbf{z}_G$, …

$^B\mathbf{z}_G$     The z-axis of frame {G} expressed in the coordinates of frame {B}. Similarly for other combinations of frames and basis vectors, e.g. $^A\mathbf{x}_B$, $^G\mathbf{y}_B$, …

$^B z_{Gy}$     The y-component of $^B\mathbf{z}_G$ as defined above

$^A_B\square$     The rotation from {A} to {B}, relative to {A}, where $\square$ is $R$, $q$, $F$, … This denotes a standard rotation.

$^{GA}_B\square$     The rotation from {A} to {B}, relative to {G}, where $\square$ is $R$, $q$, $F$, … This denotes a referenced rotation.

$^G\psi$     The fused yaw of a known frame relative to frame {G}

## Matrix, Vector and Rotation Literals

$\begin{bmatrix} \cdot & \cdot & \cdot \end{bmatrix}$     Matrix literal

$(\cdot, \cdot, \cdot)$     Vector literal (often dually treated as column vector)

$E(\cdot, \cdot, \cdot)$     ZYX Euler angles literal, e.g. $E(\frac{\pi}{2}, 0, -\frac{\pi}{4})$ is the ZYX Euler angles representation with $\psi_E = \frac{\pi}{2}$, $\theta_E = 0$ and $\phi_E = -\frac{\pi}{4}$. Similarly for other representations, e.g. $F(\cdot, \cdot, \cdot, \cdot)$ and $T(\cdot, \cdot, \cdot)$. Quaternion literals $(\cdot, \cdot, \cdot, \cdot)$ and $(\cdot, \cdot)$ are written without the preceding '$q$'.

$E_R(\cdot, \cdot, \cdot)$     ZYX Euler angles literal converted to a rotation matrix, e.g. to multiply it by another rotation matrix or vector. Similarly for all other rotation representation combinations, e.g. $F_q(\cdot, \cdot, \cdot, \cdot)$, $T_F(\cdot, \cdot, \cdot)$, …

## Basic Mathematics

| | |
|---|---|
| $\dot{\square}$ | Single time derivative of $\square$, i.e. $\frac{d\square}{dt}$ |
| $\ddot{\square}$ | Double time derivative of $\square$, i.e. $\frac{d^2\square}{dt^2}$ |
| $\mathbf{0}$ | Zero vector in $\mathbb{R}^n$, where $n$ is usually clear from context |
| $\mathbb{I}$ | Identity matrix, with dimensions that are clear from the context or explicitly specified, e.g. $\mathbb{I}_3 \equiv \mathbb{I}_{3\times3}$ |
| $\mathbb{R}^n$ | The n-dimensional real space, e.g. $\mathbb{R}^2$, $\mathbb{R}^3$ |
| $\mathbb{R}^{m\times n}$ | The set of all $m \times n$ real matrices |
| $\mathbb{R}\mathrm{P}^n$ | The n-dimensional real projective space |
| SO(3) | Special orthogonal group of dimension 3, consisting of all $3 \times 3$ real orthogonal matrices of determinant 1 |
| TR(3) | Subset of SO(3), given by all pure tilt rotations |
| $\mathcal{S}^n$ | The unit n-sphere embedded in $\mathbb{R}^{n+1}$, e.g. $\mathcal{S}^1$ is the unit circle in 2D space and $\mathcal{S}^2$ is the unit sphere in 3D space |
| $\bar{\mathcal{D}}^n(\mathbf{x}, r)$ | The filled (including its interior) closed sphere $\mathcal{S}^{n-1}$ of radius $r$ and centre $\mathbf{x}$ in $\mathbb{R}^n$, e.g. $\bar{\mathcal{D}}^2(\mathbf{0}, r)$ is the closed disc of radius $r$ in $\mathbb{R}^2$, centred at the origin $\mathbf{0}$ |
| $\bar{\mathcal{D}}^n(r)$ | Shorthand for $\bar{\mathcal{D}}^n(\mathbf{0}, r)$ |
| $\mathcal{C}^k$ | Differentiability class of order $k$, where $k = 0, 1, 2, \ldots, \infty$ |
| $\|\mathbf{v}\|$ | Euclidean 2-norm of $\mathbf{v}$, i.e. $\|\mathbf{v}\|_2$ |
| $\|\mathbf{v}\|_p$ | Vector p-norm of $\mathbf{v}$, for $p \in [1, \infty]$ |
| $\|A\|_F$ | Frobenius norm of matrix $A$ |
| $[\mathbf{v}]_\times$ | Cross product matrix, defined such that $\mathbf{v} \times \mathbf{w} = [\mathbf{v}]_\times \mathbf{w}$ |
| $\mathrm{axial}(V)$ | Inverse operation of $[\mathbf{v}]_\times = V$, i.e. $\mathbf{v} = \mathrm{axial}(V)$ |
| $\mathbf{u} \cdot \mathbf{v}$ | Vector dot product, for $\mathbf{u}, \mathbf{v} \in \mathbb{R}^n$ |
| $\mathbf{u} \times \mathbf{v}$ | Vector cross product, for $\mathbf{u}, \mathbf{v} \in \mathbb{R}^3$ |
| $\square \circ \square$ | Rotation composition, where $\square$ is $E, F, T, \ldots$ |
| $\square \oplus \square$ | Tilt vector addition where $\square$ is $P, \tilde{P}, T, \ldots$ |
| $pq$ | Quaternion multiplication, for $p, q \in \mathbb{H}$ (plain juxtaposition, no explicit multiplication symbol) |
| $\lfloor x \rfloor$ | Floor function, i.e. the greatest integer $\leq x$ |
| $\lceil x \rceil$ | Ceiling function, i.e. the lowest integer $\geq x$ |
| $\mathrm{sgn}(\cdot)$ | Sign function with range $\{-1, 0, 1\}$ and $\mathrm{sgn}(0) = 0$ |
| $\mathrm{sign}(\cdot)$ | Sign function with range $\{-1, 1\}$ and $\mathrm{sign}(0) = 1$ |
| $\min\{\ldots\}$ | Minimum element of a set |
| $\max\{\ldots\}$ | Maximum element of a set |
| $\mathrm{argmin}(\square)$ | Arguments of the minima of a function |
| $\mathrm{argmax}(\square)$ | Arguments of the maxima of a function |
| $\mathrm{sinc}(\cdot)$ | Cardinal sine function, given by $\mathrm{sinc}(x) = \frac{\sin x}{x}$ |
| $\mathrm{atan2}(y, x)$ | 2-argument arctangent function with $\mathrm{atan2}(0, 0) = 0$ |
| | |
| $\mathrm{wrap}(\cdot)$ | Wrap the angle $\cdot$ to the range $(-\pi, \pi]$ |
| $\mathrm{slerp}(q_0, q_1, u)$ | Spherical linear interpolation from $q_0$ to $q_1$ by $u$ |
| $\mathrm{coerce}(\cdot, m, M)$ | Coerce/saturate $\cdot$ to the range $[m, M]$ |

**Kinematic Pose Spaces**

| | |
|---|---|
| $\mathbf{q}$ | Vector of joint space positions |
| $\dot{\mathbf{q}}$ | Vector of joint space velocities |
| $\ddot{\mathbf{q}}$ | Vector of joint space accelerations |
| | |
| $\boldsymbol{\Phi}$ | Vector of abstract space positions |
| $\boldsymbol{\Phi}_l$ | Vector of abstract space leg positions |
| $\boldsymbol{\Phi}_a$ | Vector of abstract space arm positions |
| | |
| $\phi_{l*}$ | Abstract space leg angle $*$, where $* = x, y, z$ |
| $\phi_{f*}$ | Abstract space foot angle $*$, where $* = x, y$ |
| $\tilde{\phi}_{f*}$ | Abstract space ankle angle $*$, where $* = x, y$ |
| $\phi_{a*}$ | Abstract space arm angle $*$, where $* = x, y$ |
| $\epsilon_l$ | Abstract space leg retraction |
| $\epsilon_a$ | Abstract space arm retraction |
| | |
| $\mathbf{p}_l$ | Inverse space leg end effector (foot) position $\in \mathbb{R}^3$ |
| $\mathbf{p}_a$ | Inverse space arm end effector (hand) position $\in \mathbb{R}^3$ |
| $q_l$ | Inverse space leg end effector (foot) orientation $\in \mathbb{Q}$ |
| $q_a$ | Inverse space arm end effector (hand) orientation $\in \mathbb{Q}$ |
| | |
| $\mathbf{t}_l$ | Leg tip point $\in \mathbb{R}^3$ (part of leg tip pose space) |
| $\mathbf{t}_a$ | Arm tip point $\in \mathbb{R}^3$ (part of arm tip pose space) |
| | |
| $L_l$ | Upper and lower leg link length (assumed equal) |
| $L_a$ | Upper and lower arm link length (assumed equal) |
| $L_i$ | Inverse leg scale (scalar measure of robot hip to ankle size) |
| $L_t$ | Leg tip scale (scalar measure of robot hip to foot size) |

**Miscellaneous**

| | |
|---|---|
| $g$ | Gravitational acceleration constant, $g = 9.81 \,\mathrm{m/s^2}$ |
| $\mathbf{g}$ | Gravitational acceleration vector, $\mathbf{g} = (0, 0, -g)$ |
| $\bar{*}$ | Shorthand for $\frac{1}{2}*$, where $*$ is a rotation parameter like $\psi$, $\theta_E$ |
| $s_*$ | Shorthand for $\sin(*)$, e.g. $s_\psi \equiv \sin(\psi)$ |
| $c_*$ | Shorthand for $\cos(*)$, e.g. $c_\gamma \equiv \cos(\gamma)$ |
| $s_{\bar{*}}$ | Shorthand for $\sin\left(\frac{1}{2}*\right)$, e.g. $s_{\bar{\psi}} \equiv \sin\left(\frac{1}{2}\psi\right)$ |
| $c_{\bar{*}}$ | Shorthand for $\cos\left(\frac{1}{2}*\right)$, e.g. $c_{\bar{\alpha}} \equiv \cos\left(\frac{1}{2}\alpha\right)$ |
| $A \leftarrow B$ | Assignment operator, where $A$ receives the value of $B$ |
| $\mathcal{D}_w, \mathcal{D}_x, \mathcal{D}_y, \mathcal{D}_z$ | A symmetric four-way partition of the set of all rotations, based on which quaternion parameter has the greatest magnitude. Loose notation is used, in that we can say both $R \in \mathcal{D}_w$ and $q \in \mathcal{D}_w$, and it means the same for equivalent $R$ and $q$. Referred to as *dominant regions* (see Section 5.5.5.2). |

## Coordinate Frames

- All coordinate frames are right-handed ($\mathbf{x} \times \mathbf{y} = \mathbf{z}$)

- {G} always refers to the global coordinate frame

- {B} always refers to the body-fixed coordinate frame

- In general for coordinate frames, the x-axis points 'forwards', the y-axis points 'leftwards' and the z-axis points 'upwards'

Part I

OPENING

# INTRODUCTION

The task of bipedal locomotion exposes many facets of the concept of balance—most notably the many and varied methods by which balance can be preserved. While humans, even in early childhood, seem to effortlessly know how to stabilise their gait and best react to pushes from all directions while walking, the situation is very different for humanoid robotic platforms. For robots it must first be established by which approaches they can be made to keep their balance, before algorithms can be developed that allow them to reliably execute such strategies. It is a desire in the field of humanoid robotics for legged robots to eventually be able to move as fluidly and dynamically as their animal and human counterparts. Although this is something that we are beginning to see in the state of the art of quadruped robots, for biped robots this goal is still decently removed.

This thesis explores the generation of robust feedback-stabilised bipedal walking gaits, with the aim of using as diverse a set of strategies as possible for keeping balance—not just the standard adjustments of step size and timing.[1] It is explored how gaits and their underlying sensorimotor architecture can be constructed so that they work in particular for low-cost robots, where dealing with sensor noise and lacking actuator precision is paramount, and where the kinds of calculations and predictions one can make about the future states of the robot are limited. With possible additional restrictions on the available onboard computational resources, methods are investigated that rely predominantly on analytic calculation, and not on the solution of large-scale numerical optimisation problems. This allows the resulting methods to be as portable and efficient as possible in generating largely model-free walking motions for a wide range of bipedal robots.

The task of bipedal locomotion poses many difficulties for a robot, including having to deal with incomplete information, sensor noise, imperfect actuation, joint backlash, structural non-rigidities, uneven surfaces, external disturbances, and latencies in the sensorimotor loop. The control aspect of bipedal walking is also made difficult by the high dimensionality and significant nonlinearity of the system, as well as the floating-base nature of the trunk, relatively low controllability of the full dynamical system, and the only indirect observability of the positions and orientations of the trunk and limbs of the robot. Viewed at a fundamental level, all changes of state of the robot can only be effectuated via the foot-to-ground contact forces, but exactly these forces are neither controllable nor measurable[2] nor particularly predictable, so the resulting task is highly challenging.

---

1 With sometimes the additional application of ankle torque strategies.

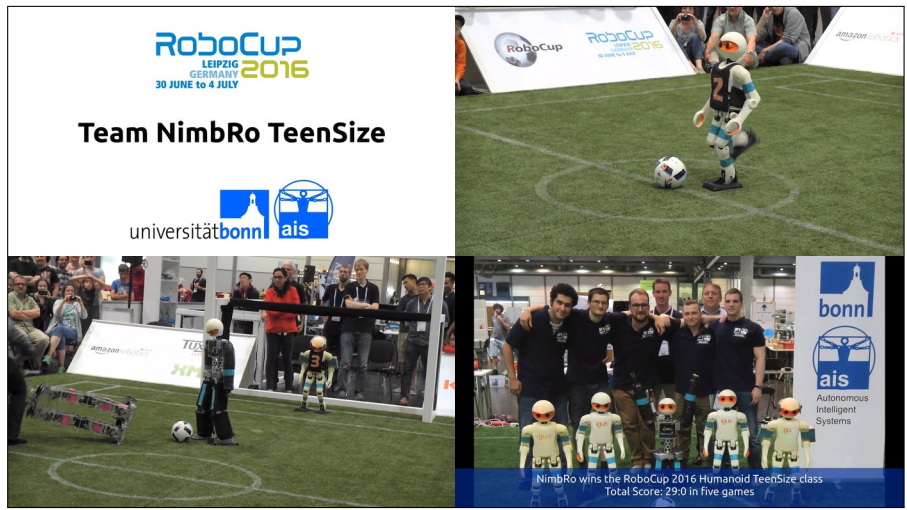

Video 1.1: Highlights of the performance of team NimbRo in the TeenSize soccer competition at RoboCup 2016 in Leipzig, Germany. The robot soccer team won the competition, and consisted of Dynaped and four igus Humanoid Open Platform robots.
https://youtu.be/G9llFqAwI-8
*RoboCup 2016: Humanoid TeenSize Soccer Winner NimbRo*

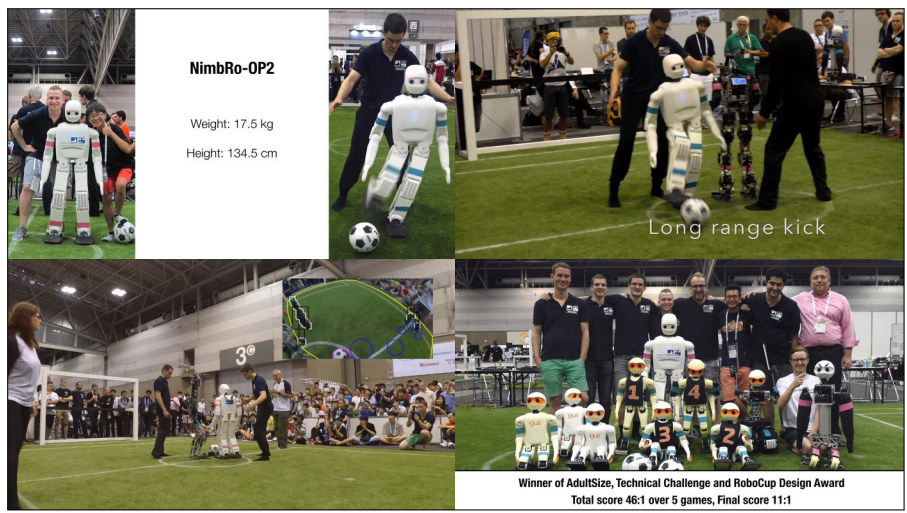

Video 1.2: Highlights of the performance of team NimbRo in the Adult-Size soccer competition at RoboCup 2017 in Nagoya, Japan. The robot soccer team won the competition, and consisted of the NimbRo-OP2 and a highly modified Copedo robot for jumping.
https://youtu.be/RG2O5OwGdSg
*RoboCup 2017 Humanoid League Winner NimbRo AdultSize*

The algorithms that are developed in this thesis to address these challenges—and specifically the requirements that these algorithms then have to fulfil—are motivated by the environment of the Robo-Cup competition. RoboCup, or specifically the RoboCup Humanoid League,[3] is a yearly international competition where teams from all over the world come together to compete against each other in robot soccer. The robots have to be fully autonomous, and fall into one of three size classes: KidSize, TeenSize and AdultSize. Videos 1.1 to 1.4 show summaries of NimbRo's winning entries to the TeenSize and AdultSize competitions over the recent years. As can be seen in the videos, the soccer fields that the robots play on consist of an uneven artificial grass surface that can be unpredictable and destabilising at times. Frequent tangles and contacts between the robots also cause the field to become a locomotion environment that is rife with a wide variety of unforeseeable disturbances (see Video 1.5). The main influence of the RoboCup competition on the methods developed in this thesis however, is from the perspective of robustness under continuous autonomy. RoboCup lends the viewpoint that a method is only useful if it does the right thing every time, in every situation, without the possibility of external intervention—and not if it only does the right thing 95% of the time.

There are two main paradigms for the implementation of bipedal robotic gaits. A common approach in the state of the art is to use a dynamics model of some kind to capture and predict the physical response of the robot, and calculate or optimise a trajectory to satisfy the required motion and balance criteria. This motion trajectory is then executed on the robot, often with a controller to reject deviations and enforce tracking, and/or under regular recomputation to adapt for differences in the real response of the robot. For low-cost and imprecise robots however, where good quality tracking and execution of a trajectory is not given, using such optimised trajectories generated directly from simplified or even whole-body dynamics models is often fraught with difficulty. Significant nonlinearities, such as joint backlash, sensor and actuator delays, irregular properties of the contact surface, and unmeasurable external disturbances, are difficult to incorporate into models. This greatly limits the predictive power of such models, and subsequently the applicability of such methods to such robots.

Small, cheap and simple robots can often easily be made to walk despite these nonlinearities using hand-crafted gaits. These gaits are usually not very flexible, and not particularly resistant to disturbances, but nevertheless find a way to make use of the natural dynamics of the robot to produce a functional semi-stable gait. This is the foundation of the second paradigm for the implementation of bipedal gaits—letting the robot find its own natural rhythm with an open-loop gait generator, and extending this generator with feedback controllers that seek to return the robot to that rhythm when there are noticeable deviations from it. The gaits developed in this thesis, namely the direct fused angle feedback gait and the tilt phase controller, both follow this paradigm, and extend their respective open-loop gaits, the Central Pattern Generator (CPG) and the Keypoint Gait Generator (KGG), with so-called corrective actions. These are targeted dynamic modifications of the generated open-loop waveforms that are designed to influence the balance of the robot in a particular way. Examples of corrective actions include tilt adjustments of the arms, orientation adjustments of the feet, and translational shifts of the robot's Centre of Mass (CoM). It should be noted that step size and timing adjustments are also considered to be corrective actions, but step size feedback, for example, is only used as a last resort for keeping balance, as amongst other things it leads to a direct non-compliance with any existing footstep plans.

---

2 No reliable ankle force-torque sensors are assumed to be available in this thesis.

3 Official website: https://www.robocuphumanoid.org

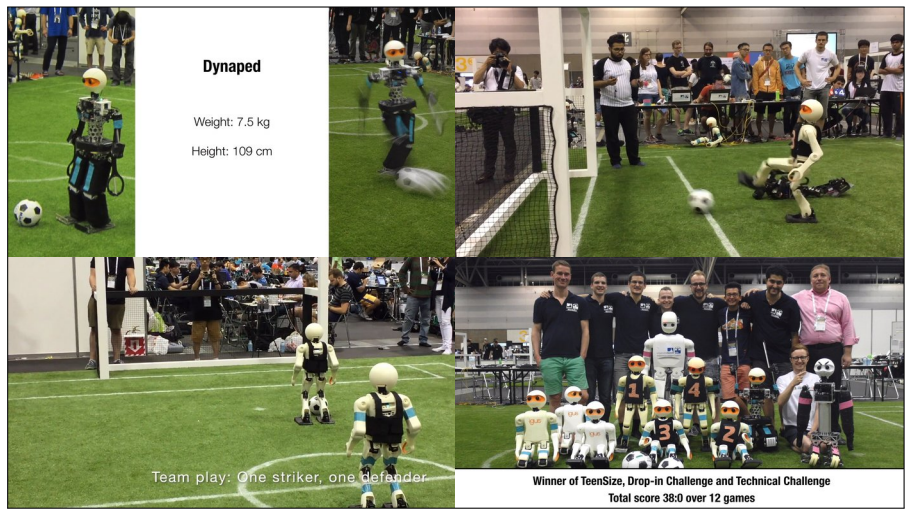

Video 1.3: Highlights of the performance of team NimbRo in the TeenSize soccer competition at RoboCup 2017 in Nagoya, Japan. The robot soccer team won the competition, and consisted of Dynaped and seven igus Humanoid Open Platform robots.
https://youtu.be/6ldHWWHfeBc
*RoboCup 2017 Humanoid League Winner NimbRo TeenSize*

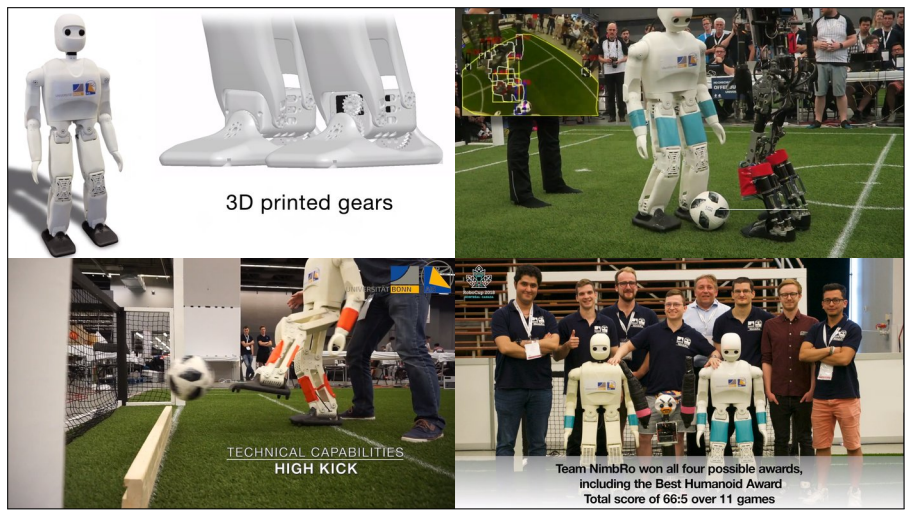

Video 1.4: Highlights of the performance of team NimbRo in the Adult-Size soccer competition at RoboCup 2018 in Montréal, Canada. The robot soccer team won the competition, and consisted of the NimbRo-OP2, NimbRo-OP2X, and a highly modified Copedo robot for jumping.
https://youtu.be/tPktQyFrMuw
*RoboCup 2018 Humanoid AdultSize Soccer Winner NimbRo*

Underpinning the two new gaits are many developments in the areas of sensor processing, state estimation, actuation and rotation theory. These developments are all presented in detail in this thesis, and in particular include methods for actuator control, sensor calibration, sensor processing, attitude estimation, stepping motion estimation and robot calibration. In terms of rotation theory, significant advancements to the state of the art are made through the introduction of a new and better concept of 'yaw', namely fused yaw. Three new rotation representations that leverage this concept of yaw—the tilt angles, fused angles and tilt phase space representations—are also introduced and used recurrently throughout the remainder of the thesis.

All of the developments and contributions made throughout this thesis have been released fully open source online (Team NimbRo, 2018a). They have been tested not only in simulation, but also predominantly on real hardware. This is important, as real hardware conditions are almost always more strenuous than simulated ones, and arguably, performance on a real robot is all that matters. Testing has been carried out on a wide variety of real robots, including the igus Humanoid Open Platform and the NimbRo-OP2. More of the tested robots are detailed in Chapter 2.

## 1.1 KEY CONTRIBUTIONS

The following key contributions are made by this thesis:

- The novel concept of fused yaw is introduced as a way of partitioning 3D rotations into their respective yaw and tilt components in a geometrically and mathematically meaningful way.

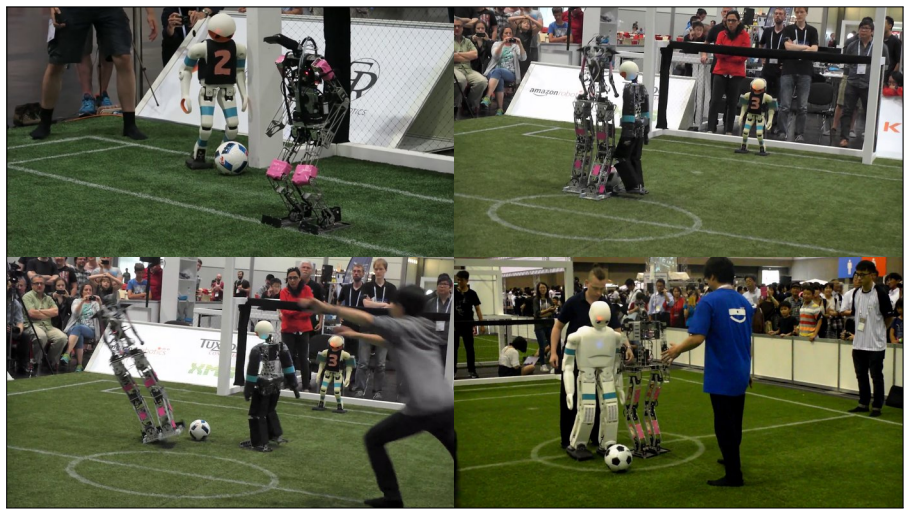

Video 1.5: Examples of three situations, in addition to the effects of the artificial grass surface, that caused disturbances to walking bipedal robots at various RoboCup competitions.
https://youtu.be/IuqGolmLi2M
*Examples of Walking Disturbances at RoboCup*

- Based on the concept of fused yaw, three new rotation representations are introduced and investigated in detail: the tilt angles, fused angles and tilt phase space representations. The latter two in particular offer definitions of 'pitch' and 'roll' that far surpass the definitions that are used in the context of Euler angles.

- A gait using corrective actions driven by fused angle feedback mechanisms is built around an open-loop central pattern-driven gait generator, and is tested on many different robotic platforms.

- A new self-stable omnidirectional gait generator called the KGG is developed. It strikes a balance between the security and simplicity of hand-crafted gaits, and the advantages of analytically computed and optimised gaits. More than just make the robot walk, the KGG directly embeds a myriad of corrective actions that can be commanded and activated by higher level controllers to systematically allow preservation of balance during walking.

- A controller for the corrective actions of the KGG is formulated based on the tilt phase space. The diversity of the actuated corrective actions arguably covers the majority of humanlike strategies by which biped robots can balance during walking.

In addition to these key contributions, the following contributions are also made by this thesis:

- The notions of referenced rotations and tilt vector addition are introduced in the context of 3D rotation theory.

- An actuator control scheme is realised that makes use of a servo motor model and feed-forward torque estimation scheme to improve the compliant tracking of dynamic joint trajectories.

- Methods for sensor and robot calibration are presented, in particular for the purposes of magnetometer and Inertial Measurement Unit (IMU) calibration and data processing.

- Methods for the online autocalibration of the magnetometer and gyroscope biases are developed.

- Two useful non-standard kinematic pose spaces are presented, and used to formulate a cost-based solution to the humanoid leg inverse kinematics problem. A generalised arm inverse kinematics method is also developed that can place any functional point of the arm on any desired ray through the shoulder.

- A 3D attitude estimator is realised that automatically uses the concept of fused yaw in the absence of magnetometer data for IMU orientation estimation purposes.

- A humanoid kinematic model is used in conjunction with the developed attitude estimator to formulate CoM state and stepping motion estimation processes.

- The capture step gait from Missura (2015) is refined in terms of its open-loop gait, state estimation and implementation. A rigorous data-based tuning approach is also documented.

- The novel *nonlinear tripendulum model* is developed for the purposes of dynamic step size adjustment.

- A number of innovative non-standard mathematical filters and functions are detailed and used throughout the various gaits and calibration algorithms in this thesis.

- All developed algorithms and methods are released fully open source online, in the C++ and sometimes also Matlab languages.

Although not a direct contribution of this thesis, further contributions to the released Humanoid Open Platform ROS Software (and associated robot platforms) were made by the author in the course of completing this thesis. These more indirect contributions include work in the areas of soccer behaviours, demonstration routines, state machines (Allgeuer and Behnke, 2013), servo communication routines, microcontroller firmwares (CM730/CM740), robot hardware design and construction. More general contributions to the ROS software framework included contributions in the areas of robot control, Gazebo simulation, real-time logging, dynamic configuration, visualisation tools, hardware interfaces, network interfaces, and general motion modules.

## 1.2    PUBLICATIONS

Parts of this thesis have been previously published in journals and conference proceedings. The source codes behind essentially all of the described methods and algorithms have also been released open source as a contribution to the community. The academic publications and hardware/software releases that contributed and form part of this thesis are listed as follows.

### 1.2.1    Core Academic Publications

**Allgeuer and Behnke (2014)**    Presents an attitude estimator, based on 3D nonlinear complementary filtering, that fuses 3-axis gyroscope, accelerometer and optionally magnetometer data into a robust quaternion estimate of orientation.

> P. Allgeuer and S. Behnke (2014). "Robust Sensor Fusion for Robot Attitude Estimation". In: *International Conference on Humanoid Robots (Humanoids)*. Madrid, Spain.

**Allgeuer and Behnke (2015)**    Introduces and thoroughly investigates the properties of the fused angles and tilt angles rotation parameterisations, motivated by the task of representing the orientation of a balancing body.

> P. Allgeuer and S. Behnke (2015). "Fused Angles: A Representation of Body Orientation for Balance". In: *International Conference on Intelligent Robots and Systems (IROS)*. Hamburg, Germany.

**Allgeuer and Behnke (2016)**    Presents a complete omnidirectional closed-loop gait based on the direct feedback of orientation deviation estimates expressed in terms of fused angles. The feedback controls the step timing, in addition to the virtual walking slope and five further types of corrective actions.

> P. Allgeuer and S. Behnke (2016). "Omnidirectional Bipedal Walking with Direct Fused Angle Feedback Mechanisms". In: *International Conference on Humanoid Robots (Humanoids)*. Cancún, Mexico.
>
> Video 13.1: `https://youtu.be/xnzJi2hTfAo`                    (page 365)

**Allgeuer and Behnke (2018a)**    Presents a multifaceted feedback controller, based on the tilt phase space, that uses step timing and eight other corrective actions to stabilise a core keypoint-based gait. The feedback controller is free of any physical model of the robot, very computationally inexpensive, and requires only a single 6-axis IMU.

> P. Allgeuer and S. Behnke (2018a). "Bipedal Walking with Corrective Actions in the Tilt Phase Space". In: *International Conference on Humanoid Robots (Humanoids)*. Beijing, China.

**Allgeuer and Behnke (2018b)**    Investigates and delineates in great detail the advantages that fused angles have over Euler angles for representing orientations in balance-related scenarios. Points of comparison include the locations of the singularities, the associated parameter sensitivities, the level of mutual independence of the parameters, and the axisymmetry of the parameters.

P. Allgeuer and S. Behnke (2018b). "Fused Angles and the Deficiencies of Euler Angles". In: *International Conference on Intelligent Robots and Systems (IROS)*. Madrid, Spain.

**Allgeuer and Behnke (2018c)**    Introduces the tilt phase space rotation parameterisation, and extensively explores the properties of this new representation. Previously unexplored properties of the general notion of tilt rotations are also presented.

P. Allgeuer and S. Behnke (2018c). "Tilt Rotations and the Tilt Phase Space". In: *International Conference on Humanoid Robots (Humanoids)*. Beijing, China.

### 1.2.2 Further Academic Publications

**Schwarz et al. (2013)**    Details the hardware of the NimbRo-OP robot, as an extension to Schwarz et al. (2012), and presents the preliminary DARwIn-OP software framework-based software release that was made for the robot. The state of development (at the time) of the ROS-based software is also addressed.

M. Schwarz, J. Pastrana, P. Allgeuer, M. Schreiber, S. Schueller, M. Missura and S. Behnke (2013). "Humanoid TeenSize Open Platform NimbRo-OP". In: *17th RoboCup International Symposium*. Eindhoven, Netherlands.

**Allgeuer and Behnke (2013)**    Details two novel behaviour control architectures for autonomous agents, namely the State Controller Library and Behaviour Control Framework. Both have associated C++ implementations, and while the former is state-based and allows for multiple future actions to be planned, the latter is behaviour-based and exploits a hierarchical structure with inhibitions to allow for dynamic transitioning.

P. Allgeuer and S. Behnke (2013). "Hierarchical and State-based Architectures for Robot Behavior Planning and Control". In: *8th Workshop on Humanoid Soccer Robots, International Conference on Humanoid Robots (Humanoids)*. Atlanta, USA.

**Allgeuer et al. (2013)**    Describes a software framework for the NimbRo-OP that is based on the Robot Operating System (ROS) middleware, providing functionality for hardware abstraction, visual perception, behaviour generation and basic soccer skills.

P. Allgeuer, M. Schwarz, J. Pastrana, S. Schueller, M. Missura and S. Behnke (2013). "A ROS-based Software Framework for the NimbRo-OP Humanoid Open Platform". In: *8th Workshop on Humanoid Soccer Robots, International Conference on Humanoid Robots (Humanoids)*. Atlanta, USA.

**Allgeuer et al. (2015)**    Details the 3D printed igus Humanoid Open Platform robot, both in terms of its mechanical design and in terms of the significant associated updates to the ROS soccer software.

P. Allgeuer, H. Farazi, M. Schreiber and S. Behnke (2015). "Child-sized 3D Printed igus Humanoid Open Platform". In: *International Conference on Humanoid Robots (Humanoids)*. Seoul, Korea.

Video 2.2: `https://youtu.be/RC7ZNXclWWY`                    (page 29)

**Allgeuer et al. (2016)**    Provides updates to Allgeuer et al. (2015) on the continued development of the igus Humanoid Open Platform.

P. Allgeuer, H. Farazi, G. Ficht, M. Schreiber and S. Behnke (2016). "The igus Humanoid Open Platform: A Child-sized 3D Printed Open-Source Robot for Research". In: *Künstliche Intelligenz* 30.3.

**Ficht et al. (2017)**    Presents the 3D printed NimbRo-OP2 robot from a mechanical, electrical and software perspective, building on the previous igus Humanoid Open Platform robot, and in particular the ROS soccer software.

G. Ficht, P. Allgeuer, H. Farazi and S. Behnke (2017). "NimbRo-OP2: Grown-up 3D Printed Open Humanoid Platform for Research". In: *International Conference on Humanoid Robots (Humanoids)*. Birmingham, UK.

Video 2.3: `https://youtu.be/WJKc56uUuF8`                    (page 31)

**Ficht et al. (2018)**    Presents the 3D printed NimbRo-OP2X robot from a mainly mechanical and software perspective, providing improvements over the NimbRo-OP2 predecessor robot, including in particular the inclusion of a Graphics Processing Unit (GPU).

G. Ficht, H. Farazi, A. Brandenburger, D. Rodriguez, D. Pavlichenko, P. Allgeuer, M. Hosseini and S. Behnke (2018). "NimbRo-OP2X: Adult-sized Open-source 3D Printed Humanoid Robot". In: *International Conference on Humanoid Robots (Humanoids)*. Beijing, China.

Video 2.4: `https://youtu.be/qmYSt3nVJpA`          (page 33)

### 1.2.3   Hardware Releases

**NimbRo-OP Hardware (2012)**     Provides 3D CAD files in both IGES and STEP formats for all the parts of the NimbRo-OP robot. In addition to the hardware design, this repository also contains the required and recommended patches of the ROBOTIS DARwIn-OP software, to allow it to run on, and control, the NimbRo-OP.

| | |
|---|---|
| **Author:** | Team NimbRo |
| **Url:** | `https://github.com/NimbRo/nimbro-op` |
| **Publications:** | Schwarz et al. (2012), Schwarz et al. (2013) |

**igus Humanoid Open Platform Hardware (2015)**     Provides 3D CAD files in the STEP format for all 3D printed plastic parts of the igus Humanoid Open Platform.

| | |
|---|---|
| **Author:** | igus GmbH |
| **Url:** | `https://github.com/igusGmbH/HumanoidOpenPlatform` |
| **Publications:** | Allgeuer et al. (2015), Allgeuer et al. (2016) |

**NimbRo-OP2[X] Hardware (2018)**     Provides specification of all the hardware aspects of the NimbRo-OP2 and NimbRo-OP2X robots, including in particular Bills of Materials (BOMs), and 3D CAD files in the STEP format for all gears and 3D printed plastic parts.

| | |
|---|---|
| **Author:** | Team NimbRo |
| **Url:** | `https://github.com/NimbRo/nimbro-op2` |
| **Publications:** | Ficht et al. (2017), Ficht et al. (2018) |

### 1.2.4   Software Releases

**NimbRo-OP ROS Software (2013)**     Provides a software framework for the NimbRo-OP robot that is based on the ROS middleware. The software provides functionality for hardware abstraction, low-level control, simple visual perception and behaviour generation, and implements basic soccer skills.

| | |
|---|---|
| **Author:** | Team NimbRo |
| **Url:** | `https://github.com/NimbRo/nimbro-op-ros` |
| **Publications:** | Allgeuer et al. (2013) |

**State Controller Library (2014)**    Provides a C++ software framework for the implementation of generalised finite state machines that allow the planning and enqueuing of future states.

| | |
|---|---|
| **Author:** | Philipp Allgeuer |
| **Url:** | `https://github.com/AIS-Bonn/state_controller_library` |
| **Publications:** | Allgeuer and Behnke (2013) |

**Behaviour Control Framework (2014)**    Provides a C++ software framework for the construction and execution of a set of behaviours that are coordinated by chained and non-chained inhibition trees.

| | |
|---|---|
| **Author:** | Philipp Allgeuer |
| **Url:** | `https://github.com/AIS-Bonn/behaviour_control_framework` |
| **Publications:** | Allgeuer and Behnke (2013) |

**Attitude Estimator (2016)**    Provides a C++ implementation of a 3D IMU fusion algorithm that is based on nonlinear passive complementary filtering and the concept of fused yaw.

| | |
|---|---|
| **Author:** | Philipp Allgeuer |
| **Url:** | `https://github.com/AIS-Bonn/attitude_estimator` |
| **Publications:** | Allgeuer and Behnke (2014) |

**Matlab/Octave Rotations Library (2018)**    Provides a library for working with 3D rotations of all kinds in Matlab and Octave. In particular, the library supports the fused angles, tilt angles and tilt phase space representations, and is intended as a test bed for rotation-related calculations and algorithms.

| | |
|---|---|
| **Author:** | Philipp Allgeuer |
| **Url:** | `https://github.com/AIS-Bonn/matlab_octave_rotations_lib` |
| **Publications:** | Allgeuer and Behnke (2015), Allgeuer and Behnke (2018b), Allgeuer and Behnke (2018c) |

**Rotations Conversion Library (2018)**    Provides a high performance C++ library for working with 3D rotations of all kinds, including in particular the fused angles, tilt angles and tilt phase space representations. This library is essentially an optimised port to C++ of the Matlab/Octave Rotations Library (2018).

| | |
|---|---|
| **Author:** | Philipp Allgeuer |
| **Url:** | `https://github.com/AIS-Bonn/rot_conv_lib` |
| **Publications:** | Allgeuer and Behnke (2015), Allgeuer and Behnke (2018b), Allgeuer and Behnke (2018c) |

Humanoid Open Platform ROS Software (2018)    Provides a ROS-based software framework for the igus Humanoid Open Platform, NimbRo-OP2 and NimbRo-OP2X robots, that vastly extends and improves on the NimbRo-OP ROS Software (2013). The framework has been updated several times since its initial release in 2015, and features a fully functional rule-compliant RoboCup soccer system as its main application. The implemented foundations, including low-level control, sensor management, and bipedal locomotion, can however easily be used for other applications.

| | |
|---|---|
| **Author:** | Team NimbRo |
| **Url:** | https://github.com/AIS-Bonn/humanoid_op_ros |
| **Publications:** | Allgeuer et al. (2015), Allgeuer et al. (2016), Ficht et al. (2017), Ficht et al. (2018) |

## 1.3 OUTLINE

The remainder of this thesis is organised as follows:

**Chapter 2:**   Introduces the robotic platforms used throughout this thesis, and the ROS software framework made to run on them.

**Chapter 3:**   Details the main robot control loop, and the actuator control scheme used to compensate the actuator commands.

**Chapter 4:**   Details the kinematic calibration procedures of the robot, as well as the IMU and magnetometer calibration routines. The autocalibration schemes of the latter two are also discussed.

**Chapters 5 to 7:**   Establish the tilt angles, fused angles and tilt phase space rotation representations, critically compare them to Euler angles, and present a vast array of useful results, conversions and visualisations involving these new representations.

**Chapter 8:**   Presents related work for the field of bipedal locomotion.

**Chapter 9:**   Presents the humanoid kinematic models in use, and the corresponding required pose space conversions.

**Chapter 10:**   Details the attitude estimator used for trunk orientation estimation, as well as the CoM state estimation scheme, and the stepping motion model.

**Chapter 11:**   Details the Central Pattern Generator (CPG), and how it is used to make robots walk open-loop.

**Chapter 12:**   Reviews the capture step gait and what has been modified for its reimplementation in the ROS software framework.

**Chapter 13:**   Presents the direct fused angle feedback controller, which is a balance controller that adds closed-loop stabilising mechanisms to the CPG.

**Chapter 14:**   Introduces the Keypoint Gait Generator (KGG), which is an analytically computed open-loop gait generator that directly incorporates the option for a wide array of corrective actions in its trajectory generation scheme.

**Chapter 15:**   Details the tilt phase controller, which is a controller that activates the corrective actions of the KGG in order to maintain balance during walking.

**Chapter 16:**   Makes closing remarks about the achievements and contributions of this thesis.

**Appendix A:**   Provides a reference for the many standard and non-standard functions and filters that are used throughout this thesis.

**Appendix B:**   Provides supporting proofs for certain equations and/or claims that are made throughout Part III that are important to understand, yet non-obvious.

While reading through these chapters, please be aware that there is:

- A List of Acronyms on page xxix,

- A useful Summary of Notation on page xxxi, and,

- A comprehensive Index of Terms on page 521.

Part II

SENSORIMOTOR MANAGEMENT

# ROBOT PLATFORMS

In addition to their testing and validation in physical simulation, the methods and algorithms developed in this thesis have been applied to a wide variety of different robotic platforms over the years. This is seen to be an essential property of any developed approach to robot control—that it is applicable and easily portable to a wide range of target robots, without significant adjustment or overfitting to the behaviour of any particular one. The robots used in this thesis, including namely the

- igus Humanoid Open Platform
- NimbRo-OP
- NimbRo-OP2
- NimbRo-OP2X
- Dynaped
- Copedo

are presented in detail in this chapter, with a specific focus on the hardware and software that is used in each.

## 2.1 ROBOT HARDWARE

The robot hardware platforms used for the validation of the methods developed in this thesis are presented as follows, starting with a description of the shared hardware elements between them.

### 2.1.1 Hardware Building Blocks

Numerous mechanical and electrical building blocks were shared between the many different generations of robots that were constructed. These include the type and model of servo motors that were used, as well as the electrical support infrastructure that was required to drive them.

#### 2.1.1.1 Servo Motors

All of the mechanical actuators used in the constructed robot platforms are ROBOTIS Dynamixel smart actuators,[1] as shown in Figure 2.1. These include servo motors from the Dynamixel:

---

[1] http://www.robotis.us/dynamixel

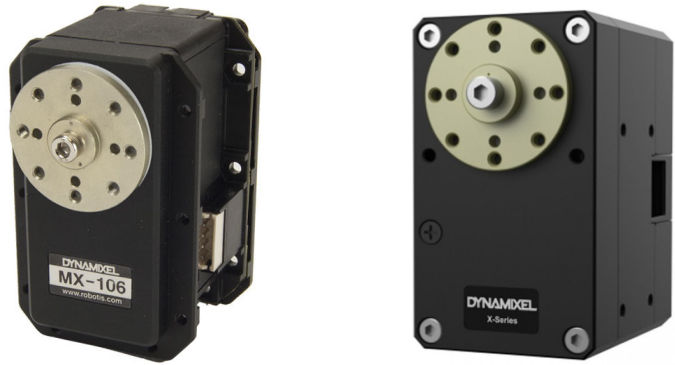

Figure 2.1: Examples of Dynamixel smart actuators, namely the MX-106 (left)
and XM540-W270 (right).
Source: https://www.trossenrobotics.com/robot-servos

- EX series (EX-106, EX-106+),

- RX series (RX-28, RX-64),

- MX series (MX-64, MX-106), and

- X series (XM540-W270) ranges.

All of these servos are modular building blocks that include a Direct
Current (DC) motor, reduction gearhead, motor driver circuit, joint
encoder, and microcontroller with serial communications ability. The
servos are designed to be mechanically used as exclusive connecting
joints in the structure of robots, and can be mechanically parallelised
to increase the torque. The servos are connected and controlled via a
serial data packet stream running the so-called Dynamixel protocol,
which allows read/write access to specific registers in the servo. This
allows commands and configuration options to be written to the servo,
and sensor data such as position, voltage and temperature to be read.
Internally, the position is controlled via a configurable Proportional-
Integral-Derivative (PID) feedback loop running at 2 kHz.

### 2.1.1.2  *CM740 Microcontroller Board*

Given that there are anywhere between 20 and 34 servos in a robot, an
interface is required between the serial connection ports of the servos
and the Personal Computer (PC) that is built into the robot. This
interface is provided by the ROBOTIS CM740 microcontroller board,
pictured in Figure 2.2, which incorporates a Cortex M3 STM32F103RE
ARM microcontroller running at 72 MHz. The CM740 connects via a
partially daisy-chained star topology to all of the servos, and provides
a gateway to route the serial communications from the servos to the
PC via USB. The CM740 was designed and manufactured by ROBOTIS
for their ROBOTIS OP2 robot, and is the direct successor of the more well-known CM730 microcontroller board, which was designed for the DARwIn-OP robot.

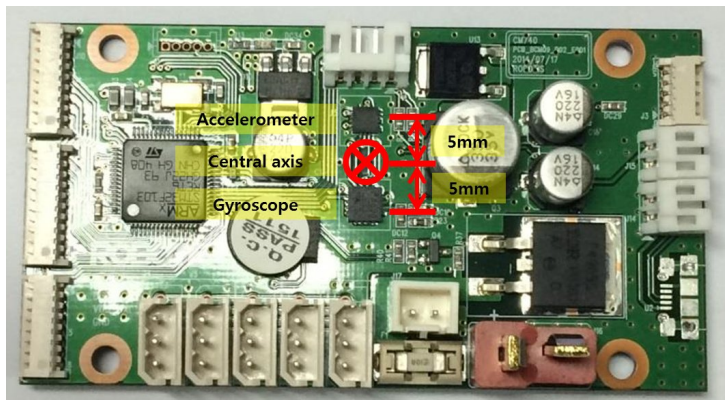

Figure 2.2: Picture of the CM740 microcontroller board showing the locations of the accelerometer and gyroscope sensors.
Source: http://support.robotis.com/en/product/robotis-op2/sub_controller(cm-740).htm

Table 2.1: CM740 microcontroller board specifications

| Type | Specification | Value |
|------|---------------|-------|
| **Microcontroller** | Chip | ARM Cortex M3 STM32F103RE |
| | Memory | 512 KB flash, 64 KB SRAM |
| | Frequency | 72 MHz |
| **Sensors** | Gyroscope | L3G4200D (3-axis) |
| | Accelerometer | LIS331DLH (3-axis) |
| | Magnetometer | HMC5883L (3-axis, optional) |
| **Interface board** | Buttons | 3 (one hardwired as reset) |
| | Lights | 5 × LEDs, 2 × RGB LEDs |
| **Extras** | Connectivity | USB, 5 × Dynamixel serial ports |
| | Other | Supply voltage sensor, buzzer |

In addition to its role as a servo communications gateway, the CM740 also includes a 3-axis accelerometer, 3-axis gyroscope, buzzer, supply voltage sensor, and ports for the connection of an interface board that contains three buttons, five Light Emitting Diodes (LEDs) and two Red-Green-Blue (RGB) LEDs. With customisation of the firmware (see Section 2.2.1), a 3-axis magnetometer can also be connected over I²C and used as an additional sensor. A power MOSFET on the board optionally allows the DC voltage supply to a subset of the servos (at most ~6–8 due to power considerations) to be turned on or off in software for safety purposes. A summary of the specifications and features of the CM740 microcontroller board is given in Table 2.1.

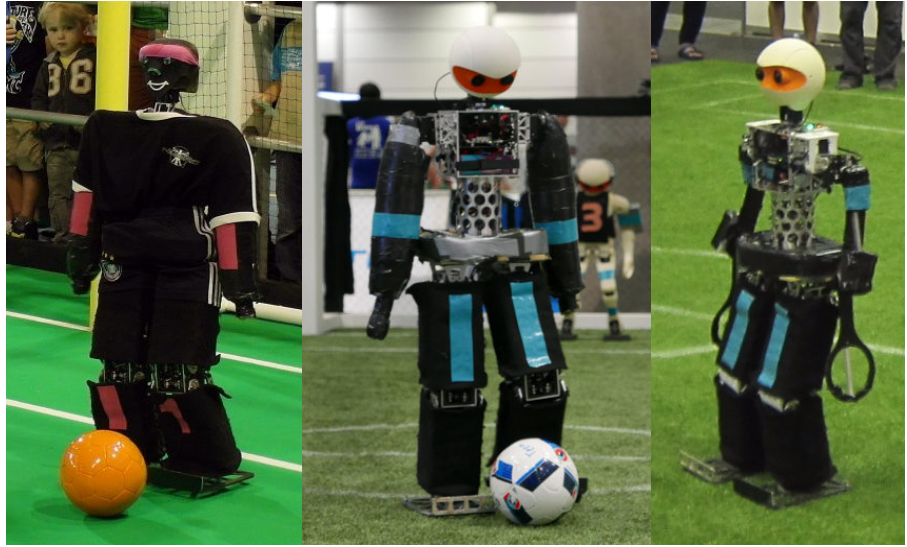

Figure 2.3: The many faces of Dynaped over the years, from left to right in chronological order. Note the changed arms in the right-most picture. Less externally visible is the change of PC and electronics from the first picture to the second.

### 2.1.2   Robot Platforms

All of the robot platforms on which the methods in this thesis were tested are custom built. Each of these platforms is presented below, in chronological order of first construction. A summary of the main specifications of each platform is also provided in Tables 2.2 and 2.3.

#### 2.1.2.1   *Dynaped*

Dynaped, shown in Figure 2.3, is an extremely robust and long-lived humanoid robot platform that was first built in 2008 specifically for the RoboCup TeenSize league (see Videos 1.1 and 1.3 on pages 4 and 6). Over the years, many gradual adjustments and upgrades were made to the robot in order to support faster PC, newer electronics including a CM740, newer software, more degrees of freedom in the arms (for getting up off the ground when fallen), and more degrees of freedom in the neck. Dynaped is constructed predominantly from milled aluminium and carbon fiber composite parts, and in its final state incorporated a 2 Degree of Freedom (DoF) 3D printed head, 3 DoF arms, and 5 DoF legs (only 5 as parallel leg kinematics are used), for a total of 18 DoF. Thanks to its use of parallel kinematics in the legs, Dynaped is mechanically very stable in the pitch direction when standing, and with four servos per pitch joint (thigh and shank pitch) per leg, the robot has a relative large amount of torque, given its low weight of just 8.0 kg. Foam padding, incorporated in strategic places around the robot frame, allows for a great resilience of the robot to impacts and falling.

Table 2.2: Robot platform specifications (Part 1)

| Platform | Spec. | Value |
|---|---|---|
| **Dynaped** | Height | 109 cm |
| | Weight | 8.0 kg |
| | DoF | 18 (5 per leg, 3 per arm, 2 in head) |
| | CPU | Intel Core i7-5500U (dual-core, 4 logical cores, 2.4–3.0 GHz) |
| | Servos | Dynamixel EX-106, RX-64, MX-64 |
| | Sensors | CM740 IMU, Logitech C905 camera fitted with a custom 150° wide-angle lens |
| | Battery | 5-cell LiPo (18.5 V, 3.8 Ah) |
| **Copedo** | Height | 114 cm |
| | Weight | 8.0 kg |
| | DoF | 17 (5 per leg, 3 per arm, 1 in head) |
| | CPU | Intel Core i7-5500U (dual-core, 4 logical cores, 2.4–3.0 GHz) |
| | Servos | Dynamixel EX-106(+), RX-64, RX-28 |
| | Sensors | 2-axis accelerometer (ADXL203), dual 1-axis gyroscopes (ADXRS300), IDS uEye UI1226LE camera fitted with a custom 150° wide-angle lens |
| | Battery | 5-cell LiPo (18.5 V, 3.8 Ah) |
| **NimbRo-OP** | Height | 90 cm |
| | Weight | 6.6 kg |
| | DoF | 20 (6 per leg, 3 per arm, 2 in head) |
| | CPU | AMD E-450 (dual-core, 2 logical cores, 1.65 GHz) |
| | Servos | 12 × MX-106, 8 × MX-64 |
| | Sensors | CM740 IMU, Logitech C905 camera fitted with a custom 180° wide-angle lens |
| | Battery | 4-cell LiPo (14.8 V, 3.6 Ah) |

Table 2.3: Robot platform specifications (Part 2)

| Platform | Spec. | Value |
|---|---|---|
| **igus Humanoid Open Platform** | Height | 92 cm |
| | Weight | 6.6 kg |
| | DoF | 20 (6 per leg, 3 per arm, 2 in head) |
| | CPU | Intel Core i7-5500U (dual-core, 4 logical cores, 2.4–3.0 GHz) |
| | Servos | 12 × MX-106, 8 × MX-64 |
| | Sensors | CM740 IMU, Logitech C905 camera fitted with a 150° wide-angle lens |
| | Battery | 4-cell LiPo (14.8 V, 3.8 Ah) |
| **NimbRo-OP2** | Height | 134.5 cm |
| | Weight | 18.5 kg |
| | DoF | 18 (5 per leg, 3 per arm, 2 in head) |
| | CPU | Intel Core i7-7700HQ (quad-core, 8 logical cores, 2.8–3.8 GHz) |
| | GPU | Nvidia GeForce GTX 1060 |
| | Servos | 34 × MX-106 |
| | Sensors | CM740 IMU, Logitech C905 camera fitted with a 150° wide-angle lens |
| | Battery | 4-cell LiPo (14.8 V, 8.0 Ah) |
| **NimbRo-OP2X** | Height | 135 cm |
| | Weight | 19.0 kg |
| | DoF | 18 (5 per leg, 3 per arm, 2 in head) |
| | CPU | Intel Core i7-8700T (hex-core, 12 logical cores, 2.4–4.0 GHz) |
| | GPU | Nvidia GeForce GTX 1050 Ti |
| | Servos | 34 × XM540-W270 |
| | Sensors | CM740 IMU, Logitech C905 camera fitted with a 150° wide-angle lens |
| | Battery | 4-cell LiPo (14.8 V, 8.0 Ah) |

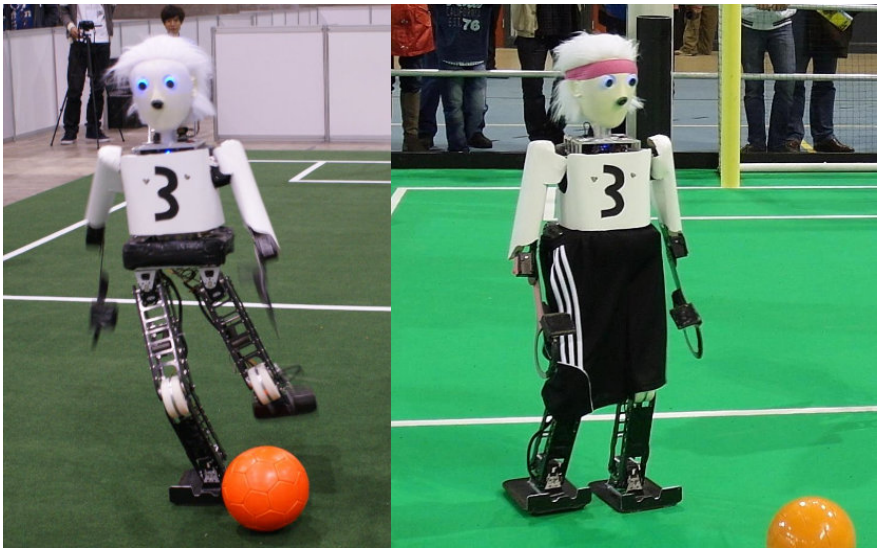

Figure 2.4: The Copedo robot in action at RoboCup competitions.

### 2.1.2.2 *Copedo*

Copedo (see Figure 2.4) is also a robot that, similar to Dynaped, was designed and manufactured from milled aluminium and carbon fiber composite parts, and intended for the RoboCup TeenSize league. Built in 2011, Copedo was decommissioned five years later in 2016 and used to construct a jumping AdultSize robot for further Robo-Cup competitions. One unique feature of Copedo was that it allowed pretensioning springs to be fitted to the legs, allowing the weight of the robot to be passively compensated to the extent that the robot could be made to balance standing up in its powered-off state. Like Dynaped, a parallel kinematic link structure was used in the legs for stability, and a total of 17 DoF were available—5 DoF per leg, 3 DoF per arm and 1 DoF in the head (yaw). Over the years the electronics and PC were updated to support a faster dual-core Central Processing Unit (CPU) (see Table 2.2), but otherwise the robot stayed much the same, weighing in at about 8.0 kg.

### 2.1.2.3 *NimbRo-OP*

Inspired by the DARwIn-OP robot and the experience gained with the Dynaped and Copedo robots, a prototype standard platform robot called the NimbRo-OP (see Figure 2.5) was developed in 2012. As an open platform, both the hardware and software designs for the robot were released open source to other researchers and the general public, in the hope to foster interest in the robot, and encourage research collaboration through community-based improvement of the platform. Following the first hardware release of the NimbRo-OP (Team NimbRo, 2012), an initial software release was completed that was heavily based on the framework released by ROBOTIS for the DARwIn-OP (Team

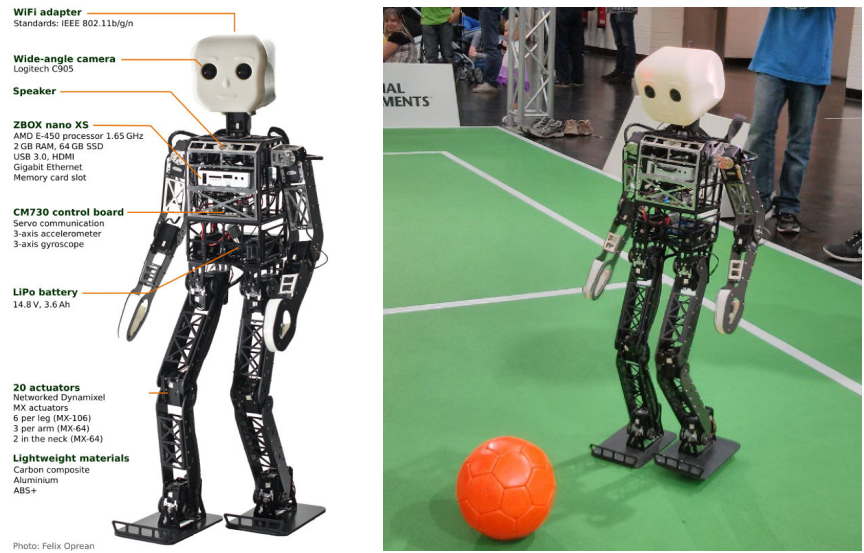

Figure 2.5: The NimbRo-OP robot, along with its hardware specifications.

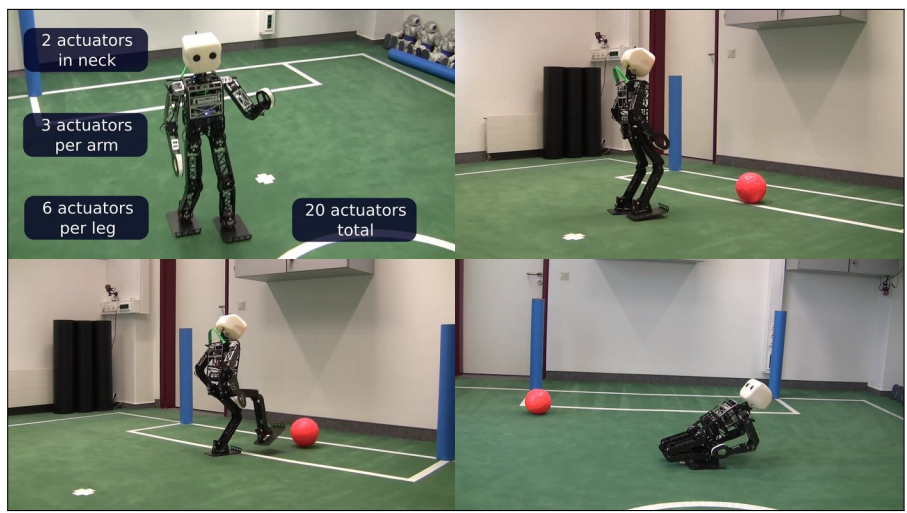

Video 2.1: Introduction video to the TeenSize open platform NimbRo-OP robot, showing the robot hardware and electronics, prone and supine get-up motions, and a soccer demonstration.
https://youtu.be/tn1uSz6YseI
*NimbRo-OP Humanoid TeenSize Open Platform Robot*

NimbRo, 2012). This was however only a transitional release, pending the shortly following release of the NimbRo-OP ROS Framework (Team NimbRo, 2013), an original custom software framework for the robot based on the Robot Operating System (ROS) middleware.

A diagram of the NimbRo-OP that highlights all of the main hardware features is shown on the left in Figure 2.5. As can be seen, the robot is manufactured mainly from milled aluminium profiles and carbon composite sheets, and has a 3D printed ABS+ plastic head. The robot is 95 cm tall and weighs a total of 6.6 kg with its battery. This combination of height and weight was chosen so as to allow the robot to be driven using only a single actuator per joint in a purely serial kinematics arrangement. This decreases the cost and complexity of the design, and allows the robot to exhibit quite large ranges of motion in its 20 degrees of freedom. ROBOTIS Dynamixel MX series actuators are used for all of the joints, of which there are two in the neck, three in each arm and six in each leg. All of the Dynamixel servos are chained together and addressed using a single one-wire Transistor-transistor Logic (TTL) serial bus, interfaced to the PC via a CM730 microcontroller board (predecessor of the CM740).

At the heart of the robot is a Zotac Zbox nano XS PC, featuring a dual-core AMD E-450 1.65 GHz processor with 2 GB Random-access Memory (RAM) and a 64 GB solid state disk, and various communication interfaces including Universal Serial Bus (USB) 3.0, High-Definition Multimedia Interface (HDMI) and Gigabit Ethernet ports. A USB WiFi adapter also allows for wireless networking. A stripped Logitech C905 camera fitted with a 180° wide-angle fisheye lens is mounted behind the right eye, and also connected to the PC via USB. The extremely wide field of view is convenient for keeping multiple objects in view at the same time.

More details on the NimbRo-OP hardware and software can be found in Schwarz et al. (2013) and Allgeuer et al. (2013). A video of the robot in action can be found in Video 2.1.

### 2.1.2.4 *Igus Humanoid Open Platform*

The igus Humanoid Open Platform (see Figure 2.6 and Video 2.2) is the standard platform that emerged from the prototype NimbRo-OP robot. It is 92 cm tall, weighs 6.6 kg and has a kinematic, electrical and sensory design very similar to that of the NimbRo-OP. Powered by a 4-cell lithium polymer battery, ROBOTIS Dynamixel MX series actuators are used for all joints. As shown in the left-most image in Figure 2.6, six MX-106 servos are used for each leg (3 in the hip, 1 in the knee and 2 in the ankle), and three MX-64 servos are used for each arm (2 in the shoulder and 1 in the elbow). Two MX-64 servos also control the pan and tilt of the head. All actuators communicate with a CM740 microcontroller board, flashed with a fully custom firmware as described later in Section 2.2.1, via a minimally daisy-chained star topology RS485 two-wire bus. The CM740 houses the 3-axis accelerometer, gyroscope and (optionally) magnetometer chips. An overview of the electrical architecture of the igus Humanoid Open Platform is given in Figure 2.7.

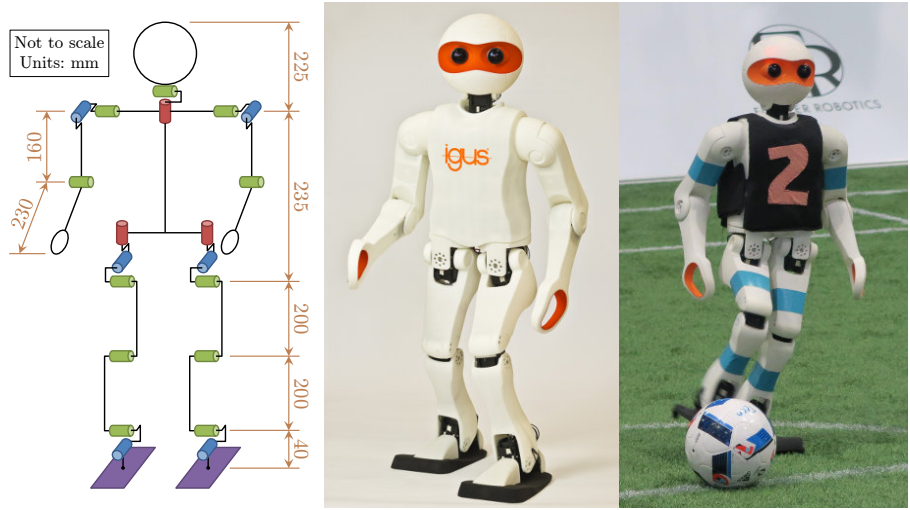

Figure 2.6: The igus Humanoid Open Platform, and its kinematic dimensions. Note that cleats underneath the feet of the robot contribute an extra 2 cm in height not shown in the diagram. The right-most picture shows the platform in action at a RoboCup competition, wearing a special-purpose security jacket to protect it from falls.

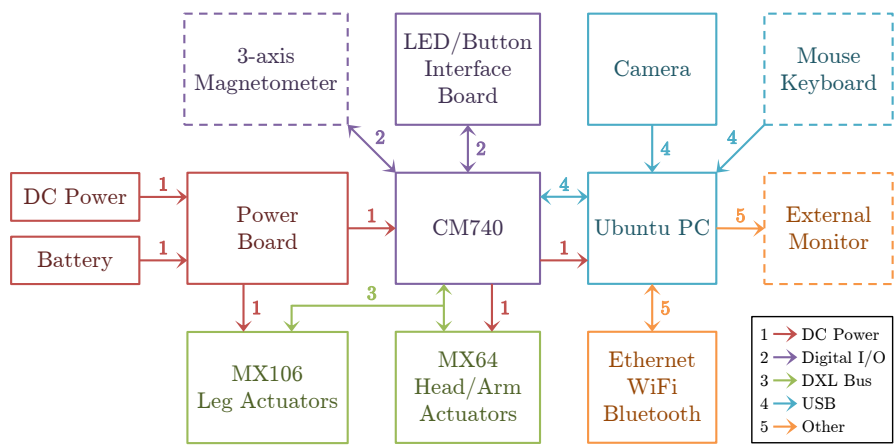

Figure 2.7: Electrical architecture of the igus Humanoid Open Platform

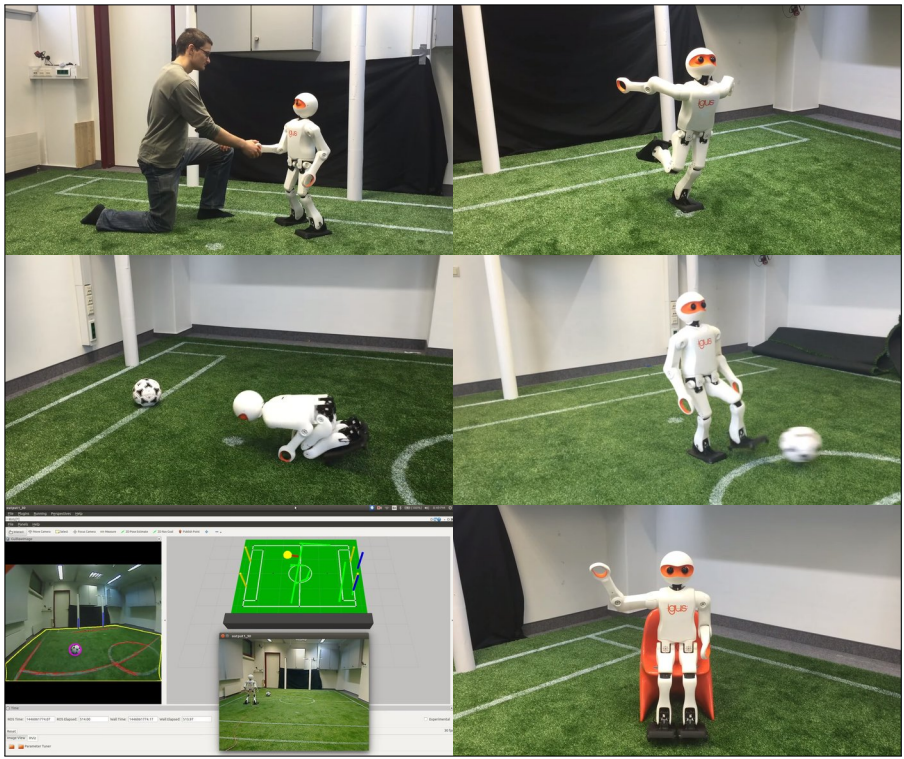

Video 2.2: Introduction video to the child-sized 3D printed igus Humanoid Open Platform, showing human-robot interaction, balancing on one foot, walking, kicking, getting up, and soccer computer vision.
https://youtu.be/RC7ZNXclWWY
*Igus Humanoid Open Platform*

For visual perception, the robot is equipped with a Logitech C905 USB camera fitted with a 150° wide-angle lens. All mechanical parts of the robot were 3D printed out of Nylon-12 (Polyamide 12, PA 12) in layer increments of less than 0.1 mm using a Selective Laser Sintering (SLS) process. This allows for high modularity, production speed, design flexibility and aesthetic appeal. There are no further supporting elements underneath the outer plastic shell. All of the electronics and sensors are housed inside the torso, apart from the camera and USB WiFi adapter, which are located in the head. The robot is nominally equipped with a dual-core Intel Core i7-5500U CPU, which has four logical cores and a base frequency of 2.4 GHz with Turbo Boost up to 3.0 GHz. Available communication interfaces include USB 3.0, HDMI, Mini DisplayPort and Gigabit Ethernet.

Other than a slight adjustment of dimensions, two noteworthy changes in the kinematics were made between the NimbRo-OP and igus Humanoid Open Platform. The feet are more strongly reinforced, increasing their rigidity to elastic deformation, and the mounting points for the ankles are further back on the foot. This reduces the effect that the backlash in the ankle pitch actuator has on the stability and balance of the robot during walking, and avoids similar effects caused by bending of the feet under unbalanced loads.

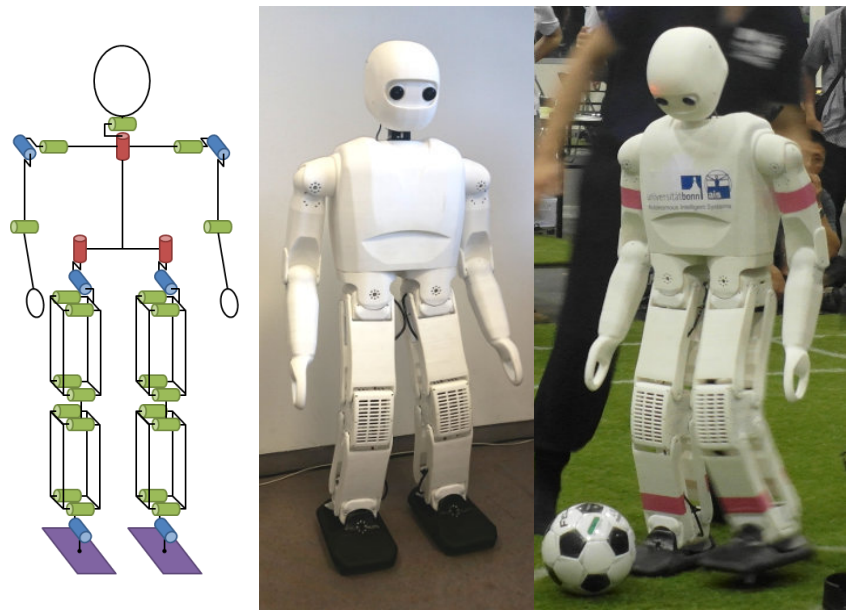

Figure 2.8: The NimbRo-OP2 robot, and its kinematic layout. Note the use of parallel kinematics in the thigh and shank links of the robot. All 16 thigh and shank servos are located in the knees, and external brass gearing is used in the hip yaw, hip roll and ankle roll joints.

More details on the igus Humanoid Open Platform hardware and software and design ideas can be found in Allgeuer et al. (2015) and Allgeuer et al. (2016). Videos of the platform in action at RoboCup 2016 and 2017 can be found in Videos 1.1 and 1.3, respectively, on pages 4 and 6.

### 2.1.2.5  *NimbRo-OP2*

The NimbRo-OP2, shown in Figure 2.8 and Video 2.3, was designed and constructed in 2017, and like the igus Humanoid Open Platform was almost entirely 3D printed from Nylon-12 (Polyamide 12, PA 12) by an industrial grade SLS printer. The mechanical design focused mainly on simplicity, cost reduction, and the achievement of part rigidity through geometric features like ribs and lips, as opposed to just through part thicknesses. The robot is 134.5 cm tall, weighs 18.5 kg, and uses a parallel kinematic link arrangement in its legs for increased stability. Powered by a 4-cell lithium polymer battery, the NimbRo-OP2 has a total of 34 ROBOTIS Dynamixel MX-106 actuators, of which each knee contains eight—four in parallel to actuate the thigh, and four in parallel to actuate the shank. The use of multiple servos per joint, sometimes additionally in combination with external brass gearing (see Figure 2.9), greatly boosts the amount of available torque, and thereby increases the range of dynamic motions that the robot can perform. In total, 13 MX-106 actuators are used in each leg (3 per hip, 8 per knee, and 2 per ankle), 3 are used in each arm (2 per shoulder, and 1 per elbow), and 2 are used in the neck (pan and tilt of the head). As for the igus Humanoid Open Platform, all actuators are connected via a minimally daisy-chained star topology RS485 two-wire bus, and communicate with the PC via a CM740 microcontroller board.

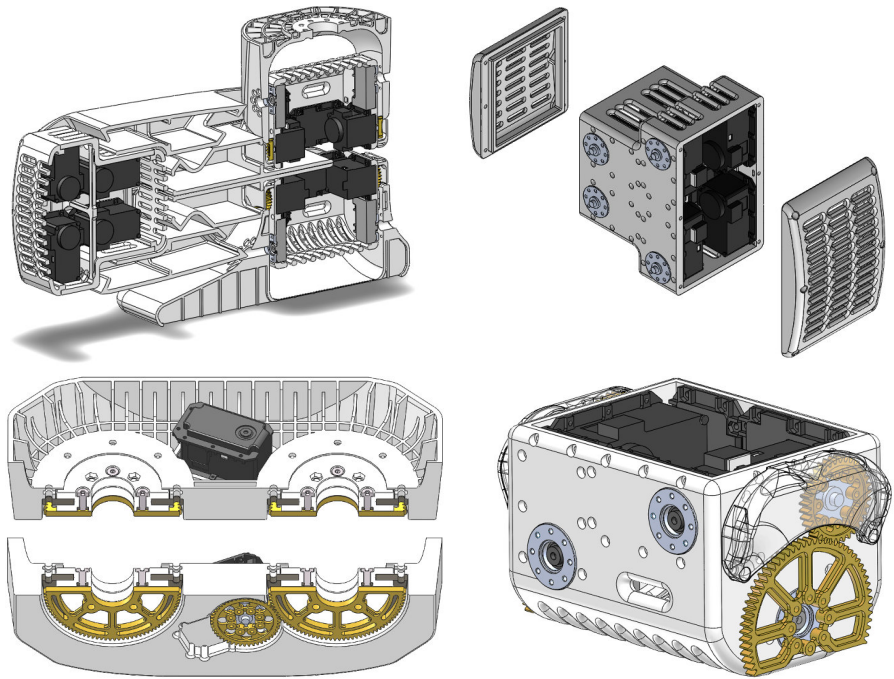

Figure 2.9: Elements of the mechanical design of the NimbRo-OP2 robot. The top row shows a cross-section through one of the legs of the robot, and an exploded view of the knee assembly. The bottom row shows the external brass gearing used in the hip yaw (top/bottom cross-sectional views on the left side), and the equivalent gearing used in the hip and ankle roll joints (right side).

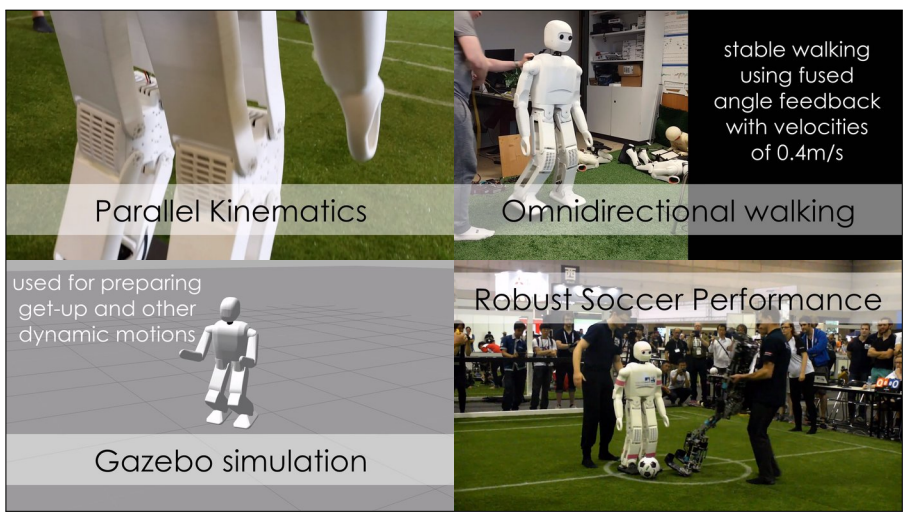

Video 2.3: Introduction video to the 3D printed AdultSize NimbRo-OP2 robot, showing details of the mechanical hardware, omnidirectional walking, simulations in Gazebo, and soccer playing abilities.
https://youtu.be/WJKc56uUuF8
*NimbRo-OP2: Grown-up 3D Printed Open Humanoid Platform for Research*

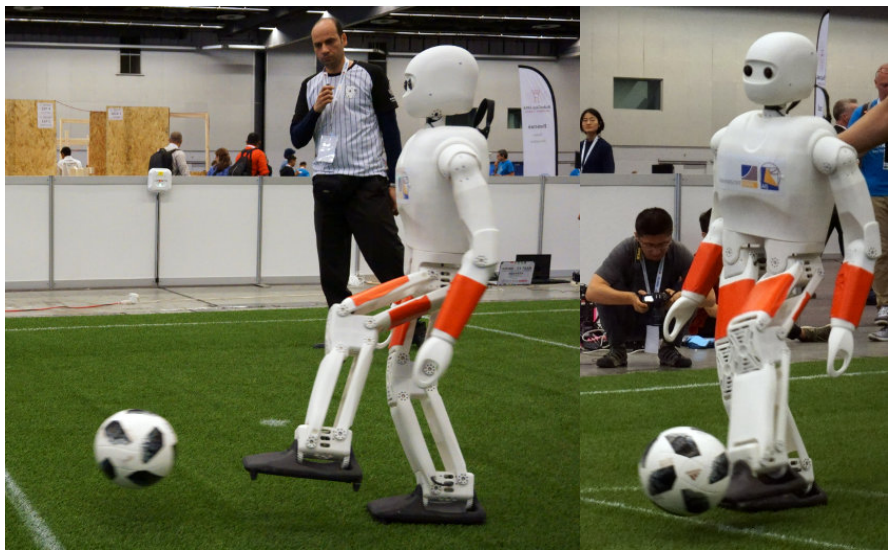

Figure 2.10: The NimbRo-OP2X robot in action at RoboCup competitions.

To perceive the environment, the robot utilises a Logitech C905 USB camera in its head, fitted with a 150° wide-angle lens. All of the remaining sensors and electronics, including the PC, are housed within the torso. The electrical architecture of the NimbRo-OP2 is very similar to the one shown in Figure 2.7 for the igus Humanoid Open Platform, with the main exception that the DC power for all servos is routed through a BTS555 high current power switch Metal-oxide-semiconductor Field-effect Transistor (MOSFET), instead of the lower capacity power board. With a quad-core Intel Core i7-7700HQ PC (8 logical cores), there is enough processing power available for even demanding applications, and thanks to the integration of a dedicated graphics card, machine learning applications are also possible.

More details on the NimbRo-OP2, in particular in terms of the mechanical and electrical designs of the robot, can be found in Ficht et al. (2017). Videos of the robot in action at RoboCup 2017 and 2018 can be found in Videos 1.2 and 1.4, respectively, on pages 4 and 6.

### 2.1.2.6 *NimbRo-OP2X*

The NimbRo-OP2X (see Figure 2.10 and Video 2.4) is the next-generation version of the NimbRo-OP2 robot. Although the appearance of the two robots is highly similar, the NimbRo-OP2X is in fact a wide-scale mechanical redesign of the NimbRo-OP2, that also includes some further electrical and hardware upgrades, like for example the inclusion of cooling fans in the torso and both knees (see Figure 2.11). The mechanical redesign was necessitated by the choice of using the newer ROBOTIS Dynamixel XM540-W270 servos, which have a different form factor and mounting options to the MX-106 servos. In addition to a redesigned gearbox, the XM540-W270 servos have a 29% increase in output torque over the MX-106 servos, and a metal casing that provides better heat dissipation and higher general mechanical durability. At a height of 135 cm and weight of approximately 19.0 kg, the same (parallel) kinematics as for the NimbRo-OP2 was chosen (see Figure 2.8), so the number of servos stayed the same at 34.

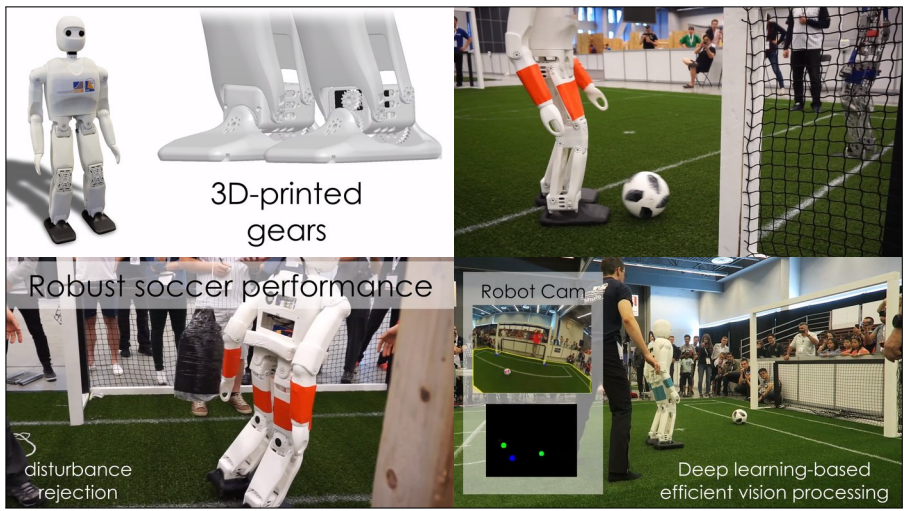

Video 2.4: Introduction video to the 3D printed AdultSize NimbRo-OP2X robot, showing the robot mechanics and specifications, walking and kicking motions, deep learning vision system, and robust soccer performance.
https://youtu.be/qmYSt3nVJpA
*NimbRo-OP2X: Adult-sized Open-source 3D Printed Humanoid Robot*

The mechanical parts are all, as before, 3D printed from Nylon 12 using an SLS process. The freedom in design that 3D printing gives, allowed the overall appearance of the NimbRo-OP2X robot to be made aesthetically rounder and smoother than that of the NimbRo-OP2. The structural components are however not the only parts of the robot that are 3D printed in the NimbRo-OP2X. The procurement and milling of the brass gears was a great bottleneck in the construction of the NimbRo-OP2, so the NimbRo-OP2X has gears that are 3D printed from an iglidur I6-PL material.[2] This material is specifically designed for gears, and accordingly has good mechanical properties for such applications—in particular a low coefficient of friction and no need for lubrication. A double helical design was chosen (see Figure 2.11) for the gears so as to minimise backlash and allow for smoother engagement (than spur gears) of the gear teeth. Specifically a *double* helical design was chosen, so that the generated helical gear axial forces counteract each other and do not apply undue stresses on the servos and servo axles. The printed plastic gears require a larger module and thickness than the equivalent NimbRo-OP2 brass gears, but are only around 54% of the weight, and most importantly, are much quicker, easier and cheaper to produce. The amount of backlash in the helical 3D gears is also comparable to the amount of backlash in the brass spur gears, despite the difference in module.

---

2 https://www.igus.de/product/14950

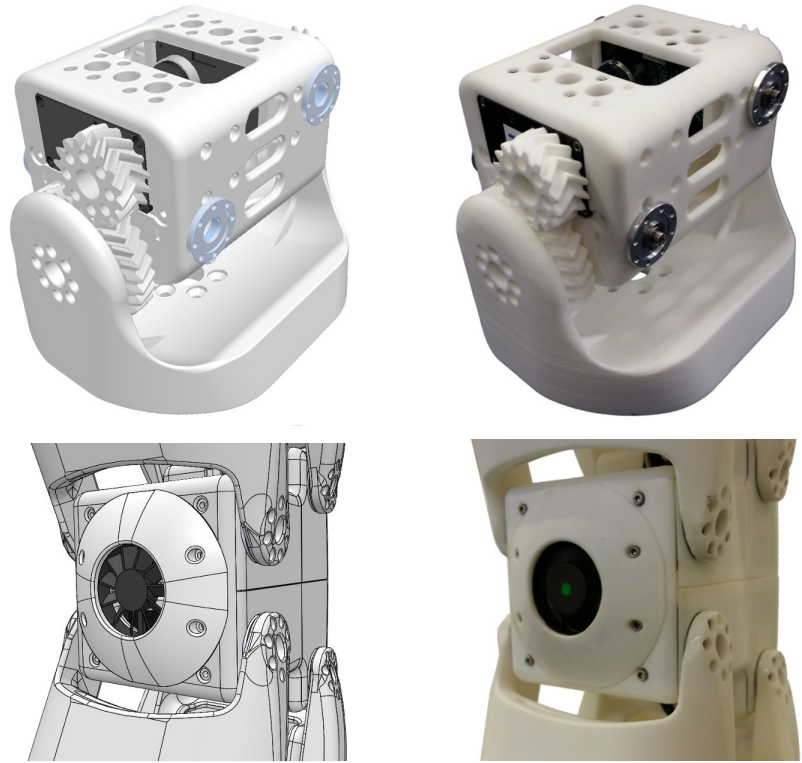

Figure 2.11: Elements of the mechanical design of the NimbRo-OP2X robot. The top row shows an inverted view of the hip assembly (left: CAD, right: 3D printed), and in particular shows the 3D printed double helical gears. The bottom row shows the back of the knee assembly, and in particular the inbuilt cooling fan.

Further improvements of the NimbRo-OP2X over the NimbRo-OP2 include the incorporation of a larger torso cavity, so that any standard Mini-ITX motherboard can be used, instead of just a limited range of small form factor PCs. This change also means that any moderately sized Graphics Processing Unit (GPU) can be fitted and used in the robot. Nominally, the NimbRo-OP2X is fitted with a hex-core Intel Core i7-8700T PC (12 logical cores), and an Nvidia GeForce GTX 1050 Ti graphics card, so an even greater amount of processing power than on the NimbRo-OP2 is available for applications.

More details on the NimbRo-OP2X, in particular in terms of the mechanical and electrical designs of the robot, can be found in Ficht et al. (2018). Footage of the robot in action at RoboCup 2018 can be found in Video 1.4 on page 6.

## 2.2 ROBOT SOFTWARE

All of the methods and approaches for sensor processing, walking, balancing, and so on, described in this thesis have been implemented in a common ROS-based software framework that runs on all of the aforementioned robots. A brief overview of the software framework is given here, starting with the custom firmware that was written for the CM740 microcontroller board.

### 2.2.1 Custom CM740 Firmware

As described in Section 2.1.1.2, the role of the CM740 is primarily to interface the PC with the Dynamixel servo bus, and to provide Inertial Measurement Unit (IMU) information to the PC software from its 3-axis gyroscope, accelerometer, and (optionally) magnetometer sensors. The CM740 comes with stock firmware from ROBOTIS that is intended for the DARwIn-OP robot, but due to a wide variety of reasons including a general lack of performance, features and reliability (and the general presence of bugs in the implementation of the Dynamixel protocol in the servos), a new fundamentally redesigned custom firmware was written by the author that solves all of these problems (Allgeuer, 2018a). Amongst other things, the new firmware led to a reduction in data latency, and an increase in data throughput for communications between the PC and the servos.

One change in particular that was made was an extension of the Dynamixel protocol for communications between the CM740 and the PC. This extension translated into significant gains in bus stability and error tolerance, as well as into time savings for bulk reads of servo data. It should be noted however, that the upgraded communications protocol is fully back-compatible with the standard Dynamixel protocol, even if PC software that was written to use the extended protocol is then obviously no longer compatible with the stock firmware.

While the IMU values of the CM740 are polled by the PC software at approximately 100 Hz with sometimes unavoidably high jitter due to software, USB and Dynamixel protocol constraints, the sensors themselves on the board are polled extremely reliably by the microcontroller at higher frequencies using interrupts, e.g. at 259.3 Hz for the gyroscope and accelerometer sensors. In order to make the most of the available sensor data and avoid any temporal aliasing effects from downsampling, an averaging scheme has been implemented for the IMU data. The high frequency IMU data that is polled by the microcontroller is placed in a small circular buffer in on-chip Static Random-access Memory (SRAM), and is emptied every time the CM740 microcontroller receives a request from the PC for the IMU data. The values returned to the PC are then the averages of the contents of the circular buffer right before they are cleared. This averaging scheme helps fight noise by effectively applying a small low-pass filter to the data, and seeing as the data is generally further filtered or numerically integrated in some way on the PC anyway, this effectively allows the information contained in the higher frequency data to be transmitted with minimal qualitative loss over a high jitter 100 Hz polling connection.

### 2.2.2   NimbRo-OP ROS Software

When Dynaped and Copedo were first constructed and used in Robo-Cup competitions, they were running a custom Windows soccer software framework on Windows XP on Sony Vaio UX Micro PCs. With the advent of the NimbRo-OP in 2012 however, a switch was made to Linux Ubuntu, and an initial software release was made as a series of patches against the ROBOTIS DARwIn-OP software framework version that was current at the time (Schwarz et al., 2012; Team NimbRo, 2012).[3] Soon after, development began on the NimbRo-OP ROS Software framework, which aimed to be a standardised humanoid robotic software framework based on C++ and the common Robot Operating System (ROS) middleware (Quigley et al., 2009).[4] The software framework came with sample applications geared towards robot soccer, but tried to be as abstracted and general as possible on the front of generic robot control, motions and walking. The software was released in November 2013 (Team NimbRo, 2013), and all further development occurred under the new name 'Humanoid Open Platform ROS Software'. The NimbRo-OP ROS Software framework is described in detail in Allgeuer et al. (2013), and more information about the behaviour control frameworks that were developed as part of the project can be found in Allgeuer and Behnke (2013).

### 2.2.3   Humanoid Open Platform ROS Software

In 2014, as the design focus shifted from the NimbRo-OP to the successor igus Humanoid Open Platform, the NimbRo-OP ROS Software was frozen in time, and development continued under the new name Humanoid Open Platform ROS Software (or sometimes, igus Humanoid Open Platform ROS Software). This C++ software framework was generalised and applied to all further constructed robots over the years, and grew to contain the implementations of all the

---

3  Associated video: https://youtu.be/tn1uSz6YseI

4  ROS website: http://www.ros.org/about-ros

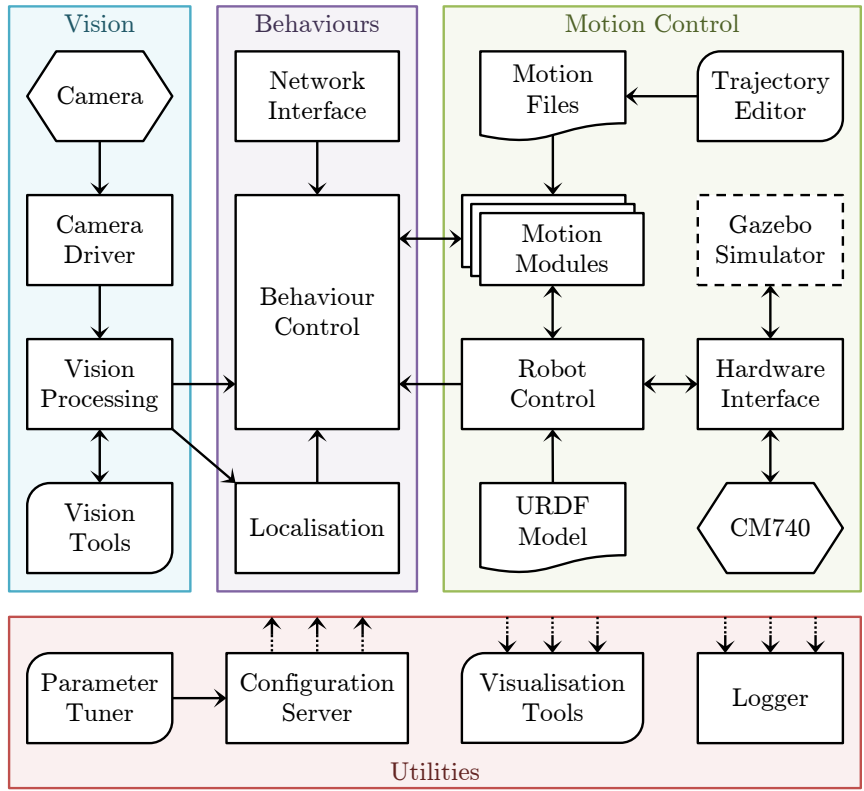

Figure 2.12: Architecture of the Humanoid Open Platform ROS Software.

algorithms and methods that were developed and are presented in this thesis. In addition to running on the newer NimbRo-OP2 and NimbRo-OP2X robots, the software was also retrofitted to Dynaped, as well as Copedo, when it was refurbished as a jumping AdultSize robot. The Humanoid Open Platform ROS Software has been released on multiple occasions to GitHub (Team NimbRo, 2018a), and has been successively detailed in numerous publications, including

- "Child-sized 3D Printed igus Humanoid Open Platform" by Allgeuer et al. (2015),

- "The igus Humanoid Open Platform: A Child-sized 3D Printed Open-Source Robot for Research" by Allgeuer et al. (2016),

- "NimbRo-OP2: Grown-up 3D Printed Open Humanoid Platform for Research" by Ficht et al. (2017),

- "NimbRo-OP2X: Adult-sized Open-source 3D Printed Humanoid Robot" by Ficht et al. (2018).

A wide range of further publications, including Allgeuer and Behnke (2013), Allgeuer and Behnke (2014), and Farazi et al. (2015), detail more specific aspects of the software.

Figure 2.12 gives an overview of the fundamental architecture of the Humanoid Open Platform ROS Software. At the heart of the entire soft-ware framework is the robot control ROS node (`robotcontrol`), which runs a nominally 100 Hz hard real-time control loop that manages sensor and actuator communications, state estimation and motion generation. A Unified Robot Description Format (URDF) model of the robot is loaded from file—so as to abstract away which robot is executing the software—and a per-launch configurable selection of motion module plugins use this information in addition to sensory perception to generate dynamic motions for the robot. One example of such a motion module is the motion player, which reads in timed sequences of keyframes from so-called motion files in YAML Ain't Markup Language (YAML) format, and can play them back, when triggered, as cubic spline trajectories in the joint space. A further plugin scheme is used for the hardware interface, which takes all output commands from the motion modules, transmits them via the CM740, and then queries the CM740 again for new sensor data for the next execution cycle of the motion modules. To support testing on non-robot PCs, drop-in replacements for the standard hardware interfaces exist that only control a virtual dummy robot, or link to a physically simulated one in a Gazebo simulator.

The higher level planning and artificial intelligence controlling the actions of the robot (and thereby controlling which motion module is activated at which time) is implemented in a separate behaviour control ROS node, which for the case of RoboCup competitions is given by the *walk and kick* node. This name derives satirically from the humble beginnings and intentions of that behaviour node. Visual perception and processing occurs in its own node, and in the case of robot soccer also ties in the self-localisation of the robot on the field. All of the nodes communicate with clearly defined message paths and types via the ROS topic management system. Request and response architectures are supported via ROS service calls and action servers.

A number of framework-wide helper nodes have also been implemented, of which the visualisation tools, configuration server and logger find the most prominent use. The configuration server is a centralised storage location and manager for software parameters, and replaces the ROS parameter server for dynamic reconfiguration of the software. A Graphical User Interface (GUI) front end called the parameter tuner is available for the configuration server, and allows all configuration parameters to be viewed and edited in real-time. As the robots are generally disconnected from visualisation tools during operation, and any encountered problems can be difficult to reproduce, a data logger node is permanently running in the background. The logger uses a round robin scheme of ROS bags to store the data, and automatically cleans up old bag files when disk space or number quotas are reached. Bag files can also manually be recorded from the plotter widget of the *rqt* visualisation window (see Figure 2.13), which is a visualisation tool for plotting live data waveforms of the

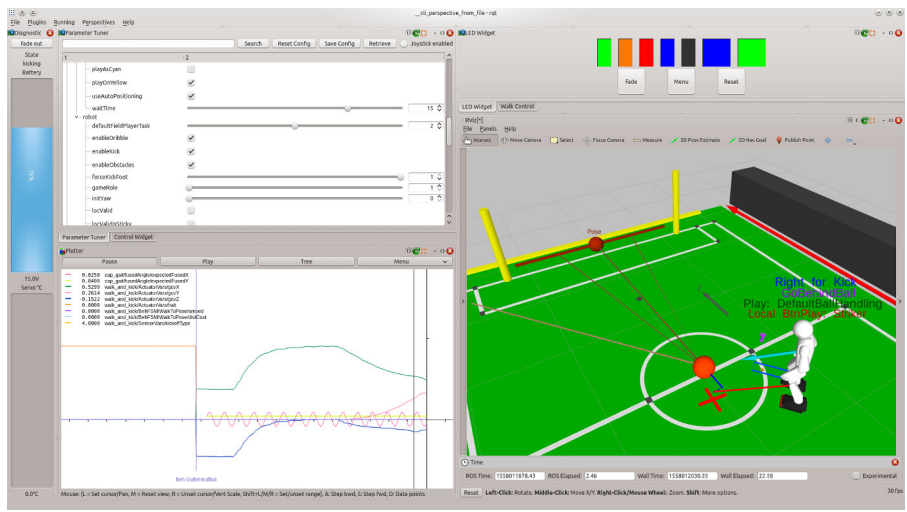

Figure 2.13: Sample layout of the rqt visualisation window of the Humanoid
Open Platform ROS Software, showing the 3D RViz visualisation
(right), LED status and button widget (top right), parameter tuner
(top left), and plotter (bottom left).

state of the robot and software. Typically, the visualisation tools and
parameter tuner are run on an external PC, but the configuration server
and logger are run on the robot.

The software was developed under the guise and target application
of humanoid robot soccer, but with suitable adaptation of the beha-
viour control, vision processing and motion module sections, software
for virtually any other application can be realised. The robot control
node that forms the core of this software framework has been run on
many robots beyond the scope of this thesis, including for instance the
hybrid wheeled-legged robot Momaro that was used for the DARPA
Robotics Challenge (DRC) competition (Schwarz et al., 2017). The flex-
ibility of the software is possible because of the strongly modular way
in which the software was written, greatly supported by the natural
modularity of ROS. For instance, in addition to the hardware interface
and motion module plugin schemes already described previously, the
gait motion module uses a further gait engine plugin scheme to allow
the smooth integration of multiple gaits into the code without any
unnecessary code duplication and/or inconsistencies.

In specific reference to the latest open source code release (Team
NimbRo, 2018a), we can summarise that

- Chapter 3 (Actuator Control) is implemented jointly by the main
  robot control node and the hardware interface that it loads,

- Chapter 4 (Sensor Calibration) is implemented in the hardware
  interface,

- Chapters 5 to 7 (Rotation Formalisms in 3D) is implemented in
  the rot_conv package, and has also been separately released as
  Allgeuer (2018c) and Allgeuer (2018d),

- Chapter 9 (Humanoid Kinematics) is implemented in the stand-alone `humanoid_kinematics` package,

- Section 10.1 (Attitude Estimation) is implemented in the `rc_utils` package, but is used in the hardware interface, and has also been separately released as Allgeuer (2016),

- Section 10.2 (Pose Estimation) is implemented in the code of each of the gait engines separately,

- Chapter 11 (A Central Pattern Generator for Walking), and Chapter 12 (Capture Step Gait), and Chapter 13 (Direct Fused Angle Feedback Controller) are all implemented in the `cap_gait` gait engine, which is loaded via the `gait` motion module,

- Chapter 14 (Keypoint Gait Generator) and Chapter 15 (Tilt Phase Controller) are implemented in the `feed_gait` gait engine, which is loaded via the `gait` motion module,

- Appendix A (Mathematical Functions and Filters) is implemented in the `rc_utils` utilities package, and is used throughout the entire remainder of the code.

This list demonstrates how all the algorithms and methods developed in this thesis have been implemented in the Humanoid Open Platform ROS Software framework and thereby released online.

# 3

## ACTUATOR CONTROL

In order for any motions generated for a robot to work as intended, robust, accurate and efficient tracking of the computed actuator trajectories is paramount. Even though the Dynamixel servo motors have an inbuilt configurable Proportional-Integral-Derivative (PID) position control loop, it is hard to tune this loop so that it works appropriately and equally in all situations. The approach of simply using very high gain settings, thereby making the robot very stiff, is not appropriate because it leads to problems with

- Overloading and overheating, leading to temporary or even permanent servo failure,

- Little to no damping in the motions, leading to motions that are not smooth, and prone to unwanted oscillation, and

- Increased self-disturbances of the robot, due to much harsher and sharper contacts—in particular impacts with the ground.

One fundamental limitation of using just the PID control loops inside each servo is that each control loop is operating completely independently. They have no global picture of the state of the robot, what the robot is trying to achieve, and what amount of torque they can each expect to require to follow their respective commanded trajectories. PID control loops are also agnostic to the specific contributing factors *why* a servo does not necessarily manage to follow its input commands, and can only react to disturbances when they have already been measured in the output, at which point it is too late to avoid them, and difficult to avoid them in a systematic way. In this chapter, we discuss the pipeline of how joint commands are generated, compensated and sent out to the servos, and describe in detail the *actuator control scheme*, which seeks to use torque estimation and feed-forward compensation (in addition to servo-internal proportional feedback) to improve actuator tracking in a systematic targeted way.

## 3.1  ROBOT CONTROL SCHEME

As described in Section 2.2.3, the robot software is structured in such a way that all motion generation and servo control occurs in a single Robot Operating System (ROS) node, namely the robot control node. This node runs a hard 100 Hz real-time control loop that generates joint commands and joint effort commands using an array of so-called *motion modules* (one of which for instance is the `gait` module).

While the joint command for each joint consists of a required angular position, velocity and acceleration, the joint effort command consists of a dimensionless scalar value in the range $[0, 1]$ that expresses how stiff the joint should be. In addition to these per-joint commands, the motion modules also return two global support coefficients, one for the left leg and one for the right leg, which are dimensionless scalar values in the range $[0, 1]$ that express the proportion of the weight of the robot that is expected to be supported by each leg. Ideally, the support coefficients should always sum to 1.0. The generated joint, joint effort and support coefficient commands are processed by the actuator control scheme, and the resulting raw servo commands, consisting of a commanded position and P gain, are sent out to the actuators via the CM740 and the Dynamixel bus. After subsequently reading in and processing the latest servo encoder and sensor (e.g. IMU) data values, the cycle repeats.

It is important to note that the 100 Hz loop is 'hard' in the sense that it is enforced that the cycles execute exactly in 10 ms increments. This means for example that if one cycle takes 11 ms to complete for whatever reason, the next one then only has 9 ms of time allocated to compensate for that. This works because every cycle ends in a call to `sleep`, and the duration of this sleep can be adjusted dynamically. The only requirement is that the majority of all cycles naturally execute in less than 10 ms. This is easily the case in the software, and is generally only breached if errors or delays occur in waiting for servo response packets on the Dynamixel bus.

### 3.1.1   Main Robot Control Loop

We now take a closer look at the main robot control loop cycle, with a focus on how the servo commands are managed, and what exact processing they are subjected to. In every 10 ms cycle of the loop, the following actions are executed in order:

1. Each loaded motion module plugin is first queried (in order) whether it should be active in this cycle, and if so, it is executed. The gait motion module, for example, returns that it should be inactive in cycles where the robot should not be walking. Each motion module can return position commands for any subset of the available joints, and can optionally also return velocity and/or acceleration commands for these joints. For each joint, the last motion module to set a position command 'wins'. If no velocity and/or acceleration commands were provided along with that position command, they are automatically calculated based on first and second derivative Savitzky-Golay filters (see Appendix A.2.1.2). The last motion module to set the two support coefficients likewise determines which final coefficients are returned.

2. Based on the final commands returned from the motion modules, the corresponding final commands for any virtual joints present in the robot model are computed. Virtual joints that alias or are the reverse of other joints can be necessary, for example, if parallel leg kinematics are present in the robot.

3. Based on the final joint commands, final support coefficients and a model of gravity, feed-forward joint torques are calculated for each joint using an inverse dynamics approach (see Section 3.2.2). These feed-forward torques estimate how much torque is required as a result of gravity and inertial effects (but not friction) to follow the commanded trajectory at the commanded position.

4. The feed-forward torques for any virtual joints (refer to Step 2) are calculated based on the feed-forward torques computed in Step 3.

5. The joint effort commands from Step 1 are slope-limited for safety (see Appendix A.2.4.2), and converted to servo P gains using a proportional scale factor.

6. Based on the final joint commands and feed-forward joint torques, appropriate raw servo position commands are calculated using the servo motor model (see Section 3.2.1). The servo motor model, amongst other things, compensates for the internal servo joint friction. Saturation is applied to the final command to ensure that the position commands do not exceed the capacity of the servo.

7. The final servo position commands and P gains are sent to the servos on the Dynamixel bus through the CM740. Given a deviation between the current servo position and the new commanded position, the servos internally apply a torque proportional to the magnitude of the deviation (and the P gain of course) to correct it.

8. The joint encoder values, and other sensor data like the Inertial Measurement Unit (IMU) data, battery voltage, and CM740 temperature, are queried from the servos and CM740 via the Dynamixel bus.

9. The acquired sensor data is processed into the forms required by the motion modules in Step 1, for example to estimate the attitude of the robot as described in Section 10.1.

10. The robot control loop sleeps until the next multiple of 10 ms is reached, and then starts again.

The robot control loop is organised the way it is, i.e. first writing then reading, as the process of reading from the Dynamixel bus is comparatively long and unpredictable, and by doing it second, the writes in Step 7 can despite this remain relatively consistently timed.

## 3.2   ACTUATOR CONTROL SCHEME

Referring to the description of the main robot control loop given just previously, the actuator control scheme encompasses the actions of Steps 3 to 7. In this section, we specifically present the servo motor model (Step 6), and how inverse dynamics can be used to perform feed-forward joint torque estimation (Step 3).

### 3.2.1   Servo Motor Model

The servo motor model is a mathematical and electromechanical model of the behaviour of the used servo motors that incorporates both Direct Current (DC) motor and mechanical friction characteristics. It is used to allow for compensated control of the servos.

#### 3.2.1.1   DC Motor Model

The Dynamixel servo motors used in the robots are built around standard brushed DC motors driven by a Pulse Width Modulation (PWM) approach. This means that we can model the motor as an effectively continuously variable voltage $V_m$ applied to an armature resistance in series with an ideal motor winding, as shown in Figure 3.1. The armature current $I$ flows through the armature resistance $R$, producing a voltage drop of $RI$ from Ohm's law, and further through the ideal motor winding, which produces the torque $\tau$ on the output shaft. As a result of the rotation of the output shaft within the magnetic field of the motor, a voltage $E$ is induced across the winding (due to electromagnetic induction) and is referred to as the back-Electromotive Force (EMF) voltage. The back-EMF voltage is proportional to the angular velocity $\omega$ of the output shaft, and is explicitly given by

$$E = k_e \omega, \tag{3.1}$$

where $k_e$ is the so-called back-EMF constant of the motor. We conclude from Kirchhoff's voltage law, as applied to Figure 3.1, that

$$V_m = RI + E \tag{3.2a}$$
$$= RI + k_e \omega. \tag{3.2b}$$

As a simple model of a DC motor, we know that the torque generated by the motor is proportional to the current flowing through its windings, i.e.

$$\tau = k_t I, \tag{3.3}$$

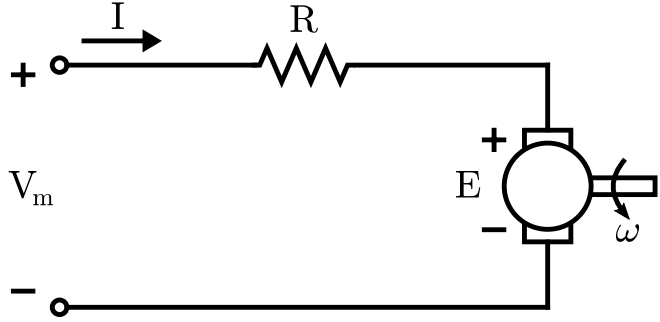

Figure 3.1: A simple electrical model of a DC motor. The voltage $V_m$ applied to the motor results in an armature current $I$ flowing through the armature windings of resistance $R$, and makes the motor spin with an angular velocity of $\omega$. A back-EMF voltage $E$ is generated as a result of the rotation of the output shaft in the magnetic field.

where $k_t$ is the so-called motor torque constant, and is generally considered to be equal to the back-EMF constant $k_e$. As a result,

$$I = \frac{\tau}{k_t},$$  (3.4)

and substituting this into Equation (3.2b) gives

$$V_m = \frac{R}{k_t}\tau + k_e\omega.$$  (3.5)

This is a basic DC motor model, prior to the consideration of gearing and friction.

### 3.2.1.2  Friction Model and Gearing

While $\omega$ is the angular velocity of the shaft of the DC motor, the effect of gearing means that the angular velocity $\dot{q}$ of the servo motor joint axis is proportionally different, i.e.

$$\omega = k_g\dot{q},$$  (3.6)

where $k_g$ is the gear reduction ratio. The output torque $\tau_q$ at the joint axis is also scaled by the effect of gearing,

$$\tau_q = k_g\tau,$$  (3.7)

and is the sum of the true output torque $\tau_d$, and the torque $\tau_f$ that is required to overcome friction. Thus, by rearrangement of Equation (3.7),

$$\tau = \frac{1}{k_g}(\tau_d + \tau_f).$$  (3.8)

From Equations (3.5) and (3.6), this leads to

$$V_m = \frac{R}{k_g k_t}\tau_d + k_g k_e \dot{q} + \frac{R}{k_g k_t}\tau_f.$$  (3.9)

The main question now is how to model the friction torque $\tau_f$ in terms of the joint angular velocity $\dot{q}$. As discussed in Schwarz and Behnke (2013), a suitable friction model involving both static and Coulomb friction terms, and an exponential transition (Stribeck curve) between them, is given by

$$\tau_f = \mathrm{sgn}(\dot{q})\big(\beta\tau_s + (1-\beta)\tau_c\big) + c_v\dot{q}, \tag{3.10}$$

where $\tau_s$ is the limit static friction torque, $\tau_c$ is the Coulomb friction torque, $c_v$ is the viscous friction constant, and $\beta$ is the interpolating factor

$$\beta = \exp\left(-\left|\frac{\dot{q}}{\dot{q}_s}\right|^{\delta}\right), \tag{3.11}$$

where $\dot{q}_s$ is the transition velocity between static and Coulomb friction, and $\delta$ is an empirical constant in the range 0.5 to 1.0 based on the material surfaces and properties. Based on Schwarz and Behnke (2013), a choice of $\dot{q}_s = 0.1\,\mathrm{rad/s}$ and $\delta = 1.0$ was made for the servos in this thesis.

Substituting Equation (3.10) into Equation (3.9) gives

$$V_m = \frac{R}{k_g k_t}\tau_d + \left(k_g k_e + \frac{Rc_v}{k_g k_t}\right)\dot{q} + \frac{R\tau_c}{k_g k_t}\mathrm{sgn}(\dot{q})(1-\beta) + \frac{R\tau_s}{k_g k_t}\mathrm{sgn}(\dot{q})\beta,$$

which, with the definition of appropriate constants $\hat{\alpha}_0$, $\hat{\alpha}_1$, $\hat{\alpha}_2$ and $\hat{\alpha}_3$, can be simplified to

$$V_m = \hat{\alpha}_0\tau_d + \hat{\alpha}_1\dot{q} + \hat{\alpha}_2\,\mathrm{sgn}(\dot{q})(1-\beta) + \hat{\alpha}_3\,\mathrm{sgn}(\dot{q})\beta. \tag{3.12}$$

Recalling from Equation (3.11) that $\beta$ is simply a function of $\dot{q}$ (the angular velocity of the servo motor joint axis), we can see that the effective voltage applied to the motor can therefore be expressed as a function of $\dot{q}$, $\tau_d$ (the true joint axis output torque) and four fixed constants.

### 3.2.1.3 *Compensated Motor Control*

The Dynamixel servo motors used in the robots are position-controlled, and have an internal position-based PID loop with configurable gains for control purposes. In order to centralise the control of the servos and overcome the limitations identified on page 41 of just using servo-level PID control loops, we nominally configure each servo to only use proportional (P) control, and find an approach to use the servo model for global control instead.

In pure proportional mode, the effective voltage applied to the motor is proportional to the battery voltage and the motor PWM duty cycle, which in turn is proportional to the position error and P gain. Thus, with incorporation of a proportionality constant $K_c$ that governs the scaling from position errors to PWM duty cycles, the effective motor voltage $V_m$ applied by the servo electronics to the DC motor at any given time is given by

$$V_m = V_b K_c K_P (q_d - q),  \tag{3.13}$$

where $V_b$ is the battery voltage, $K_P$ is the P gain, $q_d$ is the desired position of the servo, and $q$ is the current position of the servo. Therefore, if we wish for the servo to follow a trajectory where at some instant the position needs to be $q$, the velocity needs to be $\dot{q}$, and the predicted amount of torque required to follow the trajectory is $\tau_d$,[1] the required setpoint for the position P control loop in order to achieve this is given by

$$\begin{aligned}
q_d &= q + \frac{1}{V_b K_c K_P} V_m, \\
&= q + \frac{1}{V_b K_c K_P} \left( \hat{\alpha}_0 \tau_d + \hat{\alpha}_1 \dot{q} + \hat{\alpha}_2 \operatorname{sgn}(\dot{q})(1-\beta) + \hat{\alpha}_3 \operatorname{sgn}(\dot{q})\beta \right) \\
&= q + \frac{1}{V_b K_P} \left( \alpha_0 \tau_d + \alpha_1 \dot{q} + \alpha_2 \operatorname{sgn}(\dot{q})(1-\beta) + \alpha_3 \operatorname{sgn}(\dot{q})\beta \right), \tag{3.14}
\end{aligned}$$

where the factor of $K_c$ has been absorbed into the constants $\hat{\alpha}_*$ to give $\alpha_*$ (for $* = 0, 1, 2, 3$).

Equation (3.14) embodies the entire servo motor model, and corresponds to Step 6 of the robot control scheme. Given a servo motor, suitable values of $\alpha_*$ are first identified and tuned (see Section 3.2.1.4). Then, given the required $q$, $\dot{q}$ and $\tau_d$ of a joint trajectory in real-time, Equation (3.14) is evaluated (recall that $\beta$ is a function of $\dot{q}$) and the resulting $q_d$ is sent to the servo as its target position. $K_P$ is known as it is the current P gain of the servo, and is calculated from the motion module joint effort command (see Step 5 on page 43), and $V_b$ is known as it is measured by the CM740 and reported to the PC software in Step 8. Note that by design, even though $q_d$ is sent as the target position of the servo, it is *not* intended that the servo actually reaches this position. The whole servo model calculation aims to ensure that if the servo tries to reach $q_d$, then it will instead only reach $q$ (as actually desired) due to the effect of friction and external forces. The only missing step at this point is how to calculate the required $\tau_d$, also known as the feed-forward torque, given the commanded joint trajectories of all the joints of a robot. This is addressed in Section 3.2.2.

### 3.2.1.4 *System Identification*

The tuning of the $\alpha_*$ constants for the servo motors used in this thesis (mainly Dynamixel MX-106 servos) was done in Schwarz and Behnke (2013) using Iterative Learning Control (ILC). As performing motions on an entire free-standing robot is difficult and not suitably repeatable, a test bench was constructed whereby the torso of the NimbRo-OP was fixed to a table, and a realistic hip pitch trajectory was executed and evaluated for tracking accuracy. Estimates of the optimal values of $\alpha_*$ for tracking were successively refined with every execution of the trajectory. A total of 12 ILC iterations were required to converge the estimates of $\alpha_*$ to their final values, and a maximum trajectory deviation of approximately 0.02 rad was achieved. Refer to Schwarz and Behnke (2013) for more details on the method and results. The tuning results for later robots and servo types was not independently published.

---

1 That is, the predicted amount of true output torque on the joint axis required to follow the trajectory from a dynamics perspective (e.g. due to gravity, inertias, contacts, …), not including the torque that is required to overcome joint friction.

### 3.2.2   Feed-forward Torque Estimation

As indicated at the end of Section 3.2.1.3, the only remaining missing link in the pipeline of the application of the servo motor model is the calculation at every instant of the desired true output torque $\tau_d$. This torque is referred to as the feed-forward torque, and the calculation thereof is the subject of Step 3 of the robot control scheme on page 43.

#### 3.2.2.1   *Inverse Dynamics*

Given the desired positions $\mathbf{q}$, velocities $\dot{\mathbf{q}}$, and accelerations $\ddot{\mathbf{q}}$ of all the joints in a robot at some instant in time, it is possible to calculate the associated required joint torques to achieve such a state (under consideration of external force conditions) using inverse dynamics. As discussed in Section 2.2.3 and shown in Figure 2.12, a Unified Robot Description Format (URDF) model of the robot, which specifies all dimensions, offsets, link masses and inertia tensors, is available to the robot control node. This physical model is converted in software to a format that the Rigid Body Dynamics Library (RBDL) (Felis, 2017) can work with. Given any external force and loading conditions of the robot, specified as the magnitude and direction of gravity and any contact forces, the RBDL library can then use the Recursive Newton Euler Algorithm (RNEA) to calculate all corresponding internal joint torques.

#### 3.2.2.2   *Single Support Models*

One difficulty of the inverse dynamics calculation is that it is unable to automatically resolve indeterminate contacts. That is, if both feet of the robot are contacting the ground, then the amount of contact force and torque going through each foot cannot be calculated, and instead needs to be heuristically resolved. This is less a limitation per se of the RBDL library, but more a limitation of the availability of information about the dynamic state of the robot, i.e. the position, velocity and acceleration of the robot's free-floating base relative to the environment. In order to overcome this limitation, we introduce the concept of *single support models* and review the notion of *support coefficients* (see Section 3.1).

A single support model of the robot relative to a particular link is the dynamic model that assumes that the nominated link is rigidly fixed in space (with 6 DoF reaction forces and torques), while the rest of the robot can move freely. One of these single support models is created for the trunk link, as well as for each tip link, i.e. link at an end of the kinematic tree (e.g. foot, hand and head links). At each instant in time, a support coefficient in the range $[0, 1]$ is specified for each single support model, and expresses the proportion of the weight of the robot that is expected to be carried by the associated link at that time. The sum of the support coefficients of all single support models at any instant should always be 1.0 for obvious reasons. Seeing as during walking only the feet of a humanoid robot intentionally touch the ground, often only the support coefficients of the left foot and right foot are considered. These two support coefficients are often just referred to as the support coefficients of the left and right leg respectively. The single support models and associated support coefficients are the heuristic entity that allow the case of indeterminate ground contacts to be resolved.

### 3.2.2.3  *Joint Torque Estimation*

Given the desired positions $\mathbf{q}$, velocities $\dot{\mathbf{q}}$, and accelerations $\ddot{\mathbf{q}}$, the joint torques required to overcome and accelerate the link masses and inertias are first evaluated using the trunk link single support model. No gravity or other external forces are applied in this first inverse dynamics computation, as this is handled separately in a second step, and the trunk link rarely has a non-zero support coefficient anyway.

In the second step, the inverse dynamics of all the available single support models are computed in turn, with just the positions $\mathbf{q}$ and the force of gravity and any other explicitly modelled external forces applied. Based on the principle of superposition, the resulting joint torques from each computation are combined together using the support coefficients as weights, and further additively combined with the joint torques that were computed in the first step (see Figure 3.2 for a visual example). This yields the final feed-forward joint torques that are used for the input $\tau_d$ of the servo motor model. As a necessary simplification for feedback stability reasons, the gravitational acceleration vector is always assumed to point in a fixed downwards direction relative to the trunk in every single support model. Perhaps unintuitively, the error in this assumption is actually empirically less than if orientation feedback is used, due to the complex nature of ground contacts.

It may not immediately be obvious why the application of the joint velocities and accelerations in the first step needs to be separated from the remaining single support model evaluations. The core problem is that the foot-to-ground contacts are not always well-modelled by the assumption that the foot link is completely rigidly fixed in space, i.e. rigidly connected to the ground. If we consider the case that the robot commands a quick adjustment to the ankle pitch of both its feet, then in reality we know that the feet will likely take up a small part of the motion elastically, and react to the rest by temporarily lifting off the ground at one of their edges until the robot starts tipping as a reaction. These non-rigid effects become more pronounced when the feet are located on soft surfaces, like for instance artificial grass. According to the single support models of the feet however, the ankles need to generate an enormous amount of torque in order to accelerate the entire torso and weight of the robot, and be able to change their pitch. Clearly, this does not reflect reality, and overestimates the torques required in order to perform such ankle motions. In the NimbRo-OP robot, this systematic overestimation sometimes led to dangerous oscillatory instabilities, in particular because the NimbRo-OP had very flexible feet. The approach of separating the calculation of the inertial joint torques to a separate evaluation of just the trunk link single support model solved this problem.

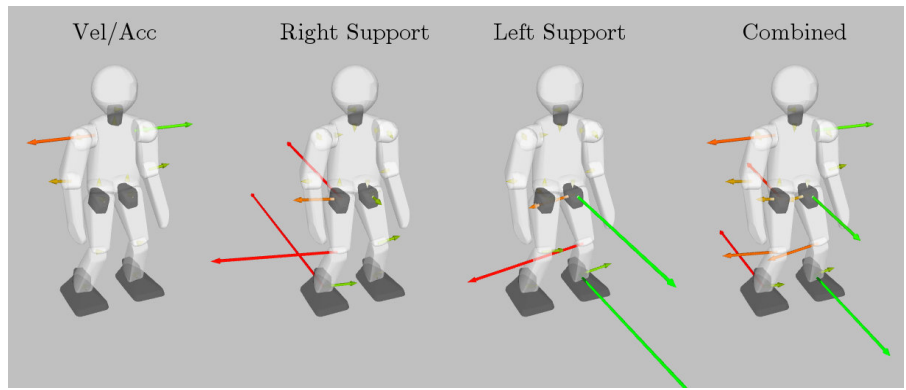

Figure 3.2: Visualisation of the feed-forward joint torque estimation method for the case of a standing robot that is accelerating its right arm forwards and left arm backwards. The left-most image shows the first step, which considers (relative to a fixed trunk) the joint velocities and accelerations, and the resulting torques required to overcome link inertias. The centre images show the torques required to overcome gravity based on the right and left foot single support models. The right-most image shows the result of combining all three previously calculated torques using super-position, where support coefficients of 0.5 are used for each foot.

### 3.2.3 Results

Figure 3.2 shows an example of the feed-forward torque estimation method being applied to a standing posture of the igus Humanoid

Open Platform, in the instant that it started lifting its right arm forwards and left arm backwards. The four 3D visualisations show the pose of the robot in the first frame of the motion, i.e. the pose where the arms are still accelerating to move in their respective directions, along with the feed-forward torques that were calculated for that instant. The left-most image shows the result of the evaluation of the trunk link single support model (trunk link fixed in space), considering only the velocities and accelerations of each joint, and the resulting inertial torques. From the right-hand rule, it can be observed from the green block arrow on the left shoulder that the sign of the torque that is estimated to be required for the shoulder pitch does in fact accelerate the arm backwards, and that a similar but opposite torque on the right shoulder accelerates the right arm forwards. Non-zero torques are also estimated, as expected, for the respective elbow pitches, as the mass of the lower arms accelerating forwards/backwards results in torque being required to hold the positions of the elbow joints.

The centre two images in Figure 3.2 show the result of the evaluation of the right and left foot single support models respectively, considering only gravity, and not the velocities and accelerations of the joints. It can be observed in each case that the joints in the respective support leg required significant torque to hold the weight of the robot, and that the knee joint in the respective free leg actually required a small amount of torque in the opposite direction to keep the lower legs in their position against gravity. The right-most image in Figure 3.2 shows the superposition of the torques calculated in the previous three, with consideration of the commanded support coefficients $\kappa_r$ and $\kappa_l$ for the right and left legs respectively. Specifically, if $\boldsymbol{\tau}_i$ corresponds to the inertia-related torques from the left-most image, and $\boldsymbol{\tau}_r$ and $\boldsymbol{\tau}_l$ correspond to the gravity-related torques from the centre two images, then the final feed-forward torques shown in the right-most image correspond to

$$\boldsymbol{\tau}_d = \boldsymbol{\tau}_i + \kappa_r \boldsymbol{\tau}_r + \kappa_l \boldsymbol{\tau}_l, \tag{3.15}$$

where in this case $\kappa_r = \kappa_l = 0.5$, as it was assumed that both legs carry an equal proportion of the weight of the robot ($\kappa_r = \kappa_l$), and

$$\kappa_r + \kappa_l = 1. \tag{3.16}$$

It can be observed from Figure 3.2 that the final combined feed-forward torques intuitively capture both the effects of gravity and the velocities and accelerations of the joints. A live demonstration of the commanded feed-forward torques during motions of an igus Humanoid Open Platform in Gazebo simulation is provided in Video 3.1.

An example of the effect of the full actuator control scheme on the commanded servo target positions is shown in Figure 3.3. The figure shows the robot in a balancing pose on its right leg, with the corresponding final feed-forward torques $\tau_d$ visualised using 3D block arrows at each joint. The left image shows the commanded joint positions from the motion modules, i.e. the servo target positions that would be sent if the actuator control scheme were disabled, and the right image shows the servo target positions that were computed using Equation (3.14) and subsequently sent out over the Dynamixel bus. The difference in leg joint positions between the left and right images can clearly be identified, and observed to correspond proportionally to the calculated amount of feed-forward torque in each of the respective joints.[2] Note that it makes sense, for example, that the actuator control scheme commands a pose for the right leg that is further forwards and outwards than the pose commanded by the motion modules, as intuitively one can see that the weight of the robot acts to press the leg back into the opposite direction again, towards the truly desired motion module pose.

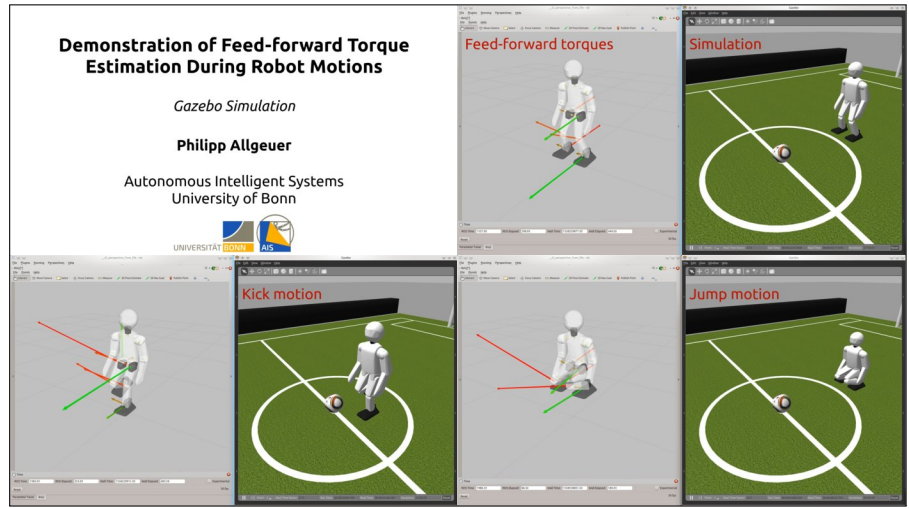

Video 3.1: Demonstration of the actuator control scheme, and in particular feed-forward torque estimation scheme, on the igus Humanoid Open Platform in Gazebo simulation.
https://youtu.be/4AQRvCpquC8
*Demonstration of Feed-forward Torque Estimation During Robot Motions*

Figure 3.4 shows plots of the tracking performance of the left knee pitch joint during a walking experiment of the igus Humanoid Open Platform. A clear difference in tracking performance can be observed depending on whether the actuator control scheme is enabled or disabled, even if due to disturbances and real world inaccuracies the tracking with the scheme enabled is still not perfect. It was observed in the walking experiments that the battery life of the robot was longer with the actuator control scheme enabled, as opposed to disabled, suggesting that it can also increase the energy efficiency of robot motions. This observation is empirically supported by the servo model parameter tuning experiments of Schwarz and Behnke (2013), in which 40 full steps of the NimbRo-OP robot consumed 189 J with the servo model disabled, and 140 J with the model enabled.

---

2 In this case, the friction terms in Equation (3.14) had no effect as the commanded pose was stationary, i.e. $\dot{q} = 0$ for all joints.

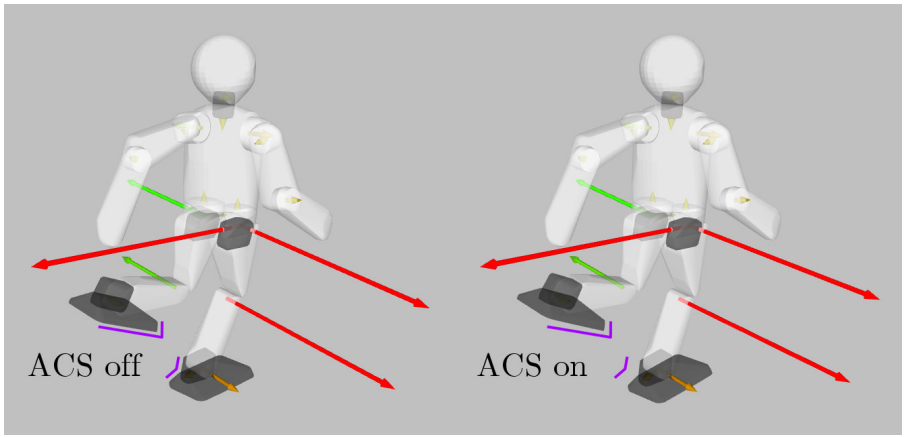

Figure 3.3: Comparison of the commanded servo target positions with and without the actuator control scheme enabled, for a robot balancing on its right leg. The left image shows the pose $\mathbf{q}$ that the robot is trying to achieve, and the right image shows the servo target positions $\mathbf{q}_d$ that are computed as a result of the calculated feed-forward torques (visualised by the 3D arrows). The target positions for the servos in the right leg are further forward than the pose $\mathbf{q}$ they are actually desired to reach (refer to the fixed purple lines), as when the weight of the robot comes into play, one can imagine that it would push the leg further backwards than commanded, towards $\mathbf{q}$.

### 3.2.4  Discussion

Together, the servo motor model (Section 3.2.1) and method for feed-forward torque estimation (Section 3.2.2) constitute the actuator control scheme. All in all, the actuator control scheme is able to compensate effectively for many factors, including battery voltage, servo P gain, gravity, joint friction, inertia, and the relative loadings of the legs. All of these factors influence how well the servos track their commanded trajectories, and by specifically considering and modelling them, the performance of the robot becomes significantly more consistent across the wide range of possible robot states and conditions.

One possible downside of the actuator control scheme is that it requires a relatively accurate and highly-tuned physical model of the robot. In the case of the robots used in this thesis, this is in general not a problem due to the use of Computer-Aided Design (CAD) and 3D printing, but it is potentially a problem for other more custom robots that do not have an accurate URDF model (with masses and inertias), and do not need one anyway for other higher level motion planning algorithms. If no masses and inertias are available for a robot, then the actuator control scheme can still be used, just with the feed-forward torques set to zero, or replaced with simplified calculated values based at each time step on the support coefficients and weight of the robot. Doing this would sacrifice the inertia compensation and the accuracy of the gravity compensation, but would still leave the friction, battery voltage and servo P gain compensation intact.

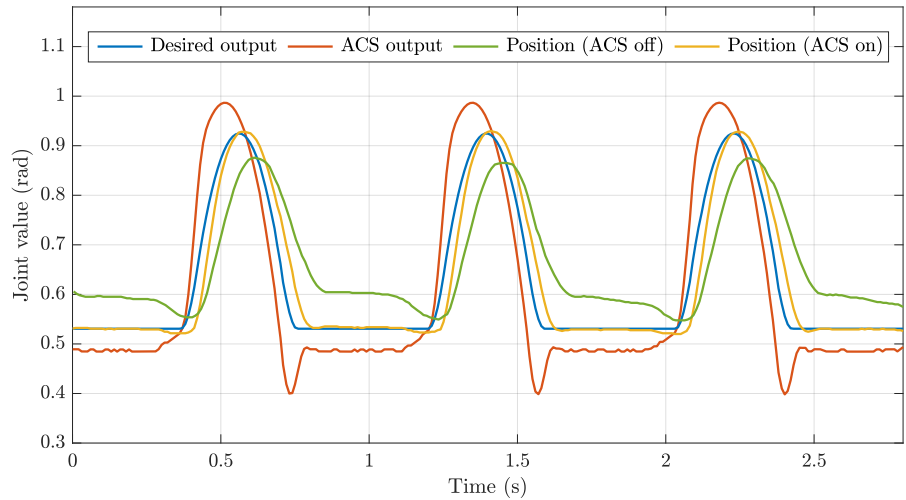

Figure 3.4: Plot of the left knee pitch tracking performance during a walking experiment on the igus Humanoid Open Platform. The desired output trajectory of the knee joint is given in blue. Passing this directly as the commanded servo targets results in the green output position waveform, which differs significantly from the desired waveform. With the actuator control scheme enabled however, the red waveform is computed in a feed-forward manner and used as the commanded servo targets instead. The resulting yellow position waveform corresponds much more closely to the intended waveform, albeit still with some signs of delay.

Positive effects of the actuator control scheme include reduced servo overheating and wear, increased battery life due to increased energy efficiency, reduced issues with impacts and self-disturbances, and the ability to increase joint compliance while maintaining good levels of trajectory tracking. One particular benefit of the control scheme is that the motions performed by the robot become more consistent across the full spectrum of possible battery voltage values. This is particularly useful when designing, for example, keyframe motions, or for that matter essentially any other motions that have a significant open-loop component to them as well.

# SENSOR CALIBRATION

As discussed in Section 3.1, all sensor data available to the robot is read in via the CM740 by the main robot control node at a rate of 100 Hz. This corresponds to Step 8 of the main robot control loop, as described on page 43. In order to properly use the available sensor data—for instance for the estimation of the orientation of the robot (see Section 10.1)—it needs to be calibrated first. In addition to the data from the joint encoders, gyroscope, accelerometer and magnetometer sensors, it is also useful to perform a kinematic calibration to ascertain the dimensions and offsets of each of the links of the robot, in addition to their masses and inertias. This helps with the dynamics calculations required for the feed-forward joint torque estimation method (see Section 3.2.2.3), and also allows general kinematic conversions to be made, such as for example the conversion from the joint space to the task space, referred to as the *inverse space* in this thesis.

Some calibration methods rely on others already having been successfully performed. For example, the gyroscope bias can only be correctly estimated if the Inertial Measurement Unit (IMU) orientation and gyroscope scale calibrations have already been completed. As a result, the full calibration procedure of a robot is somewhat sensitive to order. A list of all the sensor calibration procedures used in this thesis is given as follows, in the order that the calibrations are generally performed:

- Robot calibration
    - Joint position calibration
    - Kinematic calibration

- IMU calibration
    - IMU orientation calibration
    - Gyroscope scale calibration
    - Gyroscope bias calibration
    - Online gyroscope bias autocalibration

- Magnetometer calibration
    - Hard iron compensation
    - Soft iron compensation
    - Angle warping
    - Online hard iron autocalibration
    - Reference field vector

Note that not all calibrations are strictly necessary and/or compatible, as for example the magnetometer sensor has multiple competing calibration methods to choose from, depending on the nature and variability of the raw data, and the required use case.

## 4.1  ROBOT CALIBRATION

The first type of calibration that is performed, robot calibration, is concerned with modelling the physical and kinematic parameters of the robot, as well as the angular offsets from the servo motor positions to the joint positions.

### 4.1.1  Joint Position Calibration

The joint encoder sensors that are mounted into each Dynamixel servo motor measure the position of the joint axle in so-called *ticks* (an integral value) relative to a fixed mechanical zero position. These ticks can easily be converted to angular values in radians by multiplying by an appropriate scale factor $k_T$, but it must be considered that it is not always possible to mount the servos in such a way that the mechanical zero tick positions coincide exactly with the desired zero joint positions. As such, a joint position calibration (otherwise known as a tick offset calibration) is required to determine the tick values $\mathbf{T}_0$ at which the joint positions $\mathbf{q}$ are all zero. Then, given a measured set $\mathbf{T}$ of joint encoder tick positions, the corresponding joint angles are given by

$$\mathbf{q} = \delta_q k_T (\mathbf{T} - \mathbf{T}_0), \tag{4.1}$$

where for each joint independently $\delta_q = \pm 1$ depending on whether the corresponding servo is mounted forwards or backwards relative to the nominal direction of rotation of the joint.

The joint position calibration is nominally carried out in two different poses, the zero pose $\mathbf{q} = 0$, shown on the left in Figure 4.1, and a second custom pose that was specifically designed to allow easy and accurate identification of tick offset deviations, shown on the right in Figure 4.1. The zero pose is defined geometrically, using conditions such as the collinearity and perpendicularity of the joint axes, and/or the virtual lines between them, to define the required zero positions. For example, the zero position of the knee pitch joint is defined as the position where the hip pitch, knee pitch and ankle pitch joint axes are collinear (as viewed from the side of the leg). The second custom pose is defined by a fixed set of joint angles, and can be used in particular to more easily compare the tick offset calibrations between two equivalent robots.

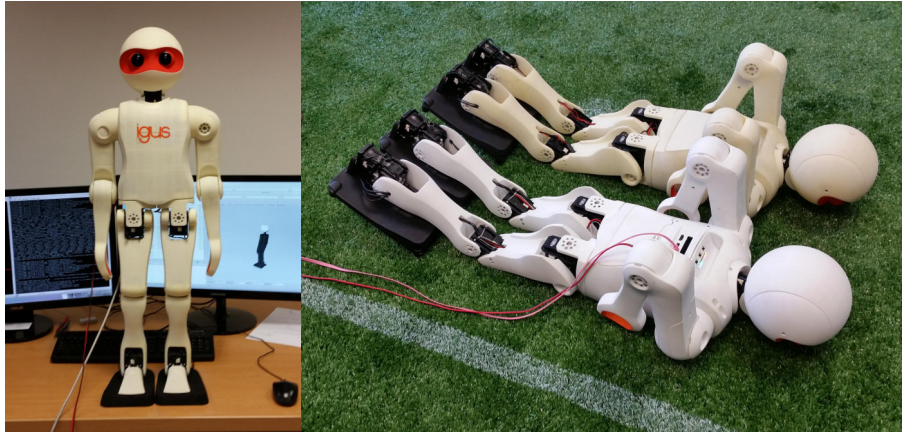

Figure 4.1: Poses used for joint position calibration. Left: The zero pose of the robot ($\mathbf{q} = \mathbf{0}$), Right: A custom pose for which deviations in joint positions can easily and accurately be identified.

### 4.1.2 Kinematic Calibration

Given a calibrated mapping from the measured joint encoder ticks to the corresponding joint angles $\mathbf{q}$ in radians (see Section 4.1.1), the software framework still needs to know the dimensions and properties of all of the robot's links in order to gain a wholesome proprioceptive awareness of the robot. Without this information, calculations like the forward and inverse kinematics of the limbs, and/or inverse dynamics of the whole robot, are not possible. In order to model the complete physical properties of the robot, a Unified Robot Description Format (URDF) model (see Figure 4.2) is created in a process referred to as kinematic calibration, or more generally, parametric robot calibration.

For robots like the igus Humanoid Open Platform, NimbRo-OP2 and NimbRo-OP2X that are 3D printed with a very high level of accuracy from a Computer-Aided Design (CAD) model, the process of kinematic calibration for the most part simply involves exporting the CAD model in the required format. Note that the material properties need to be specified in the CAD software prior to exporting, so that the link masses and inertias can be computed automatically along the way. For robots that are not 3D printed (or closely constructed from a 3D CAD model)—in this thesis only Dynaped and Copedo—the process of kinematic calibration is more complicated and involves precise manual measurement of all of the link offsets and dimensions, and best possible measurement and/or estimation of their masses and inertias. Internal joint frictions are not explicitly incorporated as part of the kinematic calibration as they are already implicitly calibrated as part of the actuator control scheme (see Section 3.2).

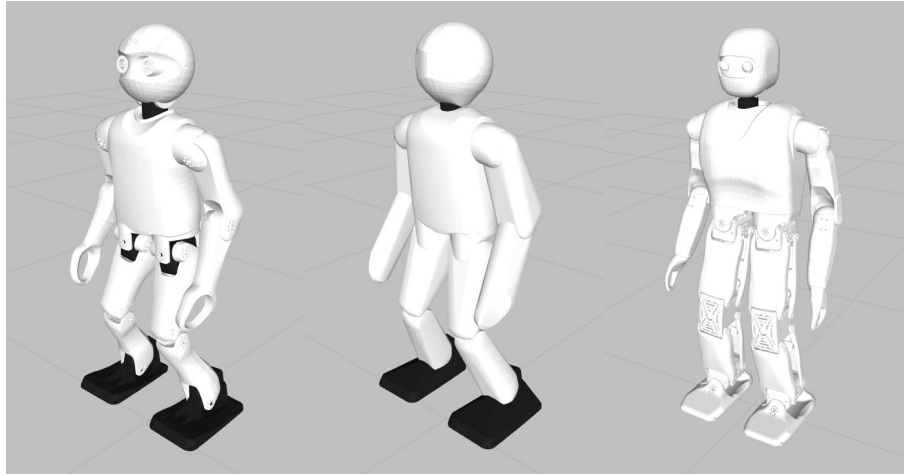

Figure 4.2: URDF models of the igus Humanoid Open Platform (left) and NimbRo-OP2X (right) robots. The centre image shows a simplified visual model of the igus Humanoid Open Platform that is used to speed up collision checking and RViz visualisations. The simplified model is completely identical to the full model in terms of its link lengths, offsets, masses and inertias, but significantly reduces the complexity of the used meshes. The NimbRo-OP2X also has a simplified URDF model, as can be seen for example towards the end of Video 1.4 (at 2:17).

## 4.2    INERTIAL MEASUREMENT UNIT CALIBRATION

One very important class of robot calibrations is given by the IMU calibration procedures. In this section, we refer to the IMU as consisting of only the 3-axis gyroscope and accelerometer sensors, as these are the two sensors that are natively integrated into the CM740, and thus rigidly connected to each other via the CM740 Printed Circuit Board (PCB). The optional 3-axis magnetometer is mounted externally to the CM740, and is calibrated separately, as described in Section 4.3.

### 4.2.1  IMU Orientation Calibration

The first aspect of the IMU that needs to be calibrated is the orientation offset of the IMU sensors relative to the main trunk link of the robot.[1] As the gyroscope and accelerometer chips are mounted in a fixed and aligned arrangement relative to one another, tuning only the offset of the *accelerometer* chip relative to the trunk suffices. This is done by sampling low-pass filtered (see Appendix A.2.2.1) values of the accelerometer at various well-defined orientations of the trunk, and computing the best-fitting estimate of $^{B}_{I}R$, the orientation of the accelerometer (i.e. IMU) frame {I} relative to the body-fixed (i.e. trunk-fixed) frame {B}.

---

1  It is assumed that the CM740 is rigidly attached to the trunk of the robot, so the required offset is constant, and thus can be tuned.

The tilt rotation component (see Section 5.4.3) of $^B_IR$ is computed from measurements of the 3-axis acceleration $^I\mathbf{a}$ when the robot trunk is manually positioned to be perfectly upright relative to gravity, i.e. so that the body-fixed and global z-axes align and point upwards. That is, if $^I\bar{\mathbf{a}}_u$ is the mean accelerometer value measured in this position, then we can naturally define

$$^I\mathbf{z}_B = \frac{^I\bar{\mathbf{a}}_u}{\|^I\bar{\mathbf{a}}_u\|},$$ (4.2)

and from this derive the required tilt rotation component. The remaining yaw rotation component of $^B_IR$ is resolved by averaging out the fused yaws calculated from the measurements of $^I\mathbf{a}$ that are obtained when the robot trunk is lying perfectly on its front, back and/or side. Joined together, the yaw and tilt rotation components together specify the complete IMU orientation offset $^B_IR$, as required.

### 4.2.2 Gyroscope Scale Calibration

Once the orientation of the IMU in the trunk is known, the magnitude of the gyroscope sensor values needs to be compensated for scale and temperature. The trunk orientation computed by the attitude estimator presented in Section 10.1 is used for this purpose. Only the gyroscope and accelerometer values are applied as inputs to the estimator, and the value of the openly integrated fused yaw component of the output (see Section 5.4.1) is monitored while the robot trunk is manually rotated about its upright z-axis (relative to gravity). The exact calibration procedure is given as follows:

1. Position the robot so that the trunk is perfectly upright relative to gravity.

2. Indicate to the software that the gyroscope scale calibration is starting.

3. **Rotation 1:** While keeping the trunk as upright as possible, yaw the robot about its (upright) z-axis by exactly $K \geq 1$ revolutions in a counterclockwise (CCW) direction. This corresponds to an applied yaw rotation of $\Psi_1 = 2\pi K$ radians.

4. Indicate to the software that the calibration midpoint has been reached.

5. **Rotation 2:** While still keeping the trunk as upright as possible, yaw the robot back to its original orientation by applying exactly $K$ revolutions in the opposite direction. This corresponds to an applied yaw rotation of $\Psi_2 = -2\pi K$ radians.

6. Indicate to the software that the original orientation has been reached again.

Based on the measured 3-axis gyroscope values and changes in fused yaw throughout both rotations, an *unbiased* estimate of a gyroscope scale factor $k_g$ can be obtained for the current measured CM740 temperature. Doing this once for a high CM740 temperature and once for a low CM740 temperature allows coerced linear interpolation (see Appendix A.1.4.1) to be used to select an appropriate scale factor $k_g(T)$ for any future measured temperature $T$. If ${}^I\hat{\boldsymbol{\Omega}}$ is a raw measured gyroscope value, the corresponding scale-corrected gyroscope value ${}^I\hat{\boldsymbol{\Omega}}_s$ is given by

$$ {}^I\hat{\boldsymbol{\Omega}}_s = k_g(T){}^I\hat{\boldsymbol{\Omega}}. \tag{4.3} $$

Note that {I} is the coordinate frame defining the x, y and z-axes of the IMU sensors, and is therefore also the frame relative to which the gyroscope measurements are expressed.

#### 4.2.2.1  *Calculation of the Gyroscope Scale Factor from Data*

We now present the method by which the gyroscope scale factor $k_g$ can be calculated from the data acquired through the gyroscope scale calibration procedure described above.

For each of the two calibration rotations, we obtain some number $n$ of measured gyroscope values ${}^I\hat{\boldsymbol{\Omega}}_i$ (for $i = 1, \ldots, n$), obtained at regular time intervals of $\Delta t$ seconds. Additionally, for each rotation we also obtain the total change in fused yaw $\Delta\psi$ that was observed by the attitude estimator during the rotation, calculated internally by 3D integration of the gyroscope values (see Section 10.1.4). Given that the trunk of the robot remains upright throughout the entire applied yaw rotations, we can observe that the component of the gyroscope measurement contributing to the estimated fused yaw at each measurement instant $i = 1, \ldots, n$ is given by

$$ \omega_{zi} = {}^I\mathbf{z}_G \bullet {}^I\hat{\boldsymbol{\Omega}}_i \tag{4.4} $$

about the (constant) global z-axis ${}^I\mathbf{z}_G$. Numerically integrating Equation (4.4) must give $\Delta\psi$, so mathematically we have that

$$ \Delta\psi = \sum_{i=1}^{n} \omega_{zi}\Delta t = \Delta t \sum_{i=1}^{n}({}^I\mathbf{z}_G \bullet {}^I\hat{\boldsymbol{\Omega}}_i), \tag{4.5} $$

where we recall that $\Delta\psi$ is the change in yaw that is *measured* by the attitude estimator. A similar formula can be derived for the change in yaw $\Psi$ that was actually *applied* to the robot (see Steps 3 and 5 of the calibration procedure), namely

$$ \Psi = \Delta t \sum_{i=1}^{n}({}^I\mathbf{z}_G \bullet {}^I\boldsymbol{\Omega}_i), \tag{4.6} $$

where ${}^I\boldsymbol{\Omega}_i$ is the true applied angular velocity at every measurement instant $i$.

The relationship between the measured and truly applied angular velocities ${}^I\hat{\boldsymbol{\Omega}}_i$ and ${}^I\boldsymbol{\Omega}_i$ (respectively) is modelled as

$$
{}^I\hat{\boldsymbol{\Omega}}_{si} \equiv k_g\,{}^I\hat{\boldsymbol{\Omega}}_i = {}^I\boldsymbol{\Omega}_i + \mathbf{b}_\Omega + \mathbf{w}_{\Omega i}, \tag{4.7}
$$

where $k_g$ is the temperature-dependent scale factor required for the gyroscope measurements, $\mathbf{b}_\Omega$ is the gyroscope bias vector (assumed to be constant during any one calibration), and $\mathbf{w}_{\Omega i}$ is zero mean sensor noise. Thus, it can be concluded from Equation (4.6) that

$$
\begin{aligned}
\Psi &= \Delta t \sum_{i=1}^{n} \big({}^I\mathbf{z}_G \boldsymbol{\cdot} {}^I\boldsymbol{\Omega}_i\big) \\
&= \Delta t \sum_{i=1}^{n} \Big({}^I\mathbf{z}_G \boldsymbol{\cdot} (k_g\,{}^I\hat{\boldsymbol{\Omega}}_i - \mathbf{b}_\Omega - \mathbf{w}_{\Omega i})\Big) \\
&= k_g \Delta t \sum_{i=1}^{n} \big({}^I\mathbf{z}_G \boldsymbol{\cdot} {}^I\hat{\boldsymbol{\Omega}}_i\big) - \Delta t \sum_{i=1}^{n} \big({}^I\mathbf{z}_G \boldsymbol{\cdot} \mathbf{b}_\Omega\big) - \Delta t \Big({}^I\mathbf{z}_G \boldsymbol{\cdot} \overbrace{\sum_{j=1}^{n} \mathbf{w}_{\Omega i}}^{\approx 0}\Big) \\
&= k_g \Delta\psi - n\Delta t \big({}^I\mathbf{z}_G \boldsymbol{\cdot} \mathbf{b}_\Omega\big),
\end{aligned} \tag{4.8}
$$

where the last step makes use of Equation (4.5), and the fact that $\mathbf{w}_{\Omega i}$ is zero mean noise. Writing out Equation (4.8) independently for each of the rotations 1 and 2 yields

$$
\Psi_1 = k_g \Delta\psi_1 - n_1 \Delta t \big({}^I\mathbf{z}_G \boldsymbol{\cdot} \mathbf{b}_\Omega\big), \tag{4.9a}
$$

$$
\Psi_2 = k_g \Delta\psi_2 - n_2 \Delta t \big({}^I\mathbf{z}_G \boldsymbol{\cdot} \mathbf{b}_\Omega\big). \tag{4.9b}
$$

Solving these linear equations simultaneously for $k_g$ gives

$$
k_g = \frac{n_1 \Psi_2 - n_2 \Psi_1}{n_1 \Delta\psi_2 - n_2 \Delta\psi_1}. \tag{4.10}
$$

This is the final required expression for the calibrated gyroscope scale factor $k_g$, which can then be used with Equation (4.3) to correct future measurements of ${}^I\hat{\boldsymbol{\Omega}}$.

### 4.2.3 Gyroscope Bias Calibration

As discussed in the previous section and shown in Equation (4.7), the gyroscope sensor is modelled as having a non-zero bias $\mathbf{b}_\Omega$ that offsets all of the scale-corrected measured angular velocities

$$
{}^I\hat{\boldsymbol{\Omega}}_s \equiv k_g(T)\,{}^I\hat{\boldsymbol{\Omega}}. \tag{4.11}
$$

That is, if ${}^I\boldsymbol{\Omega}$ is the true angular velocity of the IMU at some instant in time, it is modelled that

$$
{}^I\hat{\boldsymbol{\Omega}}_s = {}^I\boldsymbol{\Omega} + \mathbf{b}_\Omega + \mathbf{w}_\Omega, \tag{4.12}
$$

where $\mathbf{w}_\Omega$ is zero mean sensor noise. If the assumption is made that $\mathbf{b}_\Omega$ is a fixed constant over all time, then it is possible to calibrate the gyroscope bias only once, and simply subtract it from every future scale-corrected measurement ${}^I\hat{\mathbf{\Omega}}_s$, as

$$ {}^I\hat{\mathbf{\Omega}}_s - \mathbf{b}_\Omega = {}^I\mathbf{\Omega} + \mathbf{w}_\Omega. \tag{4.13} $$

If the robot (and therefore IMU) is completely stationary for some period of time, we know that ${}^I\mathbf{\Omega} = 0$, and therefore that

$$ {}^I\hat{\mathbf{\Omega}}_s = \mathbf{b}_\Omega + \mathbf{w}_\Omega. \tag{4.14} $$

As such, if we low-pass filter the obtained values of ${}^I\hat{\mathbf{\Omega}}_s$ during this period of time with a moderate 90% settling time $T_{s,M}$ (on the order of seconds), the effect of $\mathbf{w}_\Omega$ diminishes (as it is zero mean) and a highly accurate estimate of $\mathbf{b}_\Omega$ results. This process has been implemented as a gyroscope bias calibration procedure that can manually be triggered in the software. After waiting a number of seconds for the bias estimate to converge, the low-pass filtered value of ${}^I\hat{\mathbf{\Omega}}_s$ is captured and saved as the bias estimate $\mathbf{b}_\Omega$.

### 4.2.4    Online Gyroscope Bias Autocalibration

The assumption that the gyroscope bias is constant over all time is not always accurate. To overcome this, an online method of gyroscope bias autocalibration has been developed that allows even brief periods of time that the robot is stationary during operation to be automatically detected, and leveraged to improve the gyroscope bias estimate. The robot is considered to be stationary if the difference between ${}^I\hat{\mathbf{\Omega}}_s$—see Equation (4.11)—and a moderately low-pass filtered version of ${}^I\hat{\mathbf{\Omega}}_s$ (90% settling time $T_{s,M} \approx 2\,\mathrm{s}$) is less than a given threshold in norm for a configured duration of time ($\approx 1.5\,\mathrm{s}$). The norm threshold is chosen so that it is not much larger than the general maximum magnitude of sensor noise $\mathbf{w}_\Omega$, so as to ensure that the robot really is stationary when the condition is satisfied. The gyroscope bias adjustment process starts (time $t_a = 0$) every time the robot is detected to become stationary, and stops whenever the threshold is even just temporarily exceeded.

In addition to the moderate low-pass filter running on ${}^I\hat{\mathbf{\Omega}}_s$, a low-pass filter with an even slower 90% settling time ($T_{s,S} \approx 8\,\mathrm{s}$) is used in parallel, in order to improve the noise rejection ability of the estimated gyroscope bias $\mathbf{b}_\Omega$ when the robot is stationary for longer, i.e. when a larger $t_a$ has been reached. Let $\mathbf{\Omega}_M$ be the filtered angular velocity from the moderate low-pass filter, and $\mathbf{\Omega}_S$ be the equivalent output angular velocity of the slower low-pass filter, which is automatically reset to the value of $\mathbf{\Omega}_M$ in the instant when $t_a = 0$, i.e. when the adjustment process starts. For as long as the robot is stationary, in each measurement cycle the estimated gyroscope bias $\mathbf{b}_\Omega$ is interpolated linearly towards the so-called target bias $\mathbf{b}_T$, given by

$$\mathbf{b}_T = u\mathbf{\Omega}_S + (1-u)\mathbf{\Omega}_M, \tag{4.15}$$

where $u$ is 0 when the bias adjustment process starts, and interpolates up to 1 over the next $T_{s,S}$ seconds, i.e. (see Appendix A.1.4.1 for the notation)

$$u = \text{interpolateCoerced}\big([0, T_{s,S}] \to [0,1],\, t_a\big). \tag{4.16}$$

In each cycle, the bias $\mathbf{b}_\Omega$ is updated using the formula

$$\mathbf{b}_\Omega \leftarrow \mathbf{b}_\Omega + \alpha_B(\mathbf{b}_T - \mathbf{b}_\Omega), \tag{4.17}$$

where $\alpha_B \in [0,1]$ is the smoothing factor of $\mathbf{b}_\Omega$, given by

$$T_{s,B} = \text{interpolateCoerced}\big([0, t_d] \to [T_{s,slow}, T_{s,fast}],\, t_a\big), \tag{4.18a}$$

$$\alpha_B = 1 - 0.10^{\frac{\Delta t}{T_{s,B}}}, \tag{4.18b}$$

where $\Delta t$ is the measurement cycle time, $t_d$ is a configured fade duration ($\approx 1.2\,\text{s}$), and $T_{s,slow}$ and $T_{s,fast}$ are configured slow and fast gyroscope bias 90% settling times, respectively. Note that Equation (4.17) is the standard update equation for a first-order low-pass filter, but one where the smoothing factor $\alpha_B$ is continuously variable with time.

Effectively, the described autocalibration scheme attempts to extract from the measured gyroscope values as much useful information as possible, in as little time as possible, but without risking using the data too early and temporarily degrading the bias estimation accuracy. This is why the scheme initially waits a small amount of time for the robot to be stable, and then only weakly uses the data at first to improve the bias estimate, before becoming increasingly confident as further time elapses and the robot is still perceived to be stable. The use of low-pass filtering ensures that no sudden jumps in the bias estimate occur, and that the noise in the sensor can be robustly and reliably rejected. In addition to being automated and having a quicker responsiveness, these are both properties that do not apply to the manual calibration scheme presented in Section 4.2.3.

A plot of the gyroscope bias autocalibration method in action is shown in Figure 4.3. Initially, the robot is in motion, as can be seen from the gyroscope values, but after stopping and waiting a small amount of time for the motion of the torso to settle, the bias adjustment process automatically triggers at $t_a = 0\,\text{s}$, and fades the initially zero bias estimate $\mathbf{b}_\Omega = (b_{\Omega x},\, b_{\Omega y},\, b_{\Omega z})$ to its appropriate filtered measured value. After only 1.25 s, the estimate reaches its final value, and only varies very slightly thereafter due to the residual effect of sensor noise.

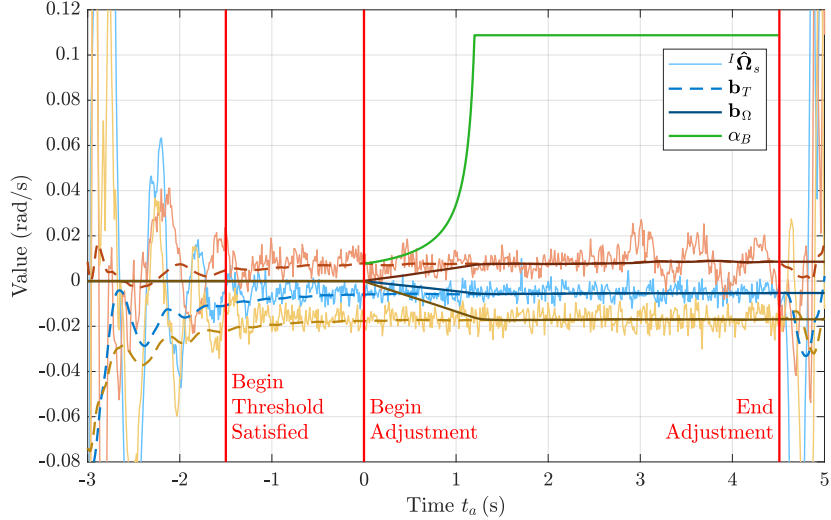

Figure 4.3: Time plots of the gyroscope autocalibration scheme in action. As indicated by the legend, the thin light lines correspond to the components of the scale-corrected gyroscope value $^I\hat{\mathbf{\Omega}}_s$, the thick dark lines correspond to the components of the estimated gyroscope bias $\mathbf{b}_\Omega$, and the dashed lines correspond to the components of the target bias $\mathbf{b}_T$, towards which $\mathbf{b}_\Omega$ is faded during the adjustment process. Note that before and after the beginning of adjustment, the value of $\mathbf{b}_T$ is just $\mathbf{\Omega}_M$, the filtered angular velocity from the moderate low-pass filter. The gyroscope bias smoothing factor $\alpha_B$ is shown for the interval where adjustment is active. After the norm threshold is violated at $t_a \approx 4.5\,\mathrm{s}$, the estimated bias $\mathbf{b}_\Omega$ is no longer adjusted, and simply retains its value.

## 4.3   MAGNETOMETER CALIBRATION

For all of the robots used in this thesis, a magnetometer sensor can optionally be mounted to the CM740 and connected to it electronically. If care is taken to mount it in a consistent orientation relative to the two remaining inertial sensors, the same IMU offset $^B_I R$ as tuned in Section 4.2.1 can be used to correct the data. If not, a further orientation offset needs to be measured and applied to the magnetometer data—to bring it to the IMU frame {I}—before transforming it to the body-fixed trunk link frame {B} using the IMU offset $^B_I R$. Either way, it is assumed for the remainder of this section that the raw magnetometer data $^B\hat{\mathbf{m}}$ is available in the trunk link frame {B}, as required.

### 4.3.1   Magnetometer Correction Pipeline

Given the raw measured magnetometer vector $^B\hat{\mathbf{m}}$ relative to the frame {B}, the data correction pipeline runs as follows, where not all steps are usually enabled at the same time:

1. A **spike filter** is applied (see Appendix A.2.4.4) to filter out any temporary large jumps in the data. Such jumps were observed to occasionally occur with the HMC5883L magnetometer chip.

2. A **hard iron offset** is applied to the data. This offset may come from a manual calibration (see Section 4.3.2), or the online hard iron autocalibration scheme (see Section 4.3.5).

3. A **soft iron correction** that also has a hard iron component in it is applied to the data.

4. A **cyclic angle warp** (see Appendix A.1.4.3) is applied to the x and y-components of the data.

5. A small **mean filter** is applied (see Appendix A.2.1.1) to smooth out part of the noise in the final data.

The calibration of the spike filter in Step 1 is relatively simple, as it only requires a slope threshold to be set, which should be exceeded if and only if an unrealistic data spike is encountered. Similarly, the calibration of the mean filter in Step 5 is also simple, as only a filter order needs to be decided upon, as a direct trade-off between the amount of noise that needs to be removed and the delay time that is induced. The calibration of the remaining filters and steps is detailed in the following sections.

### 4.3.2 Hard Iron Compensation

Ultimately, what a magnetometer sensor aims to do is measure the strength and direction of the Earth's magnetic field, and assuming this is approximately constant throughout the local operating space of the robot (e.g. throughout the room that the robot is operating in), use this information to make deductions about the orientation of the robot. Robots have many sources of their own magnetic fields however, including servos, encoders (if using Hall effect sensing), speakers and similar. These sources of local magnetic fields disturb the measurements of the magnetometer, but if the sources that have non-negligible effects on the measurements remain more or less constant in magnitude, and remain in a fixed position and orientation relative to the body-fixed frame {B}, the effect on the values ${}^{B}\hat{\mathbf{m}} \in \mathbb{R}^3$ measured by the magnetometer is to produce a constant offset. This offset is referred to as the hard iron offset, and can be identified as $\mathbf{b}_m$ in the magnetometer measurement model

$$ {}^{B}\hat{\mathbf{m}} = {}^{B}\mathbf{m}_e + \mathbf{b}_m + \mathbf{w}_m, \tag{4.19} $$

where ${}^{B}\mathbf{m}_e$ is the vector corresponding to Earth's magnetic field for the current body-fixed frame {B}, and $\mathbf{w}_m$ is zero mean sensor noise.

Although the direction of ${}^{B}\mathbf{m}_e$ changes as the robot, i.e. frame {B}, rotates, its magnitude $M_e$ (the local magnitude of the Earth's magnetic field) clearly stays constant. We conclude from Equation (4.19) that the observed magnetometer measurements ${}^{B}\hat{\mathbf{m}}$ must lie on a sphere of radius $M_e$ centred at the hard iron offset $\mathbf{b}_m$, with some random perturbations as a result of the sensor noise $\mathbf{w}_m$. Consequently, if 3-axis magnetometer data ${}^{B}\hat{\mathbf{m}}$ is recorded while the robot is manually rotated in all different directions, including for example completely upside down, a sphere can be fitted (see Appendix A.1.6.6) to estimate the hard iron offset $\mathbf{b}_m$ and the magnetic field strength $M_e$. Then, given any future measured value ${}^{B}\hat{\mathbf{m}}$, the hard iron effects can be corrected by computing

$$ {}^{B}\hat{\mathbf{m}}_c = {}^{B}\hat{\mathbf{m}} - \mathbf{b}_m. \tag{4.20} $$

Note that the magnitude $\|{}^{B}\hat{\mathbf{m}}_c\|$ of the measured magnetometer values is rarely useful, so normally the vector ${}^{B}\hat{\mathbf{m}}_c$ is normalised to give a unit vector before being further used.

Performing a full 3D calibration of the hard iron offset can be quite cumbersome and involved for larger robots that cannot be picked up and handled so easily. As such, if it is known that a full 3D hard iron offset has been calibrated at least once in the past and is still approximately correct, a simpler 2D recalibration of the hard iron offset can be performed instead of redoing an entire 3D one. The magnetometer sensor is mainly used to infer which direction the robot is facing when it is upright, so it is reasonable to refine the calibration based only on these situations. If 3-axis magnetometer data ${}^{B}\hat{\mathbf{m}}$ is recorded while the robot is upright and manually being yawed about its vertical z-axis $\mathbf{z}_B$, it can be inferred from the measurement model in Equation (4.19) that the expected locus of ${}^{B}\hat{\mathbf{m}} = \left({}^{B}\hat{m}_x, {}^{B}\hat{m}_y, {}^{B}\hat{m}_z\right)$ is a circle in the **xy** plane at some constant z-value ${}^{B}\hat{m}_z$. Thus, if a circle is fitted to the recorded data using the method described in Appendix A.1.6.6, i.e. to the recorded $\left({}^{B}\hat{m}_x, {}^{B}\hat{m}_y\right)$ data, a refined estimate of the x and y-components of $\mathbf{b}_m$ can be made without requiring a full 3D calibration. This is referred to as a 2D magnetometer hard iron offset calibration.

### 4.3.3 Soft Iron Compensation

While local magnetic fields superimposing on top of the Earth's magnetic field is handled by hard iron compensation, the local bending and distortion of existing magnetic fields due to the presence of materials with a high magnetic permeability (like iron) is handled by soft iron compensation. While the normal undistorted locus of the measured magnetometer values ${}^{B}\hat{\mathbf{m}}$ is expected to be a sphere (as discussed in the previous section), with the effect of distortions this becomes a three-dimensional ellipsoid. The process of soft iron compensation uses a symmetric $3 \times 3$ matrix $W$ to model the local magnetic distortions in the robot, leading to the measurement model

$$^{B}\hat{\mathbf{m}} = W\,{}^{B}\mathbf{m}_e + \mathbf{b}_m + \mathbf{w}_m, \tag{4.21}$$

where $\mathbf{b}_m$ is the hard iron offset as before, and $\mathbf{w}_m$ is the zero mean sensor noise. The matrix $W$ needs to be symmetric (i.e. $W^T = W$) as it should not contain any rotation component that would rotate the data away from being relative to frame {B}.[2] Given Equation (4.21) and appropriately calibrated $W^{-1}$ and $\mathbf{b}_m$ parameters, any measured magnetometer values $^{B}\hat{\mathbf{m}}$ can simultaneously be soft and hard iron corrected using the formula

$$^{B}\hat{\mathbf{m}}_c = W^{-1}\big({}^{B}\hat{\mathbf{m}} - \mathbf{b}_m\big). \tag{4.22}$$

The values of the soft iron undistortion matrix $W^{-1}$ and hard iron offset $\mathbf{b}_m$ can be jointly calibrated by fitting a generic 3D ellipsoid to the magnetometer data $^{B}\hat{\mathbf{m}}$ obtained while rotating the robot through all possible orientations. Following the method in Appendix A.1.6.7, this yields the symmetric matrix $A \in \mathbb{R}^{3\times3}$ and vector $\mathbf{c} \in \mathbb{R}^3$ such that the best-fitting ellipsoid is given by

$$\big({}^{B}\hat{\mathbf{m}} - \mathbf{c}\big)^T A \big({}^{B}\hat{\mathbf{m}} - \mathbf{c}\big) = 1. \tag{4.23}$$

As both $\mathbf{b}_m$ and $\mathbf{c}$ correspond to the centre of the fitted ellipsoid, we can immediately conclude that the required calibrated hard iron offset $\mathbf{b}_m$ is given by

$$\mathbf{b}_m = \mathbf{c}. \tag{4.24}$$

Furthermore, the three principal radii $r_1$, $r_2$, $r_3$ of the fitted ellipsoid can be determined from the eigenvalues of $A$, and used to calculate the estimated local magnetic field strength of the Earth $M_e$, using the formula

$$M_e = \sqrt[3]{r_1 r_2 r_3}. \tag{4.25}$$

From Equation (4.22) and the definition of $W$, we expect that $W^{-1}$ transforms the ellipsoidal locus of $^{B}\hat{\mathbf{m}} - \mathbf{b}_m$ to a spherical locus of $^{B}\hat{\mathbf{m}}_c$ (both loci centred at the origin), with zero net rotation. The required spherical locus of $^{B}\hat{\mathbf{m}}_c$ is defined by the equation

$$\|{}^{B}\hat{\mathbf{m}}_c\| = M_e, \tag{4.26}$$

---

2 As indicated in Section 7.1.2.2, the rotation component of a linear transformation expressed as a $3 \times 3$ matrix $A$ is $A(A^T A)^{-\frac{1}{2}}$. For the matrix $W$, we require this component of rotation to be the identity matrix $\mathbb{I}$, i.e. no rotation, so $W(W^T W)^{-\frac{1}{2}} = \mathbb{I}$ $\implies W = (W^T W)^{\frac{1}{2}} \implies W^2 = W^T W \implies W = W^T$, and so $W$ must be symmetric.

so we can derive that

$$
\begin{aligned}
1 &= \tfrac{1}{M_c^2} \left\| {}^B\hat{\mathbf{m}}_c \right\|^2 \\
&= \tfrac{1}{M_c^2} {}^B\hat{\mathbf{m}}_c^T\, {}^B\hat{\mathbf{m}}_c \\
&= \tfrac{1}{M_c^2} \left( W^{-1} \big( {}^B\hat{\mathbf{m}} - \mathbf{b}_m \big) \right)^T \left( W^{-1} \big( {}^B\hat{\mathbf{m}} - \mathbf{b}_m \big) \right) \\
&= \big( {}^B\hat{\mathbf{m}} - \mathbf{b}_m \big)^T \frac{(W^{-1})^2}{M_c^2} \big( {}^B\hat{\mathbf{m}} - \mathbf{b}_m \big),
\end{aligned}
\tag{4.27}
$$

as $W^{-1}$ is symmetric and as a result $(W^{-1})^T = W^{-1}$. Comparing this expression to Equation (4.23) yields

$$
\frac{(W^{-1})^2}{M_c^2} = A,
\tag{4.28}
$$

which trivially reduces down to

$$
W^{-1} = M_c A^{\frac{1}{2}}.
\tag{4.29}
$$

This equation gives the final required calibrated value of the soft iron undistortion matrix $W^{-1}$, and together with Equation (4.24) allows Equation (4.22) to be used for the soft iron compensation of all future incoming magnetometer data.

### 4.3.4 Angle Warping

Together, the options of using hard and soft iron compensation cover almost all situations that occur with magnetometers. Sometimes however, for instance when local distortions of the magnetic field become too nonlinear, there are residual inaccuracies even in the corrected measurements of the magnetometer, leading to distortions of the robot heading directions that are estimated using the data. Angle warping is the last-in-line mechanism to deal with these heading distortions, and tries to ensure that the estimated heading directions are accurate at least when the robot is upright.

Suppose the robot is placed on the ground in such a way that its torso is upright and facing a particular heading direction relative to the environment. We refer to this position as the zero heading pose ($\psi_0 = 0$), and denote the corrected measured magnetometer value at that pose ${}^B\hat{\mathbf{m}}_{c0} = (m_{x0}, m_{y0}, m_{z0})$. If the robot is now physically yawed CCW about its vertical z-axis, so that it is still upright but facing exactly the heading direction given by some chosen angle $\psi_1 \in (-\pi, \pi]$ relative to the zero heading pose, we can measure the corrected magnetometer value again, and record it as ${}^B\hat{\mathbf{m}}_{c1} = (m_{x1}, m_{y1}, m_{z1})$. We repeat this process until there are $N+1$ angles $\psi_0, \psi_1, \ldots, \psi_N$ and $N+1$ corresponding data points ${}^B\hat{\mathbf{m}}_{c0}, {}^B\hat{\mathbf{m}}_{c1}, \ldots, {}^B\hat{\mathbf{m}}_{cN}$, ideally spread more or less evenly across all possible heading directions of the robot.

For $i = 0, \ldots, N$, let us define

$$\tilde{\psi}_i = -\operatorname{atan2}(m_{yi}, m_{xi}). \tag{4.30}$$

These are the raw heading angles calculated from the corrected magnetometer data points $^B\hat{\mathbf{m}}_{ci}$, where importantly we recall that relative to the torso the x-axis points forwards and the y-axis points left. Note that if the heading of the robot changes in a CCW direction, the magnetometer measurement changes in a clockwise (CW) direction relative to the robot, hence the minus sign in the equation. If the hard and/or soft iron calibrations worked perfectly in removing all distortions of the magnetometer data, we would expect that (up to angle wrapping by $2\pi$)

$$\tilde{\psi}_i = \tilde{\psi}_0 + \psi_i \tag{4.31}$$

for all $i = 0, \ldots, N$. If this is not the case however, we use the concept of angle warping, and based on the method in Appendix A.1.4.3 construct an invertible piecewise linear function $f_w(\cdot)$ over the cyclic domain $(-\pi, \pi]$ that maps all $\tilde{\psi}_i$ to their respective $\tilde{\psi}_0 + \psi_i$, i.e.

$$f_w(\tilde{\psi}_i) = \tilde{\psi}_0 + \psi_i. \tag{4.32}$$

Then, given any new corrected magnetometer value

$$^B\hat{\mathbf{m}}_{cn} = (m_{xn}, m_{yn}, m_{zn}), \tag{4.33}$$

we calculate

$$M_n = \sqrt{m_{xn}^2 + m_{yn}^2}, \tag{4.34a}$$

$$\tilde{\psi}_n = -\operatorname{atan2}(m_{yn}, m_{xn}), \tag{4.34b}$$

and construct the angle-warped magnetometer output vector $^B\hat{\mathbf{m}}_w$ using the equation

$$^B\hat{\mathbf{m}}_w = \left( M_n \cos\left(-f_w(\tilde{\psi}_n)\right),\ M_n \sin\left(-f_w(\tilde{\psi}_n)\right),\ m_{zn} \right). \tag{4.35}$$

This correction process is referred to as 'angle warping', because the raw heading angle of the magnetometer measurements is being 'warped' over the cyclic domain $(-\pi, \pi]$ to be consistent with the nominal known data $\psi_i$. If no angle warping is necessary because Equation (4.31) is already true, by comparison of this equation with Equation (4.32), we can see that the fitted $f_w(\cdot)$ would end up being the identity function, and Equation (4.35) would subsequently simplify down to

$$^B\hat{\mathbf{m}}_w = (m_{xn}, m_{yn}, m_{zn}) = {}^B\hat{\mathbf{m}}_{cn}, \tag{4.36}$$

meaning that the angle warping does nothing as expected.

### 4.3.5 Online Hard Iron Autocalibration

If only a hard iron offset is being used to correct the magnetometer measurements, it is possible to use an autocalibration scheme to calibrate this offset online while the robot is operating. The method presented in this section is intended for applications where the robot is more or less upright for the majority of the time of operation, and where discernment of the robot heading direction is the most important function of the magnetometer. The main advantage of the autocalibration scheme is that small to medium shifts in the magnetometer readings can be identified and corrected online, reducing the regularity with which it is necessary to perform magnetometer recalibrations, and reducing the sensitivity that the robot has towards the quality of each calibration.

We recall from the 2D hard iron offset calibration method described in Section 4.3.2 on page 66, that the locus of the raw magnetometer measurements $^B\hat{\mathbf{m}}$ when the robot is upright is given by a horizontal circle, i.e. a circle in a plane parallel to the **xy** plane in 3D space. Let $R$ denote the radius of this circle, and $(b_{mx}, b_{my}, z_R)$ denote its centre, where we note that $b_{mx}$ and $b_{my}$ are the x and y-components of the hard iron offset $\mathbf{b}_m$, respectively. The core idea of the magnetometer autocalibration scheme is to monitor the raw magnetometer values $^B\hat{\mathbf{m}}$ when the robot is estimated to be close to upright, and to continuously incrementally update the parameters $b_{mx}$, $b_{my}$, $z_R$, and $R$ to fit the circular data that is observed. This is done, in order, as follows:

1. The raw magnetometer data is debounced so that new data points are only accepted if they differ from the last accepted data point by some minimum normed distance.

2. A weight is computed for each new debounced data point $^B\hat{\mathbf{m}}$, based on how close the robot is to being upright at the time. This is judged by a mix of the deviation between $^B\hat{m}_z$ and $z_R$, and the result of the magnetometer-less attitude estimator in the previous cycle (see Section 10.1).

3. The debounced data points of non-zero weight are placed into a circular buffer in memory.

4. When a certain number of data points have been added or a certain time limit has elapsed since the last autocalibration, the autocalibration procedure triggers and estimates new target values $\tilde{b}_{mx}$, $\tilde{b}_{my}$, $\tilde{z}_R$, $\tilde{R}$ for the calibration parameters, along with a target confidence value $\tilde{C}$ in the range $[0, 1]$ (often alternatively expressed as a value in percent, e.g. like in Figure 4.4).

5. Slope-limited first-order low-pass filters (see Appendix A.2.2.2) are used to slowly fade the current values of the four calibration parameters to the target ones calculated in the previous step.

The configured slope limits of the filters are proportional to the target confidence, and the configured time constants are inversely proportional. The $b_{mx}$ and $b_{my}$ filters are linked so that the resulting adjustment of $(b_{mx}, b_{my})$ is always strictly towards $(\tilde{b}_{mx}, \tilde{b}_{my})$, and not potentially in any other direction due to the effect of the slope limits.

The autocalibration procedure that triggers in Step 4 is a complicated and involved procedure, for which the Humanoid Open Platform ROS Software (Team NimbRo, 2018a) is the definitive resource.[3] Nevertheless, the procedure can be summarised as follows:

1. The weighted circle of best fit of the data in the circular buffer is computed (see Appendix A.1.6.6), along with the associated normalised mean residual fitting error $\hat{e}_f$. Note that the fitted circle is constrained to be parallel to the **xy** plane, i.e. with constant z-value. Its centre is given by $(c_x, c_y, c_z)$ and its radius is given by $R_f$.

2. The *angular spread* of the data around the fitted centre of the circle is quantified. If $^B\hat{\mathbf{m}}_i = (m_{ix}, m_{iy}, m_{iz})$ represents the data points in the buffer, the angular spread is given by

$$S_a = 1 - \left\| \frac{\sum w_i \hat{\mathbf{d}}_i}{\sum w_i} \right\| \in [0, 1], \tag{4.37}$$

where

$$\hat{\mathbf{d}}_i = \frac{(m_{ix}, m_{iy}) - (c_x, c_y)}{\|(m_{ix}, m_{iy}) - (c_x, c_y)\|}. \tag{4.38}$$

If all the data points are in exactly the same direction from the fitted centre, the angular spread $S_a$ is 0. If the data points are uniformly spread out around the entire circumference of the circle of best fit, i.e. in all directions, the angular spread $S_a$ is 1. Every other case is in between.

3. The *planar spread* of the data in the **xy** plane is calculated. If $\Sigma_{\mathbf{m}} \in \mathbb{R}^{2 \times 2}$ is the covariance matrix of the weighted buffered data $(m_{ix}, m_{iy})$, and $\sigma_a^2$ and $\sigma_b^2$ are the corresponding principal covariances (not necessarily aligned with the x and y-axes), then the standard deviation of the 2D data is given by an ellipse centred at the weighted mean with principal semi-axis lengths $\sigma_a$ and $\sigma_b$. The planar spread $S_p$ of the data is defined to be the equivalent radius of this ellipse, i.e. the radius of the circle that has the same area as this ellipse. In terms of the principal

---

3 Specifically, the function `computeAutoCalib()` in:
https://github.com/AIS-Bonn/humanoid_op_ros/blob/master/src/nimbro_robotcontrol/hardware/robotcontrol/src/hw/magfilter.cpp

standard deviations $\sigma_a$, $\sigma_b$ and covariance matrix $\Sigma_\mathbf{m}$, this is given by

$$S_p = \sqrt{\sigma_a \sigma_b} = \sqrt[4]{\sigma_a^2 \sigma_b^2} = \sqrt[4]{\det(\Sigma_\mathbf{m})}. \tag{4.39}$$

Note that $S_p$ is effectively a *scalar* measure of the weighted standard deviation of the 2D data about its weighted mean.

4. **Fitted target:**   Target values for the four calibration parameters $b_{mx}$, $b_{my}$, $z_R$ and $R$ are calculated based on the idea of moving the current 2D magnetometer calibration circle towards the circle in Step 1 that was empirically fitted through the buffered data, i.e.

$$(\breve{b}_{mx}, \breve{b}_{my}, \breve{z}_R, \breve{R}) = (c_x, c_y, c_z, R_f). \tag{4.40}$$

The confidence $\breve{C} \in [0, 1]$ of the fitted target is high when there is a thin wide angular spread of the magnetometer data, with low residual fitting error and a believable fitted radius. Numerically, the confidence is computed as a combination of the relative proximity between $R_f$ and the current $R$, and the angular spread $S_a$ of the data relative to a particular threshold that is computed as a function of the residual fitting error $\hat{e}_f$.

5. **Mean target:**   Target values for the four calibration parameters $b_{mx}$, $b_{my}$, $z_R$ and $R$ are calculated based on the idea of moving the current 2D magnetometer calibration circle so that its perimeter is just touching the weighted mean of the magnetometer data. If $(\bar{m}_x, \bar{m}_y, \bar{m}_z)$ is the weighted mean of the data points and $(b_{mx}, b_{my}, z_R, R)$ is the current set of calibration parameters, this corresponds to setting a target of

$$(\breve{b}_{mx}, \breve{b}_{my}) = (\bar{m}_x, \bar{m}_y) + R\hat{\mathbf{d}}_m, \tag{4.41a}$$

$$(\breve{z}_R, \breve{R}) = (\bar{m}_z, R). \tag{4.41b}$$

where

$$\hat{\mathbf{d}}_m = \frac{(b_{mx}, b_{my}) - (\bar{m}_x, \bar{m}_y)}{\|(b_{mx}, b_{my}) - (\bar{m}_x, \bar{m}_y)\|} \tag{4.42}$$

is the unit vector pointing from the weighted mean of the data to the current centre of the magnetometer calibration circle. The confidence $\breve{C} \in [0, 1]$ of the mean target is high if the data is not close to the current centre of the calibration circle, and is clustered in a small area of the measurement space, i.e. the robot did not rotate much during the recording of the data. Numerically, the confidence is computed as a combination of the planar spread $S_p$ and the proximity of the weighted mean $(\bar{m}_x, \bar{m}_y)$ to the current calibration values $(b_{mx}, b_{my})$.

6. **Final target:** The final target values of the calibration para-
   meters $b_{mx}$, $b_{my}$, $z_R$ and $R$ are calculated by weighted linear
   interpolation between the fitted and mean target values. If $\tilde{C}_f$ is
   the confidence of the fitted target and $\tilde{C}_m$ is the confidence of
   the mean target, then the interpolation factor $u \in [0, 1]$ from the
   mean target towards the fitted target is given by

   $$u = \frac{\tilde{C}_f(1 + \tilde{C}_m)}{\tilde{C}_f + \tilde{C}_m}. \tag{4.43}$$

   Using this factor, all four calibration parameters and the corres-
   ponding target confidences are linearly interpolated to give the
   final target parameters $\tilde{b}_{mx}$, $\tilde{b}_{my}$, $\tilde{z}_R$ and $\tilde{R}$. For instance, keeping
   in mind Equations (4.40) and (4.41b), the final values of $\tilde{z}_R$ and
   $\tilde{C}$ are given by

   $$\tilde{z}_R = uc_z + (1 - u)\bar{m}_z, \tag{4.44a}$$
   $$\tilde{C} = u\tilde{C}_f + (1 - u)\tilde{C}_m. \tag{4.44b}$$

   The final calculated target values are used in Step 5 on page 70 to
   adjust and improve the 2D magnetometer calibration parameters,
   and in particular account for any changes in local hard iron
   effects over time.

The autocalibration procedure is visualised in Figure 4.4. A variety of
example calibration situations are shown, along with the respective
fitted, mean and final target parameter circles, which are subsequently
used to adjust the magnetometer calibration parameters using the
slope-limited low-pass filters explained previously.

### 4.3.6 Reference Field Vector

In order for a robot to be able to orient itself in an environment, it
needs to have a fixed reference point relative to which it can judge its
heading. The fully calibrated and corrected magnetometer data is used
for this purpose. When the robot is brought to a new environment (or
the magnetometer calibration changes), a robot heading calibration
is required to set the required fixed reference point in the form of a
magnetometer reading. The robot is brought to an upright position
facing the intended heading reference direction, and when triggered,
the current corrected output of the magnetometer is saved as the so-
called reference field vector. The reference field vector is essentially a
direct estimate of ${}^G\mathbf{m}_e$—the magnetic field vector of the Earth relative
to the global frame {G} corresponding to the (upright) nominal heading
reference direction of the robot. Deviations of the current corrected
magnetometer measurements to this reference vector are used, for
example, in Section 10.1.5.1 to estimate the heading of the robot as
part of the attitude estimation process.

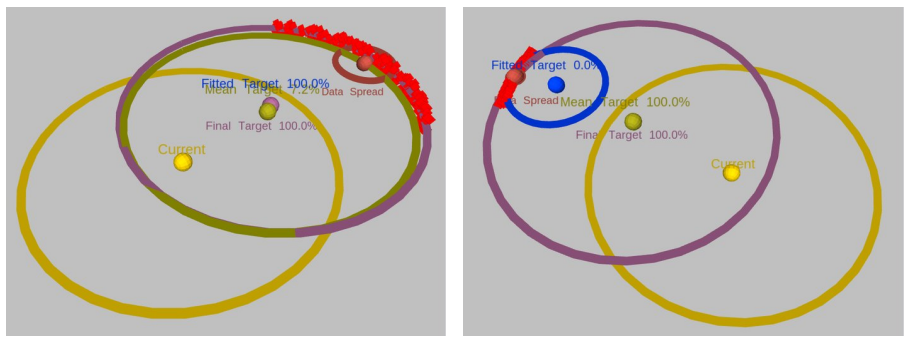

(a) A case where the fitted circle has low residual error, high angular spread, and a radius close to that of the current calibration circle. The result is great confidence in the fitted target (100%), which despite a small confidence in the mean target is then consequently used as the final target.

(b) A case where the fitted circle is unbelievable (0%) due to its radius, but the mean target has full confidence (100%) due to the small planar spread of the data and the lack of proximity between the weighted data mean and the current circle centre.

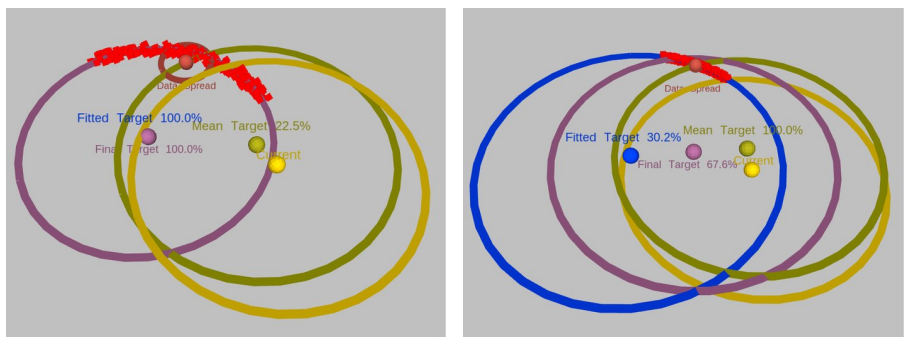

(c) A case demonstrating how the fitted target is given a certain level of priority over the mean target, due to the effects of Equation (4.43). The mean target has a low to moderate confidence, but does not contribute to the final target due to the 100% confidence of the fitted target. This is because fitted targets, by nature, are a stronger general indication of the true hard iron offset of the magnetometer than mean targets.

(d) A case where the mean target is 100% confident due to the small planar spread of the data, but the fitted target is only moderately confident due to the small angular spread of the data and the fact that the fitted radius is noticeably larger than the radius of the current calibration circle. The final target circle is an interpolation between both targets (with 67.6% confidence), pending further more decisive magnetometer data.

Figure 4.4: Example situations of the magnetometer autocalibration procedure. In each image the buffered magnetometer data points are indicated in red, the current calibration circle is indicated in yellow, and the fitted, mean and final target calibration circles are indicated in blue, dark yellow and violet, respectively.

## 4.4 SENSOR CALIBRATION IN THE BIGGER PICTURE

In Chapter 2, the robot platforms in use in this thesis, and the software framework they are running, were both presented in detail. In particular, it was explained how a main robot control loop is used to perform trajectory generation and motion control with the help of so-called motion modules, and how the output joint commands from these modules are passed to the actuator control scheme (Chapter 3) for gravity, inertia and friction compensation. After sending out the compensated joint commands to the servos, the main robot control loop was explained to continue with the acquisition of all available sensor data, before looping back to the start and repeating the process.

This chapter, Chapter 4, detailed how the raw acquired sensor data can be processed into a more useful form with the help of numerous calibration procedures, ranging from joint position calibrations for the joint encoder sensors, to IMU orientation calibrations for the gyroscope and accelerometer sensors, to hard iron offset calibrations for the magnetometer sensor. How this processed sensor data is filtered and used for higher level state estimation however (required for the purposes of feedback to the motion modules), has not yet been addressed, and is the topic of Chapter 10. In particular, Section 10.1 details how the calibrated gyroscope, accelerometer and magnetometer data can be fused together in an attitude estimation process to estimate the orientation of the robot. Before we can deal with the estimation of orientations and headings of the robot however, and use these to construct bipedal walking gaits, we first need a greater understanding of rotations and general rotation formalisms in 3D. In particular, we need to be able to answer simple questions like

**What is 'yaw' and how should we quantify it?**

**Is there more than one possible definition of yaw?**

**Which definition corresponds to our natural intuition?**

Part III of this thesis, i.e. the next three chapters, deals with these and many other related questions, including for example questions pertaining to the concepts of 'pitch' and 'roll'. Building on this gained knowledge, Part IV then develops effective algorithms for the tasks of state estimation and bipedal walking.

Part III

ROTATION FORMALISMS IN 3D

# PART CONTENTS

Due to significant use of section nesting, the following is an overview of the chapters and sections in this part that is more detailed than the main table of contents:

# REPRESENTATIONS OF 3D ROTATIONS

---

In order to arrive at a gait that uses orientation feedback to maintain stability, the orientation of the robot must first be estimated and represented in a way that is both meaningful and useful. It is shown in Chapter 10 how 3D nonlinear filtering techniques can be used to estimate the orientation of a robot, but the final result is given by four quaternions parameters in the range from $-1$ to $1$. On the face of it, without making any further conversions or calculations, it is not obvious how to interpret these four parameters so as to know basic things about the resulting estimated orientation, such as what the heading of the robot is, or how close the robot is to falling down in some direction. It is also not an easy task for a human to interpret the four parameters to understand what the exact nature of the orientation is. This chapter seeks to develop methods and concepts that can be used to extract meaning from such expressions of orientation, both for algorithmic purposes, for example to stabilise motions using orientation feedback, and for the ease of human interpretation and visualisation.

The applications of newly developed formulations of, for example, yaw are also useful deeper inside the state estimation itself. In Section 10.1.5.3, the concept of *fused yaw* is used to resolve a measured orientation of the Inertial Measurement Unit (IMU) in the case that no magnetometer is available. In this chapter, we take a systematic approach to placing the idea of fused yaw into context by developing and presenting three new highly intertwined representations of rotations, namely *tilt angles*, *fused angles* and the *tilt phase space*, that all depend fundamentally on this notion of yaw. Rotation representations, interchangeably referred to as rotation parameterisations, are simply put a way of assigning numeric values or *parameters* to a rotation. Here we are considering only the three-dimensional space of 3D rotations, so the number of parameters must be at least three. Some rotation representations consist of exactly three parameters, like Euler angles and rotation vectors, but some consist of four, like quaternions and axis-angle pairs, and others consist of even more, like rotation matrices, which consist of nine parameters. A full review of existing rotation representations is given in Section 5.3.

While this chapter has been kept relatively focused on presenting the new rotation representations, how to work with them, and what basic properties they have, a deeper investigation that somewhat completes the picture with a multitude of further properties and results is provided in Chapter 7. Meanwhile, Chapter 6 is dedicated entirely to answering the question of why the development of new rotation representations was even necessary, and why Euler angles are not an appropriate solution. All of the presented conversions and algorithms for working with tilt angles, fused angles and the tilt phase space have been implemented in both C++ and Matlab, and have been released open source for anyone to use (Allgeuer, 2018c; d), to provide support with using these new representations. The libraries are seen as a test bed for the development of numerical rotation-related algorithms.

## 5.1    NOTATION AND CONVENTIONS

Before we can begin our discussion of rotation representations, some basic notation and conventions must be introduced. All 3D coordinate frames used in this thesis are right-handed, i.e. such that

$$\mathbf{x} \times \mathbf{y} = \mathbf{z}, \tag{5.1}$$

and are denoted by capitalised letters like $A$, $B$, $C$, which are enclosed in parentheses when referring to them directly in text, like for example 'frame {A}'. Frame {G} always refers to the global coordinate frame, and if there is a robot or body of which the orientation is being expressed, frame {B} always refers to the body-fixed coordinate frame.

The convention that the global z-axis $\mathbf{z}_G$ points 'upwards' relative to the environment is used in this thesis. In almost all application scenarios, this implies that the global z-axis is taken to point in the exact opposite direction to gravity, as the assessment and control of balancing bodies in 3D in a gravitational field is the main conceived application of this work on rotations. However, in some applications involving contacts, it can also be useful to alternatively define the global z-axis based on a particular surface normal, so that concepts like yaw and tilt quantify the relative orientation of the two contacting bodies. In either case, the choice of this 'up' direction ensures that definitions such as that of yaw, pitch and roll make terminological sense in consideration of the true rotations being undergone relative to the environment. All derived rotation formulas and results could easily be rewritten using an alternative convention of which coordinate axis points 'up', if this were to be desired. Note that, in the few cases where it is relevant, the further convention that the x-axis points 'forwards' and the y-axis points 'leftwards' is followed. This can be of relevance when for example assigning a coordinate frame to a robot.

The three unit vectors corresponding to the positive coordinate axes of a frame {A} are denoted $\mathbf{x}_A$, $\mathbf{y}_A$ and $\mathbf{z}_A$, where these vectors are expressed by default in the coordinates of the global frame {G}. Expressed in the coordinates of another frame {C}, these same coordinate axes of {A} are given by ${}^C\mathbf{x}_A$, ${}^C\mathbf{y}_A$ and ${}^C\mathbf{z}_A$. Thus, in terms of notation, the following are equivalent:

$$\mathbf{x}_A \equiv {}^G\mathbf{x}_A, \tag{5.2a}$$

$$\mathbf{y}_A \equiv {}^G\mathbf{y}_A, \tag{5.2b}$$

$$\mathbf{z}_A \equiv {}^G\mathbf{z}_A. \tag{5.2c}$$

We further follow the convention that the components of coordinate axes are formed by appending $x$, $y$ or $z$ to the subscript, for example

$$^G\mathbf{z}_B = ({}^Gz_{Bx}, {}^Gz_{By}, {}^Gz_{Bz}). \tag{5.3}$$

In general, vectors are typeset in bold font like '$\mathbf{v}$', and a basis coordinate frame can be explicitly specified, e.g. ${}^A\mathbf{v}$, if the vector is not just by default in the coordinates of the global frame {G}. The vector components are then by default denoted

$$\mathbf{v} = (v_x, v_y, v_z), \tag{5.4a}$$

$$^A\mathbf{v} = ({}^Av_x, {}^Av_y, {}^Av_z). \tag{5.4b}$$

Note that unit vectors by convention are often denoted with 'hatted' variables, like '$\hat{\mathbf{e}}$', and that the vector components by default are then still denoted

$$\hat{\mathbf{e}} = (e_x, e_y, e_z). \tag{5.5}$$

Using rotation matrices as an example, the rotation from frame {G} to {B} is denoted ${}^G_BR$. This notation is kept consistent between the different rotation representations, in that for example the quaternion rotation from {G} to {B} is analogously denoted ${}^G_Bq$. If a rotation is specified to occur about a particular axis or vector, then a positive rotation is <span style="color:magenta">counterclockwise</span> (<span style="color:blue">CCW</span>) when viewed from where the vector is pointing to, and a negative rotation is <span style="color:magenta">clockwise</span> (<span style="color:blue">CW</span>).

When working with rotation representations, many identities that are derived involve plentiful instances of trigonometric functions. As such, to save equation space and ease discernment of certain overlying patterns, shorthand notation is often used to denote the $\sin(\cdot)$ and $\cos(\cdot)$ functions, namely,

$$s_* \equiv \sin(*), \tag{5.6a}$$

$$c_* \equiv \cos(*), \tag{5.6b}$$

$$s_{\bar{*}} \equiv \sin(\tfrac{1}{2}*), \tag{5.6c}$$

$$c_{\bar{*}} \equiv \cos(\tfrac{1}{2}*), \tag{5.6d}$$

where $*$ is any scalar angular variable, and $\bar{*} \equiv \tfrac{1}{2}*$. Thus, for example,

$$s_{\bar{\psi}} \equiv \sin(\bar{\psi}) \equiv \sin(\tfrac{1}{2}\psi). \tag{5.7}$$

## 5.2   MOTIVATION AND AIMS

This work on 3D rotations was motivated by the development of gait stabilisation algorithms for bipedal robot platforms, or more generally, the analysis and control of balancing bodies in 3D. Walking bipedal robots are seen to be balancing bodies as they are under constant influence of gravity, and constantly aim to keep their state of balance upright and stable, despite changing support conditions and possible disturbances. The need for the development of new representations of 3D rotations emerged from the need to answer certain basic questions about the orientations of such walking robots—questions that were not well answered by any other existing rotation parameterisations.

### 5.2.1   Core Aims: Amount of Rotation in the Major Planes

Keeping with the context of walking bipedal robots, suppose we have a robot with a body-fixed coordinate frame {B} attached to its trunk. Assuming the global orientation of this frame ${}^G_B q$ can be estimated, e.g. using IMUs and other sensors, the state of balance of the robot is entirely encoded in this rotation. It is not uncommon for a gait to treat the sagittal and lateral planes of balance relative to the robot, i.e. the forwards/backwards and sideways planes of balance, separately, so for the purposes of constructing stabilising feedback it is useful to be able to answer questions like (see Figure 5.1):

- How rotated is the robot in the **transverse** plane, i.e. what is the heading of the robot?

- How rotated is the robot in the **sagittal** plane?

- How rotated is the robot in the **lateral** plane?

We are searching for a rotation representation that, amongst other things, can provide good answers to these questions. These are not the only requirements that we are looking for however, to find the 'ideal' rotation representation for the analysis and control of balancing bodies in 3D. Our further requirements and expectations are outlined in the remainder of this section (as well as in the one that follows).

As illustrated in Figure 5.2, given any orientation of the robot, we wish to be able to answer the three questions above with three *scalar angular values* that we can write in the three figurative boxes shown. We will refer to these values as 'yaw', 'pitch' and 'roll' respectively, and together, these three values should define the entire orientation. By definition, the three values should quantify the 'amount of rotation', whatever that is chosen to mean, within the **xy**, **xz** and **yz** major planes (see Figure 5.1).[1] For ease of interpretation, we require the

---

1 This can approximately be thought of as the 'amount of rotation' about the corresponding z, y and x-axes respectively, but not quite.

and

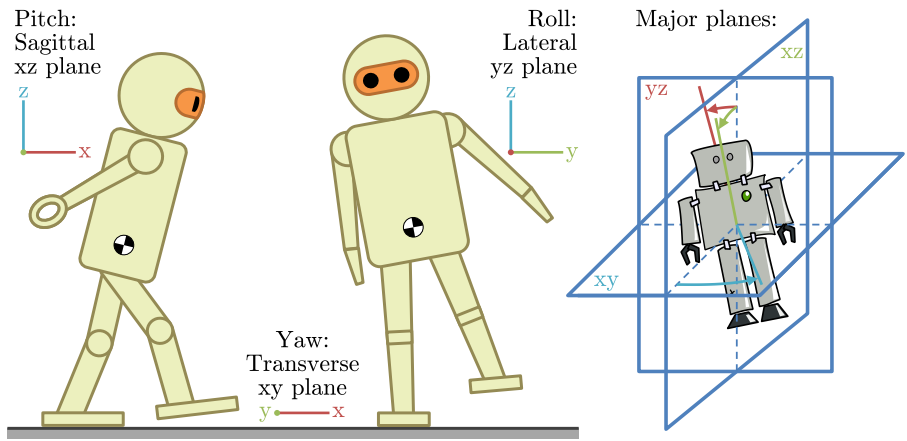

Figure 5.1: Illustration of the three major planes of balance, the sagittal **xz** plane (pitch), lateral **yz** plane (roll), and transverse **xy** plane (yaw). For the purpose of being able to analyse and control the state of balance of the robot, we wish to be able to quantify the 'amount of rotation' a robot has individually within each of these three planes, as suggested by the arrows on the right-hand image.

values to just be angles, i.e. in units of radians, and loosely speaking we should expect that bigger values lead to bigger rotations. As a final requirement (for now), in order to be intuitive and geometrically useful, the sole effect of applying a global z-rotation to the orientation of the robot should be to additively affect the yaw value. That is, if the robot is rotated by $\frac{\pi}{4}$ rad about the global positive z-axis, then only the yaw value should change, and up to angle wrapping, should change by exactly $+\frac{\pi}{4}$ rad. This property is referred to as yaw additivity.

The requirements listed so far are all relatively elemental and understandable, but in fact, other than Euler angles, no standard existing rotation representation is in a position to be able to satisfy these requirements, and to provide three values to write into the three boxes. The list of existing rotation representations is introduced in detail in the following section (see Section 5.3), but as an example, quaternions and rotation matrices clearly fail to provide any scalar angular values at all, let alone ones that directly quantify the amount of rotation in each of the three major planes, and rotation vectors in particular fail the required additivity property of yaw. Euler angles may at first *seem* to be a satisfactory solution to the given requirements, being a commonly accepted catch-all solution, but this is not entirely so. They can often enough be the correct choice for a task, such as for example the modelling of gimbals or a colocated series of joints, but too often they are chosen simply because there does not *seem* to be a reasonable alternative. This thesis provides two such alternatives and dedicates an entire chapter (see Chapter 6) to an in-depth explanation of the shortcomings of Euler angles, and why they are unsuitable for balance-related tasks. In fact, other than for specific situations like just previously mentioned, where the task itself inherently and physically involves a series of well-defined elemental rotations, it is unclear what reason there would be to ever use Euler angles at all over a construct like the tilt phase space (introduced in this chapter). The main problems with Euler angles revolve around the location of their singularities, the lack of parameter axisymmetry, and the lack of complete parameter independence. All of these points are discussed in detail in Chapter 6, with an adjoining discussion of why the representations introduced in this chapter do not have any such problems.

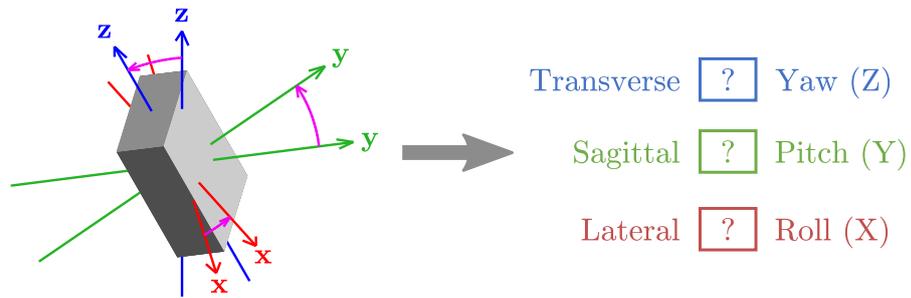

Figure 5.2: Given any 3D orientation, represented here on the left by the rotation of a rectangular prism, we wish to be able to write three scalar angular values in the three figurative boxes shown, where together the three values unambiguously define the entire orientation. The three values should relate to the 'amounts of rotation' within the three major planes, which can loosely be thought of as the amounts of rotation about the x, y and z-axes, and will be referred to as the 'roll', 'pitch' and 'yaw' of the rotation respectively.

### 5.2.2 Further Aims: Partitioning of Rotations into Yaw and Tilt

Three new rotation representations are introduced in this chapter, namely the tilt angles, fused angles and tilt phase space parameterisations. While tilt angles form an intermediary representation that intuitively demonstrates how to partition rotations into meaningful notions of *yaw* and *tilt*, namely the fused yaw and tilt rotation components, the other two representations follow suit, but reparameterise the tilt component to provide *pitch* and *roll* components, as required by the motivational scenario of the three boxes describing the amount of rotation in the three major planes (see Figures 5.1 and 5.2). The fused angles and tilt phase space parameterisations are easy ways of quantifying 3D rotations, just like Euler angles are, but ways that have been specifically developed for balance-related scenarios and mobile robotics, and solve numerous problems and imperfections that Euler angles have. They describe the state of balance of a robot in an intuitive and problem-relevant way, and in particular the fused angles parameterisation offers a useful geometric interpretation as well. Most importantly however, both new parameterisations use *concurrent* spe-cifications of pitch and roll, in that no order of rotations is imposed on the two like for Euler angles. That is, the tilt component in each case is a single atomic rotation that is quantified by the two scalar angular pitch and roll values, but no subdivision into separate pitch and/or roll rotations can logically be made. This is important as it enables the definitions of pitch and roll to be axisymmetric, a point that is discussed in more detail in Section 6.2.4.

Ultimately, when using rotation parameterisations to represent the orientation of a robot for the purpose of a balance-related task such as a bipedal gait, the greatest concern is how close the robot is to falling over. This can also be characterised as how far the robot is from being upright, so naturally the chosen way of partitioning rotations into yaw and tilt components should reflect that. Specifically, the yaw component should encapsulate the heading of the robot, i.e. the horizontal planar 360° bearing of the robot that it is deemed to be facing, and the tilt component should purely encapsulate the remaining heading-independent relation between the robot and the planar ground, i.e. how far the robot is from being upright no matter what direction it is facing. This is why the tilt component is sometimes referred to as a heading-independent balance state, as it should indicate what direction locally relative to the robot's heading the robot is falling in, if any.

How far a robot is from being upright can equally be seen as a measurement of how far the gravity vector is away from pointing straight down relative to the robot's body-fixed coordinate frame {B}. In fact, the direction of this gravity vector (i.e. $-{}^B\mathbf{z}_G$) is exactly what an accelerometer attached to the robot would measure under quasi-static conditions, and it can be seen that this vector is completely independent to any changes in heading, i.e. any applications of global z-rotations to {B}. This leads to the further aim that in order to suitably represent the heading-independent balance state, the tilt component of a rotation should be defined in such a way that there is a one-to-one correspondence in general to the set of possible measured gravity directions, and therefore also to the unit sphere of possible ${}^B\mathbf{z}_G$.

Rotations in general can always be viewed as a single rotation by an angle $\theta_a$ about a unit vector axis $\hat{\mathbf{e}} = (e_x, e_y, e_z)$ (see Section 5.3.2). This characterisation of rotations can also be used to set expectations for how the definitions of the yaw and tilt components should behave. Specifically, the z-component of $\hat{\mathbf{e}}$, namely $e_z$, encodes a measure of the proportion of the rotation that is about the z-axis, where we recall that by convention the z-axis points 'upwards'. Thus, we should expect that the tilt component of a rotation has no $e_z$ component, and that conversely, the yaw component of a rotation is purely a function of $\theta_a$ and $e_z$, and completely independent of $e_x$ and $e_y$—the proportions of the rotation about the x and y-axes.

The method of partitioning 3D rotations into fused yaw and tilt rotation components, as presented in this chapter, satisfies all of the aforementioned expectations and requirements.

### 5.2.3   Existing Applications

The tilt angles, fused angles and tilt phase space parameterisations have to date found many applications in many scenarios. Specifically in the context of this thesis, these applications include:

- An attitude estimator that, in the absence of absolute heading information, uses fused yaw to reconstruct measured orientations that are most similar to the current rotation estimate. [Section 10.1.5]

- Estimation of the local Centre of Mass (CoM) position and velocity based on estimated tilt orientations and applied foot constraints. [Section 10.2.1]

- Dead reckoning of the locomotion of a robot, using the split between fused yaw and tilt to mathematically model foot orientations and ground contacts. [Section 10.2]

- A direct fused angle feedback controller that applies corrective actions in the fused angles space to stabilise a walking robot. [Chapter 13]

- A walking motion generator with a wide array of inbuilt corrective actions, mostly defined in the tilt angles and tilt phase spaces. [Chapter 14]

- Cubic spline interpolation of a series of 3D foot orientations for the purpose of automatically calculating suitable intermediate angular velocities. [Section 14.2.4.10]

- A tilt phase feedback controller that calculates activations for a myriad of corrective actions in the tilt phase space to stabilise a walking robot. [Chapter 15]

- The Humanoid Open Platform ROS Software (2018), which incorporates the source code for almost all parts of this thesis and utilises tilt angles and fused angles in many scenarios, including for example for the implementation of Proportional-Integral-Derivative (PID) orientation feedback during the playback of keyframe-edited motions. [Section 2.2.3]

- The Matlab/Octave Rotations Library (2018), which implements a wide range of conversion and computation functions related to tilt angles, fused angles and the tilt phase space, as well as the majority of relevant existing rotation representations.

- The Rotations Conversion Library (2018), which is a fully-featured port of the aforementioned library to C++.

## 5.3 EXISTING ROTATION REPRESENTATIONS

Numerous ways of representing a rotation in three-dimensional Euclidean space have been developed and refined over the years. Many of these representations arose naturally from classical mathematics, and have found widespread use in areas such as physics, engineering and robotics. Different representations have different advantages and disadvantages, and which representation is suitable for a particular application depends on a wide range of considerations. Such considerations include:

- Ease of geometric interpretation, in particular in a form that is relevant to the particular problem,

- The range of singularity-free behaviour,

- Computational efficiency in terms of common operations such as rotation composition and vector rotation,

- Mathematical convenience, in terms of numeric and algebraic complexity and manipulability, and

- Algorithmic convenience, in the sense of a representation potentially possessing properties that can conveniently be exploited for a particular algorithm.

A wide range of existing rotation representations are reviewed in this section as a basis of comparison to the ones newly developed in this chapter. Due to the dimensionality of the space of 3D rotations, a minimum of three parameters is required for any such representation. A representation with exactly three parameters is referred to as minimal, while other representations with a greater number of parameters are referred to as redundant.

### 5.3.1 Rotation Matrices

A rotation can be represented as a linear transformation of coordinate frame basis vectors, expressed in the form of an orthogonal matrix of unit determinant. Due to the strong link between such transformation matrices and the theory of direction cosines, the name direction cosine matrix is also sometimes used. The space of all rotation matrices is called the special orthogonal group SO(3), and is defined as

$$SO(3) = \{R \in \mathbb{R}^{3 \times 3} : R^T R = \mathbb{I}, \det(R) = 1\}, \tag{5.8}$$

where $\mathbb{R}^{3 \times 3}$ is the set of all $3 \times 3$ matrices with real entries. It is important to note that the special orthogonal group SO(3) has an exact one-to-one correspondence with the space of all 3D rotations, so it is often interchangeably used to denote it.

We first note from Equation (5.8) that all rotation matrices have the property that

$$R^T = R^{-1}. \tag{5.9}$$

Rotation of a vector $\mathbf{v} \in \mathbb{R}^3$ by a rotation matrix is given by matrix premultiplication. Furthermore, for a rotation from coordinate frame {G} to {B}, we have that

$$
{}^G_B R = \begin{bmatrix} \uparrow & \uparrow & \uparrow \\ {}^G\mathbf{x}_B & {}^G\mathbf{y}_B & {}^G\mathbf{z}_B \\ \downarrow & \downarrow & \downarrow \end{bmatrix} = \begin{bmatrix} \leftarrow & {}^B\mathbf{x}_G & \rightarrow \\ \leftarrow & {}^B\mathbf{y}_G & \rightarrow \\ \leftarrow & {}^B\mathbf{z}_G & \rightarrow \end{bmatrix}, \tag{5.10}
$$

where ${}^G\mathbf{y}_B$, for example, is the column vector corresponding to the y-axis of frame {B} expressed in the coordinates of frame {G}, as discussed in Section 5.1. With nine parameters, rotation matrices are clearly a redundant parameterisation of the rotation space. They are quite useful in that they are free of singularities, easy to compose, and trivially expose the basis vectors of the fixed and rotated frames, but for many tasks they are not as computationally and/or numerically suitable as other representations. One problem in particular, is that the numerical reorthogonalisation of a nearly valid rotation matrix is a non-trivial and involved process.

### 5.3.2 Axis-angle and Rotation Vector Representations

By Euler's rotation theorem, every rotation in three-dimensional Euclidean space can be expressed as a single CCW rotation by up to $\pi$ radians about some axis. As such, every 3D rotation can be assigned a vector-scalar axis-angle pair

$$
\begin{aligned}
A &= (\hat{\mathbf{e}}, \theta_a) \\
&\in \mathcal{S}^2 \times [0, \pi] \equiv \mathbb{A},
\end{aligned} \tag{5.11}
$$

where $\hat{\mathbf{e}} = (e_x, e_y, e_z)$ is a unit vector corresponding to the axis of rotation, $\theta_a$ is the angular magnitude of the rotation, and $\mathcal{S}^2$ is the unit sphere in $\mathbb{R}^3$. The domain $\mathbb{A}$ is the set of all axis-angle pairs, where it should be noted that certain pairs correspond to the same final coordinate frame, like for example $(\hat{\mathbf{e}}, \pi)$ and $(-\hat{\mathbf{e}}, \pi)$. By convention, for numerical and/or algorithmic benefits, the identity rotation is often chosen to be represented by $\hat{\mathbf{e}} = \mathbf{0}$, even though strictly speaking $\mathbf{0} \notin \mathcal{S}^2$. If we consider values of $\theta_a$ outside of the default $[0, \pi]$ interval, the following representational equivalences hold:

$$(\hat{\mathbf{e}}, \theta_a) \equiv (\hat{\mathbf{e}}, \theta_a + 2\pi k), \tag{5.12a}$$

$$(\hat{\mathbf{e}}, \theta_a) \equiv (-\hat{\mathbf{e}}, -\theta_a), \tag{5.12b}$$

for $k \in \mathbb{Z}$ an integer. Incidentally, these equivalences demonstrate how any $(\hat{\mathbf{e}}, \theta_a) \in \mathcal{S}^2 \times \mathbb{R}$ can be collapsed into an equivalent expression in the default domain $\mathbb{A} = \mathcal{S}^2 \times [0, \pi]$. A closely related concept to the axis-angle representation is the rotation vector, given by

$$\mathbf{e} = \theta_a \hat{\mathbf{e}} \in \mathbb{R}^3. \tag{5.13}$$

Effectively, rotation vectors are 3D vectors that encode the magnitude and axis of a 3D rotation in terms of their vector norm and direction.

While the axis-angle representation is redundant, the rotation vector representation, with only three parameters, is classified as minimal. Both representations however suffer from a general impracticality of mathematical and numerical manipulation. For example, no formula for rotation composition exists that is more direct than converting to quaternions and back. Rotation vectors do indeed provide three different angular values that quantify different dimensions of the rotation, but they do not intuitively or otherwise well-define the amount of rotation in the three major planes, and the z-component does not satisfy the necessary additivity condition (see Section 5.2.1) that is required for it to constitute a usable concept of yaw.

### 5.3.3 Quaternions

Quaternions in general are an extension of the complex numbers to four dimensions. While complex numbers define the imaginary number $i$ such that $i^2 = -1$, and describe every possible complex element as a combination $a + bi$, quaternions define three such numbers $i$, $j$ and $k$ such that

$$i^2 = j^2 = k^2 = ijk = -1, \tag{5.14}$$

and describe every possible quaternion as a combination[2]

$$q = a + bi + cj + dk. \tag{5.15}$$

While this is the classic definition of general quaternions, it is more common and useful in the context of 3D rotations to view quaternions just as their four real coefficients $(a, b, c, d)$. With a suggestive switch in coefficient labels, the set of all general quaternions becomes

$$\mathbb{H} = \{q = (w, x, y, z) \in \mathbb{R}^4\}. \tag{5.16}$$

---

2 Note that as a result of Equation (5.14), one can deduce that $ij = k$, $jk = i$ and $ki = j$. Thus, using these basic rules, products of quaternions can always be simplified down again to the canonical form given in Equation (5.15).

The relation to the classic definition of quaternions is then given by

$$1 \equiv (1, 0, 0, 0) \in \mathbb{H}, \tag{5.17a}$$

$$i \equiv (0, 1, 0, 0) \in \mathbb{H}, \tag{5.17b}$$

$$j \equiv (0, 0, 1, 0) \in \mathbb{H}, \tag{5.17c}$$

$$k \equiv (0, 0, 0, 1) \in \mathbb{H}. \tag{5.17d}$$

For $q = (w, x, y, z)$, the w-component is referred to as the scalar part of $q$, and the remaining three components are referred to as the vector part of $q$. This leads to the vector notation of a quaternion $q$, namely

$$q = (w, x, y, z) = (q_0, \mathbf{q}), \tag{5.18}$$

where

$$q_0 = w \in \mathbb{R}, \tag{5.19a}$$

$$\mathbf{q} = (x, y, z) \in \mathbb{R}^3. \tag{5.19b}$$

In fact, all scalars $a \in \mathbb{R}$ and vectors $\mathbf{v} \in \mathbb{R}^3$ can be identified with their corresponding purely scalar or vector quaternions, given by

$$a \equiv (a, 0, 0, 0) \in \mathbb{H}, \tag{5.20a}$$

$$\mathbf{v} \equiv (0, v_x, v_y, v_z) \in \mathbb{H}. \tag{5.20b}$$

Pure rotations in 3D space can be represented by the set of quaternions of unit norm. The set of all quaternion rotations is thus

$$\mathbb{Q} = \{ q \in \mathbb{H} : \|q\| = 1 \}, \tag{5.21}$$

where we note in particular that for $q \in \mathbb{Q}$, this implies that

$$w^2 + x^2 + y^2 + z^2 = 1. \tag{5.22}$$

The four quaternion rotation parameters are sometimes also referred to as Euler-Rodrigues parameters. The space $\mathbb{Q}$ corresponds to the four-dimensional unit sphere $\mathcal{S}^3$, and is a double cover of the space of rotations SO(3), in that $q$ and $-q$ both correspond to the same 3D rotation. That is,

$$q \equiv -q. \tag{5.23}$$

Quaternion rotations with $w \geq 0$ can be related to their corresponding axis-angle representation $(\hat{\mathbf{e}}, \theta_a) \in \mathbb{A}$, and thereby visualised to some degree in terms of what rotation they represent, using

$$q = (q_0, \mathbf{q}) = \left( \cos \tfrac{\theta_a}{2}, \hat{\mathbf{e}} \sin \tfrac{\theta_a}{2} \right), \tag{5.24a}$$

$$= \left( \cos \tfrac{\theta_a}{2}, e_x \sin \tfrac{\theta_a}{2}, e_y \sin \tfrac{\theta_a}{2}, e_z \sin \tfrac{\theta_a}{2} \right) \in \mathbb{Q}. \tag{5.24b}$$

The use of quaternions to express rotations generally allows for very computationally efficient calculations, and is grounded by the well-established field of quaternion mathematics. A crucial advantage of the quaternion representation is that it is free of singularities. On the other hand, as previously mentioned, it is not a one-to-one mapping of the special orthogonal group, as $q$ and $-q$ both correspond to the same rotation. The redundancy of the parameterisation, as there are four parameters, also means that the unit magnitude constraint has to explicitly and sometimes non-trivially be enforced in numerical computations. Furthermore, no clear geometric interpretation of quaternions exists beyond the implicit relation to the axis-angle representation given in Equation (5.24). In light of the motivation and aims set out in Section 5.2, for applications relating to the analysis and control of balancing bodies in 3D, quaternions can be very helpful in performing state estimation (e.g. Chapter 10), but a final quaternion orientation of the trunk does not give any direct insight into the amount of rotation within the three major planes. As such, quaternions are very useful computationally, but do not by themselves provide the answers we are looking for.

### 5.3.4 Vectorial Parameterisations

All rotation parameterisations can be generally classified as being either vectorial or non-vectorial. This classification relates to whether a given set of parameters behaves like the three Cartesian coordinates of a geometric vector in 3D space or not, and whether the parameter vector furthermore behaves vectorially with respect to a change of reference frame of the rotation, that is, behaves like matrix premultiplication when the reference frame is changed. Clearly, due to the dimension of rotation matrices and the 3D space, only minimal parameterisations, which have exactly three parameters, can be vectorial. Quaternions, for example, are therefore not vectorial, but if the scalar w-component is omitted and always implicitly defined to be the unique positive value that completes the remaining three parameters to unit norm, then the so-called reduced Euler-Rodrigues parameters result, and this representation is vectorial. Recalling that $(\hat{\mathbf{e}}, \theta_a) \in \mathbb{A}$ is the axis-angle representation of a rotation, it can be seen from Equation (5.24) that the reduced Euler-Rodrigues parameters are just

$$\mathbf{q} = \hat{\mathbf{e}} \sin \tfrac{\theta_a}{2}. \tag{5.25}$$

It is easy to see that for a change of reference frame, $\hat{\mathbf{e}}$ simply changes its basis, which corresponds to matrix premultiplication, as required. In general, all three-dimensional parameterisations of the form

$$\mathbf{p} = p(\theta_a)\,\hat{\mathbf{e}}, \tag{5.26}$$

Table 5.1: Summary of the definitions of the various common vectorial parameterisations, identifying for each the respective generating function $p(\theta_a)$ and normalisation factor $\kappa$.

| Parameterisation | $p(\theta_a)$ | $\kappa$ |
|---|---|---|
| Rotation vector | $\theta_a$ | $1$ |
| Linear parameters | $\sin \theta_a$ | $1$ |
| Reduced Euler-Rodrigues | $\sin \frac{\theta_a}{2}$ | $\frac{1}{2}$ |
| Cayley-Gibbs-Rodrigues | $2 \tan \frac{\theta_a}{2}$ | $1$ |
| Wiener-Milenković | $4 \tan \frac{\theta_a}{4}$ | $1$ |

are vectorial by the same logic, where $p(\theta_a)$ is the generating function of the parameterisation, and is any odd function of the rotation angle $\theta_a$ such that

$$\lim_{\theta_a \to 0} \frac{p(\theta_a)}{\theta_a} = \kappa, \tag{5.27}$$

for some real normalisation factor $\kappa > 0$. For convenience, $p(\theta_a)$ is usually chosen to be scaled in such a way that $\kappa = 1$.

The most basic example of a vectorial parameterisation of this form is the rotation vector, which results from trivially choosing $p(\theta_a) = \theta_a$. Many other representations that fit into this mathematical framework have been independently developed over the years however, for various specific analytical and algorithmic purposes. For example, the Cayley-Gibbs-Rodrigues parameters are given by

$$\mathbf{p} = 2 \tan \frac{\theta_a}{2} \, \hat{\mathbf{e}}, \tag{5.28}$$

the Wiener-Milenković parameters, also known as the conformal rotation vector, are given by

$$\mathbf{p} = 4 \tan \frac{\theta_a}{4} \, \hat{\mathbf{e}}, \tag{5.29}$$

and the so-called linear parameters are given by

$$\mathbf{p} = \sin \theta_a \, \hat{\mathbf{e}}. \tag{5.30}$$

A summary of the various vectorial parameterisations, along with their respective $p(\theta_a)$ and $\kappa$, is shown in Table 5.1. A full discussion of vectorial parameterisations, including detailed derivations and analyses, can be found in Bauchau and Trainelli (2003) and Trainelli and Croce (2004).

The various vectorial parameterisations that have been developed in literature were usually motivated by specific problems, algorithms or equations for which they had specific advantages. For example, the Wiener-Milenković parameters have been used most extensively for the modelling of the structural dynamics of composite beams (Wang and Yu, 2017), specifically in relation to Finite Element Analysis (FEA) and wind turbine dynamics (Wang et al., 2017). The advantages that the vectorial parameterisations have in these limited scenarios usually relate to some algorithmic or state space simplification that they allow. The benefit is of mathematical and/or computational nature and is generally limited to only those specialised cases.

Table 5.2: A complete list of the possible Euler angles axis conventions, separated into Tait-Bryan angles and proper Euler angles. A convention of YZX, for example, means that the first elemental rotation is about the y-axis, followed by the z-axis, and then the x-axis. Whether the axes from the local (rotating) or global (fixed) frame are used depends on a further choice of whether the Euler angles are intrinsic or extrinsic, respectively.

| Type | Order of axis rotations |
|---|---|
| Tait-Bryan angles | XYZ, XZY, YXZ, YZX, ZXY, ZYX |
| Proper Euler angles | XYX, XZX, YXY, YZY, ZXZ, ZYZ |

### 5.3.5 Euler Angles

Instead of representing rotations as a single turn about a single axis, like for the axis-angle representation, it is also possible to extract more meaning by expressing rotations as a sequence of three rotations about three well-defined axes. Euler angles express a rotation as such a sequence of three elemental rotations, and rotate a coordinate frame about a predefined set of *coordinate axes* in a predefined order. The elemental rotations are either by convention extrinsic about the fixed global x, y and z-axes, or intrinsic about the local x, y and z-axes of the coordinate frame being rotated. For example, if the first elemental rotation has already been applied to the global frame {G} to get the intermediate frame {A}, then the next elemental rotation about, for instance, the y-axis, would be about $\mathbf{y}_G$ for extrinsic Euler angles, or $\mathbf{y}_A$ for intrinsic Euler angles (which is the more common case). Independently of whether the extrinsic or intrinsic convention is chosen, there are six possible choices for the predefined order of coordinate axis rotations where each axis is used only once (e.g. YZX), and a further six possible conventions where the first and third axes of rotation are the same (e.g. ZXZ). The former are commonly referred to as Tait-Bryan angles, and the latter are referred to as proper Euler angles. All possible Euler angles axis conventions are summarised in Table 5.2.

It is easy to see that all extrinsic Euler angles conventions are completely equivalent to the corresponding intrinsic Euler angles conventions, just with the order of axis rotations reversed. Whether the rotations are local or global just reverses the order in which the rotations are 'stacked'. Thus, without loss of generality, we only consider the intrinsic convention from here on in. We wish to evaluate Euler angles with respect to the aims set out in Section 5.2, and compare them to the new parameterisations that are developed in this thesis. As such, it is furthermore desirable to only consider axis conventions that have their three elemental rotations about all three different axes, i.e. Tait-Bryan angles, so that the amount of rotation within each of the three major planes can be quantified.

One of the aims that was specified in Section 5.2.1 was the property of yaw additivity. In the context of Euler angles, this means that if a global z-rotation is applied to a frame, then only the Euler angle corresponding to the z-axis should change, and up to angle wrapping, it should change by the exact amount of the z-rotation. This is clearly the case if and only if the axis convention places the z-rotation first. Thus, only two viable Euler angles conventions remain as candidates for analysis, the intrinsic ZYX and intrinsic ZXY Euler angles conventions. For completeness, both of these Euler angles conventions are presented in the next two sections, but unless explicitly otherwise stated, all further references to 'Euler angles' will be referring to intrinsic ZYX Euler angles. All arguments and properties that apply to the ZYX representation can however equivalently be reformulated to apply to the ZXY representation, so the choice is arbitrary.

The z, y and x-components of the Euler axis representation collectively correspond to the concepts of yaw, pitch and roll, respectively. Although initially promising, Euler angles do not suffice for the representation of the orientation of a body in balance-related scenarios. The main reasons for this are:

- The proximity of the gimbal lock singularities to normal working ranges, leading to unwanted artefacts due to the increased local parameter sensitivities in widened neighbourhoods of the singularities,

- The mutual dependence of the parameters, leading to a mixed attribution of which parameters contribute to which major planes of rotation,

- The fundamental requirement of an order of elemental rotations, leading to non-axisymmetric definitions of pitch and roll that do not correspond to each other in behaviour, and

- The asymmetry introduced by the use of a yaw definition that depends on the projection of one of the coordinate axes onto a fixed plane, leading to unintuitive non-axisymmetric behaviour of the yaw angle.

More details on all of these points are provided in Chapter 6.

**Intrinsic ZYX Euler**

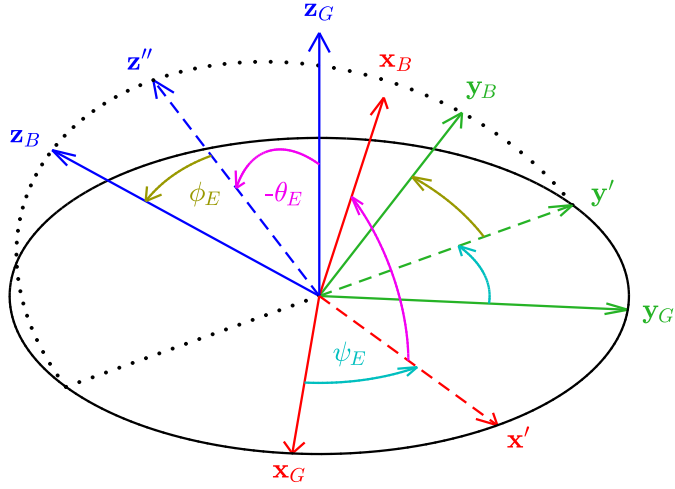

Figure 5.3: Definition of the intrinsic ZYX Euler angles convention for a rotation from {G} to {B}. The global frame {G} is first rotated CCW by the Euler yaw $\psi_E$ about $\mathbf{z}_G$, then by the Euler pitch $\theta_E$ about $\mathbf{y}'$, and finally by the Euler roll $\phi_E$ about $\mathbf{x}'' \equiv \mathbf{x}_B$. The pitch rotation is labelled $-\theta_E$ because geometrically the labelled arc in this example is CW, and not CCW.

### 5.3.5.1 *Intrinsic ZYX Euler Angles*

As usual, let {G} denote the global reference frame, and let {B} be the body-fixed frame of which the orientation is being expressed. The intrinsic ZYX Euler angles representation consists of the following sequence of three rotations, as shown in Figure 5.3:

- First a rotation by the Euler yaw $\psi_E$ about the original z-axis, which corresponds to $\mathbf{z}_G$,

- Then a rotation by the Euler pitch $\theta_E$ about the resulting intermediate y-axis, and

- Finally a rotation by the Euler roll $\phi_E$ about the once again resulting x-axis, which corresponds to $\mathbf{x}_B$.

The complete Euler angles rotation from {G} to {B} is then denoted by

$$\begin{aligned}
{}^G_B E &= (\psi_E, \theta_E, \phi_E) \\
&\in (-\pi, \pi] \times [-\tfrac{\pi}{2}, \tfrac{\pi}{2}] \times (-\pi, \pi] \equiv \mathbb{E}.
\end{aligned} \tag{5.31}$$

The representation is unique, except at the gimbal lock singularities, which correspond to $\theta_E = \pm\frac{\pi}{2}$, where for $\epsilon \in \mathbb{R}$ we have that

$$(\psi_E, \tfrac{\pi}{2}, \phi_E) \equiv (\psi_E - \epsilon, \tfrac{\pi}{2}, \phi_E - \epsilon), \tag{5.32a}$$

$$(\psi_E, -\tfrac{\pi}{2}, \phi_E) \equiv (\psi_E - \epsilon, -\tfrac{\pi}{2}, \phi_E + \epsilon). \tag{5.32b}$$

Beyond wrapping of the individual angles to the standard range of $(-\pi, \pi]$, one further representational equivalence is given by

$$(\psi_E, \theta_E, \phi_E) \equiv (\pi + \psi_E, \; \pi - \theta_E, \; \pi + \phi_E). \tag{5.33}$$

The rotation matrix $R$ corresponding to the Euler angles rotation $E = (\psi_E, \theta_E, \phi_E)$ is given by

$$R = R_z(\psi_E) R_y(\theta_E) R_x(\phi_E) \tag{5.34a}$$

$$= \begin{bmatrix} c_{\psi_E} c_{\theta_E} & c_{\psi_E} s_{\theta_E} s_{\phi_E} - s_{\psi_E} c_{\phi_E} & c_{\psi_E} s_{\theta_E} c_{\phi_E} + s_{\psi_E} s_{\phi_E} \\ s_{\psi_E} c_{\theta_E} & s_{\psi_E} s_{\theta_E} s_{\phi_E} + c_{\psi_E} c_{\phi_E} & s_{\psi_E} s_{\theta_E} c_{\phi_E} - c_{\psi_E} s_{\phi_E} \\ -s_{\theta_E} & c_{\theta_E} s_{\phi_E} & c_{\theta_E} c_{\phi_E} \end{bmatrix}, \tag{5.34b}$$

where $R_y(\cdot)$ for example is the rotation matrix corresponding to a CCW rotation by angle $\cdot$ about the y-axis, and we recall that $s_* \equiv \sin(*)$ and $c_* \equiv \cos(*)$. The reverse conversion from a rotation matrix $R$ back to the Euler angles representation $E$ is given by

$$\psi_E = \mathrm{atan2}(R_{21}, R_{11}), \tag{5.35a}$$

$$\theta_E = \mathrm{asin}(-R_{31}), \tag{5.35b}$$

$$\phi_E = \mathrm{atan2}(R_{32}, R_{33}), \tag{5.35c}$$

where $R_{ij}$ is the $i^{\text{th}}$-row $j^{\text{th}}$-column entry of the rotation matrix $R$. The conversion from Euler angles to the quaternion representation $q = (w, x, y, z)$ is given by

$$q = q_z(\psi_E) q_y(\theta_E) q_x(\phi_E) \tag{5.36a}$$

$$= (c_{\bar\psi_E} c_{\bar\theta_E} c_{\bar\phi_E} + s_{\bar\psi_E} s_{\bar\theta_E} s_{\bar\phi_E}, \; c_{\bar\psi_E} c_{\bar\theta_E} s_{\bar\phi_E} - s_{\bar\psi_E} s_{\bar\theta_E} c_{\bar\phi_E},$$
$$c_{\bar\psi_E} s_{\bar\theta_E} c_{\bar\phi_E} + s_{\bar\psi_E} c_{\bar\theta_E} s_{\bar\phi_E}, \; s_{\bar\psi_E} c_{\bar\theta_E} c_{\bar\phi_E} - c_{\bar\psi_E} s_{\bar\theta_E} s_{\bar\phi_E}), \tag{5.36b}$$

where $q_z(\cdot)$ for example is the quaternion corresponding to a CCW rotation by angle $\cdot$ about the z-axis, and we recall that $\bar* \equiv \frac{1}{2}*$. The conversion from $q = (w, x, y, z)$ back to $E = (\psi_E, \theta_E, \phi_E)$ is given by

$$\psi_E = \mathrm{atan2}(wz + xy, \tfrac{1}{2} - y^2 - z^2), \tag{5.37a}$$

$$\theta_E = \mathrm{asin}(2(wy - xz)), \tag{5.37b}$$

$$\phi_E = \mathrm{atan2}(wx + yz, \tfrac{1}{2} - x^2 - y^2). \tag{5.37c}$$

The conversion equations presented here are required later for analysis of the Euler angles representation, in particular for Chapter 6.

### 5.3.5.2  *Intrinsic ZXY Euler Angles*

The intrinsic ZXY Euler angles representation consists of the following sequence of three rotations, as shown in Figure 5.4:

- First a rotation by the ZXY Euler yaw $\psi_{\tilde E}$ about the original z-axis, which corresponds to $\mathbf{z}_G$,

### Intrinsic ZXY Euler

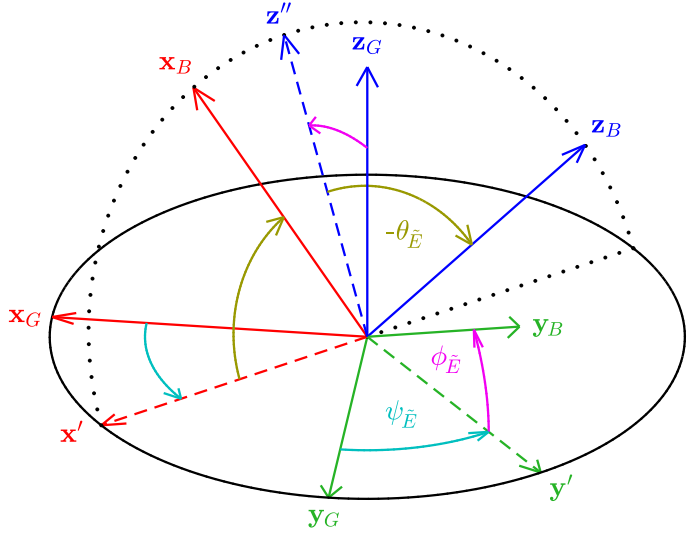

Figure 5.4: Definition of the intrinsic ZXY Euler angles convention for a rotation from {G} to {B}. The global frame {G} is first rotated CCW by the ZXY Euler yaw $\psi_{\tilde{E}}$ about $\mathbf{z}_G$, then by the ZXY Euler roll $\phi_{\tilde{E}}$ about $\mathbf{x}'$, and finally by the ZXY Euler pitch $\theta_{\tilde{E}}$ about $\mathbf{y}'' \equiv \mathbf{y}_B$. The pitch rotation is labelled $-\theta_{\tilde{E}}$ because geometrically the labelled arc in this example is CW, and not CCW.

- Then a rotation by the ZXY Euler roll $\phi_{\tilde{E}}$ about the resulting intermediate x-axis, and

- Finally a rotation by the ZXY Euler pitch $\theta_{\tilde{E}}$ about the once again resulting y-axis, which corresponds to $\mathbf{y}_B$.

The complete ZXY Euler angles rotation is then denoted by

$$
\begin{aligned}
{}^G_B\tilde{E} &= (\psi_{\tilde{E}}, \phi_{\tilde{E}}, \theta_{\tilde{E}}) \\
&\in (-\pi, \pi] \times [-\tfrac{\pi}{2}, \tfrac{\pi}{2}] \times (-\pi, \pi] \equiv \tilde{\mathbb{E}}.
\end{aligned}
\tag{5.38}
$$

The representation also suffers from two gimbal lock singularities, at $\phi_{\tilde{E}} = \pm\tfrac{\pi}{2}$, and in addition to plain angle wrapping by $2\pi$, the following representational equivalences hold, for $\epsilon \in \mathbb{R}$:

$$
(\psi_{\tilde{E}}, \tfrac{\pi}{2}, \theta_{\tilde{E}}) \equiv (\psi_{\tilde{E}} - \epsilon, \tfrac{\pi}{2}, \theta_{\tilde{E}} + \epsilon),
\tag{5.39a}
$$

$$
(\psi_{\tilde{E}}, -\tfrac{\pi}{2}, \theta_{\tilde{E}}) \equiv (\psi_{\tilde{E}} - \epsilon, -\tfrac{\pi}{2}, \theta_{\tilde{E}} - \epsilon),
\tag{5.39b}
$$

$$
(\psi_{\tilde{E}}, \phi_{\tilde{E}}, \theta_{\tilde{E}}) \equiv (\pi + \psi_{\tilde{E}}, \pi - \phi_{\tilde{E}}, \pi + \theta_{\tilde{E}}).
\tag{5.39c}
$$

The rotation matrix $R$ corresponding to the ZXY Euler angles rotation $\tilde{E} = (\psi_{\tilde{E}}, \phi_{\tilde{E}}, \theta_{\tilde{E}})$ is given by

$$
R = R_z(\psi_{\tilde{E}}) R_x(\phi_{\tilde{E}}) R_y(\theta_{\tilde{E}})
\tag{5.40a}
$$

$$
= \begin{bmatrix}
c_{\psi_{\tilde{E}}}c_{\theta_{\tilde{E}}} - s_{\psi_{\tilde{E}}}s_{\phi_{\tilde{E}}}s_{\theta_{\tilde{E}}} & -s_{\psi_{\tilde{E}}}c_{\phi_{\tilde{E}}} & c_{\psi_{\tilde{E}}}s_{\theta_{\tilde{E}}} + s_{\psi_{\tilde{E}}}s_{\phi_{\tilde{E}}}c_{\theta_{\tilde{E}}} \\
s_{\psi_{\tilde{E}}}c_{\theta_{\tilde{E}}} + c_{\psi_{\tilde{E}}}s_{\phi_{\tilde{E}}}s_{\theta_{\tilde{E}}} & c_{\psi_{\tilde{E}}}c_{\phi_{\tilde{E}}} & s_{\psi_{\tilde{E}}}s_{\theta_{\tilde{E}}} - c_{\psi_{\tilde{E}}}s_{\phi_{\tilde{E}}}c_{\theta_{\tilde{E}}} \\
-c_{\phi_{\tilde{E}}}s_{\theta_{\tilde{E}}} & s_{\phi_{\tilde{E}}} & c_{\phi_{\tilde{E}}}c_{\theta_{\tilde{E}}}
\end{bmatrix}.
\tag{5.40b}
$$

The conversion from $R$ back to $\tilde{E}$ is given by

$$\psi_{\tilde{E}} = \text{atan2}(-R_{12}, R_{22}), \tag{5.41a}$$

$$\phi_{\tilde{E}} = \text{asin}(R_{32}), \tag{5.41b}$$

$$\theta_{\tilde{E}} = \text{atan2}(-R_{31}, R_{33}). \tag{5.41c}$$

A ZXY Euler angles rotation also has the corresponding quaternion formulation

$$q = q_z(\psi_{\tilde{E}}) \, q_x(\phi_{\tilde{E}}) \, q_y(\theta_{\tilde{E}}) \tag{5.42a}$$

$$= \big( c_{\check{\psi}_{\tilde{E}}} c_{\check{\phi}_{\tilde{E}}} c_{\check{\theta}_{\tilde{E}}} - s_{\check{\psi}_{\tilde{E}}} s_{\check{\phi}_{\tilde{E}}} s_{\check{\theta}_{\tilde{E}}}, \; c_{\check{\psi}_{\tilde{E}}} s_{\check{\phi}_{\tilde{E}}} c_{\check{\theta}_{\tilde{E}}} - s_{\check{\psi}_{\tilde{E}}} c_{\check{\phi}_{\tilde{E}}} s_{\check{\theta}_{\tilde{E}}},$$
$$c_{\check{\psi}_{\tilde{E}}} c_{\check{\phi}_{\tilde{E}}} s_{\check{\theta}_{\tilde{E}}} + s_{\check{\psi}_{\tilde{E}}} s_{\check{\phi}_{\tilde{E}}} c_{\check{\theta}_{\tilde{E}}}, \; s_{\check{\psi}_{\tilde{E}}} c_{\check{\phi}_{\tilde{E}}} c_{\check{\theta}_{\tilde{E}}} + c_{\check{\psi}_{\tilde{E}}} s_{\check{\phi}_{\tilde{E}}} s_{\check{\theta}_{\tilde{E}}} \big). \tag{5.42b}$$

Finally, the conversion from $q = (w, x, y, z)$ back to $\tilde{E}$ is given by

$$\psi_{\tilde{E}} = \text{atan2}(wz - xy, \tfrac{1}{2} - x^2 - z^2), \tag{5.43a}$$

$$\phi_{\tilde{E}} = \text{asin}(2(wx + yz)), \tag{5.43b}$$

$$\theta_{\tilde{E}} = \text{atan2}(wy - xz, \tfrac{1}{2} - x^2 - y^2). \tag{5.43c}$$

### 5.3.5.3  *Interpretation of Euler Yaw as Axis Projection*

Although the ZYX and ZXY Euler yaws are each nominally defined as the first of three elemental rotations, an alternative interpretation of their definition is possible via axis projection. Taking the ZYX Euler yaw as an example, after the yaw z-rotation has been applied there is a further y-rotation of up to 90°, followed by an x-rotation. As the x-axis does not change for a local x-rotation, we deduce that the final body-fixed x-axis $\mathbf{x}_B$ is separated from the x-axis that results after the initial yaw rotation by a rotation of up to 90° about an axis in the **xy** plane. This effectively implies that the latter is the projection of the former onto the **xy** plane. Thus, given any general rotation from frame {G} to {B}, the ZYX Euler yaw is the CCW angle from $\mathbf{x}_G$ to the projection of $\mathbf{x}_B$ onto the $\mathbf{x}_G \mathbf{y}_G$ plane. This definition only breaks down if the projection of $\mathbf{x}_B$ results in the zero vector, which can be seen to be exactly the case if and only if the rotation is at gimbal lock, in which case the ZYX Euler yaw is not uniquely defined by any other definition either.

Naturally, an analogous projection-based characterisation also holds for the ZXY Euler yaw. Given any general rotation from frame {G} to {B}, the ZXY Euler yaw is the CCW angle from $\mathbf{y}_G$ to the projection of $\mathbf{y}_B$ onto the $\mathbf{x}_G \mathbf{y}_G$ plane.

## 5.4   PARTITIONING ROTATIONS INTO YAW AND TILT

As thoroughly explained in Section 5.2, we wish to find a way of partitioning any 3D rotation into separate and independent *yaw* and

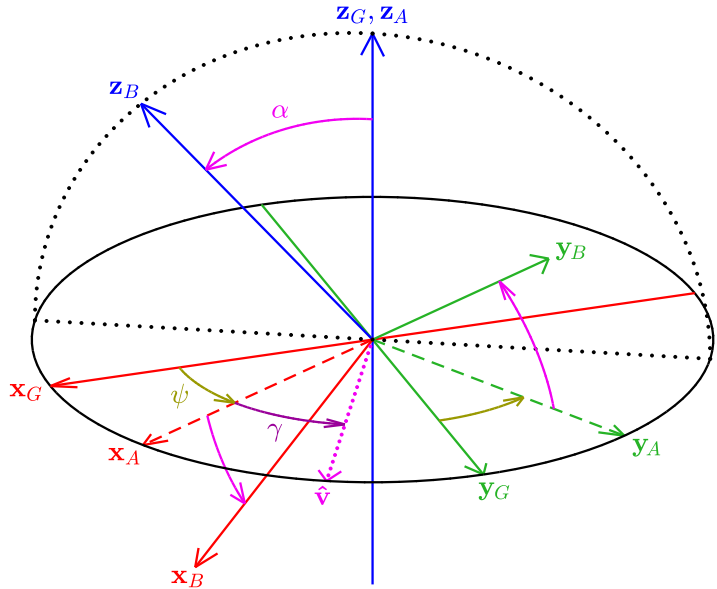

Figure 5.5: Definition of the tilt angles parameters $T = (\psi, \gamma, \alpha)$ for a rotation from {G} to {B}. The frame {G} is first rotated CCW about $\mathbf{z}_G$ by the fused yaw $\psi$ to give the intermediate frame {A}, and this frame is then tilted CCW about $\hat{\mathbf{v}}$ by the tilt angle $\alpha$ to give {B}. The tilt axis $\hat{\mathbf{v}}$ is always in the global $\mathbf{x}_G \mathbf{y}_G$ plane, and is at an angle of $\gamma$ relative to $\mathbf{x}_A$, as indicated. For aid of visualisation, note that $\mathbf{x}_B$ is pointing left-downwards out of the page, and $\mathbf{y}_B$ is pointing up-rightwards out of the page.

*tilt* components, which together uniquely define the original rotation. We wish for this partition to be intuitive and meaningful, and further wish to parameterise the tilt component by equally meaningful angular notions of *pitch* and *roll*. This section addresses how to do exactly this, and formally introduces the *tilt angles*, *fused angles* and *tilt phase space* rotation parameterisations.

### 5.4.1 Fused Yaw

We begin by defining our notion of yaw, the fused yaw of a rotation. Suppose we have the global frame {G}, and any body-fixed frame {B}, as illustrated in Figure 5.5. The rotation that we are seeking to parameterise is the one from {G} to {B}. We first define the intermediate frame {A} by rotating {B} in such a way that $\mathbf{z}_B$ rotates onto $\mathbf{z}_G$ in the most direct way possible, within the plane through the origin that contains these two vectors. Note that this rotation of {B} to get {A} is in the opposite direction to the magenta arrows in Figure 5.5 (see in particular the arrow labelled '$\alpha$'). The fused yaw $\psi \in (-\pi, \pi]$ is then given by the signed angle of the pure CCW $z$-rotation from {G} to {A}, as labelled in the figure. If $\mathbf{z}_B$ and $\mathbf{z}_G$ point in exactly opposite directions, then {A} is not uniquely defined, and this situation is referred to as the fused yaw singularity. This corresponds to all situations where the rotation from {B} to {A} has a magnitude of $\pi$ radians, i.e. 180°.

Mathematically, one can deduce that the fused yaw is given, in terms of the basic coordinate axis components of {G} and {B}, by

$$\psi = \text{wrap}\big(2\,\text{atan2}({}^{G}x_{By} - {}^{G}y_{Bx}, 1 + {}^{G}x_{Bx} + {}^{G}y_{By} + {}^{G}z_{Bz})\big), \quad (5.44)$$

where wrap($\cdot$) is a function that wraps an angle to the range $(-\pi, \pi]$ by multiples of $2\pi$. Note however that this equation is not numerically robust, and in fact fails for rotations that have an angle of exactly $\theta_a = \pi$, i.e. rotations by 180° about any axis, as atan2$(0,0)$ emerges on the right-hand side. Section 5.5.5 provides a detailed discussion of how to generalise Equation (5.44) into a globally robust expression for the fused yaw. As an alternative however, if ${}^{G}_{B}q = (w, x, y, z) \in \mathbb{Q}$ is the quaternion rotation from {G} to {B}, then the nominal mathematical definition of the fused yaw is given by

$$\psi = \text{wrap}\big(2\,\text{atan2}(z, w)\big). \quad (5.45)$$

This is the simplest and most fundamental definition of the fused yaw, and fails with atan2$(0,0)$ if and only if $w = z = 0$, which corresponds exactly to the fused yaw singularity described above. For convenience, at the fused yaw singularity the fused yaw $\psi$ is generally defined to be zero, as this makes definitions such as for example Equation (5.46) consistent with other characterisations.

### 5.4.2 Tilt Rotations

In reference to Figure 5.5, the fused yaw $\psi$ parameterises the first component of the rotation from {G} to {B}, namely the rotation from {G} to {A}, but the rotation component from {A} to {B} still remains. This component is referred to as the tilt rotation component of {B} relative to {G}, and it is easy to see that this tilt component itself has a fused yaw of zero. In fact, in general all rotations with zero fused yaw are referred to exclusively as tilt rotations, or pure tilt rotations, and the set of all tilt rotations is denoted by

$$\text{TR}(3) = \{R \in \text{SO}(3) : \psi = 0\} \subset \text{SO}(3). \quad (5.46)$$

While the fused yaw of a rotation relates to the change in heading that it induces, the tilt rotation component of a rotation relates to how much, and in what direction, the originally upright z-axis 'tips over' towards the horizontal ground plane. Three different examples of tilt rotations, as applied to an upright standing robot, are shown in Figure 5.6 for illustrative purposes of this concept.

The following are completely equivalent characterisations of the set of all (pure) tilt rotations:

### Tilt Rotations

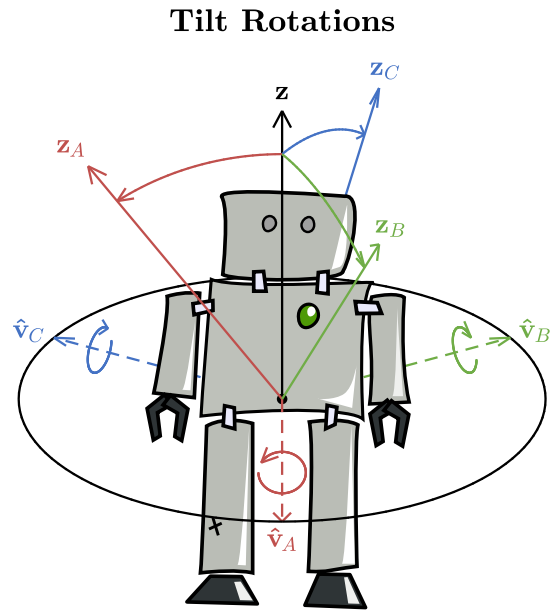

Figure 5.6: Three different examples of tilt rotations applied to an upright standing robot. The corresponding axes of rotation in the horizontal **xy** plane, $\hat{\mathbf{v}}_*$, are shown in dashed. Note that the direction of rotation is always CCW.

- All rotations that have a fused yaw of zero,

- All rotations that correspond to a single rotation about a vector in the horizontal **xy** plane,

- All rotations for which the axis of rotation has no component in the z-direction, i.e. $e_z = 0$, and,

- All rotations that have a quaternion z-component of zero.

Most of these characterisations follow naturally and easily from the definition of fused yaw in Section 5.4.1. The last one can be observed from Equation (5.45), by substituting in $z = 0$ and considering the cases separately where $w$ is positive or negative.

#### 5.4.2.1  *Z-vector Parameterisation*

Given the above definition and characterisations of tilt rotations, we now wish to be able to parameterise the two-dimensional tilt rotation component of a rotation, i.e. the component from {A} to {B}, just like the parameter $\psi$ was used to parameterise the one-dimensional yaw rotation component from {G} to {A}. The space of all tilt rotations is, as stated, two-dimensional, so two or more parameters are required for the task, or exactly two in order to be minimal. While three novel such parameterisations are introduced in this chapter, two of which specifically aim to do this in a way that reflects the angular amounts of rotation independently in the sagittal and lateral planes, we can already preliminarily parameterise tilt rotations using existing tools that we already have, namely rotation matrices and quaternions.

By definition, the tilt rotation component relates to the relative directions of the two z-axes $\mathbf{z}_G$ and $\mathbf{z}_B$, and the nature of the direct planar 3D rotation between them (see Figure 5.5). Thus, it can be observed that two immediate options for parameterising tilt rotations are just these z-axis vectors themselves, namely the global z-vector $^B\mathbf{z}_G$, and the local z-vector $^G\mathbf{z}_B$. In terms of the basic coordinate axis components of {G} and {B}, these can trivially be identified as

$$^B\mathbf{z}_G = (^Bz_{Gx}, {}^Bz_{Gy}, {}^Bz_{Gz}), \tag{5.47a}$$

$$^G\mathbf{z}_B = (^Gz_{Bx}, {}^Gz_{By}, {}^Gz_{Bz}). \tag{5.47b}$$

We note that these are unit vectors, and thus use three interdependent parameters to express the required two degrees of freedom, and that they correspond exactly to the last row and column of the rotation matrix $^G_B R$, respectively.

The local z-vector $^G\mathbf{z}_B$ is a valid parameterisation for tilt rotation components, but not in general a useful one because its x and y-components depend on the fused yaw $\psi$. This means that if the heading of a body {B} is changed, the fused yaw changes additively as required, but the local z-vector $^G\mathbf{z}_B$ also changes despite the fact that the tilt rotation component itself has not changed. The global z-vector $^B\mathbf{z}_G$ does not have this problem however, as

$$^B\mathbf{z}_G \equiv {}^B\mathbf{z}_A, \tag{5.48}$$

and therefore remains constant as expected. Thus, by convention, when referring to the z-vector parameterisation of a rotation, or just the z-vector of a rotation, this is unambiguously taken to mean the global one, namely $^B\mathbf{z}_G$. The z-vector parameterisation is meaningful and relevant in application scenarios mainly because it corresponds to the direction that an accelerometer attached to the body would measure under quasi-static conditions, as indicated in Section 5.2.2.

Given any unit global z-vector $^B\mathbf{z}_G$, the corresponding tilt rotation is uniquely defined, as the required intermediate frame {A} can be retrieved by rotating frame {B} directly onto $\mathbf{z}_G$, as per the definition in Section 5.4.1. Conversely, given a tilt rotation component, $^B\mathbf{z}_G$ is clearly uniquely defined. As such, the z-vector parameterisation is, as required, a one-to-one mapping of the space of tilt rotations, at least everywhere except for the fused yaw singularity, where $\mathbf{z}_G$ and $\mathbf{z}_B$ point in opposite directions. All tilt rotations for which this is the case are equally assigned $^B\mathbf{z}_G = (0, 0, -1)$ in terms of the z-vector parameterisation.

### 5.4.2.2  *Quaternion Parameterisation*

Being just a special type of 3D rotation, tilt rotation components can also be parameterised by their corresponding quaternion rotation from {A} to {B}, namely

$$_B^A q \equiv (w, x, y, 0), \tag{5.49}$$

where $_B^A q$ must be a unit quaternion, so

$$w^2 + x^2 + y^2 = 1. \tag{5.50}$$

Note that the above $w$, $x$ and $y$ parameters are not the same as for the quaternion representation $_B^G q$ of the *entire* rotation from {G} to {B}. Also, it should be noted that the quaternion $z$-component of tilt rotations is *always* zero, hence why the $z$-component can be omitted in Equation (5.49). Like for the $z$-vector parameterisation, the quaternion parameterisation of the tilt rotation component once again corresponds to three interdependent parameters that together express the required two degrees of freedom of the space of tilt rotations. A difference to the $z$-vector parameterisation however, is that $_B^A q$ and $-_B^A q$ are treated as equivalent parameter values. The various different tilt rotations at the fused yaw singularity also correspond to different values of the $w$, $x$ and $y$ parameters, as opposed to all corresponding to the same value, like for the $z$-vector parameterisation. This fundamental difference is discussed further in Section 5.6, also in relation to other representations and how they behave at the fused yaw singularity.

### 5.4.3  **Tilt Angles Representation**

It can be seen from the definition of fused yaw and tilt rotations that, geometrically, every rotation can be divided into a pure $z$-rotation, followed by a rotation about an axis in the horizontal **xy** plane. As shown in Figure 5.5, the former is the yaw rotation component, from {G} to {A}, and the latter is the tilt rotation component, from {A} to {B}, as previously defined. One natural way of parameterising the tilt rotation component is using the axis-angle representation, defined in Section 5.3.2. We choose an axis-angle pair $(\hat{\mathbf{v}}, \alpha) \in \mathbb{A}$ such that $\alpha \in [0, \pi]$, thus representing the tilt rotation component as a single CCW rotation by up to 180° about a vector $\hat{\mathbf{v}}$ in the horizontal **xy** plane. The angle $\alpha$ is the magnitude of the tilt rotation component, and is referred to as the tilt angle of the rotation from {G} to {B}, and the vector $\hat{\mathbf{v}}$ is referred to as the tilt axis. The CCW angle from $\mathbf{x}_A$ to $\hat{\mathbf{v}}$ about the global $z$-axis $\mathbf{z}_G$ is referred to as the tilt axis angle of {B}, and is denoted $\gamma$. Both $\mathbf{x}_A$ and $\hat{\mathbf{v}}$ are in the global **xy** plane, so $\mathbf{z}_G$ is perpendicular to them, and $\gamma$ is well-defined. The tilt angle $\alpha$, tilt axis $\hat{\mathbf{v}}$, and tilt axis angle $\gamma$ are all labelled in Figure 5.5.

It is easy to see that the tilt rotation component from {A} to {B} is completely defined by the parameter pair $(\gamma, \alpha)$. Thus, together with the fused yaw $\psi$, we arrive at the tilt angles parameterisation of the rotation from {G} to {B}, namely

$$
\begin{aligned}
{}^{G}_{B}T &= (\psi, \gamma, \alpha) \\
&\in (-\pi, \pi] \times (-\pi, \pi] \times [0, \pi] \equiv \mathbb{T}.
\end{aligned}
\tag{5.51}
$$

The tilt angles parameters corresponding to the identity rotation are given by $(0, 0, 0) \in \mathbb{T}$. It can be seen by construction that all rotations possess a tilt angles representation, but it is not always necessarily unique. Most notably, when $\alpha = 0$, the $\gamma$ parameter can be arbitrary with no effect. Mathematically, the tilt axis angle $\gamma$ and tilt angle $\alpha$ can be expressed in terms of the basic coordinate axis components of {G}, or more specifically, the components of the z-vector ${}^{B}\mathbf{z}_{G}$, as

$$
\gamma = \text{atan2}(-{}^{B}z_{Gx}, {}^{B}z_{Gy}), \tag{5.52a}
$$

$$
\alpha = \text{acos}({}^{B}z_{Gz}). \tag{5.52b}
$$

The tilt axis $\hat{\mathbf{v}}$ can be expressed in terms of $\gamma$ and the fused yaw $\psi$, as

$$
{}^{G}\hat{\mathbf{v}} = \big(\cos(\psi + \gamma), \sin(\psi + \gamma), 0\big). \tag{5.53}
$$

Relative to the intermediate frame {A}, or for tilt rotations, this is just

$$
{}^{A}\hat{\mathbf{v}} = \big(\cos\gamma, \sin\gamma, 0\big). \tag{5.54}
$$

While tilt angles themselves do not directly address all the requirements and aims that were set out in Section 5.2, they constitute a useful intermediate representation for defining and analysing other parameterisations of the tilt rotation component, and give a somewhat direct and 'low-level' view of the partition of rotations into yaw and tilt components. By optionally extending the domain of $\alpha$ to $[0, \infty)$, tilt rotations of arbitrary magnitude can also be parameterised and numerically manipulated. This, for example, is not possible with either the z-vector or quaternion parameterisations of the tilt rotation component, and can be useful for dealing with rotations that exceed 180° in magnitude, as opposed to just considering the final coordinate frame that these so-called unbounded rotations result in.

### 5.4.4   Fused Angles Representation

Although the tilt angles parameterisation is a helpful start in understanding tilt rotations, we wish to be able to parameterise the tilt rotation component in such a way that it reveals concurrent definitions of pitch and roll, as discussed in Sections 5.2.1 and 5.2.2. To do this, we consider just the tilt rotation component of a rotation from {G} to {B}, namely the component from {A} to {B}, as shown in Figure 5.7. We define the fused pitch $\theta$ as the signed angle between the global z-vector $\mathbf{z}_{G}$ $(= \mathbf{z}_{A})$ and the $\mathbf{y}_{B}\mathbf{z}_{B}$ plane, and the fused roll $\phi$ as the signed angle between $\mathbf{z}_G$ and the $\mathbf{x}_B\mathbf{z}_B$ plane, as illustrated in the figure. The signs of $\theta$ and $\phi$ are defined to be the same as the signs of $-{}^B z_{Gx}$ and ${}^B z_{Gy}$, respectively. In fact, the fused pitch and roll are exactly given by

$$\theta = \operatorname{asin}(-{}^B z_{Gx}), \tag{5.55a}$$

$$\phi = \operatorname{asin}({}^B z_{Gy}). \tag{5.55b}$$

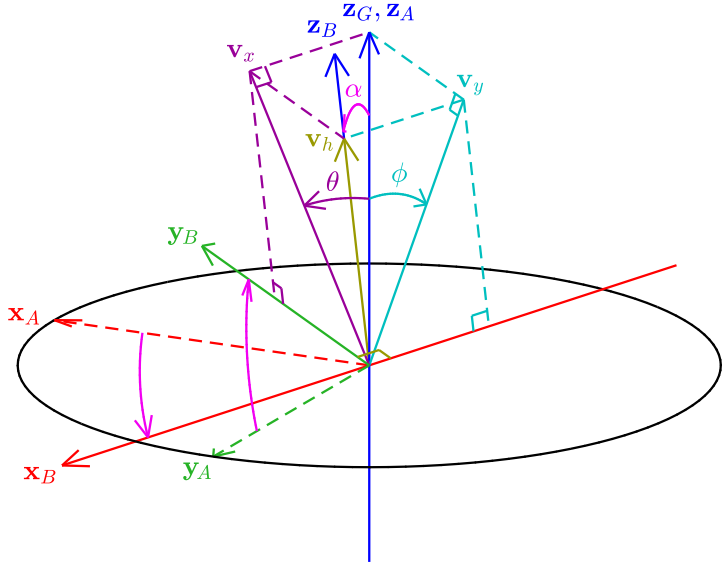

Figure 5.7: Definition of the fused angles parameters $(\theta, \phi, h)$ for a tilt rotation component from {A} to {B}. Refer to Figure 5.5 for a visual definition of the intermediate frame {A}, and the fused yaw $\psi$ used to define it. The frame {A} is rotated onto frame {B} through a tilt angle of $\alpha$ in such a way that $\mathbf{z}_A \equiv \mathbf{z}_G$ rotates directly onto $\mathbf{z}_B$. The fused pitch $\theta$ and fused roll $\phi$ are defined as the angles between $\mathbf{z}_G$ and the $\mathbf{y}_B\mathbf{z}_B$ and $\mathbf{x}_B\mathbf{z}_B$ planes respectively. The hemisphere $h$ is $+1$ if $\mathbf{z}_G$ and $\mathbf{z}_B$ are mutually in the same hemisphere, and $-1$ if not. For aid of visualisation, note that $\mathbf{x}_B$ is pointing left-downwards parallel to the page, and $\mathbf{y}_B$ is pointing up-leftwards out of the page.

Loosely speaking, rotations with positive fused pitch can be thought of as tendentially 'forwards' rotations that have a positive CCW component about the y-axis, and positive fused roll rotations can be thought of as tendentially 'rightwards' rotations that have a positive CCW component about the x-axis.

By inspection of their mathematical and geometric definitions, it can be seen that the fused pitch and roll only uniquely specify the tilt rotation component of a rotation up to the z-hemisphere, that is, whether $\mathbf{z}_B$ and $\mathbf{z}_G$ are mutually in the same unit hemisphere or not. To resolve this ambiguity, the fused hemisphere, or just hemisphere, of a rotation is defined as

$$h = \operatorname{sign}({}^B z_{Gz}), \tag{5.56}$$

where $\text{sign}(\cdot)$ is a sign function that takes on only the values $\pm 1$. Note that $\text{sign}(\cdot)$ differs from the normal sign function $\text{sgn}(\cdot)$ in that $\text{sign}(0) = 1$, whereas $\text{sgn}(0) = 0$.

Using the notion of the fused hemisphere, the triplet $(\theta, \phi, h)$ becomes a complete description of the tilt rotation component of a rotation, just like the parameter pair $(\gamma, \alpha)$ is. Thus, together with the fused yaw $\psi$, we define the fused angles parameterisation of the rotation from {G} to {B} as

$$\begin{aligned}
{}^{G}_{B}F &= (\psi, \theta, \phi, h) \\
&\in (-\pi, \pi] \times [-\tfrac{\pi}{2}, \tfrac{\pi}{2}] \times [-\tfrac{\pi}{2}, \tfrac{\pi}{2}] \times \{-1, 1\} \equiv \hat{\mathbb{F}}.
\end{aligned} \tag{5.57}$$

The fused angles parameters corresponding to the identity rotation are given by $(0, 0, 0, 1) \in \hat{\mathbb{F}}$. A hat is used for the domain $\hat{\mathbb{F}}$ in Equation (5.57), because the given definition is in fact a *superset* of the true domain $\mathbb{F}$ of the fused angles parameterisation. From Equation (5.55), it can be seen that

$$\begin{aligned}
{}^{B}z_{Gx} &= -\sin\theta, & \text{(5.58a)} \\
{}^{B}z_{Gy} &= \sin\phi. & \text{(5.58b)}
\end{aligned}$$

As ${}^{B}\mathbf{z}_{G}$ is a unit vector, we must have that

$$ {}^{B}z_{Gx}^2 + {}^{B}z_{Gy}^2 \leq 1, \tag{5.59}$$

so from Equation (5.58), we deduce that $\theta$ and $\phi$ must satisfy

$$\sin^2\theta + \sin^2\phi \leq 1. \tag{5.60}$$

This is referred to as the sine sum criterion, and given that the fused pitch and roll are both restricted to a domain of $[-\tfrac{\pi}{2}, \tfrac{\pi}{2}]$, this is completely equivalent to

$$|\theta| + |\phi| \leq \tfrac{\pi}{2}, \tag{5.61}$$

as demonstrated in Appendix B.1.1.1. The region in the fused pitch vs. fused roll space that is defined by the sine sum criterion is shown in Figure 5.8. The true domain $\mathbb{F}$ of the fused angles representation is given by the restriction of $\hat{\mathbb{F}}$ to this region. The domains of the fused yaw and hemisphere parameters remain unchanged however.

### 5.4.5   Tilt Phase Space

As an alternative to fused angles for parameterising the tilt rotation component of a rotation in a way that yields pitch and roll components, we now introduce the closely related tilt phase space parameterisation. The tilt phase space combines the ability of fused angles to quantify the amounts of rotation in the major planes, with the ability of tilt angles to optionally also represent tilt rotations over 180°. The tilt phase space also enables a more vectorial view of rotations, while still adhering to the partition of rotations into fused yaw and tilt components, as required. There are two main variants of the tilt phase space, relative and absolute, and for each case there is the choice to either consider all three parameters (3D), or only the two that quantify the tilt rotation component (2D). If referring to just 'the tilt phase space', without a qualifier, by default this refers to the relative tilt phase space, and from the context it is usually clear whether just the 2D pitch and roll parameters are being considered, or whether the 3D parameters, including the yaw, are being considered. The relative and absolute tilt phase spaces are defined as follows.

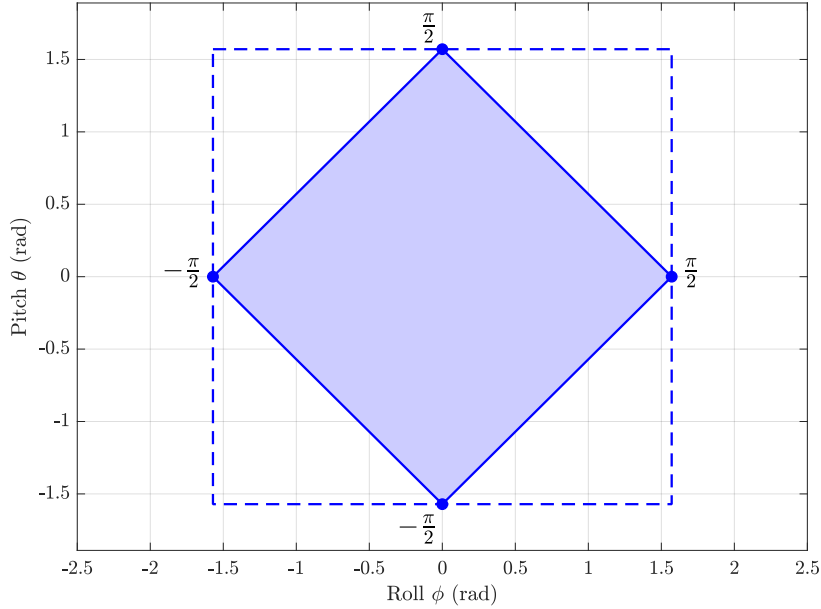

Figure 5.8: An illustration of the true domain of the fused pitch and roll. While nominally the domain is $(\phi, \theta) \in [-\frac{\pi}{2}, \frac{\pi}{2}] \times [-\frac{\pi}{2}, \frac{\pi}{2}]$, we see through the application of the sine sum criterion $|\theta| + |\phi| \leq \frac{\pi}{2}$ that the true domain is given by the shaded diamond-shaped region. The nominal square domain corresponds to $\hat{\mathbb{F}}$, and the true diamond domain corresponds to $\mathbb{F}$.

### 5.4.5.1  *Relative Tilt Phase Space*

Given the tilt angles representation ${}^G_B T = (\psi, \gamma, \alpha)$ of a rotation from {G} to {B}, the phase roll $p_x$, phase pitch $p_y$, and fused yaw $p_z$ are defined by

$$p_x = \alpha \cos \gamma, \tag{5.62a}$$

$$p_y = \alpha \sin \gamma, \tag{5.62b}$$

$$p_z = \psi. \tag{5.62c}$$

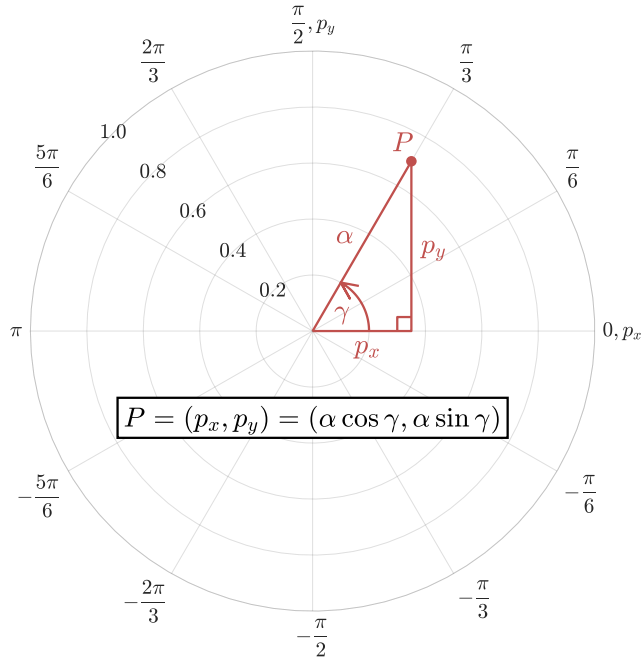

Figure 5.9: A polar plot of the 2D tilt phase space parameters $(p_x, p_y) \in \mathbb{P}^2$. The x and y-axes correspond to $p_x$ and $p_y$ respectively, and in terms of polar coordinates, $\gamma$ is the polar angle and $\alpha$ is the radius. Note that although the radial scale in this plot only goes up to 1.0, this is just an arbitrary choice, and can of course be extended to $\pi$ or more.

The 3D relative tilt phase space representation is then given by

$$^G_B P = (p_x, p_y, p_z) \in \mathbb{R}^3 \equiv \mathbb{P}^3. \tag{5.63}$$

The relative tilt phase parameters corresponding to the identity rotation are given by $(0,0,0) \in \mathbb{P}^3$. Often, when working with tilt rotations, the fused yaw component is either zero or irrelevant. In such cases, the 2D relative tilt phase space representation can be used instead, given by

$$^G_B P = (p_x, p_y) \in \mathbb{R}^2 \equiv \mathbb{P}^2, \tag{5.64}$$

which essentially just omits the fused yaw component $p_z$. The 2D formulation of the tilt phase space is the most frequently encountered one in this thesis, specifically because of its application to bipedal walking (e.g. in Chapters 14 and 15). It should be noted that $\mathbb{R}$ was used in full generality in Equations (5.63) and (5.64) for the definition of the tilt phase space domains $\mathbb{P}^2$ and $\mathbb{P}^3$, in order to naturally be able to represent rotations of all magnitudes (i.e. unbounded rotations). As is frequently the case however, if we are dealing strictly with rotations where only the final orientation of the coordinate frame matters, then the true domain of the tilt phase space would rather be

$$\mathbb{P}^3 \equiv \bar{\mathcal{D}}^2(\pi) \times (-\pi, \pi], \tag{5.65a}$$

$$\mathbb{P}^2 \equiv \bar{\mathcal{D}}^2(\pi), \tag{5.65b}$$

where $\bar{\mathcal{D}}^2(\pi)$ is the closed 2-dimensional disc of radius $\pi$ centred at the origin. That is, put more simply, in the phase pitch vs. phase roll space, $\bar{\mathcal{D}}^2(\pi)$ corresponds to a circle of radius $\pi$ and all of its interior.

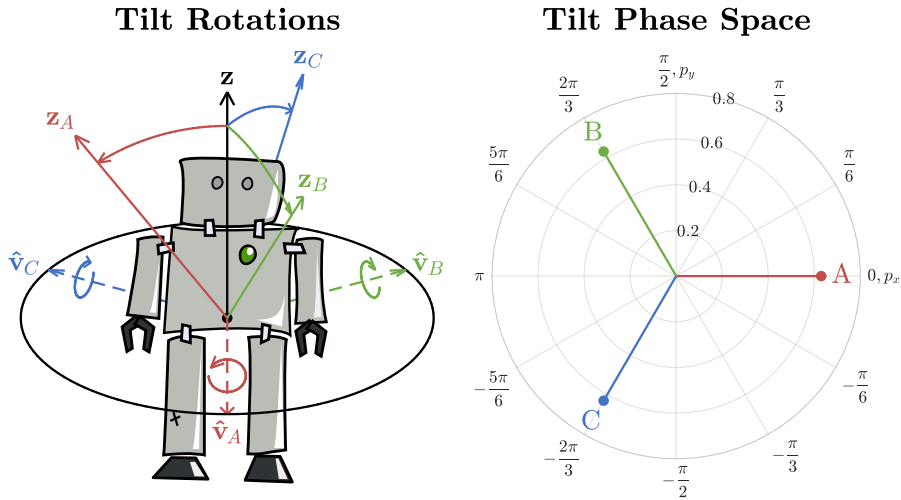

Figure 5.10: Three different examples of tilt rotations applied to an upright standing robot (left), and the corresponding points in the 2D tilt phase space (right). The respective CCW axes of rotation in the horizontal **xy** plane, $\hat{\mathbf{v}}_*$, are shown in dashed. Note, for example, that $A$ is a rotation about the positive x-axis, i.e. $\hat{\mathbf{v}}_A = (1, 0, 0)$, so its corresponding $p_y$ component is zero, and its $p_x$ component is positive (can be seen in the phase plot on the right).

Plots of phase pitch $p_y$ against phase roll $p_x$ are customarily drawn on a polar plot to highlight the relationship between the 'Cartesian' tilt phase coordinates and the 'polar' tilt angles coordinates $\alpha$ and $\gamma$, as made explicit in Figure 5.9. The circular boundary of a polar plot also well-represents the natural domain of all tilt rotations if only bounded rotations, i.e. tilt rotations of up to 180°, are considered, as this domain is circular as well, as just discussed. An example of three different pure tilt rotations, and what their respective tilt phase space coordinates are as viewed in a polar plot, is shown in Figure 5.10.

The tilt phase parameters were defined in terms of the tilt angles parameters in Equation (5.62), but as discussed in Section 5.4.3, the tilt axis angle $\gamma$ and the tilt angle $\alpha$ are not always unique in describing a particular tilt rotation component. Thus, it is important to check that the tilt phase space is nonetheless well-defined, and furthermore that it is unique, continuous and smooth (infinitely differentiable) everywhere except for at the fused yaw singularity. These are critically important properties of the tilt phase space, as they imply that the representation can for example safely be differentiated and related to angular velocities. The well-definedness, continuity and smoothness of the tilt phase space, in particular at the identity tilt rotation ${}^{G}_{B}P = (0,0) \in \mathbb{P}^2$, is demonstrated in Appendix B.1.1.2.

### 5.4.5.2    *Absolute Tilt Phase Space*

The absolute tilt phase space shares the same definition as the relative tilt phase space, only with the absolute tilt axis angle $\tilde{\gamma} = \gamma + \psi$ being used instead of $\gamma$. That is,

$$\tilde{p}_x = \alpha \cos \tilde{\gamma}, \tag{5.66a}$$

$$\tilde{p}_y = \alpha \sin \tilde{\gamma}, \tag{5.66b}$$

$$\tilde{p}_z = \psi. \tag{5.66c}$$

If a clear distinction between $\tilde{\gamma}$ and $\gamma$ is required, then $\gamma$ can be referred to as the relative tilt axis angle. The triplet

$$\begin{aligned} {}^{G}_{B}\tilde{T} &= (\psi, \tilde{\gamma}, \alpha) \\ &\in (-\pi, \pi] \times (-\pi, \pi] \times [0, \pi] \equiv \tilde{\mathbb{T}}, \end{aligned} \tag{5.67}$$

is sometimes referred to as the absolute tilt angles parameterisation of a rotation, and once again, if clear distinction is required, the standard tilt angles representation can be referred to as the relative tilt angles.

Based on Equation (5.66), the 3D absolute tilt phase space representation is given by

$$\tilde{{}^{G}_{B}\tilde{P}} = (\tilde{p}_x, \tilde{p}_y, \tilde{p}_z) \in \mathbb{R}^3 \equiv \tilde{\mathbb{P}}^3, \tag{5.68}$$

and the 2D absolute tilt phase space representation is given by

$$\tilde{{}^{G}_{B}\tilde{P}} = (\tilde{p}_x, \tilde{p}_y) \in \mathbb{R}^2 \equiv \tilde{\mathbb{P}}^2, \tag{5.69}$$

The absolute tilt phase parameters corresponding to the identity rotation are $(0,0,0) \in \tilde{\mathbb{P}}^3$, as before. Geometrically, the absolute tilt axis angle $\tilde{\gamma}$ is equivalent to the CCW angle from $\mathbf{x}_G$ to $\hat{\mathbf{v}}$ (see Figure 5.5), instead of the angle from $\mathbf{x}_A$ to $\hat{\mathbf{v}}$, as for the relative tilt axis angle $\gamma$. The relation between the absolute and relative tilt phase spaces is given by

$$\tilde{p}_x = p_x \cos \psi - p_y \sin \psi, \tag{5.70a}$$

$$\tilde{p}_y = p_x \sin \psi + p_y \cos \psi, \tag{5.70b}$$

$$\tilde{p}_z = p_z, \tag{5.70c}$$

for the conversion from relative to absolute, and

$$p_x = \phantom{-}\tilde{p}_x \cos \psi + \tilde{p}_y \sin \psi, \tag{5.71a}$$

$$p_y = -\tilde{p}_x \sin \psi + \tilde{p}_y \cos \psi, \tag{5.71b}$$

$$p_z = \tilde{p}_z, \tag{5.71c}$$

for the conversion from absolute to relative. These equations essentially correspond to simple 2D rotations by $\pm\psi$ for $(p_x, p_y)$ and $(\tilde{p}_x, \tilde{p}_y)$.

The absolute tilt phase space is only a slight variation of the relative tilt phase space, and it should be noted that for tilt rotations the two representations are identical anyway. That is,

$$p_z = \psi = 0 \implies {}^G_B\tilde{P} \equiv {}^G_BP \tag{5.72}$$

All previous arguments about bounded vs. unbounded rotations, the true bounded domain of the parameters, phase plots, continuity and smoothness and so on, hold just the same for the absolute tilt phase space. In fact, all results that hold for one of the two spaces in general correspond trivially to completely analogous results for the other.

### 5.4.5.3 *Tilt Vector Addition*

If one takes two different tilt rotations and combines them as sequential rotations, the result is in general not a tilt rotation, and depends on the order in which the two rotations are combined. In mathematical terms, this is to say that the binary operation given by standard rotation composition is not *closed* on the set of all tilt rotations, and not *commutative*. This can be problematic, if for example in an orientation feedback controller it is required to combine the contributions of multiple tilt rotation feedback paths, e.g. for PID feedback like in Section 15.2, as using standard rotation composition cannot result in an unambiguous final tilt rotation. The 2D tilt phase space provides a way of defining a useful and meaningful addition operator for tilt rotations that is closed, commutative, and like standard rotation composition, associative. This new binary operation '$\oplus$' is referred to as tilt vector addition, and for $P_1, P_2 \in \mathbb{P}^2$ is defined by

$$P_1 \oplus P_2 = (p_{x1} + p_{x2},\ p_{y1} + p_{y2}) \in \mathbb{P}^2. \tag{5.73}$$

In terms of tilt angles this is equivalent to

$$(\alpha_1 c_{\gamma_1},\ \alpha_1 s_{\gamma_1}) \oplus (\alpha_2 c_{\gamma_2},\ \alpha_2 s_{\gamma_2}) = (\alpha_1 c_{\gamma_1} + \alpha_2 c_{\gamma_2},\ \alpha_1 s_{\gamma_1} + \alpha_2 s_{\gamma_2}), \tag{5.74}$$

where the standard abbreviations $c_*$ and $s_*$ for $\cos(*)$ and $\sin(*)$ have been used. If we let $P_3 \in \mathbb{P}^2$ denote the tilt rotation corresponding to the sum $P_1 \oplus P_2$, and let $(\gamma_3, \alpha_3)$ denote the associated tilt angles parameters, then we can also write

$$(\gamma_1, \alpha_1) \oplus (\gamma_2, \alpha_2) = (\gamma_3, \alpha_3), \tag{5.75}$$

noting that this is not just elementwise vector addition, but rather shorthand for applying Equation (5.74) and then converting the result back into tilt angles. The action of tilt vector addition is illustrated in Figure 5.11.

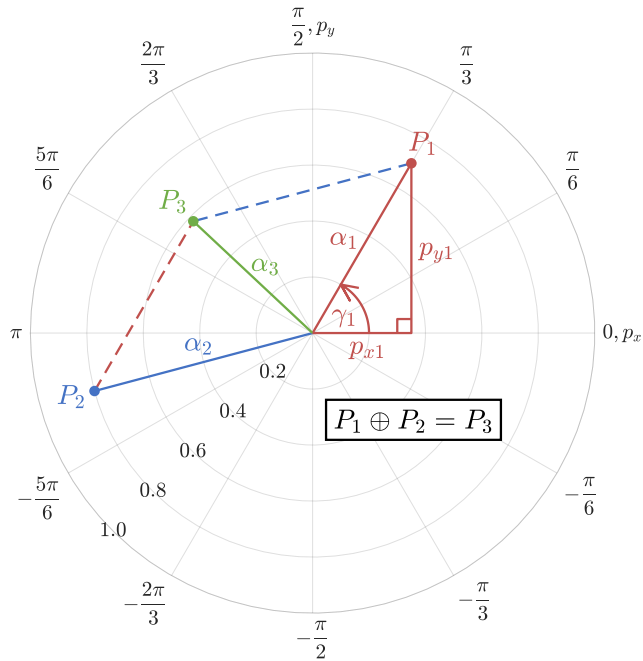

Figure 5.11: An illustration of tilt vector addition in the 2D tilt phase space, where as indicated we have that $P_1 \oplus P_2 = P_3$. When expressed in terms of the tilt phase space, it can be seen that tilt vector addition is equivalent to standard vector addition. In terms of tilt angles, we have in this case that a tilt of $0.7\,\text{rad}$ about $\gamma_1 = 60°$, added to a tilt of $0.8\,\text{rad}$ about $\gamma_2 = 195°$, gives a combined tilt of $0.5814\,\text{rad}$ about $\gamma_3 = 136.6°$.

Although tilt vector addition is nominally defined in terms of the relative tilt phase space, completely analogous definitions of tilt vector addition naturally hold for the absolute tilt phase space as well. An important observation however, which supports the self-consistency of the definition of tilt vector addition, is that regardless of whether two tilt rotations are added in the relative or absolute spaces, the outcome is the same. This means that if two tilt rotations (or tilt rotation components for rotations of equal fused yaw) are expressed in terms of their relative and absolute tilt phase space parameters and added using tilt vector addition, the resulting tilt rotations expressed as, for example, quaternions are identical.

## 5.5   ROTATION REPRESENTATION CONVERSIONS

Given the many different existing rotation representations and the new ones that have been introduced in this chapter, it is important to be able to relate them to each other and convert between them. This section summarises all relevant conversion equations, and also sheds light on the intricate connections between the various representations, as required for later parts of this thesis. Conversions from the z-vector parameterisation to other forms have been omitted, as these are implicit from the definitions that have been given in this chapter so far, e.g. like in Equations (5.52) and (5.55). Conversions to and from the tilt phase space are also in general only presented in terms of the relative tilt phase space, as Equations (5.70) and (5.71) can then be used to relate this to the absolute tilt phase space. Throughout this section, as well as throughout the remainder of this thesis, the shorthand is used that $c_* \equiv \cos(*)$, $s_* \equiv \sin(*)$ and $\bar{*} \equiv \frac{1}{2}*$.

### 5.5.1 From Tilt Angles To

Given a tilt angles rotation $T = (\psi, \gamma, \alpha) \in \mathbb{T}$, the corresponding rotation matrix is given by

$$R = R_z(\psi)R_{\hat{\mathbf{v}}}(\alpha) \tag{5.76a}$$

$$= \begin{bmatrix} c_\gamma c_{\gamma+\psi} + c_\alpha s_\gamma s_{\gamma+\psi} & s_\gamma c_{\gamma+\psi} - c_\alpha c_\gamma s_{\gamma+\psi} & s_\alpha s_{\gamma+\psi} \\ c_\gamma s_{\gamma+\psi} - c_\alpha s_\gamma c_{\gamma+\psi} & s_\gamma s_{\gamma+\psi} + c_\alpha c_\gamma c_{\gamma+\psi} & -s_\alpha c_{\gamma+\psi} \\ -s_\alpha s_\gamma & s_\alpha c_\gamma & c_\alpha \end{bmatrix}, \tag{5.76b}$$

where $R_z(\cdot)$ is a CCW rotation by angle $\cdot$ about the z-axis, and $R_{\hat{\mathbf{v}}}(\cdot)$ is a CCW rotation by angle $\cdot$ about the vector $\hat{\mathbf{v}} = (c_\gamma, s_\gamma, 0)$. The z-vector parameterisation is thus given by the bottom row

$$^B\mathbf{z}_G = (-\sin\alpha\sin\gamma,\ \sin\alpha\cos\gamma,\ \cos\alpha). \tag{5.77}$$

The quaternion corresponding to $T$, on the other hand, is given by

$$q = q_z(\psi)q_{\hat{\mathbf{v}}}(\alpha) \tag{5.78a}$$

$$= (c_{\bar\alpha}c_{\bar\psi},\ s_{\bar\alpha}c_{\bar\psi+\gamma},\ s_{\bar\alpha}s_{\bar\psi+\gamma},\ c_{\bar\alpha}s_{\bar\psi}). \tag{5.78b}$$

In specific reference to Section 5.4.2.2 and Equation (5.49), the quaternion corresponding to the tilt rotation component of a rotation is thus given by

$$^A_B q = (w, x, y, 0) \tag{5.79a}$$

$$= (\cos\tfrac{\alpha}{2},\ \sin\tfrac{\alpha}{2}\cos\gamma,\ \sin\tfrac{\alpha}{2}\sin\gamma,\ 0). \tag{5.79b}$$

The tilt phase representation of the rotation $T$ is clear by definition, i.e. from Equations (5.62) and (5.66), and aside from the fused yaw $\psi$, which is common to both representations, the fused angles $F = (\psi, \theta, \phi, h)$ can be calculated from the tilt angles parameters using

$$\sin\theta = \sin\alpha\sin\gamma, \tag{5.80a}$$

$$\sin\phi = \sin\alpha\cos\gamma. \tag{5.80b}$$

The fused hemisphere $h$ is given by

$$h = \text{sign}(\cos\alpha) \tag{5.81a}$$

$$= \begin{cases} 1 & \text{if } \alpha \le \frac{\pi}{2}, \\ -1 & \text{if } \alpha > \frac{\pi}{2}, \end{cases} \tag{5.81b}$$

where we recall that $\alpha \in [0, \pi]$. Interestingly, from Equation (5.80) we note that

$$\sin^2\theta + \sin^2\phi = \sin^2\alpha. \tag{5.82}$$

This, amongst other things, gives new insight into why the sine sum criterion in Equation (5.60) must hold.

### 5.5.2   From Fused Angles To

Given a fused angles rotation $F = (\psi, \theta, \phi, h) \in \mathbb{F}$, the tilt angles parameters can be calculated using

$$\gamma = \text{atan2}(\sin\theta, \sin\phi), \tag{5.83a}$$

$$\alpha = \text{acos}\Big(h\sqrt{1 - \sin^2\theta - \sin^2\phi}\Big), \tag{5.83b}$$

where for numerical computation one may choose to use the identity

$$1 - \sin^2\theta - \sin^2\phi \equiv \cos(\theta + \phi)\cos(\theta - \phi). \tag{5.84}$$

Most conversions involving fused angles involve the calculation of the tilt angles parameters $\gamma$ and $\alpha$, followed by the use of the appropriate tilt angles conversion formulas from Section 5.5.1. This is why the tilt angles representation is sometimes referred to as an *intermediate representation*. Despite this however, in most cases there are still slight simplifications or alternatives that can be found, either to make the calculation somewhat more direct, or to highlight mathematical links between fused angles and other representations that are useful for analysis. For instance, the rotation matrix representation of $F$ can be expressed as

$$R = \begin{bmatrix} c_\gamma c_{\gamma+\psi} + c_\alpha s_\gamma s_{\gamma+\psi} & s_\gamma c_{\gamma+\psi} - c_\alpha c_\gamma s_{\gamma+\psi} & s_\psi s_\phi + c_\psi s_\theta \\ c_\gamma s_{\gamma+\psi} - c_\alpha s_\gamma c_{\gamma+\psi} & s_\gamma s_{\gamma+\psi} + c_\alpha c_\gamma c_{\gamma+\psi} & s_\psi s_\theta - c_\psi s_\phi \\ -s_\theta & s_\phi & c_\alpha \end{bmatrix}, \tag{5.85}$$

leading to a $z$-vector parameterisation of

$${}^B\mathbf{z}_G = (-\sin\theta, \ \sin\phi, \ \cos\alpha) \tag{5.86a}$$

$$= \Big(-\sin\theta, \ \sin\phi, \ h\sqrt{1 - \sin^2\theta - \sin^2\phi}\Big). \tag{5.86b}$$

The quaternion $q$ corresponding to the fused angles $F$ can also be expressed partially in terms of the fused angles parameters using

$$q = \begin{cases} \frac{\tilde{q}_p}{\|\tilde{q}_p\|} & \text{if } h = 1, \\ \frac{\tilde{q}_n}{\|\tilde{q}_n\|} & \text{if } h = -1, \end{cases} \tag{5.87}$$

where the unnormalised positive and negative hemisphere quaternions $\tilde{q}_p$ and $\tilde{q}_n$ are given by

$$\tilde{q}_p = \big(c_{\tilde{\psi}}(1+c_\alpha),\, s_\phi c_{\tilde{\psi}} - s_\theta s_{\tilde{\psi}},\, s_\phi s_{\tilde{\psi}} + s_\theta c_{\tilde{\psi}},\, s_{\tilde{\psi}}(1+c_\alpha)\big), \tag{5.88a}$$

$$\tilde{q}_n = \big(s_\alpha c_{\tilde{\psi}},\, c_{\tilde{\psi}+\gamma}(1-c_\alpha),\, s_{\tilde{\psi}+\gamma}(1-c_\alpha),\, s_\alpha s_{\tilde{\psi}}\big). \tag{5.88b}$$

The respective quaternion norms are analytically given by

$$\|\tilde{q}_p\| = \sqrt{2(1+\cos\alpha)} = 2\cos\tfrac{\alpha}{2}, \tag{5.89a}$$

$$\|\tilde{q}_n\| = \sqrt{2(1-\cos\alpha)} = 2\sin\tfrac{\alpha}{2}. \tag{5.89b}$$

Note that both cases in Equation (5.87) can actually be used everywhere in both fused hemispheres, but the normalisation of $\tilde{q}_p$ becomes numerically sensitive towards the bottom ($\alpha = \pi$) of the negative hemisphere, and the normalisation of $\tilde{q}_n$ similarly becomes numerically sensitive towards the top ($\alpha = 0$) of the positive hemisphere. This can be seen from the quaternion norms given in Equation (5.89). It should also be noted that the value of $\alpha$ does not need to be computed in order to evaluate Equation (5.88), only $\cos\alpha$ and $\sin\alpha$, which can be obtained from Equations (5.82) and (5.83b) directly.

The conversion from fused angles to the tilt phase space is also nominally performed via calculation of $\alpha$ and $\gamma$, but the two spaces can also be more directly related using

$$p_x = \frac{\sin\phi}{\operatorname{sinc}\alpha} = \Big(\frac{\alpha}{\sin\alpha}\Big)\sin\phi, \tag{5.90a}$$

$$p_y = \frac{\sin\theta}{\operatorname{sinc}\alpha} = \Big(\frac{\alpha}{\sin\alpha}\Big)\sin\theta, \tag{5.90b}$$

where

$$\operatorname{sinc}\alpha = \begin{cases} \frac{\sin\alpha}{\alpha} & \text{if } \alpha \neq 0, \\ 1 & \text{if } \alpha = 0, \end{cases} \tag{5.91}$$

is the cardinal sine function, a smooth function of $\alpha$.

### 5.5.3 From Tilt Phase Space To

The conversions from the tilt phase space $P = (p_x, p_y, p_z) \in \mathbb{P}^3$ to all other representations mentioned in this chapter are nominally performed via the tilt angles representation, the conversion to which is given by

$$\psi = p_z, \tag{5.92a}$$

$$\gamma = \text{atan2}(p_y, p_x), \tag{5.92b}$$

$$\alpha = \sqrt{p_x^2 + p_y^2}. \tag{5.92c}$$

The only exception is the fused angles representation, which can be related to the tilt phase space directly using Equation (5.92c), followed by

$$\sin\theta = p_y \operatorname{sinc}\alpha, \tag{5.93a}$$

$$\sin\phi = p_x \operatorname{sinc}\alpha. \tag{5.93b}$$

The tilt phase space is another example of why the tilt angles representation is sometimes referred to as an *intermediate representation*.

### 5.5.4   From Quaternion To

Given the quaternion $q = (w, x, y, z) \in \mathbb{Q}$, the corresponding rotation matrix is given by

$$R = \begin{bmatrix} 1 - 2(y^2 + z^2) & 2(xy - wz) & 2(xz + wy) \\ 2(xy + wz) & 1 - 2(x^2 + z^2) & 2(yz - wx) \\ 2(xz - wy) & 2(yz + wx) & 1 - 2(x^2 + y^2) \end{bmatrix}, \tag{5.94}$$

leading to a z-vector parameterisation of

$${}^B\mathbf{z}_G = \Big( 2(xz - wy),\, 2(yz + wx),\, 1 - 2(x^2 + y^2) \Big). \tag{5.95}$$

As given in Equation (5.45), the quaternion parameters define the fused yaw $\psi$ as

$$\psi = \text{wrap}\big(2\,\text{atan2}(z, w)\big), \tag{5.96}$$

where $\text{wrap}(\cdot)$ is a function that wraps an angle to the range $(-\pi, \pi]$ by multiples of $2\pi$. The remaining tilt angles parameters are then given by

$$\gamma = \text{atan2}(wy - xz, wx + yz), \tag{5.97a}$$

$$\alpha = \text{acos}(2(w^2 + z^2) - 1) \tag{5.97b}$$

$$= 2\,\text{acos}\Big(\sqrt{w^2 + z^2}\Big), \tag{5.97c}$$

and the remaining fused angles parameters are given by

$$\theta = \text{asin}\big(2(wy - xz)\big), \tag{5.98a}$$

$$\phi = \text{asin}\big(2(wx + yz)\big), \tag{5.98b}$$

$$h = \text{sign}(w^2 + z^2 - \tfrac{1}{2}). \tag{5.98c}$$

Care has to be taken with the combination of Equation (5.96) and Equation (5.97a) near the fused yaw singularity $\alpha = \pi$, which is equivalent to $w = z = 0$. The fused yaw has an essential discontinuity there, so both equations necessarily become numerically sensitive in that region. Due to the resulting effects, the calculated tilt parameters may not accurately describe the original quaternion. Assuming a rotation is at the fused yaw singularity, more numerically consistent equations for the tilt angles parameters are $\psi = 0$, $\alpha = \pi$, and

$$\gamma = \operatorname{atan2}(y, x). \tag{5.99}$$

Once the tilt angles parameters are known, the tilt phase parameters can trivially be calculated using Equations (5.62) and (5.66). Alternatively, the tilt phase parameters can be calculated more directly from the quaternion parameters using

$$p_x = \frac{\alpha}{S}(wx + yz), \tag{5.100a}$$

$$p_y = \frac{\alpha}{S}(wy - xz), \tag{5.100b}$$

$$p_z = \operatorname{wrap}\big(2\operatorname{atan2}(z, w)\big), \tag{5.100c}$$

where

$$\alpha = \operatorname{acos}\big((w^2 + z^2) - (x^2 + y^2)\big), \tag{5.101a}$$

$$S = \tfrac{1}{2}\sin\alpha = \sqrt{(w^2 + z^2)(x^2 + y^2)}. \tag{5.101b}$$

Similar to the need for Equation (5.99), if $S = 0$ then

$$(p_x, p_y) = \begin{cases} (0, 0) & \text{if } w^2 + z^2 \geq x^2 + y^2, \\ \pi \dfrac{(x, y)}{\|(x, y)\|} & \text{otherwise.} \end{cases} \tag{5.102}$$

It should be noted that if the operand of $\operatorname{acos}(\cdot)$ in Equation (5.101a) is divided by $(w^2 + z^2) + (x^2 + y^2)$, which is nominally 1 for a unit quaternion $q = (w, x, y, z)$, then the entire conversion algorithm to the tilt phase space is quaternion magnitude independent.

### 5.5.5 From Rotation Matrix To

#### 5.5.5.1 *From Rotation Matrix to Z-vector*

The most trivial conversion of a rotation matrix $R = [R_{ij}] \in \mathrm{SO}(3)$ is to the corresponding z-vector representation of the tilt rotation component, given simply by

$$^B\mathbf{z}_G = (R_{31}, R_{32}, R_{33}). \tag{5.103}$$

In fact, as the z-vector is exactly just the bottom-most row of the rotation matrix, and directly and completely defines the tilt rotation component (see Section 5.4.2.1), all conversions from the rotation matrix $R$ to a tuple of corresponding tilt rotation parameters (e.g. to a $(\gamma, \alpha)$, $(\phi, \theta, h)$ or $(p_x, p_y)$ tuple) simultaneously demonstrate how to convert a pure z-vector to the required parameters. Obviously, in order to convert a z-vector ${}^B\mathbf{z}_G$ into a full rotation representation however, such as for example a rotation matrix or quaternion, an additional specification of fused yaw $\psi$ is required. These kinds of more complicated conversions are addressed later in Section 7.3.7, and more specifically in Section 7.3.7.1.

### 5.5.5.2  *Partitioning the Rotation Space into Dominant Regions*

In general, rotation matrices have the difficulty that they describe the three-dimensional rotation space with nine parameters, leading to a six parameter redundancy. As such, when converting a rotation matrix to another representation, there are often multiple ways in which the required information can be extracted, but in general no one way for which the calculation is globally robust. For example, if $q = (w, x, y, z) \in \mathbb{Q}$ is the quaternion representation of the rotation matrix $R \in \mathrm{SO}(3)$, and we define the four non-unit quaternions

$$q_w = \tfrac{1}{4}(1 + R_{11} + R_{22} + R_{33},\; R_{32} - R_{23},\; R_{13} - R_{31},\; R_{21} - R_{12}), \quad (5.104a)$$

$$q_x = \tfrac{1}{4}(R_{32} - R_{23},\; 1 + R_{11} - R_{22} - R_{33},\; R_{21} + R_{12},\; R_{13} + R_{31}), \quad (5.104b)$$

$$q_y = \tfrac{1}{4}(R_{13} - R_{31},\; R_{21} + R_{12},\; 1 - R_{11} + R_{22} - R_{33},\; R_{32} + R_{23}), \quad (5.104c)$$

$$q_z = \tfrac{1}{4}(R_{21} - R_{12},\; R_{13} + R_{31},\; R_{32} + R_{23},\; 1 - R_{11} - R_{22} + R_{33}), \quad (5.104d)$$

then by substituting in the formulas for the rotation matrix entries $R_{ij}$ in terms of the quaternion components, given implicitly in Equation (5.94), we can see that

$$q_w = (w^2,\; wx,\; wy,\; wz) = wq, \quad (5.105a)$$

$$q_x = (xw,\; x^2,\; xy,\; xz) = xq, \quad (5.105b)$$

$$q_y = (yw,\; yx,\; y^2,\; yz) = yq, \quad (5.105c)$$

$$q_z = (zw,\; zx,\; zy,\; z^2) = zq. \quad (5.105d)$$

Thus, if we calculate each of the four quaternions $q_w$, $q_x$, $q_y$ and $q_z$ using Equation (5.104), and then normalise them, then the output in each case is $q$ or $-q$, which are both equivalent quaternion representations for the rotation matrix $R$. The possible difference in sign comes from the fact that the normalisation step effectively divides the four quaternions by $|w|$, $|x|$, $|y|$ and $|z|$ respectively, and because, for example,

$$\frac{w}{|w|} = \pm 1, \quad (5.106)$$

for all $w \in \mathbb{R} \backslash \{0\}$. As such, combined with a normalisation step, Equations (5.104) and (5.105) define four different ways of extracting the equivalent quaternion from a rotation matrix. None of these four ways are globally robust however, as for example the normalisation step from $q_w$ to $\pm q$ is highly numerically sensitive if $w$ is close to zero, and even undefined if $w$ is exactly zero. Nonetheless, if we partition the rotation space into four different regions, denoted $\mathcal{D}_w$, $\mathcal{D}_x$, $\mathcal{D}_y$ and $\mathcal{D}_z$, where in each region the corresponding quaternion parameter is guaranteed to be far from zero, then a globally robust conversion from rotation matrices to quaternions can be achieved by always using the $q_w$ method in region $\mathcal{D}_w$, where $w$ is far from zero, the $q_x$ method in region $\mathcal{D}_x$, where $x$ is far from zero, and so on. The four regions are collectively referred to as the <span style="color:red">dominant regions</span> of the rotation space, and the individual regions are referred to as the w-dominant region ($\mathcal{D}_w$), x-dominant region ($\mathcal{D}_x$), and so on. As shown later in this section and thesis, the ability to partition the rotation space into these dominant regions allows robust formulas in terms of rotation matrix entries to be derived for a multitude of parameters and representations, including in particular the fused yaw $\psi$.

At this point, the only missing link is how to actually precisely define the dominant regions $\mathcal{D}_w$, $\mathcal{D}_x$, $\mathcal{D}_y$ and $\mathcal{D}_z$. Every possible 3D rotation must be in one of these dominant regions, and every rotation in a particular dominant region must have the corresponding quaternion parameter far from zero. Clearly, there is some freedom how these conditions can be satisfied, but the further a quaternion parameter is guaranteed to be away from zero in the corresponding dominant region, the better the robustness of the resulting conversions will be. Thus, the most obvious way of defining the dominant region sets is to map each 3D rotation to the dominant region that corresponds to the quaternion parameter with the greatest absolute value. That is, given any rotation matrix $R \in SO(3)$ with equivalent quaternion parameters $q = (w, x, y, z) \in \mathbb{Q}$, both $R$ and $q$ equivalently belong to the dominant region $\mathcal{D}_m$, where

$$m = \underset{u \in \{w,x,y,z\}}{\arg\max} |u|. \tag{5.107}$$

To give a simple example, the quaternion $(0.6, 0, -0.8, 0) \in \mathbb{Q}$ shows dominance in the y-component, as this is the maximum absolute component, and is thus an element of $\mathcal{D}_y$. Mathematically, this is simply written as $(0.6, 0, -0.8, 0) \in \mathcal{D}_y$. Equation (5.107) works as a definition of the dominant regions, as by definition,

$$|w|, |x|, |y|, |z| \leq |m| \tag{5.108}$$

which leads to

$$w^2, x^2, y^2, z^2 \leq m^2 \tag{5.109}$$

and thus from the unit norm constraint,

$$1 = w^2 + x^2 + y^2 + z^2 \leq 4m^2 \implies |m| \geq \tfrac{1}{2}. \tag{5.110}$$

It can be observed that if the previously described scheme is followed of normalising the quaternion in Equation (5.104) that corresponds to the required dominant region, then the normalisation step now effectively involves dividing by $|m|$, which is always guaranteed to be greater than $\frac{1}{2}$. This demonstrates why the approach of using dominant regions for the conversion of rotation matrices to quaternions is globally highly numerically stable.

Given a rotation matrix $R = [R_{ij}] \in SO(3)$, the problem still remains how to identify which dominant region it belongs to, as Equation (5.107) requires knowledge of the quaternion parameters. For this and other situations, the following is a list of five completely equivalent characterisations of the definition of the dominant regions, including the nominal definition:

1. Every quaternion $q$ is assigned to the dominant region corresponding to the quaternion component of maximum absolute value.

2. Every rotation matrix $R$ is assigned to $\mathcal{D}_w$ if all three pairwise sums of the diagonal entries are non-negative, and otherwise to one of the dominant regions $\mathcal{D}_x$, $\mathcal{D}_y$ and $\mathcal{D}_z$, depending on which of $R_{11}$, $R_{22}$ and $R_{33}$ is maximum, respectively. Note that the diagonal entries are really compared based on their complete signed value, and not based on just their magnitude.

3. Same as characterisation 2, only a rotation matrix $R$ is assigned to $\mathcal{D}_w$ if at most one diagonal entry is negative, and if so, if that negative entry is smaller in magnitude than the two other positive diagonal entries.

4. Given that the four elemental quaternions are

$$q_{ew} = (1, 0, 0, 0), \tag{5.111a}$$
$$q_{ex} = (0, 1, 0, 0), \tag{5.111b}$$
$$q_{ey} = (0, 0, 1, 0), \tag{5.111c}$$
$$q_{ez} = (0, 0, 0, 1), \tag{5.111d}$$

every quaternion $q$ is assigned to the closest elemental quaternion $q_{em}$ (and therefore $\mathcal{D}_m$), judged by the minimum angle in four-dimensional space between $\pm q$ and $\pm q_{e*}$ (for $* = w, x, y, z$), i.e. the minimum angle $\Theta(q, q_{e*})$ between the quaternion lines $q$ and $q_{e*}$. More complete definitions of *quaternion lines*, and the angles $\Theta(\cdot, \cdot)$ between them, are provided in Section 7.1.2.3.

5. Every rotation corresponding to a coordinate frame {B} relative to {G} is assigned to the dominant region for which the magnitude of the direct axis-angle rotation from {B} to the frame resulting when $q_{e*}$ is applied to {G}, is minimum (for $* = w, x, y, z$). Note

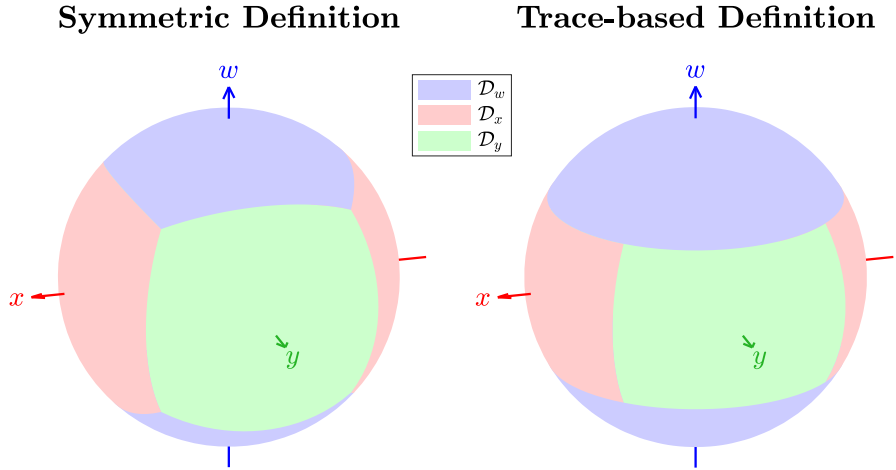

**Symmetric Definition**          **Trace-based Definition**

Figure 5.12: Visualisations of the dominant regions $\mathcal{D}_w$, $\mathcal{D}_x$ and $\mathcal{D}_y$ as patches on the quaternion unit sphere Q for the simplified case of tilt rotations, i.e. $z = 0$. The nominal symmetric definition as per Equation (5.112) is shown on the left, and the alternative trace-based definition as per Equation (5.113) is shown on the right. Both are totally valid definitions, and the full four-dimensional case can be imagined by adding a pair of diametrically opposite $\mathcal{D}_z$ patches along the fourth dimension.

that the elemental quaternion rotations $q_{e*}$ correspond to the identity rotation, along with the three 180° rotations about the x, y and z-axes respectively.

Based on characterisation 2, an algorithm for determining which dominant region $\mathcal{D}_m$ a rotation matrix $R$ belongs to, is then given by

$$\mathcal{D}_m = \begin{cases} \mathcal{D}_w & \text{if } R_{11} + R_{22}, \ R_{22} + R_{33}, \ R_{33} + R_{11} \geq 0, \\ \mathcal{D}_x & \text{else if } R_{11} \geq R_{22} \text{ and } R_{11} \geq R_{33}, \\ \mathcal{D}_y & \text{else if } R_{22} \geq R_{33}, \\ \mathcal{D}_z & \text{otherwise.} \end{cases} \tag{5.112}$$

This algorithm assigns rotations on the border between two or more dominant regions to the chronologically first such dominant region that is tested for.

As can be directly surmised from most of the characterisations given in the list above, the four dominant regions are in fact a symmetric four-way partition of the rotation space. As such, the dominant regions can be most readily visualised by the symmetric patches they form on the surface of the four-dimensional unit sphere $\mathcal{S}^3$. These can be imagined by considering the three-dimensional example case of pure tilt rotations (see left side of Figure 5.12), for which $z = 0$, and mentally extending this to four dimensions.

### 5.5.5.3  *An Alternative Definition of the Dominant Regions*

As stated previously, there is some freedom in how exactly the dominant regions are defined, as all they are doing is trading off the numerical sensitivities of various interchangeable analytically correct formulas. As such, one can choose to slightly change and simplify the first condition in Equation (5.112) to give

$$
\mathcal{D}_m = \begin{cases}
\mathcal{D}_w & \text{if } \operatorname{tr}(R) \geq 0, \\
\mathcal{D}_x & \text{else if } R_{11} \geq R_{22} \text{ and } R_{11} \geq R_{33}, \\
\mathcal{D}_y & \text{else if } R_{22} \geq R_{33}, \\
\mathcal{D}_z & \text{otherwise},
\end{cases} \tag{5.113}
$$

where

$$
\operatorname{tr}(R) = R_{11} + R_{22} + R_{33} \tag{5.114}
$$

is the trace of the matrix $R$. The new trace-based condition for $\mathcal{D}_w$ is a slight relaxation of the old symmetric condition, and in fact can be inferred from it, but comes at essentially no cost to numerical sensitivity, as from Equations (5.22) and (5.94),

$$
0 \leq \operatorname{tr}(R) = R_{11} + R_{22} + R_{33} = 4w^2 - 1 \iff |w| \geq \tfrac{1}{2}. \tag{5.115}
$$

Thus, the guarantee from Equation (5.110) that $|m| \geq \tfrac{1}{2}$ still stands with the new trace-based condition. The condition can be visualised by identifying that

$$
\operatorname{tr}(R) = 1 + 2\cos\theta_a, \tag{5.116}
$$

where $\theta_a$ is the axis-angle magnitude of the rotation $R$, and observing that as a result,

$$
\operatorname{tr}(R) \geq 0 \iff \theta_a \leq \tfrac{2\pi}{3} = 120°. \tag{5.117}
$$

In effect, the trace-based definition of $\mathcal{D}_w$ consists of all rotations of up to 120° in magnitude. The dominant region patches can be visualised, similar to before, by considering the three-dimensional example case of pure tilt rotations (see right side of Figure 5.12), and extending it mentally to 4D.

### 5.5.5.4  *From Rotation Matrix to Other Representations*

Putting together everything we have developed in the previous two sections, we can now see that a globally robust conversion from the rotation matrix $R$, to its equivalent quaternion $q$, can be succinctly written as

$$
q = \frac{\tilde{q}}{\|\tilde{q}\|}, \tag{5.118}
$$

where, making use of Equation (5.104),

$$\tilde{q} = \begin{cases} q_w & \text{if } R \in \mathcal{D}_w, \\ q_x & \text{if } R \in \mathcal{D}_x, \\ q_y & \text{if } R \in \mathcal{D}_y, \\ q_z & \text{if } R \in \mathcal{D}_z. \end{cases} \tag{5.119}$$

As required, note that these equations are both true and robust, independent of which of the two exact definitions of the dominant regions is chosen.

By direct consequence of Equation (5.45) and the conversion presented above, a globally robust expression for the fused yaw can be deduced to be

$$\psi = \text{wrap}(2\tilde{\psi}), \tag{5.120}$$

where

$$\tilde{\psi} = \begin{cases} \text{atan2}(R_{21} - R_{12},\, 1 + R_{11} + R_{22} + R_{33}) & \text{if } R \in \mathcal{D}_w, \\ \text{atan2}(R_{13} + R_{31},\, R_{32} - R_{23}) & \text{if } R \in \mathcal{D}_x, \\ \text{atan2}(R_{32} + R_{23},\, R_{13} - R_{31}) & \text{if } R \in \mathcal{D}_y, \\ \text{atan2}(1 - R_{11} - R_{22} + R_{33},\, R_{21} - R_{12}) & \text{if } R \in \mathcal{D}_z. \end{cases} \tag{5.121}$$

We note that Equation (5.121) in terms of quaternion parameters corresponds to

$$\tilde{\psi} = \begin{cases} \text{atan2}(4wz,\, 4w^2) & \text{if } R \in \mathcal{D}_w, \\ \text{atan2}(4xz,\, 4xw) & \text{if } R \in \mathcal{D}_x, \\ \text{atan2}(4yz,\, 4yw) & \text{if } R \in \mathcal{D}_y, \\ \text{atan2}(4z^2,\, 4zw) & \text{if } R \in \mathcal{D}_z. \end{cases} \tag{5.122}$$

This explains why the dominant cases come into play for the fused yaw, as we are effectively trying to cancel 4 times a quaternion parameter from each argument of $\text{atan2}(\cdot, \cdot)$, in each case. As such, in each case it is required for numerical stability that the respective quaternion parameter being cancelled is far from zero.

With the fused yaw $\psi$ known, the remaining tilt angles parameters can then be calculated as

$$\gamma = \text{atan2}(-R_{31}, R_{32}), \tag{5.123a}$$
$$\alpha = \text{acos}(R_{33}), \tag{5.123b}$$

and the remaining fused angles parameters can be calculated as

$$\theta = \text{asin}(-R_{31}), \tag{5.124a}$$
$$\phi = \text{asin}(R_{32}), \tag{5.124b}$$
$$h = \text{sign}(R_{33}). \tag{5.124c}$$

Similar to the need for Equation (5.99), care has to be taken with the combination of Equation (5.120) and Equation (5.123a) near the fused yaw singularity $\alpha = \pi$, as there is unavoidable numerical sensitivity there. This results in pairs of $\psi$ and $\gamma$ being calculated that potentially do not accurately describe the original rotation. At the fused yaw singularity itself, and only there, more numerically consistent equations for the tilt angles parameters are given by $\psi = 0$, $\alpha = \pi$, and

$$\gamma = \tfrac{1}{2}\operatorname{atan2}(R_{12}, R_{11}).\tag{5.125}$$

Once the tilt angles parameters are known, the tilt phase parameters can trivially be calculated from Equations (5.62) and (5.66). Alternatively, the tilt phase parameters can also be calculated more directly from the rotation matrix entries using

$$p_x = \phantom{-}\frac{\alpha}{A}R_{32},\tag{5.126a}$$

$$p_y = -\frac{\alpha}{A}R_{31},\tag{5.126b}$$

$$p_z = \operatorname{wrap}(2\tilde{\psi}),\tag{5.126c}$$

where $\tilde{\psi}$ is calculated from Equation (5.121) as before, and

$$\alpha = \operatorname{acos}(R_{33}),\tag{5.127a}$$

$$A = \sin\alpha = \sqrt{R_{31}^2 + R_{32}^2}.\tag{5.127b}$$

Similar to the need for Equation (5.125), if $A = 0$ then

$$C = \cos(2\gamma) = \tfrac{1}{2}(R_{11} - R_{22}),\tag{5.128a}$$

$$S = \sin(2\gamma) = \tfrac{1}{2}(R_{12} + R_{21}),\tag{5.128b}$$

$$p_x = \begin{cases} 0 & \text{if } R_{33} \geq 0, \\ \pi\sqrt{\tfrac{1}{2}(1 + C)} & \text{otherwise,} \end{cases}\tag{5.128c}$$

$$p_y = \begin{cases} 0 & \text{if } R_{33} \geq 0, \\ \operatorname{sign}(S)\,\pi\sqrt{\tfrac{1}{2}(1 - C)} & \text{otherwise.} \end{cases}\tag{5.128d}$$

In numerical calculations, the case that $A$ is exactly zero only rarely occurs however, so these equations are more just for consistency and completeness.

## 5.6    SINGULARITY ANALYSIS

When examining rotation representations, it is important to identify and precisely quantify any singularities. It was shown by Stuelpnagel (1964) that it is topologically impossible to have a one-to-one and global three-dimensional, i.e. minimal, parameterisation of the rotation group without any singular points. For an n-dimensional parameterisation to be one-to-one and global without any singular points, there must exist a differentiable one-to-one map with differentiable inverse that carries the rotation space SO(3) to the required parameter domain subset of $\mathbb{R}^n$. Thus, for $n = 3$, it must be the case for every parameterisation that either the map is not one-to-one, not differentiable, or not inverse differentiable.

Of greatest concern to the mathematical and practical applications of particular rotation representations is the location and prevalence of singular points. The possible types of singularities of rotation representations can be characterised to primarily include:

(i) Rotations that do not possess a unique parameterised representation,

(ii) Sets of parameters that do not correspond to a unique rotation,

(iii) Rotations in the neighbourhood of which the sensitivity of the rotation to parameters map is unbounded, and,

(iv) Sets of parameters in the neighbourhood of which the sensitivity of the parameters to rotation map is unbounded.

It should be noted that strictly speaking, all cyclic parameters that loop around at $\pm\pi$ break the mathematical condition of differentiability, as this requires continuity. This is not seen to result in true singularities or discontinuities however, as the cyclic parameters are clearly viewed as having a cyclic topology, for which the jump at $\pm\pi$ is not a discontinuity.

### 5.6.1 Fused Yaw Singularity

The most prominent singularity present in the new representations developed in this thesis is the fused yaw singularity, which has been mentioned before in Section 5.4.1. The geometric definition of the fused yaw $\psi$ requires a direct planar rotation to be constructed from $\mathbf{z}_B$ to $\mathbf{z}_G$, but this rotation is not uniquely defined in situations where $\mathbf{z}_B$ and $\mathbf{z}_G$ are antiparallel, i.e. when they point in opposite directions. As shown in Figure 5.13, in such situations there are multiple combinations of yaw and tilt that can be used to arrive at the final body-fixed frame {B}, leading to an ambiguity in the definition of the associated parameters. The following are all completely equivalent characterisations of rotations at the fused yaw singularity:

• Rotations for which the body-fixed $z$-axis points vertically downwards,

• Rotations by 180° about an axis in the **xy** plane, i.e. pure 180° tilt rotations,

• Rotations for which ${}^G z_{Bz} \equiv {}^B z_{Gz} = -1$, i.e. $\mathbf{z}_B = -\mathbf{z}_G$,

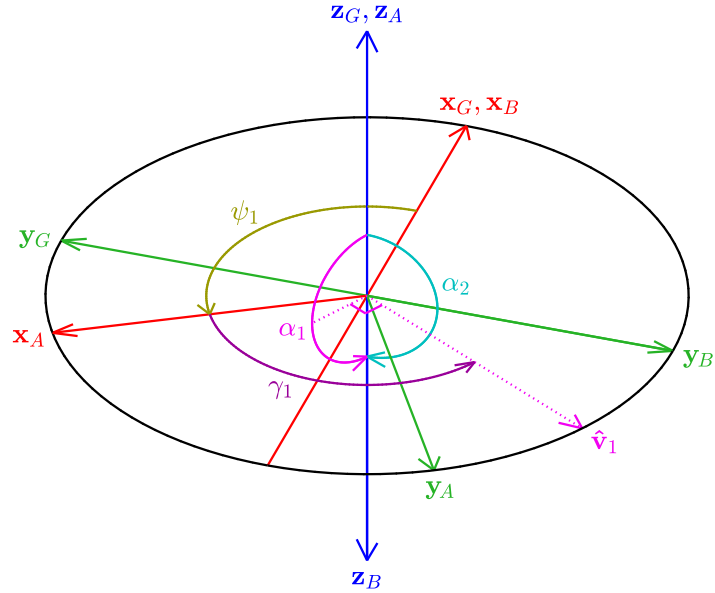

Figure 5.13: At the fused yaw singularity, i.e. $\mathbf{z}_B = -\mathbf{z}_G$, there are multiple combinations of yaw and tilt that can take a global frame {G} to the local frame {B}. In this example, a yaw rotation of $\psi_1$ followed by a tilt rotation of $(\gamma_1, \alpha_1)$ (i.e. a CCW rotation of {A} by $\alpha_1 = \pi$ radians about $\hat{\mathbf{v}}_1$), is equivalent to a yaw rotation of $\psi_2 = 0$ (i.e. no yaw rotation) followed by a tilt rotation of $(0, \alpha_2)$ (i.e. a CCW rotation of $\alpha_2 = \pi$ radians about $\mathbf{x}_G$).

- Rotation matrices for which $R_{33} = -1$,

- Quaternion rotations for which $w = z = 0$,

- Tilt angles rotations for which $\alpha = \pi$,

- Fused angles rotations for which $\theta = \phi = 0$ and $h = -1$,

- Tilt phase space rotations for which $\|(p_x, p_y)\| = \sqrt{p_x^2 + p_y^2} = \pi$.

Conceptually, the fused yaw singularity can be seen to be as 'far away' from the identity rotation as possible, as all fused yaw singular rotations are 180° away from the identity, and no two rotations in the rotation space *can* be separated by more than that. This is particularly useful for applications involving balancing bodies, as in particular for most cases of mobile robotics, a completely inverted pose is the most uncommon part of the rotation space to occur.

Based on the above characterisations of rotations at the fused yaw singularity, expressions can be developed for the formats of the various rotation representations at the fused yaw singularity. For instance:

$$T_{sing} = (\psi, \gamma, \pi), \tag{5.129a}$$

$$F_{sing} = (\psi, 0, 0, -1), \tag{5.129b}$$

$$P_{sing} = (p_x, p_y, \psi), \tag{5.129c}$$

$$q_{sing} = (0, x, y, 0) \tag{5.129d}$$

$$= \left(0, \cos(\gamma + \tfrac{\psi}{2}), \sin(\gamma + \tfrac{\psi}{2}), 0\right), \tag{5.129e}$$

where $p_x^2 + p_y^2 = \pi^2$ and $x^2 + y^2 = 1$. For rotation matrices, the format of the fused yaw singularity is given by the following expressions:

$$R_{sing} = \begin{bmatrix} R_{11} & R_{12} & 0 \\ R_{12} & -R_{11} & 0 \\ 0 & 0 & -1 \end{bmatrix} \tag{5.130a}$$

$$= \begin{bmatrix} x^2 - y^2 & 2xy & 0 \\ 2xy & y^2 - x^2 & 0 \\ 0 & 0 & -1 \end{bmatrix} \tag{5.130b}$$

$$= \begin{bmatrix} \cos(2\gamma + \psi) & \sin(2\gamma + \psi) & 0 \\ \sin(2\gamma + \psi) & -\cos(2\gamma + \psi) & 0 \\ 0 & 0 & -1 \end{bmatrix}, \tag{5.130c}$$

where $R_{11}^2 + R_{12}^2 = 1$ in Equation (5.130a), and $x^2 + y^2 = 1$ in Equation (5.130b).

As per the characterisation of singularities given at the start of this section, the fused yaw singularity is a singularity of types (i) and (iii) for tilt angles and the tilt phase space, and types (i), (ii) and (iii) for fused angles. The difference comes about from the way that the fused angles tilt rotation component parameters are equal for all fused yaw singular rotations, which is not the case for the other two representations. The singularity is of type (iii) because it constitutes an essential discontinuity in the map from the rotation space to the respective fused yaws. As a result, given any fused yaw singular rotation $R$, and any neighbourhood $U$ of $R$ in the rotation space $SO(3)$, for every $\psi \in (-\pi, \pi]$ there exists a rotation in $U$ with a fused yaw of $\psi$. Despite this discontinuity, by convention when calculating the fused yaw of a singular rotation, the value of $\psi = 0$ is used. This makes sense because the value of zero emerges naturally from the evaluation of $\text{atan2}(0, 0)$, which occurs in all formulas for $\psi$ when the input rotation is fused yaw singular, and because it is most natural to parameterise pure tilt rotations in a way that has zero yaw. Despite the convention of using $\psi = 0$ however, for tilt angles and the tilt phase space it is still nonetheless possible to unambiguously interpret sets of singular parameters with $\psi \neq 0$. This is done, for example for the parameter set $T = (\psi, \gamma, \pi) \in \mathbb{T}$, by first geometrically applying the yaw rotation that is defined by $\psi$, followed by the tilt rotation that is defined by $\gamma$ and $\pi$. This yields a well-defined and unambiguous final rotation in $SO(3)$ corresponding to the tilt angles parameters $T$.

### 5.6.2  Other Singularities

In knowledge of the nature of the fused yaw singularity and its effects also on the other parameters, the remaining singularities in the new representations developed in this thesis can be determined by examining just pure tilt rotations of up to but not including 180°. This follows directly from the general continuity of the fused yaw, and the characterisation that the fused yaw singular rotations are pure 180° tilt rotations. Consider for example the fused angles parameterisation $(\theta, \phi, h)$ of a tilt rotation. The entries $R_{ij} \in [-1, 1]$ of the associated rotation matrix are well-known to be a continuous, and in fact smooth, function of the underlying 3D rotation. As such, from Equation (5.124) and the continuity of the standard asin($\cdot$) function, it can be seen that the fused pitch $\theta$ and the fused roll $\phi$ are continuous over the entire rotation space. Furthermore, the hemisphere parameter of the fused angles representation is uniquely and unambiguously defined over the rotation space. As a result, despite its discrete and thereby technically discontinuous nature, the hemisphere parameter is not considered to be the cause of any singularities in the fused angles representation.

It can be seen from Equations (5.129e) and (5.130c) that at the fused yaw singularity, the fused yaw $\psi$ and tilt axis angle $\gamma$ are linearly dependent in the sense that only the value of $\gamma + \frac{\psi}{2}$ matters. Thus, the tilt axis angle, like the fused yaw $\psi$, also suffers from a singularity at $\alpha = \pi$ due to the fused yaw singularity. In addition to this however, similar to the nature of polar coordinates at the origin, the $\gamma$ parameter has an essential discontinuity at $\alpha = 0$. The local sensitivity of $\gamma$ around this discontinuity is not in general a problem however, as the closer $\alpha$ gets to zero, the proportionally less effect the value of $\gamma$ actually has on the rotation that the parameters represent. This is reflected, for example, by Equation (5.78b) in that all uses of $\gamma$ are premultiplied by $\sin \frac{\alpha}{2}$, which tends to zero for small values of $\alpha$. Equation (5.123a) demonstrates through the continuity of atan2($\cdot, \cdot$) away from $(0, 0)$ that the tilt axis angle $\gamma$ is in fact continuous everywhere except for the aforementioned cases of $\alpha = 0$ and $\alpha = \pi$, as these conditions correspond exactly to $R_{31} = R_{32} = 0$. Similarly, Equation (5.123b) demonstrates from the continuity of acos($\cdot$) that the tilt angle $\alpha$ is continuous everywhere, even if it can be observed that there are cusps at $\alpha = 0$ and $\pi$.

The singularity analysis for the tilt phase space is slightly more complicated, as the relation to the rotation matrix parameters is somewhat less direct, but the tilt phase space is in fact continuous and smooth everywhere away from the fused yaw singularity. Given the definition of the tilt phase space in terms of the tilt angles parameters $\gamma$ and $\alpha$, it is not immediately clear why this should be the case, but a proof involving the smoothness of the cardinal sine function has been provided in Appendix B.1.1.2.

## 5.7 SELECTED PROPERTIES OF YAW-TILT ROTATIONS

The fused angles, tilt angles and tilt phase space representations all possess a remarkable number of sometimes subtle, yet powerful, properties that turn out to be useful in many applications and contexts. A small selection of such mathematical and geometric properties are presented in this section. A continuation of many more such properties however, is presented later in Section 7.3.

### 5.7.1 Rotation Decomposition into Yaw and Tilt

As shown in detail in Section 5.4, all 3D rotations can be decomposed into their fused yaw and tilt components, where these components are two completely independent entities that together define the rotation. In reference to Section 5.4.1 and Figure 5.5, the fused yaw component is the rotation from {G} to {A}, and the tilt rotation component is the rotation from {A} to {B}. As these are just two normal successive rotations, the total rotation from {G} to {B} can therefore be expressed in terms of the yaw and tilt components using standard rotation composition. If '∘' is used to denote rotation composition equivalent to plain multiplication for rotation matrices and quaternions, then

$$R = R_f R_t \quad = R_z(\psi) R_t, \tag{5.131a}$$

$$q = q_f q_t \quad = q_z(\psi) q_t, \tag{5.131b}$$

$$T = T_f \circ T_t = T_z(\psi) \circ T_t, \tag{5.131c}$$

$$F = F_f \circ F_t = F_z(\psi) \circ F_t, \tag{5.131d}$$

$$P = P_f \circ P_t = P_z(\psi) \circ P_t, \tag{5.131e}$$

where $R_f$, for example, is the fused yaw rotation component of $R$, and is equivalent to the rotation matrix $R_z(\psi)$, which corresponds to a CCW z-rotation by $\psi$, the fused yaw of the rotations $R$, $q$, $T$ etc. Written more explicitly, the fused yaw rotation components $\square_f$ are given by

$$R_f = \begin{bmatrix} \cos\psi & -\sin\psi & 0 \\ \sin\psi & \cos\psi & 0 \\ 0 & 0 & 1 \end{bmatrix}, \tag{5.132a}$$

$$q_f = (\cos\tfrac{\psi}{2},\, 0,\, 0,\, \sin\tfrac{\psi}{2}), \tag{5.132b}$$

$$T_f = (\psi, 0, 0), \tag{5.132c}$$

$$F_f = (\psi, 0, 0, 1), \tag{5.132d}$$

$$P_f = (0, 0, \psi). \tag{5.132e}$$

The tilt rotation components $\square_t$ in Equation (5.131) are given by

$$R_t = \begin{bmatrix} c_\gamma^2 + c_\alpha s_\gamma^2 & c_\gamma s_\gamma(1 - c_\alpha) & s_\alpha s_\gamma \\ c_\gamma s_\gamma(1 - c_\alpha) & s_\gamma^2 + c_\alpha c_\gamma^2 & -s_\alpha c_\gamma \\ -s_\alpha s_\gamma & s_\alpha c_\gamma & c_\alpha \end{bmatrix}, \tag{5.133a}$$

$$q_t = (w, x, y, 0) \tag{5.133b}$$

$$= (\cos \tfrac{\alpha}{2}, \sin \tfrac{\alpha}{2} \cos \gamma, \sin \tfrac{\alpha}{2} \sin \gamma, 0), \tag{5.133c}$$

$$T_t = (0, \gamma, \alpha), \tag{5.133d}$$

$$F_t = (0, \theta, \phi, h), \tag{5.133e}$$

$$P_t = (p_x, p_y, 0), \tag{5.133f}$$

where we recall the notation that $c_* \equiv \cos *$ and $s_* \equiv \sin *$. Using Equations (5.131) to (5.133), it is trivial to construct a rotation from its yaw and tilt components if they are given in essentially any parameterisation. The only exception to this is if the tilt rotation component is given in terms of the $z$-vector parameterisation (see Section 5.4.2.1). In this case, the methods described in Sections 7.3.7.1 and 7.3.7.2 can be used.

If $R = [R_{ij}]$, $q = (w, x, y, 0)$, $T = (0, \gamma, \alpha)$ and $F = (0, \theta, \phi, h)$ all represent the same tilt rotation, then alternative formulas for the tilt rotation component $R_t$ include

$$R_t = \begin{bmatrix} R_{11} & R_{12} & -R_{31} \\ R_{12} & R_{22} & -R_{32} \\ R_{31} & R_{32} & R_{33} \end{bmatrix} \tag{5.134a}$$

$$= \begin{bmatrix} 1 - 2y^2 & 2xy & 2wy \\ 2xy & 1 - 2x^2 & -2wx \\ -2wy & 2wx & 2w^2 - 1 \end{bmatrix} \tag{5.134b}$$

$$= \begin{bmatrix} 1 - s_\gamma^2(1 - c_\alpha) & c_\gamma s_\gamma(1 - c_\alpha) & s_\alpha s_\gamma \\ c_\gamma s_\gamma(1 - c_\alpha) & 1 - c_\gamma^2(1 - c_\alpha) & -s_\alpha c_\gamma \\ -s_\alpha s_\gamma & s_\alpha c_\gamma & c_\alpha \end{bmatrix} \tag{5.134c}$$

$$= \begin{bmatrix} \cdot & \cdot & s_\theta \\ \cdot & \cdot & -s_\phi \\ -s_\theta & s_\phi & h\sqrt{1 - s_\theta^2 - s_\phi^2} \end{bmatrix}, \tag{5.134d}$$

where in Equation (5.134b) we have that $w^2 + x^2 + y^2 = 1$.

## 5.7.2   Links Between Quaternions and Fused Yaw

For rotations away from the fused yaw singularity $\alpha = \pi$, that is, for rotations where the fused yaw is well-defined and unambiguous, inspection of Equation (5.78) reveals that the $z$-component

$$z = \cos \tfrac{\alpha}{2} \sin \tfrac{\psi}{2} \tag{5.135}$$

of a quaternion $q = (w, x, y, z) \in \mathbb{Q}$ is zero if and only if the fused yaw is zero. This comes about because the $\cos \tfrac{\alpha}{2}$ term is non-zero for $\alpha \in [0, \pi)$. Expressed in terms of mathematical notation, we have that

$$\psi = 0 \iff z = 0, \tag{5.136}$$

and consequently, from Equation (5.24b), that

$$\psi = 0 \iff e_z = 0. \tag{5.137}$$

This is a remarkable and not to be undervalued property of the fused yaw, as it asserts that the fused yaw is zero if and only if there is no component of rotation about the vertical z-axis. This, by contrast, is not true for any other definition of yaw. It can further be observed that the quaternion $q_f$ corresponding to the fused yaw component of a rotation can be constructed by zeroing the x and y-components of the quaternion $q = (w, x, y, z)$ and renormalising. This leads to the equation

$$q_f = \frac{1}{\sqrt{w^2 + z^2}} (w, 0, 0, z). \tag{5.138}$$

and like Equation (5.136) makes great intuitive sense, as we are essentially zeroing the x and y-components of the axis of rotation of $q$, leaving just the z-axis yaw rotation component. Once again, Equation (5.138) and this kind of thinking does not apply to any other definition of yaw. The ability to so easily and directly calculate $q_f$ leads to one way of removing the fused yaw component of a quaternion—something that is a surprisingly common operation—using the expression

$$q_t = q_f^* q \tag{5.139a}$$

$$= \frac{1}{\sqrt{w^2 + z^2}} \big( wq + z(z, y, -x, -w) \big), \tag{5.139b}$$

where $q_f^*$ is the conjugate, and therefore inverse, of $q_f$ (recall that $q_f$ is a unit quaternion). As an alternative to Equation (5.139), the fused yaw can also be calculated directly using Equation (5.96) and manually removed using the fact that

$$q_f = q_z(\psi) = (\cos \tfrac{\psi}{2},\ 0,\ 0,\ \sin \tfrac{\psi}{2}). \tag{5.140}$$

Equations (5.138) and (5.139b) fail if and only if $w = z = 0$, which is precisely equivalent to $\alpha = \pi$, the fused yaw singularity.

### 5.7.3 Fused Yaw Under Pure Z-rotations

As described in Section 5.2.1, an essential property of any definition of yaw is yaw additivity. This states that if a global z-rotation is applied to any body-fixed frame {B}, then its yaw should change by exactly the amount of that z-rotation, up to angle wrapping about $\pm\pi$. The fused yaw $\psi$ satisfies this property, and in fact unlike any other definition of yaw, somewhat remarkably satisfies this property irrespective of whether the applied z-rotation is local (about $\mathbf{z}_B$) or global (about $\mathbf{z}_G$).

If $\Psi(\cdot)$ is an operator that returns the fused yaw of a rotation in any representation, the stated property of fused yaw can be expressed as

$$\Psi\big(R_z(\psi_z)R\big) = \Psi\big(RR_z(\psi_z)\big) = \text{wrap}\big(\Psi(R) + \psi_z\big), \qquad (5.141a)$$

$$\Psi\big(q_z(\psi_z)q\big) = \Psi\big(qq_z(\psi_z)\big) \;\; = \text{wrap}\big(\Psi(q) + \psi_z\big), \qquad (5.141b)$$

where $q_z(\psi_z)$, for example, refers to the CCW quaternion rotation by $\psi_z$ radians about the z-axis, premultiplication by $q_z(\psi_z)$ corresponds to a z-rotation that is global, and post-multiplication by $q_z(\psi_z)$ corresponds to a z-rotation that is local. As demanded by yaw additivity, when global z-rotations are applied, the additional result holds that the tilt rotation component remains completely unchanged. If the '$\circ$' operator is used to denote rotation composition for representations where this is not just multiplication, this result can be written as

$$R_z(\psi_z)R = R_z(\psi_z)\big(R_z(\psi)R_t\big) = R_z(\psi + \psi_z)R_t, \qquad (5.142a)$$

$$q_z(\psi_z)q = q_z(\psi_z)\big(q_z(\psi)q_t\big) = q_z(\psi + \psi_z)q_t, \qquad (5.142b)$$

$$T_z(\psi_z) \circ T(\psi, \gamma, \alpha) = T(\psi + \psi_z, \gamma, \alpha), \qquad (5.142c)$$

$$F_z(\psi_z) \circ F(\psi, \theta, \phi, h) = F(\psi + \psi_z, \theta, \phi, h), \qquad (5.142d)$$

$$P_z(\psi_z) \circ P(p_x, p_y, p_z) = T(p_x, p_y, p_z + \psi_z), \qquad (5.142e)$$

where $\psi$ is the fused yaw in each case prior to global z-rotation (i.e. premultiplication), $R_t$ and $q_t$ are the tilt rotation components of $R$ and $q$ respectively (see Section 5.7.1), and $T(\cdot, \cdot, \cdot)$, for example, is just notation for the tilt angles rotation with the enclosed parameters. Proofs for Equations (5.141) and (5.142) are given in Appendix B.1.1.3.

### 5.7.4 Rotation Inverses

The inverses of rotation matrices and quaternions are trivially given by

$$R^{-1} = R^T, \qquad (5.143a)$$

$$q^{-1} = q^*, \qquad (5.143b)$$

where $q^*$ denotes the conjugate of the quaternion $q$. The situation is however not always that easy, like for example for Euler angles, where the inverse of the rotation $E = (\psi_E, \theta_E, \phi_E)$ is given by

$$E^{-1} = (\psi_{Einv}, \theta_{Einv}, \phi_{Einv}), \qquad (5.144)$$

where

$$\psi_{Einv} = \text{atan2}(c_{\psi_E}s_{\theta_E}s_{\phi_E} - s_{\psi_E}c_{\phi_E}, \; c_{\psi_E}c_{\theta_E}), \qquad (5.145a)$$

$$\theta_{Einv} = -\text{asin}(c_{\psi_E}s_{\theta_E}c_{\phi_E} + s_{\psi_E}s_{\phi_E}), \qquad (5.145b)$$

$$\phi_{Einv} = \text{atan2}(s_{\psi_E}s_{\theta_E}c_{\phi_E} - c_{\psi_E}s_{\phi_E}, \; c_{\theta_E}c_{\phi_E}). \qquad (5.145c)$$

The tilt angles, fused angles and tilt phase space parameters, on the other hand, are intricately linked to their inverses. For instance, the inverse of a fused angles rotation $F = (\psi, \theta, \phi, h)$ is given by

$$F^{-1} = (-\psi, \theta_{inv}, \phi_{inv}, h), \tag{5.146}$$

where

$$\theta_{inv} = -\operatorname{asin}(c_\psi s_\theta + s_\psi s_\phi) = -\operatorname{asin}(s_\alpha s_{\psi+\gamma}), \tag{5.147a}$$

$$\phi_{inv} = \operatorname{asin}(s_\psi s_\theta - c_\psi s_\phi) = -\operatorname{asin}(s_\alpha c_{\psi+\gamma}). \tag{5.147b}$$

It is quite remarkable to note from Equation (5.146) that $\psi_{inv} = -\psi$, i.e.

$$\Psi(R^{-1}) = -\Psi(R). \tag{5.148}$$

This property of fused yaw is referred to as the <span style="color:magenta">negation through rotation inversion</span> property, and is clearly not satisfied by any variant of Euler yaw, or any other definition of yaw for that matter. For pure tilt rotations, the negation through rotation inversion property also applies to the fused pitch and roll. Specifically,

$$\psi = 0 \implies F^{-1} = (0, -\theta, -\phi, h). \tag{5.149}$$

By comparison, for zero Euler yaw $\psi_E$, the expression for the inverse rotation does not simplify as significantly:

$$\psi_E = 0 \implies \begin{aligned} E^{-1} = \big( &\operatorname{atan2}(s_{\theta_E} s_{\phi_E}, c_{\theta_E}), \\ &\operatorname{asin}(-s_{\theta_E} c_{\phi_E}), \\ &\operatorname{atan2}(-s_{\phi_E}, c_{\theta_E} c_{\phi_E}) \big). \end{aligned} \tag{5.150}$$

Given a tilt angles rotation $T = (\psi, \gamma, \alpha)$, the inverse is given by

$$T^{-1} = (-\psi, \operatorname{wrap}(\psi + \gamma - \pi), \alpha), \tag{5.151}$$

or if the domain of the resulting parameters is not so important,

$$T^{-1} = (-\psi, \psi + \gamma, -\alpha). \tag{5.152}$$

The inverses of the relative and absolute tilt phase space representations $P = (p_x, p_y, p_z)$ and $\tilde{P} = (\tilde{p}_x, \tilde{p}_y, \tilde{p}_z)$ are given by

$$P^{-1} = (-c_\psi p_x + s_\psi p_y, \; -s_\psi p_x - c_\psi p_y, \; -p_z), \tag{5.153a}$$

$$\tilde{P}^{-1} = (-c_\psi \tilde{p}_x - s_\psi \tilde{p}_y, \; s_\psi \tilde{p}_x - c_\psi \tilde{p}_y, \; -\tilde{p}_z), \tag{5.153b}$$

In light of Equations (5.70) and (5.71), by consequence it can be seen that the relative and absolute tilt phase spaces are in fact just negative inverses of each other. That is,

$$P^{-1} = -\tilde{P}, \tag{5.154a}$$

$$\tilde{P}^{-1} = -P. \tag{5.154b}$$

As $\tilde{P} \equiv P$ for pure tilt rotations, as a corollary, the tilt phase space satisfies the negation through rotation inversion property for pure tilt rotations, i.e.

$$p_z = \psi = 0 \implies P^{-1} = -P. \tag{5.155}$$

### 5.7.5  Links Between Fused Angles and Euler Angles

Even though the interpretations of the variables are quite different, and the nature of the domains do not correspond, purely mathematically it can be observed from Equations (5.35b) and (5.124a) that the ZYX Euler pitch is equal to the fused pitch, and from Equations (5.41b) and (5.124b) that the ZXY Euler roll is equal to the fused roll. That is,

$$\theta_E = \theta, \tag{5.156a}$$

$$\phi_{\tilde{E}} = \phi. \tag{5.156b}$$

As such, fused angles can be seen to—with an adaptation of the domains and geometric interpretation—unite the ZYX Euler pitch and ZXY Euler roll with a novel and meaningful concept of yaw, to form a useful representation of rotations.

There are numerous links that can be made between the (ZYX) Euler angles, fused angles and tilt angles spaces, beyond the equivalence described in Equation (5.156a). The following relations, for example, express the link between the Euler roll and the fused roll parameters:

$$\phi = \mathrm{asin}(c_{\theta_E} s_{\phi_E}), \tag{5.157a}$$

$$\phi_E = \mathrm{atan2}(s_\phi, c_\alpha). \tag{5.157b}$$

On the other hand, the tilt axis angle $\gamma$, and tilt angle $\alpha$, are given in terms of the Euler parameters by the formulas

$$\gamma = \mathrm{atan2}(s_{\theta_E}, c_{\theta_E} s_{\phi_E}), \tag{5.158a}$$

$$\alpha = \mathrm{acos}(c_{\theta_E} c_{\phi_E}), \tag{5.158b}$$

and the fused hemisphere parameter $h$ can be expressed in terms of the Euler roll as

$$h = \mathrm{sign}(c_{\phi_E}) = \begin{cases} 1 & \text{if } |\phi_E| \leq \frac{\pi}{2}, \\ -1 & \text{otherwise.} \end{cases} \tag{5.159}$$

Based on Equation (5.157a), a result similar to Equation (5.82) can be deduced for the Euler pitch and roll parameters, namely

$$s_\alpha^2 = s_{\theta_E}^2 + s_{\phi_E}^2 - s_{\theta_E}^2 s_{\phi_E}^2. \tag{5.160}$$

In terms of the yaw parameters, the relations between the Euler angles, fused angles and tilt angles parameters is given by

$$\psi_E = \text{wrap}\big(\psi + \gamma - \text{atan2}(c_\alpha s_\gamma, c_\gamma)\big), \tag{5.161a}$$

$$\psi = \text{wrap}\big(\psi_E - \gamma + \text{atan2}(c_\alpha s_\gamma, c_\gamma)\big) \tag{5.161b}$$

$$= \text{wrap}\big(\psi_E - \text{atan2}(s_\theta, s_\phi) + \text{atan2}(s_\theta c_{\phi_E}, s_{\phi_E})\big) \tag{5.161c}$$

$$= \text{wrap}\big(\psi_E - \text{atan2}(s_{\theta_E}, c_{\theta_E} s_{\phi_E}) + \text{atan2}(s_{\theta_E} c_{\phi_E}, s_{\phi_E})\big), \tag{5.161d}$$

at least away from the fused yaw and Euler yaw singularities. It can be seen from Equation (5.161) that if any of the fused pitch, Euler pitch or Euler roll parameters are zero, then the fused yaw and Euler yaw are equal. That is,

$$\phi_E = 0 \ \text{ or } \ \theta_E = \theta = 0 \ \implies \ \psi_E = \psi. \tag{5.162}$$

In fact, the relationships between each of the pitch and roll components being zero for Euler angles and fused angles is summarised by

$$\phi_E = 0 \implies \phi = 0, \tag{5.163a}$$

$$\phi = 0 \implies \phi_E = 0 \text{ or } \pi, \tag{5.163b}$$

and

$$\theta_E = \theta = 0 \implies \phi_E = \begin{cases} \phi & \text{if } h = 1, \\ \text{wrap}(\pi - \phi) & \text{if } h = -1. \end{cases} \tag{5.164}$$

These relations mainly help in visualising or checking corner cases of certain rotation properties involving fused angles and Euler angles.

## 5.8 DISCUSSION

In this section, we discuss how the developed rotation representations relate to the requirements that were set out in Section 5.2, and provide application examples for them.

### 5.8.1 Rotation Representation Aims

As described in detail in Section 5.2, the work on 3D rotations in this chapter was motivated by the analysis and control of balancing bodies in 3D, and set out to develop new rotation representations, specifically the fused angles and tilt phase space representations, to address certain gaps in the field of existing rotation representations. A condensed summary of the aims that were listed for the new representations is as follows—the developed representations should:

**Aim 1**: Partition rotations into independent yaw and tilt rotation components that are then independently parameterised,

**Aim 2**: Quantify the amount of rotation in the **xy**, **xz** and **yz** major planes as scalar angular values that can meaningfully be referred to as 'yaw', 'pitch' and 'roll',

**Aim 3**: Define a notion of yaw that satisfies yaw additivity, i.e. global z-rotations should purely additively affect the yaw,

**Aim 4**: Define notions of pitch and roll that are concurrent, i.e. with no forced order of application that prioritises one of the two,

**Aim 5**: Have a tilt rotation component that encapsulates the heading-independent balance state of a robot, i.e. how far in any direction the robot is from being upright,

**Aim 6**: Have a tilt rotation component that has a direct correspondence to the set of possible accelerometer-measured gravity directions, and,

**Aim 7**: Given that $(\hat{\mathbf{e}}, \theta_a) \in \mathbb{A}$ is the axis-angle representation, have a tilt rotation component that has no $e_z$ component, and a yaw rotation component that is purely a function of $\theta_a$ and $e_z$.

As discussed and demonstrated throughout this chapter, all of these aims have been addressed by the fused angles and tilt phase space representations. The notion of fused yaw for example, presented in Section 5.4.1, quantifies the amount of rotation in the **xy** plane (Aim 2), satisfies yaw additivity (Aim 3), and explicitly partitions rotations into independently parameterised yaw and tilt rotation components, as required by Aim 1. The fused pitch and roll $(\theta, \phi)$, and phase roll and pitch $(p_x, p_y)$, are also two different definitions of pitch and roll that are concurrent (Aim 4), as in each case there is no delineable order of rotations, and that quantify the amount of rotation in the **xz** and **yz** major planes. It should be noted that due to the required yaw additivity and yaw/tilt parameter independence, the major planes being referred to in all instances are the major planes of frame {A}, as defined in Section 5.4.1 and Figure 5.5. This makes intuitive sense, as {A} is like an untilted and upright heading-local frame, meaning that rotations in the $\mathbf{x}_A\mathbf{z}_A$ and $\mathbf{y}_A\mathbf{z}_A$ planes really do correspond to heading-local sagittal and lateral rotations, as desired. The described nature of {A} also justifies why the tilt rotation component from {A} to {B} is a heading-independent balance state (Aim 5). The reason why there is a direct correspondence (Aim 6) between tilt rotations and the set of possible gravity directions as measured by a quasi-static accelerometer, i.e. ${}^B\mathbf{z}_G$, was discussed in Section 5.4.2.1. The final remaining aim, Aim 7, can be seen to be satisfied by the definition of fused yaw and tilt, by consideration of Equations (5.24b), (5.45) and (5.79).

The main differences between the fused angles and tilt phase space representations are magnitude axisymmetry (discussed later in Section 6.2.4.3), the way that rotations in the negative fused hemisphere are handled, and the optional ability to represent unbounded rotations in the case of the tilt phase space. The tilt phase space also facilitates tilt vector addition, a commutative and unambiguous way of concurrently adding tilt rotations, which finds use in multiple practical scenarios.

### 5.8.2  Application Examples

The rotation representations developed in this chapter can be applied, with advantages, in many scenarios. Uses of the developed representations within the scope of this thesis, or in associated open source software releases, have already been listed in Section 5.2.3. As a general example however, quadrotors for instance need to tilt in the direction they wish to accelerate, so a smooth and concurrent way of representing such tilt in terms of pitch and roll, using a suitable mix of tilt angles, fused angles, and in particular the tilt phase space, can be of great benefit. Similar arguments for the use of these representations also apply to the scenarios of balance and bipedal locomotion, where the tilt rotation component is particularly relevant, because as previously mentioned, it encapsulates the entire heading-independent balance state of the robot, with no extra component of rotation about the z-axis. Indeed, due also to the parallel between tilt rotations and the values measured by accelerometer sensors (Aim 6 above), tilt rotations can be seen to be a natural and therefore practically useful split of orientations into yaw and tilt.

Figure 5.1 illustrates how for the application of bipedal walking the fused angles $\psi$, $\theta$, $\phi$ and/or the tilt phase space parameters $p_x$, $p_y$, $p_z$ can be identified as the 'roll', 'pitch' and 'yaw', and used to independently quantify the amounts of rotation in the lateral, sagittal and horizontal (i.e. heading) planes of walking, respectively. This allows the motion, stability and state of balance to be measured and controlled separately in each of these three major directions of walking. Examples of gait stabilisation schemes that work in this way include the ones presented in Chapters 12 and 13, which both operate largely independently in the sagittal and lateral planes based on the fused pitch and fused roll orientation values. Another example is the balance feedback controller presented in Chapter 15, which based on the many advantageous properties of the tilt phase space, constructs numerous pure tilt rotation feedback components. It is in the nature of feedback controllers, e.g. PID-style controllers, to produce control inputs of arbitrary magnitude based on a set of gains, so the unbounded nature of the tilt phase space allows this to be handled in a clean way to full effect. The combining of various tilt rotation feedback components is also naturally handled by the tilt phase space via tilt vector addition.

On a lower level, the representations developed in this chapter have also been used, for example, in Chapter 14 for the constraint-based generation of gait trajectories. Most notably, the tilt phase space is used to separate the yaw and tilt of the feet at each so-called gait keypoint, and then to interpolate between them using the method for orientation cubic spline interpolation described later in Section 7.3.3.3. This ensures that the yaw and tilt profiles are individually exactly as intended in the final 3D foot trajectories, especially seeing as the yaw profiles come from the commanded step sizes, and the tilts come separately from the feedback controller. Tilt vector addition is also required, because multiple feedback paths need to contribute to each final foot tilt. The summed foot tilts are not guaranteed to be in range though, so the unbounded nature of the tilt phase space helps in being robust to 'wrapping around' prior to foot tilt saturation. As such, the presented work on rotations allows for the effective and robust generation of gait trajectories using such methods.

As a final note, it is reiterated that open source software libraries in both C++ (Allgeuer, 2018d) and Matlab (Allgeuer, 2018c) have been released to support the development and use of algorithms involving the tilt angles, fused angles and/or tilt phase space representations. Although these new representations are clearly the focus, it should be noted that the released libraries equally support all standard rotation representations as well, like the Euler angles, rotation matrix and quaternion representations.

## WHY NOT EULER ANGLES?

Three new rotation representations, all based on a novel way of partitioning 3D rotations into yaw and tilt components, were introduced in the previous chapter. Detailed explanations were given as to why the representations are required, and what properties they are desired to have (see Section 5.2). A condensed list of aims was given in Section 5.8.1. Despite hints having been given as to why it is the case, one question has so far remained however—

**Why are Euler angles not good enough for the job?**

This chapter aims to clearly and resolutely answer this question by presenting a comparative analysis between Euler angles, fused angles and the tilt phase space, in specific reference to the motivations and aims that were provided in the previous chapter, and why Euler angles do not fulfil them. Further problems and illogicalities of Euler angles are also presented, in explicit contrast to how the situation is different for fused angles and the tilt phase space.

The reasons why the other existing rotation representations, like for example quaternions, do not address our rotation representation needs is relatively clear from Section 5.2, and has already been discussed previously, e.g. at the end of Section 5.3.3. It should be clearly noted however, that rotation matrices and quaternions are of course nonetheless *very* useful for all kinds of computations and algorithms in this thesis, including ones that involve the new representations. For instance, the attitude estimator in Section 10.1 is based on the fused yaw, but formulated in terms of rotation matrices and quaternions, and the final estimated orientation is expressed in terms of fused angles and/or the tilt phase space in order to be useful for the various gait feedback controllers that use it.

As an explanatory introduction to this chapter, a video of an IROS conference spotlight talk has been included in Video 6.1. This video provides a brief recap of the situation with rotation representations, the aims we set out with, and how Euler angles tie into these.

### 6.1 EULER ANGLES CONVENTIONS

This chapter relies heavily on the definitions and results that were presented in Chapter 5. To understand the ensuing comparative analysis, it is important, for example, to be familiar with Section 5.4, and in particular the definitions and properties of the selected two Euler angles conventions, given in Section 5.3.5. One should recall that due to the reverse equivalence between intrinsic and extrinsic Euler angles, the convention that the $z$-axis points 'upwards', and the desire to express rotation components about all three principal axes, the only two relevant Euler angles conventions are the intrinsic ZYX and ZXY Euler angles conventions, given respectively by

$$\tensor[^G_B]{E}{} = (\psi_E, \theta_E, \phi_E) \in (-\pi, \pi] \times [-\tfrac{\pi}{2}, \tfrac{\pi}{2}] \times (-\pi, \pi] \equiv \mathbb{E}, \qquad (6.1a)$$

$$\tensor[^G_B]{\tilde{E}}{} = (\psi_{\tilde{E}}, \phi_{\tilde{E}}, \theta_{\tilde{E}}) \in (-\pi, \pi] \times [-\tfrac{\pi}{2}, \tfrac{\pi}{2}] \times (-\pi, \pi] \equiv \tilde{\mathbb{E}}. \qquad (6.1b)$$

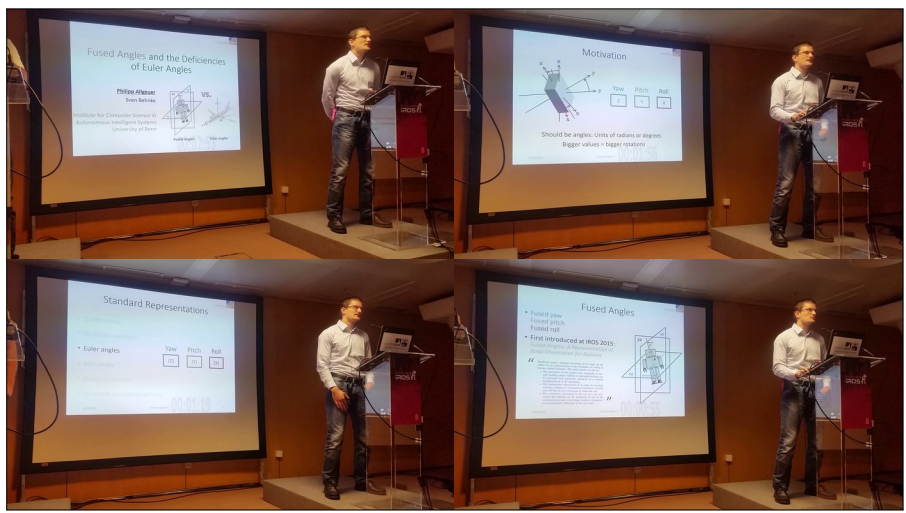

Video 6.1: A three minute spotlight talk at the 2018 International Conference on Intelligent Robots and Systems (IROS), for the paper "Fused Angles and the Deficiencies of Euler Angles". A brief overview of the motivation for a new rotation representation is given, and the main reasons why Euler angles are unsuitable are mentioned.
https://youtu.be/GVEdK0BuzG4
*3 Minute Spotlight Talk: Fused angles*

All problems or results that are derived for one of these two conventions can analogously be derived for the other, so without loss of generality, in this chapter we will mainly focus on the ZYX Euler angles convention, and refer to it ubiquitously as *the* 'Euler angles' representation. If the ZXY convention is needed, then it will be referred to explicitly as the 'ZXY Euler angles', with the 'ZXY Euler yaw', 'ZXY Euler roll' and 'ZXY Euler pitch' components, and the associated *tilded* notation will be used, as given in Equation (6.1b).

## 6.2   PROBLEMS WITH EULER ANGLES

There are many different problems with Euler angles, but before these are meticulously mathematically explored, here are some examples where one can intuitively tell[1] that something is not behaving as it really should:

---

1   In consideration of Section 5.8.1

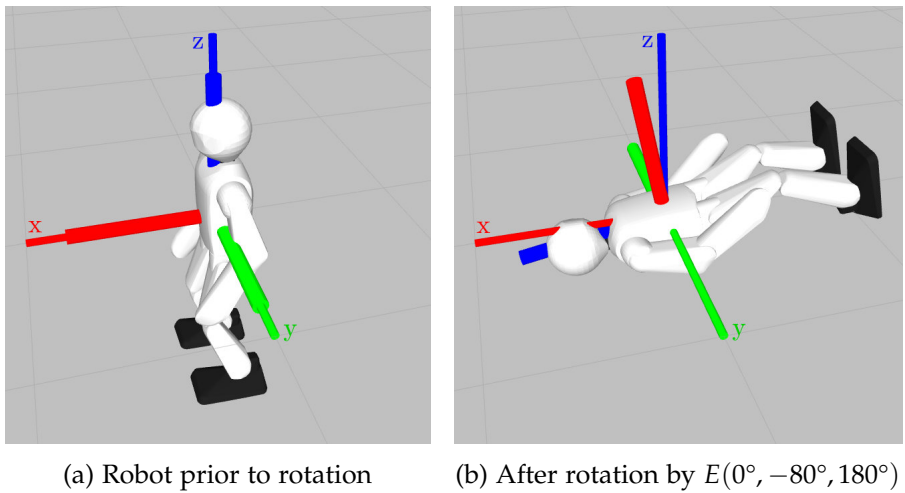

(a) Robot prior to rotation            (b) After rotation by $E(0°, -80°, 180°)$

Figure 6.1: Consider the rotation of a robot by the ZYX Euler angles rotation $E = (0°, -80°, 180°)$. If the robot is initially upright and facing to the left, as shown in (a), then the final pose of the robot is lying on its back and facing to the right, as shown in (b). Although the ZYX Euler yaw of this rotation is zero, clearly a vast change in the intuitive understanding of 'heading' has occurred, demonstrating a clear shortcoming of the ZYX Euler angle representation.

1. Consider a humanoid robot where the body-fixed z-axis points up through its head and the x-axis points forwards out of its chest, as shown in Figure 6.1a. If the yaw of a rotation is zero, then one would expect that no significant z-rotation or change in heading occurs. Nonetheless, when for example combining a Euler yaw of 0° with a Euler pitch of −80° and a Euler roll of 180°, the resulting rotation leaves the robot lying on its back facing the opposite direction to the one it started in, as shown in Figure 6.1b. The fact that a large change in heading has occurred is supported by the axis-angle representation of the rotation, which is $\pi$ radians about the vector $(0.766, 0, 0.643)$, which has a significant non-zero z-component.

2. Consider the same humanoid robot, with a pure pitch rotation of $\pm75°$ applied to it, so that it is close to lying on its front or its back. If a humanoid robot falls over and tries to get up again, these are not unexpected orientations to occur. If the heading-independent balance state were to be defined as the orientation of the robot with its Euler yaw removed (as opposed to its fused yaw), then one can observe that this balance state would be very unstable in situations like the one described. If the robot rotates just 10° locally about its z-axis, the Euler yaw would change by 34.3°. For pure pitch rotations of $\pm85°$, this increases to changes in Euler yaw of 45.1° for rotations about the z-axis of just 5°.

This sensitivity is clearly not desired within the normal working ranges of the robot.

3. While it is clear by assumption that the z-axis points 'upwards', along the direction of gravity, the choice of whether the x and y-axes point 'forwards' or 'sidewards', or in another diagonal direction, is essentially arbitrary. Thus, the concepts of pitch and roll, and the parameters expressing them, should behave analogously to each other, just in 90° differing directions. This is not the case at all for the Euler pitch and Euler roll parameters, as they do not even share the same domain ($[-\frac{\pi}{2}, \frac{\pi}{2}]$ vs. $(-\pi, \pi]$).

4. For ZYX Euler angles, the x-axis about which the final x-rotation is performed has a zero y-component relative to the initial yaw-rotated intermediate frame (i.e. the supposed 'heading-local' frame), but a non-zero z-component. This means that the Euler yaw and roll actually both contribute to the final heading of the robot, which is not desired. In fact, at the extremes when the Euler pitch is $\pm\frac{\pi}{2}$, the Euler yaw and roll become completely interchangeable, as demonstrated by Equation (5.32).

Summarised into a more generalised and abstract list, the core failings of the Euler angles representations are as follows:

A) **Singularities and Local Parameter Sensitivities:**
   The gimbal lock singularities are in close proximity to normal working ranges, making them hard to avoid, and leading to unwanted artefacts due to the increased local parameter sensitivities in widened neighbourhoods of the singularities.

B) **Mutual Independence of Rotation Parameters:**
   The Euler parameters have mutual interdependencies, leading to a mixed attribution of which parameters contribute to which major planes of rotation.

C) **Axisymmetry of Yaw:**
   The definition of the Euler yaw depends implicitly on axis projection (see Section 5.3.5.3), leading to unintuitive non-axisymmetric behaviour of the yaw angle.

D) **Axisymmetry of Pitch and Roll:**
   The definition of Euler pitch and roll requires a fundamental assumption of the order of elemental rotations, leading to non-axisymmetric definitions that do not correspond to each other in domain and/or behaviour.

Each of these core issues are investigated in detail in the following sections, with a specific focus on demonstrating why Euler angles have these problems, and fused angles and the tilt phase space do not.

Of the seven aims given in Section 5.8.1, Euler angles only satisfy two of them, Aim 3 and Aim 6. For instance, as described just above in Example 2, the problem of parameter sensitivities (Problem A) prevents the definition of a meaningful heading-independent balance state based on Euler pitch and roll, refuting Aim 5. The issues relating to axisymmetry (Problems C and D) also complicate or even violate Aims 2 and 4. The problem of parameter interdependencies (Problem B) further contribute to this, but also explain why Aims 1, 5, and 7 are not met.

### 6.2.1 Singularities and Local Parameter Sensitivities

We recall from Section 5.6 that it was shown by Stuelpnagel (1964) that it is topologically impossible to have a global one-to-one three-dimensional parameterisation of the rotation group without any singular points. That is, every three-dimensional parameterisation of the rotation space must have at least one of the following:

(i) A rotation that does not have a unique set of parameters,

(ii) A set of parameters that does not specify a unique rotation,

(iii) A rotation in the neighbourhood of which the sensitivity of the map from rotations to parameters is unbounded,

(iv) A set of parameters in the neighbourhood of which the sensitivity of the map from parameters to rotations is unbounded.

The Euler angles representation is singular at gimbal lock, i.e. when the Euler pitch $\theta_E = \frac{\pi}{2}$ or $-\frac{\pi}{2}$. Based in part on Equation (5.32), it can be seen that the Euler yaw $\psi_E$ and Euler roll $\phi_E$ both have essential discontinuities there. In reference to the characterisation of singularities given above, the singularities at both gimbal lock scenarios are of type (i) and (iii), for both $\psi_E$ and $\phi_E$. It is critical to compare this to the case for fused angles and the tilt phase space, which both only have the one fused yaw singularity in the parameter $\psi$, which occurs for $\alpha = \pi$. Thus, these two representations have a single singularity in a single parameter that does not affect the heading-independent balance state, while Euler angles have two singularities in two parameters, one of which directly affects it. As such, fused angles and the tilt phase space can represent heading-local states of balance completely without singularities, while this is not the case for Euler angles.

The fused yaw singularity is also 'maximally far' from the identity rotation, requiring a 180° tilt rotation in any direction to get there, while the two Euler angle singularities are only 90° away, which is close to, if not in, normal working ranges. In fact, the increased parameter sensitivity of the Euler yaw and roll near gimbal lock has noticeable effects even for tilt rotations of only 65° for instance. Sudden sensitive changes in Euler yaw and roll can occur if even small changes in orientation are made, even though the resulting rotation remains essentially pure pitch in nature. Suppose we consider pure pitch rotations of up to 90°, and calculate the sensitivity of the Euler yaw and roll parameters relative to small infinitesimal rotations about the local body-fixed z-axis. The result is shown in Figure 6.2, along with similarly calculated sensitivities for the fused yaw, fused roll and phase roll parameters. While the fused yaw maintains a perfect sensitivity of 1 for all pitch angles, as expected from Equation (5.141), and the fused roll and phase roll parameters slowly increase in sensitivity as the local z-rotation becomes closer to being a global x-rotation, the Euler yaw and roll can both be seen to have strongly divergent sensitivities that have noticeable effects for even moderate pitch rotations. Consequently, it can be seen that the Euler yaw component of a rotation cannot in general meaningfully be removed, as due to the highly sensitive Euler roll, even moderate tilts can experience large z-rotations in the rotation that remains, even though this should actually only be the contribution of 'pitch' and 'roll'.

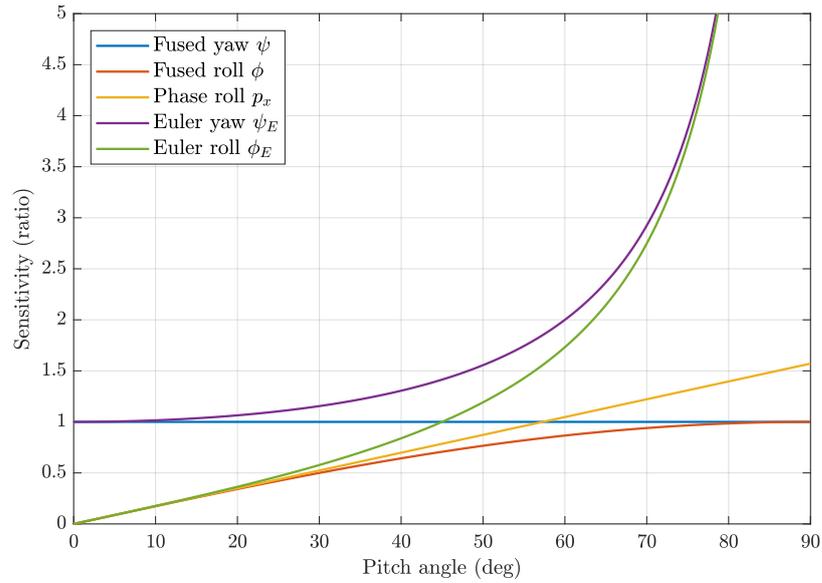

Figure 6.2: Plot of the sensitivities of the Euler angles, fused angles and tilt phase space parameters to infinitesimal local z-rotations at pure pitch orientations. Suppose $\beta \in [0, 90°]$ and we consider locally z-rotating (i.e. about its own z-axis) the pure pitch orientation $R_y(\beta)$. If the z-rotation is by some infinitesimal angle $\epsilon$, then the sensitivity of a parameter $X$ in describing $R_y(\beta)$ is given by $S_X(\beta) = |\frac{dX}{d\epsilon}|$. Clearly, high sensitivities as displayed by the Euler yaw and roll are undesirable, as for example, $S_{\phi_E}(\beta) = 3$ means that a small change in z-rotation causes a threefold change in the Euler roll $\phi_E$. We can observe that the fused angles and tilt phase space parameters have reasonable sensitivities everywhere, and in particular that the fused yaw sensitivity is perfectly unit magnitude for all $\beta$.

### 6.2.2 Mutual Independence of Rotation Parameters

In order to be able to fulfil many of the aims set out in Section 5.8.1, one necessary implicit condition is that the individual rotation parameters need to be as mutually independent as possible, and correspond intuitively and uniquely to the x, y and z-components of rotation. This is *not* the case for Euler angles, as is shown in the following subsections by direct comparison to the fused angles and tilt phase space representations.

#### 6.2.2.1 *Mutual Dependence of Yaw and Roll*

Consider the Euler angles rotation $E = (\psi_E, \theta_E, \phi_E) \in \mathbb{E}$ from a global frame {G} to the body-fixed frame {B}, consisting of a sequence of three rotations, namely about the z, y and x-rotations respectively. Let {E} denote the frame that results when {G} is initially rotated by the Euler yaw $\psi_E$. Ideally, as per Aim 5, {E} should be the reference frame for a Euler angles-based heading-independent balance state, i.e. the rotation component from {E} to {B}. One can quickly see however that this so-called balance state can actually have large z-rotation components in it, contravening that it is heading-independent in any intuitive sense. One obvious example of this has already been shown in Figure 6.1, where an example robot with zero Euler yaw can be observed to be intuitively facing the exact opposite direction to its reference identity pose. If we consider the axis of rotation of the Euler roll parameter (the third of the three elemental Euler rotations) in terms of the (supposedly) heading-local frame {E}, we can see that it is given by

$$^E\mathbf{x}_B = (\cos\theta_E,\, 0,\, -\sin\theta_E). \tag{6.2}$$

This is of course problematic, as the z-component is non-zero, meaning that the Euler roll rotation about this axis contributes to the heading of frame {B}. This effect can most clearly be seen when $|\sin\theta_E|$ is large, or in particular when

$$|\sin\theta_E| = 1 \iff \theta_E = \pm\tfrac{\pi}{2}, \tag{6.3}$$

which corresponds to gimbal lock, where Euler yaw and roll become completely interchangeable as given by Equation (5.32). The Euler pitch does not have this same problem, as relative to {E}, its axis of rotation is given by the pure

$$^E\mathbf{y}_E = (0, 1, 0). \tag{6.4}$$

Conceptually, what is happening is that part of the total 'yaw' of a rotation is always being quantified by the Euler roll parameter, in addition to the Euler yaw parameter, meaning that neither cleanly represents the component of rotation that they ideally should. The fused angles and tilt phase space representations do not have any such mutual dependencies, thanks in part to the fact that they rely on concurrent and not sequential definitions of pitch and roll. Importantly, this means that the heading-independent balance state defined by each representation (as expected) has no component of rotation about the z-axis, as indicated by Equation (5.54).

### 6.2.2.2  *Mutual Dependence of Pitch and Roll*

As the Euler pitch elemental rotation sequentially precedes the Euler roll one, the axis of rotation ${}^E\mathbf{x}_B$ of the latter is a function of $\theta_E$, as explicitly given in Equation (6.2). This creates a dependency of $\phi_E$ on $\theta_E$, which results in the Euler roll $\phi_E$ not completely capturing the intuitive sense of 'roll' all by itself. This can be seen in the bottom row of the rotation matrix in Equation (5.34), which corresponds to the z-vector

$$ {}^B\mathbf{z}_G = (-\sin\theta_E,\ \cos\theta_E \sin\phi_E,\ \cos\theta_E \cos\phi_E). \tag{6.5} $$

The unit vector ${}^B\mathbf{z}_G$, as discussed in Section 5.2.2, is a heading-independent measure of the global 'up' direction, just like a quasi-static accelerometer would measure the direction of gravity. While the x-component is a pure function of $\theta_E$, the y-component is not a function purely of $\phi_E$, as would naturally be desired. It can be seen for example from Equation (5.86) however, that

$$ {}^B\mathbf{z}_G = (-\sin\theta,\ \sin\phi,\ \cos\alpha), \tag{6.6} $$

so both of the aforementioned properties hold for fused angles. The situation is slightly more complicated for the tilt phase space, but from Equation (5.90), seeing as the phase roll and pitch parameters are just a rescaling of the sines of the fused angles parameters, an essentially similar result applies.

### 6.2.2.3  *Purity of the Axis of Rotation*

As discussed in Section 5.3.2, every 3D rotation can be expressed as a single rotation by some angle $\theta_a \in [0, \pi]$ about an axis $\hat{\mathbf{e}} = (e_x, e_y, e_z)$ in 3D space. We start by recalling from Equation (5.137) that the fused yaw is zero if and only if there is no component of rotation about the vertical z-axis, i.e.

$$ \psi = 0 \iff e_z = 0, \tag{6.7} $$

and observe from Equations (5.24b) and (5.138) that the fused yaw component of rotation $q_f$ is given by the renormalisation of

$$ \tilde{q}_f = \left(\cos\tfrac{\theta_a}{2},\ 0,\ 0,\ e_z \sin\tfrac{\theta_a}{2}\right). \tag{6.8} $$

It can consequently be seen that the fused angles and tilt phase space representations both satisfy Aim 7 (see Section 5.8.1). Conceptually, the interpretation of this result is that the notion of yaw used in both representations is intricately and purely linked to the z-component of the axis of rotation, in such a way that the former cleanly parameterises the latter. This also means that the remaining two pitch and roll parameters in each representation together cleanly parameterise the remaining two x and y-components of the axis of rotation. Neither result is true for the Euler angles representation, where for example the Euler angles rotation given in Example 1 on page 145 demonstrates that rotations free of Euler yaw can still have significant non-zero z-rotation components.

For each representation, we now look more carefully at the purity of the pitch and roll parameters in representing the x and y-components of the axis of rotation. Due to yaw additivity (Aim 3) and the desire for yaw/tilt parameter independence (Aim 1), it only makes sense to look at these components relative to the respective heading-local yaw-rotated intermediate frame, namely

- Frame {A} for the fused angles and tilt phase space representations, given by a yaw rotation of {G} by the fused yaw $\psi$,

- Frame {E} for the ZYX Euler angles representation, given by a yaw rotation of {G} by the ZYX Euler yaw $\psi_E$, and,

- Frame {Ẽ} for the ZXY Euler angles representation, given by a yaw rotation of {G} by the ZXY Euler yaw $\psi_{\tilde{E}}$.

This conforms with our previously stated expectation (Aim 2) that the pitch and roll values should quantify the amount of rotation in the **xz** and **yz** major planes of exactly these frames, respectively. Thus, we only need to examine and compare, for each representation, the nature of the axis of rotation for rotations that have zero yaw in that representation. Furthermore, as only the signed direction of the axis of rotation matters for the purpose of this discussion, we can consider any non-unit vector $\tilde{\mathbf{e}} = (\tilde{e}_x, \tilde{e}_y, \tilde{e}_z)$ defining that ray.

For fused angles and tilt phase space rotations with zero fused yaw, it can be deduced from Equations (5.54), (5.62) and (5.80) that the axes of rotation are respectively given by

$$\tilde{\mathbf{e}}_F = (\sin\phi, \sin\theta, 0), \tag{6.9a}$$

$$\tilde{\mathbf{e}}_P = (p_x, p_y, 0). \tag{6.9b}$$

On the other hand, from Equations (5.36) and (5.42), for ZYX Euler angles rotations with zero ZYX Euler yaw, and ZXY Euler angles rotations with zero ZXY Euler yaw, the axes of rotation are respectively given by

$$\tilde{\mathbf{e}}_E = (s_{\tilde{\phi}_E} c_{\tilde{\theta}_E}, \ c_{\tilde{\phi}_E} s_{\tilde{\theta}_E}, \ -s_{\tilde{\phi}_E} s_{\tilde{\theta}_E}), \tag{6.10a}$$

$$\tilde{\mathbf{e}}_{\tilde{E}} = (s_{\tilde{\phi}_{\tilde{E}}} c_{\tilde{\theta}_{\tilde{E}}}, \ c_{\tilde{\phi}_{\tilde{E}}} s_{\tilde{\theta}_{\tilde{E}}}, \ s_{\tilde{\phi}_{\tilde{E}}} s_{\tilde{\theta}_{\tilde{E}}}). \tag{6.10b}$$

For fused angles it can be seen that there is no component of rotation about the x and y-axes exactly when the fused roll $\phi$ and fused pitch $\theta$ are zero, respectively, and for the tilt phase space it can be seen that this occurs exactly when the phase roll $p_x$ and phase pitch $p_y$ are zero, respectively. For ZYX Euler angles however, the y-component is also zero when $\phi_E = \pi$, and for ZXY Euler angles, the x-component is also zero when $\theta_{\tilde{E}} = \pi$. This comes about because the $\tilde{e}_x$ and $\tilde{e}_y$ components are mixed expressions of Euler pitch and roll, instead of clean independent expressions like for fused angles and the tilt phase space, where direct one-to-one associations can be made between

$$\tilde{e}_x \longleftrightarrow \phi, p_x \tag{6.11a}$$

$$\tilde{e}_y \longleftrightarrow \theta, p_y \tag{6.11b}$$

It is also evident from the

$$\tilde{e}_z = \pm s_{\tilde{\phi}_E} s_{\tilde{\theta}_E} \tag{6.12}$$

components in Equation (6.10) that the Euler pitch and roll together contribute a component of rotation about the z-axis, as previously mentioned, which is unintuitive and indicates an impure contribution to the axis of rotation. In fact, seeing as Equation (6.10a) essentially encodes an arbitrary y-rotation followed by an x-rotation, and Equation (6.10b) encodes the exact reverse, it can be interpreted that the non-commutativity of these operations manifests itself in the opposite sign that results in the z-component $\tilde{e}_z$. Thus, as $\tilde{e}_z = 0$ for fused angles and the tilt phase space, these two new representations can conceptually be thought of as a concurrent way of combining x and y-rotations in a symmetrical and neutral way, somewhere exactly in between choosing the x-rotation to go first and choosing the y-rotation to go first.

### 6.2.2.4   *Purity of Rotation Inverses*

The fused yaw $\psi$, which is common to both the fused angles and tilt phase space representations, satisfies the remarkable negation through rotation inversion property, which asserts that the fused yaw of the inverse of a rotation is just the negative of its fused yaw, that is,

$$\psi_{inv} = -\psi. \tag{6.13}$$

This makes a lot of logical sense, as for example the inverse of an axis-angle pair $(\hat{\mathbf{e}}, \theta_a) \in \mathbb{A}$ is simply given by negating the axis of rotation $\hat{\mathbf{e}}$, and the inverse of a quaternion rotation is given by simply negating the x, y and z-components. It is also very reasonable to assert that if a particular rotation causes a change of heading in one direction, then the inverse rotation should cause the exact opposite change of heading in the other direction. This is however not the case for Euler yaw, as for example (in radians)

$$E = (\mathbf{1.2}, 0, 1.5) \implies E^{-1} = (-\mathbf{0.180}, -1.194, -1.378). \quad (6.14)$$

As a corollary to Equation (6.13), it can be seen that a rotation has no fused yaw component if and only if its inverse has no fused yaw component. This is once again quite logical, for similar reasons, but the Euler yaw does not satisfy this property, as can be seen from

$$E = (\mathbf{0}, 1.3, 0.6) \implies E^{-1} = (\mathbf{1.114}, -0.919, -1.198). \quad (6.15)$$

This is a discrepancy of more than one radian. The reason for such discrepancies can be found in Equation (5.145a), where it can be observed that the inverse Euler yaw actually depends on the values of both the Euler pitch and roll, demonstrating that the yaw parameter has mutual dependencies that it should not have. Even for rotations with zero Euler yaw, Equation (5.150) shows that all three inverse parameters are nonetheless mixed combinations of Euler pitch and roll, including notably the non-zero inverse Euler yaw. By comparison, it can be seen from Equations (5.149) and (5.155) that for rotations with zero fused yaw, the expressions for the inverse fused roll, fused pitch, phase roll and phase pitch parameters all trivially reduce to the negation through rotation inversion property. That is,

$$\psi = 0 \implies \begin{cases} \theta_{inv} = -\theta, & p_{xinv} = -p_x, \\ \phi_{inv} = -\phi, & p_{yinv} = -p_y. \end{cases} \quad (6.16)$$

### 6.2.3 Axisymmetry of Yaw

By convention (and without loss of generality), in this thesis the z-axis is chosen to point in the direction opposite to gravity. This, amongst other things, ensures that the concepts of 'roll', 'pitch' and in particular 'yaw', line up with what one would intuitively expect. The fixed choice of z-axis, however, still leaves one degree of rotational freedom open for the choice of the directions of the global x and y-axes, where we recall that these must obviously lie in the plane perpendicular to the z-axis, be perpendicular to each other, and satisfy the right-hand rule. No one choice is 'right' or 'wrong'—they are all equally valid and mathematically correct definitions of the global reference frame—so one would desire that any analysis of the orientation or balance of a body gives analogous results no matter which one is chosen. In the context of this chapter, the concept of parameter axisymmetry refers to the property that one or more rotation parameters are either:

(a) Invariant to the freedom of choice of x and y-axis, or,

(b) Vary in an intuitive rotational manner proportional to the choice.

In other words, axisymmetry refers to the notion that the rotation parameters, in order to be self-consistent, should be symmetrical about the unambiguously defined z-axis. This is a relatively natural property to desire, as, for example, the amount of yaw a rotation has should clearly transcend any arbitrary choice of which reference frame to use for analysis.

The fused yaw parameter is axisymmetric in the sense that it is invariant to the choice of global x and y-axes, i.e. type (a) axisymmetry. Consider a robot that is upright, and thereby considered to be in its identity orientation relative to the environment. If the robot undergoes any rotation, the above statement of fused yaw axisymmetry asserts that the fused yaw of this rotation is the same no matter what choice of reference global x and y-axis is made to numerically quantify the rotation. This is an important and reassuring property of the fused yaw as, given that the z-axis is unambiguously defined, any concept of yaw about the z-axis should clearly be a property of the actual physical rotation, not a property of some arbitrary choice of reference frame made solely for the purpose of mathematical analysis. This is not the case for Euler yaw however, as can easily be demonstrated as follows. Suppose we have a robot standing upright relative to the well-defined global z-axis, and suppose we define two different global reference frames:

- Frame {G₁}, such that relative to the identity pose of the robot the z-axis points upwards, the x-axis points forwards, and the y-axis points leftwards, and,

- Frame {G₂}, such that relative to the identity pose of the robot the z-axis points upwards, the y-axis points forwards and the x-axis points rightwards.

Both {G₁} and {G₂} are perfectly valid choices of reference frames that are consistent with the gravity-defined z-direction. If the robot now performs a 180° rotation about the horizontal axis 45° in between forwards and leftwards, as shown in Figure 6.3, the rotation quantified in terms of {G₁} has a Euler yaw of +90°, but the rotation quantified in terms of {G₂} has a Euler yaw of −90°, which is completely contradictory. This should not be, however, as the robot in both cases is performing exactly the same rotation relative to its environment. In terms of fused yaw, both quantifications of the rotation, i.e. relative to {G₁} and relative to {G₂}, have a yaw of zero.

### 6.2.3.1   *Mathematical Model of Yaw Axisymmetry*

We now develop a mathematical model of yaw axisymmetry and demonstrate that the fused yaw satisfies it, while the Euler yaw does not. Consider once again a robot that is upright relative to its environment, and that is in its identity orientation. Let {U} be a global coordinate frame such that $\mathbf{z}_U$ points in the direction opposite to gravity, as required, and suppose that a rotation is undergone by the robot that is numerically given by $_C^U R$, where {C} is a body-fixed frame that coincides with {U} prior to rotation. This is a fixed physical rotation of the robot relative to its environment, so it should have a unique well-defined fused yaw according to axisymmetry.

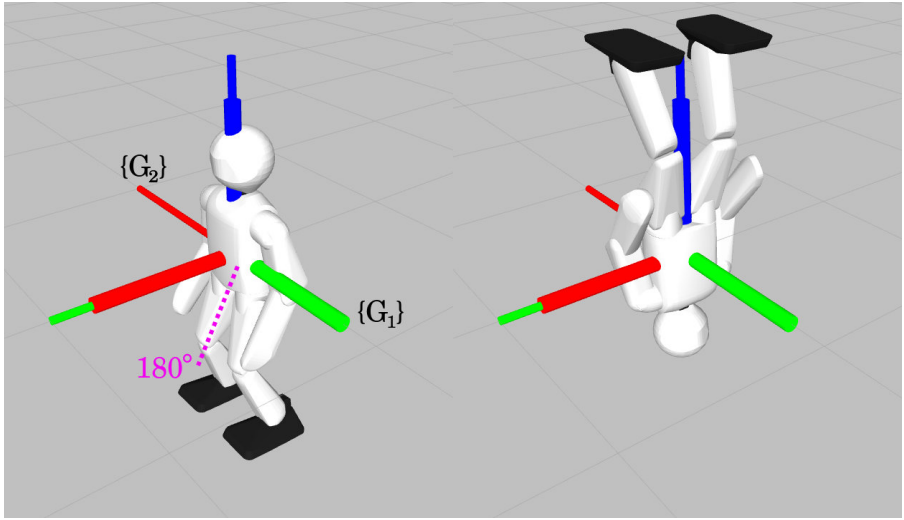

Figure 6.3: Visualisation of the 180° rotation of the robot described in the text, where prior to rotation the robot is upright (left), after rotation the robot is inverted (right), and the rotation occurs CCW about the dotted magenta line. The 180° rotation has a Euler yaw of +90° relative to {G₁} (thick axes), yet somewhat illogically has a Euler yaw of −90° relative to {G₂} (thin axes). Note that the x, y and z-axes are respectively red, green and blue, and do not rotate with the robot in the illustration.

As the z-axis is uniquely determined by the direction of gravity, every valid global coordinate system {G} that can be used as a reference frame to quantify $_C^U R$, including {U} itself, is a pure z-rotation of {U}. That is, given any valid global coordinate system {G},

$$_U^G R = R_z(\beta), \tag{6.17a}$$

$$_G^U R = R_z(-\beta), \tag{6.17b}$$

for some angle $\beta \in (-\pi, \pi]$. Given a frame {G}, a body-fixed frame {B} is attached to the robot in such a way that it coincides with {G} when the robot is initially upright, but rotates with the robot, as shown in Figure 6.4. Thus, the physical rotation undergone by the robot maps frames {U} to {C}, and {G} to {B}. As such, the rotation matrices $_C^U R$ and $_B^G R$ are simply two different ways of quantifying the exact same rotation, just with a different reference frame.

By definition, the rotation $_B^G R$ maps {U} onto {C}, so

$$_B^G R = _U^G R \, _C^U R \, _G^U R. \tag{6.18}$$

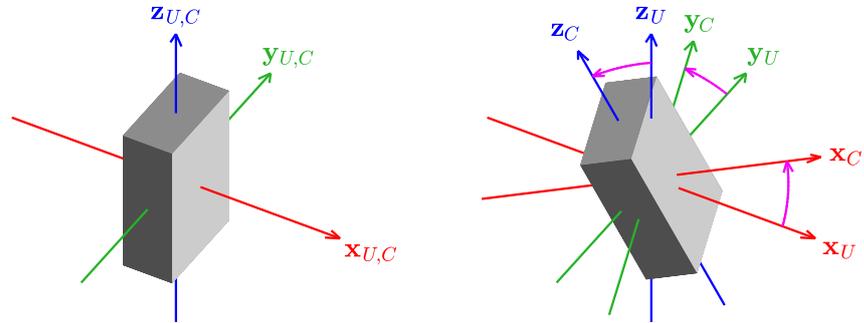

(a) Left: Prior to rotation the global frame {U} coincides with the local frame {C}, Right: Frame {C} rotates with the body, and quantifies the physical rotation relative to {U}.

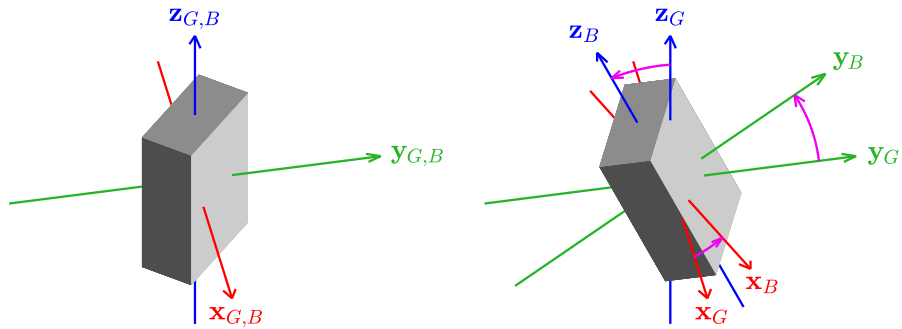

(b) Left: Prior to rotation the global frame {G} coincides with the local frame {B}, Right: Frame {B} rotates with the body, and quantifies the physical rotation relative to {G}.

Figure 6.4: Definition of frames for the investigation of parameter axisymmetry. Consider a body (represented here by a box) undergoing any physical rotation. The physical rotation can be modelled (a) as the rotation from {U} to {C}, or equivalently (b) as the rotation from {G} to {B}, where $\mathbf{z}_U \equiv \mathbf{z}_G$. Thus, $^U_C R$ and $^G_B R$ both numerically quantify the same physical rotation, despite being different rotation matrices. The axisymmetry property of fused yaw asserts that irrespective of the choice of {G}, the fused yaws of $^U_C R$ and $^G_B R$ are the same. As a result, every single physical rotation can unambiguously be assigned a fused yaw (irrespective of reference frame), which by contrast is not possible for Euler yaw.

This equation can be understood directly from the theory of *referenced rotations* (see Section 7.1.1.5), or from the observation that given any vector relative to {G}, applying the rotation $_B^G R$ is equivalent to transforming the vector to {U} coordinates using $_G^U R$, applying the rotation $_C^U R$ from {U} to {C}, and then transforming the result back to {G} coordinates using $_U^G R$. Consequently, from Equation (6.17),

$$_B^G R = R_z(\beta) \, _C^U R \, R_z(-\beta). \tag{6.19}$$

Taking the fused yaw $\Psi(\cdot)$ of both sides of Equation (6.19), and applying Equation (5.141) twice, gives

$$\begin{aligned}
\Psi(_B^G R) &= \Psi\big(R_z(\beta) \, _C^U R \, R_z(-\beta)\big) \\
&= \mathrm{wrap}\big(\beta + \Psi(_C^U R \, R_z(-\beta))\big) \\
&= \mathrm{wrap}\big(\beta + \Psi(_C^U R) - \beta\big) \\
&= \Psi(_C^U R). \tag{6.20}
\end{aligned}$$

We note that $\Psi(_C^U R)$ is clearly independent of $\beta$ and the choice of {G}, so $\Psi(_B^G R)$ must be independent as required. This demonstrates that the fused yaw of the physical rotation of the robot is invariant to the choice of reference global x and y-axis, as required, and therefore that the fused yaw satisfies type (a) parameter axisymmetry.

At this point, it can be demonstrated once again that Euler yaw violates parameter axisymmetry by considering, for example,

$$\begin{aligned}
_C^U R &= R_x(\tfrac{3\pi}{4}), \\
\beta &= \tfrac{\pi}{2}.
\end{aligned}$$

The Euler yaw of $_C^U R$ is clearly zero, as it consists of just a single counterclockwise (CCW) x-rotation by $\frac{3\pi}{4}$ radians, but from Equation (6.19),

$$\begin{aligned}
_B^G R &= R_z(\beta) \, _C^U R \, R_z(-\beta) \\
&= R_z(\tfrac{\pi}{2}) \, R_x(\tfrac{3\pi}{4}) \, R_z(-\tfrac{\pi}{2}) \\
&= \begin{bmatrix} -\frac{1}{\sqrt{2}} & 0 & \frac{1}{\sqrt{2}} \\ 0 & 1 & 0 \\ -\frac{1}{\sqrt{2}} & 0 & -\frac{1}{\sqrt{2}} \end{bmatrix}.
\end{aligned}$$

We therefore conclude that

$$\begin{aligned}
_B^G R &= R_y(\tfrac{3\pi}{4}) \\
&= E_R(\pi, \tfrac{\pi}{4}, \pi),
\end{aligned}$$

where $E_R(\cdot, \cdot, \cdot)$ is notation for the rotation matrix corresponding to the enclosed Euler angles parameters. The Euler yaw of $_B^G R$ is thus $\pi$ radians, which is totally different to the (zero) Euler yaw of $_C^U R$, despite the fact that both rotation matrices numerically quantify the same physical rotation. This proves that the Euler yaw cannot be axisymmetric.

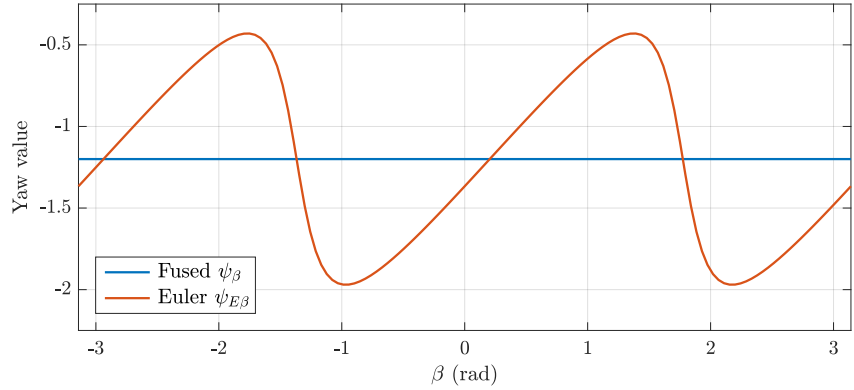

Figure 6.5: Plots of fused yaw and Euler yaw against $\beta$ for the determination of parameter axisymmetry. A $\beta$ of zero corresponds to the chosen physical rotation of $^{U}_{C}R = F_R(-1.2,\ 0.2,\ -1.3,\ 1)$. In contrast to the irregular nature of the Euler yaw, the constant nature of the fused yaw exemplifies its type (a) axisymmetry. That is, no matter what valid global reference frame is chosen to quantify $^{U}_{C}R$, the fused yaw of the rotation is the same.

### 6.2.3.2  *Visualising Yaw Axisymmetry*

As can be observed from Equation (6.17), the set of all possible choices of global reference x and y-axes corresponds directly to the set of all possible choices of $\beta$, which is in the range $(-\pi, \pi]$. Thus, given a particular physical rotation of the robot, one can plot the yaw that the rotation has relative to all possible valid global reference frames, by plotting the yaw it has for all possible values of $\beta$. This has been done, for example, in Figure 6.5 for the physical rotation $^{U}_{C}R = F_R(-1.2,\ 0.2,\ -1.3,\ 1)$, where $F_R(\cdot, \cdot, \cdot, \cdot)$ is notation for the rotation matrix corresponding to the enclosed fused angles parameters. It can clearly be seen that while the fused yaw remains constant for all $\beta$, the Euler yaw varies greatly and quite irregularly. In fact, if the same physical rotation but with $h = -1$ had been chosen, the Euler yaw would be seen to take on all possible values from $-\pi$ to $\pi$ as $\beta$ varies.

A similar observation can be made in Figure 6.6a, where surface plots of the Euler yaw and fused yaw parameters are provided for the case that an upright robot is rotated away from the vertical z-axis in every possible direction, about the vectors in the **xy** plane. Put into other words, Figure 6.6a plots, as surfaces, the Euler yaw and fused yaw parameters for all tilt rotations up to $\frac{\pi}{2}$ radians in magnitude. The x and y-coordinates of the surfaces are the x and y-components of the rotation vector representations of the respective tilt rotations. While the fused yaw demonstrates axisymmetric, and in fact constant, behaviour about the gravity-defined z-axis, the Euler yaw behaves differently depending on the direction in which the robot is tilted away from the z-axis. While requiring the surface plots in Figure 6.6a to be rotationally symmetric about the z-axis may at first seem like a different definition of axisymmetry than the one used so far, one can quickly see that plotting the yaw parameters of all tilt rotations (of a particular magnitude) with respect to a single global reference frame is in fact mathematically equivalent to plotting the yaw parameters of a single (physical) tilt rotation with respect to all possible global reference frames. This comes about because both the set of all tilt rotations of a particular magnitude, and the set of all global reference frames, are generated by z-rotational changes of basis of any one single element.

### 6.2.4    Axisymmetry of Pitch and Roll

To investigate the axisymmetry of pitch and roll for the Euler angles, fused angles and tilt phase space representations, we consider the same situation as described in Section 6.2.3.1, and illustrated in Figure 6.4. That is, suppose an upright robot undergoes a physical rotation ${}^U_C R$ relative to one particular global reference frame {U}, and that {G} is any other global reference frame such that ${}^G_B R$ models the exact same rotation. As the z-axes of both {U} and {G} are constrained by convention to point in the same opposite direction to gravity, the two frames are just separated by a single z-rotation $R_z(\beta)$, so one can write

$$
{}^G_U R = R_z(\beta), \tag{6.21}
$$

where $\beta$ is a scalar angle in the range $(-\pi, \pi]$. Given this mathematical framework, the definition of parameter axisymmetry (see page 153) for any arbitrary rotation parameter can be restated as the property that the parameter, when calculated of ${}^G_B R$, is either:

(a) Completely independent of $\beta$, or,

(b) Varies in a rotational manner with respect to $\beta$.

We start by introducing the following notation for the Euler angles, tilt angles, fused angles and tilt phase space representations of the fixed physical rotation ${}^U_C R$:

$$
{}^U_C E = (\psi_{E0}, \theta_{E0}, \phi_{E0}), \tag{6.22a}
$$

$$
{}^U_C T = (\psi_0, \gamma_0, \alpha_0), \tag{6.22b}
$$

$$
{}^U_C F = (\psi_0, \theta_0, \phi_0, h_0), \tag{6.22c}
$$

$$
{}^U_C P = (p_{x0}, p_{y0}, \psi_0), \tag{6.22d}
$$

and similarly define the following notation for the same various representations of ${}^G_B R$:

$$
{}^G_B E = (\psi_{E\beta}, \theta_{E\beta}, \phi_{E\beta}), \tag{6.23a}
$$

$$
{}^G_B T = (\psi_\beta, \gamma_\beta, \alpha_\beta), \tag{6.23b}
$$

**Euler yaw $\psi_E$**    **Fused yaw $\psi$**

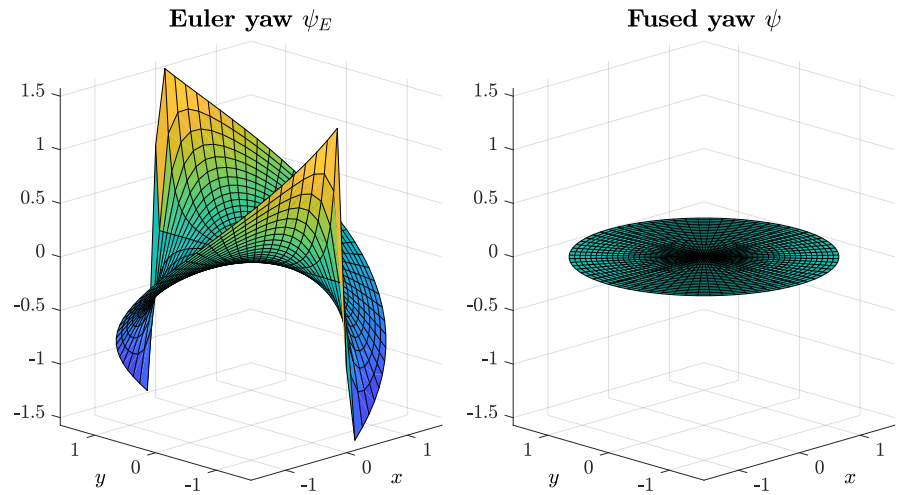

(a) 3D plots of Euler yaw and fused yaw relative to $(x, y) = (\alpha c_\gamma, \alpha s_\gamma)$.

**Euler pitch $\theta_E$**    **Euler roll $\phi_E$**

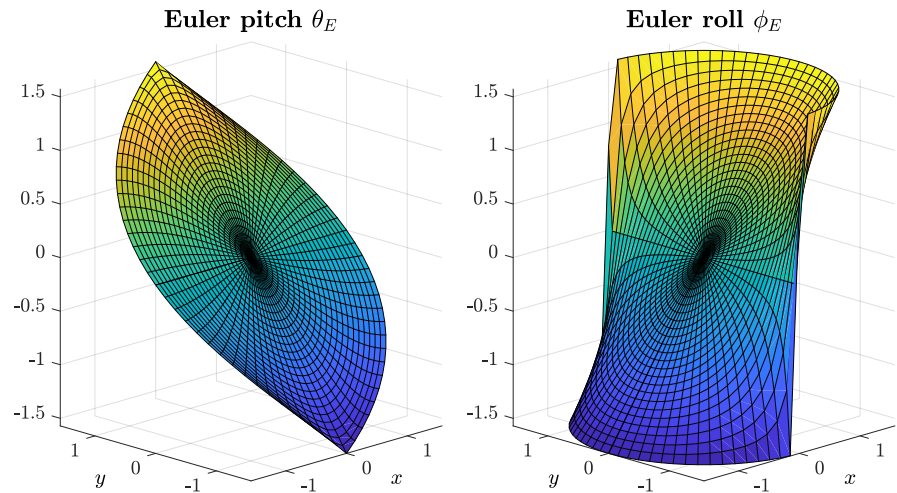

(b) 3D plots of Euler pitch and roll relative to $(x, y) = (\alpha c_\gamma, \alpha s_\gamma)$.

Figure 6.6: 3D plots of the Euler angles, fused angles and tilt phase space parameters of all pure tilt rotations up to $\frac{\pi}{2}$ in magnitude (continued in Figure 6.7). The parameters are plotted relative to the rotation vector coordinates of the associated tilt rotations, namely $(x, y) = (\alpha \cos \gamma, \alpha \sin \gamma)$. Although it may at first seem unrelated to parameter axisymmetry, plotting all tilt rotations of a particular magnitude $\alpha$ is equivalent to plotting any one such tilt rotation relative to all different possible global reference frames (i.e. all different $\beta$). Thus, for instance the rotational symmetry seen in the fused yaw plot is indicative of its type (a) axisymmetry (see page 153), while the lack of rotational symmetry in the Euler yaw plot is indicative of its lack of axisymmetry. The Euler pitch and Euler roll plots behave similarly in nature for small $\|(x, y)\| = \alpha$, but as the tilt rotations get larger, the Euler pitch and roll start to behave completely differently. For values of $\alpha$ beyond $\frac{\pi}{2}$ (beyond the pictured domain), $|\phi_E|$ even starts to exceed $\frac{\pi}{2}$ and takes on values up to $\pi$. This does not happen for the Euler pitch at all.

**Fused pitch** $\theta$        **Fused roll** $\phi$

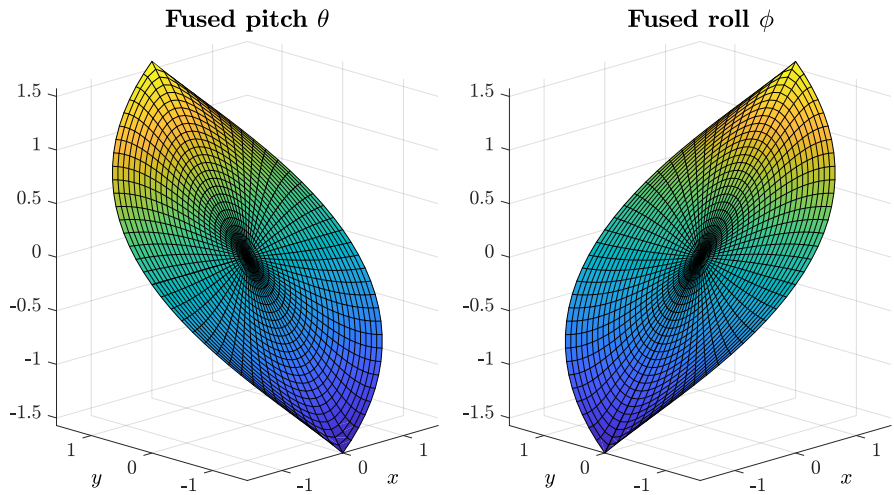

(a) 3D plots of fused pitch and roll relative to $(x, y) = (\alpha c_\gamma, \alpha s_\gamma)$.

**Phase pitch** $p_y$        **Phase roll** $p_x$

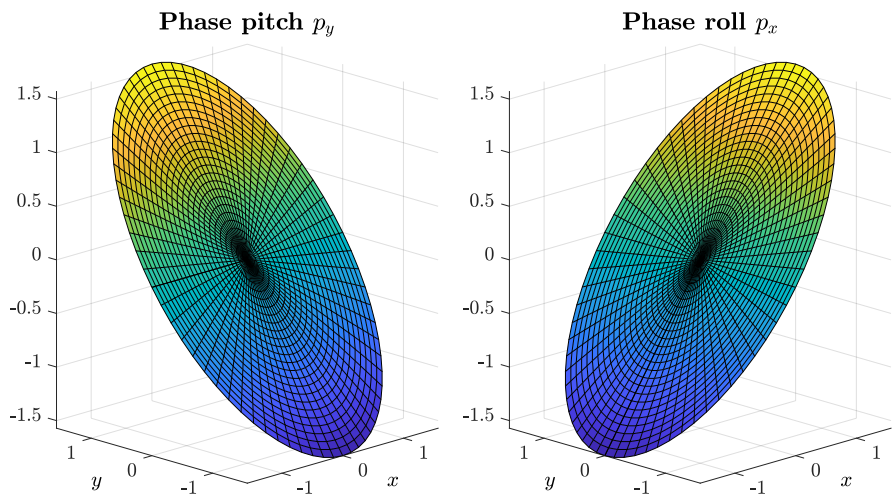

(b) 3D plots of phase pitch and roll relative to $(x, y) = (\alpha c_\gamma, \alpha s_\gamma)$.

Figure 6.7: Continuation of Figure 6.6. In contrast to the lack of correspond-ence in behaviour between the Euler pitch and roll parameters (see Figure 6.6b), the fused pitch and roll (and phase pitch and roll) parameters do clearly visibly behave identically to each other. The perfect planar circular discs that can be seen in (b) for the tilt phase space are indicative of its type (b) axisymmetry (see page 153). Refer to Visualisation C in Section 6.2.4.5 for further discussion of this figure.

$$
^G_B F = (\psi_\beta,\, \theta_\beta,\, \phi_\beta,\, h_\beta), \tag{6.23c}
$$

$$
^G_B P = (p_{x\beta},\, p_{y\beta},\, \psi_\beta). \tag{6.23d}
$$

Given this notation, the result, for example, that the tilt angle parameter $\alpha$ is type (a) axisymmetric can be proven by demonstrating that $\alpha_{\hat{\beta}}$ is independent of $\beta$, i.e. that it is a function only of the 'zero' variables $*_0$ in Equation (6.22). We already know from Section 6.2.3.1 that the fused yaw is invariant to the choice of $\beta$, so simply restating Equation (6.20) in terms of the new notation gives our first result:

$$
\psi_\beta = \psi_0. \tag{6.24}
$$

This is the mathematical embodiment of the type (a) axisymmetry of the fused yaw parameter. We now continue through all the other representations and rotation parameters, and demonstrate the type (a) or type (b) axisymmetry of each, except for the Euler angles parameters, which are demonstrated to be non-axisymmetric. The previous result, that the Euler yaw is *not* axisymmetric, can be mathematically inferred from the fact that, up to angle wrapping,

$$
\psi_{E\beta} = \psi_{E0} + \operatorname{atan2}(B, A), \tag{6.25}
$$

where

$$
A = c_\beta^2 c_{\theta_E} + s_\beta^2 c_{\phi_E} - c_\beta s_\beta s_{\theta_E} s_{\phi_E}, \tag{6.26a}
$$

$$
B = c_\beta s_\beta (c_{\theta_E} - c_{\phi_E}) - s_\beta^2 s_{\theta_E} s_{\phi_E}. \tag{6.26b}
$$

Clearly, the right-hand side of Equation (6.25) is a function of $\beta$.

### 6.2.4.1  *Axisymmetry of Tilt Angles*

From Equation (6.22b) and the conversion equation from tilt angles to rotation matrices given in Equation (5.76), we have that

$$
^U_C R = \begin{bmatrix} \cdot & \cdot & \cdot \\ \cdot & \cdot & \cdot \\ -s_{\alpha_0} s_{\gamma_0} & s_{\alpha_0} c_{\gamma_0} & c_{\alpha_0} \end{bmatrix}, \tag{6.27}
$$

where the dotted entries have been omitted for brevity. Similarly, from Equation (6.23b), we have that

$$
^G_B R = \begin{bmatrix} \cdot & \cdot & \cdot \\ \cdot & \cdot & \cdot \\ -s_{\alpha_\beta} s_{\gamma_\beta} & s_{\alpha_\beta} c_{\gamma_\beta} & c_{\alpha_\beta} \end{bmatrix}. \tag{6.28}
$$

Based on Equation (6.19) however, we also know that

$$\begin{aligned}
{}_B^G R &= R_z(\beta) \, {}_C^U R \, R_z(-\beta) \\
&= \begin{bmatrix} c_\beta & -s_\beta & 0 \\ s_\beta & c_\beta & 0 \\ 0 & 0 & 1 \end{bmatrix} \begin{bmatrix} \cdot & \cdot & \cdot \\ \cdot & \cdot & \cdot \\ -s_{\alpha_0} s_{\gamma_0} & s_{\alpha_0} c_{\gamma_0} & c_{\alpha_0} \end{bmatrix} \begin{bmatrix} c_\beta & s_\beta & 0 \\ -s_\beta & c_\beta & 0 \\ 0 & 0 & 1 \end{bmatrix} \\
&= \begin{bmatrix} \cdot & \cdot & \cdot \\ \cdot & \cdot & \cdot \\ -c_\beta s_{\alpha_0} s_{\gamma_0} - s_\beta s_{\alpha_0} c_{\gamma_0} & c_\beta s_{\alpha_0} c_{\gamma_0} - s_\beta s_{\alpha_0} s_{\gamma_0} & c_{\alpha_0} \end{bmatrix}.
\end{aligned} \tag{6.29}$$

So by comparing entries in Equations (6.28) and (6.29), we know that

$$s_{\alpha_\beta} s_{\gamma_\beta} = s_{\alpha_0}(c_\beta s_{\gamma_0} + s_\beta c_{\gamma_0}), \tag{6.30a}$$

$$s_{\alpha_\beta} c_{\gamma_\beta} = s_{\alpha_0}(c_\beta c_{\gamma_0} - s_\beta s_{\gamma_0}), \tag{6.30b}$$

$$c_{\alpha_\beta} = c_{\alpha_0}. \tag{6.30c}$$

As the $\cos(\cdot)$ function is one-to-one on the domain $[0, \pi]$, by direct consequence of Equation (6.30c),

$$\alpha_\beta = \alpha_0, \tag{6.31}$$

which demonstrates that the tilt angle $\alpha$ is type (a) axisymmetric. This insight also allows Equations (6.30a) and (6.30b) to be simplified to

$$\begin{bmatrix} \cos \gamma_\beta \\ \sin \gamma_\beta \end{bmatrix} = \begin{bmatrix} c_\beta & -s_\beta \\ s_\beta & c_\beta \end{bmatrix} \begin{bmatrix} \cos \gamma_0 \\ \sin \gamma_0 \end{bmatrix}. \tag{6.32}$$

By identifying the middle matrix as a 2D rotation matrix that rotates a vector CCW by an angle of $\beta$, one can deduce that, up to angle wrapping,

$$\gamma_\beta = \gamma_0 + \beta. \tag{6.33}$$

This expression, in particular in combination with Equation (6.32), is the mathematical embodiment of the type (b) axisymmetry of the tilt axis angle parameter $\gamma$, as it can clearly be seen that $\gamma$ varies in an intuitive rotational manner with respect to $\beta$. Thus, it can be concluded that all three tilt angles parameters are axisymmetric.

### 6.2.4.2 *Axisymmetry of Fused Angles*

The fused pitch and roll are axisymmetric in the sense that their sine ratios, $\sin \theta$ and $\sin \phi$, circumscribe a uniform circle as a function of the choice of global reference x and y-axes, i.e. as a function of $\beta$. That is, the locus of $(\sin \phi, \sin \theta)$ over all possible choices of $\beta$ is a circle, and this circle is traversed uniformly as the choice varies. This corresponds to type (b) axisymmetry, and as demonstrated later, Euler angles do not satisfy this property.

Similar to Equations (6.27) and (6.28), the rotation matrices ${}^U_C R$ and ${}^G_B R$ can, from Equation (5.85), be written as

$$
{}^U_C R = \begin{bmatrix} \cdot & \cdot & \cdot \\ \cdot & \cdot & \cdot \\ -s_{\theta_0} & s_{\phi_0} & c_{\alpha_0} \end{bmatrix}, \tag{6.34a}
$$

$$
{}^G_B R = \begin{bmatrix} \cdot & \cdot & \cdot \\ \cdot & \cdot & \cdot \\ -s_{\theta_\beta} & s_{\phi_\beta} & c_{\alpha_\beta} \end{bmatrix}. \tag{6.34b}
$$

From Equation (6.19) however, ${}^G_B R$ can alternatively be written as

$$
\begin{aligned}
{}^G_B R &= R_z(\beta)\, {}^U_C R\, R_z(-\beta) \\
&= \begin{bmatrix} c_\beta & -s_\beta & 0 \\ s_\beta & c_\beta & 0 \\ 0 & 0 & 1 \end{bmatrix} \begin{bmatrix} \cdot & \cdot & \cdot \\ \cdot & \cdot & \cdot \\ -s_{\theta_0} & s_{\phi_0} & c_{\alpha_0} \end{bmatrix} \begin{bmatrix} c_\beta & s_\beta & 0 \\ -s_\beta & c_\beta & 0 \\ 0 & 0 & 1 \end{bmatrix} \\
&= \begin{bmatrix} \cdot & \cdot & \cdot \\ \cdot & \cdot & \cdot \\ -c_\beta s_{\theta_0} - s_\beta s_{\phi_0} & c_\beta s_{\phi_0} - s_\beta s_{\theta_0} & c_{\alpha_0} \end{bmatrix}.
\end{aligned} \tag{6.35}
$$

Comparing Equations (6.34b) and (6.35) yields

$$
s_{\phi_\beta} = c_\beta s_{\phi_0} - s_\beta s_{\theta_0}, \tag{6.36a}
$$

$$
s_{\theta_\beta} = c_\beta s_{\theta_0} + s_\beta s_{\phi_0}, \tag{6.36b}
$$

which leads to the matrix equation

$$
\begin{bmatrix} \sin \phi_\beta \\ \sin \theta_\beta \end{bmatrix} = \begin{bmatrix} c_\beta & -s_\beta \\ s_\beta & c_\beta \end{bmatrix} \begin{bmatrix} \sin \phi_0 \\ \sin \theta_0 \end{bmatrix}. \tag{6.37}
$$

By once again identifying the middle matrix as a 2D rotation matrix that rotates a vector CCW by an angle of $\beta$, one can see that this equation is *exactly* the mathematical statement of the type (b) axisymmetry of fused pitch and roll. Furthermore, seeing as the fused hemisphere parameter $h$ is just the sign of $\cos \alpha$, from Equation (6.30c) we can deduce that

$$
h_\beta = h_0, \tag{6.38}
$$

i.e. that the fused hemisphere parameter is type (a) axisymmetric. Thus, it can be concluded that all four fused angles parameters, including of course the fused yaw, are axisymmetric.

The axisymmetry of the fused pitch and roll can be interpreted and visualised by viewing the respective sine ratios as quadrature sinusoid components of the associated rotation. The fused pitch $\theta$ and fused roll $\phi$ come together with the fused hemisphere $h$ to define the tilt rotation component of a rotation. The magnitude of this tilt rotation is given by the tilt angle $\alpha$, and the relative direction is given by the tilt axis angle $\gamma$. The fused pitch and roll can be thought of as a way of 'splitting up' the action of $\alpha$ into orthogonal components, much like a vector can be resolved into components relative to a coordinate frame. More precisely, the sine ratios $\sin\phi$ and $\sin\theta$ are in fact a decomposition of $\sin\alpha$ into quadrature sinusoid components, i.e. sinusoid components that are exactly 90° out of phase, and trace out a uniform circle parametrically. This quadrature nature is exemplified by the equations

$$\sin\phi = \sin\alpha\cos\gamma, \tag{6.39a}$$

$$\sin\theta = \sin\alpha\sin\gamma, \tag{6.39b}$$

and

$$\sin\alpha = \sqrt{\sin^2\phi + \sin^2\theta}, \tag{6.40a}$$

$$\gamma = \operatorname{atan2}(\sin\theta, \sin\phi), \tag{6.40b}$$

which are strongly resemblant of the standard equations that relate Cartesian coordinates to polar coordinates. The property of axisymmetry of the fused pitch and roll is equivalent to stating that the choice of global reference x and y-axis simply results in a fixed phase shift to the quadrature components, i.e. as suggested by Equation (6.33). This suggests that the nature of fused pitch and roll in expressing a rotation is a property of the actual physical rotation, not whatever arbitrary reference frame is chosen to numerically quantify it.

The uniform circular nature of the fused pitch and roll sine ratios, and how these vary axisymmetrically in a purely rotational manner with respect to $\beta$, can be seen in Figure 6.8. The locus of the sine ratio pair $(\sin\phi_\beta, \sin\theta_\beta)$ is plotted for a particular physical rotation $^U_CF$ as $\beta$ varies, and the quadrature decomposition of $\sin\alpha$ into $\sin\phi$ and $\sin\theta$ components is illustrated graphically by the orange triangle. The result that the value of $\beta$ directly additively affects the tilt axis angle $\gamma$, as stated in Equation (6.33), can also be observed.

### 6.2.4.3 *Axisymmetry of the Tilt Phase Space*

The phase roll $p_x$ and phase pitch $p_y$ are type (b) axisymmetric rotation parameters. This can most easily be seen from the axisymmetry of the tilt angles parameters, where from Equations (6.31) and (6.33) we can see that

$$p_{x\beta} = \alpha_\beta\cos\gamma_\beta = \alpha_0\cos(\gamma_0 + \beta) = \alpha_0(c_\beta c_{\gamma_0} - s_\beta s_{\gamma_0}), \tag{6.41a}$$

$$p_{y\beta} = \alpha_\beta\sin\gamma_\beta = \alpha_0\sin(\gamma_0 + \beta) = \alpha_0(s_\beta c_{\gamma_0} + c_\beta s_{\gamma_0}). \tag{6.41b}$$

This can be simplified and factored into the matrix equation

$$\begin{bmatrix} p_{x\beta} \\ p_{y\beta} \end{bmatrix} = \begin{bmatrix} c_\beta & -s_\beta \\ s_\beta & c_\beta \end{bmatrix}\begin{bmatrix} p_{x0} \\ p_{y0} \end{bmatrix}. \tag{6.42}$$

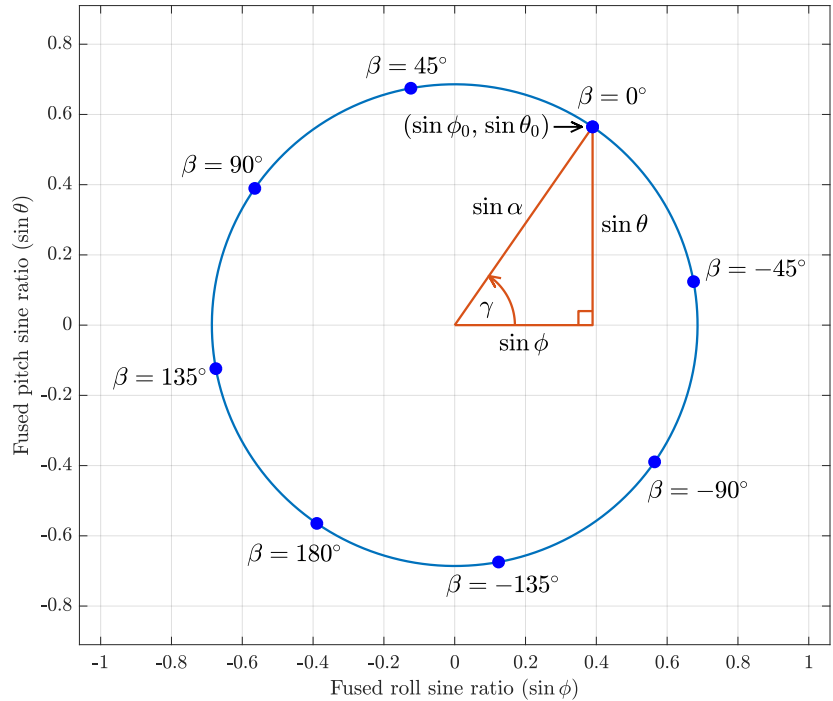

Figure 6.8: Locus of the fused roll and pitch sine ratios $(\sin\phi_\beta, \sin\theta_\beta)$ as $\beta$ varies, i.e. for all possible choices of global reference x and y-axes, for the physical rotation ${}^U_G F = F(2.0, 0.6, 0.4, 1)$. Points in $\beta$-increments of 45° are labelled, and illustrate how $\beta$ can be seen to be a positive offset to the tilt axis angle $\gamma$, as in Equation (6.33). The orange triangle demonstrates the decomposition of $\sin\alpha$ into the quadrature sinusoid components $\sin\phi$ and $\sin\theta$.

Similar to what Equation (6.37) did for the fused pitch and roll, this matrix equation epitomises the type (b) axisymmetry of the tilt phase space pitch and roll parameters. The middle matrix can clearly be identified as a 2D rotation matrix that encodes a CCW rotation by $\beta$, meaning that $(p_{x\beta}, p_{y\beta})$ varies in an intuitive rotational manner with respect to $\beta$, as required. As a result, it can be concluded that all three tilt phase space parameters are axisymmetric, as the third parameter, $p_z$, is just the fused yaw.

Like for fused angles, the tilt phase space pitch and roll parameters can also be interpreted to be quadrature sinusoid components that together describe the tilt rotation component. The phase pitch and roll parameters can be thought of as the most direct way of 'splitting up' the magnitude $\alpha$ of a tilt rotation into orthogonal components. These components correspond to quadrature sinusoids that are exactly 90° out of phase, and trace out a uniform circle parametrically. Similar to Equations (6.39) and (6.40) for fused angles, the quadrature nature of the phase pitch and roll is highlighted by the equations

$$p_x = \alpha \cos \gamma, \tag{6.43a}$$

$$p_y = \alpha \sin \gamma, \tag{6.43b}$$

and their inverse relations

$$\alpha = \sqrt{p_x^2 + p_y^2}, \tag{6.44a}$$

$$\gamma = \text{atan2}(p_y, p_x). \tag{6.44b}$$

Due to these relations, as well as Equation (6.33), the property of axisymmetry for the phase roll and pitch can be seen to be equivalent to the observation that the choice of global reference x and y-axis simply results in a fixed phase shift to the quadrature components. This observation is illustrated more clearly later in Figure 6.11, as well as in Figure 6.9, which plots the locus of the tilt phase parameters $(p_x, p_y)$ for a particular physical rotation as $\beta$ varies. In the latter figure, the quadrature decomposition of $\alpha$ into $p_x$ and $p_y$ parameters is also depicted by means of an orange triangle, and the additivity of $\beta$ to the tilt axis angle $\gamma$ can be surmised from the $\beta$ labels.

It is evident by now that fused angles and the tilt phase space share many similar positive properties. One property that the tilt phase space has, but the fused angles representation does not, is **magnitude axisymmetry**. This relates to the fact that equal angle magnitude rotations in any tilt direction (i.e. away from the z-axis) have equal norms in the 2D tilt phase space, and that this norm is directly proportional to the magnitude of tilt rotation. While the first of these properties holds for the fused angles sine ratios $(\sin \phi, \sin \theta)$, the latter does not, as the norm in the sine ratio space starts changing more slowly the closer a tilt rotation gets to a magnitude of 90°. Magnitude axisymmetry is an important property, for example in cases of rotations or orientations being used for the purposes of feedback control. In such cases, for control theoretical reasons it is often desired that rotations that are, for instance, twice as large provide exactly twice as much contribution to the feedback error term.

### 6.2.4.4 *Non-axisymmetry of Euler Angles*

To demonstrate that the Euler pitch and roll parameters are non-axisymmetric, we first turn to a simple example. Suppose we have a robot standing upright relative to the well-defined global z-axis, and two different global reference frames, {G₁} and {G₂}, as defined in the bullet point list on page 154. The x-axis of {G₁} and y-axis of {G₂} both point forwards relative to the robot, and the y-axis of {G₁} and x-axis of {G₂} point in antiparallel, i.e. opposite, directions—leftwards and rightwards relative to the robot, respectively. If the robot performs a 90° CCW rotation about the horizontal axis 60° from being forwards and 30° from being leftwards, then the rotation quantified in terms of Euler angles relative to {G$_1$} is given by

$$E_1 = (1.047, \ 1.047, \ 1.571),  \tag{6.45}$$

and in terms of Euler angles relative to {G$_2$} is given by

$$E_2 = (-0.524, \ 0.524, \ -1.571).  \tag{6.46}$$

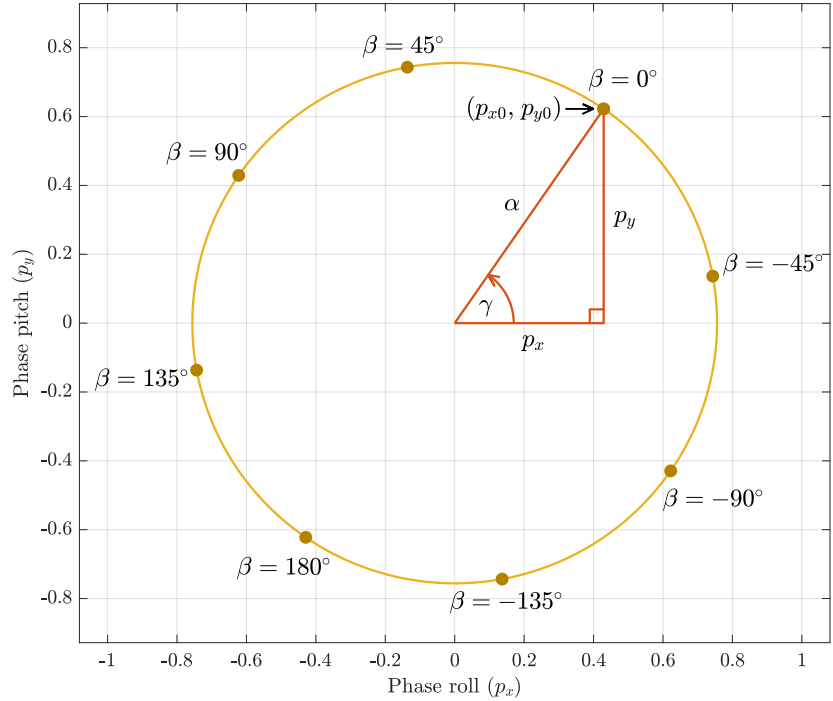

Figure 6.9: Locus of the phase roll and pitch $(p_{x\beta}, p_{y\beta})$ as $\beta$ varies, i.e. for all possible choices of global reference x and y-axes, for the same physical rotation as in Figure 6.8. Points in $\beta$-increments of 45° are labelled, and illustrate how $\beta$ can be seen to be a positive offset to the tilt axis angle $\gamma$, as in Equation (6.33). The orange triangle demonstrates the decomposition of $\alpha$ into the quadrature sinusoid components $p_x$ and $p_y$.

As $\mathbf{x}_{G_1}$ and $\mathbf{y}_{G_2}$ are by definition both the same physical vector, and as $E_1$ and $E_2$ actually both represent the same physical rotation, logically (from the concept of pitch/roll axisymmetry) one would expect that the parameter of $E_1$ that quantifies the amount of rotation about the x-axis, i.e. the Euler roll $\phi_{E1}$, is numerically the same as the parameter of $E_2$ that quantifies the amount of rotation about the y-axis, i.e. the Euler pitch $\theta_{E2}$. This is not the case however, as from above, $\phi_{E1} = 1.571$ and $\theta_{E2} = 0.524$. Similarly, the vectors $\mathbf{y}_{G_1}$ and $\mathbf{x}_{G_2}$ are negatives of each other, so logically from axisymmetry one would expect that $\theta_{E1}$ is numerically the negative of $\phi_{E2}$, but this is also not the case, as $\theta_{E1} = 1.047$ and $\phi_{E2} = -1.571$. Thus, it can be concluded that the Euler angles representation does *not* satisfy pitch/roll axisymmetry. Performing the same calculations for fused angles and the tilt phase space yields

$$\begin{aligned}
F_1 &= (0,\ 1.047,\ 0.524,\ 1), \\
F_2 &= (0,\ 0.524,\ -1.047,\ 1), \\
P_1 &= (0.785,\ 1.360,\ 0), \\
P_2 &= (-1.360,\ 0.785,\ 0).
\end{aligned} \tag{6.47}$$

It is immediately clear by inspection that pitch/roll axisymmetry is satisfied in this example, for both representations, as

$$\begin{aligned}
\phi_1 &= \theta_2 = 0.524, \\
\theta_1 &= -\phi_2 = 1.047, \\
p_{x1} &= p_{y2} = 0.785, \\
p_{y1} &= -p_{x2} = 1.360.
\end{aligned} \tag{6.48}$$

Fused yaw axisymmetry can also be observed in Equation (6.47), as

$$\begin{aligned}
\psi_1 &= \psi_2 = 0, \\
p_{z1} &= p_{z2} = 0,
\end{aligned} \tag{6.49}$$

but the same cannot be said about the Euler yaw, as

$$\psi_{E1} = 1.047 \neq -0.524 = \psi_{E2}. \tag{6.50}$$

Similar to Equation (6.25) for Euler yaw, the non-axisymmetry of the Euler pitch and roll can also be demonstrated mathematically. Using the notation from Equations (6.22) and (6.23) (see also Section 6.2.3.1 for the definition of $\beta$), the Euler pitch and roll of $^C_B R$ can be derived to be

$$\theta_{E\beta} = \operatorname{asin}\big(c_\beta s_{\theta_{E0}} + s_\beta c_{\theta_{E0}} s_{\phi_{E0}}\big), \tag{6.51a}$$

$$\phi_{E\beta} = \operatorname{atan2}\big(c_\beta c_{\theta_{E0}} s_{\phi_{E0}} - s_\beta s_{\theta_{E0}},\ c_{\theta_{E0}} c_{\phi_{E0}}\big). \tag{6.51b}$$

Clearly these expressions are neither independent of $\beta$, as would be required for type (a) axisymmetry, nor do they vary in an intuitive rotational manner with respect to $\beta$ like Equations (6.37) and (6.42) do, as would be required for type (b) axisymmetry. The non-axisymmetry of these expressions can be better visualised by plotting them against each other as a locus of $\beta$. This has been done in Figure 6.10, where the loci of the Euler angles sine ratios $(\sin \phi_E, \sin \theta_E)$, fused angles sine ratios $(\sin \phi, \sin \theta)$, and tilt phase components $(p_x, p_y)$ have been plotted for all $\beta \in (-\pi, \pi]$, for a physical rotation of

$$^U_C R = F_R(-1.2,\ 0.2,\ -1.3,\ 1).$$

The relationships between corresponding points on the three loci are shown in the figure using dotted lines. The uniform circular nature of the fused angles and tilt phase space loci are clear indications of the type (b) axisymmetry of their associated pitch and roll parameters, while the irregular non-uniform shape of the Euler angles locus clearly demonstrates the inherent non-axisymmetry of the Euler pitch and roll parameters. Conceptually, the problem of Euler angles is the fundamental requirement of a sequential order of rotations, leading to definitions of pitch and roll that do not correspond to each other in behaviour, as one then implicitly depends on the value of the other.

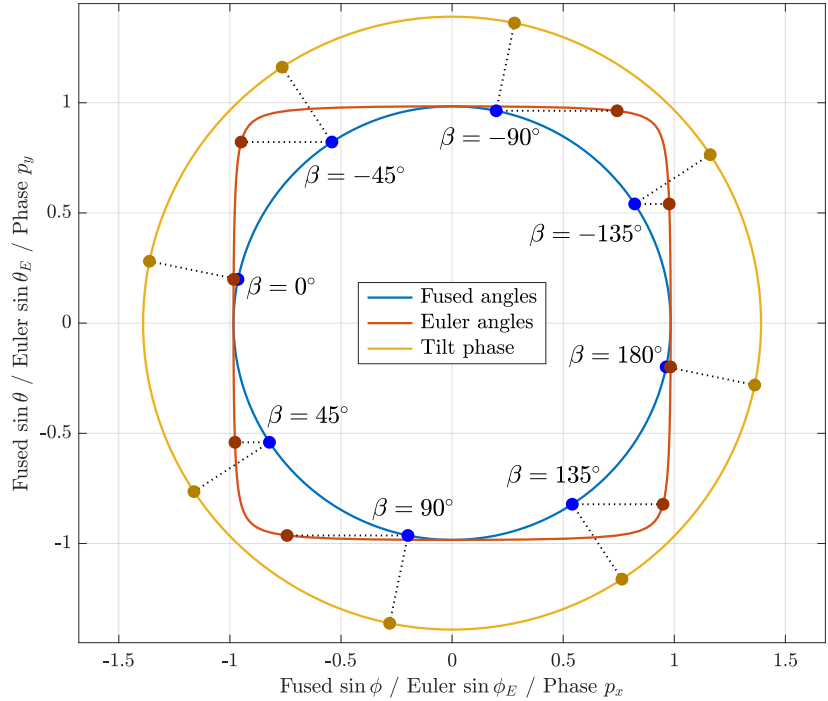

Figure 6.10: The fused angles $(\sin\phi_\beta, \sin\theta_\beta)$, Euler angles $(\sin\phi_{E\beta}, \sin\theta_{E\beta})$, and tilt phase space $(p_{x\beta}, p_{y\beta})$ loci of pitch vs. roll as $\beta$ varies, for the physical rotation ${}^U_C R = F_R(-1.2,\ 0.2,\ -1.3,\ 1)$. For each locus, points in $\beta$-increments of $45°$ are labelled, and joined to each other between the three loci using dotted lines. For $\beta = 0$ we have that ${}^U_C R = {}^G_B R$, and for instance that $(p_{x\beta}, p_{y\beta}) = (p_{x0}, p_{y0})$. The non-circularity of the Euler angles locus, as well as the non-uniform spacing of the associated keypoints, demonstrates the violation of axisymmetry for Euler pitch and roll.

### 6.2.4.5 *Visualising Pitch/Roll Axisymmetry*

Although the nature of pitch/roll axisymmetry has been somewhat visualised in previous figures, e.g. Figures 6.8 and 6.9, we now complete the picture with some further depictions of the nature of the various pitch and roll parameters.

VISUALISATION A    While the variation of fused and Euler yaw with respect to $\beta$ has already been plotted in Figure 6.5 for one particular physical rotation, we now do the same for the various different notions of pitch and roll that have been compared in this chapter. Figure 6.11 shows the variations of the fused, phase and Euler pitch and roll as $\beta$ ranges from $-\pi$ to $\pi$, for the exact same physical rotation as in Figure 6.10. While 90° out-of-phase quadrature sinusoid waveforms can be observed in the fused angles and tilt phase space plots, as discussed in Sections 6.2.4.2 and 6.2.4.3, the Euler angles plot reveals in particular the irregularity and non-axisymmetry of the Euler roll parameter.

VISUALISATION B    We know from the type (a) axisymmetry of the tilt angle $\alpha$, i.e. Equation (6.31), and the type (b) axisymmetry of the tilt axis angle $\gamma$, i.e. Equation (6.33), that all possible representations ${}_B^G R$ of a particular physical rotation ${}_C^U R$ have the same value of $\alpha$, and take on every possible value of $\gamma$. Thus, any single locus as per Figure 6.10 can be generated by plotting the respective pitch and roll parameters, for all rotations that have a tilt angle of $\alpha_0$, i.e. the same tilt angle as ${}_C^U R$. Consequently, all possible loci of pitch and roll (as $\beta$ varies) can be examined at once by plotting the level sets of constant $\alpha$ in the pitch/roll plane. This has been done for the phase pitch and roll in Figure 6.12. For fused angles and Euler angles, it turns out that it is equivalent and more correct to plot the level sets of constant $\sin \alpha$ in the pitch/roll plane, as suggested by relations like Equations (5.160) and (6.40a). The resulting plots are shown in Figure 6.13. While the fused angles and tilt phase space plots demonstrate clear axisymmetry and uniformity about the origin, the Euler angles level set contours get stretched apart into a more rounded square-like form, and for $\sin \alpha = 1$, the top and bottom edges of the square can even be considered to completely 'open up' due to the gimbal lock singularities.

For the purposes of understanding what is going on, one should note that the fused angles and Euler angles $\sin \alpha$ level set plots in Figure 6.13 are actually 'double covers' of the equivalent $\alpha$ level sets, as

$$\sin \alpha \equiv \sin(\pi - \alpha). \tag{6.52}$$

In fact, each cover corresponds exactly to one fused hemisphere $h$. Being slightly lenient with regards to gimbal lock, the top $h$-hemisphere corresponds to a Euler pitch/roll in the domain $\left[-\frac{\pi}{2}, \frac{\pi}{2}\right] \times \left[-\frac{\pi}{2}, \frac{\pi}{2}\right]$, while the bottom $h$-hemisphere corresponds to the Euler roll being in the remainder of its domain, i.e. larger than $\frac{\pi}{2}$ in magnitude. For fused angles, as $\alpha$ moves from 0 to $\pi$, the level sets start as a single point at the origin, expand uniformly to a circle of radius 1 when $\alpha = \frac{\pi}{2}$, and then shrink back down uniformly again to a point as $\alpha$ approaches $\pi$. If one were to plot the raw fused angles $(\phi, \theta)$ instead of

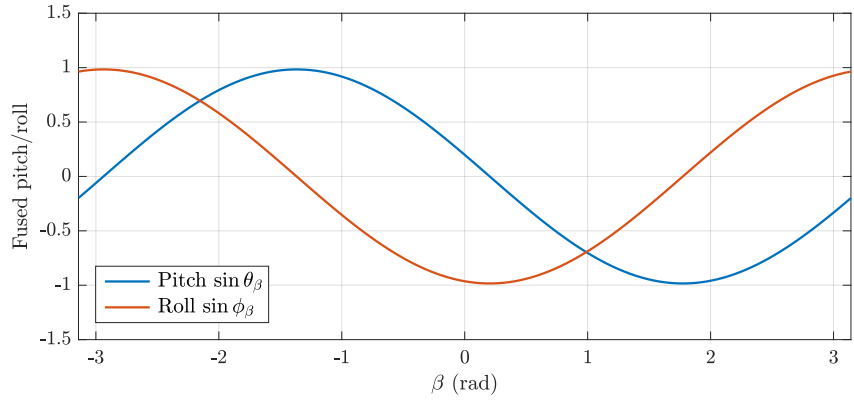

(a) Plot of the fused roll and pitch sine ratios ($\sin \phi_\beta$, $\sin \theta_\beta$) against $\beta$.

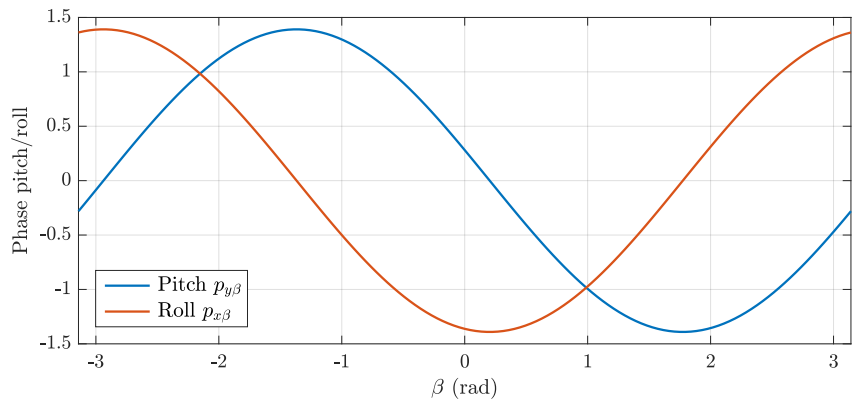

(b) Plot of the phase roll and pitch ($p_{x\beta}$, $p_{y\beta}$) against $\beta$.

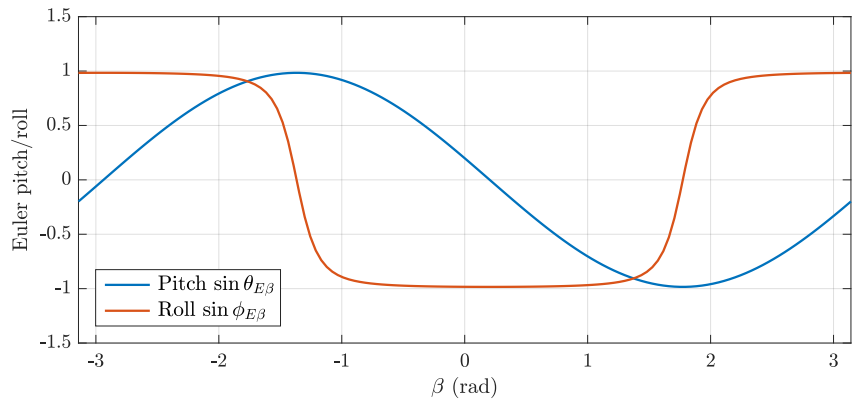

(c) Plot of the Euler roll and pitch sine ratios ($\sin \phi_{E\beta}$, $\sin \theta_{E\beta}$) against $\beta$.

Figure 6.11: Plots of fused, Euler and phase pitch and roll against $\beta$ for the determination of parameter axisymmetry (see Figure 6.5 for the yaw plot). Exactly as in Figure 6.10, a $\beta$ of zero corresponds to the chosen physical rotation of ${}^U_C R = F_R(-1.2, 0.2, -1.3, 1)$. While exact quadrature sinusoid pitch/roll waveforms can be identified in the (a) fused angles and (b) tilt phase space plots, in particular the irregularity and non-axisymmetry of the Euler roll parameter can be identified in the (c) Euler angles plot.

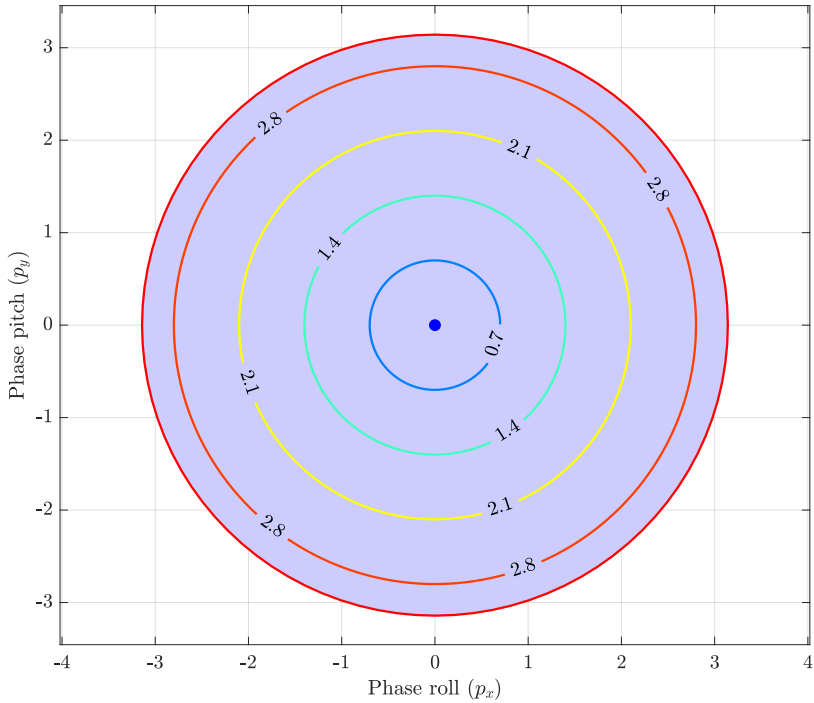

Figure 6.12: Level sets of constant $\alpha$ in the $(p_x, p_y)$ tilt phase space plane. The shaded region is the valid (bounded) 2D tilt phase space domain. The uniformly circular nature of the plot visually illustrates the axisymmetry of the phase pitch and roll.

their sine ratios $(\sin\phi, \sin\theta)$, then a similar behaviour occurs, only the level sets around $\alpha = \frac{\pi}{2}$ grow to a diamond shape and back, as shown by the contours of constant $\sin\alpha$ in Figure 7.5. For Euler angles, as $\alpha$ moves from 0 to $\pi$, the level sets shown in Figure 6.13b start expanding more or less circularly, but quickly stretch out into square form, where the top and bottom edges of the square become increasingly stretched and less densely populated with $\beta$ points. At $\alpha = \frac{\pi}{2}$, the level set essentially corresponds to two separate vertical lines, at $\sin\phi_E = \pm 1$, although it is somewhat a matter of definition how the Euler angles parameters exactly behave at gimbal lock. Beyond that, the exact reverse process of shrinking back down to the origin occurs. If one were to plot the raw Euler angles $(\phi_E, \theta_E)$ instead of their sine ratios, the level sets would initially behave somewhat qualitatively similarly, in that a dot grows approximately circularly before becoming more square-like and ending up at $\alpha = \frac{\pi}{2}$ as two vertical lines at $\phi_E = \pm\frac{\pi}{2}$. Beyond this however, the raw level sets flip outside of these lines and shrink back down to a point at $\phi_E = \pm\pi$ in a somewhat symmetrical reverse process.

**VISUALISATION C**    As discussed in Section 6.2.3.2, and shown in Figures 6.6 and 6.7 (see page 160), the axisymmetry of rotation

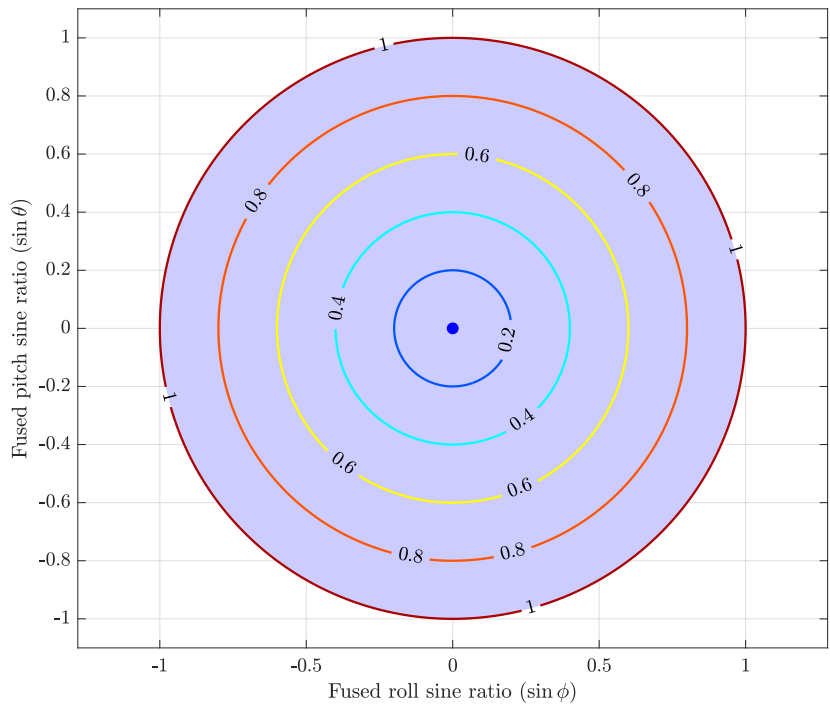

(a) Level sets of constant $\sin\alpha$ in the $(\sin\phi, \sin\theta)$ fused angles plane.

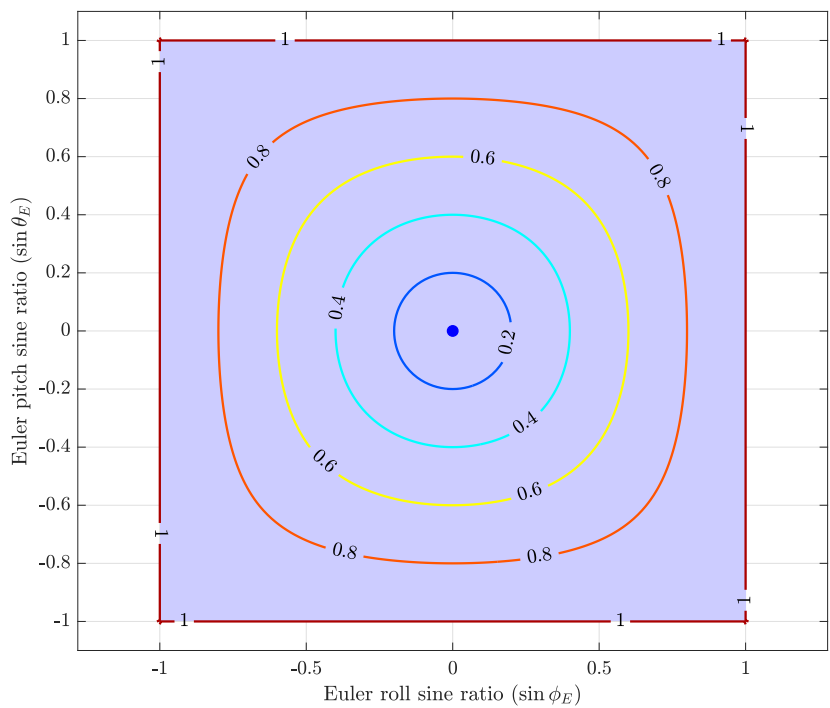

(b) Level sets of constant $\sin\alpha$ in the $(\sin\phi_E, \sin\theta_E)$ Euler angles plane.

Figure 6.13: Level sets of constant $\sin\alpha$ in the fused angles and Euler angles pitch/roll planes. The shaded region in each case corresponds to the valid pitch/roll domain. The uniformly circular nature of (a) the fused angles plot, in contrast to (b) the Euler angles plot, visually illustrates the axisymmetry of the fused pitch and roll.

parameters can be examined by constructing surface plots of their values for all tilt rotations up to $\frac{\pi}{2}$ radians in magnitude. This is because the action of tilting an upright robot away from the z-axis in all different directions is fundamentally equivalent to representing a single tilt rotation of 0 to $\frac{\pi}{2}$ radians relative to all different possible global reference frames. The interpretation of Figure 6.6a, and how it demonstrates the type (a) axisymmetry of the fused yaw and the non-axisymmetry of the Euler yaw, has already been discussed in Section 6.2.3.2. In Figures 6.7a and 6.7b, we can see the surface plots for fused pitch and roll, and phase pitch and roll, respectively. It can clearly be identified that the respective pitch and roll components are mutually symmetric in each case, and mirror each other in behaviour throughout their entire domains. Specifically, a simple 90° rotation about the z-axis maps both roll surface plots directly onto their respective pitch surface plots. This rotation is equivalent to simply reinterpreting all of the plotted tilt rotations with a $\beta$ of $\frac{\pi}{2}$, as per Equation (6.17), so this is one example of pitch/roll axisymmetry that can be directly interpreted from the figure. Looking at the Euler angles plots in Figure 6.6b, the same cannot be said for the Euler pitch and roll. The Euler roll plot shows clear distortion effects due to the gimbal lock singularities, has a much larger range of values (from $-\pi$ to $\pi$, although this cannot be seen in the figure as only $\alpha \leq \frac{\pi}{2}$ is plotted), and does not correspond at all to the pitch plot via any rotation. This illustrates why it can be said that the Euler pitch and roll parameters are *not* axisymmetric, and do *not* correspond to each other in domain and/or behaviour (refer to the statement of Problem D on page 146).

## 6.3 CONCLUSION

This chapter has demonstrated four fundamental reasons why Euler angles are not a suitable representation for representing orientations in balance-related applications, and why the fused angles and tilt phase space representations do better in all four of these areas. Euler angles may still be suitable for the analysis of physical gimbals or a collocated series of joints, but for almost all other applications, the proximity of the gimbal lock singularities to normal working ranges (greatly affecting even the non-yaw components), and the lack of parameter axisymmetry (leading to illogical and non-self-consistent behaviour of the various parameters), make Euler angles unsuitable for the task. The core problem of Euler angles is the inferior reliance on a sequential order of rotations, in particular forcing an order on the pitch and roll parameters instead of allowing them to be concurrent. This leads to mutual parameter dependencies that corrupt the 'purity' of the first and third rotation parameters, as can for example clearly be seen for the Euler yaw and roll in Figure 6.6. None of these problems are the case for the fused angles and tilt phase space representations, where for example it can be observed that the fused angles representation combines the 'uncorrupted' second rotation parameters of the ZYX and ZXY Euler angles representations respectively, with a novel and self-consistent definition of yaw, to form a genuinely useful and geometrically meaningful representation of orientation for balance.

# 7

## A DEEPER INVESTIGATION OF 3D ROTATIONS

Having already rigorously introduced the tilt angles, fused angles and tilt phase space representations in Chapter 5, and having compared them in great detail to Euler angles in Chapter 6, in this chapter we take a step back and present a deeper and more structured approach to the analysis of 3D rotations. This analysis is focused on the representations developed in this thesis, but also importantly provides context about existing rotation representations, consolidating the important known fundamentals about them into one place. Many new properties and results involving fused angles, tilt angles and the tilt phase space are provided, including for example their relation to rotational velocities, spherical linear interpolation and simple topological concepts. In addition to this, new different ways of understanding, visualising and thinking about these representations are also presented. This chapter is not critical to the understanding of the bipedal walking component of this thesis, but provides important contributions to the field of rotation theory for the interested reader, mathematician or roboticist.

## 7.1 A WALKTHROUGH OF 3D ROTATION THEORY

We start with a brief overview of the fundamentals of 3D rotation theory, before reviewing in more specific detail some more advanced properties of the rotation matrix and quaternion representations, many of which are required for later parts of this chapter and thesis. This section can be considered somewhat like a more in-depth 'related work' to the discussion of rotation representations than was already presented in Section 5.3.

### 7.1.1 Fundamental 3D Rotation Concepts

In this section, the basics of coordinate frames and vector bases are introduced, including how rotation matrices and quaternions can be defined and used to map between them. The fundamental actions of rotations on vectors and/or other rotations are also categorically listed, and generalised to the theory of *referenced rotations*, a novel contribution to basic rotation theory. The work in this section can be seen as the axiomatic foundation to all discussions of rotations in this thesis.

### 7.1.1.1   *Coordinate Frames and Basis Vectors*

From a mathematical standpoint, the everyday three-dimensional world we live in is a Euclidean space. In fact, seeing as all Euclidean spaces are equivalent (isomorphic) to one another, it can be said to be *the* three-dimensional Euclidean space. All this means however, is that every location in space can be modelled as a 'point', where the set of all points then satisfies a number of axiomatic relationships expressible in terms of the concepts of 'distance' and 'angle'. Euclidean spaces are closely tied to the concept of vector spaces however, and in fact, the definition of the former from the latter reveals that all points can be uniquely labelled by a 3-tuple of real number coordinates, i.e. something like '$(1, 2, 0.5)$'. A labelling is referred to as a Cartesian coordinate system if the coordinates are consistent with the axiomatic conditions governing Euclidean spaces and abide by certain conditions of orthogonality (i.e. the coordinate system cannot be skew). There are many ways to uniquely, consistently and orthogonally label all points in space with a 3-tuple of Cartesian coordinates however, but each possible way can be seen to correspond to a particular so-called coordinate frame. Each coordinate frame consists of a nominated point in space referred to as its origin, in addition to three mutually orthogonal (pairwise 90° apart) unit vectors $\mathbf{x}$, $\mathbf{y}$ and $\mathbf{z}$, that are usually considered geometrically to 'start' at the origin, as illustrated in Figure 7.1. As shown, given a coordinate frame, the 3-tuple (i.e. vector relative to the origin) corresponding to each point in space is given by the signed distance thereof to the $\mathbf{yz}$, $\mathbf{xz}$ and $\mathbf{xy}$ planes, respectively.

Coordinate frames are generally labelled by uppercase letters in parentheses, e.g. {A}, {B} and {C}, so by convention the associated coordinate axes or basis vectors are subscripted by the same letter, i.e. $\mathbf{x}_A$, $\mathbf{y}_A$ and $\mathbf{z}_A$ for frame {A}. Every point or vector (i.e. directed line segment from point to point) that is labelled (i.e. quantified) with respect to a particular coordinate frame, is also adorned with the respective letter. For instance, the vector from the origin of coordinate frame {A} to a particular point P in space can be denoted $^A\mathbf{v}$, but the same vector expressed relative to coordinate frame {B} is denoted $^B\mathbf{v}$. Note that $^B\mathbf{v}$ still refers to the vector from the origin of frame {A} to the point P however. The coordinate axis vectors $\mathbf{x}_A$, $\mathbf{y}_A$ and $\mathbf{z}_A$ of a frame {A} can also be expressed in terms of another frame {B} in this way. This yields the commonly seen notation $^B\mathbf{x}_A$, $^B\mathbf{y}_A$ and $^B\mathbf{z}_A$, for the basis vectors of frame {A} with respect to frame {B}.

Many of the axiomatic relationships that define Euclidean spaces relate to the concept of motions, i.e. one-to-one transformations or isometries of the Euclidean space to itself. By definition, all motions must preserve the distances and angles between any set of points in space, so that they constitute one-to-one maps between equivalent Euclidean spaces. This can loosely be interpreted to mean that no

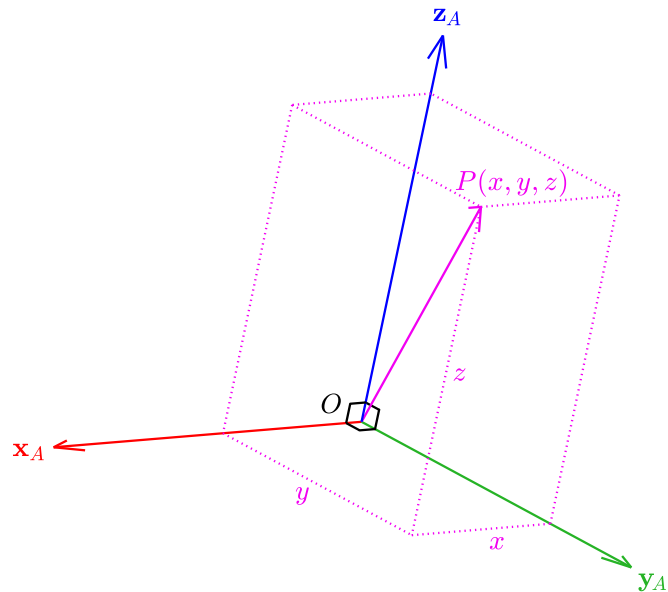

Figure 7.1: General depiction of a three-dimensional coordinate frame, in this
case designated as frame '{A}'. The three mutually perpendicular
coordinate axes or *basis vectors* are given by the unit vectors $\mathbf{x}_A$,
$\mathbf{y}_A$ and $\mathbf{z}_A$—all drawn as starting at the coordinate frame origin
$O$. Every point $P$ in space can be labelled with a 3-tuple $(x, y, z)$
of real scalar coordinates, corresponding to its signed distance to
the $\mathbf{y}_A\mathbf{z}_A$, $\mathbf{x}_A\mathbf{z}_A$ and $\mathbf{x}_A\mathbf{y}_A$ planes, respectively. The tuple $(x, y, z)$
then also denotes the vector from $O$ to $P$ relative to frame {A}.

parts of the 3D space can move relative to one other as an effect of the
motion, and that any observer attached to the moving space does not
notice any relative change (other than possibly a mirroring).

The different fundamental types of motions can be identified by
examining the fixed points they have in space, i.e. the points that map
to themselves under application of the motion. Given any two distinct
fixed points of a motion, by the preservation of distances and angles,
one can see that all points on the line joining the two fixed points are
also fixed. Likewise, given two distinct fixed lines, a similar argument
applied to all possible pairs of points between the two lines shows that
the plane containing the two fixed lines is also fixed. Furthermore, if
two distinct planes are fixed, then the entire 3D space must be fixed, i.e.
invariant to the motion. From these basic results, it can be concluded
that the set of fixed points of *every* motion must either be empty, or an
n-dimensional Euclidean subspace of the original space. That is, the
set of fixed points of a motion can either be the entire space ($n = 3$),
a single plane ($n = 2$), a single line ($n = 1$), a single point ($n = 0$), or
the empty set. The categorisation of the fundamental motion types
corresponding to each of these cases is given as follows:

**Entire space fixed:**  Corresponds to the identity motion, which maps
every point to itself.

**Single plane fixed:** Corresponds fundamentally to reflections, which swap all points in space with their perpendicularly opposite locations on the other side of the chosen fixed plane.

**Single line fixed:** Corresponds fundamentally to rotations, which turn every point by a fixed angle about the chosen fixed line.

**Single point fixed:** Corresponds fundamentally to so-called rotoreflections, which are rotations about an axis, followed by a reflection about a plane perpendicular to that axis.

**No point fixed:** Corresponds fundamentally to translations, which shift every point in space by the same distance in the same direction.

Combinations of these fundamental motion types are also possible, giving rise to:

- Glide plane operations, which are reflections about a plane, followed by a translation parallel to that plane, and,

- Screw operations, which are rotations about a line, followed by a translation parallel to that line,

both of which do not have any fixed points due to the translation component. This completes the list of fundamental motion types for the three-dimensional Euclidean space, and it is a known result that every single motion on this space is one of the aforementioned types.

In the application scenario of the analysis of the orientations of rigid bodies in 3D space, translations can be ignored as they do not affect orientation, and reflections can be ignored as they are not physically realisable by rigid bodies in the real world. Thus, removing the translation and reflection components of all seven just-listed types of Euclidean motions leaves just rotations and the identity motion (i.e. a rotation by zero radians about any axis) as the motions that need to be considered to characterise all possible orientations of a rigid body. This seems to be a very trivial and intuitively obvious statement, but what this really demonstrates is that all possible orientations of a rigid body in space can be achieved from every other orientation by a single rotation about a single fixed line, i.e. Euler's rotation theorem. Note that it can quickly be deduced that the angle of rotation needed is at most 180°, or one could simply rotate in the opposite direction.

If we now attach *coordinate frames* to the rigid bodies in space as they rotate, we can trivially deduce that all possible pairs of achievable coordinate frames can be mutually transformed into each other by a single rotation about a single fixed axis. The only exception is that we must be consistent about whether the attached coordinate frames are right-handed, i.e. ones where

$$\mathbf{x} \times \mathbf{y} = \mathbf{z}, \tag{7.1}$$

or left-handed, i.e. ones where

$$\mathbf{x} \times \mathbf{y} = -\mathbf{z}. \tag{7.2}$$

The reason for this is that coordinate frames can indeed be reflected, while real rigid bodies cannot, and reflections directly toggle which of Equation (7.1) and Equation (7.2) is true. In other words, the set of transformations between all possible coordinate frames is the set of all (possibly trivial) *rotoreflections*, while for rigid bodies it is just the set of all (possibly trivial) *rotations*. Thus, the set of all coordinate frames can be seen to be a disconnected set with two equivalent connected components, and we need to choose a convention of using just one of these halves.[1] The default convention in literature is to use only right-handed coordinate frames, and this convention is used throughout this thesis as well.

Now that it has been established that all possible right-handed coordinate frames are linked by pure rotations, and can be used to uniquely represent any orientation of a rigid body in space, a basis notation for these linking rotations can be introduced, similar to how it was done on page 178 for vectors—in particular coordinate axis basis vectors. If the rotation from a coordinate frame {A} to a coordinate frame {B} is numerically represented and quantified by a certain parameter $X$, e.g. a rotation matrix or quaternion or other, then it is adorned with the basis notation $_B^A X$. The top left prefix is the label of the frame that the rotation is relative to, and the bottom left prefix is the label of the frame that results from the rotation. This is intuitively consistent with the basis notation previously introduced for vectors.

### 7.1.1.2 *Rotation Matrices*

From the fundamental axioms of Euclidean spaces, and the knowledge that all rotations must preserve the distances and angles between any set of points, it can be concluded from basic geometrical constructs that rotations must be a linear operation.[2] That is, if the function $f(\mathbf{v})$ represents the effect of applying a particular rotation to a vector $\mathbf{v}$, we must have that

$$f(a\mathbf{v} + b\mathbf{w}) = af(\mathbf{v}) + bf(\mathbf{w}), \tag{7.3}$$

for all vectors $\mathbf{v}, \mathbf{w} \in \mathbb{R}^3$ and $a, b \in \mathbb{R}$. It can also be observed from the definition of coordinate frames that any point or vector

---

1 The equivalence of the two halves can be seen by, for example, mapping $\mathbf{x} \leftrightarrow -\mathbf{x}$ or $\mathbf{z} \leftrightarrow -\mathbf{z}$ between the right- and left-handed connected components respectively.

2 For example, $\mathbf{v}$ and $a\mathbf{v}$ are an angle of 0 or $\pi$ apart, so $f(\mathbf{v})$ and $f(a\mathbf{v})$ must be also. Thus we must have $f(a\mathbf{v}) = kf(\mathbf{v})$ for some $k \in \mathbb{R}$, where $k$ has the same sign as $a$. The distance of $a\mathbf{v}$ to the origin is $\|a\mathbf{v}\|$, so by preservation of distances we must have $\|f(a\mathbf{v})\| = \|a\mathbf{v}\| = a\|\mathbf{v}\| = a\|f(\mathbf{v})\|$. We conclude that $k = a$, and thus that $f(a\mathbf{v}) = af(\mathbf{v})$. By similar arguments involving a parallelogram construct, one can also easily deduce that $f(\mathbf{v} + \mathbf{w}) = f(\mathbf{v}) + f(\mathbf{w})$. Equation (7.3) then follows trivially.

$^A\mathbf{v} = (v_x, v_y, v_z)$ relative to a frame {A} can be expressed as the unique linear combination

$$^A\mathbf{v} = v_x\,^A\mathbf{x}_A + v_y\,^A\mathbf{y}_A + v_z\,^A\mathbf{z}_A, \tag{7.4}$$

where $^A\mathbf{x}_A$, $^A\mathbf{y}_A$ and $^A\mathbf{z}_A$ are the basis vectors of frame {A} with respect to frame {A}, which are clearly

$$^A\mathbf{x}_A = (1, 0, 0), \tag{7.5a}$$

$$^A\mathbf{y}_A = (0, 1, 0), \tag{7.5b}$$

$$^A\mathbf{z}_A = (0, 0, 1). \tag{7.5c}$$

Thus, if $f : \mathbb{R}^3 \to \mathbb{R}^3$ represents the rotation from frame {A} to another frame {B}, and the result of rotating $^A\mathbf{v}$ is given by $^A\mathbf{w}$, then

$$
\begin{aligned}
^A\mathbf{w} &= f(^A\mathbf{v}) \\
&= f(v_x\,^A\mathbf{x}_A + v_y\,^A\mathbf{y}_A + v_z\,^A\mathbf{z}_A) \\
&= v_x f(^A\mathbf{x}_A) + v_y f(^A\mathbf{y}_A) + v_z f(^A\mathbf{z}_A) \\
&= v_x\,^A\mathbf{x}_B + v_y\,^A\mathbf{y}_B + v_z\,^A\mathbf{z}_B.
\end{aligned}
\tag{7.6}
$$

The last step comes about because the rotation from {A} to {B}, by definition, must rotate the x-axis of {A} onto the x-axis of {B}, and so on. By treating $^A\mathbf{v} = (v_x, v_y, v_z)$ and $^A\mathbf{w}$ as column vectors, we observe that Equation (7.6) can be expressed more succinctly in the form of a matrix equation, namely

$$^A\mathbf{w} = \begin{bmatrix} \uparrow & \uparrow & \uparrow \\ ^A\mathbf{x}_B & ^A\mathbf{y}_B & ^A\mathbf{z}_B \\ \downarrow & \downarrow & \downarrow \end{bmatrix} {}^A\mathbf{v}. \tag{7.7}$$

We note that the matrix in this equation is independent of $^A\mathbf{v}$, and more importantly, specifies exactly how every point and vector in space is affected by the rotation. As a result, we can see that this matrix numerically *quantifies* the given rotation, and is referred to as the rotation matrix

$$^A_B R = \begin{bmatrix} \uparrow & \uparrow & \uparrow \\ ^A\mathbf{x}_B & ^A\mathbf{y}_B & ^A\mathbf{z}_B \\ \downarrow & \downarrow & \downarrow \end{bmatrix} \tag{7.8}$$

of the rotation from {A} to {B}. Recall that the top left prefix in the notation '$^A_B R$' is the label of the frame that the rotation is relative to, and the bottom left prefix is the label of the frame that results from the rotation.

The columns of $^A_B R$ are all mutually perpendicular (i.e. orthogonal) unit vectors, and therefore form an orthonormal basis of $\mathbb{R}^3$. As a direct result (from linear algebra), it can be concluded that the matrix must be orthogonal, i.e. that

$$\,^A_B R^{-1} = \,^A_B R^T. \tag{7.9}$$

Recalling our convention to only use right-handed coordinate frames, using Equation (7.1) we can further deduce that

$$
\begin{aligned}
\det\big(\,^A_B R\big) &= \begin{vmatrix} \uparrow & \uparrow & \uparrow \\ ^A\mathbf{x}_B & ^A\mathbf{y}_B & ^A\mathbf{z}_B \\ \downarrow & \downarrow & \downarrow \end{vmatrix} \\
&= (\,^A\mathbf{x}_B \times \,^A\mathbf{y}_B) \bullet \,^A\mathbf{z}_B \\
&= \,^A\mathbf{z}_B \bullet \,^A\mathbf{z}_B = \|\,^A\mathbf{z}_B\|^2 \\
&= 1.
\end{aligned}
\tag{7.10}
$$

More than just being properties, Equations (7.9) and (7.10) are in fact the exact defining conditions of rotation matrices. As such, the set of all rotation matrices is given by the special orthogonal group[3]

$$SO(3) = \{R \in \mathbb{R}^{3\times3} : R^{-1} = R^T, \det(R) = 1\}. \tag{7.11}$$

The normal convention for referencing individual entries of a matrix $R$ is used in this thesis for rotation matrices also, i.e.

$$
R = \begin{bmatrix} R_{11} & R_{12} & R_{13} \\ R_{21} & R_{22} & R_{23} \\ R_{31} & R_{32} & R_{33} \end{bmatrix}.
\tag{7.12}
$$

Equation (7.6) is a simple geometric vector equation, so it must be true no matter what basis or coordinate frame (currently frame {A}) it is numerically evaluated relative to. In particular, the equation must also be true if it is reexpressed in terms of frame {B}, i.e.

$$^B\mathbf{w} = v_x\,^B\mathbf{x}_B + v_y\,^B\mathbf{y}_B + v_z\,^B\mathbf{z}_B. \tag{7.13}$$

Clearly, from Equation (7.5), the right-hand side can be simplified to

$$^B\mathbf{w} = (v_x, v_y, v_z) = \,^A\mathbf{v}. \tag{7.14}$$

Thus, while Equation (7.7) nominally expresses how a rotation matrix rotates one vector into another relative to the same frame, i.e.

$$^A\mathbf{w} = \,^A_B R\,^A\mathbf{v}, \tag{7.15}$$

using Equation (7.14) it can also be reinterpreted as a 'change of vector basis' equation, namely

$$^A\mathbf{w} = \,^A_B R\,^B\mathbf{w}. \tag{7.16}$$

---

3 The orthogonal group $O(n)$ is the set of all $n \times n$ orthogonal matrices, and the special orthogonal group $SO(n)$ is the connected normal subgroup of index 2 thereof, of all orthogonal matrices whose determinant is $+1$ (as opposed to $-1$).

Note how the two '$B$' frames can be thought of as 'cancelling' on the forwards diagonal of the right-hand side terms, leaving over just the '$A$' as the basis of the resulting $\mathbf{w}$ vector. One consequence of Equation (7.16) is that it is now trivial to see by premultiplication that

$$^B\mathbf{w} = {}^A_B R^{-1}\, {}^A\mathbf{w},\tag{7.17}$$

and (by swapping '$A$' and '$B$') that

$$^B\mathbf{w} = {}^B_A R\, {}^A\mathbf{w}.\tag{7.18}$$

We can conclude that the inverse rotation of ${}^A_B R$ is given by

$$^A_B R^{-1} = {}^A_B R^T = {}^B_A R.\tag{7.19}$$

Using this result in combination with Equation (7.8), it becomes clear that the *rows* of ${}^A_B R$ are the basis vectors of {A} with respect to {B}.[4] Thus, in summary,

$$^A_B R = \begin{bmatrix} \uparrow & \uparrow & \uparrow \\ {}^A\mathbf{x}_B & {}^A\mathbf{y}_B & {}^A\mathbf{z}_B \\ \downarrow & \downarrow & \downarrow \end{bmatrix} = \begin{bmatrix} \leftarrow & {}^B\mathbf{x}_A & \rightarrow \\ \leftarrow & {}^B\mathbf{y}_A & \rightarrow \\ \leftarrow & {}^B\mathbf{z}_A & \rightarrow \end{bmatrix}.\tag{7.20}$$

If $\mathbf{x}$, $\mathbf{y}$ and $\mathbf{z}$ are the rows or columns of any rotation matrix, or equivalently the axes of any right-handed coordinate frame, then the following fundamental relations apply:

$$\mathbf{x} \cdot \mathbf{y} = 0, \qquad\qquad \mathbf{x} \times \mathbf{y} = \mathbf{z},\tag{7.21a}$$

$$\mathbf{y} \cdot \mathbf{z} = 0, \qquad\qquad \mathbf{y} \times \mathbf{z} = \mathbf{x},\tag{7.21b}$$

$$\mathbf{z} \cdot \mathbf{x} = 0, \qquad\qquad \mathbf{z} \times \mathbf{x} = \mathbf{y}.\tag{7.21c}$$

From a geometric argument relating directly to how vectors are resolved with respect to a coordinate frame (i.e. the geometric definition of the dot product), one can see for example that

$$^A\mathbf{x}_B = (\mathbf{x}_B \cdot \mathbf{x}_A,\ \mathbf{x}_B \cdot \mathbf{y}_A,\ \mathbf{x}_B \cdot \mathbf{z}_A).\tag{7.22}$$

As such, the rotation matrix ${}^A_B R$ in Equation (7.20) can alternatively be characterised by the formulations

$$^A_B R = \begin{bmatrix} \mathbf{x}_B \cdot \mathbf{x}_A & \mathbf{y}_B \cdot \mathbf{x}_A & \mathbf{z}_B \cdot \mathbf{x}_A \\ \mathbf{x}_B \cdot \mathbf{y}_A & \mathbf{y}_B \cdot \mathbf{y}_A & \mathbf{z}_B \cdot \mathbf{y}_A \\ \mathbf{x}_B \cdot \mathbf{z}_A & \mathbf{y}_B \cdot \mathbf{z}_A & \mathbf{z}_B \cdot \mathbf{z}_A \end{bmatrix}\tag{7.23a}$$

$$= \begin{bmatrix} \cos(\theta_{xx}) & \cos(\theta_{yx}) & \cos(\theta_{zx}) \\ \cos(\theta_{xy}) & \cos(\theta_{yy}) & \cos(\theta_{zy}) \\ \cos(\theta_{xz}) & \cos(\theta_{yz}) & \cos(\theta_{zz}) \end{bmatrix},\tag{7.23b}$$

---

4 As ${}^A_B R = {}^B_A R^T$ from Equation (7.19), and ${}^B_A R = \begin{bmatrix} {}^B\mathbf{x}_A & {}^B\mathbf{y}_A & {}^B\mathbf{z}_A \end{bmatrix}$ from Equation (7.8).

where $\theta_{zx}$ for example is the angle between the vectors $\mathbf{z}_B$ and $\mathbf{x}_A$. Equation (7.23b) in particular reveals why rotation matrices are sometimes also referred to as a direction cosine matrix, as the nine matrix entries are just the cosines of the pairwise angles between the reference and target coordinate frame axes.

The final point that needs to be considered for now is how rotation composition can be modelled using rotation matrices. Suppose a frame {A} is first rotated by $_B^A R$ to become frame {B}, and that this frame is further locally rotated by $_C^B R$ to become frame {C}. We expect the combined total rotation to be the rotation $_C^A R$ from {A} to {C}, and from Equation (7.16) we must have that

$$^A\mathbf{w} = {}_C^A R \, {}^C\mathbf{w},$$ (7.24)

for every vector $\mathbf{w} \in \mathbb{R}^3$. We note however that

$$^A\mathbf{w} = {}_B^A R \, {}^B\mathbf{w},$$ (7.25a)

$$^B\mathbf{w} = {}_C^B R \, {}^C\mathbf{w},$$ (7.25b)

so by substitution,

$$^A\mathbf{w} = {}_B^A R \, {}_C^B R \, {}^C\mathbf{w}.$$ (7.26)

By comparing this to Equation (7.24), we deduce that

$$_B^A R \, {}_C^B R = {}_C^A R.$$ (7.27)

Thus, rotation composition is equivalent to standard multiplication of the associated rotation matrices, or more carefully, standard matrix post-multiplication, if the second applied rotation is local to the frame that results from the first. Note that again, like in Equation (7.16), the two '$B$' frames can be thought of as 'cancelling' on the forwards diagonal of the left-hand side terms, leaving over just the '$A$' and the '$C$' on the right-hand side. Note also that just like rotation composition, matrix multiplication is *non-commutative*, associative and distributive over vector addition.

### 7.1.1.3  *Quaternions*

A completely different approach to how rotations can be represented and numerically manipulated comes from the theory of quaternions. A brief overview of quaternions was provided in Section 5.3.3, but we review and extend this now to demonstrate in depth how quaternions can be used to represent rotations, combine them sequentially, and apply them to arbitrary vectors.

We recall that fundamentally, every quaternion $q$ is just the combination of four real coefficients $w, x, y, z \in \mathbb{R}$, in the form

$$q = w + xi + yj + zk,$$ (7.28)

where $i$, $j$ and $k$ are three independent imaginary 'numbers' that are chosen by definition to satisfy

$$i^2 = j^2 = k^2 = ijk = -1. \tag{7.29}$$

Usually, the $i$, $j$ and $k$ variables are omitted when writing a quaternion, and the quaternion is simply written as

$$q = (w, x, y, z), \tag{7.30}$$

or alternatively,

$$q = (q_0, \mathbf{q}), \tag{7.31}$$

where the so-called scalar part $q_0$ and vector part $\mathbf{q}$ are given by

$$q_0 = w \in \mathbb{R}, \tag{7.32a}$$

$$\mathbf{q} = (x, y, z) \in \mathbb{R}^3. \tag{7.32b}$$

By definition, every vector $\mathbf{v} \in \mathbb{R}^3$ is considered to be equivalent to the quaternion that has $\mathbf{v}$ as its vector part and zero as its scalar part, i.e.[5]

$$\mathbf{v} \equiv (0, \mathbf{v}). \tag{7.33}$$

In this way, vectors can actually be viewed as simply being special cases of quaternions, and can interact with other quaternions without the need for any further definitions.

Based on Equation (7.29), all further necessary identities involving $i$, $j$ and $k$ can be derived. For instance,

$$jk = -i^2 jk = -i(ijk) = i. \tag{7.34}$$

The complete multiplication table is given by

| $\times$ | $i$ | $j$ | $k$ |
|---|---|---|---|
| $i$ | $-1$ | $k$ | $-j$ |
| $j$ | $-k$ | $-1$ | $i$ |
| $k$ | $j$ | $-i$ | $-1$ |

where for example the result of the product $ij$ is given where the row labelled $i$ and column labelled $j$ meet. It is striking to notice that

$$ji = -ij, \tag{7.35a}$$

$$ki = -ik, \tag{7.35b}$$

$$kj = -jk. \tag{7.35c}$$

---

5 Note that the right-hand side of this equation is a quaternion, and is using the notation from Equation (7.31).

These basic identities involving $i$, $j$ and $k$ can be used to naturally define a multiplication operator between quaternions. That is, if $p = (p_0, \mathbf{p}) = (w_1, x_1, y_1, z_1)$ and $q = (q_0, \mathbf{q}) = (w_2, x_2, y_2, z_2)$, then

$$
\begin{aligned}
pq &= (w_1, x_1, y_1, z_1)(w_2, x_2, y_2, z_2) \\
&= (w_1 + x_1 i + y_1 j + z_1 k)(w_2 + x_2 i + y_2 j + z_2 k) \\
&= w_1 w_2 + x_1 x_2 i^2 + y_1 y_2 j^2 + z_1 z_2 k^2 \\
&\quad + (w_1 x_2 + x_1 w_2)i + (w_1 y_2 + y_1 w_2)j + (w_1 z_2 + z_1 w_2)k \\
&\quad + (x_1 y_2 - y_1 x_2)ij + (x_1 z_2 - z_1 x_2)ik + (y_1 z_2 - z_1 y_2)jk \\
&= (w_1 w_2 - x_1 x_2 - y_1 y_2 - z_1 z_2) + (w_1 x_2 + x_1 w_2 + y_1 z_2 - z_1 y_2)i \\
&\quad + (w_1 y_2 - x_1 z_2 + y_1 w_2 + z_1 x_2)j + (w_1 z_2 + x_1 y_2 - y_1 x_2 + z_1 w_2)k. \\
&= (w_1 w_2 - x_1 x_2 - y_1 y_2 - z_1 z_2, \\
&\quad\quad w_1 x_2 + x_1 w_2 + y_1 z_2 - z_1 y_2, \\
&\quad\quad w_1 y_2 - x_1 z_2 + y_1 w_2 + z_1 x_2, \\
&\quad\quad w_1 z_2 + x_1 y_2 - y_1 x_2 + z_1 w_2).
\end{aligned}
\tag{7.36}
$$

Using the scalar-vector notation, this complicated expression can be significantly simplified to get

$$
\begin{aligned}
pq &= (p_0, \mathbf{p})(q_0, \mathbf{q}) \\
&= (p_0 q_0 - \mathbf{p} \cdot \mathbf{q}, \ p_0 \mathbf{q} + q_0 \mathbf{p} + \mathbf{p} \times \mathbf{q}).
\end{aligned}
\tag{7.37}
$$

It is easy to see from the cross product in this equation that quaternion multiplication is not commutative. In all other ways however, quaternion multiplication works exactly how one would expect, e.g. in terms of associativity, distributivity and scaling of one of the terms by a real constant.

Let us consider now how quaternions can be used for the quantification of 3D rotations. Suppose we have a rotation from some frame {A} to another frame {B}, and suppose that this rotation is given by a counterclockwise (CCW) turn of $\theta_a \in [0, \pi]$ radians about the unit vector axis $\hat{\mathbf{e}} \in \mathbb{R}^3$. Given any vector $\mathbf{v} \in \mathbb{R}^3$, let $\mathbf{w} \in \mathbb{R}^3$ be the vector that results when the rotation is applied to $\mathbf{v}$. As shown in Figure 7.2, the vectors $\mathbf{v}$ and $\mathbf{w}$ can be resolved into their components $\mathbf{v}_{\hat{e}}, \mathbf{w}_{\hat{e}}$ and $\mathbf{v}_\perp, \mathbf{w}_\perp$ parallel and perpendicular to the axis of rotation $\hat{\mathbf{e}}$, respectively, where

$$
\mathbf{v}_{\hat{e}} = (\hat{\mathbf{e}} \cdot \mathbf{v})\hat{\mathbf{e}}, \tag{7.38a}
$$

$$
\mathbf{v}_\perp = \mathbf{v} - (\hat{\mathbf{e}} \cdot \mathbf{v})\hat{\mathbf{e}}, \tag{7.38b}
$$

and

$$
\mathbf{v} = \mathbf{v}_{\hat{e}} + \mathbf{v}_\perp, \tag{7.39a}
$$

$$
\mathbf{w} = \mathbf{w}_{\hat{e}} + \mathbf{w}_\perp. \tag{7.39b}
$$

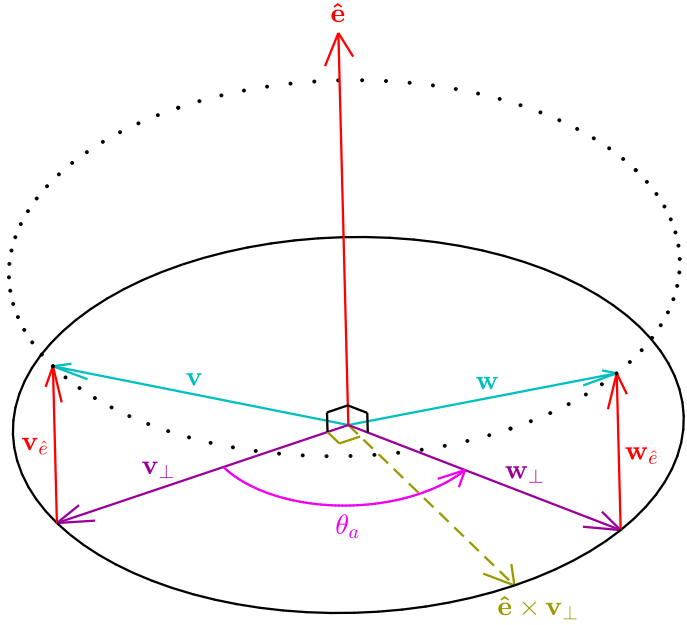

Figure 7.2: Vector diagram for the rotation of a vector $\mathbf{v} \in \mathbb{R}^3$ by an arbitrary axis-angle rotation $(\hat{\mathbf{e}}, \theta_a)$ to give $\mathbf{w} \in \mathbb{R}^3$. For the purpose of analysis, the vectors $\mathbf{v}$ and $\mathbf{w}$ are split up into their components parallel and perpendicular to $\hat{\mathbf{e}}$, from which we can see for example that $\mathbf{v}_{\hat{e}} = \mathbf{w}_{\hat{e}}$.

It can further be seen from Figure 7.2 that geometrically we must have that

$$\mathbf{v}_{\hat{e}} = \mathbf{w}_{\hat{e}}, \tag{7.40}$$

as any rotation about $\hat{\mathbf{e}}$ clearly leaves the component of $\mathbf{v}$ in the direction of $\hat{\mathbf{e}}$ untouched, and

$$\mathbf{w}_{\perp} = \mathbf{v}_{\perp} \cos\theta_a + (\hat{\mathbf{e}} \times \mathbf{v}_{\perp}) \sin\theta_a, \tag{7.41}$$

as the applied rotation makes $\mathbf{v}_{\perp}$ trace out a CCW uniform circular arc of magnitude $\theta_a$, all in the plane perpendicular to $\hat{\mathbf{e}}$. Equation (7.41) is basically a statement of the vector form of spherical linear interpolation from $\mathbf{v}_{\perp}$ towards $\hat{\mathbf{e}} \times \mathbf{v}_{\perp}$, and is based on the fact that the general parametric equation of a circle is given by $(r \cos t, r \sin t)$.

Given Equations (7.38) to (7.41), we can see that

$$\begin{aligned}
\mathbf{w} &= \mathbf{w}_{\hat{e}} + \mathbf{w}_{\perp} \\
&= \mathbf{v}_{\hat{e}} + \mathbf{v}_{\perp} \cos\theta_a + (\hat{\mathbf{e}} \times \mathbf{v}_{\perp}) \sin\theta_a \\
&= \mathbf{v}_{\hat{e}} + (\mathbf{v} - \mathbf{v}_{\hat{e}}) \cos\theta_a + (\hat{\mathbf{e}} \times (\mathbf{v} - \mathbf{v}_{\hat{e}})) \sin\theta_a \\
&= \mathbf{v} \cos\theta_a + \mathbf{v}_{\hat{e}}(1 - \cos\theta_a) + (\hat{\mathbf{e}} \times \mathbf{v}) \sin\theta_a - (\hat{\mathbf{e}} \times \mathbf{v}_{\hat{e}}) \sin\theta_a \\
&= \mathbf{v} \cos\theta_a + (1 - \cos\theta_a)(\hat{\mathbf{e}} \cdot \mathbf{v})\hat{\mathbf{e}} + (\hat{\mathbf{e}} \times \mathbf{v}) \sin\theta_a, \tag{7.42}
\end{aligned}$$

as $\hat{\mathbf{e}} \times \mathbf{v}_{\hat{e}} = 0$, as these two vectors are parallel. Using various trigonometric double-angle formulas to expand $\cos \theta_a$ and $\sin \theta_a$ yields

$$\mathbf{w} = \left(\cos^2 \tfrac{\theta_a}{2} - \sin^2 \tfrac{\theta_a}{2}\right)\mathbf{v} + 2\sin^2 \tfrac{\theta_a}{2}(\hat{\mathbf{e}} \cdot \mathbf{v})\hat{\mathbf{e}} + 2\sin \tfrac{\theta_a}{2}\cos \tfrac{\theta_a}{2}(\hat{\mathbf{e}} \times \mathbf{v}).$$

Now suppose we construct the unit quaternion $q = (q_0, \mathbf{q})$, where

$$q_0 = \cos \tfrac{\theta_a}{2}, \tag{7.43a}$$

$$\mathbf{q} = \hat{\mathbf{e}} \sin \tfrac{\theta_a}{2}. \tag{7.43b}$$

This allows us to simplify the previous equation into

$$\mathbf{w} = (q_0^2 - \|\mathbf{q}\|^2)\mathbf{v} + 2(\mathbf{q} \cdot \mathbf{v})\mathbf{q} + 2q_0(\mathbf{q} \times \mathbf{v}) \tag{7.44a}$$

$$= \mathbf{v} + q_0 \mathbf{t} + \mathbf{q} \times \mathbf{t}, \quad \text{where } \mathbf{t} = 2(\mathbf{q} \times \mathbf{v}). \tag{7.44b}$$

Being expressions of only $q_0$, $\mathbf{q}$ and $\mathbf{v}$, these two equations *embody* how unit quaternions, specifically

$$^A_B q = (\cos \tfrac{\theta_a}{2}, \ \hat{\mathbf{e}} \sin \tfrac{\theta_a}{2}) \tag{7.45a}$$

$$= (\cos \tfrac{\theta_a}{2}, \ e_x \sin \tfrac{\theta_a}{2}, \ e_y \sin \tfrac{\theta_a}{2}, \ e_z \sin \tfrac{\theta_a}{2}), \tag{7.45b}$$

can be used to completely specify arbitrary 3D rotations—in this case, the rotation from {A} to {B}. Equation (7.44a) is the traditional expanded form of the equation, and Equation (7.44b) is a more implementation-efficient form, requiring less floating point operations in code. From an analytical perspective, there is an even simpler form of the equation however. Given that the conjugate of a quaternion is given by

$$q^* = (q_0, -\mathbf{q}) = (w, -x, -y, -z), \tag{7.46}$$

one can derive from Equations (7.33) and (7.37) that

$$\begin{aligned} q\mathbf{v}q^* &= (q_0, \mathbf{q})(0, \mathbf{v})(q_0, -\mathbf{q}) \\ &= (q_0, \mathbf{q})(\mathbf{q} \cdot \mathbf{v}, \ q_0\mathbf{v} - \mathbf{v} \times \mathbf{q}) \\ &= \left(\mathbf{q} \cdot (\mathbf{v} \times \mathbf{q}), \ q_0(q_0\mathbf{v} - \mathbf{v} \times \mathbf{q}) + (\mathbf{q} \cdot \mathbf{v})\mathbf{q} + \mathbf{q} \times (q_0\mathbf{v} - \mathbf{v} \times \mathbf{q})\right) \\ &= (0, \ q_0^2\mathbf{v} + 2q_0(\mathbf{q} \times \mathbf{v}) + (\mathbf{q} \cdot \mathbf{v})\mathbf{q} - \mathbf{q} \times (\mathbf{v} \times \mathbf{q})). \end{aligned} \tag{7.47}$$

By expansion of the vector triple product $\mathbf{q} \times (\mathbf{v} \times \mathbf{q})$, this can be further simplified to[6]

$$\begin{aligned} q\mathbf{v}q^* &= (0, \ q_0^2\mathbf{v} + 2q_0(\mathbf{q} \times \mathbf{v}) + (\mathbf{q} \cdot \mathbf{v})\mathbf{q} - \mathbf{v}(\mathbf{q} \cdot \mathbf{q}) + \mathbf{q}(\mathbf{q} \cdot \mathbf{v})) \\ &= (0, \ (q_0^2 - \|\mathbf{q}\|^2)\mathbf{v} + 2(\mathbf{q} \cdot \mathbf{v})\mathbf{q} + 2q_0(\mathbf{q} \times \mathbf{v})) \\ &= (0, \mathbf{w}) \\ &= \mathbf{w}. \end{aligned} \tag{7.48}$$

---

6 Vector triple product: $\mathbf{a} \times (\mathbf{b} \times \mathbf{c}) = \mathbf{b}(\mathbf{a} \cdot \mathbf{c}) - \mathbf{c}(\mathbf{a} \cdot \mathbf{b})$ for all $\mathbf{a}, \mathbf{b}, \mathbf{c} \in \mathbb{R}^3$.

Thus, a simplified alternative equation to Equation (7.44) for applying a unit quaternion rotation $q = (q_0, \mathbf{q})$ to a vector $\mathbf{v} \in \mathbb{R}^3$, is given by

$$L_q(\mathbf{v}) = q\mathbf{v}q^*, \tag{7.49}$$

where we introduce the general notation $L_q(\cdot) : \mathbb{R}^3 \rightarrow \mathbb{R}^3$ for the vector that results when rotating the enclosed vector by the specified unit quaternion $q$. We recall that the rotation that corresponds to a unit quaternion $q$, can be interpreted geometrically in an axis-angle sense using Equation (7.45). It is trivial to see that[7]

$$L_p(L_q(\mathbf{v})) = pq\mathbf{v}q^*p^* = (pq)\mathbf{v}(pq)^* = L_{pq}(\mathbf{v}). \tag{7.50}$$

Thus, rotation composition is equivalent to standard multiplication of the associated quaternions, just as it was for rotation matrices.

In summary, the set of all quaternions is given by

$$\mathbb{H} = \{q = (q_0, \mathbf{q}) = (w, x, y, z) \in \mathbb{R}^4\}, \tag{7.51}$$

while as we have seen, the subset thereof of all quaternions that correspond to 3D rotations is given by

$$\mathbb{Q} = \{q \in \mathbb{H} : \|q\| = 1\}, \tag{7.52}$$

where in particular,

$$q_0^2 + \|\mathbf{q}\|^2 = w^2 + x^2 + y^2 + z^2 = 1. \tag{7.53}$$

Exactly two quaternions correspond to each particular physical rotation, namely $q$ and $-q$, as

$$L_q(\mathbf{v}) \equiv L_{-q}(\mathbf{v}), \tag{7.54}$$

so usually when referring to 'quaternions' in the context of rotations, only the unit quaternions $\mathbb{Q}$ are meant, and

$$q \equiv -q, \tag{7.55}$$

i.e. $q$ and $-q$ are considered to be equivalent. When specifically referring to the rotations between certain frames, the same basis notation as for rotation matrices is used. That is, $_B^A q$ is the rotation from frame {A} to frame {B}, and $_A^B q$ is the opposite (inverse) rotation.

The set of all quaternion rotations is given by the $\mathbb{Q} \cong \mathcal{S}^3$ hypersphere, i.e. the four-dimensional unit sphere, with antipodal points being considered as equivalent. As such, each 3D rotation is equivalent to a so-called quaternion line, which is a line through the origin in 4D space, and clearly intersects $\mathcal{S}^3$ at exactly one pair of antipodal points. Quaternion lines can be thought of as the linear subspace or 'direction' of a particular pair of equivalent quaternion rotations.

---

7 Note from Equations (7.37) and (7.46) that $(pq)^* = q^*p^*$ (see also Section 7.1.2.1).

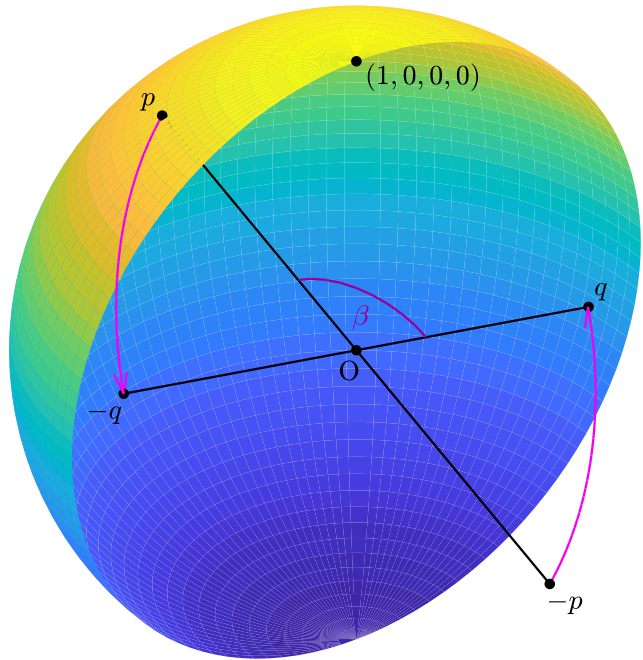

Figure 7.3: Three-dimensional slice of the quaternion hypersphere $\mathbb{Q} \cong \mathcal{S}^3 \subset \mathbb{R}^4$, showing two arbitrary quaternion lines $\pm p$ and $\pm q$. Each quaternion line corresponds to a unique rotation in 3D, and is given by a diameter through the origin $O$ joining two antipodal points on the surface of the sphere. The angle $\Theta(p, q) \in [0, \frac{\pi}{2}]$ between the two quaternion lines is defined as the minimum of the two intersection angles at $O$, i.e. the minimum of $\beta$ and $\pi - \beta$, as given in Equation (7.89) on page 199. The direct 3D axis-angle rotation from $\pm p$ to $\pm q$, i.e. from the frame defined by $\pm p$ to the frame defined by $\pm q$, is equivalently given by either of the magenta great arcs (on the surface of the sphere), as they correspond to the smaller of the angles $\beta$ and $\pi - \beta$ in this case. The magnitude $\theta_{pq}$ of the axis-angle rotation from $\pm p$ to $\pm q$ is given by $2\Theta(p, q)$.

Quaternion lines and the unit sphere can be difficult to picture in four dimensions, but for the purpose of thinking about quaternion rotations and algorithms involving them, it is often useful to imagine the three-dimensional case, i.e. the sphere $\mathcal{S}^2$, shown in Figure 7.3. In this way, the direct rotation between two orientations, i.e. quaternion lines, can for example be thought of as the shortest great arc joining them on this sphere, and this way of thinking can easily be extrapolated to four dimensions.

### 7.1.1.4 *Rotation, Vector and Basis Interactions*

As was demonstrated in Section 7.1.1.2, there are multiple ways in which rotations can be seen to interact with, and transform, vectors and other rotations. Examples included their effect as sequential local rotations, as well as changes of rotation and vector basis. The possible interactions between vectors, rotations and their respective bases are summarised in this section.

Suppose we have two frames, {G} and {B}, and can quantify the rotation from the former to the latter in terms of the corresponding rotation matrix ${}_{B}^{G}R$, or quaternion ${}_{B}^{G}q$. If this rotation maps a particular vector $\mathbf{v}$ to the vector $\mathbf{w}$, then

$$ {}^{G}\mathbf{w} = {}_{B}^{G}R\,{}^{G}\mathbf{v}, \qquad\qquad {}^{G}\mathbf{w} = L_{{}_{B}^{G}q}({}^{G}\mathbf{v}). \qquad (7.56)$$

Viewing the rotation from {G} to {B} instead as a change of reference frame (i.e. change of vector basis) gives

$$ {}^{G}\mathbf{w} = {}_{B}^{G}R\,{}^{B}\mathbf{w}, \qquad\qquad {}^{G}\mathbf{w} = L_{{}_{B}^{G}q}({}^{B}\mathbf{w}). \qquad (7.57)$$

These equations effectively describe how a vector relative to one coordinate frame can numerically be expressed with respect to another. This is a frequently required operation when working with rotations.

The next step to Equations (7.56) and (7.57) is to see how rotations interact with other rotations, instead of other vectors. A distinction must be made for this purpose between local and global rotations. Suppose again that we have the two frames {G} and {B}. In applications, {G} will frequently be the global reference frame that is attached to the environment, and {B} will often be a local reference frame that is attached to a moving body, such as a robot. If ${}_{B}^{G}R$ and ${}_{B}^{G}q$ are once again the rotation matrix and quaternion representations of the rotation from {G} to {B} respectively, then

- Rotating frame {B} about a *global axis* ${}^{G}\hat{\mathbf{e}}$ corresponds to *pre-multiplication* of ${}_{B}^{G}R$ and ${}_{B}^{G}q$ by the axis-angle rotation $({}^{G}\hat{\mathbf{e}}, \theta_a)$, and,

- Rotating frame {B} about a *local axis* ${}^{B}\hat{\mathbf{e}}$ corresponds to *post-multiplication* of ${}_{B}^{G}R$ and ${}_{B}^{G}q$ by the axis-angle rotation $({}^{B}\hat{\mathbf{e}}, \theta_a)$.

In short:

$$\text{Global axis} \Rightarrow {}^{G}\hat{\mathbf{e}} \Rightarrow \text{Premultiply}$$
$$\text{Local axis} \Rightarrow {}^{B}\hat{\mathbf{e}} \Rightarrow \text{Post-multiply}$$

In terms of mathematical equations, if the result of the total rotation in each case is labelled frame {C}, then expressed using rotation matrices the above results can be written as

$$\textbf{Global axis:} \qquad {}_{C}^{G}R = R_{{}^{G}\hat{\mathbf{e}}}(\theta_a)\,{}_{B}^{G}R \qquad (7.58a)$$

$$\textbf{Local axis:} \qquad {}_{C}^{G}R = {}_{B}^{G}R\,R_{{}^{B}\hat{\mathbf{e}}}(\theta_a) \qquad (7.58b)$$

where $R_{\hat{\mathbf{e}}}(\theta_a)$ is general notation for the rotation matrix corresponding to a CCW rotation of $\theta_a$ radians about the arbitrary vector $\hat{\mathbf{e}}$.

Equation (7.58) can be generalised in terms of basis notation to provide further insight into the interactions between successive rotations. For global axis rotations, $R_{G_{\hat{e}}}(\theta_a)$ essentially represents any rotation relative to {G} to some other frame, call it {D}. Thus, if a global rotation relative to {G} that rotates {G} onto {D} is applied to {B}, then {B} ends up as the new frame {C}, where

$$\prescript{G}{C}{R} = \prescript{G}{D}{R}\,\prescript{G}{B}{R}, \qquad\qquad \prescript{G}{C}{q} = \prescript{G}{D}{q}\,\prescript{G}{B}{q}. \qquad (7.59)$$

Similarly, for local axis rotations, $R_{B_{\hat{e}}}(\theta_a)$ essentially represents any rotation relative to {B} to some other frame, call it {C}. Thus, if a frame {B} is rotated relative to itself to end up as frame {C}, then

$$\prescript{G}{C}{R} = \prescript{G}{B}{R}\,\prescript{B}{C}{R}, \qquad\qquad \prescript{G}{C}{q} = \prescript{G}{B}{q}\,\prescript{B}{C}{q}. \qquad (7.60)$$

While nominally this equation expresses the effect of locally rotating a coordinate frame to get a new total rotation, it can also be viewed as a change of reference frame equation for rotations—$\prescript{B}{C}{R}$ expresses the orientation of frame {C} relative to frame {B}, and premultiplying by $\prescript{G}{B}{R}$ yields $\prescript{G}{C}{R}$, the orientation of the same frame relative to the reference frame {G} instead. This process is strongly reminiscent of Equation (7.57), the change of reference frame equation for vectors, and in fact both equations can be viewed as 'cancelling' the '*B*' frames across the forwards diagonal on the right-hand side, i.e. the diagonal from bottom left to top right, to give the result on the left-hand side.

### 7.1.1.5  *Referenced Rotations*

The standard basis notation $\prescript{A}{B}{R}$ for rotations is useful, and widely used in literature, but not the entire story. Suppose we have two arbitrary rotations $\prescript{G}{A}{R}$ and $\prescript{G}{B}{R}$ relative to the same global reference frame {G}.[8] For most of the ways these two rotations can be combined with each other and each other's inverses, the standard basis notation provides an interpretation of the result (see upper part of Table 7.1). For combinations like $\prescript{G}{B}{R}\,\prescript{A}{G}{R}$ however, there is no direct standard basis interpretation. In order to understand the meaning of such rotation combinations, we introduce the concept of referenced rotations. Referenced rotations are a complete generalisation of the entire standard basis notation, and are a novel contribution of this thesis.

Given two frames {A} and {B}, the rotation matrix $\prescript{A}{B}{R}$ numerically corresponds to the rotation from {A} to {B}, expressed as a rotation by some angle $\theta_a$ about some axis $\prescript{A}{}{\hat{e}}$, i.e. an axis numerically quantified relative to {A}. The rotation from {A} to {B} can however equivalently be quantified as a rotation relative to {G}, i.e. a rotation by $\theta_a$ about the same axis, but with the axis numerically quantified relative to {G}.

---

8 In this section we state all equations in terms of rotation matrices only, but equivalent statements using any other rotation representation, in particular quaternions, can of course trivially be made.

Table 7.1: List of fundamental combinations of rotations

| Expression | Eqn. | Interpretation |
|---|---|---|
| ${}^{G}_{A}R\,{}^{G}_{B}R = {}^{G}_{C}R$ | (7.59) | *Sequential rotation:* Rotation of frame {B} by the global axis rotation from {G} to {A} |
| ${}^{G}_{B}R\,{}^{G}_{A}R = {}^{G}_{D}R$ | (7.59) | *Sequential rotation:* Rotation of frame {A} by the global axis rotation from {G} to {B} |
| ${}^{A}_{G}R\,{}^{G}_{B}R = {}^{A}_{B}R$ | (7.60) | *Sequential rotation / Change of reference frame:* Local axis rotations from {A} to {G} to {B} |
| ${}^{B}_{G}R\,{}^{G}_{A}R = {}^{B}_{A}R$ | (7.60) | *Sequential rotation / Change of reference frame:* Local axis rotations from {B} to {G} to {A} |
| ${}^{G}_{A}R\,{}^{B}_{G}R = {}^{GB}_{A}R$ | (7.61) | *Referenced rotation:* Rotation from {B} to {A}, numerically quantified relative to {G} |
| ${}^{G}_{B}R\,{}^{A}_{G}R = {}^{GA}_{B}R$ | (7.61) | *Referenced rotation:* Rotation from {A} to {B}, numerically quantified relative to {G} |

This is referred to as the rotation from frame {A} to {B} *referenced by* {G}. It can be thought of as the global rotation relative to {G} that maps {A} onto {B}, and is given by

$$ {}^{GA}_{B}R = {}^{G}_{B}R\,{}^{A}_{G}R \tag{7.61a} $$

$$ = {}^{G}_{B}R\,{}^{G}_{A}R^{T}. \tag{7.61b} $$

Given the concept of this so-called *referenced rotation*, the lower part of Table 7.1 can now be completed. Referenced rotations are clearly just a generalisation of the standard basis notation, as one can trivially observe from the definition that

$$ {}^{G}_{B}R \equiv {}^{GG}_{B}R. \tag{7.62} $$

Furthermore, for any frames {A} and {G},

$$ {}^{GA}_{G}R = {}^{A}_{G}R, \tag{7.63a} $$

$$ {}^{GA}_{A}R = \mathbb{I}. \tag{7.63b} $$

Alternative mathematical formulations of the definition of referenced rotations include

$$ {}^{GA}_{B}R = {}^{G}_{A}R\,{}^{A}_{B}R\,{}^{G}_{A}R^{T} \tag{7.64a} $$

$$ = {}^{G}_{B}R\,{}^{A}_{B}R\,{}^{G}_{B}R^{T}. \tag{7.64b} $$

These can be derived directly from Equation (7.61) using the change of reference frame equation, Equation (7.60), but can also be understood geometrically in their own right. For instance, in the case of Equation (7.64a), given any vector relative to {G}, applying the referenced rotation ${}^{GA}_{B}R$ is logically equivalent to transforming the vector to {A} coordinates using ${}^G_A R^T = {}^A_G R$, applying the rotation ${}^A_B R$ from {A} to {B}, and transforming the result back to {G} coordinates using ${}^G_A R$.

Basic identities involving referenced rotations include

$$
{}^{GA}_B R \, {}^G_A R = {}^G_B R, \tag{7.65a}
$$

$$
{}^B_G R \, {}^{GA}_B R = {}^A_G R, \tag{7.65b}
$$

but these are both just special cases of the general composition of referenced rotations relation

$$
{}^{GA}_C R = {}^{GB}_C R \, {}^{GA}_B R. \tag{7.66}
$$

As expected, the inverse of a referenced rotation is given by

$$
{}^{GA}_B R^T = {}^{GB}_A R, \tag{7.67}
$$

A change of reference frame from {G} to {H} of a referenced rotation ${}^{GA}_B R$ is given by

$$
{}^{HA}_B R = {}^H_G R \, {}^{GA}_B R \, {}^H_G R^T. \tag{7.68}
$$

It is interesting to note that identities involving referenced rotations can often be thought of as 'cancelling' or 'combining' frames along the backwards diagonal of a rotation multiplication, i.e. the diagonal from the bottom right to the top left. For instance, in the fundamental definition of referenced rotations given in Equation (7.61a), the two '$G$' frames across the backwards diagonal on the right-hand side can be thought of as being combined at the top left, leaving over ${}^{GA}_B \square$ as the basis on the left-hand side. Another example is Equation (7.65a), where the '$A$' frames on the left-hand side can be thought of as cancelling across the backwards diagonal, leaving over ${}^{GG}_B R = {}^G_B R$ as the result.

Using referenced rotations, the basic interactions from Section 7.1.1.4 between rotations, and between rotations and vectors, can be expressed more generally. Equation (7.56) generalises to

$$
{}^G \mathbf{w} = {}^{GA}_B R \, {}^G \mathbf{v}, \tag{7.69}
$$

where $\mathbf{w}$ is the vector that results when $\mathbf{v}$ is rotated by the rotation relative to {G} that maps frame {A} onto frame {B}. Furthermore, if this rotation maps some frame {C} to another frame {D}, then

$$
{}^G_D R = {}^{GA}_B R \, {}^G_C R. \tag{7.70}
$$

With a permutation of the frame labels, this can be seen to be a direct generalisation of Equation (7.59).

### 7.1.2 Extended Properties of Rotation Matrices and Quaternions

As has been demonstrated so far, rotation matrices and quaternions are two of the most fundamental concepts for representing and working with 3D rotations, and are tightly interlinked with the notions of coordinate frames, coordinate bases and axes of rotation. In this section, we present further properties of these two representations, and discuss in particular those properties that are required by other parts of this thesis.

#### 7.1.2.1 *Inverse Rotation*

Due to the orthogonality of rotation matrices, the inverse of a rotation matrix ${}^{A}_{B}R \in \mathrm{SO}(3)$ is synonymous with its transpose, i.e.

$$ {}^{A}_{B}R^{-1} = {}^{A}_{B}R^{T} = {}^{B}_{A}R. \tag{7.71} $$

This means for $R, S \in \mathrm{SO}(3)$, that the standard properties of the matrix inverse and transpose hold, namely

$$ (RS)^{-1} = S^{-1}R^{-1}, \tag{7.72a} $$

$$ (RS)^{T} = S^{T}R^{T}, \tag{7.72b} $$

as well as further properties such as

$$ RR^{T} = R^{T}R = \mathbb{I}. \tag{7.73} $$

For quaternion rotations, i.e. quaternions of unit norm, the inverse of a rotation ${}^{A}_{B}q \in \mathbb{Q}$ is synonymous with its conjugate, i.e.

$$ {}^{A}_{B}q^{-1} = {}^{A}_{B}q^{*} = {}^{B}_{A}q. \tag{7.74} $$

That is, for $q = (q_0, \mathbf{q}) = (w, x, y, z) \in \mathbb{Q}$,

$$ q^{-1} = q^{*} = (q_0, -\mathbf{q}) = (w, -x, -y, -z) \in \mathbb{Q}, \tag{7.75} $$

and for $p, q \in \mathbb{Q}$, further properties of the quaternion conjugate, i.e. inverse, include

$$ (pq)^{*} = q^{*}p^{*}, \tag{7.76a} $$

$$ qq^{*} = q^{*}q = (1, 0, 0, 0). \tag{7.76b} $$

Note that $(1, 0, 0, 0) \in \mathbb{Q}$ is the identity rotation as a quaternion.

### 7.1.2.2 *Closest Valid Rotation*

Thinking of quaternion rotations as the points on the 4D unit hypersphere, it is clear that the closest valid rotation to any (possibly non-unit) quaternion $\tilde{q} \in \mathbb{H}$ is given by

$$q = \frac{\tilde{q}}{\|\tilde{q}\|}. \tag{7.77}$$

This is referred to as renormalisation, and is valid because the closest point on a sphere to any arbitrary point in space is given by the radial surface normal joining the point to the sphere. For rotation matrices, finding the closest valid $R \in SO(3)$ to an arbitrary $\tilde{R} \in \mathbb{R}^{3 \times 3}$ is referred to as reorthogonalisation, and involves calculating

$$R = \tilde{R}(\tilde{R}^T \tilde{R})^{-\frac{1}{2}}. \tag{7.78}$$

If $\tilde{R}$ is already a valid rotation matrix, then $\tilde{R}^T \tilde{R} = \mathbb{I}$, so clearly $R = \tilde{R}$. Otherwise, $\tilde{R}^T \tilde{R}$ is a real symmetric positive semidefinite matrix, so it is diagonalisable by a real orthogonal matrix $P$, and yields a diagonal matrix $D$ with non-negative entries. That is,

$$\tilde{R}^T \tilde{R} = PDP^{-1}. \tag{7.79}$$

We define $D^{-\frac{1}{2}}$ to be the diagonal matrix with the positive inverse square roots of the diagonal elements of $D$ on its own diagonal. Then, Equation (7.78) can be evaluated using

$$(\tilde{R}^T \tilde{R})^{-\frac{1}{2}} = PD^{-\frac{1}{2}}P^{-1}. \tag{7.80}$$

In numerical implementations, care has to be taken regarding the possibility of obtaining tiny non-positive entries on the diagonal of $D$ due to machine precision and floating point inaccuracies. For severely invalid $\tilde{R}$, there is also the possibility that the reorthogonalisation process returns a left-handed coordinate system, i.e. one where $\det(R) = -1$. For lack of a better option, in such cases $R$ can just be set to the identity rotation $\mathbb{I}$.

### 7.1.2.3 *Relation to Axis-Angle Representation*

As demonstrated in Section 7.1.1.1, by Euler's rotation theorem every 3D rotation can be expressed as a single rotation of up to 180° about some vector in 3D space. The standardised axis-angle representation $(\hat{\mathbf{e}}, \theta_a) \in \mathbb{A}$ captures this property, and models each rotation as a rotation of $\theta_a \in [0, \pi]$ radians about a unit vector axis $\hat{\mathbf{e}} \in \mathbb{R}^3$ (see Section 5.3.2). It is important to understand how rotations described in this way relate to rotations described by the means of quaternions and rotation matrices.

Recalling that $q \equiv -q$ for rotations, we can see from Equation (7.45) that the pair of quaternions corresponding to the axis-angle representation $(\hat{\mathbf{e}}, \theta_a) \in \mathbb{A}$ is given by

$$q = \pm \left( \cos \tfrac{\theta_a}{2}, \, \hat{\mathbf{e}} \sin \tfrac{\theta_a}{2} \right) \tag{7.81a}$$

$$= \pm \left( \cos \tfrac{\theta_a}{2}, \, e_x \sin \tfrac{\theta_a}{2}, \, e_y \sin \tfrac{\theta_a}{2}, \, e_z \sin \tfrac{\theta_a}{2} \right), \tag{7.81b}$$

where $\hat{\mathbf{e}} = (e_x, e_y, e_z) \in \mathbb{R}^3$ and $\|\hat{\mathbf{e}}\| = 1$, i.e. $\hat{\mathbf{e}} \in \mathcal{S}^2$. Keeping in mind that $\theta_a \in [0, \pi]$, the reverse conversion from a quaternion $q = (q_0, \mathbf{q}) \in \mathbb{Q}$ to its equivalent axis-angle representation is thus given by

$$\hat{\mathbf{e}} = \text{sign}(q_0) \frac{\mathbf{q}}{\|\mathbf{q}\|}, \tag{7.82a}$$

$$\theta_a = 2 \operatorname{atan2} \big( \|\mathbf{q}\|, |q_0| \big), \tag{7.82b}$$

where it should be noted that $\text{sign}(0) = 1$, and that $\hat{\mathbf{e}}$ is normally chosen to be $\mathbf{0}$ for numerical convenience when $\|\mathbf{q}\| = 0$, but theoretically can be arbitrary, as in this case $\theta_a = 0$ (identity rotation). One of the advantages of Equation (7.82) is that it is quaternion magnitude independent. This means that in code implementations it may be possible to optimise away any preceding quaternion normalisation steps, saving on computation and increasing the robustness to scale drift.

Suppose now that we have two unit quaternions $p, q \in \mathbb{Q}$, and wish to quantify the magnitude $\theta_{pq}$ of the direct axis-angle rotation between the coordinate frames that they respectively define. One option would be to construct the local rotation from one of the two frames to the other, e.g. $p^* q$, and apply Equation (7.82b) to get $\theta_a = \theta_{pq}$, but a more direct approach is to notice that

$$p^* q = (p \cdot q, \, \square, \, \square, \, \square), \tag{7.83}$$

i.e. the scalar component of $p^* q$ is equal to the quaternion dot product $p \cdot q$.[9] As a result, accounting for the fact that $p \cdot q$ might be negative, we have from Equation (7.81b) that

$$|p \cdot q| = \cos \tfrac{\theta_{pq}}{2}. \tag{7.84}$$

Consequently,

$$\theta_{pq} = 2 \operatorname{acos} |p \cdot q|, \tag{7.85}$$

where as expected, $\theta_{pq} \in [0, \pi]$, as no two coordinate frames can be more than 180° apart by direct axis-angle rotation.

The quaternions $p$ and $q$ are just normal unit vectors in $\mathbb{R}^4$, so

$$p \cdot q = \|p\| \|q\| \cos \beta = \cos \beta, \tag{7.86}$$

---

9 This is a dot product of quaternions as vectors in $\mathbb{R}^4$, and is in fact true for each of $p^* q$, $q^* p$, $q p^*$ and $p q^*$, where the first two are the local rotations from $p$ to $q$ and vice versa, and the last two are the equivalent rotations referenced by the global frame.

where $\beta \in [0, \pi]$ is the angle in 4D space between the two quaternion vectors $p$ and $q$. Observing that

$$\cos(\pi - \beta) = -\cos\beta, \tag{7.87}$$

we can deduce from Equations (7.84) and (7.86) that

$$\cos \tfrac{\theta_{pq}}{2} = |p \cdot q| = |\cos\beta| = \begin{cases} \cos\beta & \text{if } \beta \in [0, \tfrac{\pi}{2}], \\ \cos(\pi - \beta) & \text{if } \beta \in (\tfrac{\pi}{2}, \pi]. \end{cases} \tag{7.88}$$

Defining the angle

$$\Theta(p, q) = \begin{cases} \beta & \text{if } \beta \in [0, \tfrac{\pi}{2}], \\ \pi - \beta & \text{if } \beta \in (\tfrac{\pi}{2}, \pi], \end{cases} \in [0, \tfrac{\pi}{2}], \tag{7.89}$$

it follows from Equation (7.88) and $\frac{\theta_{pq}}{2} \in [0, \tfrac{\pi}{2}]$ that

$$\Theta(p, q) = \tfrac{1}{2}\theta_{pq}. \tag{7.90}$$

To put this significant result into context, consider the following alternative definitions of $\Theta(p, q)$, which can be seen to be entirely equivalent to Equation (7.89):

1. The smaller of the two angles formed at the 'X' intersection of the two quaternion lines $p$ and $q$ (see Figure 7.3 and the definition of quaternion lines on page 190),

2. The minimum angle between any combination of $\pm p$ and $\pm q$,

3. The minimum angle between any combination of $p$ and $\pm q$,

4. The angle between the quaternion vectors $p$ and $q$ if they are in some shared 4D hemisphere, otherwise the complement of this angle.

Thus, what Equation (7.90) is truly expressing is that the angle between two quaternion lines in 4D space is always exactly half the angle that the quaternion rotations are actually separated by in terms of direct axis-angle rotations in 3D space. This is a very useful characterisation to assist in gaining an intuitive and geometric understanding of how quaternions work, and in particular how the 4D quaternion unit sphere should be imagined, and can be seen to relate to real physical 3D rotations.

The situation for rotation matrices is slightly more complicated intuitively and geometrically, as they are nine-dimensional descriptions of rotation, but mathematically it can easily be derived that the rotation matrix corresponding to the CCW rotation of $\theta_a$ radians about the unit vector $\hat{\mathbf{e}} = (e_x, e_y, e_z) \in \mathcal{S}^2$ is given by

$$R_{\hat{\mathbf{e}}}(\theta_a) = \begin{bmatrix} e_x^2 C_{\theta_a} + c_{\theta_a} & e_x e_y C_{\theta_a} - e_z s_{\theta_a} & e_x e_z C_{\theta_a} + e_y s_{\theta_a} \\ e_x e_y C_{\theta_a} + e_z s_{\theta_a} & e_y^2 C_{\theta_a} + c_{\theta_a} & e_y e_z C_{\theta_a} - e_x s_{\theta_a} \\ e_x e_z C_{\theta_a} - e_y s_{\theta_a} & e_y e_z C_{\theta_a} + e_x s_{\theta_a} & e_z^2 C_{\theta_a} + c_{\theta_a} \end{bmatrix}, \quad (7.91)$$

where $c_{\theta_a} \equiv \cos\theta_a$, $s_{\theta_a} \equiv \sin\theta_a$, and

$$C_{\theta_a} = 1 - \cos\theta_a. \quad (7.92)$$

By observation from Equation (7.91), if $R \in \mathrm{SO}(3)$ is a rotation matrix corresponding to a rotation of magnitude $\theta_a$ radians,

$$\mathrm{tr}(R) = 1 + 2\cos\theta_a. \quad (7.93)$$

Thus, the angle magnitude $\theta_a \in [0, \pi]$ can be extracted from a rotation matrix $R$ using the formula

$$\theta_a = \mathrm{acos}\big(\tfrac{1}{2}(\mathrm{tr}(R) - 1)\big). \quad (7.94)$$

The extraction of the axis of rotation $\hat{\mathbf{e}} \in \mathcal{S}^2$ is more complicated, and requires an approach based on *dominant regions* (see Section 5.5.5.2) for the greatest numerical stability, namely

$$\hat{\mathbf{e}} = \frac{\tilde{\mathbf{e}}}{\|\tilde{\mathbf{e}}\|}, \quad (7.95)$$

where

$$\tilde{\mathbf{e}} = \begin{cases} \delta(R_{32}-R_{23},\ R_{13}-R_{31},\ R_{21}-R_{12}) & \text{if } R \in \mathcal{D}_w, \\ \delta(1+R_{11}-R_{22}-R_{33},\ R_{21}+R_{12},\ R_{13}+R_{31}) & \text{if } R \in \mathcal{D}_x, \\ \delta(R_{21}+R_{12},\ 1-R_{11}+R_{22}-R_{33},\ R_{32}+R_{23}) & \text{if } R \in \mathcal{D}_y, \\ \delta(R_{13}+R_{31},\ R_{32}+R_{23},\ 1-R_{11}-R_{22}+R_{33}) & \text{if } R \in \mathcal{D}_z, \end{cases} \quad (7.96\text{a})$$

$$\delta = \begin{cases} 1 & \text{if } R \in \mathcal{D}_w, \\ \mathrm{sign}(R_{32}-R_{23}) & \text{if } R \in \mathcal{D}_x, \\ \mathrm{sign}(R_{13}-R_{31}) & \text{if } R \in \mathcal{D}_y, \\ \mathrm{sign}(R_{21}-R_{12}) & \text{if } R \in \mathcal{D}_z. \end{cases} \quad (7.96\text{b})$$

If $\theta_a = \pi$, Equation (7.95) produces one of the two possible valid solutions $\pm\hat{\mathbf{e}}$ in each case. If $\theta_a = 0$, i.e. if $R$ is the identity rotation $\mathbb{I}$, a division by zero occurs and one can instead define $\hat{\mathbf{e}} = \mathbf{0}$. Strictly speaking $\mathbf{0} \notin \mathcal{S}^2$, but the axis $\hat{\mathbf{e}}$ is not uniquely defined in this case, and $\mathbf{0}$ turns out to be a convenient value for numerical implementations.

The relationships shown in Equations (7.91) and (7.95) between rotation matrices and the axis-angle representation are well formalised in terms of the Lie algebra $\mathfrak{so}(3)$ of $\mathrm{SO}(3)$. $\mathrm{SO}(3)$ is a Lie group, because it is a group under matrix multiplication, and is simultaneously a differentiable manifold with the property that the group operations of multiplication and inversion are smooth. The Lie algebra $\mathfrak{so}(3)$, on the other hand, is a vector space together with the non-associative alternating bilinear map

$$[\cdot, \cdot] : \mathfrak{so}(3) \times \mathfrak{so}(3) \to \mathfrak{so}(3)$$
$$[A, B] = AB - BA, \tag{7.97}$$

referred to as the Lie bracket (or commutator in this specific case), which must satisfy the Jacobi identity

$$[A, [B, C]] + [B, [C, A]] + [C, [A, B]] = 0. \tag{7.98}$$

While the elements of the Lie group $SO(3)$ are just the rotation matrices $R$, the Lie algebra $\mathfrak{so}(3)$ is by definition the tangent space to $SO(3)$ at the identity element $\mathbb{I}$, and thus can be shown to consist of all real skew-symmetric matrices, i.e. matrices $A$ such that

$$A^T = -A. \tag{7.99}$$

The tangent spaces at all other elements $R$ of the Lie group $SO(3)$ can be constructed by post-multiplying the elements of $\mathfrak{so}(3)$ by the respective element $R$,[10] so $\mathfrak{so}(3)$ can be thought of as completely capturing the local structure of $SO(3)$, at all elements. Note that by definition, all elements of $\mathfrak{so}(3)$, i.e. real skew-symmetric matrices, must be of the form

$$A = \begin{bmatrix} 0 & -a_z & a_y \\ a_z & 0 & -a_x \\ -a_y & a_x & 0 \end{bmatrix}, \tag{7.100}$$

for some $a_x, a_y, a_z \in \mathbb{R}$. As such, there is a one-to-one correspondence (i.e. isomorphism) between the elements of $\mathfrak{so}(3)$ and the vectors $\mathbf{a} = (a_x, a_y, a_z) \in \mathbb{R}^3$. In terms of mathematical notation, we say that

$$A = [\mathbf{a}]_\times = \begin{bmatrix} 0 & -a_z & a_y \\ a_z & 0 & -a_x \\ -a_y & a_x & 0 \end{bmatrix}, \tag{7.101}$$

is the cross product matrix[11] of $\mathbf{a} = (a_x, a_y, a_z)$, and

$$\mathbf{a} = \mathrm{axial}(A) = (A_{32}, A_{13}, A_{21}), \tag{7.102}$$

is the respective reverse operation. For numerical implementations,

$$\mathrm{axial}(A) = \tfrac{1}{2}(A_{32} - A_{23},\ A_{13} - A_{31},\ A_{21} - A_{12}). \tag{7.103}$$

---

10 See Equation (7.105) for a better idea why this is true.

11 Note that in addition to $A^T = -A$, we can also see for every skew-symmetric matrix that $A^3 = -\|A\|_2^2 A = -\|\mathbf{a}\|^2 A$. If $\mathbf{a}$ is a unit vector, this implies that $A^3 = -A$, which is a useful result for simplifying Taylor series expansions. See also Equation (7.113).

The $\begin{bmatrix} \cdot \end{bmatrix}_\times$ operator gets its name from the fact that

$$\begin{bmatrix} \mathbf{a} \end{bmatrix}_\times \mathbf{b} = \mathbf{a} \times \mathbf{b}, \tag{7.104}$$

for all vectors $\mathbf{b} \in \mathbb{R}^3$. As a noteworthy corollary, the Lie algebra $\mathfrak{so}(3)$ is in fact isomorphic to the vector space $\mathbb{R}^3$ with the Lie bracket given by the cross product operator.

It can now be better understood why the elements of $\mathfrak{so}(3)$ are given by Equation (7.99). Given any rotation matrix $R$, every infinitesimally neighbouring rotation matrix can be obtained by instantaneously (i.e. for an infinitesimally small amount of time) applying an angular velocity $\mathbf{\Omega}$ to the coordinate axis vectors $\mathbf{x}$, $\mathbf{y}$ and $\mathbf{z}$, i.e. the column vectors of $R$. From basic angular velocity formulas, the corresponding (tangential) rates of change of the axis vectors are given by $\mathbf{\Omega} \times \mathbf{x}$, $\mathbf{\Omega} \times \mathbf{y}$ and $\mathbf{\Omega} \times \mathbf{z}$. Putting this into matrix format, we have that for $\mathbf{\Omega} \in \mathbb{R}^3$, the rate of change of $R$ is

$$W = \begin{bmatrix} \uparrow & \uparrow & \uparrow \\ \mathbf{\Omega} \times \mathbf{x} & \mathbf{\Omega} \times \mathbf{y} & \mathbf{\Omega} \times \mathbf{z} \\ \downarrow & \downarrow & \downarrow \end{bmatrix} = \begin{bmatrix} \mathbf{\Omega} \end{bmatrix}_\times \begin{bmatrix} \uparrow & \uparrow & \uparrow \\ \mathbf{x} & \mathbf{y} & \mathbf{z} \\ \downarrow & \downarrow & \downarrow \end{bmatrix} = \begin{bmatrix} \mathbf{\Omega} \end{bmatrix}_\times R, \tag{7.105}$$

where $W$ is consequently the element of the tangent space of $\mathrm{SO}(3)$ at $R$ that corresponds to the angular velocity $\mathbf{\Omega}$. Setting $R = \mathbb{I}$ gives the required result that $\mathfrak{so}(3)$ consists of all possible real skew-symmetric matrices $\begin{bmatrix} \mathbf{\Omega} \end{bmatrix}_\times$. A more mathematically abstract approach of demonstrating the same result is to identify that for a matrix $A$ to be in the tangent space of $\mathrm{SO}(3)$ at $\mathbb{I}$, it must satisfy for infinitesimally small $\Delta t$ (i.e. in limit as $\Delta t$ approaches zero)

$$(\mathbb{I} + A\Delta t)(\mathbb{I} + A\Delta t)^T = \mathbb{I}. \tag{7.106}$$

Note that this expression is effectively just asserting the orthogonality condition of $\mathbb{I} + A\Delta t$, and expands to

$$\mathbb{I} + A^T \Delta t + A\Delta t + AA^T \Delta t^{2}\!\!\nearrow^{0} = \mathbb{I}, \tag{7.107}$$

where in limit as $\Delta t \to 0$ the $\Delta t^2$ component is insignificant compared to the other components. Thus,

$$A^T \Delta t + A\Delta t = 0, \tag{7.108}$$

and as a result,

$$A^T = -A. \tag{7.109}$$

This is a simple way of seeing that all elements of $\mathfrak{so}(3)$ must be skew-symmetric, as previously stated in Equation (7.99).

There are standard ways of mapping between Lie groups and their respective Lie algebras, namely the so-called exponential and logarithmic maps. In the case of $\mathrm{SO}(3)$, these correspond to the standard matrix exponential and matrix logarithm functions. The matrix exponential function maps the elements of $\mathfrak{so}(3)$ to the rotation matrices in SO(3), and is defined by

$$\exp : \mathfrak{so}(3) \to SO(3)$$

$$\exp(A) = \mathbb{I} + A + \frac{A^2}{2!} + \frac{A^3}{3!} + \cdots \tag{7.110}$$

If we set $A = \left[\theta_a \hat{\mathbf{e}}\right]_\times$ then in fact the matrix exponential maps $A$ *exactly* to the rotation matrix $R_{\hat{\mathbf{e}}}(\theta_a)$ corresponding to the axis-angle rotation $(\hat{\mathbf{e}}, \theta_a) \in \mathbb{A}$. As such, the exponential map can indirectly be considered as a map from the set of possible rotation vectors $\theta_a \hat{\mathbf{e}}$ to the set of possible rotation matrices, as follows,

$$\theta_a \hat{\mathbf{e}} \xrightarrow{\;[.]_\times\;} \begin{bmatrix} 0 & -\theta_a e_z & \theta_a e_y \\ \theta_a e_z & 0 & -\theta_a e_x \\ -\theta_a e_y & \theta_a e_x & 0 \end{bmatrix} \xrightarrow{\;\exp(\cdot)\;} R_{\hat{\mathbf{e}}}(\theta_a)$$

where as previously defined $R_{\hat{\mathbf{e}}}(\theta_a)$ is the rotation matrix that rotates $\theta_a$ radians CCW about the unit vector $\hat{\mathbf{e}}$. Analytically, it can be derived from Equation (7.110) that

$$R_{\hat{\mathbf{e}}}(\theta_a) = \exp\left(\left[\theta_a \hat{\mathbf{e}}\right]_\times\right) \tag{7.111a}$$

$$= \mathbb{I} + \sin\theta_a \left[\hat{\mathbf{e}}\right]_\times + (1 - \cos\theta)\left[\hat{\mathbf{e}}\right]_\times^2. \tag{7.111b}$$

This equation is known as Rodrigues' rotation formula, and can be seen to be directly equivalent to the previous Equation (7.91). The geometric proof of this result based on vectors and cross products is in fact one of the two missing links that can be used to demonstrate why the exponential map works as it does. The other missing link,

$$\exp\left(\left[\theta_a \hat{\mathbf{e}}\right]_\times\right) = \mathbb{I} + \sin\theta_a \left[\hat{\mathbf{e}}\right]_\times + (1 - \cos\theta_a)\left[\hat{\mathbf{e}}\right]_\times^2, \tag{7.112}$$

can be established by considering the Taylor series expansions of $\cos\theta_a$ and $\sin\theta_a$, and by observing that

$$\left[\hat{\mathbf{e}}\right]_\times^3 = -\left[\hat{\mathbf{e}}\right]_\times. \tag{7.113}$$

Rodrigues' rotation formula nicely demonstrates how rotation matrices can be partitioned into their symmetric and skew-symmetric parts. Abbreviating $R \equiv R_{\hat{\mathbf{e}}}(\theta_a)$, the parts are respectively given by

$$\tfrac{1}{2}(R + R^T) = \mathbb{I} + (1 - \cos\theta_a)\left[\hat{\mathbf{e}}\right]_\times^2, \tag{7.114a}$$

$$\tfrac{1}{2}(R - R^T) = \sin\theta_a \left[\hat{\mathbf{e}}\right]_\times. \tag{7.114b}$$

Given the definition of the matrix exponential function, the matrix logarithm

$$\log : SO(3) \to \mathfrak{so}(3)$$

naturally maps everything back in the opposite direction, i.e.

$$\theta_a \hat{\mathbf{e}} \xleftarrow{\text{axial}(\cdot)} \begin{bmatrix} 0 & -\theta_a e_z & \theta_a e_y \\ \theta_a e_z & 0 & -\theta_a e_x \\ -\theta_a e_y & \theta_a e_x & 0 \end{bmatrix} \xleftarrow{\log(\cdot)} R_{\hat{\mathbf{e}}}(\theta_a)$$

In general however,

$$R_{\hat{\mathbf{e}}}(\theta_a) = \exp(A) \tag{7.115}$$

has infinitely many solutions for $A$, namely

$$A_k = \big[ (\theta_a + 2\pi k)\hat{\mathbf{e}} \big]_\times \quad \text{for } k \in \mathbb{Z}, \tag{7.116}$$

so the principal logarithm of minimal Frobenius norm is taken, namely

$$A_0 = \big[ \theta_a \hat{\mathbf{e}} \big]_\times, \tag{7.117}$$

where $\theta_a \in [0, \pi]$ by definition. This is because the required Frobenius norm is given by

$$\|A_k\|_F = \sqrt{2} \, \|(\theta_a + 2\pi k)\hat{\mathbf{e}}\| = \sqrt{2} \, |\theta_a + 2\pi k|, \tag{7.118}$$

so $k = 0$ clearly minimises $\|A_k\|_F$.[12] All that remains now is how to actually calculate the principal logarithm of a rotation matrix. Based on Equation (7.114b), one method frequently quoted in literature is

$$\log(R) = \begin{cases} \dfrac{\theta_a}{2\sin\theta_a}\big(R - R^T\big) & \text{if } \theta_a \in (0, \pi), \\ 0 & \text{if } \theta_a = 0, \end{cases} \tag{7.119}$$

where $\theta_a$ is first computed using Equation (7.94). This method is however neither globally numerically stable, nor favourable, and even completely fails for $\theta_a = \pi$, i.e. rotations by 180°. If the final aim is to extract $\theta_a$ and $\hat{\mathbf{e}}$ separately, the method developed by the author in Equations (7.94) to (7.96) is significantly superior, as it leverages the robust concept of dominant regions.

### 7.1.2.4  *Relation to Angular Velocities*

The time derivatives of the quaternion and rotation matrix representations can conveniently be related to the angular velocities that they correspond to. Given a quaternion rotation $^G_B q \equiv q \in \mathbb{Q}$ and an arbitrary angular velocity vector $\mathbf{\Omega} \in \mathbb{R}^3$ in units of rad/s, the angular velocity $\mathbf{\Omega}$ applied to $q$ results in the quaternion velocity

$$\dot{q} \equiv \frac{dq}{dt} = \tfrac{1}{2} \, ^G\mathbf{\Omega} q = \tfrac{1}{2} q \, ^B\mathbf{\Omega}, \tag{7.120}$$

---

12 If $\theta_a = \pi$, then $k = 0$ and $k = -1$ are equally good solutions. This corresponds to the equivalence of rotating 180° one way or the other about a vector.

where the vectors ${}^G\boldsymbol{\Omega}$ and ${}^B\boldsymbol{\Omega}$ are being interpreted as quaternions as per Equation (7.33). The quaternion velocity $\dot{q}$ by definition must always be perpendicular to the original quaternion $q$ in 4D space, just like every vector in the tangent plane to a sphere must always be perpendicular to the corresponding radial vector. Thus,

$$\dot{q} \bullet q = 0, \tag{7.121}$$

where '$\bullet$' is the dot product. To convert from a quaternion velocity $\tilde{q} \in \mathbb{H}$ to the corresponding angular velocity, we first convert it to a valid quaternion velocity that satisfies Equation (7.121) by calculating

$$\dot{q} = \tilde{q} - (\tilde{q} \bullet q)q. \tag{7.122}$$

Note that this formula naturally assumes that $q$ is a unit quaternion. The corresponding global and local angular velocities are then respectively given by

$$ {}^G\boldsymbol{\Omega} = 2\dot{q}q^*, \tag{7.123a}$$

$$ {}^B\boldsymbol{\Omega} = 2q^*\dot{q}, \tag{7.123b}$$

where $\boldsymbol{\Omega} \equiv (0, \boldsymbol{\Omega}) \in \mathbb{H}$ is once again being used for both equations. Equation (7.122) ensures that the w-components of the right-hand side expressions in Equation (7.123) are both zero. In fact, the only effect of *not* applying Equation (7.122) is the generation of non-zero w-components in Equation (7.123), so if these non-zero components are just discarded, then the validation step of $\dot{q}$ can actually be skipped.

The case for rotation matrices is somewhat similar. Given a rotation matrix ${}^G_B R \equiv R \in SO(3)$ and an arbitrary angular velocity vector $\boldsymbol{\Omega} \in \mathbb{R}^3$ in units of rad/s, the corresponding rotation matrix velocity is given by

$$\dot{R} \equiv \frac{dR}{dt} = \left[{}^G\boldsymbol{\Omega}\right]_\times R = R\left[{}^B\boldsymbol{\Omega}\right]_\times, \tag{7.124}$$

where $\left[\,\cdot\,\right]_\times$ is the cross product matrix operator as defined in Equation (7.101). By virtue of Equations (7.20) and (7.104), Equation (7.124) can alternatively be formulated as

$$\dot{R} = \begin{bmatrix} \uparrow & \uparrow & \uparrow \\ {}^G\boldsymbol{\Omega}\times{}^G\mathbf{x}_B & {}^G\boldsymbol{\Omega}\times{}^G\mathbf{y}_B & {}^G\boldsymbol{\Omega}\times{}^G\mathbf{z}_B \\ \downarrow & \downarrow & \downarrow \end{bmatrix} \tag{7.125a}$$

$$= \begin{bmatrix} \leftarrow & -{}^B\boldsymbol{\Omega}\times{}^B\mathbf{x}_G & \rightarrow \\ \leftarrow & -{}^B\boldsymbol{\Omega}\times{}^B\mathbf{y}_G & \rightarrow \\ \leftarrow & -{}^B\boldsymbol{\Omega}\times{}^B\mathbf{z}_G & \rightarrow \end{bmatrix}. \tag{7.125b}$$

In order for a particular $\dot{R}$ to be valid, it by definition must be an element of the tangent space of $SO(3)$ at $R$, which is given by

$$T_R^{SO(3)} = \{AR \in \mathbb{R}^{3\times3} : A \in \mathbb{R}^{3\times3}, A^T = -A\} \tag{7.126a}$$

$$= \{AR \in \mathbb{R}^{3\times3} : A \in \mathfrak{so}(3)\}. \tag{7.126b}$$

Thus, any matrix $\check{R} \in \mathbb{R}^{3 \times 3}$ can be turned into its closest valid rotation matrix velocity by extracting the skew-symmetric part of $\check{R}R^T$ and post-multiplying by $R$, i.e.

$$\dot{R} = \tfrac{1}{2}\Big(\check{R}R^T - (\check{R}R^T)^T\Big)R. \qquad (7.127)$$

The corresponding global and local angular velocities are then respectively given by

$$^G\mathbf{\Omega} = \mathrm{axial}\big(\dot{R}R^T\big), \qquad (7.128a)$$

$$^B\mathbf{\Omega} = \mathrm{axial}\big(R^T\dot{R}\big). \qquad (7.128b)$$

These two equations are in fact still true if Equation (7.127) is skipped, but only as long as Equation (7.103) is used to evaluate the axial($\cdot$) function.

As a final note, if we wish to calculate the angular velocity corresponding to a finite difference between two rotations, then for either quaternions or rotation matrices the general formula is the same. If we have two frames {A} and {B} relative to the global frame {G}, then the global rotation from {A} to {B} is given by the *referenced rotation*

$$^{GA}_{\phantom{GA}B}R = {}^G_BR\,{}^G_AR^T, \qquad\qquad {}^{GA}_{\phantom{GA}B}q = {}^G_Bq\,{}^G_Aq^*. \qquad (7.129)$$

If $(\hat{\mathbf{e}}, \theta_a) \in \mathbb{A}$ (see Section 5.3.2) is the axis-angle representation of this referenced rotation, and the rotation is performed in $\Delta t$ seconds, the equivalent finite difference angular velocity is given by

$$^G\mathbf{\Omega} = \tfrac{1}{\Delta t}\theta_a\hat{\mathbf{e}}, \qquad (7.130)$$

i.e. the corresponding rotation vector divided by $\Delta t$. The required axis-angle representation of the referenced rotation can be calculated as described previously in Section 7.1.2.3. Note that if {G} $\equiv$ {A}, Equation (7.130) reduces to

$$^A\mathbf{\Omega} = \tfrac{1}{\Delta t}\theta_a\hat{\mathbf{e}}, \qquad (7.131)$$

where $(\hat{\mathbf{e}}, \theta_a)$ is the axis-angle representation of ${}^A_BR$.

### 7.1.2.5  *Continuous Real Powers of Rotations*

If $q \in \mathbb{Q}$ is the quaternion rotation that corresponds to a CCW rotation of $\theta_a$ radians about the vector $\hat{\mathbf{e}}$, then we know that

$$\begin{aligned}
q^{-1} &\Rightarrow \text{CW rotation about } \hat{\mathbf{e}} \text{ by } \theta_a \text{ radians}\\
(1,0,0,0) = q^0 &\Rightarrow \text{Rotation about } \hat{\mathbf{e}} \text{ by zero radians, i.e. identity}\\
q = q^1 &\Rightarrow \text{CCW rotation about } \hat{\mathbf{e}} \text{ by } \theta_a \text{ radians}\\
qq = q^2 &\Rightarrow \text{CCW rotation about } \hat{\mathbf{e}} \text{ by } 2\theta_a \text{ radians}
\end{aligned}$$

By an inductive argument it follows that for all integers $n \in \mathbb{Z}$, the quaternion $q^n \in \mathbb{Q}$ is equivalent to a CCW rotation of $n\theta_a$ radians about the vector $\hat{\mathbf{e}}$. By careful logical extension to the real numbers, we can see that the quaternion $q = q_{\hat{\mathbf{e}}}(\theta_a) = \left(\cos\frac{\theta_a}{2}, \hat{\mathbf{e}}\sin\frac{\theta_a}{2}\right)$ taken to the continuous real power of $u \in \mathbb{R}$ is given by

$$q^u = q_{\hat{\mathbf{e}}}(u\theta_a) = \left(\cos\frac{u\theta_a}{2}, \hat{\mathbf{e}}\sin\frac{u\theta_a}{2}\right). \tag{7.132}$$

For non-integer $u$, there are actually multiple or even infinitely many possible 'solutions' to $q^u$, but always choosing this principal value uniquely results in a continuous and smooth map from $u \in \mathbb{R}$ to $q^u \in \mathbb{Q}$ such that the latter is consistent with its definition for integer $u = n$. It can be seen from Equation (7.132) that taking a quaternion to a power can be thought of as scaling its angle of rotation $\theta_a$. As such, it follows that it functions geometrically as spherical linear interpolation/extrapolation between the identity rotation and the quaternion $q$ (see Section 7.3.5).

The case for rotation matrices is quite similar to that for quaternions in the sense that the continuous real powers of a rotation matrix $R \equiv R_{\hat{\mathbf{e}}}(\theta_a)$ are principally given by

$$R^u = R_{\hat{\mathbf{e}}}(u\theta_a), \tag{7.133}$$

even though hypothetically for non-integer $u \in \mathbb{R}$ there are multiple or infinitely many possible 'solutions'. Converting to the axis-angle representation and back again is a fairly indirect method of calculating $R^u$ however. Instead, an eigendecomposition can be used to diagonalise $R$—which is guaranteed to be possible because all normal and therefore also orthogonal matrices are diagonalisable by a unitary matrix—after which the principal $u^{\text{th}}$ powers of the resulting diagonal matrix elements can be taken. Stitching this back together with the pre- and post-multiplying unitary diagonalising matrices yields the required matrix power $R^u$.

## 7.2 VISUALISING FUSED ANGLES

As we can recall from Chapter 5, the development of the *fused angles* and *tilt phase space* rotation representations was motivated by the lack of an existing 3D rotation formalism that naturally deals with the dissolution of a complete rotation into parameters that are specifically and geometrically relevant to the balance of a body, and that does not introduce order-based asymmetry in the parameters. Vital to the definition (see Sections 5.4.4 and 5.4.5) of both new parameterisations was the idea of *fused yaw*, and how that can be used to cleanly partition rotations into meaningful *yaw* and *tilt* components (see Section 5.4.1). In this section, we present a rigorous geometrical approach to defining fused angles, explore in more detail how the *sine sum criterion* limits the pitch vs. roll domain, and visualise the definition of fused angles as the intersection of 3D cone-shaped level sets.

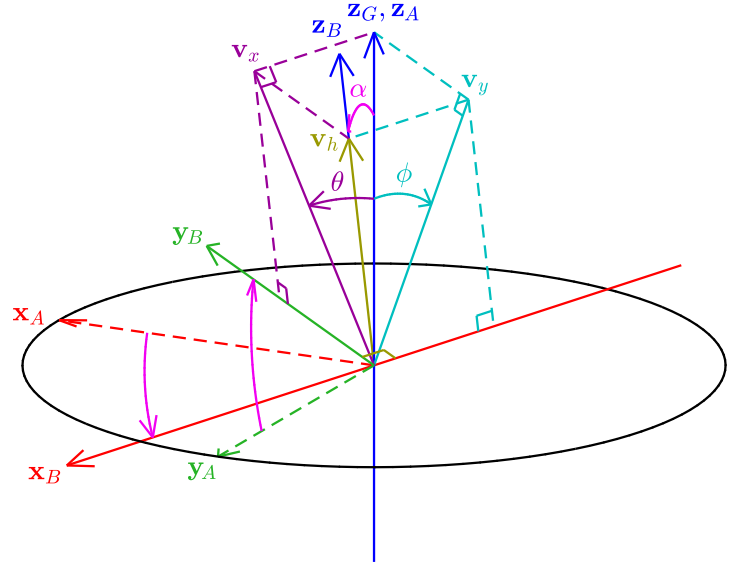

Figure 7.4: Reviewed definition of the fused angles parameters $(\theta, \phi, h)$ for a tilt rotation component from {A} to {B}. Refer to Figure 5.5 for a visual definition of the intermediate frame {A}, and the fused yaw $\psi$ used to define it. The frame {A} is rotated onto frame {B} through a tilt angle of $\alpha$ in such a way that $\mathbf{z}_A \equiv \mathbf{z}_G$ rotates directly onto $\mathbf{z}_B$. The fused pitch $\theta$ and fused roll $\phi$ are defined as the signed angles between $\mathbf{z}_G$, and $\mathbf{v}_x$ and $\mathbf{v}_y$ respectively. The hemisphere $h$ is $+1$ if $\mathbf{z}_B$ and $\mathbf{v}_h$ are parallel, and $-1$ if they are anti-parallel. For aid of visualisation, note that $\mathbf{x}_B$ is pointing left-downwards parallel to the page, and $\mathbf{y}_B$ is pointing up-leftwards out of the page.

### 7.2.1 Geometric Definition of Fused Angles

Suppose we have the 3D rotation from a global frame {G} to a local frame {B}. We recall from Section 5.4.1 that we define the intermediate frame {A} by rotating {B} in such a way that $\mathbf{z}_B$ rotates onto $\mathbf{z}_G$ in the most direct way possible, i.e. within the plane through the origin that contains these two vectors. The rotation from {G} to {A} is then a pure z-rotation, i.e. the yaw rotation component of the complete rotation from {G} to {B}, and the rotation from {A} to {B} is the tilt rotation component, as illustrated in Figure 5.5 (see page 103). The fused yaw $\psi \in (-\pi, \pi]$ is trivially defined as the CCW magnitude of the yaw rotation component, but it remains to be shown how the tilt rotation component can be rigorously geometrically described by the fused pitch, fused roll and hemisphere parameters.

Like in Section 5.4.4, let us consider only the tilt rotation component of a rotation, i.e. the component from {A} to {B}, as shown in Figure 7.4. Let $\mathbf{v}_x$ and $\mathbf{v}_y$ be the projections of the $\mathbf{z}_G$ vector onto the $\mathbf{y}_B \mathbf{z}_B$ and $\mathbf{x}_B \mathbf{z}_B$ planes respectively. We define the fused pitch of {B} as the angle $\theta \in [-\frac{\pi}{2}, \frac{\pi}{2}]$ between $\mathbf{z}_G$ and $\mathbf{v}_x$ of sign

$$-\operatorname{sign}(^B z_{Gx}) \equiv -\operatorname{sign}(^B v_{yx}).  \quad (7.134)$$

By logical completion, the magnitude of $\theta$ is taken to be $\frac{\pi}{2}$ when $\mathbf{v}_x = \mathbf{0}$. We similarly define the fused roll of {B} as the angle $\phi \in [-\frac{\pi}{2}, \frac{\pi}{2}]$ between $\mathbf{z}_G$ and $\mathbf{v}_y$ of sign

$$\operatorname{sign}(^B z_{Gy}) \equiv \operatorname{sign}(^B v_{xy}).  \quad (7.135)$$

The magnitude of $\phi$ is taken to be $\frac{\pi}{2}$ when $\mathbf{v}_y = \mathbf{0}$. As $\theta$ and $\phi$ alone can be seen to only specify a tilt rotation up to the $z$-hemisphere, that is, whether $\mathbf{z}_B$ and $\mathbf{z}_G$ are in the same 3D hemisphere or not, a fused hemisphere parameter $h$ is introduced. If $\mathbf{v}_h$ is the projection of $\mathbf{z}_G$ onto $\mathbf{z}_B$, then the hemisphere $h$ is defined to be positive $(+1)$ if $\mathbf{v}_h$ and $\mathbf{z}_B$ are parallel, and negative $(-1)$ if they are anti-parallel. That is,

$$h \equiv \operatorname{sign}(^B v_{hz}) \equiv \operatorname{sign}(^B z_{Gz}) \equiv \operatorname{sign}(^G z_{Bz}).  \quad (7.136)$$

Note that as previously indicated in this chapter, $\operatorname{sign}(\cdot)$ is defined such that $\operatorname{sign}(0) = 1$, as opposed to the standard mathematical sign function $\operatorname{sgn}(\cdot)$, which obeys $\operatorname{sgn}(0) = 0$. Given this final hemisphere parameter, the complete fused angles representation is then

$$\begin{aligned} ^G_B F &= (\psi, \theta, \phi, h) \\ &\in (-\pi, \pi] \times [-\tfrac{\pi}{2}, \tfrac{\pi}{2}] \times [-\tfrac{\pi}{2}, \tfrac{\pi}{2}] \times \{-1, 1\} \equiv \hat{\mathbb{F}}, \end{aligned}  \quad (7.137)$$

as before in Section 5.4.4. It can be observed from the presented geometric definitions that the tilt rotation component parameters depend only on $^B \mathbf{z}_G$, and that they are invariant, as crucially desired, to global yaw $z$-rotations of {B}.

## 7.2.2  Fused Angles Domain

Although both the fused pitch $\theta$ and fused roll $\phi$ are permitted to have values in the range $[-\pi, \pi]$, we can observe from Equation (5.58) and the fact that $^B \mathbf{z}_G$ is a unit vector, that

$$\begin{aligned} \sin^2 \theta + \sin^2 \phi &= (-\sin \theta)^2 + (\sin \phi)^2 \\ &= {}^B z_{Gx}^2 + {}^B z_{Gy}^2 \\ &= 1 - {}^B z_{Gz}^2. \end{aligned}  \quad (7.138)$$

Therefore, as $^B z_{Gz}^2 \geq 0$,

$$\sin^2 \theta + \sin^2 \phi \leq 1.  \quad (7.139)$$

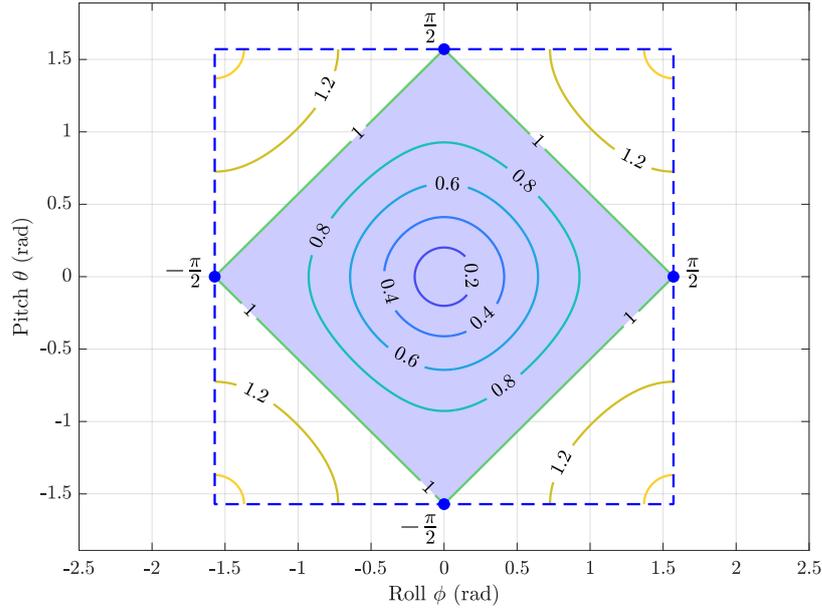

Figure 7.5: Level sets of the function $f_s(\phi, \theta) = \sqrt{\sin^2 \theta + \sin^2 \phi} = \sin \alpha$, restricted to the dashed square domain $(\phi, \theta) \in [-\frac{\pi}{2}, \frac{\pi}{2}] \times [-\frac{\pi}{2}, \frac{\pi}{2}]$. As indicated by the shaded region, the sine sum criterion $f_s(\phi, \theta) \leq 1$ is equivalent to the simpler inequality $|\phi| + |\theta| \leq \frac{\pi}{2}$, and defines the true domain of the fused pitch and roll parameters (i.e. the $\phi$ and $\theta$ components of the set $\mathbb{F}$).

This is referred to as the <span style="color:red">sine sum criterion</span>, and implies for example that $\theta$ and $\phi$ cannot both be $\frac{\pi}{2}$ at the same time, as

$$\sin^2 \tfrac{\pi}{2} + \sin^2 \tfrac{\pi}{2} = 2 \nleq 1. \tag{7.140}$$

As shown in Appendix B.1.1.1, Equation (7.139) is precisely equivalent to

$$|\theta| + |\phi| \leq \tfrac{\pi}{2}. \tag{7.141}$$

Given that by definition $\theta, \phi \in [-\frac{\pi}{2}, \frac{\pi}{2}]$ and $\alpha \in [0, \pi]$, where $\alpha$ is the *tilt angle* of a rotation, this equivalence can be visualised by plotting the level sets of the multivariate function

$$f_s : [-\tfrac{\pi}{2}, \tfrac{\pi}{2}] \times [-\tfrac{\pi}{2}, \tfrac{\pi}{2}] \to \mathbb{R}$$
$$f_s(\phi, \theta) = \sqrt{\sin^2 \theta + \sin^2 \phi} = \sin \alpha, \tag{7.142}$$

and identifying the region where $f_s(\phi, \theta) \leq 1$, as shown in Figure 7.5. The resulting diamond-shaped area in the figure can be seen to be the true domain of the fused roll and pitch parameters. As a result, we define the true domain $\mathbb{F}$ of the entire fused angles representation as the restriction of $\hat{\mathbb{F}}$ to this area, i.e. by the sine sum criterion. Formally, this yields

$$\mathbb{F} = \{F = (\psi, \theta, \phi, h) \in \hat{\mathbb{F}} : |\theta| + |\phi| \leq \tfrac{\pi}{2}\}. \tag{7.143}$$

If referring to just the fused angles tilt rotation parameters, i.e. $(\theta, \phi, h)$, then we define

$$\mathbb{F}_t = \left\{ (\theta, \phi, h) \in [-\tfrac{\pi}{2}, \tfrac{\pi}{2}] \times [-\tfrac{\pi}{2}, \tfrac{\pi}{2}] \times \{-1, 1\} : |\theta| + |\phi| \leq \tfrac{\pi}{2} \right\}. \quad (7.144)$$

### 7.2.3 Fused Angles as the Intersection of Cones

For the rotation from a frame {G} to a frame {B}, it can be seen from Equation (5.86b) that ${}^{B}\mathbf{z}_G$ completely defines the tilt rotation component of the rotation, and is given by the well-defined multivariate function

$$f_z : \mathbb{F}_t \to \mathcal{S}^2, \ (\theta, \phi, h) \mapsto {}^{B}\mathbf{z}_G$$

$$f_z(\theta, \phi, h) = \left( -\sin\theta, \ \sin\phi, \ h\sqrt{1 - \sin^2\theta - \sin^2\phi} \right). \quad (7.145)$$

Suppose for each $\hat{\theta} \in [-\tfrac{\pi}{2}, \tfrac{\pi}{2}]$ we construct the image set

$$f_z(\hat{\theta}, *, *) = \left\{ f_z(\theta, \phi, h) = {}^{B}\mathbf{z}_G \in \mathcal{S}^2 : \theta = \hat{\theta} \right\}, \quad (7.146)$$

i.e. the combined image of all fused angles tuples $(\theta, \phi, h) \in \mathbb{F}_t$ that have a fused pitch of $\hat{\theta}$. By definition, every vector ${}^{B}\mathbf{z}_G$ in the set $f_z(\hat{\theta}, *, *)$ must have the signed angle $\hat{\theta}$ between it and the $\mathbf{y}_B \mathbf{z}_B$ plane. As such, it follows that the set must be a circle about the $\mathbf{x}_B$ axis on the surface of the unit sphere $\mathcal{S}^2$. If we consider the surface that the unit vector ${}^{B}\mathbf{z}_G$ tracing out that circle would sweep however, we can picture the set of all ${}^{B}\mathbf{z}_G$ with a fused pitch of $\hat{\theta}$ as an open-ended cone. An example of such a fused pitch cone is shown in Figure 7.6a. It can clearly be observed that for all points on the cone, the angle between the corresponding vector ${}^{B}\mathbf{z}_G$ and the $\mathbf{y}_B \mathbf{z}_B$ plane is $\hat{\theta}$.

Similar image sets can also be constructed for $\hat{\phi} \in [-\tfrac{\pi}{2}, \tfrac{\pi}{2}]$ and $\hat{h} \in \{-1, 1\}$, that is,

$$f_z(*, \hat{\phi}, *) = \left\{ f_z(\theta, \phi, h) = {}^{B}\mathbf{z}_G \in \mathcal{S}^2 : \phi = \hat{\phi} \right\}, \quad (7.147a)$$

$$f_z(*, *, \hat{h}) = \left\{ f_z(\theta, \phi, h) = {}^{B}\mathbf{z}_G \in \mathcal{S}^2 : h = \hat{h} \right\}. \quad (7.147b)$$

The former is also an open-ended cone, but about $\mathbf{y}_B$, as shown in Figure 7.6b, and the latter is either the complete upper hemisphere for $\hat{h} = +1$ (see Figure 7.6c), or the complete lower hemisphere for $\hat{h} = -1$ (see Figure 7.6d). When viewing all of these image sets, referred to collectively as <span style="color:red">fused cones</span>, it is important to remember that the z-vectors that comprise them are the *global* z-vectors $\mathbf{z}_G$ relative to the rotating frame {B}, *not* the local z-vectors $\mathbf{z}_B$ relative to the fixed global frame {G}, as one might quickly intuitively think.

Knowing how the fused cones are defined and what they look like, we can now see that the specification of a fused angles tilt rotation $(\hat{\theta}, \hat{\phi}, \hat{h}) \in \mathbb{F}_t$ is nothing other than a definition of ${}^{B}\mathbf{z}_G$ by means of finding the unique intersection of the three associated image sets

$${}^{B}\mathbf{z}_G \simeq f_z(\hat{\theta}, *, *) \cap f_z(*, \hat{\phi}, *) \cap f_z(*, *, \hat{h}). \quad (7.148)$$

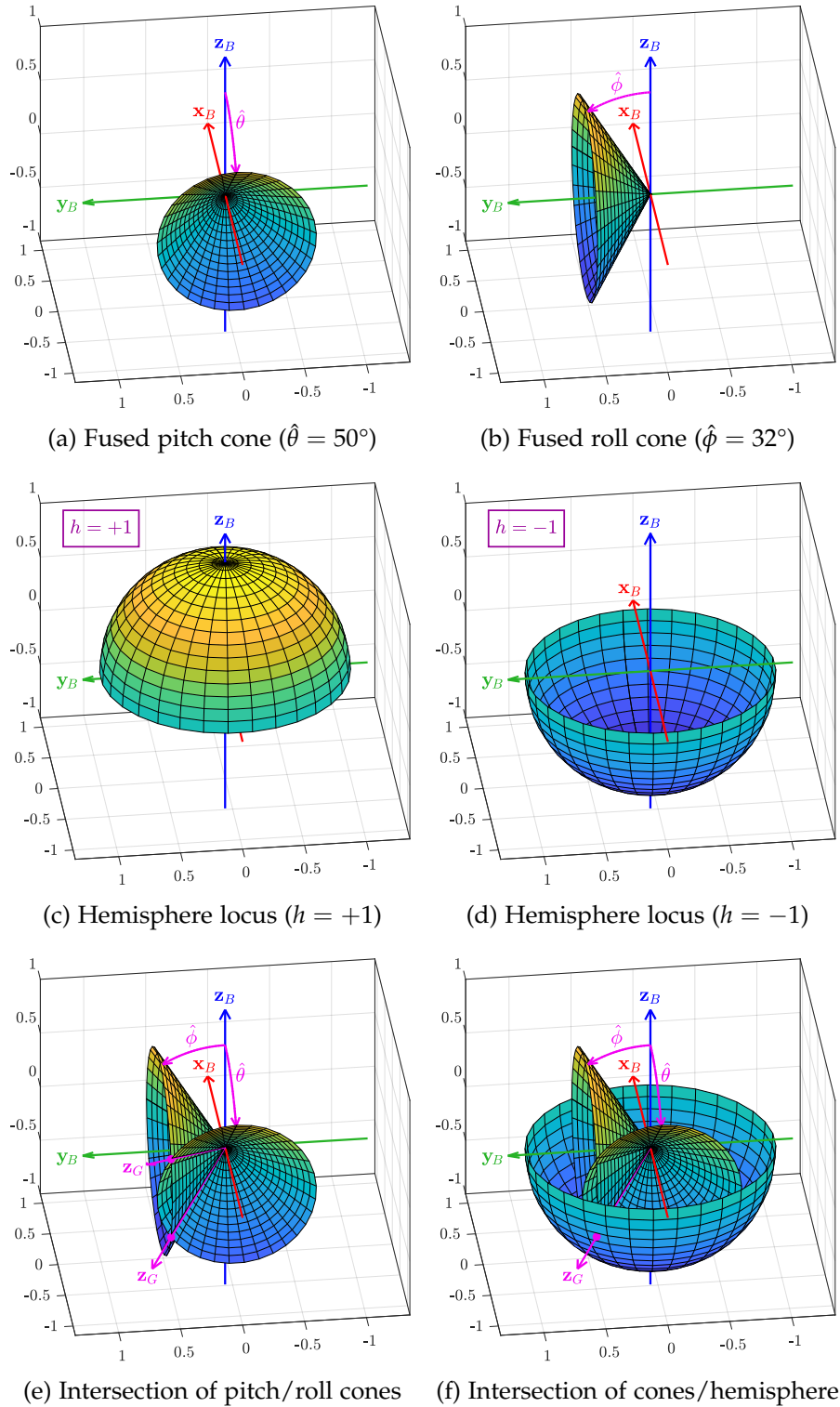

(a) Fused pitch cone ($\hat{\theta} = 50°$)

(b) Fused roll cone ($\hat{\phi} = 32°$)

(c) Hemisphere locus ($h = +1$)

(d) Hemisphere locus ($h = -1$)

(e) Intersection of pitch/roll cones

(f) Intersection of cones/hemisphere

Figure 7.6: Plots of the swept locus of ${}^{B}\mathbf{z}_G$ for various constant fused angles parameters. Plot (e) shows how a locus of constant fused pitch intersects a locus of constant fused roll at two points, one corresponding to $h = +1$ and one corresponding to $h = -1$. In (f), intersecting this again with the corresponding hemisphere locus gives the final unique specification of ${}^{B}\mathbf{z}_G$.

This kind of three-way intersection, and how it uniquely resolves a global z-vector ${}^B\mathbf{z}_G$, is illustrated in Figure 7.6f. At this point, it can clearly be geometrically understood why a hemisphere parameter is necessary for the fused angles representation. If only two fused cones corresponding to a particular fused pitch and roll are intersected, as shown in Figure 7.6e, two possible solutions for ${}^B\mathbf{z}_G$ result, as in general the two cones will always, if any, have two points of intersection—one in the upper hemisphere and one in the lower hemisphere. If a fused pitch cone and a fused roll cone fail to intersect, this is precisely equivalent to a violation of the sine sum criterion, and hence an invalid specification of $\theta$ and $\phi$. The half-apex angles of the fused pitch and roll cones are geometrically given by $\frac{\pi}{2} - |\theta|$ and $\frac{\pi}{2} - |\phi|$, so clearly a cone intersection occurs if and only if

$$(\tfrac{\pi}{2} - |\theta|) + (\tfrac{\pi}{2} - |\phi|) \geq \tfrac{\pi}{2}. \tag{7.149}$$

This trivially simplifies down to

$$|\theta| + |\phi| \leq \tfrac{\pi}{2}, \tag{7.150}$$

which is exactly the sine sum criterion.

## 7.3 FURTHER PROPERTIES OF YAW-TILT ROTATIONS

A selection of mathematical and geometric properties of the fused angles, tilt angles and tilt phase space representations was previously presented in Section 5.7. We continue here in this section with further, partially more advanced properties, and properties that are important, but not necessarily core to the understanding of the bipedal walking part of this thesis.

### 7.3.1 Axiomatic Emergence of the Tilt Phase Space

Let us consider the general arbitrary case of a parameterisation $(X, Y)$ (in some domain $\mathbb{D} \subseteq \mathbb{R}^2$) of the tilt rotation component of a general 3D rotation. That is, recalling that $\mathrm{TR}(3) \subset \mathrm{SO}(3)$ denotes the set of all 3D tilt rotations, let us consider any map

$$f_p : \mathrm{TR}(3) \to \mathbb{D}, \; R \mapsto (X, Y) \tag{7.151}$$

that allows us to assign a well-defined pair of real scalar values to every single three-dimensional tilt rotation. In order for the parameterisation $(X, Y)$ to be able to meet the motivation and aims set out in Section 5.2 without the need for any further parameters, and be generally intuitive and mathematically and geometrically useful, we formulate three simple, readily acceptable axioms:

1. The map $f_p$ should be one-to-one, onto and continuous, up to clearly defined equivalences and the adaptations to the standard topology of $\mathbb{R}^2$ that these equivalences require.

2. Pure CCW rotations by $\sigma \in (-\pi, \pi)$ radians about the x-axis should map to $(\sigma, 0) \in \mathbb{D}$, and pure CCW rotations by $\sigma$ about the y-axis should map to $(0, \sigma) \in \mathbb{D}$. That is,

$$f_p\big(R_x(\sigma)\big) = (\sigma, 0), \tag{7.152a}$$

$$f_p\big(R_y(\sigma)\big) = (0, \sigma). \tag{7.152b}$$

3. Z-rotations by $\beta \in (-\pi, \pi]$ of the definition of the reference frames used to express a rotation (see Figure 6.4) should be equivalent to rotating $(X, Y) \in \mathbb{R}^2$ by $-\beta$ about the origin. That is, if $^G_B R \in \mathrm{TR}(3)$ is any tilt rotation and {U} is the frame {G} z-rotated CCW by $\beta$ radians, then

$$f_p\big(^{UG}_B R\big) = \begin{bmatrix} \cos\beta & \sin\beta \\ -\sin\beta & \cos\beta \end{bmatrix} f_p\big(^G_B R\big). \tag{7.153}$$

Although on first sight Axiom 3 may seem complicated, it is really nothing more than a basic statement of parameter axisymmetry, and simply ensures that the parameterisation $(X, Y)$ behaves logically equivalently no matter which arbitrary global reference frame is chosen for the purpose of analysis.

To give an example of what is meant by 'equivalences' in Axiom 1, consider the example case that $X$ is the ZYX Euler roll $\phi_E \in (-\pi, \pi]$. While the standard topology of $\mathbb{R}^2$ would declare $f_p$ discontinuous, as $X = \phi_E$ jumps from $\pi$ to $-\pi$ and vice versa in certain parts of the rotation space, clearly this is not a true discontinuity. To rigorously mathematically cater for this, we define the domain of $X$ in this case to be $[-\pi, \pi]$, declare $X = \pi$ and $X = -\pi$ to be *equivalent* and thereby synonymous points for each individual $Y$, and extend the topology on $\mathbb{D}$ to also include all open sets that straddle the jump at $X = \pm\pi$ in some way.[13] Then, under consideration of $\mathbb{D}$ with this extended topology, $f_p$ is mathematically continuous. This is a simple example of how equivalences can affect the bijectiveness and continuity of $f_p$, but one should be aware that less trivial equivalences can potentially be required for more general parameterisations.

The listed axioms are modest and few in number, but it can readily be demonstrated that the tilt phase space parameterisation is in fact the only possible solution for $f_p$. The tilt phase space can thus be considered to emerge axiomatically from these complete rotation fundamentals. Mathematically, $X \equiv p_x$, $Y \equiv p_y$ and $\mathbb{D} \equiv \bar{\mathcal{D}}^2(\pi)$, the closed disc of radius $\pi$ in $\mathbb{R}^2$ about the origin, and the required equivalence relation for Axiom 1 is given by identifying all pairs of antipodal points on the circular outer boundary of $\mathbb{D}$. Interestingly, the strong property of magnitude axisymmetry does not feature in any of the axioms directly, but results from the combination of Axioms 2 and 3. Fused angles can be seen to not satisfy the axioms, as for example in Axiom 2, if $\sigma$ exceeds $\frac{\pi}{2}$ in magnitude the fused pitch and roll start decreasing again towards zero instead of continuing on to $\pi$.

---

13 Mathematically speaking, we are defining an equivalence relation $\sim$ on $\mathbb{D}$, and replacing the codomain $\mathbb{D}$ of $f_p$ with the corresponding quotient space $\check{\mathbb{D}}$, the elements of which are the equivalence classes of $\mathbb{D}$. The associated quotient topology on $\check{\mathbb{D}}$ then consists of all sets with an open preimage under the canonical projection map that maps each element to its equivalence class.

### 7.3.2 Links Between Rotation Representations

As was seen in Section 5.5, the various developed rotation representations have many links to each other, as well as to existing rotation formalisms. Some further mathematical links are presented in this section.

#### 7.3.2.1 *Links Between Fused Angles and the Tilt Phase Space*

Just like the fused roll $\phi$ and fused pitch $\theta$ can be thought of as quantifying the amount of tilt rotation about the x and y-axes respectively, the tilt phase space parameters $(p_x, p_y) \in \mathbb{P}^2$ also do exactly that. Formulated more carefully, from Equation (5.92), a tilt rotation of pure $p_x$ corresponds by definition to a $\gamma$ of 0 or $\pi$, and hence corresponds to a pure x-rotation, as the tilt axis $\hat{\mathbf{v}}$ (see Figure 5.5) is then parallel or antiparallel to the x-axis. Similarly, a tilt rotation of pure $p_y$ corresponds to a $\gamma$ of $\pm\frac{\pi}{2}$, and hence corresponds to a pure y-rotation. We know from Equation (5.80) that

$$\sin\phi = \sin\alpha \cos\gamma, \tag{7.154a}$$

$$\sin\theta = \sin\alpha \sin\gamma. \tag{7.154b}$$

For rotations that are not too far from the identity, we know that $\phi$, $\theta$ and $\alpha$ are small in magnitude. Thus, applying the small angle approximation

$$\sin\epsilon \approx \epsilon \quad \text{for } |\epsilon| \ll 1, \tag{7.155}$$

to the angles $\phi$, $\theta$ and $\alpha$ in Equation (7.154) yields

$$\phi \approx \alpha \cos\gamma, \tag{7.156a}$$

$$\theta \approx \alpha \sin\gamma. \tag{7.156b}$$

From the definition of the tilt phase space in Equation (5.62), it can immediately be concluded, for rotations not too far from the identity, that

$$\phi \approx p_x, \tag{7.157a}$$

$$\theta \approx p_y. \tag{7.157b}$$

In fact, $p_x$ and $p_y$ are from a mathematical perspective the $\mathcal{O}(\alpha^3)$ Taylor series approximations of $\phi$ and $\theta$ respectively:

$$\phi = p_x + \mathcal{O}(\alpha^3), \tag{7.158a}$$

$$\theta = p_y + \mathcal{O}(\alpha^3). \tag{7.158b}$$

Equivalently, the tilt phase space parameters $(p_x, p_y)$ can be considered to be the linearisation of the fused angles parameters $(\phi, \theta)$ about the identity rotation $(0, 0)$.

It can be interpreted from the stated mathematical links between the two representations that the 2D tilt phase parameters mimic fused angles for tilt rotations of small magnitudes, but increase linearly to infinity for large magnitudes. This is unlike $\phi$ and $\theta$, which loop around to always correctly represent the resulting orientation, even if more than $\pi$ radians was traversed to arrive there. For example, $(\gamma, \pi)$, $(\gamma, -\pi)$ and $(\gamma, 3\pi)$ all correspond to the same orientation in terms of fused angles, but are completely different tilt phase rotations if unbounded rotations are being considered. Put concisely, the tilt phase space is for unbounded tilt rotations, i.e. tilt rotations of more than 180°, what fused angles is for bounded tilt rotations—a way of concurrently quantifying in an axisymmetric manner the amount of rotation about each of the coordinate axes.

The relative differences between the fused angles and tilt phase space parameters are plotted in Figure 7.7 as functions of $p_x$ and $p_y$. The plotted differences are expressed as ratios of the tilt rotation magnitude $\alpha$, where for example a value of $0.1 = 10\%$ corresponds to a difference of $0.1\alpha$ between, for instance, $p_x$ and $\phi$. It can be observed from the figure that the errors in the approximations $p_x \approx \phi$ and $p_y \approx \theta$ are in fact notably less than the errors that would be mathematically involved in just assuming, for example, $\sin \alpha \approx \alpha$, showing how close the fused angles and tilt phase space representations are for even medium-sized rotations.

### 7.3.2.2    *Links Between Fused Angles and Rotation Matrices*

The correct and most efficient conversions between the fused angles and rotation matrix representations are given in Section 5.5. There are some other links between fused angles and rotation matrices however, that may be useful in other scenarios. For instance, the rotation matrix $R$ corresponding to the fused angles rotation $F = (\psi, \theta, \phi, h)$, nominally given by Equation (5.85), can be tweaked to the form

$$R = \begin{bmatrix} c_\gamma c_{\gamma+\psi} + c_\alpha s_\gamma s_{\gamma+\psi} & s_\gamma c_{\gamma+\psi} - c_\alpha c_\gamma s_{\gamma+\psi} & s_\psi s_\phi + c_\psi s_\theta \\ c_\gamma s_{\gamma+\psi} - c_\alpha s_\gamma c_{\gamma+\psi} & s_\gamma s_{\gamma+\psi} + c_\alpha c_\gamma c_{\gamma+\psi} & s_\psi s_\theta - c_\psi s_\phi \\ -s_\theta & s_\phi & h\sqrt{1 - s_\theta^2 - s_\phi^2} \end{bmatrix}, \tag{7.159}$$

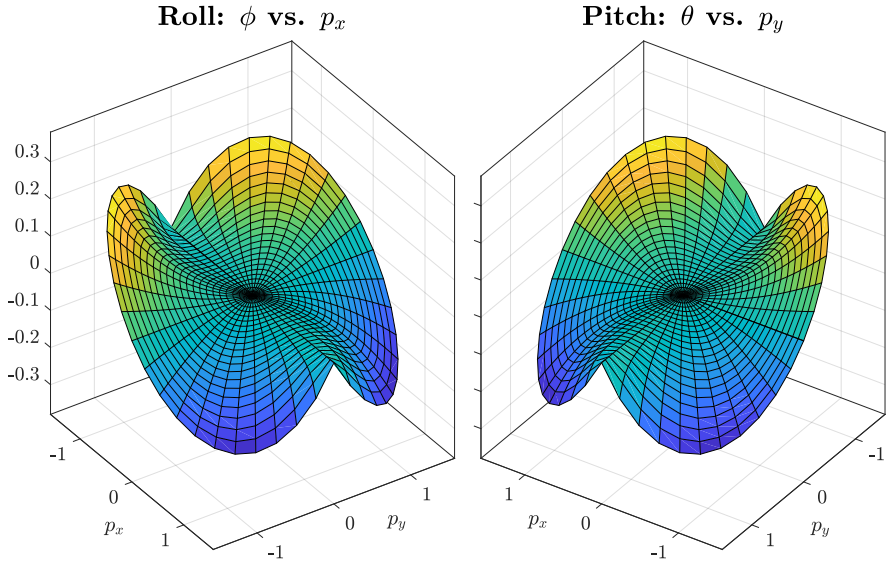

**Roll: $\phi$ vs. $p_x$**    **Pitch: $\theta$ vs. $p_y$**

Figure 7.7: Plots of the relative difference between the fused angles and tilt phase space parameters, expressed as a ratio of the tilt rotation magnitude $\alpha$—that is, plots of $\frac{1}{\alpha}(\phi - p_x)$ (left) and $\frac{1}{\alpha}(\theta - p_y)$ (right) against the 2D tilt phase space parameters $(p_x, p_y)$. For $\alpha = 1 \approx 57.3°$, the maximum relative difference is just 7.1%, compared to 15.9% for the equivalent small angle approximation $\sin \alpha \approx \alpha$. For $\alpha = \frac{\pi}{2} = 90°$, these numbers are 21.1% and 36.3%, respectively.

where $T = (\psi, \gamma, \alpha)$ is the corresponding tilt angles rotation. The conversion of $R$ back to the fused angles tilt rotation parameters $(\theta, \phi, h)$ is then given by

$$\theta = \operatorname{asin}(-R_{31}) = \operatorname{asin}(c_\psi R_{13} + s_\psi R_{23}), \tag{7.160a}$$

$$\phi = \operatorname{asin}(R_{32}) \quad = \operatorname{asin}(s_\psi R_{13} - c_\psi R_{23}), \tag{7.160b}$$

$$h = \operatorname{sign}(R_{33}). \tag{7.160c}$$

Equation (5.120) remains the only robust expression for the fused yaw $\psi$ based on $R$.

### 7.3.2.3  Links to Vectorial Parameterisations

Several of the possible representations for tilt rotations, in particular the ones developed in this thesis, can, as it turns out, be understood within the generalised context of vectorial parameterisations. Given the axis-angle representation $(\hat{\mathbf{e}}, \theta_a) \in \mathbb{A}$ of a rotation, we recall from Section 5.3.4 that the general form of a vectorial parameterisation is given by

$$\mathbf{p} = p(\theta_a)\,\hat{\mathbf{e}}, \tag{7.161}$$

for some odd scalar generating function $p(\theta_a)$ that satisfies

$$\lim_{\theta_a \to 0} \frac{p(\theta_a)}{\theta_a} = \kappa > 0. \tag{7.162}$$

A list of common vectorial parameterisations was given in Table 5.1. For tilt rotations, we know from the tilt angles representation $(\gamma, \alpha)$ that

$$\hat{\mathbf{e}} = (\cos \gamma, \sin \gamma, 0), \tag{7.163a}$$

$$\theta_a = \alpha. \tag{7.163b}$$

Thus, the general form for the vectorial parameterisation of a tilt rotation is given by

$$\mathbf{p} = \big(p(\alpha) \cos \gamma, \, p(\alpha) \sin \gamma, \, 0\big). \tag{7.164}$$

Suppose we now define the 'roll' and 'pitch' angles $(X, Y) \in \mathbb{R}^2$ to parameterise the tilt rotation, where

$$p(X) = p(\alpha) \cos \gamma, \tag{7.165a}$$

$$p(Y) = p(\alpha) \sin \gamma. \tag{7.165b}$$

The choice to define $X$ and $Y$ in this way makes logical and intuitive sense, as it can be thought of as taking the 'magnitude' of the tilt rotation in some vectorial space implicitly defined by $p(\cdot)$, resolving this magnitude vectorially into x and y-components, and then converting these components back to the 'angle space' via $p^{-1}(\cdot)$.

We can immediately see that if we take $p(\alpha) = \alpha$, i.e. the rotation vector parameterisation, that $X$ and $Y$ correspond to the tilt phase parameters. That is,

$$p(\alpha) = \alpha \implies \begin{cases} X \equiv p_x \\ Y \equiv p_y \end{cases} \tag{7.166}$$

We can also see by selecting the linear vectorial parameters (see Table 5.1) that we arrive at the fused angles representation, i.e.

$$p(\alpha) = \sin \alpha \implies \begin{cases} X \equiv \phi \\ Y \equiv \theta \end{cases} \tag{7.167}$$

If the generating function $p(\alpha) = \sin \frac{\alpha}{2}$ is chosen, corresponding to the reduced Euler-Rodrigues vectorial parameterisation, we notice that the parameterisation $\mathbf{p}$ is equivalent to the vector part of the quaternion representation of the tilt rotation. That is, if $q = (q_0, \mathbf{q}) = (w, x, y, 0)$ for $w \geq 0$ is the said quaternion representation, then from Equations (5.25) and (7.161),

$$\mathbf{p} = p(\alpha) \, \hat{\mathbf{e}} = \hat{\mathbf{e}} \sin \frac{\alpha}{2} = \mathbf{q}. \tag{7.168}$$

This demonstrates how even the quaternion representation of a tilt rotation can be related to an underlying vectorial parameterisation.

We continue for the case that $p(\alpha) = \sin\frac{\alpha}{2}$ by observing that the $x$ and $y$ quaternion parameters of a tilt rotation (for $w \geq 0$) are given by

$$x = p(\alpha)\cos\gamma = \sin\tfrac{\alpha}{2}\cos\gamma, \tag{7.169a}$$

$$y = p(\alpha)\sin\gamma = \sin\tfrac{\alpha}{2}\sin\gamma. \tag{7.169b}$$

Comparing this to Equation (7.165), we can see that the general vectorial framework within which we have placed the quaternion, fused angles and tilt angles parameterisations suggests that it would for instance be possible to define a completely new representation based on the reduced Euler-Rodrigues parameters, given by

$$\phi_q = p^{-1}(x) = 2\operatorname{asin}(x) = 2\operatorname{asin}\bigl(\sin\tfrac{\alpha}{2}\cos\gamma\bigr) \in (-\pi, \pi], \tag{7.170a}$$

$$\theta_q = p^{-1}(y) = 2\operatorname{asin}(y) = 2\operatorname{asin}\bigl(\sin\tfrac{\alpha}{2}\sin\gamma\bigr) \in (-\pi, \pi]. \tag{7.170b}$$

The complete new rotation representation would then be given by the quaternion angles $(\psi, \theta_q, \phi_q)$—a tuple consisting of the fused yaw $\psi$, quaternion pitch $\theta_q$, and quaternion roll $\phi_q$. Whether this definition of a rotation representation would be useful in real life applications however, would still need to be established.

### 7.3.3 Properties of Tilt Vector Addition

The notion of tilt vector addition, in the specific context of the tilt phase space, was introduced in Section 5.4.5.3. In this section, some further properties and applications of tilt vector addition are explored and presented.

#### 7.3.3.1 *Interpretation of Tilt Vector Addition*

The definition of tilt vector addition given in Section 5.4.5.3 can be intuitively related to the unambiguous commutative way in which angular velocities, unlike angular rotations, can be added. Essentially, the tilt phase space (and the intended application of balance) provides the robust mathematical framework within which the methods of commutative rotation combination that are usually reserved for infinitesimal rotations, can be meaningfully and correctly applied to all tilt rotations.

Suppose we wish to add together two tilt rotations, $(\gamma_1, \alpha_1)$ and $(\gamma_2, \alpha_2)$, to yield a third tilt rotation $(\gamma_3, \alpha_3)$. Using tilt vector addition, this addition is mathematically written as

$$(\gamma_1, \alpha_1) \oplus (\gamma_2, \alpha_2) = (\gamma_3, \alpha_3), \tag{7.171}$$

where from Equation (5.74) we can see that

$$\alpha_3 \cos\gamma_3 = \alpha_1 \cos\gamma_1 + \alpha_2 \cos\gamma_2, \tag{7.172a}$$

$$\alpha_3 \sin\gamma_3 = \alpha_1 \sin\gamma_1 + \alpha_2 \sin\gamma_2. \tag{7.172b}$$

The tilt rotation specified by the parameter pair $(\gamma_1, \alpha_1)$ is equivalent to applying an angular velocity of $^G\boldsymbol{\Omega}_1$ for $t$ seconds, where

$$^G\boldsymbol{\Omega}_1 = \tfrac{1}{t}(\alpha_1 \cos \gamma_1, \; \alpha_1 \sin \gamma_1, \; 0). \tag{7.173}$$

Similarly, the tilt rotation $(\gamma_2, \alpha_2)$ is equivalent to applying an angular velocity of $^G\boldsymbol{\Omega}_2$ for $t$ seconds, where

$$^G\boldsymbol{\Omega}_2 = \tfrac{1}{t}(\alpha_2 \cos \gamma_2, \; \alpha_2 \sin \gamma_2, \; 0). \tag{7.174}$$

Thus, if both angular velocities are applied simultaneously, the resulting total angular velocity is given by $^G\boldsymbol{\Omega}_t$, where

$$
\begin{aligned}
^G\boldsymbol{\Omega}_t &= \, ^G\boldsymbol{\Omega}_1 + \, ^G\boldsymbol{\Omega}_2 \\
&= \tfrac{1}{t}(\alpha_1 \cos \gamma_1 + \alpha_2 \cos \gamma_2, \; \alpha_1 \sin \gamma_1 + \alpha_2 \sin \gamma_2, \; 0) \\
&= \tfrac{1}{t}(\alpha_3 \cos \gamma_3, \; \alpha_3 \sin \gamma_3, \; 0).
\end{aligned}
\tag{7.175}
$$

Given that this resulting angular velocity is still applied for $t$ seconds, it is clear from the forms of Equations (7.173) and (7.174) that this summed angular velocity does indeed correspond to the tilt rotation $(\gamma_3, \alpha_3)$, as expected. It is important to note that tilt rotations, just like angular velocities, are unbounded, so this interpretation of tilt vector addition as the combination of angular velocities extends naturally to tilt rotations of all magnitudes, including greater than $\pi$ radians.

### 7.3.3.2   *Vector Space of Tilt Rotations*

Tilt vector addition, in terms of the tilt phase space, was defined in Equation (5.73) as

$$
\begin{aligned}
P_1 \oplus P_2 &= (p_{x1}, p_{y1}) \oplus (p_{x2}, p_{y2}) \\
&= (p_{x1} + p_{x2}, \; p_{y1} + p_{y2}) \in \mathbb{P}^2,
\end{aligned}
\tag{7.176}
$$

where $P_1, P_2 \in \mathbb{P}^2$ are two arbitrary 2D tilt phase space rotations. Recalling that

$$\mathbb{P}^2 \equiv \mathbb{R}^2, \tag{7.177}$$

it is easy to see that $\oplus$ is closed, associative and commutative over $\mathbb{P}^2$, and that $(0,0) \in \mathbb{P}^2$ is an identity element[14] that observes

$$(0,0) \oplus (p_x, p_y) = (p_x, p_y) \oplus (0,0) = (p_x, p_y). \tag{7.178}$$

Also, introducing the notation $\ominus(p_x, p_y)$ for the <span style="color:red">inverse of a tilt rotation</span> with respect to tilt vector addition $\oplus$, we can see from Equation (5.155) that $\ominus(p_x, p_y) \in \mathbb{P}^2$ always exists, and is equal to

$$\ominus(p_x, p_y) = (-p_x, -p_y). \tag{7.179}$$

---

14  Note that $(0,0) \in \mathbb{P}^2$ also corresponds exactly to the identity tilt rotation in 3D space.

Thus, it can be concluded that $(\mathbb{P}^2, \oplus)$ is an abelian group.

For $\lambda \in \mathbb{R}$, we define the scalar multiplication of a 2D tilt phase by $\lambda$ in the standard way, i.e.

$$\lambda(p_x, p_y) = (\lambda p_x, \lambda p_y). \tag{7.180}$$

Identifying that clearly (for $\lambda, \eta \in \mathbb{R}$ and $P, S \in \mathbb{P}^2$),

$$1P = P, \tag{7.181a}$$

$$\lambda(\eta P) = (\lambda \eta)P, \tag{7.181b}$$

$$(\lambda + \eta)P = \lambda P + \eta P, \tag{7.181c}$$

$$\lambda(P + S) = \lambda P + \lambda S, \tag{7.181d}$$

this completes $\mathbb{P}^2$ as a *vector space* over the field $\mathbb{R}$, where each 'vector' corresponds to a unique unbounded tilt rotation, expressed nominally in terms of the 2D tilt phase space. The vector space is referred to as the vector space of tilt rotations, and is isomorphic to $\mathbb{R}^2$, as suggestively written in Equation (7.177). The vector space of absolute tilt rotations is similarly defined. By definition, the additive inverses in these vector spaces always correspond to the true 3D inverses of the corresponding tilt rotations, and the additive identity vector $(0,0)$, i.e. the *zero vector*, corresponds to the true identity tilt rotation.

Given that tilt rotations can be formulated as a vector space as such, many useful properties, results, concepts and algorithms come for 'free'. For instance, combining tilt vector addition with scalar multiplication leads to a trivial definition of the mean of a set of tilt rotations. Other possibly useful corollaries of having a vector space include results involving differentiation, integration, metrics and general linear algebra.

### 7.3.3.3    *Tilt-based Orientation Interpolation*

Having a vector space of tilt rotations also allows cubic spline interpolation to be performed, with the guarantee that all produced intermediate rotations are indeed tilt rotations. Other methods of orientation spline interpolation—involving for example the logarithmic and exponential map and working with the Lie algebra $\mathfrak{so}(3)$—in general do not have this critical property, are significantly more computationally expensive, and cannot deal with rotations above 180°, albeit for the benefit of sometimes being bi-invariant (Kang and Park, 1999). In the inherently asymmetrical situation where there is a clear notion of 'up' however, bi-invariance is not seen as a decisive advantage, especially when observing that tilt phase space cubic spline interpolation is in fact invariant about the 'up' z-axis anyway, which for balance is the aspect of invariance that counts most. It should be noted that the optimal minimum angular acceleration interpolating curve between orientations is an involved three-dimensional, fourth-order nonlinear two-point boundary value problem, does not in general admit a closed form solution (Kang and Park, 1999), and is thus frequently not an option.

Table 7.2: Parameterisation map definitions for each rotation representation

| Representation | Rotation $D$ | Domain $\mathbb{D}$ | See eqn. |
|---|---|---|---|
| Rotation matrix | $R = [R_{ij}]$ | SO(3) | (7.11) |
| Quaternion | $q = (w, x, y, z)$ | $\mathbb{Q}$ | (7.52) |
| Axis-angle | $A = (\hat{\mathbf{e}}, \theta_a)$ | $\mathbb{A}$ | (5.11) |
| ZYX Euler angles | $E = (\psi_E, \theta_E, \phi_E)$ | $\mathbb{E}$ | (5.31) |
| ZXY Euler angles | $\tilde{E} = (\psi_{\tilde{E}}, \phi_{\tilde{E}}, \theta_{\tilde{E}})$ | $\tilde{\mathbb{E}}$ | (5.38) |
| Tilt angles | $T = (\psi, \gamma, \alpha)$ | $\mathbb{T}$ | (5.51) |
| Tilt phase space | $P = (p_x, p_y, p_z)$ | $\mathbb{P}^3$ | (5.63), (5.65a) |
| Fused angles | $F = (\psi, \theta, \phi, h)$ | $\mathbb{F}$ | (7.137), (7.143) |

The interpolation between tilt rotations expressed in the fused angles space is not as easy as in the tilt phase space, due to the binary hemisphere parameter $h$. The direct interpolation of fused angles in just the positive tilt hemisphere however, away from the hemisphere boundary, is an option that for many less stringent applications can produce completely satisfactory results. Nonetheless, it should be noted that such direct fused angles interpolation is not necessarily advised, and is not mathematically invariant, even just about the 'upright' z-axis.

### 7.3.4   Special Rotation Forms and Formats

In this section, mostly for general reference, we provide a list of the standard equivalences and standard rotation forms that exist for the various rotation representations that are used in this thesis.

#### 7.3.4.1   *Equivalences and Standard Forms*

In a similar spirit to Section 7.3.1, let us consider the map

$$f_d : \mathbb{D} \to SO(3), \; D \mapsto R \tag{7.182}$$

that defines some arbitrary parameterisation $D \in \mathbb{D}$ of the 3D rotation space SO(3). Example definitions of $D$ and $\mathbb{D}$ for the various parameterisations that are used in this thesis is given in Table 7.2. For each unique 3D rotation $R \in SO(3)$, we define the set of all parameter tuples that map to it as being *equivalent*, and can thereby define an equivalence relation $\sim$ on $\mathbb{D}$ such that

$$D_1 \sim D_2 \iff f_d(D_1) = f_d(D_2). \tag{7.183}$$

The collection of all non-trivial relations $D_1 \sim D_2$ (i.e. ones where $D_1 \neq D_2$) is referred to as the set of parameter equivalences of a particular rotation representation.[15] Given a complete definition of $\sim$, it is possible to construct the quotient space

$$\check{\mathbb{D}} = \mathbb{D}/\sim = \left\{ \check{\mathcal{D}} = \{D_1 \in \mathbb{D} : D_1 \sim D_2\} : D_2 \in \mathbb{D} \right\}, \qquad (7.184)$$

which has as its elements all parameter tuple equivalence classes with respect to $\sim$, and is equipped with the appropriate corresponding quotient topology. By definition, each equivalence class $\check{\mathcal{D}} \in \check{\mathbb{D}}$ can be mapped to a unique 3D rotation in SO(3). Thus, if we define an unambiguous nominal representative element $D_s \in \check{\mathcal{D}}$ for each class $\check{\mathcal{D}} \in \check{\mathbb{D}}$, we can see that every parameter tuple in $\mathbb{D}$ can be unambiguously mapped to its corresponding representative element $D_s$, referred to as its standard form, which in turn can be mapped uniquely to a rotation in SO(3).

ROTATION MATRIX    The task of defining $\sim$, and the nominal representative element of each resulting equivalence class, is up to the specific definition of each rotation representation individually. For example, rotation matrices are totally unique and have no non-trivial parameter equivalences, and hence no equivalence classes consisting of anything other than exactly one element. This is obvious because the map $f_d : \mathbb{D} \to SO(3)$ in the case of rotation matrices is really just the identity map $f_d : SO(3) \to SO(3)$, and so

$$R_1 \sim R_2 \iff R_1 = R_2. \qquad (7.185)$$

QUATERNION    Quaternions, on the other hand, have the single equivalence

$$q \sim -q, \qquad (7.186)$$

for all $q \in \mathbb{Q}$. Thus, every single equivalence class corresponding to a true rotation in SO(3) consists of exactly two sign-opposite quaternions, and we choose the one with $w \geq 0$ to be the standard form. More precisely, given any $q = (w, x, y, z) \in \mathbb{Q}$, the standard form of $q$ is given by $q$ or $-q$, depending on in which of the two quaternions the first non-zero component from left to right is positive. When working with quaternions, usually the strict concept of the standard form is not required in analysis, but equations or algorithms requiring $w \geq 0$ have for example already been encountered in this thesis, e.g. Equations (5.24) and (7.168).

AXIS-ANGLE    The parameter equivalences of the axis-angle representation have already been covered in Section 5.3.2, but we summarise

---

them again here in equivalence notation. If $(\hat{\mathbf{e}}, \theta_a) \in \mathcal{S}^2 \times [0, \pi] \equiv \mathbb{A}$ and we for numerical convenience allow $\hat{\mathbf{e}} = \mathbf{0}$ for the identity rotation, then

$$(\hat{\mathbf{e}}, 0) \sim (\mathbf{0}, 0), \tag{7.187a}$$

$$(\hat{\mathbf{e}}, \pi) \sim (-\hat{\mathbf{e}}, \pi). \tag{7.187b}$$

If we were to take a more relaxed initial domain $(\hat{\mathbf{e}}, \theta_a) \in \mathcal{S}^2 \times \mathbb{R}$, then we would get the additional equivalences

$$(\hat{\mathbf{e}}, \theta_a) \sim (-\hat{\mathbf{e}}, -\theta_a), \tag{7.188a}$$

$$(\hat{\mathbf{e}}, \theta_a) \sim (\hat{\mathbf{e}}, \theta_a + 2\pi k), \tag{7.188b}$$

where $k \in \mathbb{Z}$ is any integer. The definition of the standard form follows naturally from the above equations, first using Equation (7.188) to get to the domain $\mathbb{A}$, and then using Equation (7.187) to collapse the remaining equivalence classes at $\theta_a = 0$ and $\theta_a = \pi$ down to $\hat{\mathbf{e}} = 0$, and the $\pm\hat{\mathbf{e}}$ with its first non-zero component positive, respectively.

EULER ANGLES    The cases for both conventions of Euler angles that are considered in this thesis have also already been presented, in Sections 5.3.5.1 and 5.3.5.2 respectively, but are reviewed here more rigorously in terms of equivalences. Given any ZYX Euler angles parameter set $(\psi_E, \theta_E, \phi_E) \in (-\pi, \pi] \times [-\frac{\pi}{2}, \frac{\pi}{2}] \times (-\pi, \pi] \equiv \mathbb{E}$ and any $\epsilon \in \mathbb{R}$,

$$(\psi_E, \tfrac{\pi}{2}, \phi_E) \sim (\operatorname{wrap}(\psi_E - \epsilon), \tfrac{\pi}{2}, \operatorname{wrap}(\phi_E - \epsilon)), \tag{7.189a}$$

$$(\psi_E, -\tfrac{\pi}{2}, \phi_E) \sim (\operatorname{wrap}(\psi_E - \epsilon), -\tfrac{\pi}{2}, \operatorname{wrap}(\phi_E + \epsilon)). \tag{7.189b}$$

Treating ZYX Euler angles more loosely as being $(\psi_E, \theta_E, \phi_E) \in \mathbb{R}^3$, we get the further equivalences

$$(\psi_E, \theta_E, \phi_E) \sim (\pi + \psi_E, \pm\pi - \theta_E, \pi + \phi_E), \tag{7.190a}$$

$$(\psi_E, \theta_E, \phi_E) \sim (\psi_E + 2\pi k_1, \theta_E + 2\pi k_2, \phi_E + 2\pi k_3), \tag{7.190b}$$

where $k_1, k_2, k_3 \in \mathbb{Z}$. The situation is similar for the ZXY Euler angles convention, where if $(\psi_{\tilde{E}}, \phi_{\tilde{E}}, \theta_{\tilde{E}}) \in (-\pi, \pi] \times [-\frac{\pi}{2}, \frac{\pi}{2}] \times (-\pi, \pi] \equiv \tilde{\mathbb{E}}$,

$$(\psi_{\tilde{E}}, \tfrac{\pi}{2}, \theta_{\tilde{E}}) \sim (\operatorname{wrap}(\psi_{\tilde{E}} - \epsilon), \tfrac{\pi}{2}, \operatorname{wrap}(\theta_{\tilde{E}} + \epsilon)), \tag{7.191a}$$

$$(\psi_{\tilde{E}}, -\tfrac{\pi}{2}, \theta_{\tilde{E}}) \sim (\operatorname{wrap}(\psi_{\tilde{E}} - \epsilon), -\tfrac{\pi}{2}, \operatorname{wrap}(\theta_{\tilde{E}} - \epsilon)), \tag{7.191b}$$

and if $(\psi_{\tilde{E}}, \phi_{\tilde{E}}, \theta_{\tilde{E}}) \in \mathbb{R}^3$, then additionally

$$(\psi_{\tilde{E}}, \phi_{\tilde{E}}, \theta_{\tilde{E}}) \sim (\pi + \psi_{\tilde{E}}, \pm\pi - \phi_{\tilde{E}}, \pi + \theta_{\tilde{E}}), \tag{7.192a}$$

$$(\psi_{\tilde{E}}, \phi_{\tilde{E}}, \theta_{\tilde{E}}) \sim (\psi_{\tilde{E}} + 2\pi k_1, \phi_{\tilde{E}} + 2\pi k_2, \theta_{\tilde{E}} + 2\pi k_3). \tag{7.192b}$$

For both the ZYX and ZXY conventions, the standard form is achieved by bringing the parameters into their nominal ranges (i.e. $\mathbb{E}$, $\tilde{\mathbb{E}}$), and then if $\theta_E, \phi_{\tilde{E}} = \pm\frac{\pi}{2}$, choosing the equivalent parameter tuple with $\psi_E, \psi_{\tilde{E}} = 0$.

TILT ANGLES   Given any tuple of tilt angles $(\psi, \gamma, \alpha) \in (-\pi, \pi] \times (-\pi, \pi] \times [0, \pi] \equiv \mathbb{T}$, the following parameter equivalences hold:

$$(\psi, \gamma, 0) \sim (\psi, 0, 0), \tag{7.193a}$$

$$(\psi, \gamma, \pi) \sim \big(\text{wrap}(\psi - 2\epsilon), \text{ wrap}(\gamma + \epsilon), \pi\big), \tag{7.193b}$$

where $\epsilon \in \mathbb{R}$, and we note that a special case of Equation (7.193b) is

$$(\psi, \gamma, \pi) \sim (\psi, \text{wrap}(\gamma \pm \pi), \pi). \tag{7.194}$$

Relaxing the domain to $(\psi, \gamma, \alpha) \in \mathbb{R}^3$ yields the further equivalences

$$(\psi, \gamma, \alpha) \sim (\psi, \gamma \pm \pi, -\alpha), \tag{7.195a}$$

$$(\psi, \gamma, \alpha) \sim (\psi + 2\pi k_1, \ \gamma + 2\pi k_2, \ \alpha + 2\pi k_3). \tag{7.195b}$$

After Equation (7.195) has been used to ensure that a particular parameter set is in the nominal domain $\mathbb{T}$, the standard form for the tilt angles representation is given by choosing the equivalent parameter tuple with $\gamma = 0$ if $\alpha = 0$, and $\psi = 0$, $\gamma \in (-\frac{\pi}{2}, \frac{\pi}{2}]$ if $\alpha = \pi$.

TILT PHASE SPACE   As the tilt phase space is so closely related to the tilt angles representation, in particular in terms of its definition, the parameter equivalences of the tilt phase space can be derived from those just presented for the tilt angles representation. Given a parameter set $(p_x, p_y, p_z) \in \bar{\mathcal{D}}^2(\pi) \times (-\pi, \pi]$ and $\epsilon \in \mathbb{R}$, the sole resulting parameter equivalence is given by

$$\|(p_x, p_y)\| = \pi \implies$$
$$(p_x, p_y, p_z) \sim (c_\epsilon p_x - s_\epsilon p_y, \ s_\epsilon p_x + c_\epsilon p_y, \ \text{wrap}(p_z - 2\epsilon)), \tag{7.196}$$

where we are clearly considering the *bounded* interpretation of the tilt phase space here, as all that matters is the corresponding rotation matrix that is assigned by $f_d$. Note that the $p_x$ and $p_y$ components on the right-hand side in Equation (7.196) are just getting rotated CCW in 2D by $\epsilon$ radians, and that as a special case,

$$\|(p_x, p_y)\| = \pi \implies (p_x, p_y, p_z) \sim (-p_x, -p_y, p_z). \tag{7.197}$$

If we consider $(p_x, p_y, p_z) \in \mathbb{R}^3$, but still in a bounded sense, then the wrapping of various angles by $2\pi$ further contributes the equivalence

$$(\hat{p}_x, \hat{p}_y) = \lambda(p_x, p_y), \ \|(\hat{p}_x, \hat{p}_y) - (p_x, p_y)\| = 2\pi k_1$$
$$\implies (p_x, p_y, p_z) \sim (\hat{p}_x, \hat{p}_y, \text{wrap}(p_z + 2\pi k_2)), \tag{7.198}$$

for $\lambda \in \mathbb{R}$ and $k_1, k_2 \in \mathbb{Z}$. This expression at first looks complicated, but is essentially just saying that angle wrapping of $p_z$ is possible, and that every point $(p_x, p_y)$ is equivalent to the point $(\hat{p}_x, \hat{p}_y)$ obtained by shifting some multiple of $2\pi$ radians radially inwards or outwards.[16] The latter of these two equivalences can be identified as simply corresponding to angle wrapping of the tilt angle $\alpha$. The standard form of a tilt phase tuple $(p_x, p_y, p_z) \in \mathbb{R}^3$ can be derived by first applying Equation (7.198) to bring it to the domain $\bar{\mathcal{D}}^2(\pi) \times (-\pi, \pi]$, and then applying Equation (7.196) to ensure on the boundary of $\bar{\mathcal{D}}^2(\pi)$ that $p_z = 0$, and that the first non-zero element of $(p_x, p_y)$, if present, is positive. This standard form is naturally consistent with the definition of the standard form for the tilt angles representation.

FUSED ANGLES    Given any tuple of fused angles $(\psi, \theta, \phi, h) \in \mathbb{F}$, where $\mathbb{F}$ is defined as in Equation (7.143), i.e. in adherence to the sine sum criterion, the sole parameter equivalence is given by

$$|\theta| + |\phi| = \tfrac{\pi}{2} \implies (\psi, \theta, \phi, 1) \sim (\psi, \theta, \phi, -1). \tag{7.199}$$

Note that the situation at the fused yaw singularity is not incorporated as an equivalence here, as it depends on the exact technical definition of the fused angles representation that is chosen there. Nominally at the singularity, every possible fused angles parameter set $F_{sing} = (\psi, 0, 0, -1)$ is taken to simultaneously represent all possible fused yaw singular rotations in a many-to-many manner, as even if an explicit mapping were to be chosen, it would still just be an arbitrary choice and not alleviate the problems of an essential discontinuity. The motivation behind the many-to-many mapping is for convenience in numerical implementations, so as not to overspecify the expected result when multiple would be valid, and floating point errors make exact numerical comparisons error-prone. If we relax the sine sum criterion but still keep $|\theta|, |\phi| \leq \tfrac{\pi}{2}$, i.e. have any tuple $(\psi, \theta, \phi, h) \in (-\pi, \pi] \times [-\tfrac{\pi}{2}, \tfrac{\pi}{2}] \times [-\tfrac{\pi}{2}, \tfrac{\pi}{2}] \times \{-1, 1\} \equiv \hat{\mathbb{F}}$, then we get the additional equivalence

$$S = \sqrt{s_\theta^2 + s_\phi^2} \geq 1 \implies$$
$$(\psi, \theta, \phi, h) \sim \left(\psi, \operatorname{asin}\left(\tfrac{1}{S} s_\theta\right), \operatorname{asin}\left(\tfrac{1}{S} s_\phi\right), h\right), \tag{7.200}$$

which is essentially just radially shrinking $(\sin\theta, \sin\phi)$, if necessary, to enforce the sine sum criterion. Further relaxing the fused angles domain to $(\psi, \theta, \phi, h) \in \mathbb{R}^3 \times \{-1, 1\}$ yields

$$(\psi, \theta, \phi, h) \sim (\psi, \pm\pi - \theta, \phi, -h), \tag{7.201a}$$

$$(\psi, \theta, \phi, h) \sim (\psi, \theta, \pm\pi - \phi, -h), \tag{7.201b}$$

$$(\psi, \theta, \phi, h) \sim (\psi + 2\pi k_1, \theta + 2\pi k_2, \phi + 2\pi k_3, h), \tag{7.201c}$$

for $k_1, k_2, k_3 \in \mathbb{Z}$. The standard form of a fused angles parameter tuple $(\psi, \theta, \phi, h)$ is achieved by collapsing it into the nominal domain $\mathbb{F}$ using Equations (7.200) and (7.201), and then setting $h = 1$ if the fused pitch and roll are exactly on the domain boundary, i.e. $|\theta| + |\phi| = \frac{\pi}{2}$.

---

16 This applies to the tilt phase origin $(p_x, p_y) = (0, 0)$ also, which is equivalent to every point on every circle of radius $2\pi k$ in the 2D tilt phase space, for $k \in \mathbb{Z}$.

### 7.3.4.2  *Format of Pure XYZ Rotations*

In the course of our discussion of 3D spatial rotations, we have frequently encountered the need to notationally refer to pure x, y and/or z-rotations, i.e. pure axis rotations, in various representations. The general convention that has been used is that if

$$\square = R, q, T, F, P, \ldots$$

then for $\beta \in \mathbb{R}$ we denote

$\square_x(\beta) \Rightarrow$ CCW rotation by $\beta$ about the x-axis in representation $\square$,

$\square_y(\beta) \Rightarrow$ CCW rotation by $\beta$ about the y-axis in representation $\square$,

$\square_z(\beta) \Rightarrow$ CCW rotation by $\beta$ about the z-axis in representation $\square$.

Just for general reference, in this section we explicitly give the numerical definitions of the various pure axis rotations for each rotation representation. For instance, for rotation matrices we have that

$$R_x(\beta) = \begin{bmatrix} 1 & 0 & 0 \\ 0 & c_\beta & -s_\beta \\ 0 & s_\beta & c_\beta \end{bmatrix}, \tag{7.202a}$$

$$R_y(\beta) = \begin{bmatrix} c_\beta & 0 & s_\beta \\ 0 & 1 & 0 \\ -s_\beta & 0 & c_\beta \end{bmatrix}, \tag{7.202b}$$

$$R_z(\beta) = \begin{bmatrix} c_\beta & -s_\beta & 0 \\ s_\beta & c_\beta & 0 \\ 0 & 0 & 1 \end{bmatrix}, \tag{7.202c}$$

where we recall that $c_\beta \equiv \cos \beta$ and $s_\beta \equiv \sin \beta$. For quaternions,

$$q_x(\beta) = \left( \cos \tfrac{\beta}{2}, \sin \tfrac{\beta}{2}, 0, 0 \right), \tag{7.203a}$$

$$q_y(\beta) = \left( \cos \tfrac{\beta}{2}, 0, \sin \tfrac{\beta}{2}, 0 \right), \tag{7.203b}$$

$$q_z(\beta) = \left( \cos \tfrac{\beta}{2}, 0, 0, \sin \tfrac{\beta}{2} \right). \tag{7.203c}$$

Pure axis rotations in the tilt angles representation are given by

$$T_x(\beta) = (0, 0, \beta), \tag{7.204a}$$

$$T_y(\beta) = (0, \tfrac{\pi}{2}, \beta), \tag{7.204b}$$

$$T_z(\beta) = (\beta, 0, 0), \tag{7.204c}$$

while, keeping in mind the parameter equivalences for fused angles listed in Section 7.3.4.1, the corresponding fused angles representations are given by

$$F_x(\beta) = (0, 0, \beta, 1), \tag{7.205a}$$

$$F_y(\beta) = (0, \beta, 0, 1), \tag{7.205b}$$

$$F_z(\beta) = (\beta, 0, 0, 1). \tag{7.205c}$$

Finally, in the tilt phase space the pure axis rotations are given by

$$P_x(\beta) = (\beta, 0, 0), \tag{7.206a}$$

$$P_y(\beta) = (0, \beta, 0), \tag{7.206b}$$

$$P_z(\beta) = (0, 0, \beta). \tag{7.206c}$$

### 7.3.5   Spherical Linear Interpolation

Given two frames {A} and {B} relative to a global frame {G}, it is frequently of interest to be able to interpolate smoothly and directly between these two frames, and thereby generate intermediate frames at arbitrary points along the interpolation. Various methods of performing such an interpolation are available, but each method has distinct advantages and disadvantages, in particular based on whether the method is

(a) **Torque minimal**, meaning that it follows a 'shortest' path through the space of rotations to get from {A} to {B},

(b) **Constant speed**, meaning that the path is traced out with constant magnitude of angular velocity in 3D space, and/or,

(c) **Commutative**, meaning that interpolating part of the way from frame {A} to frame {B}, and then from there part of the way to another frame {C}, is equivalent to interpolating towards {C} first, and then towards {B}.

The most obvious method for interpolating between two frames is to construct the most direct axis-angle rotation possible from one to the other, and then smoothly increasing the angle of rotation while keeping the axis of rotation constant. By Euler's rotation theorem, this process is unique and well-defined, and only runs into slight ambiguity if the angle of rotation is 180° (the maximum possible), as then there is a choice of which direction to rotate about the rotation axis. The described method is known as spherical linear interpolation (slerp), and can quickly be seen to be torque minimal and constant speed, but not commutative. Alternatives to slerp include normalised linear interpolation (nlerp), which involves linear interpolation and renormalisation in the quaternion space, and log space linear interpolation (log lerp), which involves using the log map (see page 203) to convert to the rotation vector space and linearly interpolating there. The former (nlerp) is torque minimal and commutative but not constant speed, while the latter (log lerp) is commutative but neither torque minimal nor constant speed, even though the angular speed is nevertheless usually quite close to being constant. It should be noted that no method of rotation interpolation exists that satisfies all three desired properties, so each is a trade-off between the various possible features.

### 7.3.5.1 *Quaternion Formulation*

In terms of quaternions, spherical linear interpolation corresponds to the interpolation along great arcs of the four-dimensional unit sphere $\mathbb{Q} \cong \mathcal{S}^3$. Great arcs are circular arcs on the surface of a sphere that share the same centre as the sphere, and constitute the so-called geodesics of a sphere, as they correspond to the set of all 'shortest paths' from one point to another. Given two quaternions $q_0, q_1 \in \mathbb{Q}$, and an interpolation parameter $u \in [0, 1]$, where $u = 0$ corresponds to the quaternion $q_0$ and $u = 1$ corresponds to the quaternion $q_1$, the slerp interpolation method is given by

$$\text{slerp}(q_0, q_1, u) = (q_1 q_0^{-1})^u q_0 = q_0 (q_0^{-1} q_1)^u. \tag{7.207}$$

Note that we can extend the domain of $u$ to all of $\mathbb{R}$, resulting in a method for extrapolation as well as interpolation. The quaternion $q_1 q_0^{-1}$ in Equation (7.207) is the relative rotation from $q_0$ to $q_1$, and taking it to the power of $u$ is like linearly interpolating the angle parameter of the corresponding axis-angle representation, as can be seen from Equation (7.132). An exact definition of what it means to take a quaternion to a continuous real power is provided in Section 7.1.2.5. In terms of basis notation, if we are interpolating from a frame {A} to a frame {B}, both of which are expressed relative to a global frame {G}, then

$$q_0 \equiv {}^G_A q, \tag{7.208a}$$

$$q_1 \equiv {}^G_B q, \tag{7.208b}$$

and the intermediate frame {C} corresponding to the interpolating parameter $u \in \mathbb{R}$ is given by

$$ {}^G_C q = \left( {}^G_B q \, {}^A_G q \right)^u {}^G_A q = {}^G_A q \left( {}^A_G q \, {}^G_B q \right)^u, \tag{7.209}$$

which simplifies to

$$ {}^G_C q = {}^{GA}_{\;\;B} q^u \, {}^G_A q = {}^G_A q \, {}^A_B q^u, \tag{7.210}$$

where ${}^{GA}_{\;\;B} q$ is a *referenced rotation*. Both of the expressions for slerp given in Equation (7.210) can easily be interpreted and verified geometrically to be doing the right thing, based on the theory of global and local axis rotations (see page 192).

---

**Algorithm 7.1:** Quaternion spherical linear interpolation (slerp)

---

**Input:** Quaternions $q_0, q_1 \in \mathbb{Q}$, interpolation parameter $u \in \mathbb{R}$
**Output:** Interpolated quaternion $q_u \in \mathbb{Q}$

1: **function** SLERP($q_0$, $q_1$, $u$)
2:     $d \leftarrow q_0 \cdot q_1$
3:     **if** $d < 0$ **then**
4:         $q_1 \leftarrow -q_1$
5:         $d \leftarrow -d$
6:     **end if**
7:     **if** $d \geq 1 - \epsilon$ **then**        ▷ For `double`, $\epsilon = 5 \times 10^{-9}$ is appropriate
8:         $\tilde{q}_u \leftarrow (1 - u)q_0 + uq_1$
9:     **else**
10:        $\tilde{q}_u \leftarrow \sin\big((1-u)\Omega\big)q_0 + \sin(u\Omega)q_1$
11:    **end if**
12:    $q_u \leftarrow \dfrac{\tilde{q}_u}{\|\tilde{q}_u\|}$
13:    **return** $q_u$
14: **end function**

---

Analytically, an alternative but equivalent formulation of slerp is given by

$$\text{slerp}(q_0, q_1, u) = \left(\frac{\sin\big((1-u)\Omega\big)}{\sin\Omega}\right)q_0 + \left(\frac{\sin u\Omega}{\sin\Omega}\right)q_1, \qquad (7.211)$$

where $\Omega = \text{acos}(q_0 \cdot q_1)$, and we require that $\Omega \in [0, \frac{\pi}{2}]$, i.e. $q_0 \cdot q_1 \geq 0$. If $q_0 \cdot q_1 < 0$, we simply negate $q_1$ prior to applying Equation (7.211), as then with the negated $q_1$ (which still represents the same rotation) we have that $q_0 \cdot q_1 \geq 0$ as required. If $q_1$ were not negated in this situation, the resulting 3D interpolation would go the 'long way around' from $q_0$ to $q_1$, as opposed to the shortest and most direct way. The property

$$q_0 \cdot q_1 \geq 0 \qquad (7.212)$$

is usually referred to as $q_0$ and $q_1$ being 'mutually in the same 4D hemisphere' as each other, as it is equivalent to the condition that each quaternion is contained in the 4D hemisphere centred at the other. Nonetheless, a more direct interpretation of Equation (7.212) is that it specifies that the angle in 4D space between the quaternion vectors $q_0$ and $q_1$ is at most 90°.

For numerical implementations, Equation (7.211) has sensitivity issues when $q_0$ and $q_1$ are very close to each other, as in that case $\sin\Omega$ is very close to zero. As a result, in practice it is common to use nlerp when $\Omega$ is close to zero, as nlerp is numerically stable and indistinguishable from slerp for such cases. The resulting numerical algorithm is shown in Algorithm 7.1. It should be noted that this algorithm, just like the analytic form in Equation (7.211), does not fail (like for example pure nlerp does) if the coordinate frames that $q_0$ and $q_1$ represent are 180° apart. If this is the case, the signs of $q_0$ and $q_1$ decide which way around the interpolation goes.

One important property of spherical linear interpolation is that it is bi-invariant. This is to say that it is both left-invariant and right-invariant, which is respectively defined to mean that

$$\text{slerp}(\hat{q}q_0, \hat{q}q_1, u) = \hat{q}\,\text{slerp}(q_0, q_1, u), \tag{7.213a}$$

$$\text{slerp}(q_0\hat{q}, q_1\hat{q}, u) = \text{slerp}(q_0, q_1, u)\hat{q}, \tag{7.213b}$$

for all $\hat{q} \in \mathbb{Q}$. Bi-invariance is a useful and intuitive property of spherical linear interpolation, as it effectively asserts that interpolating between any two coordinate frames in 3D space gives the same results no matter what reference frame it is numerically evaluated with respect to. A proof of Equation (7.213) is provided in Appendix B.1.2.1.

### 7.3.5.2   *Rotation Matrix Formulation*

Although it is not generally useful for numerical implementations, analytically, spherical linear interpolation can also be expressed in terms of rotation matrices. The interpolation from a rotation matrix $R_0 \in \text{SO}(3)$ to the rotation matrix $R_1 \in \text{SO}(3)$, indexed by the parameter $u \in \mathbb{R}$, is given by

$$\text{slerp}(R_0, R_1, u) = (R_1 R_0^T)^u R_0 = R_0 (R_0^T R_1)^u. \tag{7.214}$$

where $u = 0$ corresponds to $R_0$, and $u = 1$ corresponds to $R_1$. An exact definition of what it means to take a rotation matrix to the continuous real power $u \in \mathbb{R}$ is provided in Section 7.1.2.5. Once again, Equation (7.214) can be reexpressed in terms of coordinate frames and basis notation. If $R_0$ defines frame {A} and $R_1$ defines frame {B} relative to some global reference frame {G}, i.e.

$$R_0 \equiv {}^G_A R, \tag{7.215a}$$

$$R_1 \equiv {}^G_B R, \tag{7.215b}$$

then the intermediate frame {C} that corresponds to the interpolation by parameter $u \in \mathbb{R}$ is given by

$$^G_C R = \left({}^G_B R\, {}^A_G R\right)^u {}^G_A R = {}^G_A R \left({}^A_C R\, {}^G_B R\right)^u, \tag{7.216}$$

which simplifies to

$$^G_C R = {}^{GA}_B R^u\, {}^G_A R = {}^G_A R\, {}^A_B R^u. \tag{7.217}$$

where ${}^{GA}_B R$ is a *referenced rotation*.

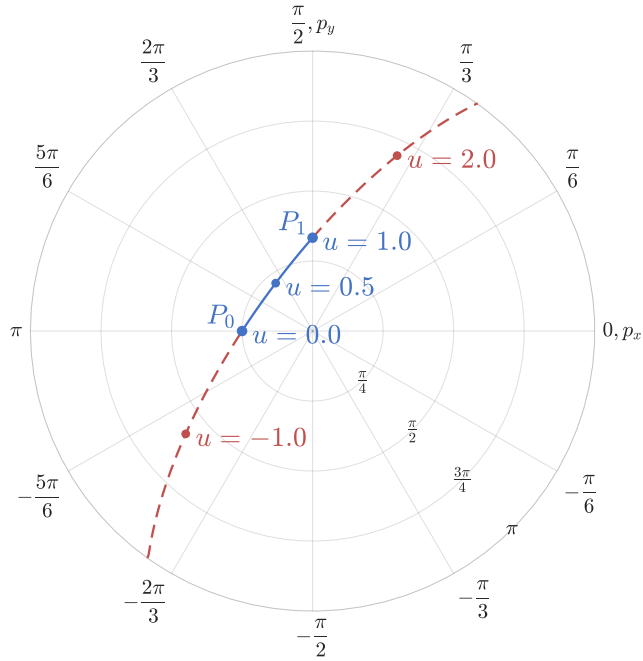

Figure 7.8: Spherical linear interpolation (in blue) between pure tilt rotations in the tilt phase space, namely from $P_0$ a roll rotation of $-\frac{\pi}{4} = -45°$, to $P_1$ a pitch rotation of $\frac{\pi}{3} = 60°$. The red dashed line shows how extrapolation can be achieved by using values of $u$ outside of $[0, 1]$. Note that the dashed line wraps around to the corresponding antipodal point on the boundary of the tilt phase space. The fused yaws of the interpolated rotations do not need to be plotted as by Equation (7.218) they are zero everywhere, as they are zero for $P_0$ and $P_1$.

### 7.3.5.3 *Slerp Between Tilt Rotations*

If spherical linear interpolation is applied between any two tilt rotations, it can be demonstrated that all possible resulting interpolated frames will always be tilt rotations as well. That is, for $q_0, q_1 \in \mathbb{Q}$ and $u \in \mathbb{R}$,

$$\Psi(q_0) = \Psi(q_1) = 0 \implies \Psi\big(\mathrm{slerp}(q_0, q_1, u)\big) = 0, \qquad (7.218)$$

where $\Psi(\cdot)$ is the operator that returns the fused yaw of a rotation in any representation. We can interpret this result as saying that the direct axis-angle rotation between any two tilt rotations produces only tilt rotations as its intermediate (and extrapolated) rotations. This is useful because it means that one can always easily and cleanly interpolate between tilt rotations, without affecting the fused yaw, as illustrated in Figure 7.8. The result is much less trivial than it may seem, as for example by comparison, the interpolation between rotations of zero Euler yaw by no means produces further rotations of zero Euler yaw, or even necessarily close to it. A proof of Equation (7.218) is provided in Appendix B.1.2.2.

Suppose now that we have two arbitrary rotations $q_0, q_1 \in \mathbb{Q}$, of equal, but not necessarily zero, fused yaw $\psi$. From Equation (5.131b), we know that we can write

$$q_0 = q_z(\psi) q_{t0}, \tag{7.219a}$$

$$q_1 = q_z(\psi) q_{t1}, \tag{7.219b}$$

for some tilt rotations $q_{t0}, q_{t1} \in \mathbb{Q}$. Thus, from left-invariance,

$$\begin{aligned} \operatorname{slerp}(q_0, q_1, u) &= \operatorname{slerp}(q_z(\psi) q_{t0}, \, q_z(\psi) q_{t1}, \, u) \\ &= q_z(\psi) \operatorname{slerp}(q_{t0}, q_{t1}, u). \end{aligned} \tag{7.220}$$

From Equation (7.218) however, we can see that $\operatorname{slerp}(q_{t0}, q_{t1}, u)$ must be a pure tilt rotation, so clearly $\operatorname{slerp}(q_0, q_1, u)$ must also have a fused yaw of $\psi$, due to the $q_z(\psi)$ component in Equation (7.220). We can summarise this result as

$$\Psi(q_0) = \Psi(q_1) = \psi \implies \Psi\big(\operatorname{slerp}(q_0, q_1, u)\big) = \psi. \tag{7.221}$$

In practice, what this means is that the spherical linear interpolation between any two rotations of the same fused yaw will always produce intermediate rotations with exactly the same fused yaw. This is clearly a direct generalisation of the previous result about tilt rotations.

### 7.3.5.4 *Metrics in Terms of Tilt Angles*

As spherical linear interpolation generates trajectories that correspond to the geodesics of a sphere, i.e. the 'shortest paths' between points on the quaternion sphere $\mathbb{Q} \cong \mathcal{S}^3$, it is a closely related problem to define what 'shortest' even means. From a mathematical perspective, the notions of 'distance' and 'separation' in a space are formalised by the concept of metrics. A metric on the rotation space $\operatorname{SO}(3)$ is simply a function

$$d : \operatorname{SO}(3) \times \operatorname{SO}(3) \to [0, \infty) \tag{7.222}$$

such that for all $x, y, z \in \operatorname{SO}(3)$,

1. $d(x,y) \geq 0$      (7.223a)
2. $d(x,y) = 0 \iff x = y$      (7.223b)
3. $d(x,y) = d(y,x)$      (7.223c)
4. $d(x,z) \leq d(x,y) + d(y,z)$      (7.223d)

The distance between any two rotations $R_1, R_2 \in \operatorname{SO}(3)$ with respect to the metric $d$ is then given by the non-negative real scalar $d(R_1, R_2)$. For convenience and brevity of notation, we allow other rotation representations to be passed to metric functions as well, and all it means is that both arguments are converted to rotation matrices first, and then passed to the metric. For quaternions, for example, this necessitates that

$$d(\pm q_1, \pm q_2) = d(q_1, q_2), \tag{7.224}$$

as otherwise $d$ would not be well-defined.

Suppose we have two rotations $R_1, R_2 \in \mathrm{SO}(3)$, and their equivalent quaternion representations $q_1, q_2 \in \mathbb{Q}$. If $(\hat{\mathbf{e}}, \theta_a) \in \mathbb{A}$ is the direct axis-angle rotation from $R_1$ to $R_2$, i.e. the geodesic great arc on the surface of $\mathbb{Q}$ that is traced out by spherical linear interpolation from $q_1$ to $q_2$, then the most natural metric is given by the Riemannian metric

$$d_R(R_1, R_2) = \|\log(R_1^T R_2)\|_F \tag{7.225a}$$
$$= 2 \arccos(|q_1 \cdot q_2|) \tag{7.225b}$$
$$= \theta_a. \tag{7.225c}$$

Essentially, this metric gives the 3D angle magnitude between two rotations, i.e. when treated as coordinate frames, and is equivalent to the magnitude of rotation performed by slerp to get from one of the rotations to the other. An alternative metric that is very computationally efficient if using quaternions, but not boundedly equivalent[17] to $d_R$ (Huynh, 2009), is given by

$$d_L(R_1, R_2) = 1 - |q_1 \cdot q_2| \tag{7.226a}$$
$$= 1 - \cos \tfrac{\theta_a}{2}. \tag{7.226b}$$

Both $d_L$ and $d_R$ are bi-invariant, i.e. if $d \equiv d_L, d_R$ and $\hat{q} \in \mathbb{Q}$ then

$$d(\hat{q} q_1, \hat{q} q_2) = d(q_1 \hat{q}, q_2 \hat{q}) = d(q_1, q_2), \tag{7.227}$$

and both can be computed directly from the equivalent tilt angles representations $T_1 = (\psi_1, \gamma_1, \alpha_1)$ and $T_2 = (\psi_2, \gamma_2, \alpha_2)$ of $q_1$ and $q_2$ respectively, by noting that

$$q_1 \cdot q_2 = \cos \tfrac{\alpha_1}{2} \cos \tfrac{\alpha_2}{2} \cos \Delta\bar{\psi} + \sin \tfrac{\alpha_1}{2} \sin \tfrac{\alpha_2}{2} \cos(\Delta\bar{\psi} + \Delta\gamma), \tag{7.228}$$

where $\Delta\bar{\psi} = \tfrac{1}{2}(\psi_1 - \psi_2)$ and $\Delta\gamma = \gamma_1 - \gamma_2$. Equations (7.225b) and (7.226a) can then be used as before to compute $d_R$ and $d_L$.

### 7.3.6   Rotational Velocity Conversions

When working with trajectories in either the 2D tilt rotation space or the full 3D rotation space—like for example for the purposes of cubic spline interpolation (see Section 7.3.3.3)—it is necessary to be able to relate rotational velocities between their various possible forms. In Section 7.1.2.4 it has already been shown how angular velocity vectors $\boldsymbol{\Omega} \in \mathbb{R}^3$ can be converted to and from their equivalent quaternion and rotation matrix velocities $\dot{q}$ and $\dot{R}$, respectively. Noting that in general dotted variables are notationally used to refer to the time derivatives of the corresponding base variable, e.g.

---

[17] Two metrics $d_A$ and $d_B$ are said to be boundedly equivalent if two positive real numbers $c_1, c_2 > 0$ exist such that $c_1 d_A(R_1, R_2) \leq d_B(R_1, R_2) \leq c_2 d_A(R_1, R_2)$ for all $R_1, R_2 \in \mathrm{SO}(3)$.

$$\dot{\psi} \equiv \frac{d\psi}{dt}, \tag{7.229}$$

in this section we now consider the <span style="color:red">rotational velocity</span> conversions between the following representations:

- Angular velocity vector ${}^G\boldsymbol{\Omega} \in \mathbb{R}^3$ (where {G} is the global frame)

- Tilt angles velocity $\dot{T} = (\dot{\psi}, \dot{\gamma}, \dot{\alpha})$

- Absolute tilt angles velocity $\dot{\tilde{T}} = (\dot{\psi}, \dot{\tilde{\gamma}}, \dot{\alpha})$

- Relative tilt phase velocity $\dot{P} = (\dot{p}_x, \dot{p}_y, \dot{p}_z)$

- Absolute tilt phase velocity $\dot{\tilde{P}} = (\dot{\tilde{p}}_x, \dot{\tilde{p}}_y, \dot{\tilde{p}}_z)$

- Reduced fused angles velocity $\dot{F} = (\dot{\psi}, \dot{\theta}, \dot{\phi})$

Rotational velocity conversions need to performed at a particular rotation, as the tangent space to the manifold of rotations is different at each point. In this section, we nominally perform all conversions at the arbitrary rotation $T = (\psi, \gamma, \alpha)$.[18] To begin, we recall that

$$\psi = p_z = \tilde{p}_z, \tag{7.230a}$$
$$\tilde{\gamma} = \gamma + \psi. \tag{7.230b}$$

It follows directly that

$$\dot{\psi} = \dot{p}_z = \dot{\tilde{p}}_z, \tag{7.231a}$$
$$\dot{\tilde{\gamma}} = \dot{\gamma} + \dot{\psi}. \tag{7.231b}$$

Thus, in particular due to Equation (7.231a), for all rotational velocity conversions between $\dot{T}$, $\dot{\tilde{T}}$, $\dot{P}$, $\dot{\tilde{P}}$ and $\dot{F}$, we only need to focus on what happens to the non-yaw components. Furthermore, Equation (7.231b) completely describes the conversion between relative and absolute tilt angles velocities $\dot{T}$ and $\dot{\tilde{T}}$, so this also does not need to be explicitly covered below.

---

18 Or equivalently, at the corresponding representations $P = (p_x, p_y, p_z)$, $\tilde{P} = (\tilde{p}_x, \tilde{p}_y, \tilde{p}_z)$ and $F = (\psi, \theta, \phi, h)$. The tilt angles parameters $T = (\psi, \gamma, \alpha)$ are just the ones that are required the most often in the equations in this section.

### 7.3.6.1  *Between Tilt Phase Velocities*

Keeping Equation (7.231a) in mind, the full conversion from a relative tilt phase velocity $\dot{P}$ to its corresponding absolute tilt phase velocity $\check{P}$ is given by

$$\check{p}_x = c_\psi \dot{p}_x - s_\psi \dot{p}_y - \tilde{p}_y \dot{p}_z, \tag{7.232a}$$

$$\check{p}_y = s_\psi \dot{p}_x + c_\psi \dot{p}_y + \tilde{p}_x \dot{p}_z. \tag{7.232b}$$

The reverse conversion from absolute back to relative is given by

$$\dot{p}_x = \phantom{-}c_\psi \check{p}_x + s_\psi \check{p}_y + p_y \check{p}_z, \tag{7.233a}$$

$$\dot{p}_y = -s_\psi \check{p}_x + c_\psi \check{p}_y - p_x \check{p}_z. \tag{7.233b}$$

We recall at this point that for example, $p_x = \alpha \cos\gamma$ and $\tilde{p}_y = \alpha \sin\tilde{\gamma}$.

### 7.3.6.2  *From Tilt Angles Velocities To*

Given an arbitrary tilt angles velocity $\dot{T} = (\dot{\psi}, \dot{\gamma}, \dot{\alpha})$, the corresponding relative and absolute tilt phase space velocities are given by

$$\dot{p}_x = c_\gamma \dot{\alpha} - s_\gamma \alpha \dot{\gamma}, \qquad \check{p}_x = c_{\tilde{\gamma}} \dot{\alpha} - s_{\tilde{\gamma}} \alpha \dot{\tilde{\gamma}}, \tag{7.234a}$$

$$\dot{p}_y = s_\gamma \dot{\alpha} + c_\gamma \alpha \dot{\gamma}, \qquad \check{p}_y = s_{\tilde{\gamma}} \dot{\alpha} + c_{\tilde{\gamma}} \alpha \dot{\tilde{\gamma}}. \tag{7.234b}$$

The corresponding fused angles velocities are given by

$$\dot{\theta} = \frac{1}{c_\theta}(s_\gamma c_\alpha \dot{\alpha} + s_\phi \dot{\gamma}), \tag{7.235a}$$

$$\dot{\phi} = \frac{1}{c_\phi}(c_\gamma c_\alpha \dot{\alpha} - s_\theta \dot{\gamma}), \tag{7.235b}$$

and the corresponding angular velocity vector ${}^G\boldsymbol{\Omega}$ can be calculated using

$${}^G\boldsymbol{\Omega} = \left(c_{\tilde{\gamma}} \dot{\alpha} - s_{\tilde{\gamma}} s_\alpha \dot{\gamma}, \; s_{\tilde{\gamma}} \dot{\alpha} + c_{\tilde{\gamma}} s_\alpha \dot{\gamma}, \; \dot{\psi} + (1 - c_\alpha)\dot{\gamma}\right), \tag{7.236}$$

where we recall that $\tilde{\gamma} = \gamma + \psi$.

### 7.3.6.3  *From Tilt Phase Velocities To*

Given an arbitrary relative tilt phase velocity $\dot{P} = (\dot{p}_x, \dot{p}_y, \dot{p}_z)$, or absolute tilt phase velocity $\check{P} = (\check{p}_x, \check{p}_y, \check{p}_z)$, we can reverse Equation (7.234) and calculate the corresponding tilt angles velocity parameters using

$$\dot{\gamma} = \frac{1}{\alpha}\left(c_\gamma \dot{p}_y - s_\gamma \dot{p}_x\right), \tag{7.237a}$$

$$\dot{\tilde{\gamma}} = \frac{1}{\alpha}\left(c_{\tilde{\gamma}} \check{p}_y - s_{\tilde{\gamma}} \check{p}_x\right), \tag{7.237b}$$

$$\dot{\alpha} = c_\gamma \dot{p}_x + s_\gamma \dot{p}_y, \tag{7.237c}$$

$$\phantom{\dot{\alpha}} = c_{\tilde{\gamma}} \check{p}_x + s_{\tilde{\gamma}} \check{p}_y. \tag{7.237d}$$

We know that the $\gamma$ and $\tilde{\gamma}$ tilt axis angle parameters have essential singularities at $\alpha = 0$, so as expected, their respective velocities as given by Equations (7.237a) and (7.237b) are infinite in this case.

When converting between tilt phase velocities and angular velocities, it is natural to try to convert via tilt angles velocities. This causes unnecessary problems with the aforementioned tilt axis angle singularity however, as neither the tilt phase representation nor the angular velocity actually has a singularity at $\alpha = 0$, but the tilt angles representation does (in the $\gamma$ and $\tilde{\gamma}$ parameters). By mathematically combining the required conversion equations however, and taking care of the resulting removable singularity at $\alpha = 0$, robust conversions between tilt phase velocities and angular velocities can be achieved. Specifically, combining Equations (7.236) and (7.237) yields a conversion from the tilt phase velocities $\dot{P}$ and $\breve{P}$ to the angular velocity ${}^G\mathbf{\Omega}$, given by

$$
{}^G\mathbf{\Omega} = \bigl(c_{\tilde{\gamma}}\dot{\alpha} - s_{\tilde{\gamma}}SG,\ s_{\tilde{\gamma}}\dot{\alpha} + c_{\tilde{\gamma}}SG,\ \dot{p}_z + CG\bigr), \tag{7.238}
$$

where $\dot{\alpha}$ is calculated using Equations (7.237c) and (7.237d), and

$$
S = \operatorname{sinc}\alpha, \tag{7.239a}
$$

$$
G = c_\gamma \dot{p}_y - s_\gamma \dot{p}_x \tag{7.239b}
$$

$$
= c_{\tilde{\gamma}}\breve{p}_y - s_{\tilde{\gamma}}\breve{p}_x - \alpha \breve{p}_z, \tag{7.239c}
$$

$$
C = \begin{cases} \frac{1 - c_\alpha}{\alpha} & \text{if } \alpha \neq 0, \\ 0 & \text{if } \alpha = 0. \end{cases} \tag{7.239d}
$$

$S$ and $C$ are smooth functions of $\alpha$, as the removable singularity at the origin in each case has been remedied by manual definition (see definition of $\operatorname{sinc}(\cdot)$, the *cardinal sine function*). This leads to a globally robust and smooth expression for ${}^G\mathbf{\Omega}$, including in particular also for $\alpha = 0$. Although it at first may not seem like it, at $\alpha = 0$ the right-hand side of Equation (7.238) reduces to

$$
\alpha = 0 \implies {}^G\mathbf{\Omega} = (c_\psi \dot{p}_x - s_\psi \dot{p}_y,\ s_\psi \dot{p}_x + c_\psi \dot{p}_y,\ \dot{p}_z) \tag{7.240a}
$$

$$
= (\breve{p}_x,\ \breve{p}_y,\ \breve{p}_z), \tag{7.240b}
$$

and is thus independent of $\gamma$. This is important, and makes sense, as $\gamma$ is singular at $\alpha = 0$ and can therefore take on any value there, but ${}^G\mathbf{\Omega}$ is unique and well-defined. To state Equation (7.240b) slightly more succinctly,

$$
\alpha = 0 \implies {}^G\mathbf{\Omega} = \breve{\dot{P}}. \tag{7.241}
$$

This result exemplifies the strong link between angular velocities and the tilt phase space, as was also discussed in the context of tilt vector addition in Section 7.3.3.1.

If we wish to convert from a tilt phase velocity $\dot{P}$ or $\breve{P}$ to a fused angles velocity, then we first convert absolute to relative (if required) using Equation (7.233), and then apply

$$
\dot{\theta} = \tfrac{1}{c_\theta}\bigl(\dot{p}_y \operatorname{sinc}\alpha + s_\gamma \dot{\alpha}(c_\alpha - \operatorname{sinc}\alpha)\bigr), \tag{7.242a}
$$

$$
\dot{\phi} = \tfrac{1}{c_\phi}\bigl(\dot{p}_x \operatorname{sinc}\alpha + c_\gamma \dot{\alpha}(c_\alpha - \operatorname{sinc}\alpha)\bigr), \tag{7.242b}
$$

where $\dot{\alpha}$ is once again evaluated using Equations (7.237c) and (7.237d).

### 7.3.6.4  *From Fused Angles Velocities To*

Given an arbitrary reduced fused angles velocity $\dot{F} = (\dot{\psi}, \dot{\theta}, \dot{\phi})$, the corresponding tilt angles velocity $\dot{T}$ is simply given by

$$\dot{\gamma} = \tfrac{1}{s_\alpha}(c_\gamma c_\theta \dot{\theta} - s_\gamma c_\phi \dot{\phi}), \tag{7.243a}$$

$$\dot{\alpha} = \tfrac{1}{c_\alpha}(s_\gamma c_\theta \dot{\theta} + c_\gamma c_\phi \dot{\phi}). \tag{7.243b}$$

To convert to a relative tilt phase velocity $\dot{P}$ instead, we use

$$\dot{p}_x = \tfrac{1}{\operatorname{sinc}\alpha}\big(c_\phi \dot{\phi} - c_\gamma \dot{\alpha}(c_\alpha - \operatorname{sinc}\alpha)\big), \tag{7.244a}$$

$$\dot{p}_y = \tfrac{1}{\operatorname{sinc}\alpha}\big(c_\theta \dot{\theta} - s_\gamma \dot{\alpha}(c_\alpha - \operatorname{sinc}\alpha)\big), \tag{7.244b}$$

where $\dot{\alpha}$ is evaluated using Equation (7.243b). The corresponding absolute tilt phase velocity $\dot{\tilde{P}}$ is then given by applying Equation (7.232).

Based on Equation (7.236), the conversion from a fused angles velocity $\dot{F}$ to an angular velocity vector ${}^G\Omega$ is given by

$$ {}^G\Omega = \big(c_{\hat\gamma}\dot{\alpha} - s_{\hat\gamma}\hat{G},\ s_{\hat\gamma}\dot{\alpha} + c_{\hat\gamma}\hat{G},\ \dot{\psi} + \hat{C}\hat{G}\big), \tag{7.245}$$

where $\dot{\alpha}$ is once again evaluated using Equation (7.243b), and

$$\hat{G} = c_\gamma c_\theta \dot{\theta} - s_\gamma c_\phi \dot{\phi}, \tag{7.246a}$$

$$\hat{C} = \begin{cases} \frac{1 - c_\alpha}{s_\alpha} & \text{if } \alpha \neq 0, \\ 0 & \text{if } \alpha = 0. \end{cases} \tag{7.246b}$$

Similar to the conversion from tilt phase velocities to angular velocities given in Equation (7.238), Equation (7.245) avoids the tilt axis angle singularity at $\alpha = 0$ by a mix of cancellation and manual definition.

### 7.3.6.5  *From Angular Velocity Vector To*

It can be observed from Equations (7.236), (7.238) and (7.245) that the angular velocity vector ${}^G\Omega$ is in general a linear combination of the individual parameter velocities. For instance, Equation (7.236) can be written as

$$ {}^G\Omega = (0, 0, 1)\dot{\psi} + (-s_{\hat\gamma}s_\alpha,\ c_{\hat\gamma}s_\alpha,\ 1 - c_\alpha)\dot{\gamma} + (c_{\hat\gamma}, s_{\hat\gamma}, 0)\dot{\alpha}, \tag{7.247}$$

and so by treating ${}^G\Omega$ and $\dot{T} = (\dot{\psi}, \dot{\gamma}, \dot{\alpha})$ as column vectors, can be written as the matrix equation

$$ {}^G\Omega = \begin{bmatrix} 0 & -s_{\hat\gamma}s_\alpha & c_{\hat\gamma} \\ 0 & c_{\hat\gamma}s_\alpha & s_{\hat\gamma} \\ 1 & 1 - c_\alpha & 0 \end{bmatrix} \begin{bmatrix} \dot{\psi} \\ \dot{\gamma} \\ \dot{\alpha} \end{bmatrix} \equiv V\dot{T}. \tag{7.248}$$

Away from any parameter singularities that affect the invertibility of $V$, we thus have

$$\dot{T} = V^{-1}\,{}^{G}\boldsymbol{\Omega}. \tag{7.249}$$

If we label the row vectors of $V^{-1}$ as follows,

$$V^{-1} = \begin{bmatrix} \leftarrow \mathbf{v}_{\psi} \rightarrow \\ \leftarrow \mathbf{v}_{\gamma} \rightarrow \\ \leftarrow \mathbf{v}_{\alpha} \rightarrow \end{bmatrix} \tag{7.250}$$

we can therefore conclude that

$$\dot{\psi} = {}^{G}\boldsymbol{\Omega} \cdot \mathbf{v}_{\psi}, \tag{7.251a}$$

$$\dot{\gamma} = {}^{G}\boldsymbol{\Omega} \cdot \mathbf{v}_{\gamma}, \tag{7.251b}$$

$$\dot{\alpha} = {}^{G}\boldsymbol{\Omega} \cdot \mathbf{v}_{\alpha}, \tag{7.251c}$$

where

$$\mathbf{v}_{\psi} = \tfrac{1}{1+c_{\alpha}}\big(s_{\alpha}s_{\tilde{\gamma}},\ -s_{\alpha}c_{\tilde{\gamma}},\ 1+c_{\alpha}\big), \tag{7.252a}$$

$$\mathbf{v}_{\gamma} = \tfrac{1}{s_{\alpha}}\big(-s_{\tilde{\gamma}},\ c_{\tilde{\gamma}},\ 0\big), \tag{7.252b}$$

$$\mathbf{v}_{\alpha} = \big(c_{\tilde{\gamma}},\ s_{\tilde{\gamma}},\ 0\big). \tag{7.252c}$$

It follows directly from Equation (7.231b) that $\dot{\tilde{\gamma}} = {}^{G}\boldsymbol{\Omega} \cdot \mathbf{v}_{\tilde{\gamma}}$, where

$$\mathbf{v}_{\tilde{\gamma}} = \mathbf{v}_{\gamma} + \mathbf{v}_{\psi} \tag{7.253a}$$

$$= \tfrac{1}{s_{\alpha}}\big(-c_{\alpha}s_{\tilde{\gamma}},\ c_{\alpha}c_{\tilde{\gamma}},\ s_{\alpha}\big). \tag{7.253b}$$

It can be seen that the calculation of $\dot{\psi}$ via $\mathbf{v}_{\psi}$ has problems only at the fused yaw singularity $\alpha = \pi$, and that $\mathbf{v}_{\tilde{\gamma}}$ and $\mathbf{v}_{\gamma}$ fail only at the tilt axis angle singularities $\alpha = 0, \pi$. The latter of these two problems is mitigated by the tilt phase space, for which we can derive that

$$\dot{p}_{x} = {}^{G}\boldsymbol{\Omega} \cdot \mathbf{v}_{p_{x}}, \qquad\qquad \dot{\tilde{p}}_{x} = {}^{G}\boldsymbol{\Omega} \cdot \mathbf{v}_{\tilde{p}_{x}}, \tag{7.254a}$$

$$\dot{p}_{y} = {}^{G}\boldsymbol{\Omega} \cdot \mathbf{v}_{p_{y}}, \qquad\qquad \dot{\tilde{p}}_{y} = {}^{G}\boldsymbol{\Omega} \cdot \mathbf{v}_{\tilde{p}_{y}}, \tag{7.254b}$$

where, for $S = \operatorname{sinc}\alpha$,

$$\mathbf{v}_{p_{x}} = \big(c_{\gamma}c_{\tilde{\gamma}} + \tfrac{1}{S}s_{\gamma}s_{\tilde{\gamma}},\ c_{\gamma}s_{\tilde{\gamma}} - \tfrac{1}{S}s_{\gamma}c_{\tilde{\gamma}},\ 0\big), \tag{7.255a}$$

$$\mathbf{v}_{p_{y}} = \big(s_{\gamma}c_{\tilde{\gamma}} - \tfrac{1}{S}c_{\gamma}s_{\tilde{\gamma}},\ s_{\gamma}s_{\tilde{\gamma}} + \tfrac{1}{S}c_{\gamma}c_{\tilde{\gamma}},\ 0\big), \tag{7.255b}$$

$$\mathbf{v}_{\tilde{p}_{x}} = \big(c_{\tilde{\gamma}}^{2} + \tfrac{c_{\alpha}}{S}s_{\tilde{\gamma}}^{2},\ c_{\tilde{\gamma}}s_{\tilde{\gamma}}\big(1 - \tfrac{c_{\alpha}}{S}\big),\ -\alpha s_{\tilde{\gamma}}\big), \tag{7.255c}$$

$$\mathbf{v}_{\tilde{p}_{y}} = \big(c_{\tilde{\gamma}}s_{\tilde{\gamma}}\big(1 - \tfrac{c_{\alpha}}{S}\big),\ s_{\tilde{\gamma}}^{2} + \tfrac{c_{\alpha}}{S}c_{\tilde{\gamma}}^{2},\ \alpha c_{\tilde{\gamma}}\big). \tag{7.255d}$$

We note that $S \neq 0$ away from the fused yaw singularity, so $\tfrac{1}{S}$ is also smooth on this domain, and, critically, there is no cusp or irregularity at $\alpha = 0$. For the fused pitch and roll velocities, even the fused yaw singularity is mitigated, as

$$\dot{\theta} = {}^{G}\mathbf{\Omega} \cdot \mathbf{v}_{\theta}, \tag{7.256a}$$

$$\dot{\phi} = {}^{G}\mathbf{\Omega} \cdot \mathbf{v}_{\phi}, \tag{7.256b}$$

where

$$\mathbf{v}_{\theta} = \tfrac{1}{c_{\phi}}\left(c_{\alpha}s_{\gamma}c_{\tilde{\gamma}} - c_{\gamma}s_{\tilde{\gamma}},\ c_{\alpha}s_{\gamma}s_{\tilde{\gamma}} + c_{\gamma}c_{\tilde{\gamma}},\ 0\right), \tag{7.257a}$$

$$\mathbf{v}_{\phi} = \tfrac{1}{c_{\phi}}\left(c_{\alpha}c_{\gamma}c_{\tilde{\gamma}} + s_{\gamma}s_{\tilde{\gamma}},\ c_{\alpha}c_{\gamma}s_{\tilde{\gamma}} - s_{\gamma}c_{\tilde{\gamma}},\ 0\right), \tag{7.257b}$$

but problems arise at the hemisphere boundary if either the fused pitch $\theta$ or fused roll $\phi$ is exactly $\frac{\pi}{2}$.

Some of the various derived angular velocity resolution vectors $\mathbf{v}_{*}$ (for $* = \psi, \gamma, \dots$) also have alternate forms in terms of the rotation matrix and quaternion representations, namely $R = [R_{ij}] \in \mathrm{SO}(3)$ and $q = (w, x, y, z) \in \mathbb{Q}$. The alternate forms are given by

$$\mathbf{v}_{\psi} = \tfrac{1}{1 + R_{33}}\left(R_{13},\ R_{23},\ 1 + R_{33}\right) \tag{7.258a}$$

$$= \tfrac{1}{w^2 + z^2}\left(wy + xz,\ yz - wx,\ w^2 + z^2\right), \tag{7.258b}$$

$$\mathbf{v}_{\gamma} = \tfrac{1}{R_{13}^2 + R_{23}^2}\left(-R_{13},\ -R_{23},\ 0\right) \tag{7.258c}$$

$$= \tfrac{1}{2K}\left(-wy - xz,\ wx - yz,\ 0\right), \tag{7.258d}$$

$$\mathbf{v}_{\alpha} = \tfrac{1}{\sqrt{R_{13}^2 + R_{23}^2}}\left(-R_{23},\ R_{13},\ 0\right) \tag{7.258e}$$

$$= \tfrac{1}{\sqrt{K}}\left(wx - yz,\ wy + xz,\ 0\right), \tag{7.258f}$$

$$\mathbf{v}_{\tilde{\gamma}} = \tfrac{1}{R_{13}^2 + R_{23}^2}\left(-R_{33}R_{13},\ -R_{33}R_{23},\ R_{13}^2 + R_{23}^2\right), \tag{7.258g}$$

$$\mathbf{v}_{\theta} = \tfrac{1}{\sqrt{R_{11}^2 + R_{21}^2}}\left(-R_{21},\ R_{11},\ 0\right) \tag{7.258h}$$

$$= \tfrac{1}{\sqrt{L}}\left(-2(wz + xy),\ 1 - 2(y^2 + z^2),\ 0\right), \tag{7.258i}$$

$$\mathbf{v}_{\phi} = \tfrac{1}{\sqrt{R_{12}^2 + R_{22}^2}}\left(R_{22},\ -R_{12},\ 0\right) \tag{7.258j}$$

$$= \tfrac{1}{\sqrt{M}}\left(1 - 2(x^2 + z^2),\ 2(wz - xy),\ 0\right), \tag{7.258k}$$

where

$$K = (w^2 + z^2)(x^2 + y^2), \tag{7.259a}$$

$$L = 1 - 4(xz - wy)^2, \tag{7.259b}$$

$$M = 1 - 4(wx + yz)^2. \tag{7.259c}$$

It should be noted that while the vectors $\mathbf{v}_{\psi}$ and $\mathbf{v}_{\tilde{\gamma}}$ have a $z$-component of exactly 1, the vectors $\mathbf{v}_{\alpha}$, $\mathbf{v}_{\theta}$ and $\mathbf{v}_{\phi}$ are unit vectors, and thus truly resolve a component of the ${}^{G}\mathbf{\Omega}$ vector in 3D space.

Examples of the vectors $\mathbf{v}_{*}$ from the equations in this section are illustrated in Figure 7.9. It can be seen in Figure 7.9a for instance that,

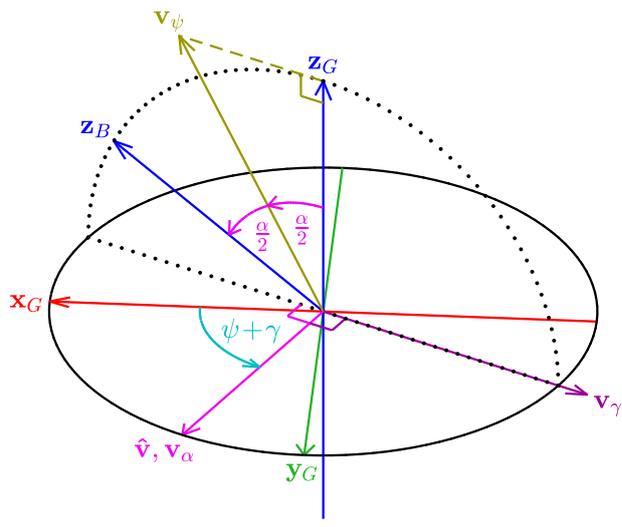

(a) Tilt angles resolution vectors $\mathbf{v}_\psi$, $\mathbf{v}_\gamma$ and $\mathbf{v}_\alpha$

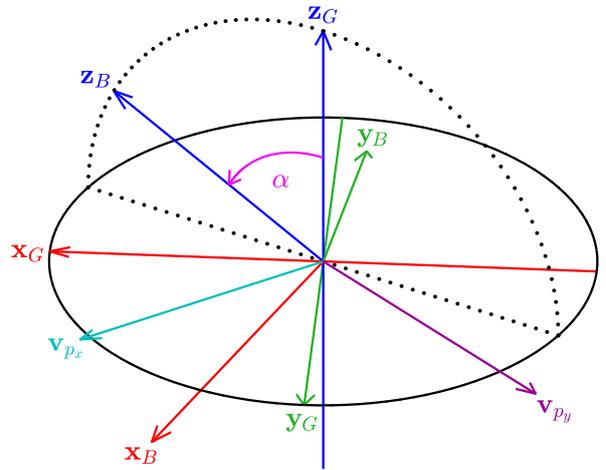

(b) Relative tilt phase space resolution vectors $\mathbf{v}_{p_x}$ and $\mathbf{v}_{p_y}$

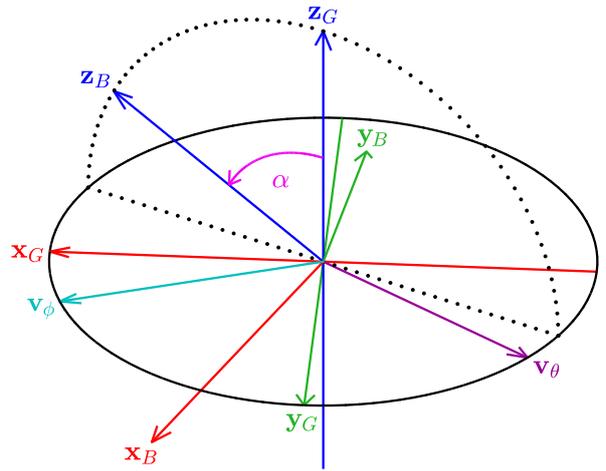

(c) Fused angles resolution vectors $\mathbf{v}_\theta$ and $\mathbf{v}_\phi$

Figure 7.9: Examples of the angular velocity resolution vectors $\mathbf{v}_*$ for the rotation ${}^G_B T = (0.7,\ 0.4,\ 1.1)$. Note that $\mathbf{x}_B$ points left-downwards out of the page, and $\mathbf{y}_B$ points up-rightwards out of the page. All of the resolution vectors are in the $\mathbf{x}_G\mathbf{y}_G$ plane, except for $\mathbf{v}_\psi$.

geometrically, $\mathbf{v}_\psi$ is along the angle bisector of $\mathbf{z}_G$ and $\mathbf{z}_B$—the two vectors that define the plane of the tilt rotation component. The vector $\mathbf{v}_\alpha$ then corresponds geometrically to the unit normal vector of this plane, as expected from the definition of $\alpha$, and is in fact orthogonal to all three of the remaining tilt angles vectors $\mathbf{v}_\psi$, $\mathbf{v}_\gamma$ and $\mathbf{v}_{\hat\gamma}$.

Although it is expected that the fused yaw parameter $p_z \equiv \psi$ of the tilt phase space representation has a problem with its angular velocity resolution vector $\mathbf{v}_\psi$ at the fused yaw singularity, it is (at least at first) somewhat less expected that $\mathbf{v}_{p_x}$ and $\mathbf{v}_{p_y}$ do also. The problem mathematically at the fused yaw singularity is that the map from tilt phase velocities to angular velocities, i.e. $V_p \in \mathbb{R}^{3\times3}$ where

$$^G\mathbf{\Omega} = V_p \dot P, \tag{7.260}$$

loses rank, and thereby loses its invertibility. As a result, the set of all possible valid $^G\mathbf{\Omega}$ becomes only two-dimensional, and every $^G\mathbf{\Omega}$ in that set is mapped to by an infinite number of possible $\dot P$ vectors. In fact, every valid $^G\mathbf{\Omega}$ is mapped to exactly by some one-dimensional linear subset of $\mathbb{R}^3$. To visualise this, consider the fused yaw singular rotation $q_x(\pi) = (0, 1, 0, 0) \in \mathbb{Q}$, and how tilt phase velocities applied to this rotation relate to angular velocity vectors. Any velocity in the $p_x$ parameter clearly contributes cleanly to the x-component of $^G\mathbf{\Omega}$, i.e. $\omega_x$, while any velocities in either the $p_y$ or $p_z$ parameters *both* contribute solely to the z-component of $^G\mathbf{\Omega}$, i.e. $\omega_z$. As a result, we can see that no tilt phase parameter in this situation contributes at all to $\omega_y$. If a pure $\omega_y$ angular velocity were to be applied to the rotation $q_x(\pi)$, then this would require $p_z$ to jump from $0$ to $\pi$, and $(p_x, p_y)$ to jump from $(\pi, 0)$ to $(0, \pm\pi)$, which clearly does not correspond to any possible finite tilt phase velocity.

### 7.3.6.6 *Angular Velocities of Pure Tilt Rotations*

If we consider any trajectory of rotations in TR(3), i.e. any trajectory of rotations where every intermediate rotation is a pure tilt rotation,[19] then we know that we must have $\psi = 0$ and $\dot\psi = 0$ at every point. All tilt rotations have a quaternion z-parameter of zero, so we know from Equations (7.251a) and (7.258b) that

$$0 = \dot\psi = {}^G\mathbf{\Omega} \cdot \mathbf{v}_\psi \tag{7.261}$$

$$= {}^G\mathbf{\Omega} \cdot \tfrac{1}{w^2+z^2}\big(wy + xz,\; yz - wx,\; w^2 + z^2\big) \tag{7.262}$$

$$= {}^G\mathbf{\Omega} \cdot \tfrac{1}{w^2}\big(wy,\; -wx,\; w^2\big). \tag{7.263}$$

Multiplying through by $w$, we conclude that at every point along the tilt rotation trajectory we must have

$$^G\mathbf{\Omega} \cdot (y, -x, w) = 0, \tag{7.264}$$

---

19 Every spherical linear interpolation between a pair of tilt rotations is an example of such a trajectory of tilt rotations.

where we note that $(y, -x, w)$ is a unit vector. This equation can be considered to characterise in terms of angular velocities the two-dimensional tangent space to the differentiable manifold of tilt rotations. In terms of quaternion velocities, the tangent space to TR(3) at the tilt quaternion $q = (w, x, y, 0)$ is given by

$$T_q^{\text{TR(3)}} = \left\{(w, x, y, 0), (0, 0, 0, 1)\right\}^\perp, \tag{7.265}$$

where $W^\perp$ denotes the orthogonal complement space of a set of vectors $W$, i.e. the linear space consisting of all vectors that are perpendicular to every vector in $W$. It should be noted however, that the tangent space in Equation (7.265) can also be expressed more explicitly as

$$T_q^{\text{TR(3)}} = \text{span}\left\{(x, -w, 0, 0), (y, 0, -w, 0), (0, y, -x, 0)\right\}, \tag{7.266}$$

where we specify three linearly dependent spanning vectors, as any one of them may be zero depending on the exact value of $q$.

### 7.3.7 Constructing Rotations from Components

We recall from Section 5.4, and in particular Section 5.4.2.1, that it is possible to partition any 3D rotation into yaw and tilt rotation components, and that the tilt rotation component can then be parameterised using either local or global z-vectors, i.e. ${}^G\mathbf{z}_B$ or ${}^B\mathbf{z}_G$, respectively. In this section, we consider the reverse process, and detail ways of constructing complete 3D rotations from any given combination of fused yaw and z-vector tilts. Note that in each case, the reconstructed rotation is from the global frame {G} to the local frame {B}.

#### 7.3.7.1 *Composition of Yaw and Global Z-vector*

Suppose we are given the desired fused yaw $\psi$ and global z-vector ${}^B\mathbf{z}_G$ of a rotation, and wish to calculate the full representation of the corresponding 3D rotation. We divide up the cases based on which final rotation representation is required. Note that the tilt phase space has been omitted here, as it trivially follows from the corresponding calculations for the tilt angles representation.

TILT ANGLES    From Equation (5.52), we have that

$$\gamma = \text{atan2}\left(-{}^B z_{Gx}, {}^B z_{Gy}\right), \tag{7.267a}$$

$$\alpha = \text{acos}\left({}^B z_{Gz}\right). \tag{7.267b}$$

Recalling that $\psi$ is given, the required full tilt angles representation is then simply ${}^G_B T = (\psi, \gamma, \alpha)$.

FUSED ANGLES   From Equations (5.55) and (5.56), we have that

$$\theta = \text{asin}(-^Bz_{Gx}), \tag{7.268a}$$

$$\phi = \text{asin}(^Bz_{Gy}), \tag{7.268b}$$

$$h = \text{sign}(^Bz_{Gz}). \tag{7.268c}$$

Recalling once again that $\psi$ is given, the required full fused angles representation is then simply $^G_BF = (\psi, \theta, \phi, h)$.

QUATERNION   Given $\psi$ and $^B\mathbf{z}_G$, the w and z-components of the corresponding full quaternion rotation are given by

$$N_{wz} = \tfrac{1}{2}(1 + {}^Bz_{Gz}), \tag{7.269a}$$

$$w = \sqrt{N_{wz}} \cos \tfrac{\psi}{2}, \tag{7.269b}$$

$$z = \sqrt{N_{wz}} \sin \tfrac{\psi}{2}. \tag{7.269c}$$

Unscaled x and y-components of rotation can then be calculated using

$$\tilde{x} = {}^Bz_{Gx}z + {}^Bz_{Gy}w, \tag{7.270a}$$

$$\tilde{y} = {}^Bz_{Gy}z - {}^Bz_{Gx}w. \tag{7.270b}$$

Being careful of the fused yaw singularity, the final quaternion is then

$$^G_Bq = \begin{cases} (0, \cos \tfrac{\psi}{2}, \sin \tfrac{\psi}{2}, 0) & \text{if } \tilde{x} = \tilde{y} = 0, \\ (w, A\tilde{x}, A\tilde{y}, z) & \text{otherwise,} \end{cases} \tag{7.271}$$

where

$$A = \sqrt{\frac{1 - N_{wz}}{\tilde{x}^2 + \tilde{y}^2}}. \tag{7.272}$$

In the special case that we are dealing with a tilt rotation ($\psi = 0$), the entire process can be simplified down to

$$w = \sqrt{\tfrac{1}{2}(1 + {}^Bz_{Gz})}, \tag{7.273a}$$

$$^G_Bq = \begin{cases} (0, \cos \tfrac{\psi}{2}, \sin \tfrac{\psi}{2}, 0) & \text{if } w = 0, \\ (w, \tfrac{1}{2w}{}^Bz_{Gy}, -\tfrac{1}{2w}{}^Bz_{Gx}, 0) & \text{otherwise.} \end{cases} \tag{7.273b}$$

ROTATION MATRIX   The most numerically safe way of constructing the rotation matrix $^G_BR$ from a specification of $\psi$ and $^B\mathbf{z}_G$, is to construct the corresponding quaternion first (with the method above), and then converting the result to a rotation matrix using Equation (5.94). If $\psi = 0$ however, and $^B\mathbf{z}_G$ is not too close to the fused yaw singularity $^Bz_{Gz} = -1$, we can alternatively use

$$^G_BR = \begin{bmatrix} R_{11} & R_{12} & -^Bz_{Gx} \\ R_{12} & R_{22} & -^Bz_{Gy} \\ {}^Bz_{Gx} & {}^Bz_{Gy} & {}^Bz_{Gz} \end{bmatrix}, \tag{7.274}$$

where

$$R_{11} = 1 - \frac{{}^B z_{Gx}^2}{1 + {}^B z_{Gz}}, \tag{7.275a}$$

$$R_{22} = 1 - \frac{{}^B z_{Gy}^2}{1 + {}^B z_{Gz}}, \tag{7.275b}$$

$$R_{12} = -\frac{{}^B z_{Gx}\,{}^B z_{Gy}}{1 + {}^B z_{Gz}}. \tag{7.275c}$$

### 7.3.7.2  *Composition of Yaw and Local Z-vector*

Suppose now that instead of a global z-vector, we are given a local z-vector ${}^G\mathbf{z}_B$ in combination with a desired fused yaw $\psi$. The most suitable methods of rotation reconstruction in this case, for the various available rotation representations, are presented below. Once again, the tilt phase space has been omitted, as its formulas follow trivially from the corresponding calculations for the tilt angles representation.

TILT ANGLES    From Equations (5.83a), (7.160a) and (7.160b), we can derive that

$$\gamma = \text{atan2}\big({}^G z_{Bx} c_\psi + {}^G z_{By} s_\psi,\; {}^G z_{Bx} s_\psi - {}^G z_{By} c_\psi\big). \tag{7.276}$$

As ${}^B z_{Gz} \equiv {}^G z_{Bz}$, we know from Equation (5.52b) that

$$\alpha = \text{acos}\big({}^G z_{Bz}\big), \tag{7.277}$$

so the required full tilt angles representation is then ${}^G_B T = (\psi, \gamma, \alpha)$.

FUSED ANGLES    From Equation (7.160), we can see that

$$\theta = \text{asin}\big({}^G z_{Bx} c_\psi + {}^G z_{By} s_\psi\big), \tag{7.278a}$$

$$\phi = \text{asin}\big({}^G z_{Bx} s_\psi - {}^G z_{By} c_\psi\big), \tag{7.278b}$$

$$h = \text{sign}\big({}^G z_{Bz}\big). \tag{7.278c}$$

Recalling the $\psi$ is given, the required full fused angles representation is then simply ${}^G_B F = (\psi, \theta, \phi, h)$.

QUATERNION    Given $\psi$ and ${}^G\mathbf{z}_B$, the w and z-components of the corresponding full quaternion rotation are given by

$$N_{wz} = \tfrac{1}{2}(1 + {}^G z_{Bz}), \tag{7.279a}$$

$$w = \sqrt{N_{wz}}\,\cos\tfrac{\psi}{2}, \tag{7.279b}$$

$$z = \sqrt{N_{wz}}\,\sin\tfrac{\psi}{2}. \tag{7.279c}$$

Unscaled x and y-components of rotation can then be calculated using

$$\bar{x} = {}^{G}z_{Bx}z - {}^{G}z_{By}w, \tag{7.280a}$$

$$\bar{y} = {}^{G}z_{By}z + {}^{G}z_{Bx}w. \tag{7.280b}$$

Being careful of the fused yaw singularity, the final quaternion is then

$$
{}^{G}_{B}q = \begin{cases} (0, \cos\frac{\psi}{2}, \sin\frac{\psi}{2}, 0) & \text{if } \bar{x} = \bar{y} = 0, \\ (w, A\bar{x}, A\bar{y}, z) & \text{otherwise,} \end{cases} \tag{7.281}
$$

where

$$A = \sqrt{\frac{1 - N_{wz}}{\bar{x}^2 + \bar{y}^2}}. \tag{7.282}$$

In the special case that we are dealing with a tilt rotation ($\psi = 0$), the entire process can be simplified down to

$$w = \sqrt{\tfrac{1}{2}(1 + {}^{G}z_{Bz})}, \tag{7.283a}$$

$$
{}^{G}_{B}q = \begin{cases} (0, \cos\frac{\psi}{2}, \sin\frac{\psi}{2}, 0) & \text{if } w = 0, \\ (w, -\frac{1}{2w}{}^{G}z_{By}, \frac{1}{2w}{}^{G}z_{Bx}, 0) & \text{otherwise.} \end{cases} \tag{7.283b}
$$

ROTATION MATRIX    The most numerically safe way of constructing the rotation matrix ${}^{G}_{B}R$ from a specification of $\psi$ and ${}^{G}\mathbf{z}_B$, is to construct the corresponding quaternion first (with the method above), and then converting the result to a rotation matrix using Equation (5.94). If $\psi = 0$ however, and ${}^{G}\mathbf{z}_B$ is not too close to the fused yaw singularity ${}^{G}z_{Bz} = -1$, we can alternatively use

$$
{}^{G}_{B}R = \begin{bmatrix} R_{11} & R_{12} & {}^{G}z_{Bx} \\ R_{12} & R_{22} & {}^{G}z_{By} \\ -{}^{G}z_{Bx} & -{}^{G}z_{By} & {}^{G}z_{Bz} \end{bmatrix}, \tag{7.284}
$$

where

$$R_{11} = 1 - \frac{{}^{G}z_{Bx}^2}{1 + {}^{G}z_{Bz}}, \tag{7.285a}$$

$$R_{22} = 1 - \frac{{}^{G}z_{By}^2}{1 + {}^{G}z_{Bz}}, \tag{7.285b}$$

$$R_{12} = -\frac{{}^{G}z_{Bx}\,{}^{G}z_{By}}{1 + {}^{G}z_{Bz}}. \tag{7.285c}$$

### 7.3.7.3  Composition of Mismatched Yaw and Tilt

So far, in Sections 7.3.7.1 and 7.3.7.2, we have been combining specifications of yaw and tilt that are both relative to the same frame {G}. Suppose now that we wish to construct a rotation ${}^{G}_{B}q \in \mathbb{Q}$ that has a particular fused yaw ${}^{G}\psi$ relative to {G}, but a tilt rotation component of $^Hq_t \in \mathbb{Q}$ relative to another frame {H}. More specifically, suppose we are given a fused yaw $^G\psi$, the relative rotation $^G_Hq = (w_H, x_H, y_H, z_H)$ between the two global reference frames {G} and {H}, and any rotation $^H_Cq = (w_C, x_C, y_C, z_C)$ relative to {H} that has the required tilt rotation component $^Hq_t$. Clearly, the frame {B} that satisfies both $^G\psi$ and $^Hq_t$ must be a simple z-rotation of {C} about $\mathbf{z}_H$. The remaining question however, is what magnitude this z-rotation should have to cause {B} to have the desired fused yaw relative to {G}.

If $^{HC}_Bq$ is the *referenced rotation* that denotes the required z-rotation from {C} to {B} relative to {H}, we know from Equation (7.65a) (replace $GA$ with $HC$, and $R$ with $q$) that

$$^H_Bq = {}^{HC}_Bq \, {}^H_Cq, \tag{7.286}$$

i.e. $^{HC}_Bq$, when applied relative to {H}, rotates frame {C} to frame {B}. Thus, premultiplying by $^G_Hq$ to express {B} relative to {G} yields

$$^G_Bq = {}^G_Hq \, {}^{HC}_Bq \, {}^H_Cq. \tag{7.287}$$

As a z-rotation, we know that $^{HC}_Bq$ must be of the form

$$^{HC}_Bq = (\cos\tfrac{\beta}{2}, \, 0, \, 0, \, \sin\tfrac{\beta}{2}), \tag{7.288}$$

for $\beta \in (-\pi, \pi]$, so

$$^G_Bq = (w_H, x_H, y_H, z_H)(\cos\tfrac{\beta}{2}, \, 0, \, 0, \, \sin\tfrac{\beta}{2})(w_C, x_C, y_C, z_C). \tag{7.289}$$

Expanding the right-hand side of Equation (7.289)—which is solely a function of $\beta$—and solving

$$\Psi(^G_Bq) = {}^G\psi, \tag{7.290}$$

then leads to the total solution

$$^{HC}_Bq = \frac{1}{\sqrt{E^2 + F^2}} \, (E, 0, 0, F), \tag{7.291}$$

where, using the notation $c_{\bar\psi} = \cos\left(\tfrac{1}{2}{}^G\psi\right)$ and $s_{\bar\psi} = \sin\left(\tfrac{1}{2}{}^G\psi\right)$,

$$a = x_H x_C + y_H y_C, \qquad\qquad b = x_H y_C - y_H x_C, \tag{7.292a}$$
$$c = w_H z_C + z_H w_C, \qquad\qquad d = w_H w_C - z_H z_C, \tag{7.292b}$$
$$E = (d+a)c_{\bar\psi} - (b-c)s_{\bar\psi}, \qquad F = (d-a)s_{\bar\psi} - (b+c)c_{\bar\psi}. \tag{7.292c}$$

Relative to frames {H} and {G}, frame {B} is subsequently given by Equations (7.286) and (7.287).

The presented method for constructing {B} fails if and only if $E = F = 0$. Provably, this occurs exactly when

$$\alpha_H + \alpha_C = \pi, \tag{7.293}$$

where $\alpha_H$ and $\alpha_C$ are the tilt angles of $^G_H q$ and $^H_C q$, respectively. In this case, every {B} that has the required tilt rotation component relative to {H} has the same fused yaw relative to {G}, namely $\Psi\left(^G_C q\right)$. As such, if $^G\psi = \Psi\left(^G_C q\right)$ there are infinitely many solutions—of which the frame {C} itself is one—but otherwise there are no solutions.

### 7.3.8    Relation to Topological Concepts

The study of 3D rotations and rotation representations ties in nicely with a variety of topological theorems and concepts. In addition to the ones already mentioned in this chapter (e.g. in Sections 7.1.2.3 and 7.3.1), a small selection of such results are presented in this section.

#### 7.3.8.1    *Rotation Space Topologies*

Topologies are important when studying any kind of mathematical space, as they tell us how the elements of a space are related, which kinds of elements are in the neighbourhood of each other, and how the space 'joins up' into a manifold. Topologies help us understand a space—how it behaves, what properties it has, what it looks like, and what other spaces it is similar (homeomorphic) to in a continuous way. For instance, if we consider the space SO(3) of all 3D rotations, in particular in the form of rotation matrices, we know how to understand each element individually as a 3D rotation in space. However, if we consider every rotation matrix as just some 'point' on a higher-dimensional manifold, it is at first quite difficult to understand what this rotation space manifold should look and behave like—all we know is that rotations that are close in real life should be close together as points on the manifold.

We recall that we were able to understand the manifold of all possible quaternion representations as the four-dimensional unit sphere $\mathcal{S}^3$ (see Figure 7.3). This hypersphere is a three-dimensional manifold embedded in four-dimensional space, just like a circle is a one-dimensional line embedded in a two-dimensional plane. Quaternions are a double cover of SO(3) however, as every distinct pair of quaternions $q$ and $-q$ corresponds to the same physical rotation. Thus, knowing that no other such correspondences occur, from a topological perspective we can see that SO(3) must be structurally identical to the space that results when we start with the 3-sphere, and then join together every pair of antipodal (diametrically opposite) points. Stated more precisely, if we consider $\mathcal{S}^3$ with the standard Euclidean topology, i.e. the standard topology of $\mathbb{R}^4$ restricted to this manifold, and topologically identify all pairs of antipodal points, then SO(3) is homeomorphic (in fact diffeomorphic) to the space that results, which from standard topology we know to be the real projective space $\mathbb{R}P^3$. Thus, based on our knowledge of this standard topological space, we can gain new appreciation for the structure and order of the rotation space SO(3).

The real projective space $\mathbb{R}P^n$ is a compact and smooth manifold of dimension $n$, and can be constructed in any of the following topologically equivalent ways:

A) The space of all lines through the origin $\mathbf{0}$ in $\mathbb{R}^{n+1}$,

B) The quotient space of $\mathbb{R}^{n+1} \setminus \{\mathbf{0}\}$ under the equivalence relation $\mathbf{x} \sim \lambda \mathbf{x}$, for $\lambda \neq 0$,

C) The space formed by identifying all antipodal points of $\mathcal{S}^n$ in $\mathbb{R}^{n+1}$,

D) The space formed by identifying all antipodal points on the $\mathcal{S}^{n-1}$ bounding equator of just one hemisphere of $\mathcal{S}^n$ in $\mathbb{R}^{n+1}$,

E) The space formed by identifying all antipodal points on the boundary of the closed unit disc $\bar{\mathcal{D}}^n(1)$ in $\mathbb{R}^n$.

As special cases, $\mathbb{R}P^1$ is referred to as the *real projective line*, and is topologically equivalent to a circle or loop, and $\mathbb{R}P^2$ is referred to as the *real projective plane*.

Now that we know the topological structure of the entire rotation space SO(3), we can continue to look at the smaller spaces that result when we partition rotations into their respective yaw and tilt components. Clearly, the space of all *yaw rotation components* $\psi$ is $(-\pi, \pi]$, and loops around continuously at $\pm \pi$. Thus, we can see that it corresponds to a circular topology, which is equivalently both the real projective line $\mathbb{R}P^1$, and Euclidean $\mathcal{S}^1$. The space of all *tilt rotation components* TR(3), on the other hand, is more difficult to interpret. If we follow the same approach as for SO(3) above, and look at the quaternion representations instead of the rotation matrix representations, we can see that the general tilt rotation form is exactly

$$q = (w, x, y, 0), \tag{7.294}$$

for all

$$w^2 + x^2 + y^2 = 1. \tag{7.295}$$

We can see as a result that the space of all tilt rotation quaternions corresponds exactly to the standard sphere $\mathcal{S}^2$ embedded in $\mathbb{R}^4$. Like before however, this is still a double cover of TR(3) due to the equivalence of $\pm q$. Thus, by identifying $q$ and $-q$ topologically on $\mathcal{S}^2$, we can see from Definition C that TR(3) is equivalent to the real projective plane $\mathbb{R}P^2$. This is a remarkable result, and further demonstrates why the partition of rotations into fused yaw and tilt is so natural, as the entire $\mathbb{R}P^3$ rotation space is being divided up into two topologically similar, but lower-dimensional, $\mathbb{R}P^1$ and $\mathbb{R}P^2$ components. Despite being only a two-dimensional manifold, $\mathbb{R}P^2$ famously cannot be embedded into 3D space (i.e. represented in $\mathbb{R}^3$ without self-intersections), but for visualisation purposes it can be *immersed* in $\mathbb{R}^3$, as shown in Figure 7.10. This figure shows Boy's surface, and is how we can imagine the *structure* of the space of all tilt rotations. If we were able to mentally add a dimension and some projective complexity to this surface, this would also be how we can visualise the structure of the full rotation space SO(3).

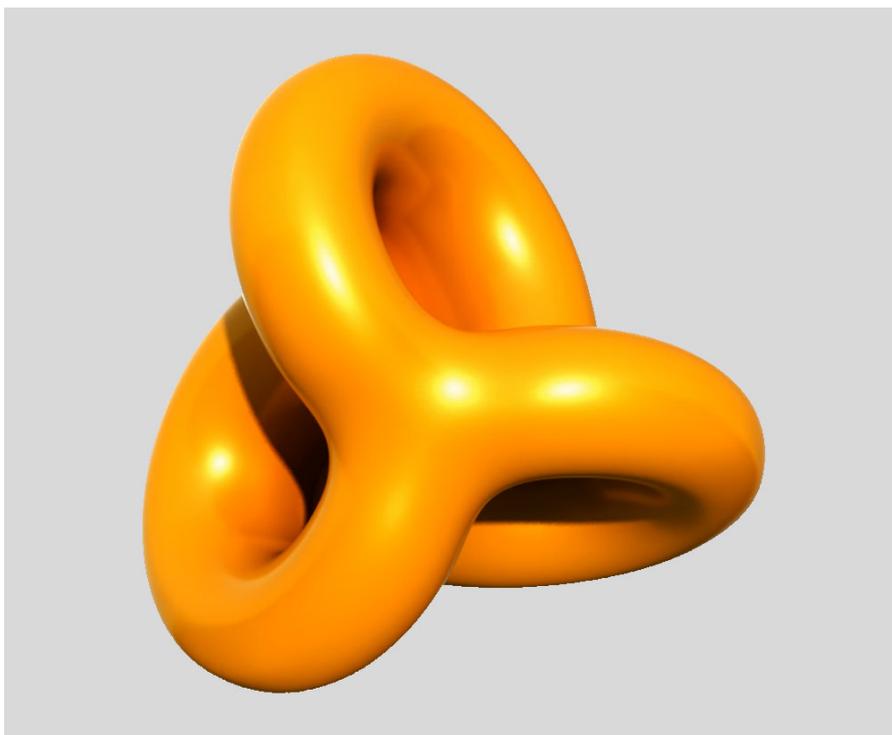

Figure 7.10: Visualisation of a possible (self-intersecting) immersion of Boy's surface into $\mathbb{R}^3$. The surface is single-sided, like a higher-dimensional Möbius strip, and can be thought of as an 'inner chamber' (centre of the image) with three separate intertwined openings to the 'outside'.
Source: https://youtu.be/9gRx66xKXek

If we look now at the various possible parameterisations of the space of tilt rotations, as given in Section 5.4, we can see based on parameter equivalences and the corresponding quotient spaces (see Section 7.3.4.1) that the topologies corresponding to the tilt angles, tilt phase space and quaternion parameterisations are all $\mathbb{RP}^2$. For tilt angles this result is not immediately obvious, but first applying the equivalence

$$(\gamma, 0) \sim (0, 0) \tag{7.296}$$

to the initially open-ended cylindrical space $(\gamma, \alpha) \in (-\pi, \pi] \times [0, \pi]$ yields something akin to an open-ended cone. Further closing up the remaining open end with

$$(\gamma, \pi) \sim (\text{wrap}(\gamma + \pi), \pi), \tag{7.297}$$

i.e. by identifying antipodal points on the circle $(\gamma, \pi)$ for $\gamma \in (-\pi, \pi]$, yields $\mathbb{RP}^2$ then as required, by Definition D. If we look at the $z$-vector and fused angles parameterisations of TR(3) however, we notice after applying the required parameter equivalences that their topologies are both in fact equivalent to Euclidean $\mathcal{S}^2$, and not $\mathbb{RP}^2$. The fundamental difference is how these representations deal with 'closing up' the tilt rotations space at the fused yaw singularity. While for example the tilt phase space only joins up pairs of antipodal points on its $\alpha = \pi$ boundary, the $z$-vector parameterisation joins up this entire boundary to form the single point

$$^{B}\mathbf{z}_G = (0, 0, -1), \tag{7.298}$$

and the fused angles parameterisation does the same to form the single point

$$(\theta, \phi, h) = (0, 0, -1). \tag{7.299}$$

This alleviates all kinds of problems at or near the fused yaw singularity, but in some sense sacrifices expressive power in return. The power of the study of topology is that such similarities and differences between representations can so clearly be seen, despite their greatly varying definitions and nature.

As a final note, having knowledge of the topologies of 3D rotation spaces allows us to be more mathematically rigorous about what we mean by the concept of continuity. As a topology is nothing other than a definition of all the sets in a space that are considered to be open, it implicitly also defines which functions to and from the space are considered to be continuous. For instance, if we consider the tilt rotation phase space domain $\bar{\mathcal{D}}^2(\pi)$ and were to look at this space with the standard Euclidean topology, we would find that all trajectories that pass through $\alpha = \pi$ are discontinuous. This is clear because at some point each such trajectory jumps from one point on the boundary of $\bar{\mathcal{D}}^2(\pi)$ to the diametrically opposite one, and according to the Euclidean topology these two boundary points are separable by open sets. From the viewpoint of the real projective topology however, this is not the case, as the points are identified and all open sets that contain one of the points by definition also contain the other. Conveniently, we can think of the open sets in the real projective topology as the sets that result when open patches on Boy's surface (minus the self-intersections) are mapped back to $\bar{\mathcal{D}}^2(\pi)$.

### 7.3.8.2 *Yaw and Tilt as a Hopf Fibration*

The way in which 3D rotations are partitioned into fused yaw and tilt components can be seen to emerge naturally from the mathematical concept of the Hopf fibration. The Hopf fibration is a continuous way of mapping the 3-sphere $\mathcal{S}^3$ onto the 2-sphere $\mathcal{S}^2$ in a many-to-one manner, such that the preimage of every point in $\mathcal{S}^2$ is a circle in $\mathcal{S}^3 \subset \mathbb{R}^4$. Identifying circles in $\mathbb{R}^4$ as being equivalent to $\mathcal{S}^1$, we write

$$\mathcal{S}^1 \hookrightarrow \mathcal{S}^3 \xrightarrow{\ h\ } \mathcal{S}^2, \tag{7.300}$$

where we interpret this to mean that the *fiber space* $\mathcal{S}^1$ (i.e. each circle in $\mathcal{S}^3$) is embedded in the *total space* $\mathcal{S}^3$, and the <span style="color:purple">Hopf map</span>

$$h : \mathcal{S}^3 \to \mathcal{S}^2 \tag{7.301}$$

maps each of the fibers (circles) in $\mathcal{S}^3$ to a unique point in the *base space* $\mathcal{S}^2$. There is no unique way of defining the Hopf map, but one set of possible definitions is of the form

$$h_{\hat{\mathbf{v}}} : q \mapsto L_q(\hat{\mathbf{v}}), \tag{7.302}$$

where $q \in \mathbb{Q} \cong \mathcal{S}^3$, and $\hat{\mathbf{v}} \in \mathcal{S}^2$ is some fixed unit vector. Essentially, the Hopf map $h_{\hat{\mathbf{v}}}$ maps every quaternion $q$ to the unit vector that it rotates the fixed unit vector $\hat{\mathbf{v}}$ to. Clearly, the map is different for every choice of $\hat{\mathbf{v}} \in \mathcal{S}^2$, and there is a one-dimensional subspace of quaternions that maps $\hat{\mathbf{v}}$ to any particular $\hat{\mathbf{w}} \in \mathcal{S}^2$. In fact, it can be demonstrated that this subspace, i.e. the preimage $h_{\hat{\mathbf{v}}}^{-1}(\hat{\mathbf{w}}) \subset \mathbb{Q}$, is a circle in $\mathbb{Q}$ for each $\hat{\mathbf{w}}$, and is referred to as the *fiber* over $\hat{\mathbf{w}}$ of the Hopf map with respect to $\hat{\mathbf{v}}$.

Let us consider the specific Hopf map for $\hat{\mathbf{v}} = \hat{\mathbf{e}}_z \equiv (0,0,1)$, but with the slight modification that the conjugate (i.e. inverse) of $q$ is used to rotate $\hat{\mathbf{v}}$ instead of $q$ itself. That is, let us consider the map

$$h_z : \mathbb{Q} \to \mathcal{S}^2, \ q \mapsto L_{q^*}(\hat{\mathbf{e}}_z). \tag{7.303}$$

Clearly, despite not being in the form given by Equation (7.302), this is still a valid Hopf fibration, as the preimage of every point in the codomain is still a circle in $\mathbb{Q}$. This is because $h_z$ can be considered to be the composition of the map $q \mapsto q^*$ with the Hopf map $h_{\hat{\mathbf{e}}_z}$, and the former bidirectionally maps circles to circles. We recall from Equation (5.131b) that we can express every quaternion $q$ as the combination of a fused yaw z-rotation $q_z(\psi)$ and a tilt rotation $q_t$, i.e.

$$q = q_z(\psi) q_t. \tag{7.304}$$

Thus, we can see that

$$\begin{aligned}
h_z(q) &= L_{q^*}(\hat{\mathbf{e}}_z) \\
&= L_{(q_z(\psi) q_t)^*}(\hat{\mathbf{e}}_z) \\
&= L_{q_t^* q_z(-\psi)}(\hat{\mathbf{e}}_z) \\
&= L_{q_t^*}\big(L_{q_z(-\psi)}(\hat{\mathbf{e}}_z)\big) \\
&= L_{q_t^*}(\hat{\mathbf{e}}_z).
\end{aligned} \tag{7.305}$$

We can interpret this result as saying that $h_z(q)$ depends exactly only on the tilt rotation component of $q$, and is completely independent of the yaw rotation component $\psi$. In fact, we can quickly see if $q \equiv{}^G_B q$, that

$$h_z(q) = L_{q_t^!}(\hat{\mathbf{e}}_z) = {}^B\mathbf{z}_G, \tag{7.306}$$

which is exactly the global z-vector parameterisation of the tilt rotation component of $q$. We conclude that the Hopf map $h_z$ implicitly demonstrates how to partition 3D rotations into their respective fused yaw and tilt rotation components. The yaw rotation component corresponds to the fibers $h_z^{-1}({}^B\mathbf{z}_G) \cong \mathcal{S}^1 \cong (-\pi, \pi]$, and the tilt rotation component corresponds to the vectors $h_z(q) = {}^B\mathbf{z}_G$, which form the space $\mathcal{S}^2$. Even though the fibers can be seen to represent the yaw rotation component, it should be noted that the Hopf map however makes no explicit suggestion how this component can or should be parameterised.

The Hopf fibration, and the way in which it partitions the quaternion rotation space $\mathbb{Q} \cong \mathcal{S}^3$ into circles, can be visualised in 3D by means of stereographic projection. If we consider the projection map

$$\begin{aligned} p : \mathbb{Q}\backslash\{(1,0,0,0)\} &\to \mathbb{R}^3 \\ (w, x, y, z) &\mapsto \tfrac{1}{1-w}(x, y, z), \end{aligned} \tag{7.307}$$

we can see that with the exception of one point, every quaternion $q \in \mathbb{Q} \subset \mathbb{R}^4$ maps continuously to a unique point in $\mathbb{R}^3$. This remarkably allows us to visualise essentially all of $\mathbb{Q}$ in standard 3D space. Even more remarkably, the projection map $p$ maps circles to circles (except for circles through $(1, 0, 0, 0)$, which map to lines in $\mathbb{R}^3$), so it can be concluded that the circular fibers of the Hopf fibration stereographically project to circles in $\mathbb{R}^3$ (and one line), as illustrated in Figure 7.11. Furthermore, as the fibers partition $\mathbb{Q}$, and $p$ is one-to-one, the projected circles can be concluded to partition $\mathbb{R}^3$. As a fun fact, every pair of projected circular fibers in $\mathbb{R}^3$ is interlinked, i.e. every single circle passes through every single other circle exactly once, and the single projected line that results from the fiber through $(1, 0, 0)$ also passes through every single other circular fiber. More information on the Hopf fibration, including in particular animated visualisations of the fibers, can be found at the footnoted link.[20]

---

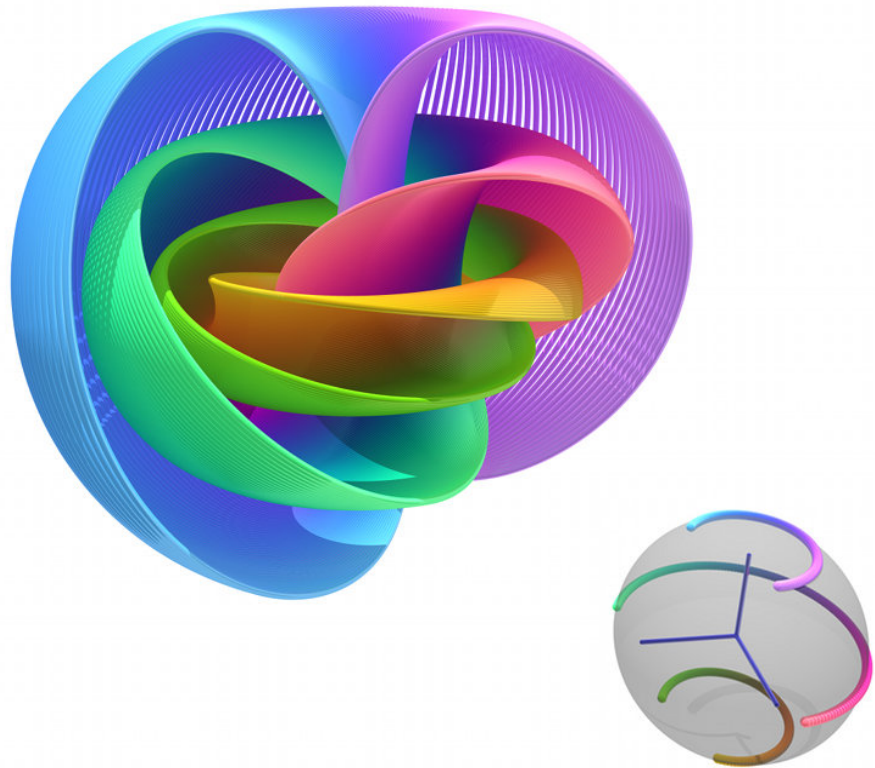

Figure 7.11: Visualisation of the Hopf fibration via stereographic projection of $\mathcal{S}^3$ to $\mathbb{R}^3$, and further compression of $\mathbb{R}^3$ into a ball. Various unit vectors $\hat{\mathbf{w}} \in \mathcal{S}^2$ are shown as points on the small sphere on the bottom right, and their corresponding (compressed) circular fibers are shown in the corresponding colour in the main image. Note, remarkably, that every pair of fibers is interlinked.

Source: https://commons.wikimedia.org/wiki/File:Hopf_Fibration.png

See also: https://youtu.be/AKotMPGFJYk

Part IV

BIPEDAL WALKING

## RELATED WORK

In this chapter, some of the most prominent and relevant examples of the state of the art in balanced bipedal walking are presented, to establish a background of the methods that exist in literature for such applications.

### 8.1 ZMP-BASED GAIT GENERATION

Many modern bipedal gaits are, in a wide variety of ways, based on the concept of the Zero Moment Point (ZMP). The ZMP of a bipedal robot is the point within the support polygon of the robot, i.e. within the convex hull of the ground contact patches, about which the net resultant ground reaction wrench has zero moment parallel to the plane of contact (Vukobratović and Borovac, 2004). If such a point does not exist because it occurs outside the support polygon, the point is instead referred to as the 'fictitious' ZMP, but often for simplicity this technical distinction is not made. The ZMP is of interest to situations of walking and balancing as it can be used to formulate a (conservative) stability criterion. If during a walking motion the actual physical ZMP remains within the support polygon at all times, the motion is guaranteed to be stable and the robot will not fall.

Early methods of ZMP-based gait stabilisation and balance control, such as for example the one used for the ASIMO robot (Hirai et al., 1998), used ZMP feedback control loops that effectively monitored the difference between the measured and desired ZMP points, and applied foot position, posture and body trajectory adjustments to counteract any unbalanced tipping moments. The desired ZMP in this case was calculated from the ideal walking pattern using a detailed physical model of the robot. Later methods of ZMP-based gait generation constructed reference ZMP trajectories based on desired walking motions, and used these to compute reference trajectories for the Centre of Mass (CoM). These trajectories were then tracked as closely as possible, generally with the use of inverse kinematics and an appropriate control law.

Given a piecewise polynomial ZMP trajectory, Harada et al. (2004) analytically derived a formulation for the reference CoM trajectory. This was later extended by Morisawa et al. (2007) to allow modifications of foot placement, while still maintaining the analytic solution approach. Harada et al. (2006) presented a method for simultaneously and analytically planning CoM and ZMP trajectories in real-time by connecting newly calculated trajectories to previous ones at the begin-ning of double support. Morisawa et al. (2009) extended this approach
to allow footstep locations to be treated as free variables, allowing
reactive steps for push recovery to be realised. This was integrated
with a state feedback controller in Morisawa et al. (2010) so that dis-
turbances could be rejected either with step adaptations or disturbance
suppression methods depending on their severity.

Yi et al. (2011) presented a gait for the DARwIn-OP robot that
uses a step controller that plans a ZMP trajectory two steps into the
future, and generates corresponding CoM trajectories in closed form
directly from the Linear Inverted Pendulum Model (LIPM) equations.
Jerk minimisation is not considered, and the jerk of the CoM is in
fact discontinuous in the middle of each support transition, but this
does not seem to have a significant effect on the stability of the gait.
A push recovery controller is integrated with the walking controller,
and allows ankle, hip and stepping strategies to be applied when
disturbances are detected. The activations of the individual push
recovery strategies are learnt over many real-world trials within a
reinforcement learning setting.

### 8.1.1  Preview Control

A basis for many works is ZMP tracking with preview control, first
introduced in the context of bipedal walking by Kajita et al. (2003).
A series of footsteps are planned and used to define a reference ZMP
trajectory with the use of suitable heuristics. This trajectory is often
locally fixed once it has been computed, but often also recomputed
online in response to tracking errors and/or disturbances (Nishiwaki
and Kagami, 2010; Tedrake et al., 2015). Based on the Linear Inverted
Pendulum Model described in Kajita et al. (2001), an optimal preview
controller is constructed that utilises state feedback, integral tracking
error feedback, and preview action based on the ZMP reference for
a given future time window (Kajita et al., 2003). This controller acts
to minimise ZMP tracking error under consideration of the planned
future steps, and allows the robot to walk in relatively controlled and
non-dynamic situations.

Many variations and follow-up works of the original ZMP preview
control idea have been published over the years. For instance, Kim et al.
(2019) added a control performance model that approximates the track-
ing performance of the CoM (including the effects of the implemented
low-level tracking controller) using a mass-spring-damper between
the desired CoM and real CoM. This slightly relieves the dependence
on high performance tracking ability of the robot, and increases the
applicability of preview control to compliant robots. Kajita et al. (2018)
also investigated the use of Spatially Quantised Dynamics (SQD) in
combination with preview control for stretched-knee bipedal walking.
While the lateral motions of the CoM are handled by a preview control approach, the sagittal walking trajectories are first designed kinematically, before being quantised spatially based on the hip joint position, and subsequently refined using Differential Dynamic Programming (DDP) to respect the nonlinear spatially quantised LIPM equations. During execution of the thereby attained walking motions, the SQD LIPM equations are once again used to return deviating measured states to the desired trajectory. The real benefit of using the SQD formulation is admittedly unclear to the authors of the paper themselves (Kajita et al., 2018). They also identify a lacking theoretical background for their use of local optimisation for feedback control, and need to permanently manually enforce non-zero sagittal reference velocities in order to avoid division-by-zero singularities arising from their spatial quantisation.

Variable step timing is rarely addressed by ZMP-based gait generation methods, especially in terms of online replanned reactions to unforeseen disturbances. Kryczka et al. (2015) have presented a ZMP-based method for incorporating some level of timing feedback into the trajectory optimisation process (mainly only lateral pushes towards the stance foot). ZMP preview control as per Kryczka et al. (2013) is used to generate the nominal gait pattern using the LIPM. In addition to this, CoM and ZMP feedback controllers are used to overcome the compliance of the robot. If during walking the error in the CoM position exceeds a certain threshold, the gait trajectory is replanned from the current state using nonlinear optimisation of the following two step timings and the following sagittal and lateral step positions. The aim of the replanned gait trajectories is to return the robot to the nominal CoM position and velocity that it should have at the beginning of a subsequent step. The replanned gait patterns are executed using a preview control scheme, and are spliced into the nominal gait patterns using interpolation over a small time window.

When a disturbance to a stationary robot is detected, Urata et al. (2011) optimise two future footstep locations and one future footstep time, selecting them from a number of computed ZMP/CoM trajectory pairs for which the CoM trajectory does not diverge. The CoM trajectories are calculated explicitly using singular LQ preview regulation, based on the corresponding heuristically generated ZMP reference. The executed motion plan is fixed to always consist of exactly three steps, but this is sufficient to produce very convincing disturbance rejection, albeit on specially designed hardware that is able to execute the planned motions with extreme fidelity.

### 8.1.2 Model Predictive Control

Given a method for calculating a reference ZMP trajectory, an alternative approach to optimising a CoM trajectory to match this ZMP trajectory is given by Model Predictive Control (MPC). A landmark paper in this direction is Wieber (2006), in which a Linear Model Predictive Control scheme is developed that solves a continuous sequence of finite-horizon optimal control problems, and in each cycle only executes the first time step of the resulting calculated CoM trajectory (with the help of a tracking controller). While not the first to use MPC in the setting of bipedal locomotion (Azevedo et al., 2002), the innovation was to use simple LIPM dynamics and ZMP reference trajectories (generated as for ZMP preview control) for the purpose of optimisation. This allows the crux Quadratic Program (QP) to be solved analytically using matrix manipulations instead of numerical techniques, and thereby allows the scheme to work in real-time. In each time step, the optimal control problem tries to minimise CoM jerk and deviations from the ZMP reference trajectory, and as a form of feedback uses the current measured robot configuration as the initial state. While this already allows certain robustness to disturbances, Wieber (2006) also investigated reduction of the QP to minimising just the CoM jerk, but under explicit consideration of ZMP location constraints. This formulation of the QP no longer has an analytic solution, but allowed for more aggressive ZMP/CoM trajectory pairs, leading to greater disturbance rejection abilities.

To overcome the restriction of having fixed footstep locations, Diedam et al. (2008) extended the work of Wieber (2006) to include the translational locations of the footsteps in the optimisation task. Nominal footstep locations are generated as before, but instead of being used directly, these are used to penalise (via the cost function) the generation of steps that deviate from them. Additional inequality constraints are added to the MPC to ensure that only feasible steps are produced. Stephens and Atkeson (2010) also included footstep locations in their linear MPC scheme, which was targeted at push recovery for a torque-controlled robot. Their method however was constrained to push recovery from standing to standing in a fixed number of steps (generally one) with fixed timing.

Allowing non-fixed footstep yaws and heights of the CoM during walking introduces significant nonlinearities to the Linear Inverted Pendulum Model. Griffin and Leonessa (2019) proposed an MPC scheme based on the time-varying Divergent Component of Motion (DCM) dynamics (Hopkins et al., 2014) that allows both height (predefined CoM height trajectories) and foot orientation changes during walking. Smooth DCM trajectories with fixed step times are generated using DCM acceleration as the control input, and executed using reverse-time numerical integration and feedback control. Piecewise linear approximations of the constraints, in particular in relation to the foot orientations, leads to a Mixed-Integer Quadratically Constrained Program (MIQCP) that cannot quite be solved fast enough to be real-time.

Further examples of works that consider free foot placements and time-variant CoM heights include Kuindersma et al. (2014) and Brasseur et al. (2015), both of which conservatively bound nonlinearities using linear functions in order to allow more efficient solution methods. Building on the latter of these two works, Pajon and Wieber (2019) introduced a linear MPC scheme that incorporates a 3D capturability constraint for greater passive safety of the generated walking motions. All generated future trajectories of the robot CoM and Centre of Pressure (CoP) are constrained to end at their preview horizon with a dynamically balanced convergent stopping motion, i.e. an exponential continuation of the previewed CoM trajectory that in limit $t \to \infty$ settles to a stable position that involves the CoM stopping somewhere over the final support foot.

The main issue with bipedal gaits based on the Model Predictive Control method is their high computational intensity. A mix of linearisations, approximations and design choices must generally be made to reduce the complexity of the problem to a tractable level, but even so, a large amount of computation is required in every cycle, only to be thrown away after executing merely the very first time step of the calculated trajectory. Some other walking methods, for example as described by Feng et al. (2015), replan their calculated CoM trajectories only once per single support phase, but this to some extent sacrifices immediate responsiveness to disturbances.

## 8.2 CAPTURE POINT AND DIVERGENT COMPONENT OF MOTION

The concept of capture points and capture regions (the set of all capture points) was first introduced by Pratt et al. (2006). Most generally, a capture point of a robot in a certain state is a point on the ground that if stepped on brings the robot to a complete stable stop. Although complicated to calculate for general bipedal robots, for simple models like the LIPM there is a unique capture point that can easily be calculated based on the CoM position and velocity. Push recovery of a real robot based on the capture point was first demonstrated by Pratt et al. (2009). Later, Englsberger et al. (2011) presented a landmark walking gait based on stabilisation of the divergent component of the LIPM capture point dynamics.

Koolen et al. (2012) presented a treatise on the $N$-step capturability of walking robots as it relates to three simple gait models based on the 3D LIPM, and refined the definition of the capture point to include the instantaneous capture point. This is essentially the same as the original definition of the capture point, only without the restrictions of dynamic reachability, i.e. without consideration of step length and time limitations. Pratt et al. (2012) demonstrated a gait based on the instantaneous capture point. The current capture region is first estimated based on the current state and capturability analysis of the 3D LIPM with finite-sized feet, before a desired footstep is calculated based on that. A state machine is used to break down the gait cycle into nine different stages, and appropriate leg motions are generated to perform the required footstep using individual swing and stance leg controllers. Griffin et al. (2017) incorporated some notion of step timing adjustment into their planned capture point trajectories for walking. Utilising a QP optimisation scheme, adjustments to an existing planned capture point trajectory were undertaken to produce plans that reduced the amount of required knee bend (while remaining kinematically feasible).

Assuming flat ground and a constant CoM height, Koolen et al. (2016) presented a gait for the Atlas robot that generates instantaneous capture point trajectories at the beginning of each single support phase, and calculates desired Centroidal Moment Pivot (CMP) trajectories using a Proportional-Integral (PI) control law aimed at capture point tracking. The desired CMP is used to construct a desired linear momentum rate of change of the robot, which is subsequently unified with contact information and constraints in a low-level QP aimed at providing joint accelerations and ground reaction wrenches to the inverse dynamics used to calculate the required joint torques. Although a constant CoM height is assumed, the momentum-based control framework was successfully used both on rough terrain and stairs.

Parallel to the development of the instantaneous capture point and its applications to bipedal walking, Takenaka, Matsumoto and Yoshiike (2009) developed the essentially equivalent notion of the Divergent Component of Motion (DCM), and used it as a relaxed boundary constraint for transitioning between cyclic gait patterns on the ASIMO robot. Takenaka, Matsumoto, Yoshiike and Shirokura (2009) extended this method of pattern generation to running by including consideration of vertical and flywheel models in parallel to the three-mass Linear Inverted Pendulum Model already used for forwards locomotion planning. In addition to enforcing continuous ZMP and DCM locations at the transition points between the calculated gait patterns (by adding trapezoidal offsets to the planned ZMP waveforms), continuous flywheel moments were also enforced to control the inclination of the torso. Takenaka, Matsumoto, Yoshiike, Hasegawa et al. (2009) include details of the balance controller used to stabilise the upper body position during running of the ASIMO robot. In addition to joint angle, ground reaction force and body inclination control schemes, an implemented model ZMP control scheme uses horizontal and rotational sagittal accelerations of the torso based on feedback to stabilise the robot. This results in possible adaptations of step placement and timing in order to keep the robot balanced while satisfying all kinematic constraints.

The concepts of the instantaneous capture point and DCM were extended to 3D by Englsberger et al. (2015), and lead to the development of the notions of the Enhanced Centroidal Moment Pivot (eCMP) and Virtual Repellent Point (VRP). The 3D DCM is a point in space a certain distance in front of the CoM in the direction that it is instantaneously moving. That is,

$$\boldsymbol{\xi} = \mathbf{x} + b\dot{\mathbf{x}}, \tag{8.1}$$

where $\mathbf{x} \in \mathbb{R}^3$ is the position of the CoM, $\dot{\mathbf{x}} \in \mathbb{R}^3$ is its instantaneous velocity, and $b$ is a time-constant. The eCMP is a point that encodes the sum of all external forces $\mathbf{F}_{ext}$ (excluding gravity), and is located behind the CoM, in the exact opposite direction to which $\mathbf{F}_{ext}$ points. Specifically,

$$\mathbf{r}_{ecmp} = \mathbf{x} - \frac{b^2}{m} \mathbf{F}_{ext}, \tag{8.2}$$

where $m$ is the mass of the robot. The VRP is equivalent to the eCMP, only with consideration of the force of gravity as well, i.e.

$$\mathbf{r}_{vrp} = \mathbf{x} - \frac{b^2}{m} \mathbf{F}_{tot}, \tag{8.3}$$

where $\mathbf{F}_{tot} = \mathbf{F}_{ext} + \mathbf{F}_g$. One can observe that the eCMP and VRP only differ in their z-coordinate, by the constant height $b^2 g$, and that the resultant DCM dynamics simply become

$$\dot{\boldsymbol{\xi}} = \frac{1}{b}(\boldsymbol{\xi} - \mathbf{r}_{vrp}). \tag{8.4}$$

This can be interpreted as the VRP point always exponentially 'pushing away' the DCM along the ray that connects them. Englsberger et al. (2015) use this fact to plan reference DCM trajectories for walking on uneven terrain based on VRP points that are chosen to be fixed at a constant height above the required footsteps. These are then tracked using a force-based DCM tracking control law based on modifications of the VRP. The associated reference CoM trajectories are calculated from the DCM trajectories using the convergent component of the CoM dynamics. Extensions that include continuous double support phases and heel-to-toe eCMP trajectories (as opposed to a fixed eCMP at the foot centre) are also presented.

Many modern methods of gait generation and control rely on the Divergent Component of Motion. Englsberger et al. (2017) and Mesesan et al. (2018) use polynomial splines to generate VRP and DCM trajectories analytically in closed form, and generate multi-step reference trajectories based on this. Englsberger et al. (2017), in particular, provide an explicit mechanism for push recovery via step adjustment, and propose a momentum-based disturbance observer for estimating and overcoming continuous perturbations. While only providing simulation results, real-robot experiments with the TORO robot were later published in Englsberger et al. (2018), in the context of a QP-based whole-body control framework. Khadiv et al. (2016) also presented a method for incorporating step adjustments into a DCM planning and tracking scheme. Its application to a simulated robot with passive ankles is demonstrated, but it was found that its performance was highly dependent on the manually chosen step timing.

Hopkins et al. (2014) extended the concept of the DCM to a time-varying version, and used it to construct stepping trajectories with generic variability in the CoM height. Stair climbing using a pre-planned CoM height trajectory was demonstrated, albeit only in simulation. The scheme runs in real-time, but no provision is made for real-time step adjustment or trajectory replanning as a response to disturbances. Caron et al. (2019) used a whole-body admittance controller in conjunction with a DCM observer and feedback controller to perform stair climbing with the HRP-4 robot in an Airbus factory. Caron (2019) investigated the applicability of a whole-body admittance control strategy in combination with the Variable-Height Inverted Pendulum (VHIP) model. A linear feedback controller is devised for the nonlinear VHIP model that, where feasible, acts similar to linear control of the DCM, and utilises a CoM height variation strategy otherwise (when the ZMP reaches the edge of the support polygon). Henze et al. (2016) developed a passivity-based whole-body controller in purely Cartesian formulation that was used, amongst other things, for multi-contact balance control. The controller was later extended by Mesesan et al. (2019) to a generalised tasks formulation, and applied to walking of the TORO robot over compliant mattresses and pebble beds. Walking on grass and flat terrain with edge contacts was also demonstrated, albeit with relatively static and long step times of 1.2–1.7 s.

## 8.3 ROBOCUP WALKING APPROACHES

The RoboCup Humanoid Soccer competition is an environment where the analytical or academic beauty of a method is completely second to its proven robustness and functionality on the field. Interestingly, there are very few examples of successful RoboCup teams that use walking approaches similar to any of the finely crafted methods described above.[1] There could be many reasons for this, ranging from time and resources of the teams, to the necessity of having a method that can be retuned on short notice before any given soccer match, but one significant factor is likely the robot hardware. Most robots participating in the RoboCup Humanoid League do not even have the quality of actuators and sensors that would be required to reliably track a planned CoM or DCM trajectory, let alone with much higher step frequencies, highly dynamic changes of walking plans,

---

1 With notable exception, for example, of Team THORwIn (McGill et al., 2015).

moving obstacles, and completely unpredictable ground, ball and push disturbances.

Team Rhoban (Allali et al., 2019) won the KidSize league three years in a row using a walking engine `QuinticWalk` that extended their previous open source gait engine `IKWalk` (Rouxel et al., 2015) to use $5^{th}$ order polynomial splines, continuous gait command accelerations, and explicit double support phases (Allali et al., 2018).[2] The generated gait trajectories are executed open-loop, without consideration of any ZMP stability criteria or modelling of the dynamics, but a stabilisation module is used to defer commanding of the support transitions until it is deemed to be the right time based on the foot force sensors. This is used as a form of timing feedback to help avoid desynchronisation of the desired and actual lateral oscillations of the robot.

The widely successful Standard Platform League (SPL) RoboCup team B-Human (Röfer et al., 2019) has in the last years used a gait based on the rUNSWift Nao gait (Hengst, 2014). A manually tuned open-loop gait is stabilised using ankle feedback in the pitch direction, and timing feedback in the lateral direction. The ankle feedback is implemented in the form of pitch offsets that are proportional to low-pass filtered gyroscope measurements. The timing feedback is implemented by estimating the lateral location of the CoP using foot force sensors, and adjusting the start of the leg lifting profile to always occur right before the CoP crosses the zero line. During gameplay, an online learning scheme monitors the peaks in the observed gyroscope measurements and uses this to successively guess and refine suitable gains for the ankle pitch feedback scheme.

The SPL RoboCup team Nao Devils (Hofmann et al., 2018) uses a walk engine based directly on the original preview control gait by Kajita et al. (2003). One slight modification to the original method is the introduction of an additional cart-spring-damper to the model, forming the so-called Flexible Linear Inverted Pendulum Model (Urbann et al., 2015). Although original walking results were qualitatively moderate,[3] the method was successively refined over following years to improve performance. For instance, the delay in responsiveness of walking intent induced by the 1 s preview window was reduced by resetting the ZMP generation after each step (previously this was done only when tracking errors exceeded a configured threshold). A Proportional-Derivative (PD) controller controlling all leg joints except the hip yaw based on the gyroscope and angle sensors was also added.

Team THORwIn of the AdultSize league (formally Team DARwIn of the KidSize league) were widely successful in both leagues from the years 2011 to 2015 (McGill et al., 2015). Originally using the gait developed for the DARwIn-OP (see Yi et al., 2011 on page 258), for the THOR robot they extended it to a hybrid locomotion controller that can switch online between two different walking controllers— a ZMP preview controller using linear quadratic optimisation and a simpler ZMP controller based on a closed-form solution to the LIPM equations (Yi et al., 2013). Smaller push disturbances were dealt with through reactive ankle pitch offsets and a mode whereby the robot stops walking and lowers its CoM to dampen out oscillations.

---

2 The dominant team of the TeenSize and AdultSize leagues in recent years was Team NimbRo, which used exactly the approaches to walking presented in this thesis.

3 Video published alongside Urbann et al. (2015): https://youtu.be/q3byPqjQ5a0

Although it has only seen isolated uses at RoboCup competitions, a distinct and impressive approach to balanced walking has been developed by Missura of team NimbRo in the form of the capture step framework (Missura, 2015). This approach adjusts both the step position and timing to preserve balance, based on the prediction of the CoM trajectory using the LIPM. The main advantage of this method is that it does not require forces, torques or the ZMP to be measured, making it more suitable for low-cost robots. More details of this method, including a discussion of both its features and shortcomings, can be found in Chapter 12.

## 8.4  DISCUSSION

It is a general problem of most ZMP preview control, MPC, capture point and DCM-based gaits that good actuator tracking performance and sensor feedback is required to make it work. This limits the applicability of such approaches to high quality hardware, and/or smaller robots that have favourable torque-to-weight ratios, especially if this is in combination with proportionally large feet (e.g. the Nao robot). Many state of the art approaches also require CoP/ZMP measurement through integrated force-torque sensors as well as precise physical models to make the planned motions feasible. These criteria are often not met when using lower cost robots such as the igus Humanoid Open Platform, in part due to lacking sensors, but mainly due to the limited robot stiffness and actuation quality.

With respect to the state of the art in bipedal walking, the following generalised observations can be made:

- Only select few methods can deal with true dynamic and real-time online adjustments of step timing for the purpose of balance. This is a problem, as in the face of even small disturbances lateral oscillations can lead to significant falls.[4]

- Many methods are evaluated using experiments that only perform a few forwards steps at a time (possibly over uneven terrain or stairs) and then stop. While this demonstrates that locomotion is fundamentally possible with the method, it is difficult to ascertain how robust the method truly is, and whether it would for example be suitable for real-life dynamically disturbed omni-directional walking over the full duration of a 10 minute soccer half-time (e.g. as required for RoboCup).

---

4  Good demonstration: https://youtu.be/l9uvBD9zmsw

- Many state of the art methods that rely on tracking of generated motion plans take relatively slow (but large) steps (>1 s per step), as the motion of the robot generally needs to be quite controlled for the tracking to work well.

- Publications frequently show experiments in simulation only, with real-robot experiments lagging behind or only possible with great subsequent effort and adaptation. Englsberger et al. (2013) and the TORO robot is an example of this, with real-robot TORO walking experiments only being published five years later in Englsberger et al. (2018).

By contrast, the methods for walking presented in this thesis are:

- Targeted at low-cost robots with an Inertial Measurement Unit (IMU) and joint encoders, but no options for force sensing and/or meaningful CoP/ZMP localisation,

- Targeted at robots with position-controlled actuators (limiting the compliance and controllability of the interactions the robot can have with the ground),

- Equipped with step timing feedback in combination with many other stabilising mechanisms to ensure preservation of balance,

- Based on the idea of stabilising a core semi-stable open-loop gait with the use of feedback mechanisms,

- Highly dynamic,[5] in that they allow the robots to go to their very limits of balance while still trying to recapture control, and,

- Aimed at allowing continuous walking for long periods of time (more than 10 minutes), even in the face of significant unknown and uncategorised disturbances.

---

5 As opposed to the frequently observed 'quasi-static' walking style seen on more expensive platforms.

# 9

## HUMANOID KINEMATICS

Before we can present the state estimation (Chapter 10) and subsequent approaches to bipedal walking (Chapter 11 and onwards), we must first develop standardised kinematic models of the robot that allow poses of the robot to be quantified and converted, to and from the joint space and task space. In addition to the standard joint and inverse pose spaces, two new pose spaces are introduced in this chapter—the abstract space and tip space—to allow better and easier planning of motion trajectories. Conversion algorithms between all four of these pose spaces are provided and discussed in terms of how they deal with singularities and cases of multiple possible solutions. As a definitive reference, an implementation of all of the pose spaces and conversion algorithms presented in this chapter can be found in the `humanoid_kinematics` package, released as part of the Humanoid Open Platform ROS Software (available here[1]).

## 9.1 ROBOT KINEMATICS POSE SPACES

The four different kinematics pose spaces that are used in this thesis can be evenly divided into two categories, namely

A) The angular space representations, given by the joint space and the abstract space, and,

B) The task space representations, given by the inverse space and the tip space.

Each of these pose spaces is detailed in a corresponding subsection below. Where relevant, note that by convention the x-axis is always chosen to point 'forwards', the y-axis is always chosen to point 'leftwards', and the z-axis is always chosen to point 'up'.

### 9.1.1 Joint Space

The joint space representation $\mathbf{q}$ of the pose of a robot is given by the vector of all scalar joint angles $\theta_*$ that are present in the robot. Figure 9.1 shows the assumed layout of all of these joints for the arms and legs of the robot. This layout is referred to as the joint kinematic model of the robot. Various definitions of coordinate frames (see Table 9.1), lengths and offsets (see Table 9.2), and joint axes $\hat{\mathbf{a}}_*$ form part of the model, including in particular the hip offsets $h_x$, $h_y$ and foot offsets $f_x$, $f_y$, $f_z$. All joints by convention have their counterclockwise (CCW) direction of rotation about their corresponding x, y or z-axis of rotation considered as positive, and their zero positions are as defined in Section 4.1.1 (and as illustrated in Figure 9.1). Note that the upper and lower links in each respective limb are assumed to be of equal length. This kinematic property is common in robot designs, and is true for all robots used in this thesis.

---

1 https://github.com/AIS-Bonn/humanoid_op_ros/tree/master/src/nimbro/hardware/humanoid_kinematics

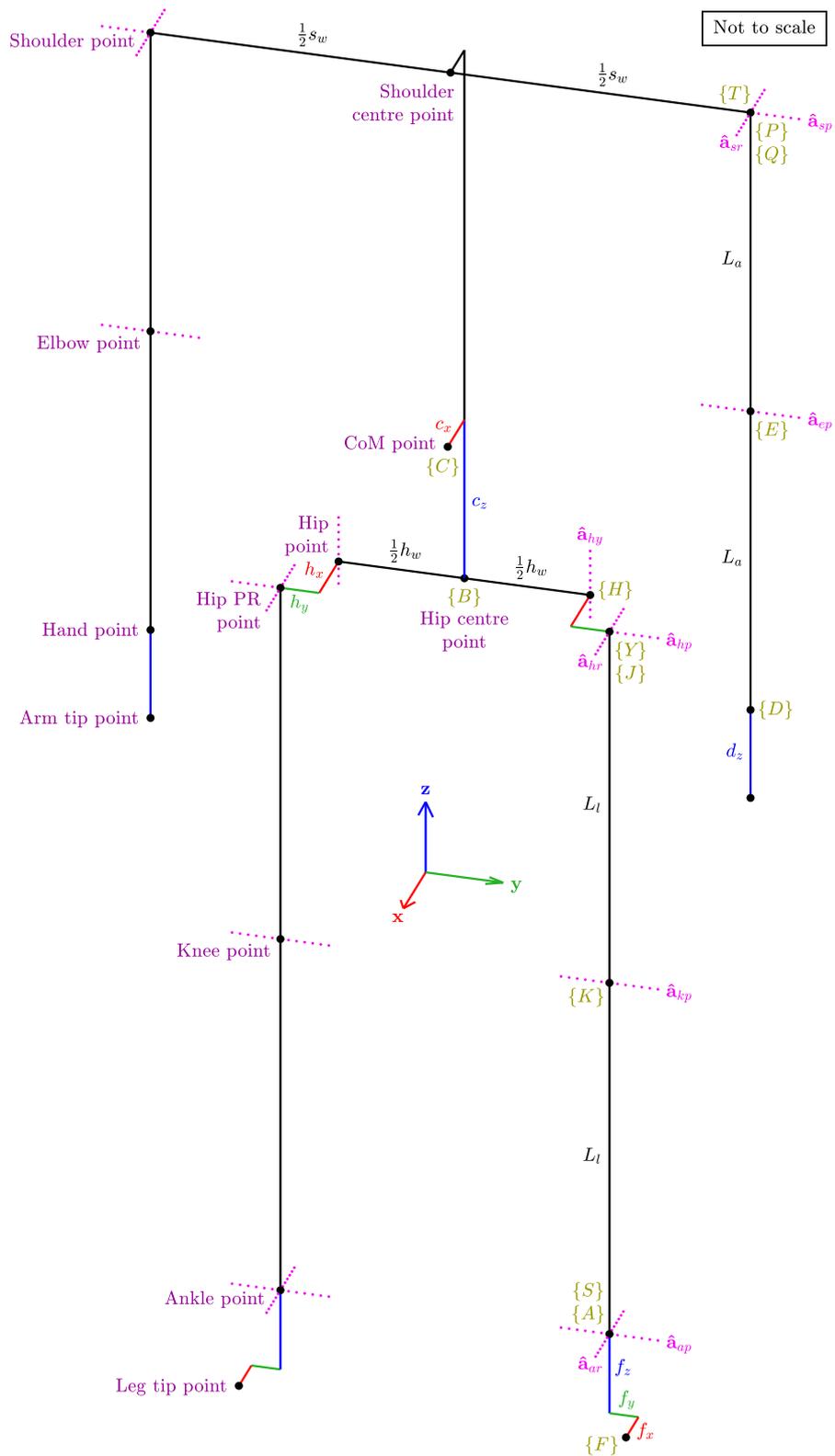

Figure 9.1: Definition of the joint kinematic model, i.e. the joint space pose representation, where the robot is depicted as facing out of the page. The definitions of the various kinematic coordinate frames {∗} can be seen, in addition to the axes of rotation $\hat{\mathbf{a}}_*$, the named points shown in purple, and the modelled link lengths and offsets.

Table 9.1: Joint space coordinate frames of the robot

| Frame | Location | Aligned with |
|-------|----------|-------------|
| {T} | Shoulder point | Trunk frame |
| {P} | Shoulder point | {T} rotated by $\theta_{sp}$ about $\hat{\mathbf{a}}_{sp}$ |
| {Q} | Shoulder point | {P} rotated by $\theta_{sr}$ about $\hat{\mathbf{a}}_{sr}$ |
| {E} | Elbow point | {P} rotated by $\theta_{sr}$ about $\hat{\mathbf{a}}_{sr}$ |
| {D} | Hand point | {E} rotated by $\theta_{ep}$ about $\hat{\mathbf{a}}_{ep}$ |
| {C} | CoM point | Trunk frame |
| {B} | Hip centre point | Trunk frame |
| {H} | Hip point | Trunk frame |
| {Y} | Hip PR point | {H} rotated by $\theta_{hy}$ about $\hat{\mathbf{a}}_{hy}$ |
| {J} | Hip PR point | {Y} rotated by $\theta_{hr}$ about $\hat{\mathbf{a}}_{hr}$ |
| {K} | Knee point | {J} rotated by $\theta_{hp}$ about $\hat{\mathbf{a}}_{hp}$ |
| {S} | Ankle point | {K} rotated by $\theta_{kp}$ about $\hat{\mathbf{a}}_{kp}$ |
| {A} | Ankle point | {S} rotated by $\theta_{ap}$ about $\hat{\mathbf{a}}_{ap}$ |
| {F} | Leg tip point | {A} rotated by $\theta_{ar}$ about $\hat{\mathbf{a}}_{ar}$ |

Table 9.2: Kinematic lengths and offsets of the robot

| Name | Sym. | Definition |
|------|------|-----------|
| Arm link length | $L_a$ | Length of the upper and lower arm links |
| Shoulder width | $s_w$ | Distance between the shoulder points |
| Hand z-offset | $d_z$ | Downward z-offset from the hand point to the arm tip point |
| Leg link length | $L_l$ | Length of the upper and lower leg links |
| Hip width | $h_w$ | Distance between the hip points |
| CoM x-offset | $c_x$ | Forward x-offset from the hip centre point to the assumed fixed CoM point |
| CoM z-offset | $c_z$ | Upward z-offset from the hip centre point to the assumed fixed CoM point |
| Hip x-offset | $h_x$ | Forward x-offset from the hip point to the hip PR point |
| Hip y-offset | $h_y$ | Outward y-offset from the hip point to the hip PR point |
| Foot x-offset | $f_x$ | Forward x-offset from the ankle point to the leg tip point |
| Foot y-offset | $f_y$ | Outward y-offset from the ankle point to the leg tip point |
| Foot z-offset | $f_z$ | Downward z-offset from the ankle point to the leg tip point |

As can be seen from Figure 9.1 and Table 9.1, for each arm and leg the assumed order of joint rotations is given by

$$
\begin{aligned}
\text{Shoulder pitch} &\Rightarrow \text{Rotate by } \theta_{sp} \text{ about the y-axis } \hat{\mathbf{a}}_{sp} = \mathbf{y}_T \\
\text{Shoulder roll} &\Rightarrow \text{Rotate by } \theta_{sr} \text{ about the x-axis } \hat{\mathbf{a}}_{sr} = \mathbf{x}_P \\
\text{Elbow pitch} &\Rightarrow \text{Rotate by } \theta_{ep} \text{ about the y-axis } \hat{\mathbf{a}}_{ep} = \mathbf{y}_E
\end{aligned}
$$

$$
\begin{aligned}
\text{Hip yaw} &\Rightarrow \text{Rotate by } \theta_{hy} \text{ about the z-axis } \hat{\mathbf{a}}_{hy} = \mathbf{z}_H \\
\text{Hip roll} &\Rightarrow \text{Rotate by } \theta_{hr} \text{ about the x-axis } \hat{\mathbf{a}}_{hr} = \mathbf{x}_Y \\
\text{Hip pitch} &\Rightarrow \text{Rotate by } \theta_{hp} \text{ about the y-axis } \hat{\mathbf{a}}_{hp} = \mathbf{y}_J \\
\text{Knee pitch} &\Rightarrow \text{Rotate by } \theta_{kp} \text{ about the y-axis } \hat{\mathbf{a}}_{kp} = \mathbf{y}_K \\
\text{Ankle pitch} &\Rightarrow \text{Rotate by } \theta_{ap} \text{ about the y-axis } \hat{\mathbf{a}}_{ap} = \mathbf{y}_S \\
\text{Ankle roll} &\Rightarrow \text{Rotate by } \theta_{ar} \text{ about the x-axis } \hat{\mathbf{a}}_{ar} = \mathbf{x}_A
\end{aligned}
$$

where each rotation is local to the frame that results from the previous rotation. It can also be seen from Figure 9.1 that the joint kinematic model defines a multitude of named points (in purple) throughout the entire kinematics of the robot. These named points are useful to have (and reference), when developing algorithms related to motions of the limbs, and include for example the hip centre point, hip points, hip PR points, ankle points and leg tip points. The leg tip points are a fixed point on the bottom of each foot that approximate the idealised point of contact between that foot and the ground. Essentially, differences in foot height and the sizes of steps taken are assumed to be completely characterised by the relative motions of these points. In the case of the robots used in this thesis, the leg tip points are specified to be the geometric centre of the contact patch of each foot, as the contact patch is perfectly rectangular for each of the robots.

### 9.1.2 Abstract Space

The abstract space is a pose representation that was specifically developed for humanoid robots in the context of walking and balancing. It is similar to the joint space in that its description of the robot pose consists predominantly of scalar angular values, but it is different in that it focuses on the orientation and length of so-called arm and leg centre lines instead of individual joint positions. Examples of such limb centre lines for the arms and legs of the robot are shown in Figure 9.2. Note that no matter what the pose of the robot is, the arm centre line is always the line joining the shoulder point to the hand point, and the leg centre line is always the line joining the hip PR point to the ankle point. In each case, it can be observed that the length of the centre line is a direct measure of the amount of rotation in the corresponding elbow or knee.

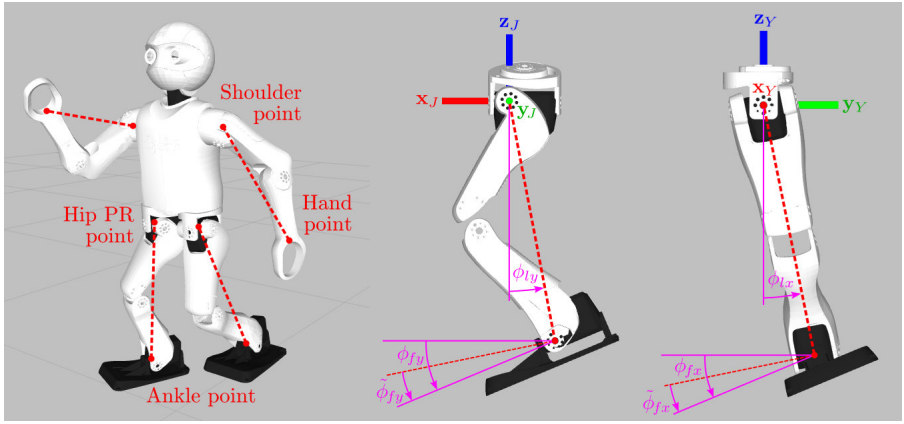

Figure 9.2: Limb centre lines (left) and their geometric interpretation for both the arms and the legs. The right two images show the geometric interpretation of the leg angles $(\phi_{lx}, \phi_{ly})$, foot angles $(\phi_{fx}, \phi_{fy})$, and ankle angles $(\tilde{\phi}_{fx}, \tilde{\phi}_{fy})$, where it should be noted that the legs are shown relative to frames {J} and {Y} respectively. Note also that the right-most image shows a pose where $\phi_{fy} = 0$, as otherwise the leg and foot rotations would not be in the same plane, and thereby become harder to visualise.

Figure 9.3 shows the full abstract kinematic model of the arms and legs of the robot. It should be noted that the defined named points, link offsets (see Table 9.2), and coordinate frames (see Table 9.3) are consistent with the joint kinematic model depicted in Figure 9.1. Given the arm link length $L_a$ and the leg link length $L_l$ (see Table 9.2), the assumed order of abstract space rotations and translations is given by

Arm angle Y $\Rightarrow$ Rotate by $\phi_{ay}$ about the y-axis $\hat{\mathbf{a}}_{ay} = \mathbf{y}_T$

Arm angle X $\Rightarrow$ Rotate by $\phi_{ax}$ about the x-axis $\hat{\mathbf{a}}_{ax} \equiv \hat{\mathbf{a}}_{sr} = \mathbf{x}_P$

Arm retraction $\Rightarrow$ Translate by $2L_a(1 - \epsilon_a)$ down the axis $\hat{\mathbf{a}}_{a\epsilon} = \mathbf{z}_N$

Leg angle Z $\Rightarrow$ Rotate by $\phi_{lz}$ about the z-axis $\hat{\mathbf{a}}_{lz} = \mathbf{z}_H$

Leg angle X $\Rightarrow$ Rotate by $\phi_{lx}$ about the x-axis $\hat{\mathbf{a}}_{lx} = \mathbf{x}_Y$

Leg angle Y $\Rightarrow$ Rotate by $\phi_{ly}$ about the y-axis $\hat{\mathbf{a}}_{ly} = \mathbf{y}_J$

Leg retraction $\Rightarrow$ Translate by $2L_l(1 - \epsilon_l)$ down the axis $\hat{\mathbf{a}}_{l\epsilon} = \mathbf{z}_L$

Foot angle Y $\Rightarrow$ Rotate by $\phi_{fy} - \phi_{ly}$ about the y-axis $\hat{\mathbf{a}}_{fy} = \mathbf{y}_L$

Foot angle X $\Rightarrow$ Rotate by $\phi_{fx} - \phi_{lx}$ about the x-axis $\hat{\mathbf{a}}_{fx} = \mathbf{x}_A$

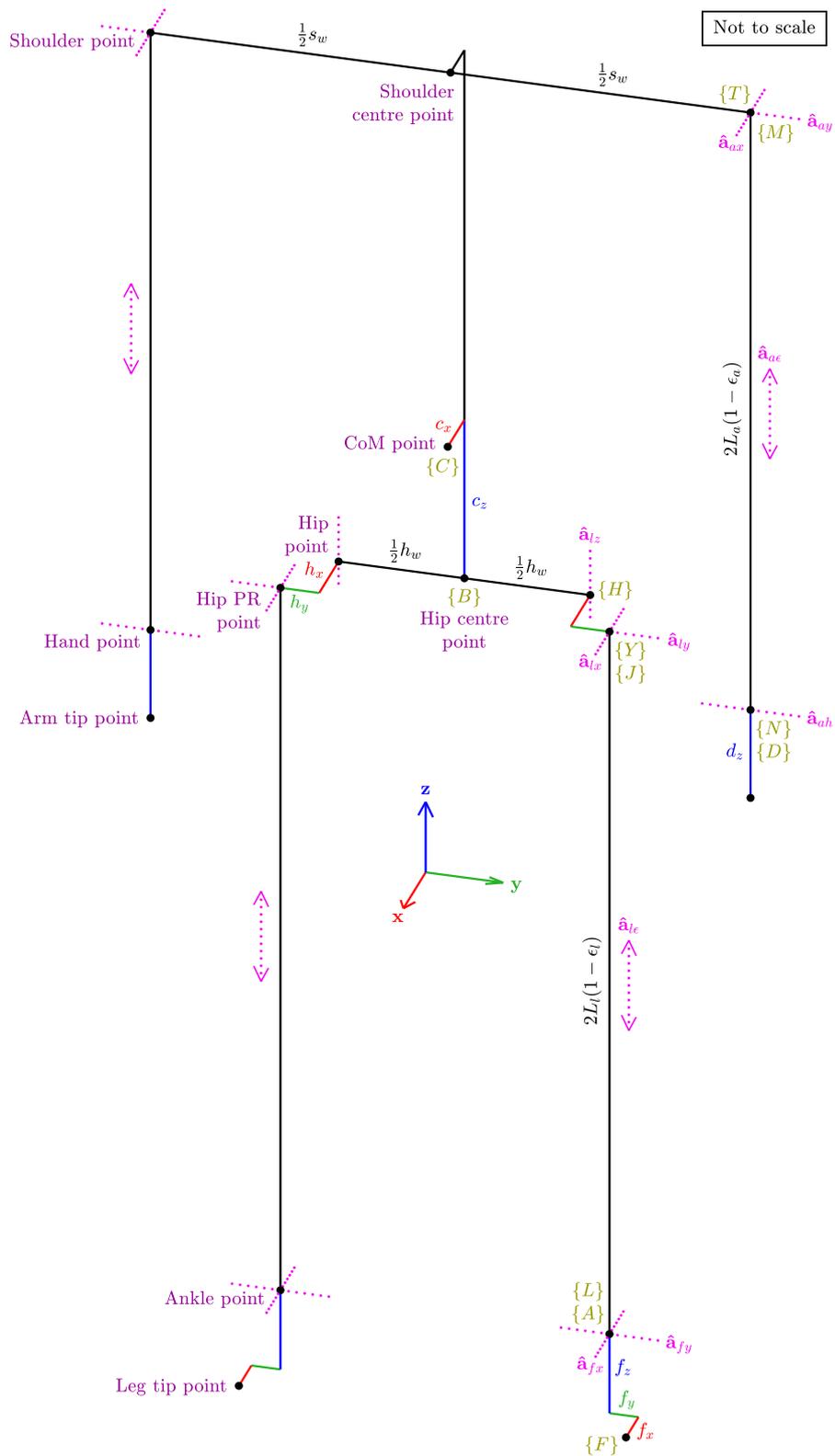

Figure 9.3: Definition of the abstract kinematic model, i.e. the abstract space representation, where the robot is depicted as facing out of the page. The definitions of the various kinematic coordinate frames {∗} can be seen, in addition to the axes of rotation $\hat{\mathbf{a}}_*$, the named points shown in purple, and the modelled link lengths and offsets.

Table 9.3: Abstract space coordinate frames of the robot

| Frame | Location | Aligned with |
|-------|----------|--------------|
| {T} | Shoulder point | Trunk frame |
| {M} | Shoulder point | {T} rotated by $\phi_{ay}$ about $\hat{\mathbf{a}}_{ay}$ |
| {N} | Hand point | {M} rotated by $\phi_{ax}$ about $\hat{\mathbf{a}}_{ax} \equiv \hat{\mathbf{a}}_{sr}$, and translated $2L_a(1-\epsilon_a)$ along $\hat{\mathbf{a}}_{a\epsilon}$ |
| {D} | Hand point | {N} rotated by $-\alpha_a$ about $\hat{\mathbf{a}}_{ah}$, See Equation (9.7a) for $\alpha_a$ |
| {C} | CoM point | Trunk frame |
| {B} | Hip centre point | Trunk frame |
| {H} | Hip point | Trunk frame |
| {Y} | Hip PR point | {H} rotated by $\phi_{lz}$ about $\hat{\mathbf{a}}_{lz}$ |
| {J} | Hip PR point | {Y} rotated by $\phi_{lx}$ about $\hat{\mathbf{a}}_{lx}$ |
| {L} | Ankle point | {J} rotated by $\phi_{ly}$ about $\hat{\mathbf{a}}_{ly}$, and translated $2L_l(1-\epsilon_l)$ along $\hat{\mathbf{a}}_{l\epsilon}$ |
| {A} | Ankle point | {L} rotated by $\phi_{fy} - \phi_{ly}$ about $\hat{\mathbf{a}}_{fy}$ |
| {F} | Leg tip point | {A} rotated by $\phi_{fx} - \phi_{lx}$ about $\hat{\mathbf{a}}_{fx}$ |

where each transformation is local to the frame that results from the one before. The complete abstract space pose for an arm is then given by the vector

$$\boldsymbol{\Phi}_a = (\phi_{ax}, \phi_{ay}, \epsilon_a), \tag{9.1}$$

and for a leg is given by

$$\boldsymbol{\Phi}_l = (\phi_{lx}, \phi_{ly}, \phi_{lz}, \phi_{fx}, \phi_{fy}, \epsilon_l). \tag{9.2}$$

The combination of two abstract arm poses $\boldsymbol{\Phi}_a$ and abstract leg poses $\boldsymbol{\Phi}_l$ in a single vector yields the abstract pose $\boldsymbol{\Phi}$ of the entire robot, just like $\mathbf{q}$ is the joint pose vector of the entire robot. Note that the parameters $\phi_{fx}$ and $\phi_{fy}$ are called the foot angle X and foot angle Y respectively, and that the (infrequently used) dependent parameters

$$\tilde{\phi}_{fx} = \phi_{fx} - \phi_{lx}, \tag{9.3a}$$

$$\tilde{\phi}_{fy} = \phi_{fy} - \phi_{ly}, \tag{9.3b}$$

are referred to as the ankle angle X and ankle angle Y.[2] The geometric interpretations of the leg angles $\phi_{l*}$, foot angles $\phi_{f*}$, and ankle angles $\tilde{\phi}_{f*}$, and how they are related to each other, are elucidated in the right image in Figure 9.2.

---

2 As these angles are numerically equivalent to how much rotation occurs in the ankles in the abstract kinematic model.

The parameters $\epsilon_a, \epsilon_l \in [0,1]$ are referred to as the arm and leg retraction parameters respectively, and are equal to 0 when the corresponding limb is fully extended, and 1 when it is fully retracted (i.e. when the respective knee or elbow pitch is 180°). The retraction parameter is effectively the distance that a limb is shorter than its fully extended length, expressed as a ratio of the fully extended length. By definition, $\epsilon_a = \epsilon_l = 0$ for the zero pose, as expected.

The abstract pose parameters can easily be converted to and from their corresponding joint pose parameters. The conversion from the joint space to the abstract space for the arms is given by

$$\alpha_a = \tfrac{1}{2}\operatorname{wrap}(-\theta_{ep}), \qquad \phi_{ax} = \theta_{sr}, \qquad (9.4a)$$

$$\epsilon_a = 1 - \cos\alpha_a, \qquad \phi_{ay} = \theta_{sp} - \alpha_a, \qquad (9.4b)$$

where $\operatorname{wrap}(\cdot)$ is a function that wraps an angle to the range $(-\pi, \pi]$ by multiples of $2\pi$. For the legs, the same conversion is given by

$$\alpha_l = \tfrac{1}{2}\operatorname{wrap}(\theta_{kp}), \qquad (9.5)$$

followed by

$$\phi_{lx} = \theta_{hr}, \qquad \phi_{fx} = \theta_{ar} + \theta_{hr}, \qquad (9.6a)$$

$$\phi_{ly} = \theta_{hp} + \alpha_l, \qquad \phi_{fy} = \theta_{ap} + \theta_{hp} + 2\alpha_l, \qquad (9.6b)$$

$$\phi_{lz} = \theta_{hy}, \qquad \epsilon_l = 1 - \cos\alpha_l. \qquad (9.6c)$$

The reverse conversion from the abstract space back to the joint space is given for the arms by

$$\alpha_a = \operatorname{acos}(1 - \epsilon_a), \qquad \theta_{sp} = \phi_{ay} + \alpha_a, \qquad (9.7a)$$

$$\theta_{ep} = -2\alpha_a, \qquad \theta_{sr} = \phi_{ax}. \qquad (9.7b)$$

For the legs, the same conversion is given by

$$\alpha_l = \operatorname{acos}(1 - \epsilon_l), \qquad (9.8)$$

followed by

$$\theta_{hy} = \phi_{lz}, \qquad \theta_{kp} = 2\alpha_l, \qquad (9.9a)$$

$$\theta_{hr} = \phi_{lx}, \qquad \theta_{ap} = \phi_{fy} - \phi_{ly} - \alpha_l, \qquad (9.9b)$$

$$\theta_{hp} = \phi_{ly} - \alpha_l, \qquad \theta_{ar} = \phi_{fx} - \phi_{lx}. \qquad (9.9c)$$

### 9.1.3  Inverse Space

When designing motion trajectories for the limbs of a robot, it is useful to be able to interpret and set the poses of the robot via the 3D position and orientation of the corresponding end effectors, i.e. hand and foot links. The inverse space provides this functionality, and allows the configuration of the robot to be viewed from a task space perspective. The pose of each leg is represented by a tuple

$$(\mathbf{p}_l, q_l) \in \mathbb{R}^3 \times \mathbb{Q}, \tag{9.10}$$

and the pose of each arm is represented by a tuple

$$(\mathbf{p}_a, q_a) \in \mathbb{R}^3 \times \mathbb{Q}, \tag{9.11}$$

where

  $\mathbf{p}_l \Rightarrow$ 3D position vector of the ankle point in the body-fixed frame {B}, relative to its position in the zero pose

  $\mathbf{p}_a \Rightarrow$ 3D position vector of the hand point in the body-fixed frame {B}, relative to its position in the zero pose

  $q_l \Rightarrow$ Quaternion orientation of the foot-fixed frame {F} relative to the body-fixed frame {B}, i.e. ${}_F^B q$

  $q_a \Rightarrow$ Quaternion orientation of the hand-fixed frame {D} relative to the body-fixed frame {B}, i.e. ${}_D^B q$

Clearly, by definition, when the robot is standing upright and straight with all of its joint angles equal to zero, the inverse space position $\mathbf{p}_*$ of all four limbs is $(0, 0, 0)$, and the inverse space orientation $q_*$ in each case is the identity rotation $(1, 0, 0, 0)$.

As the robots used in this thesis only have 3 Degrees of Freedom (DoFs) in each arm (pitch → roll → pitch), it turns out that the hand position $\mathbf{p}_a$ and hand orientation $q_a$ are in general completely dependent parameters. That is, for essentially all hand positions $\mathbf{p}_a$, there is only one unique orientation $q_a$ with which the hand can be positioned at $\mathbf{p}_a$ (within the constraints of the kinematics). As such, the inverse arm pose is often truncated to consist of just the hand position $\mathbf{p}_a$, instead of the full tuple $(\mathbf{p}_a, q_a)$. The inverse leg pose $(\mathbf{p}_l, q_l)$ stays as it is however, as the leg kinematics does indeed have the full 6 DoF required to ensure that in general all ankle positions $\mathbf{p}_l$ can be reached with all foot orientations $q_l$.

While the joint and abstract spaces are dimensionless in the sense that the same values applied to identically shaped robots of different sizes produce exactly the same pose, this is not true for the inverse space. If a robot that is 30 cm tall moves its foot by 10 cm, this is a completely different pose than if a robot that is 180 cm moves its foot by the same amount. To combat this dimensionality of inverse leg poses, the inverse leg scale parameter $L_i$ is introduced, given by

$$L_i = 2L_l. \tag{9.12}$$

$L_i$ is a fixed constant for each robot, and corresponds to the vertical distance from the hip point to the ankle point in the zero pose. Expressing ankle positions $\mathbf{p}_l$ in units of $L_l$ is one way of making the inverse leg poses as dimensionless as the joint and abstract poses. This allows poses expressed for one robot to more easily be used on other robots, even if the other robots are of different size.

### 9.1.4  Tip Space

The tip space is similar to the inverse space, with the only difference being that instead of quantifying the positions of the hand and ankle points, the positions of the arm and leg tip points are quantified (see Figure 9.1). The arm and leg tip poses are given by the tuples

$$(\mathbf{t}_a, q_a) \in \mathbb{R}^3 \times \mathbb{Q}, \tag{9.13a}$$

$$(\mathbf{t}_l, q_l) \in \mathbb{R}^3 \times \mathbb{Q}, \tag{9.13b}$$

where $q_a$ and $q_l$ are the same quaternion orientations as before, and

$\mathbf{t}_a \Rightarrow$ 3D position vector of the arm tip point in the body-fixed frame {B}, relative to its position in the zero pose

$\mathbf{t}_l \Rightarrow$ 3D position vector of the leg tip point in the body-fixed frame {B}, relative to its position in the zero pose

Like for the inverse pose, the arm tip pose is often truncated to consist of only the arm tip position $\mathbf{t}_a$, and in the zero pose both $\mathbf{t}_a$ and $\mathbf{t}_l$ are by definition equal to $(0, 0, 0)$. A leg tip scale parameter $L_t$ is also introduced, given by

$$L_t = 2L_l + f_z, \tag{9.14}$$

to allow leg tip poses to be expressed in a dimensionless manner (see Table 9.2 for the definition of $f_z$). Numerically, $L_t$ corresponds to the vertical distance from the hip point to the leg tip point when the robot is in the zero pose.

The conversions between the inverse and tip spaces are relatively trivial, and can be summarised by the equations

$$\mathbf{t}_a = \mathbf{p}_a + L_{q_a}(\mathbf{d}), \qquad \mathbf{t}_l = \mathbf{p}_l + L_{q_l}(\mathbf{f}), \tag{9.15a}$$

$$\mathbf{p}_a = \mathbf{t}_a - L_{q_a}(\mathbf{d}), \qquad \mathbf{p}_l = \mathbf{t}_l - L_{q_l}(\mathbf{f}), \tag{9.15b}$$

where

$$\mathbf{d} = (0, 0, -d_z), \tag{9.16a}$$

$$\mathbf{f} = (f_x, \delta f_y, -f_z), \tag{9.16b}$$

are respectively the hand offset vector and foot offset vector. It should be recalled from Equation (7.49) on page 190 that $L_q(\mathbf{v})$ is the notation for the vector that results when a vector $\mathbf{v}$ is rotated by the quaternion $q \in \mathbb{Q}$. Note that the limb sign variable

$$\delta = \begin{cases} 1 & \text{if left limb,} \\ -1 & \text{if right limb,} \end{cases} \tag{9.17}$$

was used in Equation (9.16b) to ensure that the offset $f_y$ is applied in the correct outwards direction for each leg.

As mentioned previously, the tip space is of particular interest for the legs, because the leg tip points model the idealised point of contact between the foot and the ground. Thus, when task space trajectories are planned for the feet, especially in the light of soft ground contacts and regularly changing support conditions, the tip space is used.

## 9.2 KINEMATIC CONVERSIONS BETWEEN SPACES

Given that four different spaces are used to represent the poses of the robot, it is useful to be able to convert between all of them. The conversions (respectively) between the joint space and abstract space, and inverse space and tip space, have already been presented in the previous section, so it only remains to provide conversions between the joint and inverse spaces.

### 9.2.1 Forward Kinematics

The conversion from the joint space to the inverse space is known as the forward kinematics, and is relatively simple to perform by successively applying the required rotations and offsets, as dictated by the order of the kinematic chain. For instance, given the joint angles $(\theta_{sp}, \theta_{sr}, \theta_{ep})$ representing the pose of an arm, the corresponding inverse pose $(\mathbf{p}_a, q_a)$ can be calculated using

$$q_a = {}^T_E q \, q_y(\theta_{ep}), \tag{9.18a}$$

$$\mathbf{p}_a = (0, 0, 2L_a) - L_{{}^T_E q}\big((0, 0, L_a)\big) - L_{q_a}\big((0, 0, L_a)\big), \tag{9.18b}$$

where

$$ {}^T_E q = q_y(\theta_{sp}) \, q_x(\theta_{sr}), \tag{9.19}$$

is the total amount of rotation that occurs in the shoulder, and $q_y(\theta_{sp})$, for instance, is the quaternion corresponding to a y-rotation by $\theta_{sp}$ radians. The forward kinematics of a leg can be computed in a similar manner, but clearly requires more steps due to the larger number of joints and offsets in the corresponding kinematic chain.

### 9.2.2 Leg Inverse Kinematics

The conversion of a limb pose from the inverse space to the joint space is known as the inverse kinematics. The inverse kinematics of a leg is significantly more complex than its forward kinematics, as it requires the simultaneous solution of multiple trigonometric equations, and is made especially complex by the inclusion of the hip offsets $h_x$ and $h_y$ in the joint kinematic model (see Figure 9.1). The hip offsets make the trigonometric equations more difficult to solve, and result in greater problems with inverse poses having multiple possible joint space solutions. The offsets cannot simply be omitted, because this would sacrifice the generality of the model, and robots such as the igus Humanoid Open Platform have a significantly non-zero $h_x$ component that would then go unmodelled.

Given an arbitrary inverse leg pose $(\mathbf{p}_l, q_l)$, we start by computing the two possible solutions for the hip yaw $\theta_{hy}$. Recalling that {F} is the frame fixed to the foot, and that by definition

$$q_l \equiv {}^B_F q, \tag{9.20}$$

we can compute (from $q_l$) the unit vector

$$^B\mathbf{x}_F = (x_{Fx}, x_{Fy}, x_{Fz}), \tag{9.21}$$

that corresponds to the x-axis of {F} with respect to the body-fixed frame {B}. Then, given the vector

$$^H\mathbf{p}_A = \mathbf{p}_l + (h_x, \delta h_y, -2L_l), \tag{9.22}$$

which corresponds to the position of the ankle point relative to the hip point, we compute the dot products

$$B = {}^H\mathbf{p}_A \bullet (0, -x_{Fz}, x_{Fy}), \tag{9.23a}$$

$$C = {}^H\mathbf{p}_A \bullet (-x_{Fz}, 0, x_{Fx}). \tag{9.23b}$$

Note that the limb sign variable $\delta \in \{-1, 1\}$ from Equation (9.17) was used in Equation (9.22) to ensure that the offset $h_y$ is applied in the correct outwards direction for each leg. Based on Equation (9.23), the two possible solutions for the hip yaw $\theta_{hy}$ can be calculated to be

$$\theta_{hy} = \text{wrap}(\xi + D), \tag{9.24a}$$

$$\text{and} \quad \theta_{hy} = \text{wrap}(\xi + \pi - D), \tag{9.24b}$$

where

$$\xi = \text{atan2}(B, C), \tag{9.25a}$$

$$A = \sqrt{B^2 + C^2}, \tag{9.25b}$$

$$D = \text{asin}\left(\frac{x_{Fz}\delta h_y}{A}\right). \tag{9.25c}$$

If a non-zero hip y-offset ($h_y$) is present, and $|x_{Fz}h_y| > A$, it can be observed from Equation (9.25c) that no suitable value of $D$ can be calculated. This is a result of the inverse kinematics problem not having any solutions in this case, and can be remedied by coercing the operand in Equation (9.25c) to the range $[-1, 1]$ before applying the arcsine function. This produces the closest feasible inverse kinematics solution to the desired input. It can further be observed that explicit singularities of the kinematics occur when $A = 0$, which is equivalent to ${}^H\mathbf{p}_A$ and ${}^B\mathbf{x}_F$ either being parallel, or both having a zero z-component.[3] At these singularities, if there are any solutions at all, there are infinitely many of them. One example of an $A = 0$ singularity is when the ankle point is in front of the robot at the same height as the hip point (${}^H p_{Az} = 0$), and the foot is horizontally flat relative to the torso ($x_{Fz} = 0$). In this case, it can be observed that the hip roll and ankle roll axes coincide, and can be rotated arbitrarily without affecting the position and orientation of the foot (only the knee moves). Singular poses where $A = 0$ are not normally encountered during the generation of any motions that revolve around upright standing postures of the robot, like walking for example, so sanity checking of the produced poses and coercion of the allowed z-height of the ankle, for instance, is sufficient to avoid any problems with the inverse kinematics.

Away from singularities, given the two possible solutions for the hip yaw $\theta_{hy}$ calculated using Equation (9.24), each solution can be used to generate a further two possible sets of solutions for the remaining five joint angles in the leg.[4] The procedure by which this is done is quite mathematically involved,[5] but works by first calculating the corresponding set of five abstract space parameters (as $\theta_{hy} \equiv \phi_{lz}$) for one solution, and then constructing the second solution using the transformation

$$\phi_{lx} \leftarrow \mathrm{wrap}(\pi + \phi_{lx}), \tag{9.26a}$$

$$\phi_{ly} \leftarrow \mathrm{wrap}(\pi - \phi_{ly}), \tag{9.26b}$$

$$\phi_{fy} \leftarrow -\phi_{fy}, \tag{9.26c}$$

where the foot angle X ($\phi_{fz}$) and leg retraction ($\epsilon_l$) parameters stay unchanged. Ultimately, this entire process leads to four distinct joint space solutions to the given inverse kinematics problem.

The scheme by which one of the four inverse kinematics joint space solutions is chosen is a critical component of ensuring that all reasonable inverse space trajectories are continuous when converted to the joint space. This is less trivial than it first seems, due to the highly varied nature of possible inverse space trajectories, and the observation that the continuity property sometimes requires inexact solutions to be chosen over available exact solutions (more details

---

3  In fact, $A = 0$ is equivalent to ${}^H\mathbf{p}_A \times {}^B\mathbf{x}_F$ having zero x and y-components, i.e. being a vector that is zero or purely in the z-direction. This can be seen to be equivalent to the two stated conditions.

4  Solutions where the knee bends 'backwards' are ignored, otherwise for each value of $\theta_{hy}$ there would be *four* different solutions for the remaining five parameters.

5  Refer to `InvKinCalcAbsPose()` in the following file for a code implementation: `https://github.com/AIS-Bonn/humanoid_op_ros/blob/master/src/nimbro/hardware/humanoid_kinematics/src/serial/serial_kinematics.cpp`

below). The following two costs are evaluated for each of the four possible joint space solutions:

$$C_\infty = \max\{|w_y\theta_{hy}|, |w_r\theta_{hr}|, |w_p\theta_{hp}|\}, \tag{9.27a}$$

$$C_1 = |w_y\theta_{hy}| + |w_r\theta_{hr}| + |w_p\theta_{hp}|, \tag{9.27b}$$

where $\theta_{hy}$, $\theta_{hr}$ and $\theta_{hp}$ are the hip yaw, roll and pitch respectively, and the weights $w_y$, $w_r$ and $w_p$ account for the fact that not all hip joints have the same ideal ranges of motion. As an example, for the igus Humanoid Open Platform, weights of $w_y = w_r = 1$ and $w_p = \frac{2}{3}$ were chosen. The joint space solutions are prioritised first by minimum $C_\infty$, then by minimum $C_1$ in cases of remaining equality. This promotes primarily joint pose configurations that stay within the joint limits, and are furthermore as 'close' as possible to the identity zero pose. The joint pose with the lowest such total cost is taken as the solution of the inverse kinematics, and is referred to as the *canonical solution*. Note that the canonical solution may not be an exact solution to the inverse kinematics if the workspace boundaries are reached by the input inverse pose. Inexact solutions are ones that do not strictly satisfy the inverse pose input, but are the closest feasible joint poses for that given $\theta_{hy}$. Importantly, inexactness is not a reason to discard a possible solution, as due to the modelled hip offsets it is possible that two of the four solutions are exact and two are inexact, but the exact solutions are completely infeasible due to joint limits, and would cause discontinuities. It is also important that only the hip angles contribute to the calculated cost, as otherwise larger rotations of just the foot can cause total flipping of the leg. By not including the ankle joint angles in the costs, the leg pose remains stable, and joint coercion robustly handles situations of excessive foot rotation.

### 9.2.3 Arm Inverse Kinematics

The inverse kinematics of the arms is much simpler than the inverse kinematics of the legs, as the number of degrees of freedom is lower. Given a desired hand position $\mathbf{p}_a$ (recall that the hand position is specified relative to the position of the hand point in the zero pose), the position of the hand point relative to the body-fixed shoulder frame {T} is given by

$$^T\mathbf{p}_D = \mathbf{p}_a - (0, 0, 2L_a), \tag{9.28}$$

where we recall from Table 9.2 that $L_a$ is the arm link length. If the magnitude of this position vector is larger than $2L_a$, the desired hand point is outside the reachable workspace of the arm, and the vector $^T\mathbf{p}_D$ needs to be rescaled so that it has a magnitude of exactly only $2L_a$. Then, the arm base angle $\alpha_a$—which corresponds to the positive equal base angles of the isosceles triangle formed by the upper arm, lower arm and arm centre line (see Figure 9.4)—is given by

$$\alpha_a = \text{acos}\left(\frac{\|{}^T\mathbf{p}_D\|}{2L_a}\right).$$ (9.29)

The required joint angles of the arm are then calculated using

$$\theta_{ep} = -2\alpha_a,$$ (9.30a)

$$\theta_{sr} = \text{asin}\left(\frac{2L_a {}^T p_{Dy}}{\|{}^T\mathbf{p}_D\|^2}\right),$$ (9.30b)

$$\theta_{sp} = \text{atan2}\left({}^T p_{Dx}, {}^T p_{Dz}\right) - \text{atan2}\left(\sin\alpha_a, -\cos\theta_{sr}\cos\alpha_a\right).$$ (9.30c)

The arm is relatively simple kinematically, at just 3 DoF, but exactly this level of 'underactuation' results in significant regions where the inverse kinematics has no solution. For a given arm base angle $\alpha_a$, which from Equations (9.4a) and (9.7a) can be seen to be equal to

$$\alpha_a = \tfrac{1}{2}\,\text{wrap}(-\theta_{ep}) = \text{acos}(1 - \epsilon_a),$$ (9.31)

there is no solution for placing the hand point at a Cartesian point that subtends an angle less than $\alpha_a$ at the shoulder point relative to the double-ended y-axis $\mathbf{y}_T$. For each $\alpha_a$, the set of all infeasible inverse arm poses (i.e. hand positions $\mathbf{p}_a$) is thus given by the interior of a double-ended cone of apex angle $2\alpha_a$ about the y-axis $\mathbf{y}_T$, centred at the shoulder point. Infeasible poses can be dealt with by coercing the operand of arcsine in Equation (9.30b) to $[-1, 1]$, or by automatically reducing the value of $\alpha_a$ to make the requested inverse pose feasible. In the latter case, the hand point is then at least in the desired direction relative to the shoulder point, but the arm will be more elongated than intended.

### 9.2.4 Generalised Arm Inverse Kinematics

Often, the primary purpose of the arms in a generated motion is to shift the centre of gravity of the robot, and/or to act as dynamic reaction masses. As such, in many situations it is more important that the Centre of Mass (CoM) of an arm is pointing in the right direction, than that the hand point is in some exact position. To cater for this way of specifying the required pose of an arm, a generalised inverse kinematics algorithm has been developed that takes as input

- The unit vector

$$^T\hat{\mathbf{r}} = (r_x, r_y, r_z),$$ (9.32)

  defining the trunk-fixed ray through the shoulder point on which some point of interest $K$ should be placed, where $K$ is defined locally relative to the pitch- and roll-rotated shoulder frame {Q} of the arm (see Table 9.1),

- The vector

$$^{Q}\mathbf{p}_{K}(\alpha_{a}) = \left(^{Q}p_{Kx},\ 0,\ ^{Q}p_{Kz}\right), \tag{9.33}$$

  specifying the point of interest $K$ relative to the local arm frame {Q}, as a function of the arm base angle $\alpha_a$,

- The nominal arm base angle $\alpha_{nom}$ to use for the pose, and,

- The maximum allowed shoulder roll $\theta_{srm}$ to use for the pose.

The maximum shoulder roll specification $\theta_{srm}$ has priority over the nominal arm base angle $\alpha_{nom}$, and if required, forces a selection of $\alpha_a$ that is less than $\alpha_{nom}$, in order for $\theta_{srm}$ to be satisfied. If the $\theta_{srm}$ specification still cannot be satisfied, at least the $\alpha_a$ that minimises the shoulder roll, i.e. $\alpha_a = 0$, should be selected.

If we define the unit vector $\hat{\mathbf{p}}(\alpha_a)$ as

$$\hat{\mathbf{p}}(\alpha_a) = \frac{^{Q}\mathbf{p}_{K}(\alpha_a)}{\|^{Q}\mathbf{p}_{K}(\alpha_a)\|} = \left(p_x(\alpha_a),\ 0,\ p_z(\alpha_a)\right), \tag{9.34}$$

i.e. the desired 'direction' of the point of interest $K$ relative to {Q}, the generalised inverse kinematics problem can be summarised as needing to find three joint angles $\theta_{sp}$, $\theta_{sr}$ and $\theta_{ep}$ such that

$$^{T}\hat{\mathbf{r}} = {_{Q}^{T}}R\,\hat{\mathbf{p}}(\alpha_a) \tag{9.35a}$$

$$= R_y(\theta_{sp})\,R_x(\theta_{sr})\,\hat{\mathbf{p}}(\alpha_a), \tag{9.35b}$$

subject to $|\theta_{sr}| \leq \theta_{srm}$, and nominally

$$\alpha_a = \tfrac{1}{2}\operatorname{wrap}(-\theta_{ep}) = \alpha_{nom}. \tag{9.36}$$

Note that $R_x(\theta_{sr})$ and $R_y(\theta_{sp})$ are the rotation matrices corresponding to CCW rotations by $\theta_{sr}$ and $\theta_{sp}$ about the x and y-axes respectively. Examples of possible definitions of the point of interest $K$, along with the corresponding definitions of $\hat{\mathbf{p}}(\alpha_a)$, are shown in Figure 9.4. The third example in the figure, in which $K$ corresponds to the CoM of the arm under the assumption that both links are approximately equal in mass,[6] is the main intended application of the generalised inverse arm kinematics. In this case, as indicated in the figure,

$$p_x(\alpha_a) = \frac{\sin 2\alpha_a}{\sqrt{10 + 6\cos 2\alpha_a}}, \qquad p_z(\alpha_a) = \frac{-(3 + \cos 2\alpha_a)}{\sqrt{10 + 6\cos 2\alpha_a}}. \tag{9.37}$$

Expanding Equation (9.35b) yields

$$r_x = p_z(\alpha_a)\cos\theta_{sr}\sin\theta_{sp} + p_x(\alpha_a)\cos\theta_{sp}, \tag{9.38a}$$

$$r_y = -p_z(\alpha_a)\sin\theta_{sr}, \tag{9.38b}$$

$$r_z = p_z(\alpha_a)\cos\theta_{sr}\cos\theta_{sp} - p_x(\alpha_a)\sin\theta_{sp}. \tag{9.38c}$$

---

6  It is also implicitly assumed that the CoM of each individual link is approximately at its geometric centre, i.e. halfway down the link.

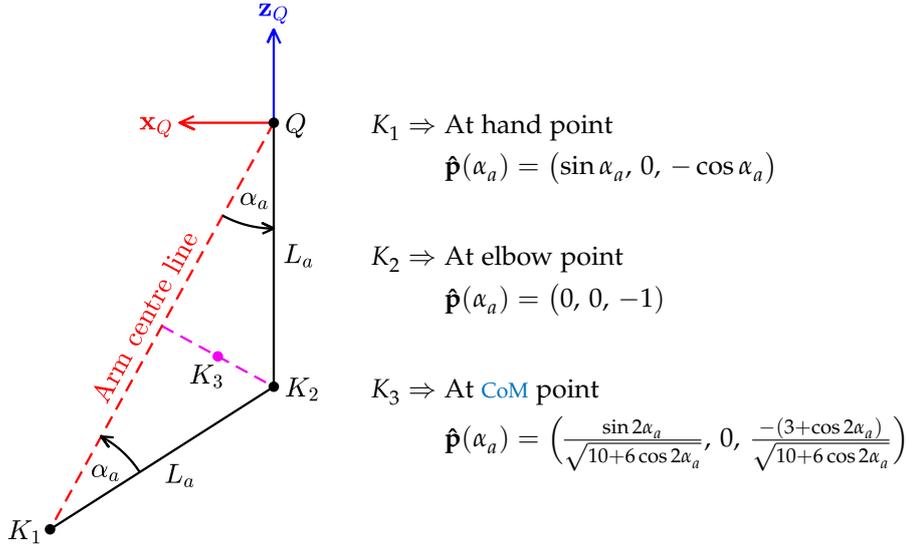

$K_1 \Rightarrow$ At hand point
$$\hat{\mathbf{p}}(\alpha_a) = (\sin\alpha_a,\ 0,\ -\cos\alpha_a)$$

$K_2 \Rightarrow$ At elbow point
$$\hat{\mathbf{p}}(\alpha_a) = (0,\ 0,\ -1)$$

$K_3 \Rightarrow$ At CoM point
$$\hat{\mathbf{p}}(\alpha_a) = \left(\frac{\sin 2\alpha_a}{\sqrt{10+6\cos 2\alpha_a}},\ 0,\ \frac{-(3+\cos 2\alpha_a)}{\sqrt{10+6\cos 2\alpha_a}}\right)$$

Figure 9.4: Examples of three possible points of interest $K$ for the generalised arm inverse kinematics algorithm, along with the corresponding definitions of the unit vector $\hat{\mathbf{p}}(\alpha_a)$, which is allowed to be a function of the arm base angle $\alpha_a$. The CoM point $K_3$ is computed under the simplifying assumption that the CoMs of the upper and lower arms are at their respective midpoints, i.e. the midpoints of $QK_2$ and $K_2K_1$, respectively.

Given a value of $\alpha_a$ (nominally $\alpha_{nom}$), these equations can be solved simultaneously to yield the required solution

$$\theta_{ep} = -2\alpha_a, \tag{9.39a}$$

$$\theta_{sr} = \operatorname{asin}\left(-\frac{r_y}{p_z(\alpha_a)}\right), \tag{9.39b}$$

$$\theta_{sp} = \operatorname{atan2}(r_x, r_z) - \operatorname{atan2}(p_x(\alpha_a),\ p_z(\alpha_a)\cos\theta_{sr}). \tag{9.39c}$$

If the required arm CoM pose is infeasible for the given $\alpha_a$, the operand in Equation (9.39b) goes outside the interval $[-1, 1]$, and the arcsine operation becomes invalid. This could be remedied by coercing the operand in Equation (9.39b) to $[-1, 1]$, but seeing as we must choose $\alpha_a$ such that

$$|\theta_{sr}| \le \theta_{srm} \le \tfrac{\pi}{2}, \tag{9.40}$$

we know anyway that the infeasible case never comes into play, as

$$|\theta_{sr}| \le \theta_{srm} \le \tfrac{\pi}{2} \implies |\sin\theta_{sr}| \le \sin\theta_{srm}$$
$$\implies \left|\frac{r_y}{p_z(\alpha_a)}\right| \le \sin\theta_{srm} \le 1. \tag{9.41}$$

In fact, we *apply* the required constraint on the shoulder roll exactly by solving Equation (9.41), i.e. by solving

$$|p_z(\alpha_a)| \ge \frac{|r_y|}{\sin\theta_{srm}} \equiv P, \tag{9.42}$$

to yield a constraint on the arm base angle $\alpha_a$. If $p_z(\alpha_a)$ is as given in Equation (9.37), then Equation (9.42) is true if $\alpha \leq \alpha_{max}$, where

$$\alpha_{max} = \tfrac{1}{2}\operatorname{acos}(M), \tag{9.43}$$

and

$$M = \begin{cases} 1, & \text{if } P \geq 1, \\ 3(P^2 - 1) + P\sqrt{9P^2 - 8}, & \text{if } \tfrac{\sqrt{8}}{3} \leq P < 1, \\ -\tfrac{1}{3}, & \text{otherwise.} \end{cases} \tag{9.44}$$

Thus, the required value of $\alpha_a$ to use with Equation (9.39) is given by

$$\alpha_a = \min\{\alpha_{nom}, \alpha_{max}\}. \tag{9.45}$$

This completes the generalised arm inverse kinematics algorithm. Note that for clarity of presentation, hard coercion of $\alpha_a$ to $\alpha_{max}$ is shown, but in implementation soft coercion (see Appendix A.1.2.3) can be used for smoother arm trajectories.

### 9.2.5 Velocity Space Conversions

Given that it is now known how to convert between poses expressed in all four pose spaces, it is also relevant for trajectory generation purposes to be able to convert between the corresponding velocities. This is relatively simple to calculate and express in terms of Jacobian matrices, so only the main conversion equations for the legs are explicitly presented here.[7] The pose velocities in the joint, abstract, inverse and tip spaces are respectively given by

$$\dot{\mathbf{q}}_l = (\dot{\theta}_{hy}, \dot{\theta}_{hr}, \dot{\theta}_{hp}, \dot{\theta}_{kp}, \dot{\theta}_{ap}, \dot{\theta}_{ar}) \in \mathbb{R}^6, \tag{9.46a}$$

$$\dot{\boldsymbol{\Phi}}_l = (\dot{\phi}_{lx}, \dot{\phi}_{ly}, \dot{\phi}_{lz}, \dot{\phi}_{fx}, \dot{\phi}_{fy}, \dot{e}_l) \in \mathbb{R}^6, \tag{9.46b}$$

$$(\dot{\mathbf{p}}_l, \boldsymbol{\Omega}_l) \in \mathbb{R}^6, \tag{9.46c}$$

$$(\dot{\mathbf{t}}_l, \boldsymbol{\Omega}_l) \in \mathbb{R}^6, \tag{9.46d}$$

where $\boldsymbol{\Omega}_l$ is the angular velocity vector corresponding to the rate of change of the foot orientation $q_l$.

The conversions between the joint and abstract space velocities are relatively simple (and sparse), and can be expressed as the matrix equations

$$\dot{\boldsymbol{\Phi}}_l = J_{AJ}\dot{\mathbf{q}}_l, \tag{9.47a}$$

$$\dot{\mathbf{q}}_l = J_{JA}\dot{\boldsymbol{\Phi}}_l, \tag{9.47b}$$

---

7 All velocity conversions not explicitly presented here can be found in the code at: https://github.com/AIS-Bonn/humanoid_op_ros/blob/master/src/nimbro/hardware/humanoid_kinematics/src/serial/serial_kinematics.cpp

where $J_{AJ}$, $J_{JA}$ are the appropriate Jacobian matrices. Given $\theta_{kp} \in [0, \pi]$ and

$$A = \sin(\tfrac{1}{2}\theta_{kp}) \tag{9.48a}$$

$$= \sqrt{\epsilon_l(2 - \epsilon_l)}, \tag{9.48b}$$

the required Jacobian matrices are given by

$$J_{AJ} = \begin{bmatrix} 0 & 1 & 0 & 0 & 0 & 0 \\ 0 & 0 & 1 & \frac{1}{2} & 0 & 0 \\ 1 & 0 & 0 & 0 & 0 & 0 \\ 0 & 1 & 0 & 0 & 0 & 1 \\ 0 & 0 & 1 & 1 & 1 & 0 \\ 0 & 0 & 0 & \frac{1}{2}A & 0 & 0 \end{bmatrix}, \tag{9.49a}$$

$$J_{JA} = \begin{bmatrix} 0 & 0 & 1 & 0 & 0 & 0 \\ 1 & 0 & 0 & 0 & 0 & 0 \\ 0 & 1 & 0 & 0 & 0 & -\frac{1}{A} \\ 0 & 0 & 0 & 0 & 0 & \frac{2}{A} \\ 0 & -1 & 0 & 0 & 1 & -\frac{1}{A} \\ -1 & 0 & 0 & 1 & 0 & 0 \end{bmatrix}. \tag{9.49b}$$

Naturally, we have that

$$J_{JA} = J_{AJ}^{-1}, \tag{9.50a}$$

$$J_{AJ} = J_{JA}^{-1}. \tag{9.50b}$$

As the foot angular velocity $\boldsymbol{\Omega}_l$ is shared between the inverse and tip velocity spaces, the complete conversion equations between these two spaces is simply given by

$$\dot{\mathbf{t}}_l = \dot{\mathbf{p}}_l + \boldsymbol{\Omega}_l \times L_{q_l}(\mathbf{f}), \tag{9.51a}$$

$$\dot{\mathbf{p}}_l = \dot{\mathbf{t}}_l - \boldsymbol{\Omega}_l \times L_{q_l}(\mathbf{f}), \tag{9.51b}$$

where $\mathbf{f} \in \mathbb{R}^3$ is the foot offset vector from Equation (9.16b).

The only missing link in the chain of velocity conversions is now how to convert from the joint and abstract velocity spaces to the inverse and tip velocity spaces, and back again. We examine the case of converting between the abstract and inverse velocity spaces. Treating all vectors as column vectors, the forward kinematics velocity conversion from the abstract space to the inverse space is given by

$$\begin{bmatrix} \dot{\mathbf{p}}_l \\ \boldsymbol{\Omega}_l \end{bmatrix} = J_{IA} \dot{\boldsymbol{\Phi}}_l, \tag{9.52}$$

where $J_{IA}$ is the $6 \times 6$ Jacobian matrix

$$J_{IA} = \begin{bmatrix} \hat{\mathbf{a}}_{lx} \times \mathbf{p}_{YA} & \hat{\mathbf{a}}_{ly} \times \mathbf{p}_{YA} & \hat{\mathbf{a}}_{lz} \times \mathbf{p}_{HA} & \mathbf{0} & \mathbf{0} & 2L_l \hat{\mathbf{a}}_{l\epsilon} \\ \hat{\mathbf{a}}_{lx} - \hat{\mathbf{a}}_{fx} & \mathbf{0} & \hat{\mathbf{a}}_{lz} & \hat{\mathbf{a}}_{fx} & \hat{\mathbf{a}}_{fy} & \mathbf{0} \end{bmatrix}, \tag{9.53}$$

and $\mathbf{p}_{YA}$, for example, is the vector from the origin of frame {Y} (i.e. the hip PR point) to the origin of frame {A} (i.e. the ankle point), as per Table 9.1. The reverse inverse kinematics velocity conversion is given by

$$\dot{\boldsymbol{\Phi}}_l = J_{AI} \begin{bmatrix} \dot{\mathbf{p}}_I \\ \boldsymbol{\Omega}_I \end{bmatrix}, \tag{9.54}$$

where

$$J_{AI} = J_{IA}^{-1}. \tag{9.55}$$

In numerical implementations, the inverse of the matrix $J_{IA}$ does not actually need to be computed, as it is more stable to directly numerically solve Equation (9.52) for $\dot{\boldsymbol{\Phi}}_l$, using decompositions of $J_{IA}$ like the Householder QR decomposition.

## 9.3   ADAPTING TO OTHER KINEMATICS

So far, this chapter has presented the robot kinematics for the case of a serial kinematics model of the robot. Although this model applies natively to robots like the NimbRo-OP and igus Humanoid Open Platform, it clearly does not apply to all possible robot configurations. In particular, parallel kinematics robots can be fitted with a serial kinematics model, but the model will at best be a good approximation that breaks down for larger deviations from the zero pose. As such, humanoid_kinematics—the Robot Operating System (ROS) package that implements all of the kinematics models and algorithms discussed in this chapter—was designed with the option of multiple layers of abstraction in mind. With significant use of inheritance and templating, the package supports the creation and parallel use of multiple different kinematic models, that are then runtime drop-in replaceable. It is thereby possible to write motion generation (e.g. gait) code that knows which pose spaces exist and which parameters they are made of, but not their exact kinematic-specific definitions, and how they are to be converted. This is a basic level of abstraction that can make motion generation code more generic, but the true power of the humanoid_kinematics package is its ability to *totally* abstract away all information other than the fact that there are four pose spaces called the joint, abstract, inverse and tip spaces. This allows motion generation algorithms like the *keypoint gait generator* (Chapter 14) to be implemented in such a way that they are almost completely agnostic of the humanoid kinematic model in use—whether serial kinematics, or parallel kinematics, or any other. This complete level of abstraction allows motion generation schemes to be implemented that are extremely portable and applicable to any type of robot, which is of great value.

# STATE ESTIMATION

In Chapter 4, we discussed in detail how the acquired sensor data is calibrated, filtered and processed to ensure that it is maximally useful in measuring the state of the robot. The acquired data included elements like the joint encoder positions, as well as the gyroscope, accelerometer and magnetometer values. A kinematic calibration of the robot was also discussed (see Section 4.1.2), which involved the creation of a dynamic model of the robot in Unified Robot Description Format (URDF), along with the subsequent tuning of the humanoid kinematic parameters given in Table 9.2, so that poses of the robot can be expressed and converted between all four available pose spaces (see Section 9.1).

In this chapter, we discuss how the calibrated and corrected sensor data is fused together into higher level estimates of the state of the robot. In particular, we discuss how the orientation (also referred to as the attitude) of the robot is estimated from the Inertial Measurement Unit (IMU) data, as well as how this estimated orientation is then used to estimate the Centre of Mass (CoM) state and walking odometry of the robot. All of these higher level state estimates are used by the various motion modules, including in particular the three implemented gaits, for the purpose of future state prediction and feedback.

## 10.1 ATTITUDE ESTIMATION

Attitude estimation is the task of constructing an estimate of the full 3D orientation of a body relative to a global fixed frame, based on a finite history of sensor measurements, e.g. gyroscope, accelerometer and magnetometer measurements. The body in question is often a robot, but in principle it can correspond to any object that is equipped with the sensors necessary for the estimation task. With the advent of low-cost inertial sensors—particularly those based on Micro-Electro-Mechanical Systems (MEMS)—the field of application for attitude estimation techniques has greatly widened, extending into the field of low-cost robotics. With low cost sensors and processors however, it is crucial that any estimation algorithms are able to run computationally efficiently, and are able to function with high noise inputs without excessively sacrificing estimator response. In addition to low estimator latency, orientation-independent mathematical and numerical stability are also desirable. For balance-related applications like bipedal walking, it is also important that the magnetometer does not influence the non-heading components of the estimated result, as these are the components that are most relevant for balance feedback, and the magnetometer is generally a less robust sensor than the gyroscope and accelerometer due to the high prevalence of magnetic inconsistencies, disturbances and distortions.

An attitude estimator that fulfils the aforementioned criteria is presented in this section. The estimator is available as part of the Humanoid Open Platform ROS Software[1], or completely separately as a generic portable C++ library (Allgeuer, 2016). All of the algorithms and cases discussed in this section are implemented in the release(s), and have been tested both in simulation and on all of the real humanoid platforms listed in Section 2.1.2, including in particular for many years under the strenuous conditions of the RoboCup competition.

### 10.1.1   Related Work

Much effort has been made in the past to develop algorithms for the reconstruction of attitude in aeronautical environments. This work was largely in relation to the Attitude and Heading Reference Systems (AHRS) required for aeronautical applications, with examples being the works of Gebre-Egziabher et al. (2004) and Munguía and Grau (2014). Other classic works in the area of attitude estimation, such as Vaganay et al. (1993) and Balaram (2000), have focused more on robotics and control applications, but do not specifically address the issues encountered with low-cost IMU systems. A comprehensive survey of modern nonlinear filtering methods for attitude estimation was undertaken by Crassidis et al. (2007). Almost all of the surveyed advanced filtering techniques relied on some form of the Extended Kalman Filter (EKF), with various modifications being used to improve particular characteristics of the filter, often convergence. Such EKF filters can be seen to be computationally expensive however, when considering implementation on embedded targets such as microcontrollers. It is often also difficult to provide a guarantee of filter robustness (Euston et al., 2008), and difficult to ensure that the magnetometer measurements do not significantly affect the estimated non-heading components of orientation.

Alternative to the general stream of development of EKF filtering is the concept of complementary filtering. This builds on the well-known linear single-input single-output (SISO) complementary filters, and extends these in a nonlinear fashion to the full 3D orientation space. Such filters have favourable frequency response characteristics, and seek to fuse low frequency attitude information with high frequency attitude rate data. Prominent examples of generalised complementary filters include the works of Jensen (2011) and Mahony et al. (2008). A different approach again is taken by Madgwick et al. (2011), who

---

formulate a numerical filter directly in the quaternion space, and numerically integrate quaternion velocities $\dot{q}$ with the use of renormalisation to estimate the required orientation. The method, developed in the context of wearable devices for rehabilitation robotics, does not respect magnetometer independence, and uses a non-ideal gradient descent numerical approximation for orientation reconstruction that converges over multiple cycles. An empirical comparison of the Madgwick filter, a complementary filter and an EKF-based approach is performed in Cavallo et al. (2014).

The orientation estimation problem addressed in this thesis relates specifically to the design of an attitude estimator that can function with noisy low-cost sensors, and that is simple and efficient enough to be implemented at high loop rates on low-power embedded targets, such as microcontrollers. To this end, the work presented by Mahony et al. (2008) was used as a basis for the developed attitude estimator. A central problem in applying this work however, is that a method is required for reconstructing an instantaneous 3D orientation 'measurement' directly from the sensor measurements (see Section 10.1.5). This is a complex optimisation problem that generally requires a suboptimal solution algorithm for computational feasibility reasons (Mahony et al., 2008). Literature does not elucidate a clear solution to this problem—in particular not in an explicit form—and not in a way that can function robustly in all cases. One of the main contributions of the work presented here lies in the development of an algorithm for the robust calculation of such instantaneous orientation measurements. Other contributions include the novel use of fused yaw (see Section 5.4.1) in an estimator, the integration of a *quick learning* scheme (Section 10.1.6.1), and the explicit extension of the attitude estimator to cases of reduced sensory information (Section 10.1.6).

### 10.1.2  Problem Definition and Notation

The goal of the attitude estimation process is to calculate an estimate of the rotation of a body relative to a global reference frame, based on observations acquired through sensory perception. Such sensory perception can include accelerometers, gyroscopes, magnetometers, Global Positioning System (GPS), visual perception and/or Light Detection and Ranging (LIDAR). The types of sensors considered for the task in this thesis are the ones that are typically found in IMU systems, and are typically available in low-cost variants for mobile robotic systems, i.e. gyroscope, accelerometer and magnetometer sensors. No matter which sensors are used however, it is always a stringent requirement that the estimator is globally stable, and is able to function equally well throughout the entire orientation space.

We define {G} to be the global reference frame relative to which the orientation of the body is estimated, and {B} to be the body-fixed frame that rotates with the body, and therefore with the sensors that provide the observational input to the attitude estimator. It is assumed that {B} is defined in such a way that its z-axis points 'upwards' (and x-axis points 'forwards') relative to the body, and correspondingly, that {G} is defined in such a way that its z-axis points 'upwards' relative to the world, i.e. in the direction opposite to gravity. Importantly, this means that the gravity vector can be written as

$$^G\mathbf{g} = (0, 0, -g),    \tag{10.1}$$

where $g = 9.81\,\text{m/s}^2$. These are the standard definitions of {G} and {B} that were used throughout the rotations part of this thesis (Part III). All further notation that was introduced and used in the rotations part is also considered to carry across to this chapter, including for instance the rotation basis notation

$$^G_B R = \begin{bmatrix} \uparrow & \uparrow & \uparrow \\ ^G\mathbf{x}_B & ^G\mathbf{y}_B & ^G\mathbf{z}_B \\ \downarrow & \downarrow & \downarrow \end{bmatrix} = \begin{bmatrix} \leftarrow & ^B\mathbf{x}_G & \rightarrow \\ \leftarrow & ^B\mathbf{y}_G & \rightarrow \\ \leftarrow & ^B\mathbf{z}_G & \rightarrow \end{bmatrix},    \tag{10.2}$$

where for example $^B\mathbf{z}_G$ is the unit z-vector of frame {G}, expressed in the coordinates of frame {B}. A quick-reference summary of the relevant notation is provided in the Notation section on page xxxi. In addition to the notation outlined there, note that throughout this chapter we refer to two vectors as

- **Parallel** if they are linearly dependent and a positive multiple of each other,

- **Antiparallel** if they are linearly dependent and a negative multiple of each other, and

- **Collinear** if they are either parallel or antiparallel.

As a final note, the concepts of *fused yaw* and *tilt* are used in many places within this chapter, and should be reviewed, if necessary, from Sections 5.4.1 and 5.4.2.

### 10.1.3    Sensor Inputs

Using the notation and definitions introduced in the previous section, the problem considered in this paper can be more precisely reformulated as being the task of robustly calculating an online estimate for $^G_B q$ (or $^G_B R$), given arbitrary time sequences of 3-axis gyroscope, accelerometer and magnetometer data. Keeping in mind the sensor and calibration models introduced in Chapter 4, the format and type of data provided by each of these sensors is assumed to be modelled as follows.

### 10.1.3.1 *Gyroscope*

This sensor is assumed to provide a measure ${}^B\mathbf{\Omega}_y \in \mathbb{R}^3$ of the angular velocity of the body, in the coordinates of frame {B}. The conversion of the raw data from the IMU frame {I} to the body-fixed frame {B} is performed using the calibrated IMU orientation offset ${}^B_I R$ (see Section 4.2.1). As given by Equation (4.12), after scale correction and temperature compensation, the gyroscope measurement is assumed to be affected by a largely time-invariant gyroscope bias $\mathbf{b}_\Omega$, along with zero mean sensor noise $\mathbf{w}_\Omega$. The bias $\mathbf{b}_\Omega$ is managed using the gyroscope bias calibration and online gyroscope bias autocalibration schemes (see Sections 4.2.3 and 4.2.4), and the sensor noise $\mathbf{w}_\Omega$ is not explicitly abated using low-pass filtering, as the gyroscope data is directly integrated as part of the estimation process, and it is the nature of integration to dampen out high frequency components of a signal. The final corrected and calibrated angular velocity measurement

$$ {}^B\mathbf{\Omega}_y \in \mathbb{R}^3, \tag{10.3} $$

is used as the gyroscope input to the attitude estimator, and is expressed in the units of rad/s.

### 10.1.3.2 *Accelerometer*

This sensor is assumed to provide a measure ${}^B\tilde{\mathbf{a}} \in \mathbb{R}^3$ of the proper acceleration of the body. This is the inertial coordinate acceleration being experienced by the body, together with the effect of gravitational acceleration. The latter term is assumed to dominate the measured proper acceleration. Cases where this assumption is violated, like for example in collisions, are implicitly filtered out by the low-pass dynamics of the estimator. Grouping the inertial acceleration components with a zero mean noise term $\mathbf{w}_a$, yields the measurement model

$$ {}^B\tilde{\mathbf{a}} = -{}^B_G R \, {}^G\mathbf{g} + \mathbf{w}_a. \tag{10.4} $$

The minus sign comes from the definition of proper acceleration, as an object at rest on the Earth's surface, for instance, has a proper acceleration of $g = 9.81\,\text{m/s}^2$ upwards, but the gravity vector ${}^G\mathbf{g}$ nominally points downwards, as in Equation (10.1). The raw accelerometer measurements are brought from the IMU frame {I} to the body-fixed frame {B} using the IMU orientation offset ${}^B_I R$ (see Section 4.2.1), and are subsequently low-pass filtered using a mean filter (see Appendix A.2.1.1) to combat the sensor noise. The resulting measured proper acceleration is then

$$ {}^B\tilde{\mathbf{a}} \approx -{}^B_G R \, {}^G\mathbf{g} = {}^B_G R \,(0,0,g) \tag{10.5a} $$

$$ = g\,{}^B\mathbf{z}_G. \tag{10.5b} $$

Thus, for each filtered accelerometer measurement ${}^B\tilde{\mathbf{a}}$, we can construct an estimate

$$ {}^B\tilde{\mathbf{z}}_G = \frac{{}^B\tilde{\mathbf{a}}}{\|{}^B\tilde{\mathbf{a}}\|} \in \mathcal{S}^2, \tag{10.6} $$

of the positive global z-vector ${}^B\mathbf{z}_G$, and use this as the accelerometer-based input to the attitude estimator.

### 10.1.3.3 *Magnetometer*

The measurement model and calibration process of the magnetometer sensor is relatively complex, and was described in detail in Section 4.3. The magnetometer correction pipeline (see Section 4.3.1) involved spike and mean filtering the data, as well as the application of hard and/or soft iron corrections, and/or cyclic angle warping for final fine-tuning. It is assumed that the final extracted magnetometer measurement ${}^B\tilde{\mathbf{m}}$ is expressed relative to the body-fixed frame {B}, and is a direct measure of the strength and direction of the Earth's magnetic field ${}^G\mathbf{m}_e = (m_{ex},\, m_{ey},\, m_{ez})$, i.e.

$$ {}^B\tilde{\mathbf{m}} \approx {}^B_G R\, {}^G\mathbf{m}_e \in \mathbb{R}^3. \tag{10.7} $$

### 10.1.4 **Complementary Filtering**

The method of attitude estimation presented in this chapter is based on the concept of complementary filtering.

### 10.1.4.1 *1D Linear Complementary Filter*

A simple preliminary approach to the attitude estimation problem is to separate the problem into each of its independent axes of rotation. This can work for body rotations close to the upright identity pose, but does not extend well to the whole orientation space. Nevertheless, the 1D filtering approach demonstrates well the concept of linear complementary filtering. Taking for example the pitch direction of rotation, one can express the filter equations as

$$ \dot{\hat{\theta}} = \omega_y - \hat{c}_\omega + k_p(\theta_y - \hat{\theta}) \tag{10.8a} $$

$$ \dot{\hat{c}}_\omega = -k_i(\theta_y - \hat{\theta}), \tag{10.8b} $$

where

$\hat{\theta}\ \Rightarrow$ Estimated pitch angle of the body (output of the filter)

$\omega_y\ \Rightarrow$ Angle rate measurement in the pitch direction, based on the gyroscope

$\hat{c}_\omega\ \Rightarrow$ Estimated integral term as an offset to $\omega_y$

$\theta_y\ \Rightarrow$ An instantaneous measurement of the pitch angle, based solely on the accelerometer (and possibly magnetometer)

$k_p \Rightarrow$ Proportional gain indicating the influence of $\theta_y$ on $\hat{\theta}$

$k_i \Rightarrow$ Integral gain indicating the influence of $\theta_y$ on $\hat{c}_\omega$

Proportional-Integral (PI) filter equations similar to Equation (10.8) can also be formulated for the roll and yaw directions, where, if no magnetometer is used, the $\theta_y - \hat{\theta}$ error term for the yaw case needs to be left as zero, as the accelerometer alone cannot yield information about the yaw of the body.

Effectively, the PI compensation closes the loop on the type I system, forming a linear second-order system with zero theoretical steady state error to step inputs. The linear complementary filter combines the high-pass rate data with the low-pass position data to form a high bandwidth estimate of the system state. However, despite possessing positive filter attributes, the assumption that each axis behaves independently places a severe limitation on the usability of the filter for attitude estimation. A core issue is that the angular velocity about one axis generally affects the rotation about all axes, and to differing amounts depending on the orientation of the body. The three independent 1D filters also do not clearly indicate how the three estimated outputs can be unambiguously and meaningfully combined into a total estimation of the 3D orientation of the body.

### 10.1.4.2 *Extension to 3D Nonlinear Filtering*

In light of the limitations of the 1D complementary filter, it is desirable to formulate a complementary filter that operates on the full 3D rotation space, ideally retaining the positive frequency attributes of the linear filter. Mahony et al. (2008) introduced three such nonlinear filters, the *direct*, *passive* and *explicit complementary filters*. The main difference between the three filters is that while the direct complementary filter uses the instantaneous inertial sensor data to transform the gyroscope measurements in the update equation, the passive complementary filter uses the current filter estimate, and the explicit complementary filter uses an update technique that operates directly on the sensor measurement vectors.

A key design decision of the attitude estimator presented here is that the magnetometer measurements should not have any direct influence on the attitude estimate, other than to resolve the yaw, i.e. heading. The reason for this is to reduce instabilities in the output pitch and roll components, and to alleviate the requirement of performing a magnetometer calibration in order for these components of the estimate to function correctly. This is impossible to achieve with the explicit complementary filter, and so the only filter in Mahony et al. (2008) to provide a solution to the problem of constructing an instantaneous orientation measurement from sensor data was found to be unsuitable. Comparison of the direct and passive filters also led to the conclusion that the feed-forward nature of the direct formulation was unsuitable due to high frequency noise considerations. Consequently, the attitude estimator presented in this chapter is built around the core of the nonlinear passive complementary filter.

### 10.1.4.3   *3D Nonlinear Passive Complementary Filter*

As indicated in Figure 10.1, we define frame {E} as the frame corresponding to the current estimate

$$
{}^{G}_{E}\hat{q} \equiv \hat{q} \in \mathbb{Q} \tag{10.9}
$$

of the orientation of the body-fixed frame {B}, relative to the global frame {G}.[2] In every update cycle, given the current accelerometer and magnetometer-based sensor measurements ${}^{B}\tilde{\mathbf{z}}_{G}$ and ${}^{B}\tilde{\mathbf{m}}$ (and if needed also $\hat{q}$), the first task is to construct a full 3D measured quaternion orientation

$$
{}^{G}_{B}q_{y} \equiv q_{y} \in \mathbb{Q}, \tag{10.10}
$$

that is consistent with these measurements. The frame that results from this reconstruction of an orientation from the sensor measurements is labelled frame {B̃}, and within the accuracy of the sensors and respective calibrations should correspond to the frame {B}. The deviation of the measured quaternion $q_{y}$ from the current estimated quaternion $\hat{q}$ can be calculated as

$$
\tilde{q} = \hat{q}^{*} q_{y}, \tag{10.11}
$$

and is referred to as the *error quaternion*

$$
{}^{E}_{B}\tilde{q} \equiv \tilde{q} \equiv (\tilde{q}_{0}, \tilde{\mathbf{q}}) \in \mathbb{Q}. \tag{10.12}
$$

The rotation axis of this error quaternion rotates the current orientation estimate towards the orientation given by the sensor measurements, and leads to the corrective error feedback term

$$
\mathbf{\Omega}_{e} = 2\tilde{q}_{0}\tilde{\mathbf{q}}. \tag{10.13}
$$

This term is an angular velocity in $\mathbb{R}^{3}$ that rotates $\hat{q}$ towards $q_{y}$, with an angular speed of exactly the sine of the angle between these two orientations. Using the corrective error feedback term, the nonlinear 3D passive complementary filter equations can be written as

$$
\dot{\hat{q}} = \tfrac{1}{2}\hat{q}(\mathbf{\Omega}_{y} - \hat{\mathbf{c}}_{\Omega} + k_{p}\mathbf{\Omega}_{e}), \tag{10.14a}
$$

$$
\dot{\hat{\mathbf{c}}}_{\Omega} = -k_{i}\mathbf{\Omega}_{e}, \tag{10.14b}
$$

where the main filter equation is based directly on Equation (7.120), and we recall from Equation (7.33) that quaternions and vectors can

---

2  To aid understanding, if the output quaternion $\hat{q}$ of the attitude estimator is perfectly accurate, i.e. $\hat{q} = {}^{G}_{B}q$, then clearly {E} = {B}, so {E} is the attitude estimator's estimate of the frame {B}.

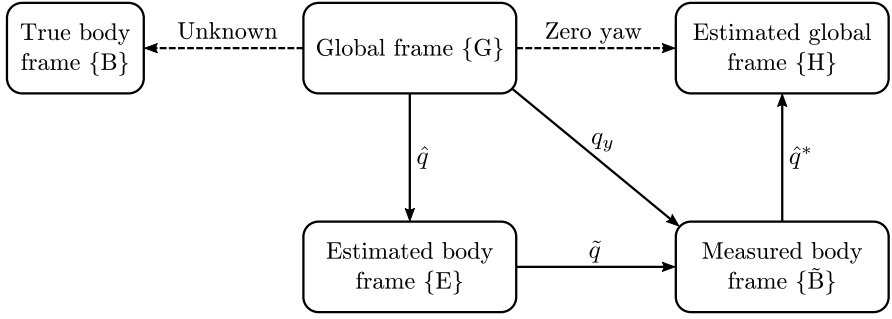

Figure 10.1: Overview of the coordinate frame definitions for the 3D non-linear passive complementary filter, and measured quaternion orientation resolution methods.

be directly multiplied with each other by making the vectors into quaternions themselves by adding a zero scalar part (w-component). The variables involved in the filter equations are given by

$\hat{q}$ ⇒ Estimated quaternion orientation of the body (output of the filter)

$\boldsymbol{\Omega}_y$ ⇒ Measured angular velocity of the body based on the gyroscope, as given by Equation (10.3)

$\hat{\mathbf{c}}_{\Omega}$ ⇒ Estimated integral term as an offset to $\boldsymbol{\Omega}_y$

$\boldsymbol{\Omega}_e$ ⇒ Corrective error feedback angular velocity term, as given by Equation (10.13)

$k_p$ ⇒ Proportional gain indicating the influence of $\boldsymbol{\Omega}_e$ on $\hat{q}$

$k_i$ ⇒ Integral gain indicating the influence of $\boldsymbol{\Omega}_e$ on $\hat{\mathbf{c}}_{\Omega}$

The PI gains $k_p$ and $k_i$ of the filter should be tuned to provide non-oscillatory yet responsive transients, as limited by sensor noise. Note that mathematically, Equation (10.14a) simply computes the quaternion velocity $\dot{\hat{q}}$ that is equivalent to the effect of the angular velocity

$$\boldsymbol{\Omega}_y - \hat{\mathbf{c}}_{\Omega} + k_p \boldsymbol{\Omega}_e \tag{10.15}$$

being applied to the quaternion $\hat{q}$. The direct parallel of the 3D filter update equations to the case for the 1D complementary filter in Equation (10.8) is clear to see by equivalating

$$\hat{q} \leftrightarrow \hat{\theta}, \qquad\qquad \boldsymbol{\Omega}_y \leftrightarrow \omega_y, \tag{10.16a}$$

$$\hat{\mathbf{c}}_{\Omega} \leftrightarrow \hat{c}_{\omega}, \qquad\qquad \boldsymbol{\Omega}_e \leftrightarrow \theta_y - \hat{\theta}. \tag{10.16b}$$

Given the filter equations in Equation (10.14), in each time step the variables $\hat{q}$ and $\hat{\mathbf{c}}_{\Omega}$ are updated based on numerical trapezoidal integration as a function of the measured elapsed time $\Delta t$ since the previous time step. Given the nominal update cycle time $\Delta t_n$ of the estimator, it is recommended for robustness purposes to coerce $\Delta t$ to some interval scaled by $\Delta t_n$, in order to avoid large jumps in the estimator states when lags occur, and ensure greater correctness of the gyroscope angular velocity integration in general. In the implementation of the attitude estimator as part of the Humanoid Open Platform ROS Software, the measured elapsed time in each cycle is coerced to the range

$$\Delta t \in \big[0.8\Delta t_n,\ 2.2\Delta t_n\big], \tag{10.17}$$

as these are known to be realistic limits for the true cycle time in all normal cases.

The stability of the passive complementary filter is discussed in detail in Mahony et al. (2008). Theoretical analysis demonstrates that there is a measure zero set in the space of all possible measured rotation and bias errors such that equilibrium exists despite lack of convergence. The equilibrium is unstable however, and the error is locally exponentially stable in all other cases. This set consists of all error states such that $\hat{\mathbf{e}}_\Omega$ is constant and equal to $\boldsymbol{\Omega}_y$, and $\tilde{q}$ is a rotation by $\pi$ radians. This pathological set is of no concern however, as it is never reached in any practical situation. Even intentional initialisation of the filter to such an equilibrium state in simulated experiments did not prove to be a problem, as mere arithmetic floating point errors were enough for the divergent dynamics near the pathological set to take over, and force the estimator away from this unstable equilibrium.

### 10.1.5 Measured Quaternion Orientation Resolution Methods

Given the 3D nonlinear passive complementary filter presented in the previous section, the core task that remains is to specify how the measured quaternion orientation $q_y$ is to be constructed from the current accelerometer and magnetometer-based measurements $^B\tilde{\mathbf{z}}_G$ and $^B\tilde{\mathbf{m}}$ respectively. The various approaches to doing this are referred to as the measured quaternion orientation resolution methods, and can, if needed, also use the current estimated orientation $\hat{q}$ as an input—for instance to ensure in the case of a lack of information that the generated $q_y$ is as close as possible to $\hat{q}$. The various resolution methods, all of which are implemented and used in the released attitude estimator code, are described in this section.

#### 10.1.5.1 *Magnetometer Resolution Method*

As can be seen from Equation (10.11), the calculation of the error quaternion $\tilde{q}$, requires knowledge of $q_y$, the instantaneous measured orientation best fitting the sensor measurements $^B\tilde{\mathbf{z}}_G$ and $^B\tilde{\mathbf{m}}$. In general, these two measurements suffice to construct a unique rotation $q_y$ that best fits the given data. If not, $\hat{q}$ is taken as a further input in the place of $^B\tilde{\mathbf{m}}$, and one of the resolution methods described in the following sections is used instead.[3] No matter what resolution method is used however, it is a design decision that the quaternion $q_y \equiv {}^G_B q_y$ must *always* be calculated in such a way that it satisfies

$$^B \mathbf{z}_G = {}^B \tilde{\mathbf{z}}_G, \tag{10.18}$$

i.e. $q_y$ must always be consistent with the direction of gravity measured by the accelerometer, with only the heading component being decided by either $^B \tilde{\mathbf{m}}$ or $\hat{q}$.

For the *magnetometer resolution method*, we use $^B \tilde{\mathbf{m}}$ to resolve the heading of $q_y$. In order to do this, we need a calibrated value of $^G \mathbf{m}_e$—the magnetic field vector of the Earth relative to the nominated global reference frame {G}. The calibration of this vector, known as the reference field vector, was discussed in Section 4.3.6. For the purposes of the magnetometer resolution method, only the x and y-components of the reference field vector are required, i.e. $m_{ex}$ and $m_{ey}$, and these are used only to resolve the heading of $q_y$.

The quaternion $q_y$ is calculated by first constructing the rotation matrix

$$^G_B R_y = \begin{bmatrix} \leftarrow & ^B \mathbf{x}_G & \rightarrow \\ \leftarrow & ^B \mathbf{y}_G & \rightarrow \\ \leftarrow & ^B \mathbf{z}_G & \rightarrow \end{bmatrix}, \tag{10.19}$$

and then converting this to quaternion form using Equation (5.118). The required value of $^B \mathbf{z}_G$ is already known from Equation (10.18), so it only remains to calculate suitable mutually orthogonal axis vectors $^B \mathbf{x}_G$ and $^B \mathbf{y}_G$. From Equation (10.7), ideally we would wish to be able to find $^B \mathbf{x}_G$ and $^B \mathbf{y}_G$ such that

$$^B \tilde{\mathbf{m}} = {}^G_B R_y^T {}^G \mathbf{m}_e, \tag{10.20}$$

but this is not necessarily possible, so instead we minimise the angular difference between the left-hand side and right-hand side vectors of this equation. The angular difference can be seen to be minimised when the respective projections of the two vectors onto the plane perpendicular to $^B \mathbf{z}_G$ are parallel. The projection of $^B \tilde{\mathbf{m}}$ can be seen to be

$$^B \hat{\mathbf{m}} = {}^B \tilde{\mathbf{m}} - \left( {}^B \tilde{\mathbf{m}} \bullet {}^B \mathbf{z}_G \right) {}^B \mathbf{z}_G, \tag{10.21}$$

and observing that

$$^G_B R_y^T {}^G \mathbf{m}_e = m_{ex} {}^B \mathbf{x}_G + m_{ey} {}^B \mathbf{y}_G + m_{ez} {}^B \mathbf{z}_G, \tag{10.22}$$

the required projection of $^G_B R_y^T {}^G \mathbf{m}_e$ can be seen to be

$$^B \hat{\mathbf{m}}_e = m_{ex} {}^B \mathbf{x}_G + m_{ey} {}^B \mathbf{y}_G. \tag{10.23}$$

---

3 The *fused yaw resolution method* is strongly recommended over the *ZYX yaw resolution method* in all situations.

Solving these equations for the condition that $^{\tilde{B}}\hat{\mathbf{m}}$ and $^{\tilde{B}}\hat{\mathbf{m}}_e$ are parallel (i.e. positive multiples of one another) yields

$$^{\tilde{B}}\mathbf{x}_G = \frac{^{\tilde{B}}\tilde{\mathbf{x}}_G}{\|^{\tilde{B}}\tilde{\mathbf{x}}_G\|}, \qquad\qquad ^{\tilde{B}}\mathbf{y}_G = \frac{^{\tilde{B}}\tilde{\mathbf{y}}_G}{\|^{\tilde{B}}\tilde{\mathbf{y}}_G\|}, \qquad (10.24)$$

where

$$^{\tilde{B}}\hat{\mathbf{u}} = {}^{\tilde{B}}\hat{\mathbf{m}} \times {}^{\tilde{B}}\mathbf{z}_G, \qquad\qquad (10.25a)$$

$$^{\tilde{B}}\tilde{\mathbf{x}}_G = m_{ex}\,^{\tilde{B}}\hat{\mathbf{m}} + m_{ey}\,^{\tilde{B}}\hat{\mathbf{u}}, \qquad (10.25b)$$

$$^{\tilde{B}}\tilde{\mathbf{y}}_G = m_{ey}\,^{\tilde{B}}\hat{\mathbf{m}} - m_{ex}\,^{\tilde{B}}\hat{\mathbf{u}}. \qquad (10.25c)$$

Equation (10.19) is then used as previously described to calculate $^{G}_{\tilde{B}}R_y$ and subsequently $q_y$. Note that the z-component $m_{ez}$ of the reference field vector is not required at any point in the calculations. The magnetometer resolution method fails due to a division by zero if $m_{ex} = m_{ey} = 0$, or if $^{\tilde{B}}\mathbf{z}_G$ and $^{B}\tilde{\mathbf{m}}$ are collinear. Both of these situations should never reasonably occur (away from the Earth's north and south poles), as it corresponds to the Earth's magnetic field being vertical in the global fixed frame—a generally unexpected case.

### 10.1.5.2   *ZYX Yaw Resolution Method*

If the general magnetometer resolution method fails to produce a valid output, the magnetometer measurement is discarded, and the measured orientation $q_y$ is instead constructed from $^{\tilde{B}}\mathbf{z}_G$ and $\hat{q}$. The latter is required as the former alone is insufficient to be able to calculate a unique $q_y$, and the latter can be used to ensure that $q_y$ is 'as close as possible' to $\hat{q}$, thereby only minimally affecting the estimate in the then-uncontrolled heading dimension (i.e. yaw).

As indicated in Figure 10.1, we define the frame {H} to be the frame {B̃} rotated by the inverse of $\hat{q}$. That is, {H} corresponds to the current estimated orientation of the global fixed frame. Note that this will not be identical to {G} in general, as $\hat{q}$ and $q_y$ generally differ, even if only slightly. The aim of the *ZYX yaw resolution method* is to find a suitable rotation matrix $^{G}_{\tilde{B}}R_y$, such that the ZYX Euler yaw (see Section 5.3.5.1) of {H} with respect to {G} is zero. This is equivalent to saying that the normalised projection of $^{\tilde{B}}\mathbf{x}_H$ onto the $\mathbf{x}_G\mathbf{y}_G$ plane should be parallel, and hence equal to, the unit vector $^{\tilde{B}}\mathbf{x}_G$. The corresponding unit vector $^{\tilde{B}}\mathbf{y}_G$ can then be computed to complete the basis. Letting $\hat{q} = (\hat{w}, \hat{x}, \hat{y}, \hat{z})$, the unit vector $^{\tilde{B}}\mathbf{x}_H$ can be calculated from Equation (5.94) to be

$$^{\tilde{B}}\mathbf{x}_H = 2\big(\tfrac{1}{2} - \hat{y}^2 - \hat{z}^2,\ \hat{x}\hat{y} - \hat{w}\hat{z},\ \hat{x}\hat{z} + \hat{w}\hat{y}\big). \qquad (10.26)$$

The required unit vectors ${}^{\mathcal{B}}\mathbf{x}_G$ and ${}^{\mathcal{B}}\mathbf{y}_G$ are thus

$$
{}^{\mathcal{B}}\mathbf{x}_G = \frac{{}^{\mathcal{B}}\tilde{\mathbf{x}}_G}{\|{}^{\mathcal{B}}\tilde{\mathbf{x}}_G\|}, \qquad\qquad {}^{\mathcal{B}}\mathbf{y}_G = \frac{{}^{\mathcal{B}}\tilde{\mathbf{y}}_G}{\|{}^{\mathcal{B}}\tilde{\mathbf{y}}_G\|}, \qquad (10.27)
$$

where

$$
{}^{\mathcal{B}}\tilde{\mathbf{x}}_G = {}^{\mathcal{B}}\mathbf{x}_H - \left({}^{\mathcal{B}}\mathbf{x}_H \boldsymbol{\cdot} {}^{\mathcal{B}}\mathbf{z}_G\right){}^{\mathcal{B}}\mathbf{z}_G, \qquad (10.28a)
$$

$$
{}^{\mathcal{B}}\tilde{\mathbf{y}}_G = {}^{\mathcal{B}}\mathbf{z}_G \times {}^{\mathcal{B}}\tilde{\mathbf{x}}_G. \qquad (10.28b)
$$

As before, Equation (10.19) is then used to calculate $q_y$ via ${}^{\mathcal{G}}_{\mathcal{B}}R_{y}$. The ZYX yaw resolution method fails only if ${}^{\mathcal{B}}\mathbf{z}_G$ and ${}^{\mathcal{B}}\mathbf{x}_H$ are collinear. This occurs if the error rotation from {G} to {H} is at gimbal lock in terms of the ZYX Euler angles. It is important to note that failure of this method depends only on this error quaternion rotation, and not in any way on the absolute rotations $\hat{q}$ and $q_y$. As a result, the algorithm is equally stable in all global orientations of the body, as desired.

If the algorithm fails, a backup algorithm that zeros the ZXY Euler yaw (see Section 5.3.5.2) of {H} with respect to {G} is employed instead. Analogously to Equation (10.26), in that case

$$
{}^{\mathcal{B}}\mathbf{y}_H = 2\left(\hat{x}\hat{y} + \hat{w}\hat{z}, \; \tfrac{1}{2} - \hat{x}^2 - \hat{z}^2, \; \hat{y}\hat{z} - \hat{w}\hat{x}\right), \qquad (10.29)
$$

after which Equation (10.27) is used again, with

$$
{}^{\mathcal{B}}\tilde{\mathbf{y}}_G = {}^{\mathcal{B}}\mathbf{y}_H - \left({}^{\mathcal{B}}\mathbf{y}_H \boldsymbol{\cdot} {}^{\mathcal{B}}\mathbf{z}_G\right){}^{\mathcal{B}}\mathbf{z}_G, \qquad (10.30a)
$$

$$
{}^{\mathcal{B}}\tilde{\mathbf{x}}_G = {}^{\mathcal{B}}\tilde{\mathbf{y}}_G \times {}^{\mathcal{B}}\mathbf{z}_G. \qquad (10.30b)
$$

Given that the previous algorithm failed, this algorithm is guaranteed never to, hence completing the ZYX yaw resolution method.

### 10.1.5.3  *Fused Yaw Resolution Method*

As discussed in Section 6.2, the concept of Euler yaw is not particularly advantageous when it comes to applications relating to balance and heading. As also discussed in Section 6.2, the fused yaw on the other hand is, so a resolution method based on this notion of yaw has also been developed. The *fused yaw resolution method* is quite similar in idea to the ZYX yaw method, only instead of zeroing the ZYX Euler yaw of {H} with respect to {G}, it zeros the fused yaw. The first notable distinction here to the ZYX yaw method is that having zero relative fused yaw is in fact a mutual relationship, as the inverse of a rotation has the exact negative of its fused yaw (see Section 5.7.4). The second notable distinction is that the notion of fused yaw is more closely related to quaternions than ZYX Euler yaw, and so a convenient direct quaternion formulation exists.

The z-vector $^H\mathbf{z}_G = (z_{Gx}, z_{Gy}, z_{Gz})$ of the true global frame {G} with respect to the estimated global frame {H} can be calculated to be

$$
\begin{aligned}
^H\mathbf{z}_G &= L_{^H_{\tilde{B}}q}(^{\tilde{B}}\mathbf{z}_G) \\
&= L_{\hat{q}}(^{\tilde{B}}\mathbf{z}_G) \\
&= \hat{q}\,^{\tilde{B}}\mathbf{z}_G\,\hat{q}^*.
\end{aligned}
\tag{10.31}
$$

Temporarily treating quaternions as column vectors in $\mathbb{R}^4$, the fused yaw resolution method can then be summarised mathematically as

$$
q_y = \frac{\tilde{q}_y}{\|\tilde{q}_y\|},
\tag{10.32}
$$

where

$$
\tilde{q}_y = \begin{bmatrix}
1 + z_{Gz} & -z_{Gy} & z_{Gx} & 0 \\
z_{Gy} & 1 + z_{Gz} & 0 & -z_{Gx} \\
-z_{Gx} & 0 & 1 + z_{Gz} & -z_{Gy} \\
0 & z_{Gx} & z_{Gy} & 1 + z_{Gz}
\end{bmatrix} \hat{q}.
\tag{10.33}
$$

From inspection it can be seen that the only case in which the algorithm fails is if $^H\mathbf{z}_G = (0, 0, -1)$. This occurs if the rotation from {G} to {H} (or equivalently, vice versa) is at the fused yaw singularity, in which case it can be derived from Equation (10.11) that the error quaternion $\tilde{q} \equiv {}^E_{\tilde{B}}q$ must be a rotation by $\pi$ radians. We recall however, from page 298, that this is the exact condition of the only unstable equilibrium point, i.e. problem case, of the passive complementary filter itself. Thus, we conclude that unlike the ZYX yaw method, the use of the fused yaw resolution method ensures that there is only a single error condition for which any part of the total estimation process yields suboptimal results. Furthermore, this one error condition is when $\tilde{q}$ is at an exact antipode of the identity rotation—a case that in practical situations is never reached. Nevertheless, for reasons of completeness and robustness, the above algorithm falls back to zeroing the ZYX yaw if it fails. This is computed using Equations (10.26) to (10.28), and is guaranteed not to fail if the fused yaw algorithm failed. It is important to note that the fused yaw resolution method is equally stable in all global orientations of the body, as its action depends only on the deviation between the two global frames {G} and {H}, and not on where in the rotation space the current attitude estimate $\hat{q}$ actually lies.

### 10.1.6 Extensions to the Estimator

Given the complete attitude estimator as described in the chapter thus far, some extensions can be made to improve performance, or allow estimation with less than the full required 9 axes of data.

### 10.1.6.1    *Quick Learning*

It is desired for the attitude estimator to settle quickly from large estimation errors, yet simultaneously provide adequate general noise rejection. To this end, *quick learning*[4] is proposed as a method to help achieve this. Quick learning allows two sets of PI gains to be tuned—one set that provides suitably fast transient response for quick convergence, and one set that provides good tracking and noise rejection for the long term. Given a desired quick learning time, a parameter $\lambda \in [0, 1]$ is then used to fade linearly between these two sets of gains over this time, where the final faded gains are the ones that provide good tracking and noise rejection. The gain fading scheme is mathematically given by

$$(k_p, k_i) = \lambda (k_p^{nom}, k_i^{nom}) + (1 - \lambda)(k_p^{quick}, k_i^{quick}). \qquad (10.34)$$

Quick learning can be triggered at any time, including automatically when the estimator starts, and is disabled when $\lambda$ subsequently returns to 1.0 via a linear ramp.

### 10.1.6.2    *Estimation with Two-axis Acceleration Data*

If only two-axis **xy** accelerometer data is available, then the missing z-component of ${}^B\tilde{\mathbf{a}}$ can be calculated by solving $\|{}^B\tilde{\mathbf{a}}\| = g$, where $g$ is the magnitude of gravitational acceleration. Letting ${}^B\tilde{\mathbf{a}} = (a_x, a_y, a_z)$, this yields

$$a_z = \sqrt{\max\{g^2 - a_x^2 - a_y^2,\, 0\}}. \qquad (10.35)$$

This however, only allows for orientation estimates in the positive z-hemisphere, as the sign of the missing $a_z$ component has to be assumed. For many applications though, like bipedal walking, this can be sufficient. Note that for optimal results, a scale calibration of the accelerometer may be necessary, and/or a sensor-specific calibration of the value of $g$.

### 10.1.6.3    *Estimation with Reduced Magnetometer Data*

Two-axis **xy** magnetometer data can still be used for ${}^B\tilde{\mathbf{m}}$ if the third unknown component is left to zero. Due to the projection operation in Equation (10.21), this in general still produces satisfactory results. The calibration process of the reference field vector ${}^G\mathbf{m}_e$ also remains the same, as the z-component $m_{ez}$ is not required for the orientation resolution algorithm anyway. If magnetometer data is only available in terms of an absolute heading angle $\psi$, then the required three-axis data can be constructed using

$$^B\tilde{\mathbf{m}} = (\cos\psi, \sin\psi, 0), \qquad (10.36)$$

---

4 This extension to the estimator is referred to as 'quick learning' mostly for historical reasons, and would more accurately be summarised as a 'gain scheduling scheme'.

and used as before.

### 10.1.6.4 *Estimation without Magnetometer Data*

If no magnetometer data is available in a system, the attitude estimator can still be used without any degradation in the estimation quality of the pitch and roll dimensions by setting $^B\tilde{\mathbf{m}}$ and $^G\mathbf{m}_e$ to zero. In this case, the magnetometer resolution method directly falls through, and the estimation relies solely on the selected yaw-based orientation resolution method. Due to the yaw-zeroing approach that is used, the open-loop yaw produced by the estimator remains stable with each update of $q_y$. The component of $\boldsymbol{\Omega}_y$ in the instantaneous direction of $\mathbf{z}_G$ however, does not have any feedback via the corrective error feedback term $\boldsymbol{\Omega}_e$ in this case, so small constant global yaw velocities in $\hat{q}$ can result. Essentially, this is because the integration in Equation (10.14) of the component of $\boldsymbol{\Omega}_y - \hat{\mathbf{c}}_\Omega$ in the direction of $\mathbf{z}_G$ is then 'unconstrained', leading to yaw drifts. Such yaw drifts are unavoidable without magnetometer data however, as no sensory information is then available that can be used to prevent it.

The effect of the yaw drifts on the output quaternion $\hat{q}$ can be eliminated for convenience by removing the fused yaw component of $\hat{q}$ (to give $\hat{q}_t$), using an equation like Equation (5.139). Note that the fused yaw is the one and only notion of yaw for which it is perfectly valid to perform this operation, as when the fused yaw component of a rotation is removed, the tilt rotation component remains, and tilt rotations have the unique property of having a direct one-to-one correspondence to the set of possible accelerometer-measured gravity directions, as discussed in Sections 5.2.2 and 5.8.1. In addition to not being a strictly valid operation, removing the ZYX Euler yaw has sensitivity issues (see Section 6.2.1), and leads to unexpected behaviour near the not uncommon scenario of pitch rotations by $\frac{\pi}{2}$ radians.

### 10.1.7 **Experimental Results**

Over the years, thorough experimentation and testing of the proposed attitude estimator (and corresponding C++ implementation) has been performed on multiple different platforms, as well as in simulation. In the context of this thesis, we provide results of the attitude estimator running on a real igus Humanoid Open Platform robot (see Section 2.1.2.4), using specifically the fused yaw resolution method, for mathematical and performance reasons.

Video 10.1 shows the visual recording of an experiment in which an igus Humanoid Open Platform robot was manually rotated in the pitch, roll and yaw directions, followed by various combinations of the three, including laying the robot face down on the floor and picking it up again. Two attitude estimators—one allowed to use magnetometer data and one not—were run in parallel to a single instance of the angle estimator that was used by Missura (2015) for the purpose of capture steps. The output of the attitude estimator with magnetometer data is shown in the video by means of a 3D robot model in RViz. The model rotates live relative to the visualised grid frame, to show the current estimated robot orientation at all times. Adjoining live data plots can also be seen, indicating the fused angles output of all three orientation estimation methods at once. It can generally be observed from the video that the estimated RViz orientation is virtually indistinguishable from the true motion of the robot, keeping in mind that the former naturally does not show the effect of any translations.

The data that was captured during Video 10.1 is shown in Figure 10.2. As an initial observation, it can be identified that the fused pitch and roll waveforms are nearly identical for the two instances of the attitude estimator. This is expected, as it was one of the core aims of the estimator that the magnetometer *strictly* only affects the fused yaw component of the estimation, and not the balance-critical fused pitch and roll components. Note that a similar statement for the Euler pitch and roll would be invalid (and not expected), as the ZYX Euler roll parameter contains components of 'yaw', as discussed in Section 6.2.2.1. In contrast to the near-inseparable pitch and roll waveforms, the two estimated yaw waveforms can be observed to have a near-constant offset to each other. This is expected, as the estimator that is not using magnetometer data has no sensor data that allows it to discern absolute heading, so it effectively expresses its estimated yaws relative to the pose the robot was in when it was switched on and the estimator started running. Nonetheless, importantly, it can be seen that the final estimated fused yaws for both attitude estimators are very close to the ones that were estimated at the beginning of the experiment. This is expected, as the robot at the end of the experiment was facing in the approximate same direction as it was facing at the beginning of the experiment.

Initially, as the robot is rotated independently about its local y, x and z-axes, it can be seen that the overwhelming estimation responses come in the fused pitch, roll and yaw parameters, respectively. After this, a complicated rotation sequence ensues that effectively rotates the robot by $\approx 90°$ clockwise (CW). It can be extracted from the yaw estimation data that the two attitude estimators quantified this rotation as $-89.6°$ and $-90.6°$, respectively. Given the general circular non-uniformity of the magnetometer measurements and the somewhat approximate nature of the 90° rotation, these two values are quite close to their expected value. The experiment continues by indirectly rotating the robot back to facing forward, and then tipping it forwards onto the ground. It can be observed in the video that the foot of the robot is inadvertently pulled by the right foot of the experimenter as the robot is being placed on the ground. This can be seen to directly explain the disturbances that appear at time $t = 35.5\,\text{s}$.

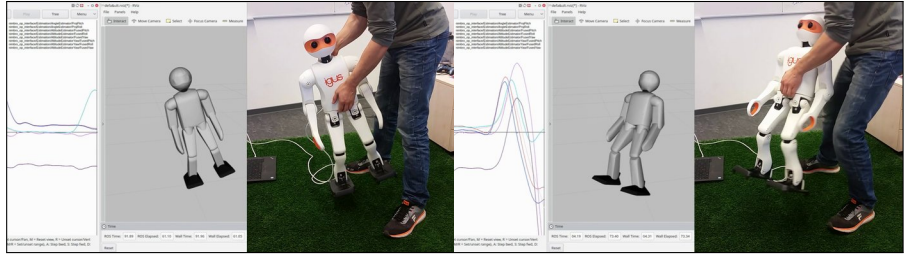

Video 10.1: Recording of the attitude estimation experiment, comparing the
        estimated orientation of the robot in RViz with the real orientation
        of the robot. The captured data is plotted in Figure 10.2.
        https://youtu.be/LEkEiFzAVrE
        *Attitude Estimation Experiment (RViz vs Real)*

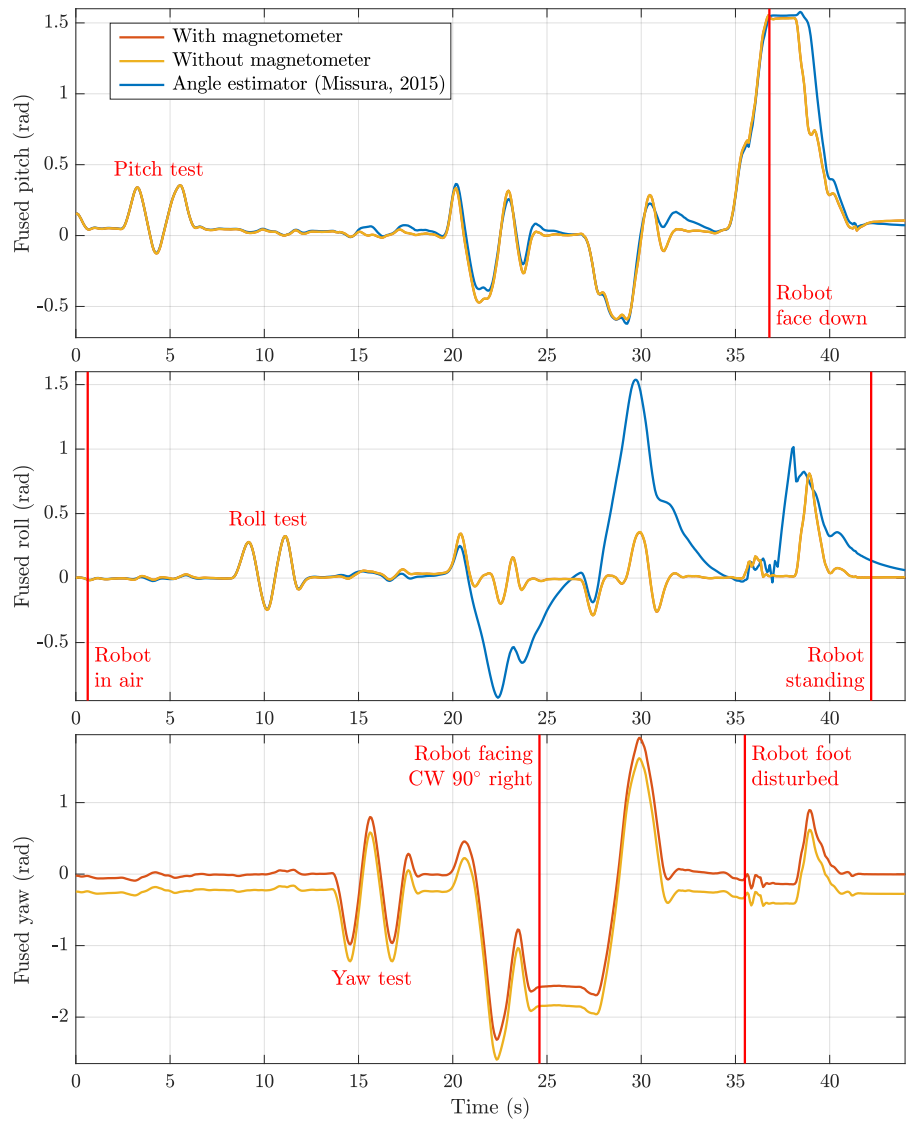

Figure 10.2: Plots of the attitude estimation data captured in Video 10.1. The
         outputs of the attitude estimation with and without magneto-
         meter measurements are compared to the angle estimator used
         in Missura (2015) for the purpose of capture steps.

While the pitch and roll values[5] that were estimated by the angle estimator (Missura, 2015) corresponded quite closely to the values estimated by the attitude estimators for very basic motions (i.e. the pitch and roll tests), as soon as these basic motions were combined into more complex ones, the angle estimator severely struggled. While Video 10.1 visually proves that the pitch and roll values calculated by the attitude estimators are quite accurate, it can be seen that as of $t = 20$ s, the values calculated by the angle estimator severely diverge from these. At $t = 22.4$ s for example, the estimated angle estimator roll is $-53.2°$, and at $t = 29.7$ s the estimated roll even reaches $88.1°$. Looking at the video, these two estimated rolls are clearly nonsensical, especially seeing as they occur *prior* to the robot being laid down on the ground. A different problem can also be identified around the time $t = 37$ s, where the estimated roll becomes somewhat unstable, as evidenced by the sharp jagged edges in the waveform there and the subsequent increase of the estimated roll to $58.2°$ despite the fact that no significant rotation is occurring during that time (see Video 10.1). After the robot is lifted upright again and placed back into a standing position, it can be seen that the angle estimator needs many seconds to recover, as was also the case the previous times the roll had a problem during the experiment. A core problem of the angle estimator is that it fails to capture local z-rotations. This is a non-ignorable issue, as they occur all the time during real walking.

Figure 10.3 shows the results of a further experiment where the robot was rotated $360°$ about the global z-axis, and placed back down on the ground.[6] Once again, a constant offset between the resulting estimated fused yaws was observed, as expected, and the final fused yaws correlated very closely to the initial ones. The sudden jumps in the middle of the plot are attributed to the fused yaws rolling over from $+\pi$ to $-\pi$. Comparing the initial and final estimated fused yaws reveals that the attitude estimator with magnetometer data saw a total change of $-1.2°$, while the estimator without magnetometer data saw a total change of $-0.9°$. These two results are in relative close correlation to each other, and it is thinkable that the true rotation that was undergone by the robot was actually only accurate to $\pm 1°$, seeing as ultimately the performed rotation was judged visually (using a fixed straight edge as a guide) by the experimenter.

While Figure 10.3 demonstrates that the short- to medium-term performance of the estimated fused yaw is reliable and stable, even without magnetometer data, the long-term performance still needs to be ascertained. This has been done in Figure 10.4, which shows the results of a long-term yaw drift test for a robot that was left untouched in a standing position for more than 10 minutes. It can be seen that the attitude estimator with magnetometer data is not prone to yaw drift, as a properly calibrated magnetometer provides an absolute reference of heading that can be used to maintain a consistent estimate of fused yaw. The attitude estimator without magnetometer data however, relies solely on 3D open-loop integration of the gyroscope to provide a reasonable estimate of fused yaw, and is thus vulnerable to drift.[7] Normally, online gyroscope bias autocalibration (see Section 4.2.4) is always enabled, but initially in Figure 10.4 when it was disabled, a yaw drift of $0.031°/s$ ($32.3 s$ per degree) was observed. This is because the gyroscope bias is not truly constant over time, so a one-time gyroscope bias calibration always eventually leads to residual errors that cause yaw drift.[8] When the online gyroscope bias autocalibration scheme was enabled at time $t = 273.4 s$ however, the yaw drift immediately halted, and reduced to below the measurable threshold of $0.0001°/s$ ($0.36°/h$). It is important to note that while the autocalibration scheme only activates when the robot is stationary, it nevertheless ensures that up to the very instant the robot starts moving again, the estimated gyroscope bias is completely correct.

---

5 Note that the angle estimator only estimates pitch and roll, and does not provide any estimate of yaw.

6 Note that the robot was in fact overrotated (around time $t = 13$ s), and brought back to the required pose after that. The final adjustments to the robot happened slowly in the time period from $t = 16$ s to $20$ s.

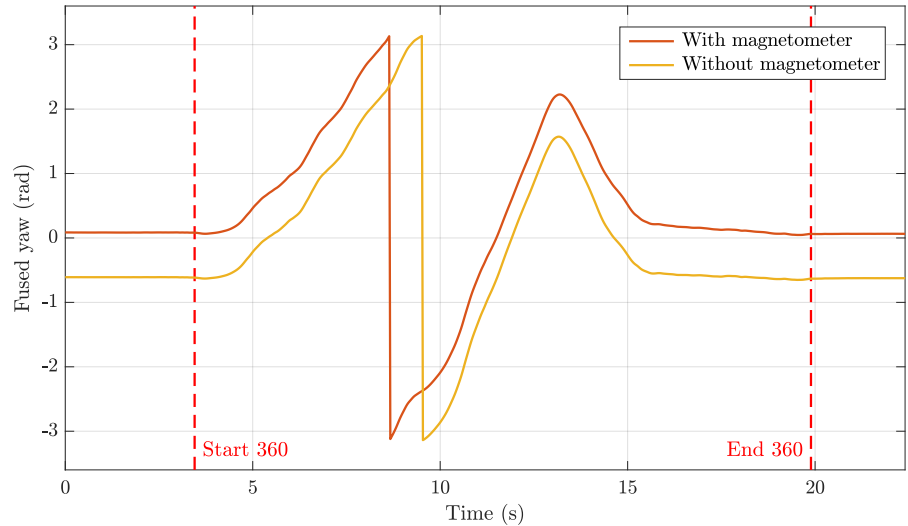

Figure 10.3: Estimated fused yaw with and without magnetometer data for a 360° yaw rotation of the robot. The rotation was performed visually by the experimenter with the aid of a fixed straight edge, and the resulting changes in yaw were measured to be −361.2° and −360.9°, respectively. Given the level of agreement between these results, it is thinkable that the true rotation performed by the experimenter was only accurate to ±1°.

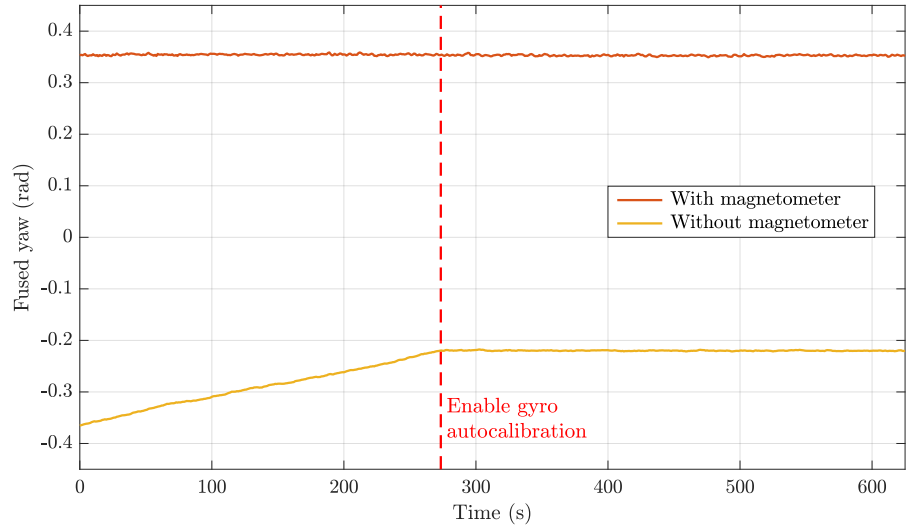

Figure 10.4: Plot of the long-term yaw estimation accuracy of the attitude estimator, demonstrating the possibility of yaw drift in the case of magnetometer-free estimation. The robot was kept stationary in a standing position for the whole experiment, and at $t = 273.4\,\text{s}$ online gyroscope bias autocalibration (see Section 4.2.4) was enabled. While the yaw drift prior to autocalibration was ≈0.031°/s, afterwards the yaw was stable at <0.0001°/s drift.

It can be observed from Figure 10.4 that the fused yaw estimated using magnetometer data is noisier than the fused yaw estimated without. This is because the magnetometer is an inherently noisy sensor, that when the robot is moving actually gets even noisier due to the magnetic and electromagnetic disturbances caused by the servos. Deviations of the local magnetic field from the Earth's magnetic field are also frequent occurrences in buildings, where electrical mains and the presence of larger metal objects and structures are commonplace. Ultimately, the use of magnetometer data in the attitude estimator allows absolute estimates of heading to be made, at the cost of increased noise and decreased short-term stability of the fused yaw.

The effect of quick learning (refer to Section 10.1.6.1) is shown in Figure 10.5. While it is nominally only active in the first few seconds after initialisation, it can also be relevant if the attitude estimator is set to a particular orientation, or reset. In the experiment in the figure, a running attitude estimator was reset twice to the identity rotation, once with quick learning enabled, and once with it disabled. The difference in time that it took for the estimator to converge to the orientation of the robot with quick learning disabled is very apparent, at around $8.8 s$ as compared to $0.36 s$. The effect of the increased gains during quick learning can be seen to temporarily increase the amount of noise in the estimated fused pitch and roll around the $t = 5-6\,\text{s}$ mark, but as $\lambda$ returns to 1 and the gains return to normal, this can be seen to smoothen out again. Note that the robot was being held in the air during the experiment, so it is not expected that the robot's orientation is perfectly constant throughout. With quick learning, the knowledge that the current orientation estimate (after reset) is likely not accurate allows larger gains to be reasonably employed to reduce the convergence time of the estimator, while ensuring all the while that the gains during normal operation can remain small enough that the excessive influence of accelerometer and magnetometer noise can be avoided.

---

7  The estimated fused pitch and roll cannot drift in any situation, as the accelerometer data provides an absolute reference of tilt. The accelerometer data however cannot be used to resolve heading, as a change in heading leaves the measured gravitational acceleration constant.

8  Note that it would be possible to simply 'ignore' gyroscope measurements smaller than a given threshold, and this would be an easy way to 'cheat' an eternally stable fused yaw, but this only helps as long as the robot is completely stationary, and does not do anything to solve the problem that the gyroscope bias is still wrong when the robot subsequently moves.

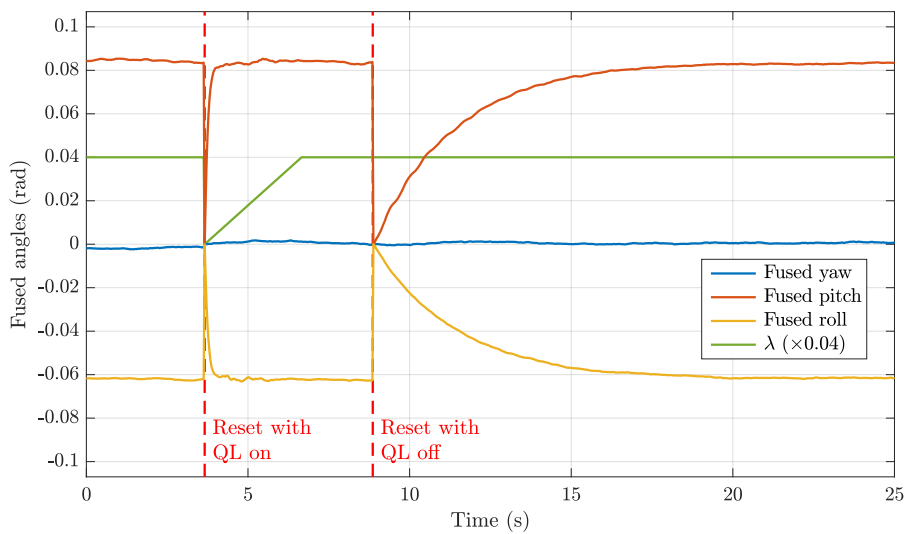

Figure 10.5: The effect of quick learning on the settling time of the attitude estimator. At times $t = 3.66\,\text{s}$ and $t = 8.86\,\text{s}$, the attitude estimator is reset to the identity rotation, and thereby forced to converge again to the true orientation of the robot. Quick learning is enabled ($\lambda = 0$) for the first reset, but disabled for the second. The resulting difference in settling time (0.36 s vs. 8.8 s) is apparent.

The attitude estimator was designed to be able to run at high loop rates on embedded hardware, so as to minimise estimation and possible control feedback latencies. The C++ attitude estimator library code was tested on a Personal Computer (PC) with a 2.40 GHz Intel i5-2430M processor. On a single Central Processing Unit (CPU) core, the average execution time of the estimator over 100 million cycles was found to be 127.6 ns for the magnetometer method, 144.3 ns for the ZYX yaw method, and 112.3 ns for the fused yaw method. It is to be expected that the fused yaw method takes comparatively less time, as it does not in general require a rotation matrix to quaternion conversion, unlike the other two methods. From these results it is confidently anticipated that the algorithm is efficient enough to be implemented at high execution rates on a low-cost microcontroller, where floating and/or fixed point operations are comparatively more expensive than on a PC.

### 10.1.8 Discussion

The presented attitude estimator has been used, completely unchanged, on a wide variety of different robots for many years, including in the demanding environment of the RoboCup competition. In the early years, when magnetometers were still allowed in the competition, the magnetometer resolution method was the main method in use. As of 2016 however, when magnetometers were disallowed, the fused yaw method took its place, and proved to work very well in estimating the pitch and roll of the trunk of the robot at all times. Paired up with the gyroscope scale calibration (see Section 4.2.2) and online gyroscope bias autocalibration scheme (see Section 4.2.4), it was also possible to derive a relatively reliable and low-drift estimate of the yaw, despite it essentially being dependent on open-loop integration. The scale calibration of the gyroscope also contributed in allowing lower values of the proportional gain $k_p$ to be chosen, leading to reduced delay times when tracking sharp changes in orientation, and generally superior transient response characteristics.

One of the most important features of the attitude estimator is that no matter which measured orientation resolution method is used, the filter is equally stable in all global body orientations, and only demonstrates potential non-convergent behaviour on a pathological set that is of no practical concern. Extensions to the filter also allow for reliable attitude estimation in situations of reduced sensory data, and the quick learning feature allows for shorter settling times from large estimation errors when required.

The output of the estimator can be used flexibly depending on the situation, in the form of either

- $\hat{q} \in \mathbb{Q}$, the full quaternion orientation estimate of the robot,

- $\hat{q}_t \in \mathbb{Q}$, the (pitch/roll) tilt rotation component of the robot, expressed as a zero z-component quaternion,

- $\hat{F} = (\psi, \theta, \phi, h) \in \mathbb{F}$, the individual fused angles parameters of $\hat{q}$ (see Section 5.4.4), or,

- $\hat{P} = (p_x, p_y, p_z) \in \mathbb{P}^3$, the individual tilt phase space parameters of $\hat{q}$ (see Section 5.4.5).

In these various forms, the estimated orientation output of the attitude estimator can be used for all kinds of feedback purposes, including in particular the analysis and control of balancing biped robots.

## 10.2  POSE ESTIMATION

The most direct application of the estimated orientation of the robot is for the purpose of pose estimation. This involves estimating the position and velocity of the CoM of the robot, in addition to how far this CoM point moves over time as the robot takes steps.

### 10.2.1  CoM State Estimation

Given the joint space humanoid kinematics model introduced in Section 9.1.1, the measured joint positions $\hat{\mathbf{q}}$ of the robot, and the estimated orientation ${}^{G}_{B}\hat{q} \in \mathbb{Q}$ of the trunk, the pose and orientation of the robot can be modelled in 3D (see Figure 10.6) and used to calculate the Cartesian coordinates of the CoM relative to the support foot. Each foot has a *leg tip frame* {F} defined for it by the joint kinematic model (see Figure 9.1). The position and orientation of this frame relative to the body-fixed frame {B} (located at the hip centre point) can be calculated for each leg by converting the measured joint positions $\hat{\mathbf{q}}$ for that leg to the corresponding leg tip pose

$$(\mathbf{t}_l, q_l) \in \mathbb{R}^3 \times \mathbb{Q}, \tag{10.37}$$

where

$$q_l \equiv {}^{B}_{F}q_l. \tag{10.38}$$

The Cartesian position of the leg tip point, i.e. the origin of {F}, relative to frame {B} is then

$${}^{B}\mathbf{p}_F = \mathbf{t}_l + \left(h_x, \, \delta(h_y + \tfrac{1}{2}h_w), \, -2L_l\right). \tag{10.39}$$

where $\delta \in \{-1, 1\}$ is the limb sign as per Equation (9.17). Given the estimated orientation ${}^{G}_{B}\hat{q}$ of the trunk from the attitude estimation, we conclude that

$${}^{G}\mathbf{p}_{BF} = {}^{G}_{B}\hat{q} \, {}^{B}\mathbf{p}_F, \tag{10.40}$$

where ${}^{G}\mathbf{p}_{BF}$ is the displacement from the origin of frame {B} to the origin of frame {F} relative to the global reference frame {G}. As the ground is assumed to be flat and parallel to the $\mathbf{x}_G\mathbf{y}_G$ plane, the leg that has the more negative z-component of ${}^{G}\mathbf{p}_{BF}$ is clearly the one that is more likely to be the main current support foot. As such, we define the support leg sign $\delta_s \in \{-1, 1\}$ to be the limb sign of the leg with the lower ${}^{G}\mathbf{p}_{BF}$ z-value. Hysteresis is used for the support leg transitions to ensure that $\delta_s$ remains stable close to support transitions.

Given the current detected support leg (as defined by $\delta_s$), the *support foot floor frame* {F̃} is defined to be the frame that results when {F} is rotated in the shortest and most direct way so that it becomes flat relative to the ground, i.e.

$$\mathbf{z}_{\tilde{F}} = \mathbf{z}_G. \tag{10.41}$$

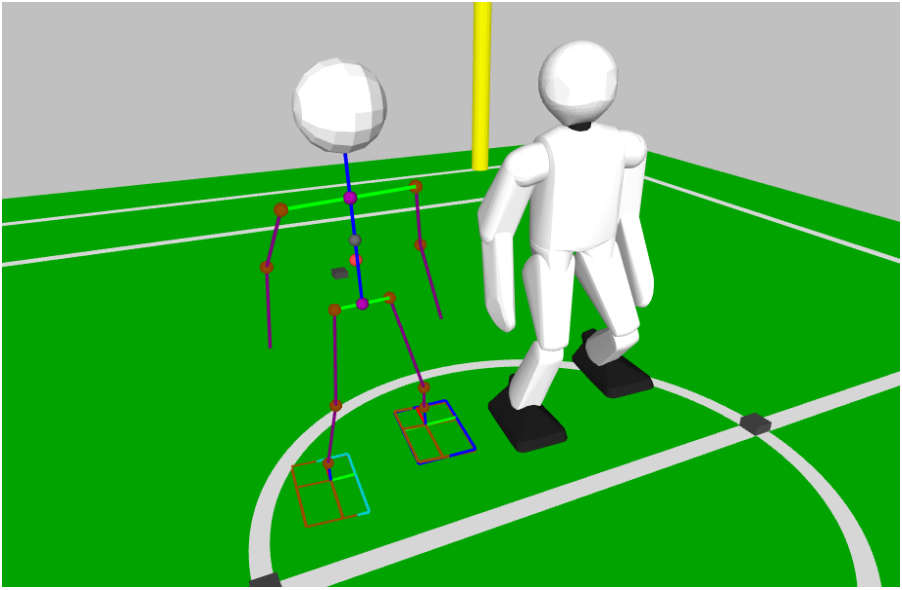

Figure 10.6: A simplified visualisation of the humanoid kinematics model of the igus Humanoid Open Platform, used for the purposes of CoM state estimation and gait odometry estimation via the stepping motion model. The assumed CoM point is marked by the orange sphere, and the current support foot floor frame is marked in cyan (only partially visible in the image, see the inner half of the modelled right foot). The support foot floor frame is the result of removing the tilt rotation component of the orientation of the leg tip frame {F} of the support foot, so as to make it parallel to the assumed flat ground.

This is equivalent to defining $_F^Gq$ as the pure yaw quaternion that results when the tilt rotation component of

$$_F^Gq = {}_B^G\hat{q}\,{}_F^Bq_l \tag{10.42}$$

is removed. One can think of this operation as 'tilting' {F} about an axis in the $\mathbf{x}_G\mathbf{y}_G$ plane until it is flat. With $_F^Gq$ known, the orientation of the CoM frame {C} (see Figure 9.1) relative to the support foot floor frame can be calculated to be

$$_C^{\tilde{F}}q = {}_F^Gq^*\,{}_B^G\hat{q}. \tag{10.43}$$

Observing that

$$^B\mathbf{p}_{\tilde{F}} = {}^B\mathbf{p}_F, \tag{10.44a}$$

$$^B\mathbf{p}_C = (c_x,\,0,\,c_z), \tag{10.44b}$$

the displacement of the CoM relative to the support foot floor frame is then

$$^{\tilde{F}}\mathbf{p}_C = L_{_C^{\tilde{F}}q}\big(^B\mathbf{p}_C - {}^B\mathbf{p}_F\big). \tag{10.45}$$

In this way, the position $^{\bar{F}}\mathbf{p}_C$ and orientation $^{\bar{F}}_C\hat{q}$ of the modelled CoM point relative to the support foot can be calculated in each cycle, in addition to the estimated support leg sign $\delta_s$.

In some applications, like for example the capture step gait in Chapter 12, estimates of not only the CoM positions, but also the CoM velocities are required. This is achieved by passing $^{\bar{F}}\mathbf{p}_C$ through a Savitzky-Golay derivative filter (see Appendix A.2.1.2). An equivalently sized Savitzky-Golay smoothing filter is also used to ensure that the CoM position trajectories are suitably noise-free, and match up in terms of induced delay with the velocity waveforms. Together, the outputs of the two Golay filters, in addition to the support limb sign $\delta_s$, form the output of the CoM state estimation.

### 10.2.2 Stepping Motion Model

If we assume that the support foot floor frame $\{\bar{\mathrm{F}}\}$ is fixed relative to the ground for as long as the corresponding foot is the support leg, we can begin to model the long-term walking odometry of the robot using the kinematic model that was used for the CoM state estimation. In each cycle, when a new estimated trunk orientation $^G_B\hat{q}$ and measured joint position vector $\hat{\mathbf{q}}$ become available, the modelled pose of the robot is updated so that the orientation of the body-fixed frame $\{\mathrm{B}\}$ becomes the new $^G_B\hat{q}$, and the positions of the joints become the new $\hat{\mathbf{q}}$. A corrective global yaw rotation and 3D translational shift are then applied to the model to ensure that the support foot floor frame $\{\bar{\mathrm{F}}\}$ remains completely unchanged during this entire update procedure. If as a result of the update the free foot ends up lower in height than the support foot, with hysteresis a support foot exchange is triggered, as described in Section 10.2.1, and the robot will begin to be modelled to move relative to the other foot instead. This entire procedure is known as the stepping motion model, and is illustrated in Figure 10.7. Note that the support foot link (i.e. frame $\{\mathrm{F}\}$) *can* rotate relative to the support foot floor frame $\{\bar{\mathrm{F}}\}$, but only in the form of a tilt rotation. Nevertheless, if the sensor measurements are accurate and well-calibrated, we expect that most of the time the deviation between $\{\mathrm{F}\}$ and $\{\bar{\mathrm{F}}\}$ will only be slight, due to the semi-rigid contact of the foot with the floor.

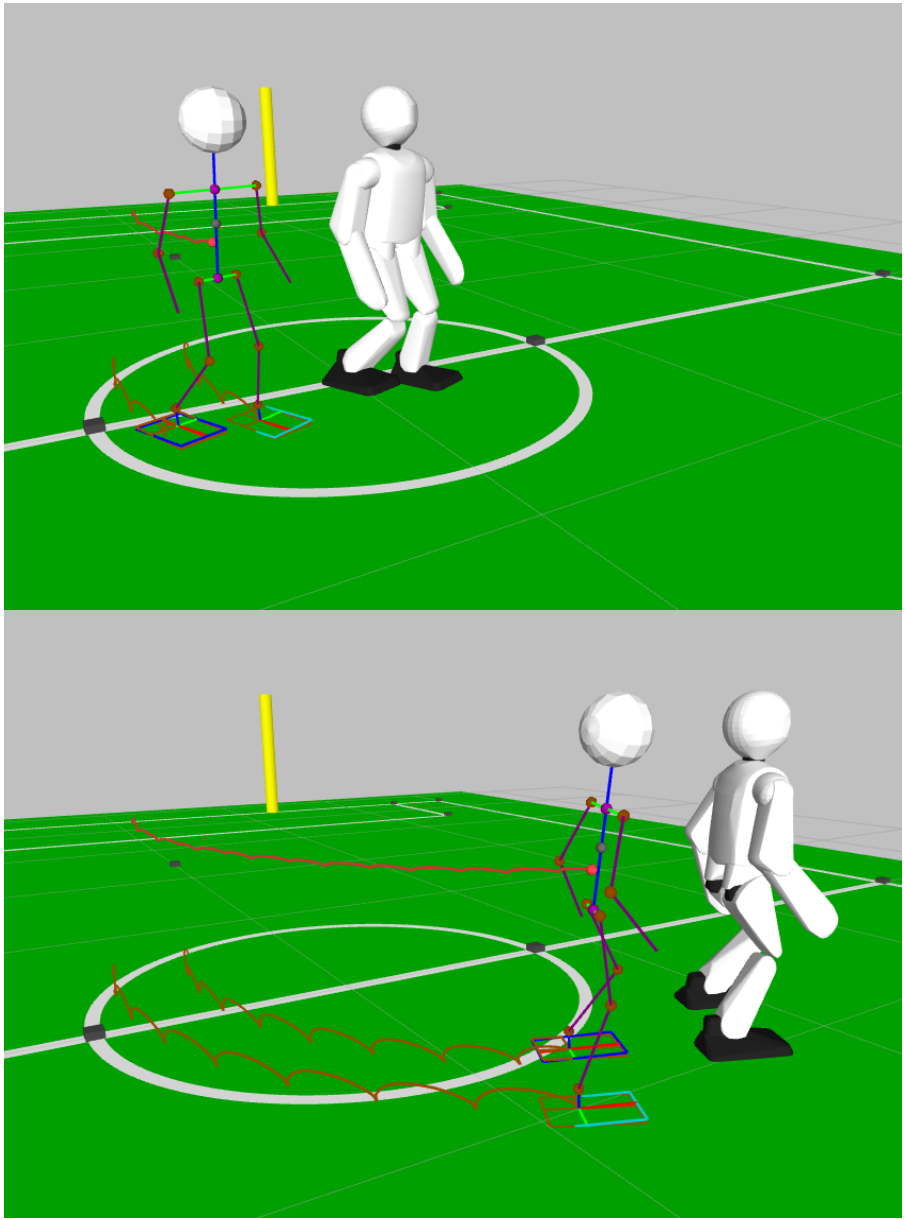

Figure 10.7: Demonstration of the stepping motion model for the purpose of dead reckoning of the gait odometry. The upper trail shows the time history of the CoM point, and the two lower trails show the time histories of the leg tip points. For visualisation purposes, the mesh model of the robot is shown at a constant global y-offset to the kinematic model.

# A CENTRAL PATTERN GENERATOR FOR WALKING

For most bipedal robot platforms, completely open-loop walking—for at least short periods of time—is reasonably possible with a moderate level of stability. This is because it is usually possible to design walking motions that take advantage of the small basin of passive stability provided by the foot stiffnesses and support polygons. The Central Pattern Generator (CPG) is one such walking motion, and is sufficiently generic and tuneable that it has proven to work, and be effective on, a wide range of different robots. The general approach to walking used by the bipedal gaits in this thesis is to start with an open-loop gait core, and build around it stabilising feedback mechanisms that adjust the gait step sizes, timing, and other specifics of the generated waveforms. The CPG that is introduced in this chapter is used as the basis of the capture step gait (Chapter 12) and the direct fused angle feedback controller gait (Chapter 13), and has been implemented in the `cap_gait` gait engine of the Humanoid Open Platform ROS Software.[1]

The CPG is based on the general ideas that were present in the CPG used by Missura (2015), but with many modifications and improvements that were mainly intended to allow the generator to be more flexible and work on a wider variety of robots. The main differences to the older CPG, also described in Missura and Behnke (2013b), include:

- Changes to the leg retraction profiles to transition more smoothly between swing and support phases,

- The addition of a double support phase for greater walking stability and passive oscillation damping,

- The addition of a trim factor for the angle relative to the ground at which the feet are lifted during stepping,

- The integration of a dynamic pose blending algorithm to enable smoother transitions to and from walking,

- The incorporation of support coefficient waveforms, for use with the actuator control scheme (see Section 3.2),

- The introduction of a leaning strategy based on the rate of change of the commanded gait velocity, and

- The use of hip motions instead of leg angle motions for the gait command velocity-based leaning strategies.

---

1 https://github.com/AIS-Bonn/humanoid_op_ros/blob/master/src/nimbro/motion/gait_engines/cap_gait/src/cap_gait.cpp

## 11.1   CPG GAIT INTERFACES

The Central Pattern Generator (CPG) in its purest form can simply be considered to be a complex high-dimensional function with certain inputs and outputs, i.e. a function like

$$f_{CPG} : \{\mathbf{v}_g, \mu_i\} \mapsto \{\mathbf{q}_o, \boldsymbol{\xi}, \kappa_l, \kappa_r\} \tag{11.1}$$

where the individual input and output variables are as described in the following subsections.

### 11.1.1   CPG Gait Inputs

One of the two inputs to the open-loop CPG gait is the dimensionless Gait Command Velocity (GCV) vector

$$\mathbf{v}_g = (v_{gx}, v_{gy}, v_{gz}) \in [-1, 1]^3. \tag{11.2}$$

The components of the input GCV vector $\mathbf{v}_g$ specify the desired walking speed of the robot in the sagittal, lateral and yaw rotation directions respectively, as a ratio of the corresponding maximum allowed speeds. For instance, full speed forwards walking corresponds to a GCV of $\mathbf{v}_g = (1, 0, 0)$. The raw input GCV vector is first scaled, if required, to ensure that its p-norm satisfies

$$\|\mathbf{v}_g\|_p = \left( |v_{gx}|^p + |v_{gy}|^p + |v_{gz}|^p \right)^{\frac{1}{p}} \leq 1, \tag{11.3}$$

and the resulting vector is slope-limited to ensure that no sharp changes in walking speed can occur, yielding the internal GCV vector

$$\mathbf{v}_i = (v_{ix}, v_{iy}, v_{iz}). \tag{11.4}$$

In general, a value of $p$ in the range $[2, 2.5]$ was found to be suitable for the tested robots. The p-norm limit ensures, for stability reasons, that the robot cannot for instance try to walk full speed forwards and sidewards at the same time, and the slope limit ensures that no jumps or step changes can occur in the CPG outputs.

   The second input to the CPG gait is the gait phase variable

$$\mu_i \in (-\pi, \pi]. \tag{11.5}$$

This variable parameterises the gait cycle that is generated by the CPG in a direct and deterministic way, and for regular open-loop walking starts at 0 or $\pi$ at the beginning of walking, and is updated after each time step $\Delta t$ using

$$\mu_i \leftarrow \text{wrap}(\mu_i + \pi f_g \Delta t), \tag{11.6}$$

where $f_g$ is the configured nominal gait frequency in units of steps per second. The wrap($\cdot$) function ensures that the gait phase stays strictly in the range $(-\pi, \pi]$ at all times. As shown in Figure 11.1, each value of $\mu_i$ corresponds to a fixed instant of the gait cycle. For example,

$$\mu_i = 0 \;\Rightarrow\; \text{Foot strike of the right leg, beginning of double support}$$

$$\mu_i = \pi \;\Rightarrow\; \text{Foot strike of the left leg, beginning of double support}$$

It should be noted that by convention, the swing phase of the left leg and support phase of the right leg occur in the phase range $\mu_i \in (0, \pi]$, and the *support phase* of the left leg and *swing phase* of the right leg occur in the phase range $\mu_i \in (-\pi, 0]$. As the arms generally swing opposite to the legs during walking, the swing and support phase definitions for the arms are exactly opposite to those for the legs.

To simplify the situation for the trajectory generation calculations, the concept of limb phase is introduced, which for each individual limb is defined to be

$$\nu_i = \begin{cases} \mu_i & \text{if the swing phase is in } (0, \pi], \\ \text{wrap}(\mu_i + \pi) & \text{if the swing phase is in } (-\pi, 0]. \end{cases} \quad (11.7)$$

Essentially, $\nu_i$ is the same as the gait phase $\mu_i$ for each limb, just possibly shifted by $\pi$ radians to ensure that the swing phase is always in the range $\nu_i \in (0, \pi]$. The synchronisation of the swing phases with respect to the limb phases $\nu_i$ allows the mathematical formulas behind the CPG gait to be more easily and efficiently stated and calculated.

### 11.1.2 CPG Gait Outputs

The outputs of the CPG gait are exactly those required by the actuator control scheme (see Section 3.2). As described in Section 3.1, in each execution cycle this corresponds to

- The vector of joint position commands $\mathbf{q}_o \in \mathbb{R}^N$, specifying the desired angular position of each joint (for $N$ joints),

- The vector of joint effort commands $\boldsymbol{\xi} \in [0, 1]^N$, specifying how stiff each joint should be, and,

- The required support coefficients $\kappa_l$ and $\kappa_r$ of the left and right legs respectively, specifying the proportion of the weight of the robot that is expected to be supported by each leg.

The vector $\mathbf{q}_o$ can be seen to correspond directly to the desired joint space pose of the robot (see Section 9.1.1), and like the support coefficients $\kappa_l$ and $\kappa_r$, it is calculated dynamically based on the GCV and gait phase inputs. The joint effort command vector $\boldsymbol{\xi}$, on the other hand, is generally assigned a manually configured constant value, and is thus trivial to compute in each cycle.

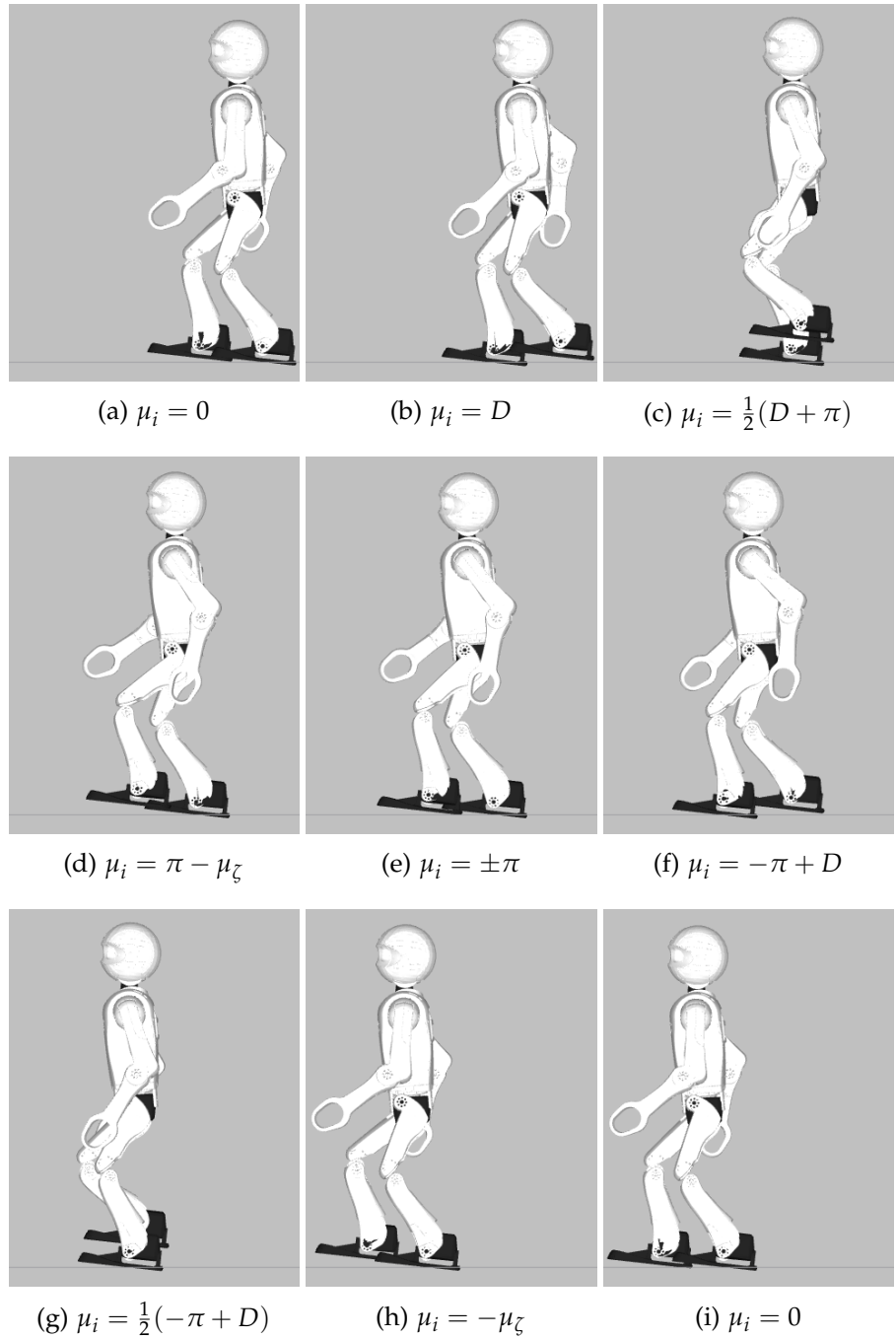

(a) $\mu_i = 0$    (b) $\mu_i = D$    (c) $\mu_i = \frac{1}{2}(D + \pi)$

(d) $\mu_i = \pi - \mu_\zeta$    (e) $\mu_i = \pm\pi$    (f) $\mu_i = -\pi + D$

(g) $\mu_i = \frac{1}{2}(-\pi + D)$    (h) $\mu_i = -\mu_\zeta$    (i) $\mu_i = 0$

Figure 11.1: Snapshots of full speed forwards walking of the CPG gait at various gait phases. (a)–(b) and (e)–(f) correspond to the two double support phases, during which the relative height of the feet changes only very little. (c) and (g) correspond to the points of maximum foot lifting, and (d) and (h) correspond to the point of maximum forwards swing. (i) shows the robot back to a gait phase of zero, ready to take the next two steps. $D$ is the double support phase length, and $\mu_\zeta$ is the swing stop phase offset.

### 11.1.3  Provisions for Closed-loop Feedback

As discussed at the beginning of this chapter, we wish to construct the CPG in such a way that it is possible for higher level balance controllers to add stabilising feedback to the generated walking motions, with the aim of making them fundamentally more robust. The CPG thus makes provisions in particular for the inclusion of timing and step size feedback.

#### 11.1.3.1  *Timing Feedback*

A balance feedback controller can apply timing feedback to the CPG through the specification of a gait frequency offset $f_{go}$, time to step $t_s$, and corresponding support leg sign $\delta_t \in \{-1, 1\}$.[2] The $t_s$ parameter specifies the amount of time that should be taken by the CPG until the foot opposite to $\delta_t$ should next be commanded to strike the ground (i.e. reach the gait instants shown in Figures 11.1e and 11.1i). The current support foot according to the current gait phase $\mu_i$ is given by

$$\delta_c = \begin{cases} -1 & \text{if } \mu_i \in (0, \pi], \\ 1 & \text{if } \mu_i \in (-\pi, 0]. \end{cases} \tag{11.8}$$

Thus, the amount of gait phase $\mu_r \in [0, 2\pi)$ that needs to be covered in the specified duration of $t_s$ seconds is

$$\mu_r = \begin{cases} -\mu_i & \text{if } \delta_t = \delta_c = 1, \\ \pi - \mu_i & \text{if } \delta_t = -1, \\ 2\pi - \mu_i & \text{if } \delta_t = 1 \text{ and } \delta_c = -1. \end{cases} \tag{11.9}$$

This corresponds to an instantaneous gait frequency, in steps per second, of

$$\tilde{f}_g = \frac{\mu_r}{\pi t_s} + f_{go}, \tag{11.10}$$

where the effect of the gait frequency offset parameter $f_{go}$ has also been included. After application of a maximum gait frequency bound $f_{g,max}$, i.e.

$$f_g = \text{coerce}\big(\tilde{f}_g, 0, f_{g,max}\big), \tag{11.11}$$

Equation (11.6) can be used as normal to update the gait phase in each cycle. The final instantaneous predicted time to step is

$$\hat{t}_s = \frac{\mu_r}{\pi f_g}. \tag{11.12}$$

Note that a feedback controller may return a different $t_s$ in each cycle, so the amount that $\mu_i$ is incremented by with each update may vary continuously, and differ at each instant throughout a single step.

---

2 See Equation (9.17) for the definition of the *limb sign* of a leg.

### 11.1.3.2  *Step Size Feedback*

In the framework of the CPG, step size feedback is realised via modification of the internal GCV vector $\mathbf{v}_i$. Balance feedback controllers can either simply apply additive offsets to the value of $\mathbf{v}_i$, or intercept its calculation prior to slope limiting (but after p-norm limiting), and calculate an arbitrary desired value $\tilde{\mathbf{v}}_i$ based on that intermediate value and the current balance state. In the latter case, the slope limiting is skipped because otherwise the ability of the controller to make quick balance recovery steps is inhibited. To prevent discontinuities in the gait outputs from occurring as a result, the internal GCV vector $\mathbf{v}_i$ is made persistent and updated in each cycle using the equation

$$\mathbf{v}_i \leftarrow \mathbf{v}_i + \frac{\Delta t}{\hat{t}_s}(\tilde{\mathbf{v}}_i - \mathbf{v}_i),  \tag{11.13}$$

where $\hat{t}_s \geq \Delta t$ is the final predicted time to step as in Equation (11.12), and $\Delta t$ is the nominal cycle time. This equation effectively adjusts the current internal GCV so that it reaches $\tilde{\mathbf{v}}_i$ exactly at the end of the current step.

No matter which method of step size adjustment is used, the final value of $\mathbf{v}_i$ is the one that is passed on to the internals of the CPG, and used to generate the required joint waveforms.

## 11.2  CPG MOTION GENERATION

The walking motions generated by the Central Pattern Generator (CPG) are based around a configured halt pose of the robot. This is a manually tuned pose that corresponds to a centred and balanced upright standing posture of the robot, as shown in Figure 11.2. The halt pose is usually configured in terms of the abstract space, as it is more intuitive to make adjustments in that space, but can easily be converted to the joint pose if required for certain calculations. When walking of the robot is triggered, the CPG first checks whether the robot is in a standing posture, and if so, triggers the following series of events:

(a) **Pre-gait blending** is applied to bring the robot to the required halt pose. This involves trapezoidal velocity spline interpolation (see Appendix A.1.5.2) for the joint position commands, and linear spline interpolation (see Appendix A.1.5.1) for the joint effort commands and support coefficients.

(b) The **gait phase** is initialised to 0 or $\pi$ depending on which leg should step first, and generation of the CPG motion waveforms starts. The input GCV $\mathbf{v}_g$ is forced to be zero for the first 2 or so steps, to make sure that the robot can reach its regular walking limit cycle before starting to walk away from its current location.

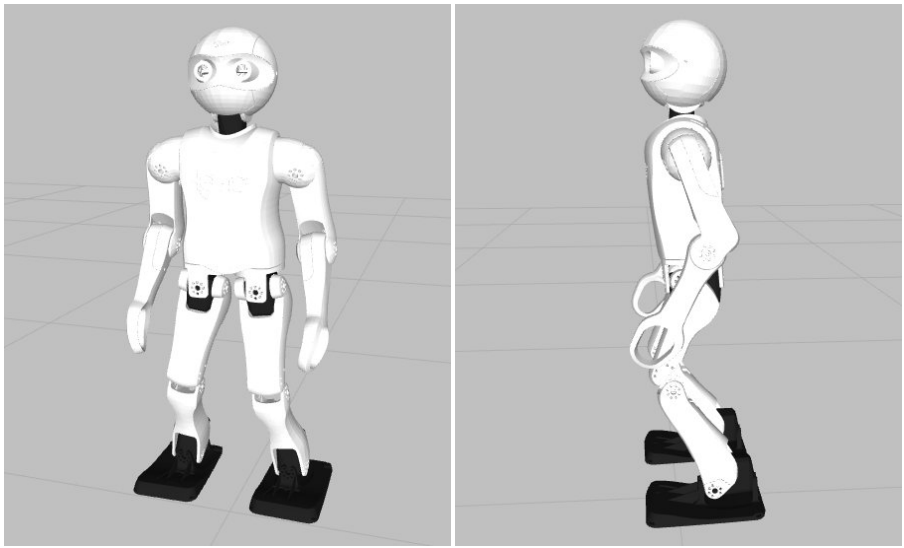

Figure 11.2: Example of the CPG gait halt pose used for the igus Humanoid Open Platform.

(c) Smooth sinusoidal **pose blending** is applied over the duration of the very first step to smoothly fade the commanded outputs from the halt pose to the generated CPG waveforms.

(d) The robot **walks** with the CPG waveforms for as long as required until the trigger to stop walking comes.

(e) When the robot is triggered that it should **stop walking**—and at least two steps have been taken—it forces the input GCV $\mathbf{v}_g$ to zero, and waits for the internal GCV $\mathbf{v}_i$ to become zero as well as a result, or at least suitably close to zero. The generation of the CPG waveforms is then halted the next time the gait phase passes either $0$ or $\pi$.

(f) Smooth sinusoidal **pose blending** is applied over a duration equivalent to the time of one step, to smoothly fade the commanded outputs of the CPG waveforms back to the halt pose.

When the final pose blending back to the halt pose finishes, the CPG gait signals that the robot is no longer walking, and allows other motion modules to take over control of the robot if necessary.

The specifics of how the CPG waveforms are generated as a function of the internal GCV vector $\mathbf{v}_i$ and gait phase $\mu_i$ are addressed separately for the leg and arm motion profiles in the following sections.

### 11.2.1 Leg Motion Profile

The generation of the CPG leg motions begins for each leg with the configured halt pose in the abstract space, and continues as follows:

(i) The halt pose of the robot is additively modified by the **abstract leg motion** function, which incorporates the required basic walking motion elements, such as for example leg lifting, leg swing and hip swing.

(ii) The abstract pose is **soft-coerced** (see Appendix A.1.2.3) to safe limits to ensure that no excessive internal GCV or faulty balance feedback controller can cause dangerous motions of the robot.

(iii) The coerced abstract pose is converted to the **inverse space** and additively modified by the **inverse leg motion** function.

(iv) The resulting inverse pose is converted to the **joint space** to give the final CPG joint position waveforms—at least prior to the effect, for instance, of the pose blending described in Steps (c) and (f) of the list in the previous section.

(v) The required **support coefficient** waveforms are calculated as a function of the gait/limb phase.

Recall that the joint effort commands are kept constant at suitably configured values, and so do not need to be dealt with explicitly here.

The entire leg motion profile is a direct function of the internal GCV vector $\mathbf{v}_i = (v_{ix}, v_{iy}, v_{iz})$ and the gait phase $\mu_i$. Depending on which leg the motion is being generated for however, the required support phase may either be in the range $\mu_i \in (-\pi, 0]$ or $\mu_i \in (0, \pi]$. In order to avoid unnecessary complications, we express the leg motion profile in terms of the limb phase $\nu_i$ instead of the gait phase $\mu_i$. Recall from Equation (11.7) that the limb phase is $\pi$ radians out of phase for the two legs, and is defined so that the swing and support phases of both legs occur at exactly the same $\nu_i$. This greatly simplifies the equations for the leg motion profile.

Table 11.1 shows the various limb phase keypoints that are used in the generation of the leg motions. $D$ is the configured length of each double support phase in units of gait phase, and $\mu_\zeta$, known as the *swing stop phase offset*, is the amount of gait phase prior to foot strike that the leg swing waveform should reach its maximum. Cross-references to the images in Figure 11.1 are provided in the table for visualisation purposes. In each case, having a limb phase of $\nu_i$ corresponds to the state of the left leg in the image corresponding to the gait phase of $\mu_i = \nu_i$. Illustrative details of how the generated leg waveforms relate to the limb phase keypoints can be found later in Figure 11.3.

### 11.2.1.1    *Abstract Leg Motion*

As indicated in Step (i) of the leg motion profile, the *abstract leg motion* function takes as input the configured halt pose of the robot in the abstract space representation, and for each leg applies additive modifications to the individual abstract leg parameters[3]

$$\boldsymbol{\Phi}_l = (\phi_{lx}, \phi_{ly}, \phi_{lz}, \phi_{fx}, \phi_{fy}, \epsilon_l).$$ (11.14)

Table 11.1: Characterisations of the limb phase keypoints of the CPG

| Fig. | Limb phase ($\nu_i$) | Characterisation |
|------|----------------------|------------------|
| (a) | 0 | Support phase $\to$ Double support phase |
| (b) | $D$ | Double support phase $\to$ Swing phase <br> Start of foot lifting and leg swing |
| (c) | $\frac{1}{2}(D + \pi)$ | Point of maximum foot lifting |
| (d) | $\pi - \mu_\zeta$ | End of leg swing |
| (e) | $\pm\pi$ | End of foot lifting <br> Swing phase $\to$ Double support phase |
| (f) | $-\pi + D$ | Double support phase $\to$ Support phase |
| (g) | $\frac{1}{2}(-\pi + D)$ | Middle of support phase |

LEG LIFTING  The first modifications to be applied are that of leg lifting and leg pushing. During the swing phase, a sinusoidal waveform is applied to the leg retraction parameter to lift the foot and place it back down again at $\nu_i = \pi$. Similarly, during the support phase, a sinusoidal waveform is applied to the leg retraction parameter to push the foot into the ground and have it return at $\nu_i = 0$. In terms of mathematical equations, this can be expressed as

$$\epsilon_l \leftarrow \epsilon_l + \Delta\epsilon_{ll},$$ (11.15)

where

$$\Delta\tilde{\epsilon}_{ll} = \begin{cases} \epsilon_{sh} \sin\left(\frac{\pi(\nu_i - D)}{\pi - D}\right) & \text{if } \nu_i \in [D, \pi], \\ -\epsilon_{ph} \sin\left(\frac{\pi(\nu_i - D + \pi)}{\pi - D}\right) & \text{if } \nu_i \in [-\pi + D, 0], \\ 0 & \text{otherwise,} \end{cases}$$ (11.16)

and $\Delta\epsilon_{ll}$ is obtained from $\Delta\tilde{\epsilon}_{ll}$ by applying *sine function fillets* to the lifting and pushing half-sinusoids (see Appendix A.1.4.4, and the difference between the solid and dotted red lines in Figure 11.3). Note that $\epsilon_{sh}$ is the desired dimensionless *step height*, and $\epsilon_{ph}$ is the desired dimensionless *push height*, where

$$\epsilon_{sh} = k_{sh} + k_{shx}|\nu_{ix}| + k_{shy}|\nu_{iy}|,$$ (11.17a)

$$\epsilon_{ph} = k_{ph} + k_{phx}|\nu_{ix}|,$$ (11.17b)

---

3 See Section 9.1.2 for definitions of the abstract space parameters.

for some configured constants $k_*$. As a result of Equation (11.17), the robot lifts its feet higher the faster it walks. In addition to the modification of the leg retraction parameter, a secondary leg lifting component is applied to the leg angle Y, i.e.

$$\phi_{ly} \leftarrow \phi_{ly} + k_{lly}\Delta\epsilon_{ll}, \tag{11.18}$$

for some gain $k_{lly}$. The purpose of this secondary component is to allow adjustment of the angle at which the robot lifts its feet and places them back down on the ground again. This allows for a certain level of fine-tuning of the leg swing profile of the robot.

LEG SWING    Leg swing is the component of the CPG waveforms that allows the robot to take steps in the sagittal, lateral and rotational directions as opposed to just walking on the spot. It is based on a dimensionless sinusoid-linear notion of swing that is defined mathematically as

$$\zeta_s = \begin{cases} -\cos\left(\frac{\pi(\nu_i - D)}{\pi - \mu_\zeta - D}\right) & \text{if } \nu_i \in (D, \pi - \mu_\zeta], \\ 1 - 2\left(\frac{\nu_i - \pi + \mu_\zeta}{\pi + \mu_\zeta + D}\right) & \text{if } \nu_i \in (\pi - \mu_\zeta, \pi], \\ 1 - 2\left(\frac{\nu_i + \pi + \mu_\zeta}{\pi + \mu_\zeta + D}\right) & \text{if } \nu_i \in (-\pi, D], \end{cases} \tag{11.19}$$

where $\zeta_s$ is referred to as the *swing angle*. Figure 11.3 shows a plot of the swing angle $\zeta_s$ against the limb phase $\nu_i$, and indicates how the quick sinusoidal forwards swing during the swing phase is alternated with the slow linear backwards swing during the single and double support phases. Given the dimensionless swing angle $\zeta_s$, the leg swing components in the lateral, sagittal and rotational directions are applied, respectively, using the formulas

$$\phi_{lx} \leftarrow \phi_{lx} + k_{sy}v_{iy}\zeta_s, \tag{11.20a}$$

$$\phi_{ly} \leftarrow \phi_{ly} - k_{sx}v_{ix}\zeta_s, \tag{11.20b}$$

$$\phi_{lz} \leftarrow \phi_{lz} + k_{sz}v_{iz}\zeta_s, \tag{11.20c}$$

In order to avoid collisions between the feet during the double support phases, the lateral and rotational separations of the feet are increased with increasing GCV. This is referred to as *lateral pushout* and *rotational V-pushout*, and is applied using the formulas

$$\phi_{lx} \leftarrow \phi_{lx} + \delta\big(k_{lpx}|v_{ix}| + k_{lpy}|v_{iy}| + k_{lpz}|v_{iz}|\big), \tag{11.21a}$$

$$\phi_{lz} \leftarrow \phi_{lz} + \delta\big(k_{rpz}|v_{iz}|\big), \tag{11.21b}$$

where $\delta \in \{-1, 1\}$ is the limb sign,[4] and $k_*$ are configured constants. Together, Equations (11.20) and (11.21) complete the contribution of leg swing to the generated CPG waveforms.

---

4 As with the other leg motion formulas, Equation (11.21) is applied to both legs, but in this case requires an adjustment in the opposite direction for each leg.

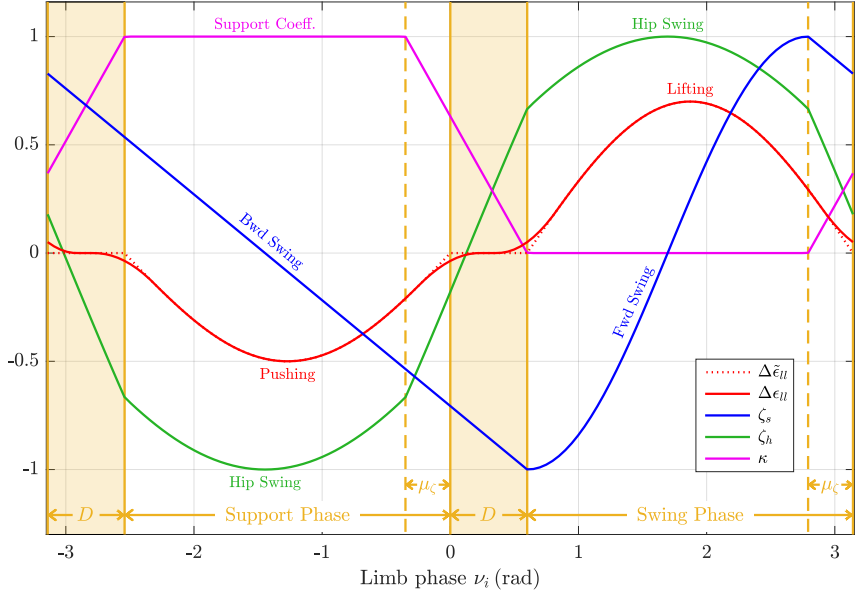

Figure 11.3: Plots of the abstract space CPG waveform components against the limb phase $\nu_i$ for the left leg. The waveforms for the right leg are exactly identical, only that the hip swing is shifted in phase by $\pi$ radians. Some parameters of the waveforms have been exaggerated in the plot for visualisation purposes, like for example $\epsilon_{sh}$, $\epsilon_{ph}$ and $\mu_\zeta$.

**HIP SWING**    In order to assist the transfer of body weight from one leg of the robot to the other during the gait cycle, lateral swaying motions of the hips are used. Mathematically, these swaying motions are generated by applying the overlapping sum of two opposing half-sinusoids to the leg angle X. The dimensionless *hip swing* variable $\zeta_h$, plotted in Figure 11.3 for the left leg, is first calculated using

$$\zeta_h = \max\{\zeta_{hR}, 0\} - \max\{\zeta_{hL}, 0\},\tag{11.22}$$

where

$$\zeta_{hR} = \sin\left(\frac{\pi\,\text{wrap}_0(\mu_i + \mu_\zeta)}{\pi + \mu_\zeta + D}\right),\tag{11.23a}$$

$$\zeta_{hL} = \sin\left(\frac{\pi\,\text{wrap}_0(\mu_i + \pi + \mu_\zeta)}{\pi + \mu_\zeta + D}\right),\tag{11.23b}$$

$\mu_i$ is the input gait phase, and $\text{wrap}_0(\cdot)$ is a function that wraps an angle to the range $[0, 2\pi)$ by multiples of $2\pi$. The required adjustment of the leg angle X parameters is then given by

$$\phi_{lx} \leftarrow \phi_{lx} + \zeta_h\big(k_{hs} + k_{hsx}|v_{ix}| + k_{hsy}|v_{iy}|\big),\tag{11.24}$$

for configured constants $k_*$.

**LEANING**    The final components to be added to the abstract space CPG waveforms are sagittal and lateral leaning. If

$$\mathbf{v}_i = (v_{ix},\ v_{iy},\ v_{iz})\tag{11.25}$$

is the internal Gait Command Velocity (GCV) as before, the corresponding gait command acceleration

$$\mathbf{a}_i = (a_{ix}, a_{iy}, a_{iz}) \tag{11.26}$$

is first calculated using a Savitzky-Golay derivative filter (see Appendix A.2.1.2) with subsequent slope limiting (see Appendix A.2.4.2). The required *hip leaning angles* $\zeta_{lx}$ and $\zeta_{ly}$ are then calculated as

$$\zeta_{lx} = k_{hxz} v_{ix} v_{iz}, \tag{11.27a}$$
$$\zeta_{ly} = -(k_{hvx} v_{ix} + k_{hax} a_{ix} + k_{hvz}|v_{iz}|), \tag{11.27b}$$

for configured constants $k_*$. The lateral hip leaning angle $\zeta_{lx}$ is responsible for making the robot lean into turns when walking with a significant sagittal component. The sagittal hip leaning angle $\zeta_{ly}$ on the other hand makes the robot lean forwards or backwards when it has a significant turning component, or is walking or accelerating significantly sagittally. The sagittal hip leaning due to $\zeta_{ly}$ is incorporated into the generated CPG waveforms by applying

$$\phi_{ly} \leftarrow \phi_{ly} + \zeta_{ly}, \tag{11.28a}$$
$$\phi_{fy} \leftarrow \phi_{fy} + \zeta_{ly}, \tag{11.28b}$$

This is effectively equivalent to applying $\zeta_{ly}$ as an offset to the hip pitch in the joint space. The situation is similar for the lateral hip leaning parameter $\zeta_{lx}$, in that it is effectively applied as a hip roll using

$$\phi_{lx} \leftarrow \phi_{lx} + \zeta_{lx}, \tag{11.29a}$$
$$\phi_{fx} \leftarrow \phi_{fx} + \zeta_{lx}. \tag{11.29b}$$

This is not the whole story however, as without further compensation the changed hip rolls would result in an undesirable change in relative foot height, as shown in Figure 11.4. To avoid this, an offset is applied to the leg retraction of the leg that is given by the limb sign equal to $\text{sign}(\zeta_{lx})$. Mathematically, the corresponding leg retraction update equation is given by

$$\epsilon_l \leftarrow \epsilon_l + k_{he}|\sin \zeta_{lx}|, \tag{11.30}$$

for a suitably tuned value of $k_{he}$.

### 11.2.1.2 *Inverse Leg Motion*

Unlike the abstract leg motion function, the *inverse leg motion* function does not have an open-loop contribution to the CPG waveforms, even if its existence is nonetheless required in general by the balance feedback controllers that build on the CPG. For instance, the direct fused angle feedback controller (see Chapter 13) uses the inverse leg motion function to apply Cartesian shifts to the Centre of Mass (CoM) of the robot. The same feedback controller also implements its so-called *virtual slope walking* (see Section 13.3.5) in the inverse space. These motion components are discussed where they are introduced in the context of the corresponding balance controller.

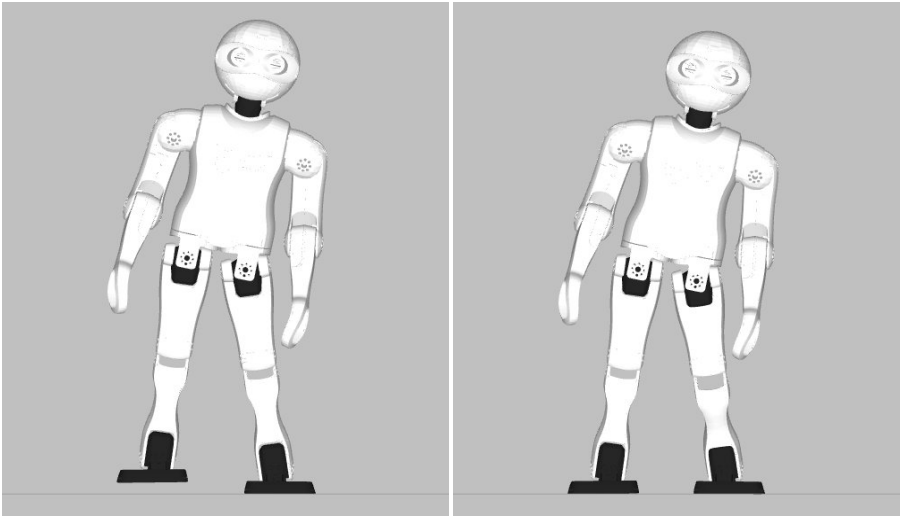

Figure 11.4: Lateral hip leaning without (left) and with (right) leg retraction adjustment. Note the significant difference in relative foot height that occurs if the leg retraction is not adjusted proportional to the sine of the magnitude of leaning.

### 11.2.1.3 *Support Coefficients*

Like most other components of the leg motion profile, the commanded support coefficients $\kappa_l$ and $\kappa_r$ are generated individually for each leg as a direct function of the limb phase $\nu_i$. The support coefficient $\kappa$ corresponding to a leg is zero during the period of sinusoid forwards swing,[5] and linearly transitions to unity during the double and single support phases. An illustration of the resulting trapezoidal support coefficient waveform is given in Figure 11.3. Mathematically, the support coefficient $\kappa$ of a leg is given by

$$\kappa = \begin{cases} 1 & \text{if } \nu_i \in (-\pi + D, -\mu_\zeta], \\ 1 - \frac{\nu_i + \mu_\zeta}{D + \mu_\zeta} & \text{if } \nu_i \in (-\mu_\zeta, D], \\ 0 & \text{if } \nu_i \in (D, \pi - \mu_\zeta], \\ \frac{\text{wrap}(\nu_i + \mu_\zeta - \pi)}{D + \mu_\zeta} & \text{if } \nu_i \in (\pi - \mu_\zeta, \pi] \cup (-\pi, -\pi + D]. \end{cases} \quad (11.31)$$

Note that the phase durations of the linear fades between 0 and 1 are exactly $D + \mu_\zeta$, and that at all times the support coefficients of the left and right legs do indeed sum to unity as expected.

---

5 As given in Equation (11.19), the sinusoidal forwards swing occurs in the limb phase range $\nu_i \in (D, \pi - \mu_\zeta]$.

### 11.2.2  Arm Motion Profile

The generation of the CPG arm motions begins for each arm with the configured halt pose in the abstract space, and continues as follows:

(i) The halt pose is additively modified by the **abstract arm motion** function, which primarily incorporates arm swinging motions to the generated waveforms.

(ii) The abstract pose is **soft-coerced** (see Appendix A.1.2.3) to safe limits, to ensure that no excessive internal GCV or faulty balance feedback controller can cause dangerous motions of the robot.

(iii) The resulting pose is converted to the **joint space** to give the final CPG joint position waveforms.

Recall that the joint effort commands are kept constant at suitably configured values, and so do not need to be dealt with explicitly here.

The arm motion profile, like the leg motion profile, is a direct function of the internal GCV vector $\mathbf{v}_i = (v_{ix}, v_{iy}, v_{iz})$ and the limb phase $\nu_i$, where it should be noted that the limb phase of each arm is equivalent to the limb phase of the *opposite* leg. This definition makes sense because during normal walking, the pairs of opposing arms and legs usually swing forwards and backwards in a synchronised manner, so as not to produce significant unbalanced inertial moments in the yaw direction.

#### 11.2.2.1  *Abstract Arm Motion*

The *abstract arm motion* function works much like the equivalent function for the legs, and generates mainly just the required sagittal arm swinging motions of the robot. The same limb phase keypoints as outlined in Table 11.1 are used, and in fact keeping in mind the definition of the limb phase $\nu_i$ for the arms, the *swing angle* $\zeta_s$ from Equation (11.19) can be directly reused to define the arm swing update equation

$$\phi_{ay} \leftarrow \phi_{ay} - \zeta_s(k_{as} + k_{asx}v_{ix}), \tag{11.32}$$

for configured constants $k_*$. In addition to the generation of the arm swing waveforms, the abstract arm motion function is used by the balance feedback controllers that build on the CPG as an opportunity to apply reactive arm feedback motions. An example of this can be found in Section 13.3, in the context of the direct fused angle feedback controller.

### 11.3  EXPERIMENTAL RESULTS

The Central Pattern Generator (CPG) has been successfully implemented on all of the robots listed in Section 2.1.2. The level of stability that the open-loop gait has varies with the mechanical quality of the robot and whether serial or parallel leg kinematics are present, but all tested robots were able to walk with the CPG with a moderate to high quality. Clearly, without any balance feedback there is a limit to the level of disturbances (and self-disturbances) that can be passively rejected by the gait, but the self-stability of the CPG was found to be sufficient in all robots for moderate durations of walking over flat terrains. For conservative input GCV vectors, the most common mode of failure is if a timing offset is induced in the balance of the robot, and the robot ends up trying to lift a leg that it is currently standing on.

Due to the moderately large number of parameters of the CPG, it may at first seem like it is difficult to tune, but seeing as most of the parameters were intentionally designed to be dimensionless and highly independent in terms of their effect, this is not truly the case. The process of tuning starts with the tuning of the halt pose, i.e. the base pose of the gait (see Figure 11.2). The halt pose should be a stable and symmetric standing pose of the robot, with a small yet non-zero amount of leg retraction. In the halt pose, the balance of the robot should be suitably centred above the support polygon defined by its feet, and in the case of serial kinematics robots, should incorporate a slight forwards tilt of the torso. The leg lifting and hip swing parameters are first tuned to achieve stable walking on the spot, before experiments with pure steady-state sagittal, lateral and rotational walking motions are used to tune the leg swing and arm swing parameters. The leaning parameters come last as they depend on the behaviour of the robot when it is accelerating in the sagittal direction, which requires tests with changing GCVs. Once the parameters of the CPG have been tuned for one robot, carrying them across to another robot is usually a simple task, and requires only very minimal changes to the parameters, if at all, for the gait to work. Usually, even between robots of different shapes and sizes, only the halt pose needs to be adjusted, and at most the configured step height parameters as well.

Figure 11.5 shows plots of the generated joint space waveforms for the left and right limbs of an igus Humanoid Open Platform robot. When the CPG gait is triggered at time $t = 0.65$ s, no pre-gait blending is required (see Step (a) of the CPG motion generation process on page 322), as the robot is already in the halt pose. After $\pi$ radians (i.e. 417 ms) of pose blending, and an overlapping 1.2 s of forced zero internal GCV time, a step change in $\mathbf{v}_g$ occurs to the commanded value of $(0.7, 0.4, 0.4)$. When this happens, the slope limiting of the internal GCV $\mathbf{v}_i$ ensures that it remains continuous in its approach of $\mathbf{v}_g$, as can be seen from the resulting linear ramps in the top plot of the figure. After a few seconds of steady-state limit cycle walking, at $t = 6.3$ s, the trigger comes that the robot should stop walking. It cannot do this immediately, as it is in the middle of taking a large step and is

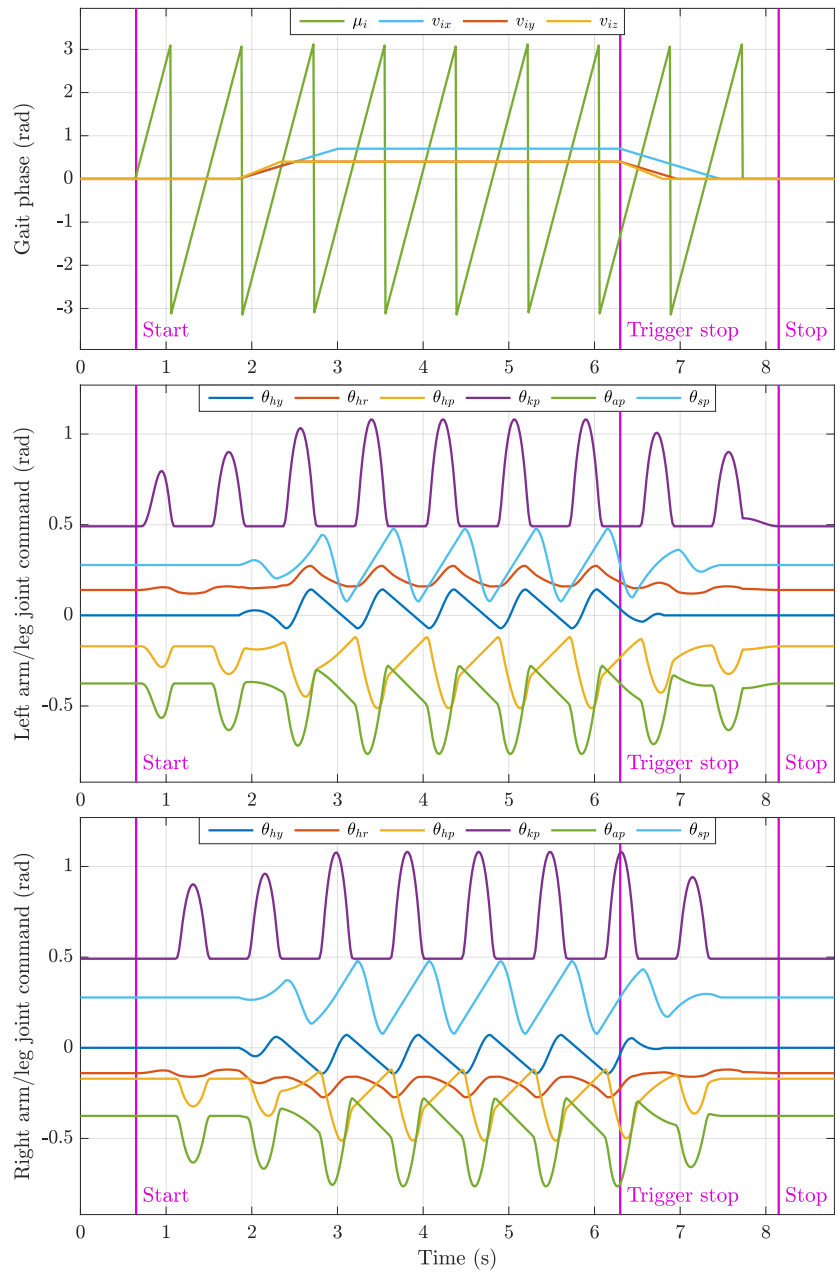

Figure 11.5: Sample output waveforms of the Central Pattern Generator (CPG) for an input GCV of $\mathbf{v}_g = (0.7, 0.4, 0.4)$ and a nominal gait frequency of $f_g = 2.4\,\text{Hz}$. The CPG gait is activated at time $t = 0.65\,\text{s}$, and receives the trigger to stop walking at time $t = 6.3\,\text{s}$. After a further two steps, during which the robot slows down to an internal GCV of zero, the robot stops walking and fades back to the halt pose. Note how the waveforms for the left limbs (middle plot) are in general in exact antiphase to those for the right limbs (bottom plot). This can most clearly be identified in the plots of $\theta_{kp}$, the knee pitch angles. The top plot shows the values of the gait phase $\mu_i$ and internal GCV vector $\mathbf{v}_i$ used for the generation of the waveforms. The effect of the initial GCV zero time, in addition to the effect of the GCV slope limiting, can be seen in the plot.

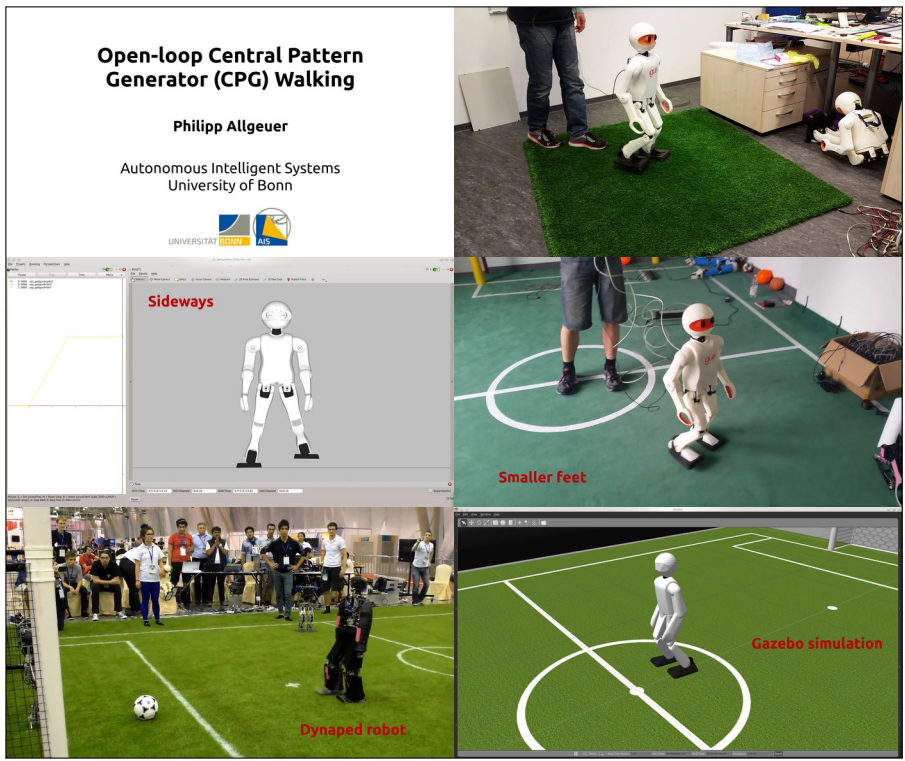

Video 11.1: Demonstration of the open-loop Central Pattern Generator (CPG)
gait on the igus Humanoid Open Platform, Dynaped and Nim-
bRo-OP2 robots. The gait is demonstrated in kinematic simula-
tion, physical simulation, and on the real robot hardware.
https://youtu.be/ksJRwGlSuTM
*Open-loop Central Pattern Generator (CPG) Walking*

currently walking with a significantly non-zero GCV, but over the next
2–3 steps, the internal GCV is faded linearly back down to zero. Once
$\mathbf{v}_i$ arrives at zero, at the next culmination of a step ($\mu_i = 0$ or $\pi$) the
CPG waveforms are stopped, and after 417 ms of further pose blending
back to the halt pose, the CPG gait releases control of the robot.

A video of the CPG gait running on various kinematic simulations,
physical simulations and real robots is provided in Video 11.1. The
quality of the gait without feedback is not perfect, and the robots are
prone to falling if disturbed, but the CPG serves as a reliable foundation
on which robust gaits like the capture step gait (see Chapter 12) and
direct fused angle feedback controller (see Chapter 13) can be built.

# CAPTURE STEP GAIT

---

The Central Pattern Generator (CPG) works well in generating open-loop walking motions for a robot, but the resulting gait is not as robust as ultimately desired for applications like the RoboCup competition. In such applications, where strong disturbances, irregular terrain and pushing are rife, a more reliable and preferably actively balanced gait is required in order to enable a high level of robot performance. The capture step gait[1] is exactly such a gait, and extends the CPG with the so-called *capture step balance controller*. First introduced by Missura and Behnke (2013a), and further evolved in the PhD dissertation of Missura (2015), the core idea of the capture step controller is the preservation of balance through predictive changes in step size and timing based on the Linear Inverted Pendulum Model (LIPM). The integration of the capture step feedback with the CPG gait presented in this thesis is relatively simple, as provisions for exactly this kind of feedback were already made in Section 11.1.3.

Missura (2015) identifies four main components of his presented approach to bipedal walking, known as the *capture step framework*. These are the

- **State estimation**, responsible for estimating the Centre of Mass (CoM) state and support leg sign,

- **Footstep control** algorithm, responsible for calculating suitable step times and sizes for balanced walking given the current state,

- **Motion generator**, responsible for generating the joint wave-forms required to effectuate the commanded step times and sizes, and,

- **Bipedal robot** hardware and control approach, responsible for translating the calculated joint commands into real motions of the robot.

In the context of this thesis, almost all of these components have been replaced (or at least significantly overhauled) relative to what was used in Missura (2015). For instance, the *state estimation* component has been upgraded to use the attitude estimator from Section 10.1, a more correct model of the foot-to-ground contacts (and concept of robot heading) using the notion of fused yaw (see Section 10.2), and a different type of filtering and velocity estimation for the final estimated

---

1 https://github.com/AIS-Bonn/humanoid_op_ros/blob/master/src/nimbro/motion/gait_engines/cap_gait/src/cap_gait.cpp

CoM state (see Section 10.2.1). The *motion generator* component has been upgraded and replaced with the CPG gait presented in the previous chapter, and the control approach used for the *bipedal robot* hardware component has been replaced to use the robot control scheme (see Section 3.1) and actuator control scheme (see Section 3.2).

The only component of Missura's capture step framework that saw only moderate upgrades and changes in its adaptation for use in this thesis is the *footstep control* algorithm. While some changes were made to refine the implementation of the balance prediction equations and make the calculations more robust to resetting and outlier cases, the fundamental idea of the LIPM prediction equations was not changed. As such, the description of the capture step controller provided in this chapter is reduced to just the main ideas and equations, as Missura and Behnke (2013a) and Missura (2015) still constitute good references for the used approach.

## 12.1  CAPTURE STEP CONTROLLER

Formally, the *footstep control* algorithm, referred to in this thesis as the capture step controller, can be thought of as a function

$$f_{CSC} : \{\delta_s, \mathbf{c}, \dot{\mathbf{c}}, \tilde{\mathbf{v}}_g\} \mapsto \{\delta_t, t_s, \tilde{\mathbf{v}}_i\} \tag{12.1}$$

that maps certain inputs to certain outputs, where

- $\delta_s \in \{-1, 1\}$ is the current estimated support leg sign[2] as per the CoM state estimation (see Section 10.2.1),

- $\mathbf{c} = (c_x, c_y)$ and $\dot{\mathbf{c}} = (\dot{c}_x, \dot{c}_y)$ are the estimated x and y positions and velocities, respectively, of the CoM point relative to the current estimated support foot floor frame {F̄} (see Section 10.2.1),

- $\tilde{\mathbf{v}}_g = (\tilde{v}_{gx}, \tilde{v}_{gy}, \tilde{v}_{gz})$ is the input Gait Command Velocity (GCV) to the CPG after p-norm limiting, but prior to slope limiting (see Section 11.1.3.2),

- $\delta_t \in \{-1, 1\}$ is the support leg sign corresponding to the commanded outputs (see Section 11.1.3.1),

- $t_s$ is the commanded *time to step* with respect to $\delta_t$, i.e. the exact time that should be taken until the CPG next commands foot strike for the foot opposite to $\delta_t$ (see Section 11.1.3.1), and,

- $\tilde{\mathbf{v}}_i = (\tilde{v}_{ix}, \tilde{v}_{iy}, \tilde{v}_{iz})$ is the commanded value of the internal GCV vector $\mathbf{v}_i$ of the CPG, up to the end-of-step fading given by Equation (11.13) (see Section 11.1.3.2).

---

2 Recall from Equation (9.17) that a limb sign of 1 corresponds to the left leg, and a limb sign of −1 corresponds to the right leg.

It should be noted that $f_{CSC}$ does not actually correspond to a function in the traditional mathematical sense of the word, because it contains an element of 'memory' in it due to the use of a predictive filter (see Section 12.1.2), but for all intents and purposes, this functional model of the capture step controller suffices.

### 12.1.1 Linear Inverted Pendulum Model

The capture step controller models the behaviour of the robot using the Linear Inverted Pendulum Model (LIPM). The mass of the robot is assumed to be concentrated at its CoM point—where we recall from Figure 9.1 that this is modelled as a fixed point relative to the trunk—and is assumed to 'tip over' with constant z-height under the influence of gravity relative to a particular pendulum base point.

#### 12.1.1.1 1D Pendulum Model

The basic one-dimensional LIPM is illustrated in Figure 12.1a. Assuming the pendulum base point $p$ and height $h$ of the robot stay constant, the future motion of the CoM point is modelled by the linearised Ordinary Differential Equation (ODE)

$$\ddot{x} = C^2(x - p), \tag{12.2}$$

where $x$ is the horizontal position of the CoM relative to the pendulum base point, and $C$ is the pendulum constant. Analytically, the value of $C$ can be expected to be

$$C \approx \sqrt{\frac{g}{h}}, \tag{12.3}$$

where $g = 9.81 \, \text{m/s}^2$, but the value is usually freely tuned instead to fit the observed behaviour of the robot.

Given suitable initial conditions $(x_0, \dot{x}_0)$, Equation (12.2) has a closed-form solution for the prediction of the CoM state at some future time $t \geq 0$, namely

$$x(t, x_0, \dot{x}_0) = p + (x_0 - p)\cosh(Ct) + \tfrac{1}{C}\dot{x}_0 \sinh(Ct), \tag{12.4a}$$

$$\dot{x}(t, x_0, \dot{x}_0) = \dot{x}_0 \cosh(Ct) + C(x_0 - p)\sinh(Ct). \tag{12.4b}$$

The times $t_x$, $t_{\dot{x}}$ at which the CoM reaches a particular future position $x$ or velocity $\dot{x}$, respectively, is given by

$$t_x(x, x_0, \dot{x}_0) = \tfrac{1}{C}\log\left(\tfrac{1}{k_1}\left(x - p \pm \Delta_x\right)\right), \tag{12.5a}$$

$$t_{\dot{x}}(\dot{x}, x_0, \dot{x}_0) = \tfrac{1}{C}\log\left(\tfrac{1}{Ck_1}\left(\dot{x} \pm \Delta_{\dot{x}}\right)\right), \tag{12.5b}$$

where

$$k_1 = (x_0 - p) + \tfrac{1}{C}\dot{x}_0, \tag{12.6a}$$

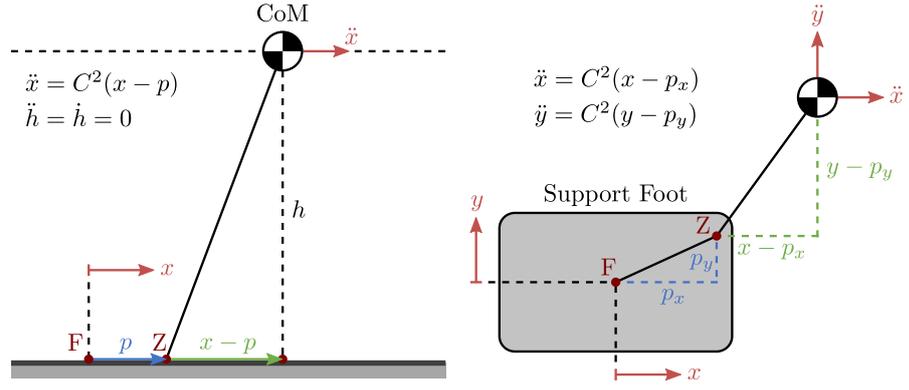

(a) The 1D Linear Inverted Pendulum Model shown from the side, where the x-axis points to the right and the z-axis points up. F is the centre of the support foot, and Z is the modelled ZMP. The height of the CoM is assumed to stay constant.

(b) The 2D Linear Inverted Pendulum Model shown from the top, where the x-axis points to the right, the y-axis points up, and the z-axis points out of the page. The 2D LIPM is equivalent to two orthogonal 1D LIPMs.

Figure 12.1: Diagrams of the 1D and 2D Linear Inverted Pendulum Models.

$$k_2 = (x_0 - p) - \tfrac{1}{C}\dot{x}_0, \tag{12.6b}$$

$$\Delta_x = \text{sign}(k_1)\sqrt{(x - p)^2 - k_1 k_2}, \tag{12.6c}$$

$$\Delta_{\dot{x}} = \text{sign}(k_1)\sqrt{\dot{x}^2 + C^2 k_1 k_2}. \tag{12.6d}$$

As a special case, the time at which the CoM reaches the apex of its trajectory, i.e. $\dot{x} = 0$, is given by

$$t_a(x_0, \dot{x}_0) = \tfrac{1}{C} \log\left(\sqrt{\tfrac{k_2}{k_1}}\right), \tag{12.7}$$

where the returned time can be negative, or undefined if no apex is ever reached.

Naturally, Equations (12.4) to (12.7) are only valid if the modelled pendulum remains free of disturbances throughout its entire trajectory. If this is the case, an invariant referred to as the pendulum orbital energy can be formulated, which remains constant throughout the entire past, present and future motion of the pendulum. The orbital energy is given by the function

$$E(x, \dot{x}) = \tfrac{1}{2}\left(\dot{x}^2 - C^2(x - p)^2\right). \tag{12.8}$$

Note that from Equation (12.2),

$$\frac{d}{dt}\left(E(x, \dot{x})\right) = \dot{x}\ddot{x} - C^2(x - p)\dot{x} = \dot{x}\left(\ddot{x} - C^2(x - p)\right) = 0, \tag{12.9}$$

so $E(x, \dot{x})$ remains constant as required.

### 12.1.1.2  *2D Pendulum Model*

The one-dimensional LIPM can model the motion of the CoM point of the robot in one particular direction, but assuming that the CPG keeps the height of the robot above the ground more or less constant, the motion of the CoM has two degrees of freedom. As such, two uncoupled 1D LIPMs are used in parallel to model the sagittal and lateral motions of the robot. That is,

$$\begin{bmatrix} \ddot{x} \\ \ddot{y} \end{bmatrix} = \begin{bmatrix} C^2 & 0 \\ 0 & C^2 \end{bmatrix} \begin{bmatrix} x - p_x \\ y - p_y \end{bmatrix}. \tag{12.10}$$

The x dimension corresponds to the sagittal plane of motion of the CoM, and the y dimension corresponds to the lateral plane. The point $\mathbf{p} = (p_x, p_y)$ is seen to correspond to the Zero Moment Point (ZMP), and acts like the base of the modelled two-dimensional pendulum, as indicated in Figure 12.1b. As intuitively expected, changes in ZMP influence the resulting acceleration of the CoM. Note that due to the assumption of sagittal and lateral independence, all of the equations introduced in the previous section can be applied to either of the two LIPMs separately to make predictions of the future CoM state.

### 12.1.2  **RX-MX-TX Predictive Filter**

Given a particular two-dimensional CoM position $(x, y)$ and CoM velocity $(\dot{x}, \dot{y})$, it is possible to predict the entire future trajectory of the CoM right up until the next step is taken. When the step happens, the values of the CoM state variables and ZMP point experience a step change, but afterwards, the equations from Section 12.1.1.1 can be used again for future state prediction relative to the new support foot. As such, if a predictive model is developed for when and where the robot should nominally step during a trajectory, and what happens to the CoM state when the step occurs, a reference trajectory for balanced walking can be generated, where we note that the nominal GCV vector is also an input to the reference trajectory. The predictive model can be evaluated arbitrarily far in the future, as every time the modelled nominal step condition is true (which always occurs prior to 'falling'), the model changes its support leg sign and continues with the next step.

Suppose that at the beginning of walking with the CPG, a predictive model as described above is initialised to the current CoM state of the robot. If the predictive model and LIPM are both accurate in describing the behaviour of the robot (and no external disturbances are present), one would expect that all future measured support leg signs $\delta_s$, and CoM states $\mathbf{c} = (c_x, c_y)$ and $\dot{\mathbf{c}} = (\dot{c}_x, \dot{c}_y)$, would align perfectly with the predicted support leg signs and CoM states from the predictive model.[3] In reality, clearly this would not be true for long though, as a walking bipedal robot is somewhat of a chaotic system that cannot be modelled perfectly. If feedback from the measured CoM states is regularly blended into the predictive model however, the states can remain indeed remain aligned. This is the idea behind the RX-MX-TX predictive filter.

Let us refer to the described predictive model as the MX model. The so-called RX model refers to an equivalent filter that is always just set to the current estimated CoM state from the state estimation, i.e. $\mathbf{c}$, $\dot{\mathbf{c}}$ and $\delta_s$. A third so-called TX model represents the state of the MX model cast a constant amount of time into the future, so as to circumvent actuator and sensor latency. At every time step of the execution of the gait,

- The RX model is updated to the latest values of the CoM state estimation (see Section 10.2.1), and if a step was just observed to occur, i.e. if $\delta_s$ just changed its value, the CoM state velocities are adjusted by the relative fused yaws of the feet in order to maintain consistency.

- The MX model (i.e. the predictive model), which currently holds the MX CoM state from the last execution cycle of the gait, is forwarded $\Delta t$ seconds into the future, in order to predict the state of the robot at the current time instant.

- If the support leg sign $\delta_m$ of the updated MX model state is equal to the support leg sign $\delta_s$ of the RX model, the MX model CoM state is linearly interpolated towards the RX model state in a blending process referred to as *adaptation*. The blending interpolation factor $b \in [0,1]$ is continuously variable throughout each step, and is calculated based on the

  - *Step noise suppression* factor, which is a Gaussian function that is close to zero near the modelled and/or predicted instants of foot strike, and,

  - *Expected deviation* factor, which is proportional to the normed difference between the RX and MX model states.

  The final calculated interpolation factor $b$ is limited to a given threshold $b_{max}$ to avoid overly sudden changes to the MX state.

- The MX model CoM state is used to calculate a suitable time to step $t_s$, ZMP location $\mathbf{p}$, and step size GCV $\tilde{\mathbf{v}}_i$ (see Section 12.1.3).

- The TX model is set to the state of the MX model, and is subsequently predicted $t_l$ seconds into the future, where $t_l$ is the approximated actuator/sensor latency time, commonly in the 10–25 ms range. Note that if required, the TX model—like the MX model—may toggle its support leg sign $\delta_t$ during its prediction step. The variables calculated in the previous step are carried across to the TX model and updated accordingly as required. As indicated by Equation (12.1), the final updated values of the support leg sign $\delta_t$, time to step $t_s$, and step size GCV $\tilde{\mathbf{v}}_i$ constitute the outputs of the capture step controller.

---

3 Recall that both the predictive model and CPG gait receive the same input GCV vector.

This entire process is referred to as the RX-MX-TX *predictive filter*. For more details on this filter, refer to Chapter 8.2 of Missura (2015).

As discussed at the beginning of this section, the predictive model encompasses not only the use of the LIPM equations for the prediction of future CoM states, but also the prediction of *when* steps nominally should be taken (as a function of the internal GCV $\mathbf{v}_i$) during normal CPG walking. This is referred to as the stepping time model, and is governed by four main parameters:

- $\alpha$, the lateral inwards distance from the support foot leg tip point to the *apex point* of the lateral swing of the CoM towards the support foot,

- $\delta$, the lateral inwards CoM location relative to the support foot leg tip point at the instant of foot strike during normal walking,

- $\omega$, the lateral inwards CoM location relative to the support foot leg tip point at the instant of foot strike of the leading foot during maximum speed sideways walking,

- $\sigma$, the sagittal CoM velocity at the step apex for the maximum sagittal walking velocity.

In order to calculate the instant of the next nominal step, $\delta$ and $\omega$ are first interpolated based on the magnitude of the lateral GCV component $v_{iy}$. Depending on whether the current step is the leading sideways step or the trailing one, either this interpolated value or $\delta$ is chosen as the nominal support exchange point $s_y$, and the predicted time for the CoM to reach this point (after passing its apex) is taken as the nominal reference time to step. Clearly, if the robot is disturbed, the optimal step sizes and timing may need to be adjusted away from these nominal reference values in order to preserve balance, but in all cases it is the aim of the capture step controller to return the robot to nominal walking as soon as possible.

### 12.1.3 Capture Step Timing and Step Size Adjustment

Given any MX model

- CoM position $\mathbf{c} = (c_x, c_y)$,

- CoM velocity $\dot{\mathbf{c}} = (\dot{c}_x, \dot{c}_y)$, and

- Support leg sign $\delta_m$,

it is the aim of the capture step controller to compute

- A suitable (remaining) step time $T$,

- A feasible ZMP point $\mathbf{p} = (p_x, p_y)$ that steers the CoM towards its target end-of-step state $s_y$ in the corresponding time $T$, and

- A feasible step size $\mathbf{F} = (F_x, F_y)$ that ensures that the CoM returns to its nominal reference trajectory (e.g. a lateral CoM apex of $\alpha$) in the following step.

For more strongly disturbed CoM states, it is not always possible to meet all of these criteria within a single step, but the capture step controller does the best it can do within the constraints of the allowed step size and support polygon (which the ZMP must stay inside). The capture step controller works as follows:

A) The **lateral ZMP offset** $p_y$ is computed so that the CoM reaches its nominal support exchange location $s_y$ (refer to $\delta$ and $\omega$ of the stepping time model) at the nominal time. The nominal time is calculated from the elapsed step time and the nominal total step time, which in turn is evaluated using $\alpha$, $s_y$, $C$ and Equation (12.5a).

B) The **lateral ZMP offset** $p_y$ is coerced to the support polygon of the support foot, to avoid tipping of the support foot relative to the ground.

C) Based on $c_y$, $\dot{c}_y$ and $p_y$, the **step time** $T$ is computed to be the time until the nominal support exchange location $s_y$ is predicted to be reached. The step time $T$ can differ from the nominal time calculated in Step A if $p_y$ was coerced in Step B. If the nominal support exchange location is never reached because the robot is predicted to tip outwards over its support foot—a situation referred to as *crossing*—a fixed step time of 2 s is maintained, as it is unclear when the robot will return to a position it can make a step in, if at all. If the robot has insufficient lateral orbital energy, and consequently is not predicted to reach $s_y$ at all, the step time $T$ is calculated using Equation (12.7) as the time to the apex of the CoM trajectory. This situation is referred to as *stalling*.

D) If the **step time** calculated in Step C leads to a predicted end-of-step sagittal CoM position in excess of a particular configured bound, the step time $T$ is reduced to ensure that the bound is not exceeded. This ensures that if the robot is disturbed in both the lateral and sagittal directions, the robot will not wait so long for the lateral balance to recover that the sagittal balance becomes irrecoverable.

E) The **lateral step size** $F_y$ is calculated such that the apex of the lateral CoM trajectory of the step after the current one is predicted to be at exactly $\alpha$. This corresponds to a return to the reference CoM trajectory, and is computed by predicting the lateral end-of-step CoM velocity using Equation (12.4b) (based on $c_y$, $\dot{c}_y$, $p_y$ and $T$), and solving for the required footstep location $F_y$ based on the orbital energy expression in Equation (12.8).

F) The **sagittal ZMP offset** $p_x$ is computed based on $c_x$, $\dot{c}_x$ and $T$ so that the sagittal end-of-step CoM state is at its nominal location $s_x$. The nominal end-of-step CoM location $s_x$ is calculated from the sagittal component $v_{ix}$ of the GCV, the maximum sagittal end-of-step CoM velocity $\sigma$, and the nominal total step time from Step A.

G) The **sagittal ZMP offset** $p_x$ is coerced to the support polygon of the support foot, to avoid tipping of the support foot relative to the ground.

H) The **sagittal step size** $F_x$ is calculated by first evaluating the predicted sagittal end-of-step CoM velocity based on $c_x$, $\dot{c}_x$, $p_x$ and $T$. The step size $F_x$ is then chosen to be the sagittal step size that would nominally result in exactly that end-of-step CoM velocity, i.e. if it was performed as a nominal step in the nominal time. This method of generating $F_x$ ensures that the CoM state at the end of the step is one that would occur during nominal walking with that step size, and hence effectively returns the robot to its reference trajectory (albeit with a GCV different to commanded).

I) The **output** time to step $t_s$ is given by the calculated step time $T$, the output step size GCV $\tilde{\mathbf{v}}_i$ is calculated from $\mathbf{F} = (F_x, F_y)$, and the modelled ZMP location is given by $\mathbf{p} = (p_x, p_y)$.

This completes the description of the capture step controller. For the equations corresponding to each of the steps, refer to Missura and Behnke (2013a) and Missura (2015), or the source code of the implementation.[4]

### 12.1.4 Capture Step Tuning

Given the combined complexity of the LIPM, stepping time model, and general predictive filter, it is important that the correct procedure for the tuning of the capture step controller is applied. It is strictly necessary that the joint offsets, attitude estimation, CoM state estimation and CPG gait are completely tuned before the tuning of the capture

---

4 https://github.com/AIS-Bonn/humanoid_op_ros/blob/master/src/nimbro/
motion/gait_engines/cap_gait/src/contrib/LimpModel.cpp

step controller is begun. If this is the case, the following tuning steps should then be applied in order:

(i) The symmetry of the fused roll estimated by the attitude estimator (and used by the CoM state estimation) should first be verified. This is done by letting the robot walk on the spot for 10–15 steps while facing one particular direction, and then the same number of steps (on the same spot) facing 180° in the opposite direction. If the tuning of the CPG is satisfactory, the robot should in both cases walk in a rhythmical, upright, and laterally symmetric manner, and spend observably equal amounts of time on each foot with each step. If the Inertial Measurement Unit (IMU) sensors and attitude estimator are correctly calibrated, one should observe during the 20–30 overall recorded steps that the average estimated fused roll is zero. If this is pronouncedly not the case, a recalibration of the sensors is likely required. If the mean of the fused roll is close to zero but not negligibly close, a small offset to the fused roll can potentially be applied prior to the CoM state estimation to ensure that the mean of the fused roll is perfectly zero.

(ii) The values of $\alpha$ and $\delta$, and offsets to the estimated CoM position $\mathbf{c} = (c_x, c_y)$, are estimated by repeating the walking on the spot experiment from the previous step. The positions of the CoM trajectory apexes are averaged out separately for the left and right legs to give $\hat{\alpha}_l$ and $\hat{\alpha}_r$ respectively, and the CoM positions immediately prior to the observed support exchanges are averaged out to give $\hat{\delta}_l$ and $\hat{\delta}_r$. Ideally, we would like to set $\alpha = \hat{\alpha}_l = \hat{\alpha}_r$, $\delta = \hat{\delta}_l = \hat{\delta}_r$ and a CoM offset Y of $c_{oy} = 0$, but it is not generally the case that the observed $\alpha$ and $\delta$ values are perfectly equal for both legs. We compensate for this by defining

$$\alpha = \tfrac{1}{2}(\hat{\alpha}_l + \hat{\alpha}_r), \tag{12.11a}$$

$$\delta = \tfrac{1}{2}(\hat{\delta}_l + \hat{\delta}_r), \tag{12.11b}$$

$$c_{oy} = \tfrac{1}{2}(\hat{\alpha}_l - \hat{\alpha}_r), \tag{12.11c}$$

where we note by design that,

$$\hat{\alpha}_r + c_{oy} = \hat{\alpha}_l - c_{oy}. \tag{12.12}$$

The CoM offset $c_{oy}$ is added to all future measurements of $c_y$ to ensure that the values of $\alpha$ and $\delta$ are symmetric. The sagittal CoM offset $c_{ox}$ is set to be the negative of the mean of the $c_x$ values observed during the walking experiment. This ensures that in the future, the values of $c_x$ have a mean of zero when walking on the spot as expected.

(iii) Given the tuned values of $\alpha$ and $\delta$, and the nominal gait frequency $f_g$ of the CPG, the required pendulum constant $C$ can be calculated by solving (for $p = 0$)

$$x\left(\tfrac{1}{2f_g}, \alpha, 0\right) = \delta, \tag{12.13}$$

where the left-hand side is expanded using Equation (12.4a). This yields the expression

$$C = 2f_g \operatorname{acosh}\left(\tfrac{\delta}{\alpha}\right). \tag{12.14}$$

(iv) Based on the end-of-step CoM positions and velocities observed during pure maximum speed forwards and sidewards walking, suitable values for $\omega$ and $\sigma$ are inferred. The value of $\omega$ is expected to be somewhat (but not considerably) larger than the value of $\delta$.

(v) Until now, the acquired walking data has only been from the raw tuned CPG itself, without the predictive filter and capture step controller. We now test the parameters tuned so far by enabling the predictive filter and driving the gait through the MX and TX models, but keeping the adaptation (blending of the MX model towards the RX model), step size and timing adjustments disabled. This essentially means that the MX model is just eternally being forwarded into the future based on the LIPM and stepping time models, and thereby generating the nominal reference trajectories for the input GCVs that the gait is receiving. The MX model is predicted into the future to form the TX model as usual, and the time to step outputs from the TX model are used to pace the gait at the natural stepping frequency of the LIPM model. Assuming that Equation (12.14) was applied correctly, this natural stepping frequency should correspond exactly to the nominal gait frequency $f_g$ of the CPG.

(vi) With a configured actuator/sensor latency time $t_l$ of zero, the robot is made to walk perfectly rhythmically on the spot as per Step (v). The average time between the support transitions of the TX model and the RX model is measured, and configured as the new latency time $t_l$. The value of $t_l$ should be fine-tuned until the support transitions of the RX and MX models are observed to be synchronised.

(vii) After configuring the adaptation parameters to suitable values, the blending of the MX model towards the RX model can be enabled in conjunction with enabling the timing adjustment component of the capture step controller. This is the first point in time that the CPG gait is being run with feedback, albeit not yet with step size feedback. It should be checked that the RX and MX models never desynchronise, even if the robot is disturbed, and that the capture step timing adjustments work in helping the robot maintain its balance when disturbed while walking on the spot.

(viii) The robot is made to walk around with a non-zero GCV, with just the timing feedback enabled. Ensuring that maximum speed walking is reached at some point in each direction, the maximum open-loop step sizes taken in the x and y-directions (as calculated by the model) are recorded and used as future limits for the step size adjustment algorithm. The maximum sagittal end-of-step CoM position is also recorded, and used for the limiting of the step time $T$ (see Step D on page 342).

(ix) After checking (for safety reasons) that the step sizes calculated by the TX model during balanced walking with a non-zero GCV are close in magnitude to the step sizes that are otherwise performed open-loop, the step size adaptation of the capture step controller can be enabled. By testing the reaction of the robot to light, medium and more severe pushes, the lateral and sagittal ZMP limits of the support polygon are tuned. These ZMP limits strike the balance between when the capture step controller considers a disturbance serious enough to merit a reactive step, and when the controller predicts that the passive stability of the robot is sufficient to smoothen it out. The values of $\omega$ and $\sigma$ may need to be fine-tuned for proper performance of the step size adaptation scheme.

This completes the tuning process of the capture step controller.

## 12.2  EXPERIMENTAL RESULTS

A video of the tuned capture step controller running on a NimbRo-OP2X robot is shown in Video 12.1. The general walking performance of the robot with the controller is first demonstrated, followed by an extensive array of lateral and sagittal pushes. While the controller visually only makes sparing adjustments during general omnidirectional walking, the effect of the capture steps in enhancing push recovery ability is quite pronounced.

Still images of one particular lateral push are shown in Figure 12.2. It can be seen that after the robot is pushed, it begins to tilt over towards its right side. At this stage, if there was enough lateral energy for the robot to be on a crossing trajectory (see page 342), the robot would have no option but to wait and hope that it does return to upright after all (and recompute appropriate footstep locations and timings only once it is on the returning trajectory). The push in Figure 12.2 was not severe enough for crossing however, and after a brief lateral tilting phase (top row, third image), the robot returned to upright and took a relatively large step (top row, fourth image) so that the ensuing lateral tilt after the step was essentially normal again (bottom row, first image). Some residual lateral energy remained, so the next step was marginally enlarged (bottom row, second image) to bring the robot back to its ideal walking limit cycle (bottom row, fourth image).

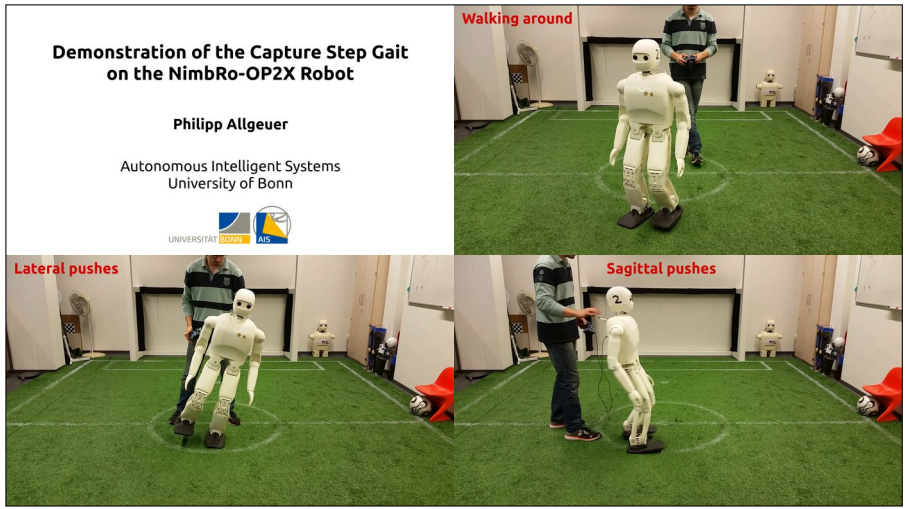

Video 12.1: A demonstration of the capture step gait (in ROS) on the Nimb-bRo-OP2X robot. Changes and improvements from the original were made to the state estimation, motion generator and hardware control components, as well as to a lesser extent the footstep control algorithm.
https://youtu.be/gSPeC3xO9YM
*Demonstration of the Capture Step Gait on the NimbRo-OP2X Robot*

This entire process can be observed more closely in the plots shown in Figure 12.3, which correspond exactly to the images shown in Figure 12.2. After the push occurs at $t = 1\,\text{s}$, the resulting measured RX CoM position $c_y$ can be seen to fall lower than expected. This can most clearly be observed right after the support transition at $t = 1.1\,\text{s}$, where there is a notable gap between the lateral CoM positions of the RX and MX models (red and blue lines). Recall that $c_y = 0$ is the modelled tipping point of the robot in the lateral direction, so the smaller $c_y$ value of the RX model indicates that the robot is closer to tipping than expected. Due to adaptation, i.e. blending of the MX model towards the RX model, over the next half-second the gap between the two models closes, and the predicted time to step $t_s$ (bottom plot) increases to account for the new estimated state. The desired step size $F_y$ is also increased to account for the predicted extra momentum that the robot will have when returning to upright, and the commanded lateral GCV $\tilde{v}_{iy}$ is increased linearly in such a way that it reaches the target step size exactly at the predicted end of the step. This process can be observed to continue in the next step, and to a lesser extent in the next step after that. After the support transition at $t = 3.2\,\text{s}$ however, it can be seen that no further changes in step size and timing are necessary, so the time to step curve reverts to a plain linear ramp, and the lateral GCV $\tilde{v}_{iy}$ is ramped down to zero.

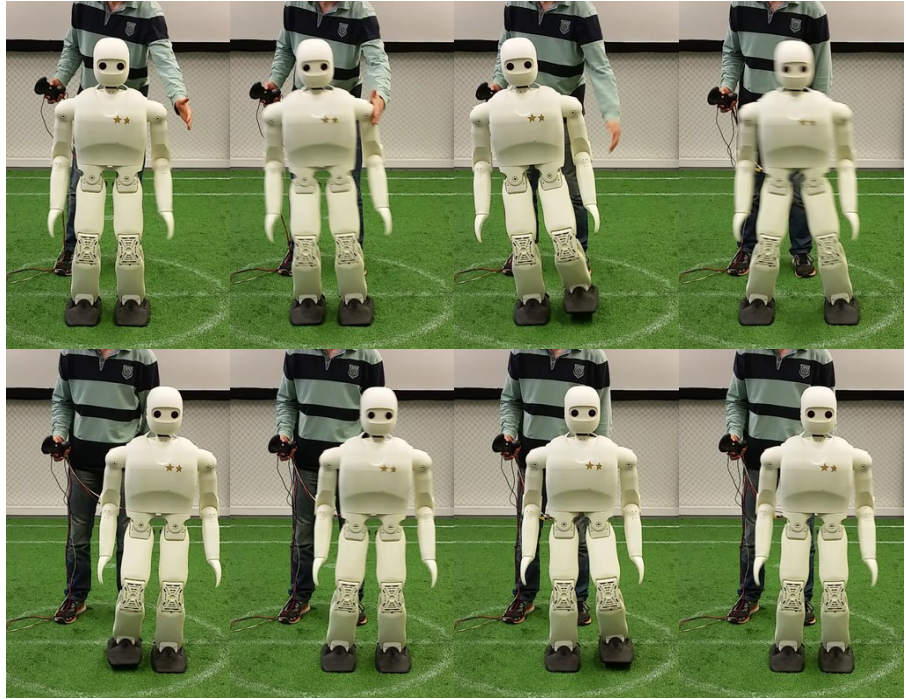

Figure 12.2: Image sequence of a lateral push on the NimbRo-OP2X robot. The capture step controller initially slows down the gait to wait for the robot to return towards the centre again, at which point it takes a large sideways step to absorb the extra lateral energy that it has gained through the push. A marginally enlarged follow-up step dissipates the remainder of the surplus energy, before subsequently returning to normal upright walking. Refer to Figure 12.3 for plots corresponding to this lateral push.

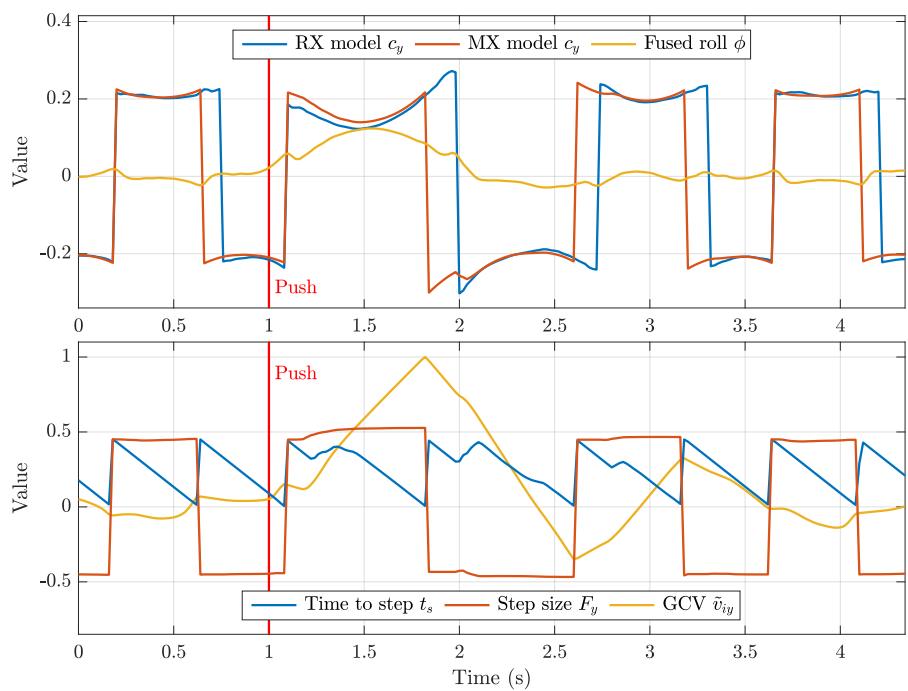

Figure 12.3: Plots corresponding to the lateral push of the NimbRo-OP2X robot shown in Figure 12.2.

Similar still images and plots for the case of a backwards sagittal push are shown in Figures 12.4 and 12.5 respectively. Referring to Figure 12.4, after the push (or rather pull) is applied to the robot (top row, second image), the large negative change in RX $c_x$ causes the MX model to adapt, resulting in a significant negative step size target $F_x$. The resulting quick first step (top row, third image) does not suffice for recapturing the balance of the robot, so a larger second step is performed (top row, fourth image) to halt the backwards tilt of the robot. Although the robot still oscillates forwards and backwards a bit over the next two steps (bottom row, second image), no significant further step size adaptations are required to return the robot to stable walking (bottom row, fourth image). This observation is confirmed by the plots in Figure 12.5. After the first two capture steps are executed, the calculated target step size $F_x$ quickly decreases in magnitude again, bringing the commanded sagittal GCV $\tilde{v}_{ix}$ along with it. Note that it is not in the nature of the capture step controller to produce truly zero step size adaptations when the robot is walking nominally. This is a downside of the formulation of the capture step controller (and associated predictive filter) that is avoided, for instance, by the step size adaptation scheme of the tilt phase controller (see Section 15.2.10).

The step timing given by the capture step controller was used successfully in multiple RoboCup competitions (see Videos 1.3 and 1.4), in particular in conjunction with the feedback mechanisms of the direct fused angle feedback controller (see Chapter 13). As discussed in the following section however, the capture step controller step size adjustments were not used directly in competition (2015–2018), mainly due to their lacking total reliability and robustness, but also because of their salient tendency to break robots (e.g. as experienced by Team NimbRo at RoboCup 2019).

## 12.3 DISCUSSION

The capture step controller demonstrates how despite great simplification of the dynamics of the robot, the resulting feedback control loop is still able to stabilise the robot in the face of strong disturbances and pushes that necessitate reactive stepping. One of the main strengths of the chosen approach is that there is no limitation to the type of disturbances that can be dealt with by the controller, assuming that the underlying CPG gait is still able to physically take the required reactive steps. This is because any disturbance that causes a deviation of the CoM state immediately results in a new planned footstep location that acts to restore the balance of the robot. There is no necessity for the capture step controller to estimate when exactly a disturbance occurs, what kind of disturbance it is, or how strong the disturbance is.

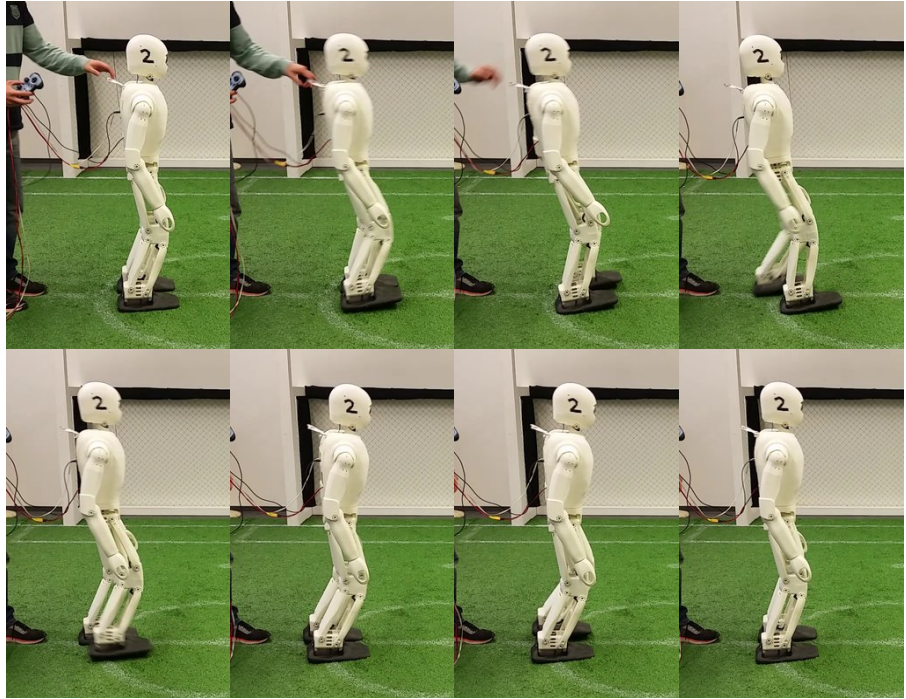

Figure 12.4: Image sequence of a sagittal push on the NimbRo-OP2X robot. After detecting the instability of the robot caused by the push, the capture step controller quickly commands two backward steps to recapture the balance of the robot. A further two steps are required before the small remaining sagittal oscillations are then also successfully dissipated. Refer to Figure 12.5 for plots corresponding to this sagittal push.

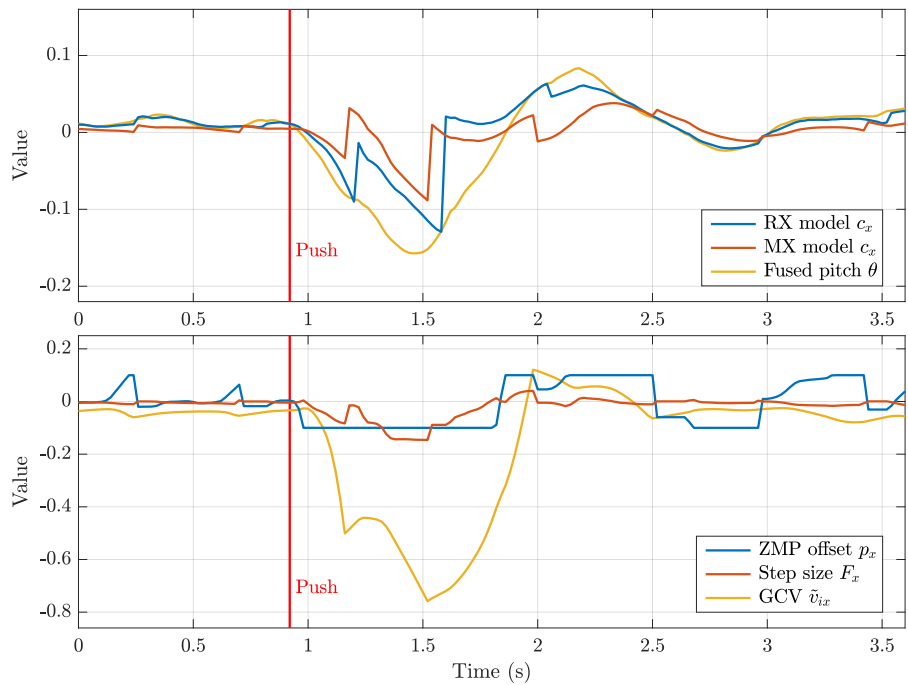

Figure 12.5: Plots corresponding to the sagittal push of the NimbRo-OP2X robot shown in Figure 12.4.

A clear distinction between the capture step controller and many other approaches to walking is that despite the fact that a prediction of the trajectory of the CoM is made to the end of the current step (and further to the apex of the following step for the purpose of lateral step size adjustment), there is no real requirement that the robot then actually follows that trajectory, and no effort is made to force it to track that trajectory. Clearly it is *desired* for the model to closely represent the future behaviour of the robot, but ultimately it is the model that needs to be adapted and tuned to follow the behaviour of the robot, not the behaviour of the robot that needs to be adapted to act according to the model. This is distinctly different to many ZMP-based gait generation methods that use a model to compute a target trajectory for the robot, and then try to force the robot to execute and track the calculated trajectory as closely as possible. As models can rarely comprehensively and accurately model the behaviour of a robot, this does not leave any room for the robot to find and take advantage of its own natural dynamics. One further difference of the capture step controller to many ZMP-based methods is that it is completely analytic and expressible in closed form, with computation times on the order of only a few microseconds.

There are many advantages of using the capture step controller for preserving the balance of a walking bipedal robot, but also some disadvantages. One strength of the controller is the RX-MX-TX predictive filter, which explicitly handles latencies in the actuator/sensor loop by always predicting the gait commands a certain time $t_l$ into the future. At the same time however, the predictive filter is also one of the main problems of the capture step controller when it comes to the relationship between the RX and MX models. Despite the effect of adaptation, there is nothing that actually *forces* the RX and MX models to stay synchronised, and for some robots this can be a big problem. If the RX and MX models lose synchronisation, there are significant portions of the gait cycle where the RX and MX models have different support leg signs. This severely impairs the ability of the robot to perform meaningful timing and step size adjustments, and the adaptation cannot work effectively in reuniting the two models. The problem of RX-MX synchronisation is more pronounced the more the natural behaviour of a robot deviates from the tuned LIPM, and in general it can be said that the capture step controller is highly sensitive to having its parameters tuned in a very accurate and balanced manner.

One further issue of the pure capture step approach is that it is difficult to get it to work effectively and robustly on robots that do not have a large amount of passive stability in their dynamics and in their open-loop gait. This is, for example, pronouncedly the case for the igus Humanoid Open Platform, and most other large low-cost robots without parallel leg kinematics. Robots of this kind were never tested by Missura (2015), and in particular no joint attempt to apply the original capture step framework to the igus Humanoid Open Platform was met with robust success.

Regarding the results of this thesis, while the capture step timing was used reliably to much success, and reasonable adjustments of the step sizes were also achieved, the robustness of the step size adjustments was the main limiting factor for applications like the Robo-Cup competition. The tendential instability of the serial leg kinematics in the sagittal direction of the robot makes the ZMP bounds of the support polygon a very blurry line to tune. At best it was possible to tune values that would make the robot engage in reactive stepping sometimes when it actually should not have, but then also sometimes not when it really should have. Despite this balanced form of tuning, it still occasionally occurred that the MX model would command an overly large step size in the sagittal direction due to a disturbance, and through the resulting instability of the model, quickly escalate with further steps, causing the robot to fall in a manner dangerous to the robot hardware. The possibility of such occurrences happening was not seen as tolerable in situations like RoboCup competitions, so step size adjustments were usually disabled at such events.

The context of the RoboCup competition highlighted a further limitation of the capture step controller. The fundamental idea of the capture step approach is to negate all disturbances purely via the means of reactive stepping. When walking along a precise desired trajectory, such as for example during a soccer game for the execution of ball handling skills, the step size adjustments were seen to interfere with the intended path of the robot, as well as its ability to stop walking when and where needed. This observation motivated the development of the two other gaits in this thesis, which view step size adjustments not as a first response, but only as a last resort for the preservation of balance, after all other thinkable ways of influencing the balance of the robot have already been tapped. These newer gaits also learnt from the difficulty of tuning the capture step controller, and effectively stayed free of any complex interplays of models, such as can be seen in the RX-MX-TX predictive filter.

# DIRECT FUSED ANGLE FEEDBACK CONTROLLER

As discussed in the previous chapter in Section 12.3, the capture step controller works well in adding push recovery ability to the Central Pattern Generator (CPG), but has difficulty choosing appropriately between action and inaction for common moderate disturbances of the robot. The capture step controller also uses step size adjustments far too readily, when, in order not to interfere with the desired walking trajectories of the robot, changes in step size should only be used as a last resort.

This chapter presents a novel balance controller, referred to as the direct fused angle feedback controller. It does not make any notable adjustments to the step sizes, but nonetheless has a strong stabilising effect on the open-loop CPG walking motions of the robot, through the use of timing adjustments and so-called corrective actions, as indicated in Figure 13.1. Corrective actions are inline modifications to the waveforms generated by the CPG, and are computed as a function of the trunk orientation feedback received from the attitude estimation in the form of *fused angles* (see Section 5.4.4). As the direct fused angle controller works by mechanisms other than step size adjustment, aside from the choice of which method of timing adjustment to use, it can coexist entirely and be used in conjunction with the capture step controller if desired. This option for coexistence is demonstrated in the released implementation of both controllers (and the CPG gait), as all three of them are implemented in the same gait engine.[1]

It is demonstrated by the direct fused angle feedback controller in this chapter that genuinely good walking results can be achieved using relatively simple feedback mechanisms, with only minimal modelling of the robot, and without measuring or controlling forces or torques. Only joint positions and a 6-axis Inertial Measurement Unit (IMU) are required for this method to work, making it flexible, portable and reliable for robots of all sizes, including in particular those with low-cost actuators and sensors. The main take-away message of the direct fused angle controller is that simple robot-agnostic feedback mechanisms are enough to make a robot walk if the sensor and feedback chains are carefully constructed, and that the feedback mechanisms can easily be run in parallel with more complicated step size adjustment schemes like the capture step controller, if desired, for greater overall walking robustness.

---

1 https://github.com/AIS-Bonn/humanoid_op_ros/tree/master/src/nimbro/
motion/gait_engines/cap_gait

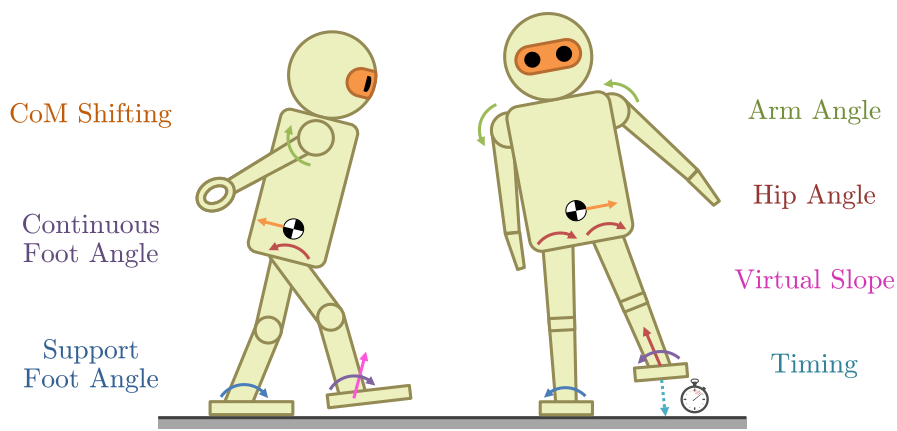

CoM Shifting

Continuous
Foot Angle

Support
Foot Angle

Arm Angle

Hip Angle

Virtual Slope

Timing

Figure 13.1: Summary of the timing feedback and corrective actions imple­mented as part of the direct fused angle feedback controller. For more details on each individual corrective action, refer to Section 13.2 and Figures 13.2 and 13.3.

## 13.1    GAIT STRUCTURE

The direct fused angle controller takes as input the estimated fused pitch $\theta_B$ and fused roll $\phi_B$ (see Section 5.4.4) of the trunk of the robot, as estimated by the attitude estimator (see Section 10.1). Based on the deviations of these two angles from their nominal expected waveforms, the controller calculates a gait frequency offset $f_{go}$ in Hertz for the purpose of CPG timing feedback as per Equation (11.10) (see Section 11.1.3.1), and implements the required corrective actions using additional additive CPG waveform components in the abstract arm motion, abstract leg motion and inverse leg motion functions (see Section 11.2). The final calculated CPG joint commands are then passed to the actuator control scheme as normal.

## 13.2    CORRECTIVE ACTIONS

Numerous different corrective actions have been implemented in the direct fused angle controller. Corrective actions can be thought of as motion primitives that are dynamically weighted and superimposed on top of the open-loop CPG gait waveforms, so as to affect or bias the balance of the robot in some way. Six different corrective actions (illustrated in Figures 13.2 and 13.3) have been implemented in a mix of the abstract and inverse pose spaces (refer to Sections 9.1.2 and 9.1.3). The implemented corrective actions are as follows:

1. **Arm angle:** The abstract arm angles $\phi_{ax}$, $\phi_{ay}$ are adjusted to bias the balance of the robot and produce reaction moments that help counterbalance transient instabilities. As an example, if both arms are moved backwards, the balance of the robot is tendentially shifted backwards as a result.

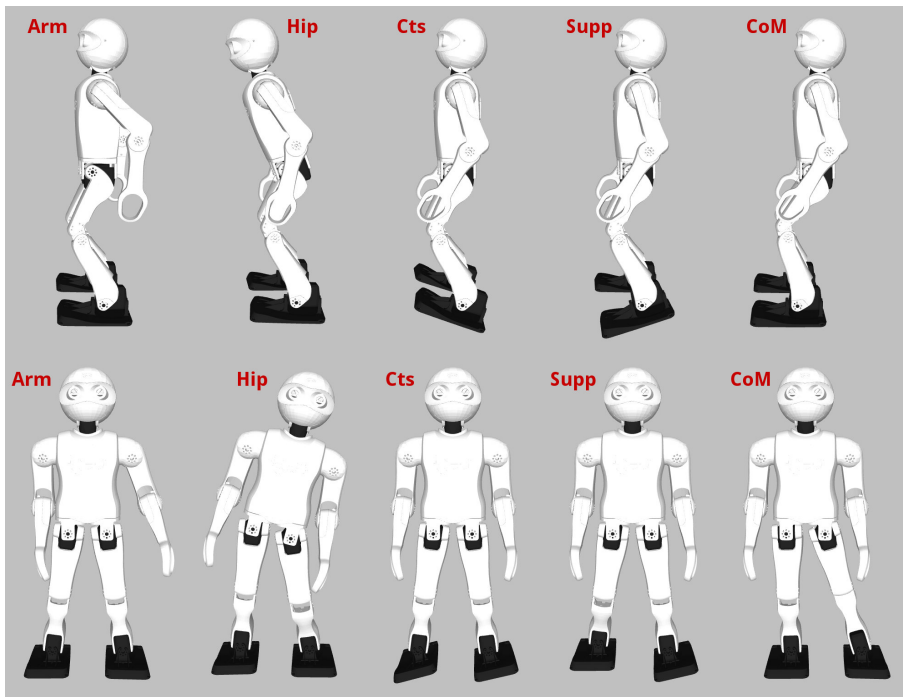

Figure 13.2: The implemented corrective actions in both the sagittal (top row) and lateral (bottom row) planes—from left to right in both cases, the arm angle, hip angle, continuous foot angle, support foot angle, and CoM shifting corrective actions. See Figure 13.3 for the virtual slope corrective action. The actions in this figure have been exaggerated for the purposes of clearer illustration.

2. **Hip angle:** The torso of the robot is tilted within the lateral and sagittal planes to induce lean in a particular direction. As for the lateral leaning component of the CPG, the leg retraction parameter of one of the legs is trimmed to ensure that no disparity in foot height occurs as a result (see Equation (11.30) and Figure 11.4).

3. **Continuous foot angle:** Continuous offsets are applied to the abstract foot angles $\phi_{fx}$, $\phi_{fy}$ to bias the tilt of the entire robot from the feet up. For example, if the feet are both offset so that the toe area points further downwards, the balance of the robot is biased in the backwards direction.

4. **Support foot angle:** Gait phase-dependent offsets are applied to the abstract foot angles $\phi_{fx}$, $\phi_{fy}$ of the support foot, in order to induce ground reaction forces that push the balance of the robot in the opposite direction. The offsets are faded in linearly starting at the instant of foot strike, and are faded out linearly just in time for leg lift-off. As such, the support foot angle offsets are only applied to each foot during their single and double support phases, and not during their swing phase.

5. **CoM shifting:** The inverse kinematic positions of the feet relative to the torso are adjusted in the horizontal plane to shift the position of the Centre of Mass (CoM). The centring of the robot's mass above its support polygon is thereby adjusted, influencing its balance. As an example of such influence, shifting the CoM to the right for instance increases the proportion of time spent on the right foot during walking.

6. **Virtual slope:** The inverse kinematic positions of the feet relative to the torso are adjusted in the vertical direction in a gait phase-dependent manner to lift the feet more at one swing extremity. This can be thought of as what the robot would need to do in order to walk up or down a slope. Figure 13.3 shows the effect of the virtual slope corrective action for a robot that is both walking and tilted sagittally forwards.

## 13.3  FUSED ANGLE FEEDBACK MECHANISMS

Each of the aforementioned corrective actions (and timing feedback) are activated and driven based on an array of feedback mechanisms that use the estimated fused pitch and roll of the torso as their source. Each feedback mechanism calculates its contribution to the activations of the corrective actions as a function of the deviations of the fused pitch $\theta_B$ and fused roll $\phi_B$ angles from their nominal limit cycle values. The nominal limit cycles of the fused angles are modelled as parameterised sine functions of the gait phase $\mu_i$, and the fused angle deviations $d_\theta, d_\phi$ are defined as the difference between $\theta_B$, $\phi_B$ and their current expected values $\theta_{exp}$, $\phi_{exp}$. That is,

$$d_\theta = \theta_B - \theta_{exp}, \tag{13.1a}$$

$$d_\phi = \phi_B - \phi_{exp}, \tag{13.1b}$$

where

$$\theta_{exp} = k_{eoy} + k_{emy} \sin(\mu_i - k_{epy}), \tag{13.2a}$$

$$\phi_{exp} = k_{eox} + k_{emx} \sin(\mu_i - k_{epx}), \tag{13.2b}$$

where $k_*$ are constants that are tuned based on CPG walking experiments to define the expected fused angle waveforms.

An overview of the complete feedback mechanism calculation pipeline, from the fused angle deviations $d_\theta$, $d_\phi$ through to the resulting corrective actions and timing adjustment, is shown in Figure 13.4. The algorithmic details of the various feedback mechanisms are discussed in the following subsections. While the timing adjustment ends with the calculation of a gait frequency offset $f_{go}$, and the virtual slope ends with the calculation of the required inverse kinematic adjustment of the foot locations, the remaining three feedback

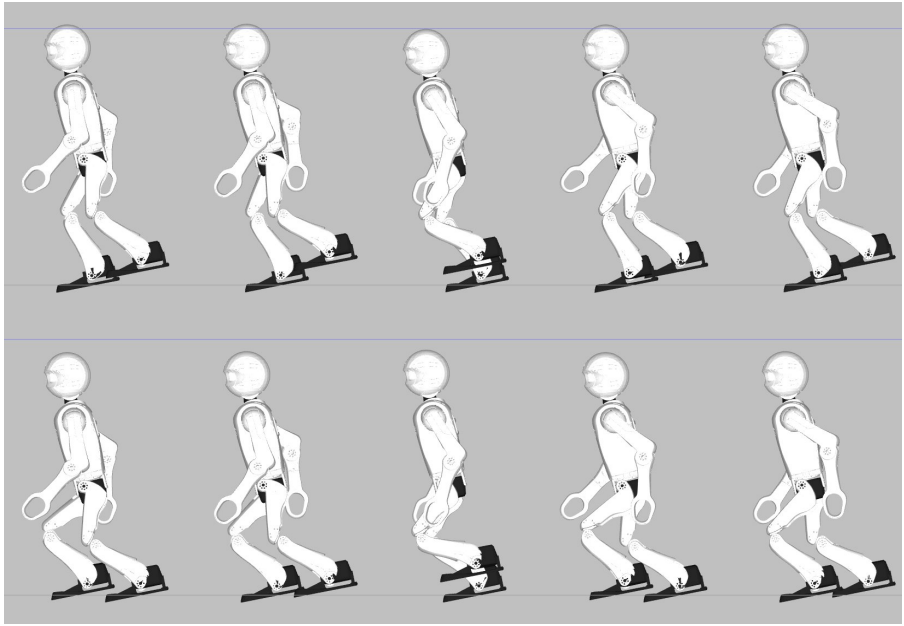

Figure 13.3: Illustration of the effect of the virtual slope corrective action for a robot that is both walking and tilted sagittally forwards. The top row of images shows a sequence of frames captured as the left leg is making a step with the pure CPG waveforms. From left to right, the frames are at the gait phase instants of commanded right foot strike, left foot lift-off, mid-swing, left foot strike, and right foot lift-off. Clearly, at the intended instants of support exchange the feet are nowhere close to level with each other, and in actual fact prematurely strike the ground before each step is complete. In addition to impeding forwards locomotion and causing unnecessary destabilisation of the robot, this also leads to significant oscillations in the CoM height of the robot. The second row shows the same CPG walking waveforms with the effect of the virtual slope corrective action included. The support transitions work as expected and the CoM height oscillations are reduced. The lower overall CoM height of the resulting waveforms is unavoidable, as otherwise leg extension limits would be reached during the gait cycle.

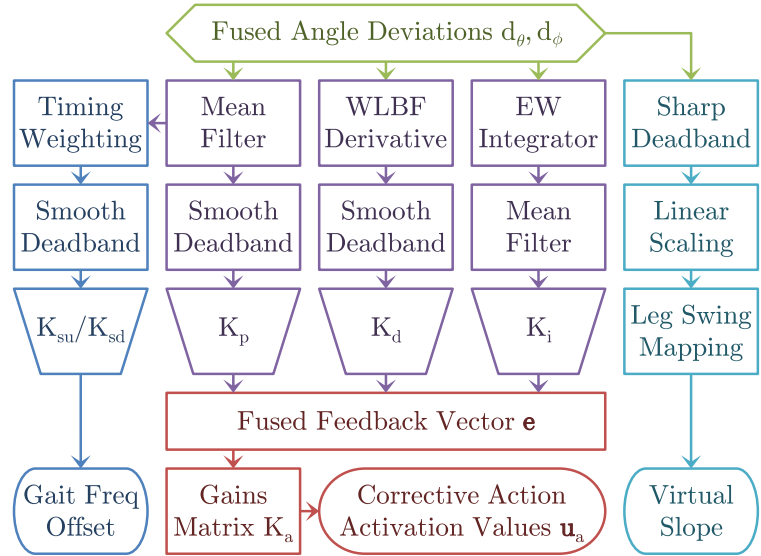

Figure 13.4: Overview of the fused angle feedback calculation pipeline.

mechanisms—namely the proportional, derivative and integral feedback mechanisms—are responsible for the activation of the remaining five corrective actions (i.e. the ones in Figure 13.2). This is done in each execution cycle of the direct fused angle controller through the calculation of a fused feedback vector

$$\mathbf{e} = \begin{bmatrix} e_{Px} & e_{Py} & e_{Dx} & e_{Dy} & e_{Ix} & e_{Iy} \end{bmatrix}^T, \tag{13.3}$$

which summarises the amount of proportional, derivative and integral feedback there should be in the lateral (x) and sagittal (y) directions at that instant of time. The calculated fused feedback vector $\mathbf{e}$ is converted to a vector $\mathbf{u}_a$ of corrective action activation values using the equation

$$\mathbf{u}_a = K_a \mathbf{e}, \tag{13.4}$$

where $K_a$ is the configured (constant) corrective action gains matrix. Note that a single $5 \times 6$ gains matrix $K_a$ is indicated here for generality and mathematical brevity, but in practice a full tune involves only 10–14 non-zero gains in the matrix. These intended feedback paths are explicitly listed throughout the following subsections. The entries of $\mathbf{u}_a$ give the magnitude and sign with which each of the available corrective actions—excluding the virtual slope, which is calculated and added separately (see Section 13.3.5)—are applied to the abstract and inverse poses generated by the CPG gait. This is the means by which the robot acts to keep its balance.

To ensure that the pose of the robot stays within suitable joint limits, soft coercion (a form of smooth saturation, see Appendix A.1.2.3) is applied to the final calculated abstract arm angles, leg angles, foot angles, and inverse kinematic adjustments to the CoM.[2] Soft coercion is used instead of standard hard coercion (i.e. instead of simply applying hard limiting bounds), because soft coercion is continuously differentiable (of class $\mathcal{C}^1$) and results in smoother saturation behaviour of the limbs, with less self-disturbances of the robot.

### 13.3.1 Fused Angle Deviation Proportional Feedback

The most fundamental and direct form of fused angle feedback is the proportional corrective action feedback. As with nearly all of the implemented feedback mechanisms, the proportional feedback operates in both the lateral and sagittal planes. The fused angle deviation values $d_\theta$, $d_\phi$ are first passed through a mean filter (see Appendix A.2.1.1) to mitigate the effects of noise. Smooth deadband (see Appendix A.1.3.3) is then applied to the output to inhibit corrective actions when the robot is close to its intended trajectory, and thereby avoids oscillations due to hunting. Smooth deadband was chosen for this application (as opposed to standard sharp deadband, see Appendix A.1.3.1) to soften the transitions between action and inaction, and to avoid any unnecessary self-disturbances and/or unexpected oscillatory activation-deactivation cycles. These can occur if sharp deadband is used, due to its strongly asymmetrical behaviour around its threshold points. The proportional fused angle feedback values $e_{P_x}$ and $e_{P_y}$ from Equation (13.3) are calculated by scaling the results of the smooth deadband by the proportional fused angle deviation gain $K_p$.

The proportional fused angle feedback mechanism is intended for activating the *arm angle*, *support foot angle*, and *hip angle* corrective actions, as these have the most direct effect on the transient behaviour of the robot.

### 13.3.2 Fused Angle Deviation Derivative Feedback

The proportional feedback works well in preventing falls when the robot is disturbed, but if used alone has the tendency to produce potentially unstable low frequency oscillations of the robot, in particular in the sagittal direction due to compliance effects in the ankles. To enhance the transient disturbance rejection performance of the robot, corresponding derivative feedback terms are incorporated to add damping to the system. If the robot has a non-zero angular velocity, this component of the feedback reacts 'earlier' to cancel out the velocity before a large fused angle deviation ensues, and the proportional feedback has to combat the disturbance instead.

---

2 Recall the soft coercion that is performed in Step (ii) on page 324, and Step (ii) on page 330. In addition to these pre-existing steps of the CPG, the inverse kinematic offset to the foot locations resulting from the CoM shifting is also soft-coerced.

The derivative fused angle feedback values $e_{Dx}$ and $e_{Dy}$ from Equation (13.3) are computed by passing the fused angle deviations $d_\phi$ and $d_\theta$ through a Weighted Line of Best Fit (WLBF) derivative filter (see Appendix A.2.4.1), applying smooth deadband, and then scaling the results by the derivative fused angle deviation gain $K_d$. The smooth deadband, like for the proportional case, is to ensure that no corrective actions are taken if the robot is within a certain threshold of its normal walking limit cycle, and ensures that the transition from inaction to action and back is smooth.

A Weighted Line of Best Fit (WLBF) filter observes the last $N$ data points of a signal, in addition to assigning confidence weights to each of the data points, and performs weighted linear least squares regression to fit a line to the data, with the data measurement timestamps being used as the independent variable. The linear fit evaluated at the current time gives a smoothed data estimate, and the fitted linear slope gives a smoothed estimate of the derivative of the data stream. WLBF derivative filters have a number of advantages, in particular when compared to the alternative of computing the numerical derivative of a signal and applying a low pass filter. WLBF filters are more robust to high frequency noise and outliers in the data for the same level of responsiveness to input transients, and have the advantage of inherently and easily supporting non-constant time separations between data points in a stable manner, even if two data points are very close in time. The use of weights in WLBF filters also allows some data points to be given higher confidences than others in a probabilistic manner. For example, in regular parts of the gait cycle where noise and errors in the measurements are expected, lower weights can be used to help reject noise and disturbances in the sensor measurements. To the knowledge of the author, the use of weighted linear least squares regression in this online fashion for the purposes of data smoothing and derivative estimation, in particular in the context of walking gaits, has not previously been published, so no references can be provided.

The derivative fused angle feedback values $e_{Dx}$ and $e_{Dy}$, are intended for activating the *arm angle*, *support foot angle* and *hip angle* corrective actions, and serve to allow better loop shaping for the transient response to disturbances.

### 13.3.3   Fused Angle Deviation Integral Feedback

The proportional and derivative fused angle feedback mechanisms are able to produce significant improvements in walking stability, but situations may arise where continued corrective actions are required to stabilise the robot, due to for example asymmetries in the robot. The continual control effort and resulting periodic steady state errors are undesired. The purpose of the integral feedback is to slowly converge to offsets to the gait halt pose (see Section 11.2) that minimise the need for control effort during general walking.

The fused angle deviations $d_\phi$ and $d_\theta$ are first integrated over time using an Exponentially Weighted (EW) integrator (see Appendix A.2.3.1), a type of 'leaky integrator'. This kind of integrator incrementally computes the sum

$$I[n] = x[n] + \alpha x[n-1] + \alpha^2 x[n-2] + \cdots \qquad (13.5)$$

where $\alpha \in [0, 1]$ is the so-called history time constant, and $x[\cdot]$ is the data to integrate. The sum is most conveniently computed using the difference equation

$$I[n] = x[n] + \alpha I[n-1]. \qquad (13.6)$$

If $\alpha = 0$, the integrator simply returns the last data value, but if $\alpha = 1$ the output is the same as that of a classical integrator, so the value of $\alpha$ effectively trims the amount of memory that the integrator has. A suitable value for $\alpha$ is computed from the desired half-life time $T_h$, which is a measure of the decay time of the integrator, using

$$\alpha = 0.5^{\frac{\Delta t}{T_h}}, \qquad (13.7)$$

where $\Delta t$ is the nominal time step. An EW integrator is used instead of a standard integrator for flexibility and stability reasons, to combat integral windup, and because old data eventually needs to be 'forgotten' while walking, because situations can change. The alternative of keeping the last $N$ fused angle deviations in a buffer and integrating over the buffer is less efficient, less continuous in the output and less flexible, in addition to being vulnerable to aliasing effects.

As indicated in Figure 13.4, the outputs of the two fused angle deviation EW integrators (one for $d_\theta$ and one for $d_\phi$) are passed through respective mean filters (see Appendix A.2.1.1) to allow the trade-off between settling time and level of noise rejection to be trimmed. The smoothed outputs of the mean filters are then scaled by the integral fused angle deviation gain $K_i$ to give the integral fused angle feedback values $e_{Ix}$ and $e_{Iy}$, as required for the calculation of the fused feedback vector $\mathbf{e}$ in Equation (13.3). The integral fused angle deviation feedback is intended for activating a mix of the *arm angle*, *hip angle*, *continuous foot angle*, and *CoM shifting* corrective actions, although to preserve greater independence between the feedback mechanisms, primarily the last two of these four are recommended.

### 13.3.4 Fused Angle Deviation Timing Feedback

The Proportional-Integral-Derivative (PID) feedback mechanisms act to return the robot to the expected fused angle limit cycles during walking. For larger disturbances this is not always possible, practicable and/or desired from a timing perspective. For instance, if a robot is pushed so far laterally that it is only resting on the outer edge of its support foot, then no matter what fused angle feedback mechanisms are applied, it is infeasible to return the robot to its centred and upright position in the same time as during normal walking. This generally results in the robot attempting to perform its next step anyway, despite being tilted, and falling as a result. To deal with this mode of failure, timing feedback has been implemented. This form of feedback adjusts the rate of progression of the gait phase as a function of the fused roll deviation $d_\phi$. Steps taken by the robot can thereby be sped up or slowed down based on the lateral balance state of the robot.

The required sign of the timing feedback depends on the current gait phase-dependent support leg sign, and only minimal timing adjustments are desired during the double support phase due to the sensor noise associated with foot strike and the transferral of the robot's weight from one foot to the other. As a result, the timing feedback is constructed by taking the mean filtered fused roll deviation $\bar{d}_\phi$ from the proportional feedback pipeline, and weighting it by a saturated oscillatory gait phase-dependent expression to give

$$\tilde{e}_{Tx} = \bar{d}_\phi \, \text{coerce}\left(-K_{tw}\sin(\mu_i - \tfrac{1}{2}D), -1, 1\right), \tag{13.8}$$

where coerce($\cdot$) is standard hard coercion (see Appendix A.1.2.1), $\mu_i$ is the gait phase, $D$ is the double support phase length, and $K_{tw} \in [1, \infty)$ is a gain that adjusts the shape of the oscillatory weighting curve. To avoid unnecessary disturbances to the gait frequency during normal walking, smooth deadband (see Appendix A.1.3.3) is applied to the calculated weighted deviation $\tilde{e}_{Tx}$ to give the final fused angle timing feedback value $e_{Tx}$. The required gait frequency offset $f_{go}$ (see Section 11.1.3.1) is then given by

$$f_{go} = \begin{cases} K_{su}e_{Tx} & \text{if } e_{Tx} \geq 0, \\ K_{sd}e_{Tx} & \text{otherwise,} \end{cases} \tag{13.9}$$

where $K_{su}$ and $K_{sd}$ are configured speed-up and slow-down gains. The calculated gait frequency offset $f_{go}$ is then used in the standard CPG gait frequency update equation, given by Equation (11.10).

### 13.3.5 Virtual Slope Feedback

If the robot is tilted sagittally in the direction that it is walking, it can happen that a foot prematurely strikes the ground during its swing phase. This is undesirable and a cause for destabilisation of the robot. The virtual slope corrective action adjusts the inverse kinematic height of the feet in a gait phase-dependent manner as a function of the estimated fused pitch $\theta_B$ and the sagittal component $v_{ix}$ of the internal

**Gait Command Velocity** (GCV). This is used to effectively simulate as if the robot was walking up or down a slope, hence the name of the corrective action. For example, if the robot is walking forwards and tilted forwards, the robot with virtual slope active will lift its feet higher at the front of its swing phases, ensuring that the legs do indeed reach full forward swing before establishing ground contact.

The required sagittal virtual slope $V_s$ is calculated at each point in time by first applying sharp deadband (see Appendix A.1.3.1) to the measured fused pitch $\theta_B$ of the trunk, then linearly scaling the result by one of two gains, depending on whether the robot is tilted in the same direction that it is walking or not. The sagittal virtual slope feedback value $e_{Vy}$ is then calculated using the formula

$$e_{Vy} = V_s v_{ix}, \tag{13.10}$$

where $v_{ix}$ is the sagittal component of the current internal GCV vector $\mathbf{v}_i$. The calculated value of $e_{Vy}$ is taken as the maximum required magnitude of inverse kinematic height adjustment of the feet during the current step. Recalling from Equation (11.19) that $\zeta_s$ is the leg-dependent *swing angle* used in the generation of the CPG waveforms, the exact required inverse kinematic height adjustment at every instant in time is given independently for both legs by the formula

$$p_{lz} \leftarrow p_{lz} + \tfrac{1}{2}\big(\zeta_s + \operatorname{sign}(e_{Vy})\big), \tag{13.11}$$

where $p_{lz}$ is the z-component of the inverse space ankle position vector $\mathbf{p}_l$ (see Section 9.1.3), and $\operatorname{sign}(\cdot)$ is a sign function such that $\operatorname{sign}(0) = 1$.

### 13.3.6 Tuning of the Feedback Mechanisms

The process of tuning the feedback mechanisms is greatly simplified by their considerable independence. The individual feedback mechanisms have clearly observable and direct effects that can be precisely isolated and tested, so arguably the process of tuning the proposed gait is quicker and easier than it would be for most model-based approaches that do not somehow work out of the box. For instance, in experiments it was always observed that the direct fused angle controller is *significantly* easier to tune for a robot than the capture step controller.

The process of tuning starts with the Proportional-Derivative (PD) gains, as these are most responsible for the transient disturbance response characteristics of the robot. The most effective corrective actions in each of the sagittal and lateral planes of motion are identified for each robot, and the gain ranges that produce noticeable effects without risking oscillations or instabilities are also experimentally established. The choice of corrective actions may be influenced by external objectives, such as for example the desire to minimise trunk angle deviations from upright. In this case, the hip angle feedback components may be omitted for example.

Once the transient response has been tuned, a suitable integral feedback half-life time is chosen based on the rate at which the robot should adapt to changes in its environment, and gains are chosen that bring the activations of the associated integral corrective actions to the desired range. Timing is then considered, with the speed-up and slow-down gains being selected to avoid premature stepping, and to instantaneously halt the gait phase when the robot reaches a certain nominal lateral angular deviation. The virtual slope mechanism is mostly parameter-free, and the sole pair of linear scaling factors are chosen to provide the desired margin of clearance of the foot from the ground during maximum forwards walking.

## 13.4  EXPERIMENTAL RESULTS

The proposed gait has been implemented and evaluated experimentally on all of the robots listed in Section 2.1.2, including in particular all seven igus Humanoid Open Platform robots that were constructed over the years, as well as the Dynaped robot. As a matter of cross-validation, the direct fused angle controller was observed to work particularly well on Dynaped, despite the fact that essentially the same controller parameters and gains matrix $K_a$ were used as on the igus Humanoid Open Platform. This demonstrates the independence and portability of the feedback controller, which is a great strength of the gait that derives from its essentially model-free nature.

Overall, the feedback mechanisms were observed to make a significant difference to the walking ability of the tested robots—as demonstrated in detail in Video 13.1 for each feedback mechanism individually—with walking often not even being possible for extended periods of time without them. The feedback mechanisms also equipped the robots with disturbance rejection capabilities that were not present otherwise. Reliable omnidirectional walking speeds of 21 cm/s were achieved for the igus Humanoid Open Platform, and 36 cm/s was achieved for Dynaped. Both of these walking speeds were measured on an artificial grass surface of blade length 32 mm.

Plots of experimental results demonstrating the efficacy of the feedback mechanisms are shown in Figure 13.5. In Figure 13.5a, the robot walked twice onto an unexpected 1.5 cm step change in floor height, the first time with proportional and derivative feedback enabled, and the second time with them disabled. It can be seen that with the feedback enabled the robot is able to avoid falling—albeit with a large steady state error in the fused pitch—while without feedback the robot falls immediately. Adding in integral feedback, configured to activate both the CoM shift and continuous foot angle corrective actions, produces the results in Figure 13.5b. It can be seen that the relatively large initial error in the fused pitch is rejected within about 5 s, with the robot converging to upright walking despite being on an uneven floor. Note that the residual fused pitch limit cycles towards the end of the plot are an effect of the robot's feet not being equally positioned on the step change in floor height.

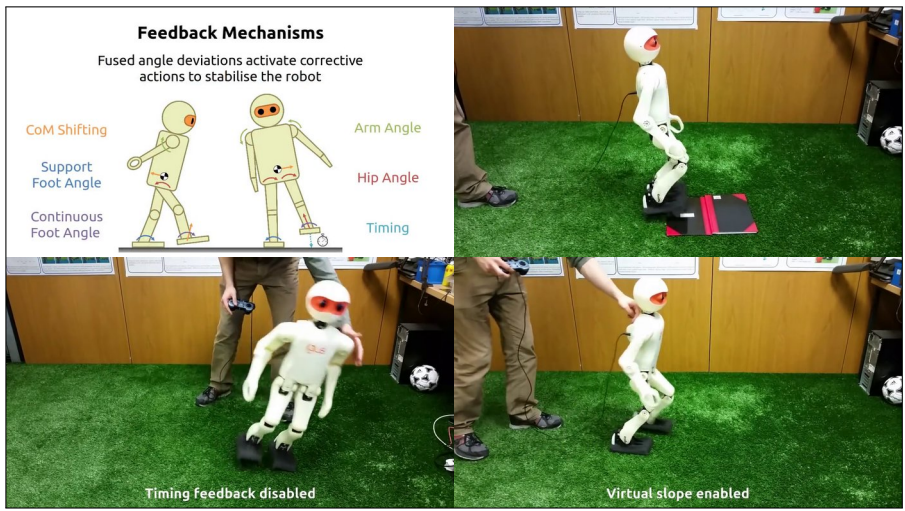

Video 13.1: Walking experiments demonstrating the feedback mechanisms implemented as part of the direct fused angle feedback controller. These are the exact experiments (for the most part) for which the plots are provided in Figure 13.5.
https://youtu.be/xnzJi2hTfAo
*Omnidirectional Bipedal Walking with Direct Fused Angle Feedback Mechanisms*

Figure 13.5c shows the effect of timing feedback. Timing feedback activates when there are significant deviations in the lateral direction between the measured and expected orientations of the robot. A plot comparing the measured and expected fused pitch and roll values for stable balanced walking is shown in Figure 13.6. Despite initial appearances, the deviations from expected are actually quite small ($< 1.1°$), and only larger deviations should trigger changes to the nominal step timing. In Figure 13.5c, the robot was subjected to two lateral pushes while walking in place, the first with timing feedback enabled, and the second without. It can be seen at $t = 3.1$ s that the gait phase slows down in response to the push, allowing the fused roll to return to its expected limit cycle within the following 2.5 s. An identical push at $t = 8.2$ s, without timing feedback enabled, causes the robot to fall.

Figure 13.5d illustrates the effect of the virtual slope corrective action. The robot was continuously pushed forwards while it was walking forwards, first with the virtual slope feedback enabled, and then without. In the first case, the adjustments to the inverse kinematic height of the feet ensured that the robot could continue to walk

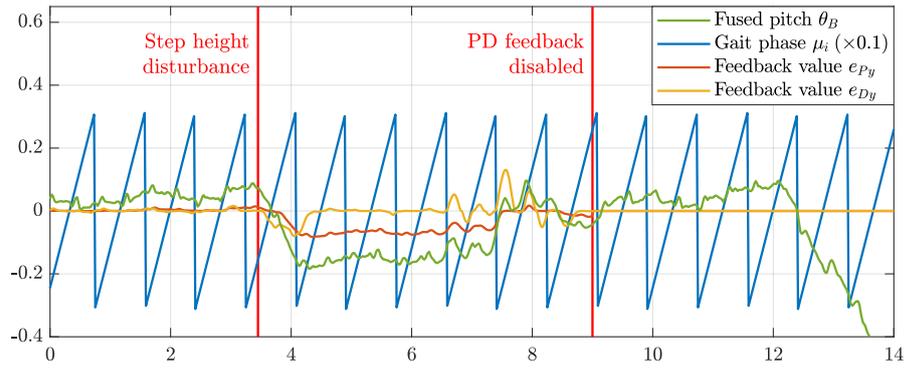

(a) Effect of PD feedback when walking onto a step change in floor height.

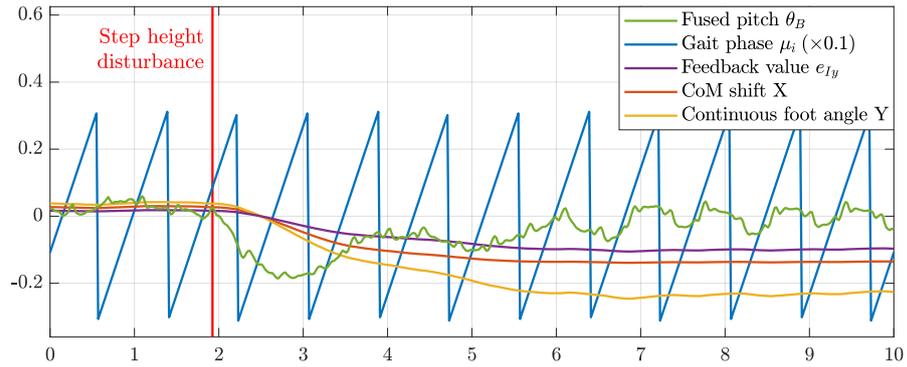

(b) Effect of I feedback when walking onto a step change in floor height.

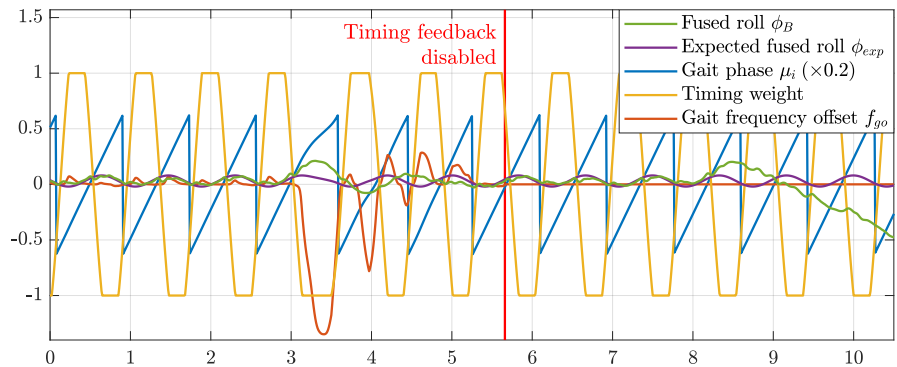

(c) Effect of timing feedback on the lateral transient response to pushes.

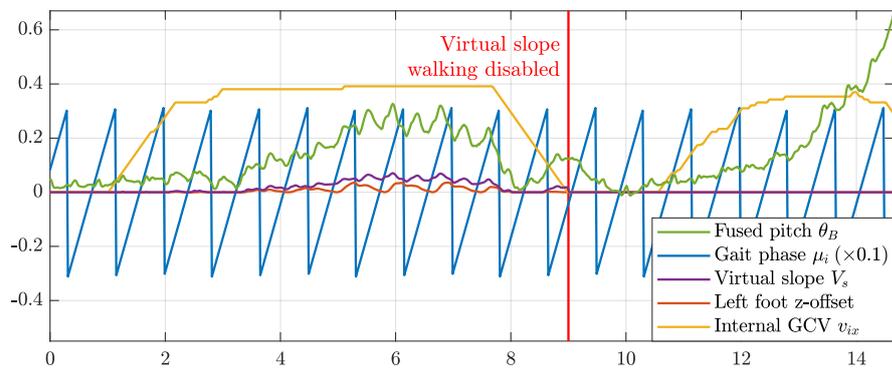

(d) Effect of virtual slope feedback while walking and being pushed forwards.

Figure 13.5: Plots of experimental results showing the effects on the stability
of the robot of the five implemented feedback mechanisms.

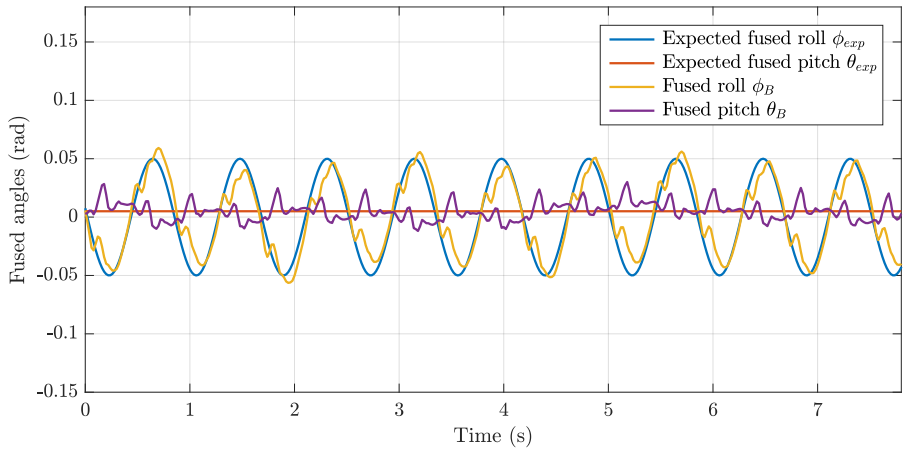

Figure 13.6: Plot of the fused angle waveforms $\theta_B$, $\phi_B$, along with the expected fused angle waveforms $\theta_{exp}$, $\phi_{exp}$, during on-spot walking. The close correlation between the expected and true orientations of the robot can be observed. Note that the plot is quite enlarged, so the deviations from expected in this case are actually all less than $\approx 0.02\,\text{rad} = 1.1°$.

forwards and regain balance once it was no longer being pushed. In the latter case however, the feet started to collide with the ground in front of the robot, preventing the robot from taking its intended step sizes, and causing significant self-destabilisation. The robot was not able to get its feet underneath its CoM again, and subsequently fell shortly after.

A more detailed idea of how the PD corrective actions activate in the face of push disturbances is given in Figure 13.7. The top plot shows the response of an igus Humanoid Open Platform robot that was pushed forwards while walking on the spot on artificial grass, while the bottom plot shows an equivalent backwards push. In both cases, the fused angle deviation proportional and derivative feedback components were enabled, and used to activate the sagittal arm angle and support foot angle corrective actions. Prior to both pushes, it can be observed from the plots that no significant activations of the feedback mechanisms were occurring. As soon as the destabilisation caused by the pushes was detected however, i.e. via the ensuing deviation caused in the measured fused pitch, the arms and feet sprung into action to counteract the disturbance and return the robot to a stable walking limit cycle. Note that the support foot angle activations alternate between the left and right feet as the robot takes steps, as it only makes sense to adjust the orientation of the foot that is actually on the ground. Although the activations reacted very quickly after the push, and spiked to a 'large' value, the derivative feedback component ensures that as the robot dissipates energy and starts to return towards upright, the magnitude of the activations quickly drops again, and in fact even briefly reverses so as to prevent excessive overshoot. Sagittal overshoot is a common problem of the open-loop CPG. It causes the pitch of the robot to oscillate back and forth in an only lightly damped manner, and makes the robot more sensitive to any further disturbances. This is actively prevented by the feedback controller, leading to greater walking stability.

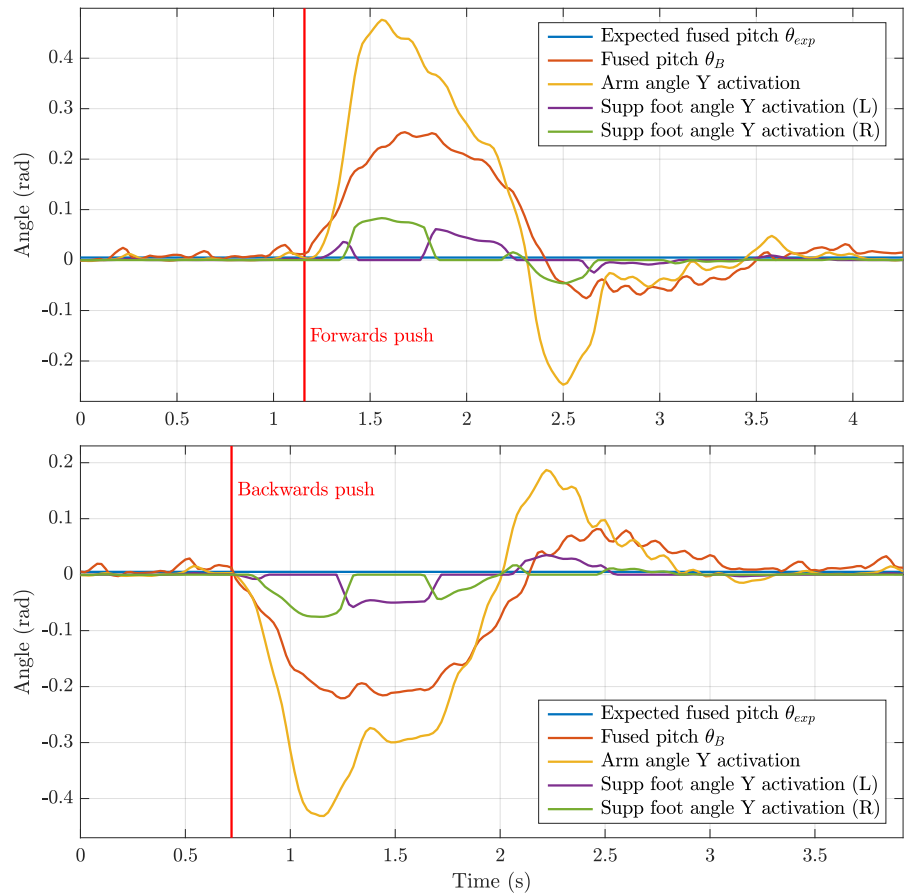

Figure 13.7: Plots of the response of the direct fused angle feedback controller to sagittal pushes. The top plot shows a forwards push during which the robot reaches a maximum forwards tilt of 0.253 rad, while the bottom plot shows a backwards push that results in a maximum backwards tilt of 0.22 rad. Both disturbances were rejected primarily via the arm angle and support foot angle corrective actions, the activations of which are shown in the plot.

A more holistic qualitative demonstration of the proposed gait (i.e. the CPG gait combined with the direct fused angles feedback controller) and its resistance to disturbances can be found in Video 13.2. Experiments and real-life situations are shown for multiple different robot platforms, including Dynaped, the igus Humanoid Open Platform, NimbRo-OP2 and NimbRo-OP2X, and impressions of walking performance in simulation are also given. Notable aspects of the video include the wide variety of conditions and disturbances that the gait is confronted with, as well as the fast maximum walking speeds that were achieved on the NimbRo-OP2X ($\approx$59 cm/s).

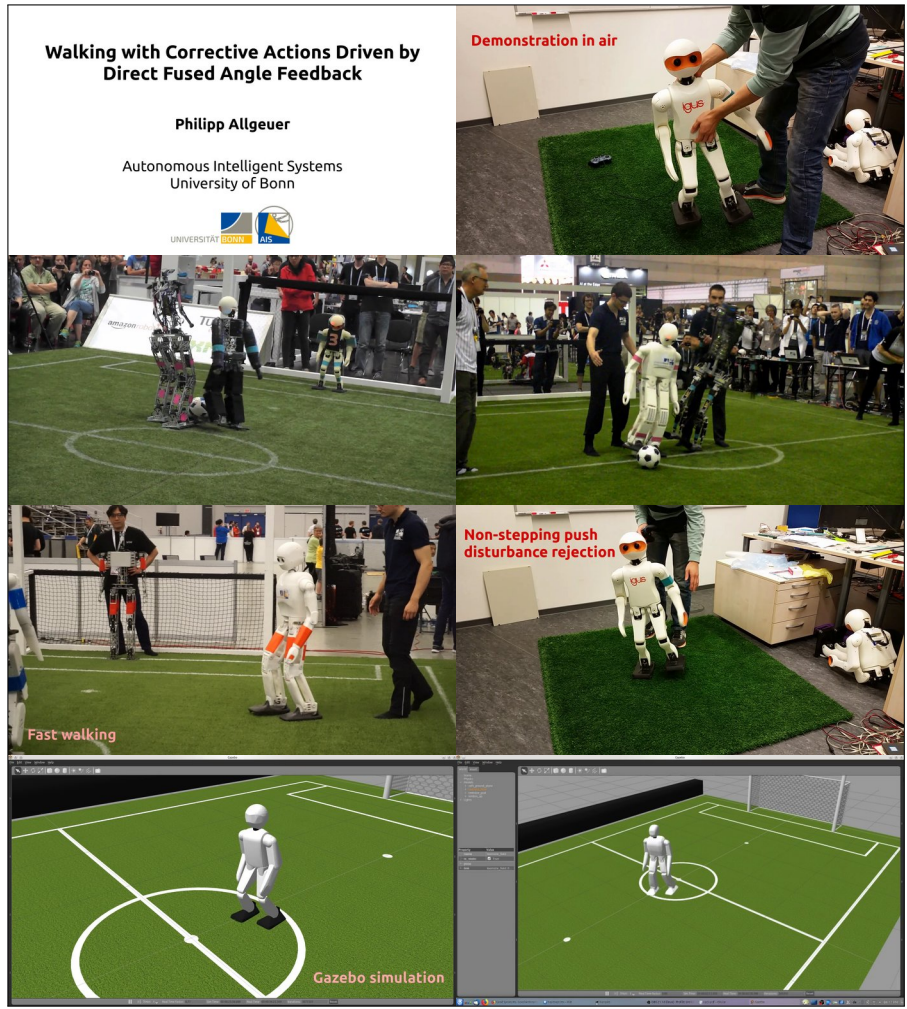

Video 13.2: Demonstration of the closed-loop direct fused angle feedback gait. Simple corrective actions based on the estimated fused angles of the torso are used to maintain the balance of the robot, even in the face of disturbances. The gait is demonstrated on the igus Humanoid Open Platform, Dynaped, NimbRo-OP2 and Nim-bRo-OP2X robots, as well as in simulation.
https://youtu.be/DvxZJVVRdyE
*Walking with Corrective Actions Driven by Direct Fused Angle Feedback*

Table 13.1: Number of withstood simulated pushes (out of 20) when walking on the spot with the direct fused angle feedback controller

| Impulse (s N) | 0.3 | 0.6 | 0.9 | 1.2 | 1.6 | 2.0 |
|---|---|---|---|---|---|---|
| **Open-loop** | 20 | 13 | 11 | 6 | 5 | 3 |
| **Closed-loop** | 20 | 20 | 20 | 19 | 17 | 13 |

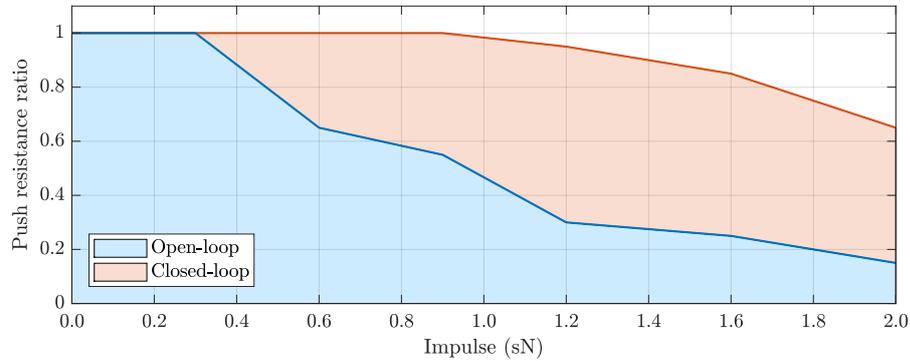

Figure 13.8: Plot of the ratio of withstood pushes against push impulse magnitude for a simulated igus Humanoid Open Platform walking on the spot with the Central Pattern Generator and direct fused angle feedback controller. The performance of the robot with and without the controller activated is contrasted. The raw data corresponding to the plot is given in Table 13.1.

Despite (by design) not using adaptations of step size for the preservation of balance, large improvements to the push recovery ability of the robot are achieved by the direct fused angle feedback controller. This level of improvement has been quantified using controlled push experiments in Gazebo simulation. A simulated igus Humanoid Open Platform was made to walk on the spot with the feedback controller either turned on or off, and 20 pushes of equal impulse magnitude but random direction[3] were applied at regular time intervals. If a robot fell over due to a push, it was manually reset to upright in time for the next push. The number of successful pushes, i.e. pushes where the robot did not fall over, even with delayed consequence, were recorded for the open-loop and closed-loop gait for push impulses in the range of $0.3$–$2.0$ s N. The numerical results of the experiment are listed in Table 13.1, and the corresponding videos of the actual experiments are provided in Video 13.3. Figure 13.8 provides a graphical summary of the obtained results, and what it means for the comparative stability of the open-loop and closed-loop gaits. Overall, it can be observed that the feedback controller makes a clear difference to the stability of the robot. The effect is most noticeable for higher push impulses, where for example a push impulse of $1.6$ s N resulted in only 5 successful pushes for the open-loop gait, but 17 for the closed-loop gait. Note that for reference of scale, a push impulse of $2.0\,\mathrm{s\,N}$ is expected to cause an instantaneous change in CoM velocity of about $30-40\,\mathrm{cm/s}$ for the igus Humanoid Open Platform, which is considerable given its CoM height of about $55\,\mathrm{cm}$.

---

3 The pushes were restricted to be 'horizontal' in the sense that they are parallel to the ground plane, but random in terms of their exact heading.

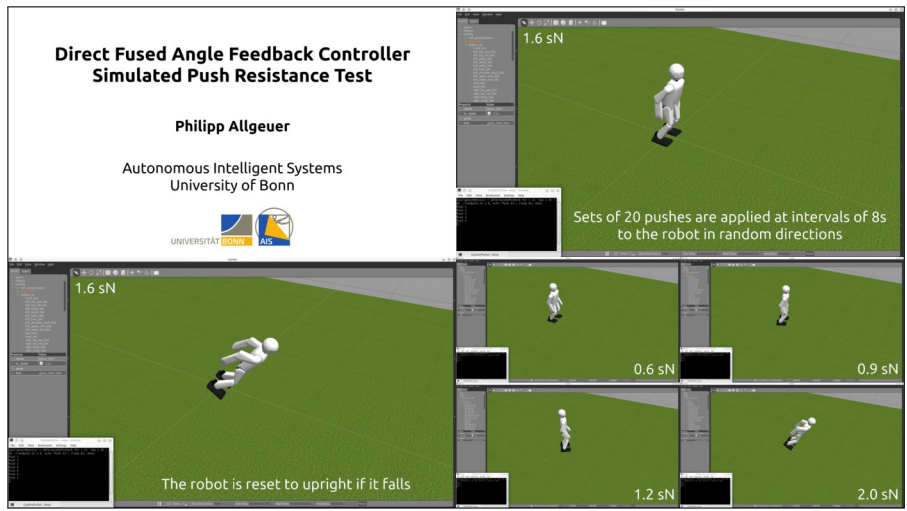

Video 13.3: Push resistance test performed on an igus Humanoid Open Platform in simulation. The robot is walking on the spot with the CPG gait, and is disturbed in random directions by impulses of various magnitudes. The effectiveness of the direct fused angle controller is evaluated by comparing it to open-loop performance.
https://youtu.be/c6zlCK4nFG0
*Direct Fused Angle Feedback Controller Simulated Push Resistance Test*

Push experiments were also performed on real hardware using the igus Humanoid Open Platform. As push impulses cannot easily be set or measured in real-life push experiments, a different approach was used, whereby the robot was pushed many times over many minutes of walking, and the corresponding push responses were extracted, synchronised, overlaid, and finally classified as either successful or unsuccessful. The results of this process can be seen in Figure 13.9. Separate plots are provided for forwards and backwards pushes, and for the open-loop and closed-loop gaits (i.e. with and without the direct fused angle feedback controller). Note that the lines in the plot terminate whenever the robot started receiving assistance due to falling (red lines), or when a subsequent push was started (blue lines). It can clearly be observed from the plots that the feedback controller has a strong effect in reducing the duration that a disturbance is experienced, and the amount that the fused pitch oscillates when it returns to nominal. The controller is able to efficiently leverage the effects of the corrective actions to more quickly dissipate the disturbance energy without causing any extra oscillations or disruptions. Note that the increase in maximum fused pitch that can successfully be recovered from is only modest when comparing open-loop and closed-loop performance. This is because the tipping point of the robot is somewhat mechanically fixed, and can only be slightly altered by moving the arms. The magnitude of the push impulse that it takes to *get* the robot to this maximum fused pitch is significantly higher however (for closed-loop), as it first has to overcome the corrective actions to even reach the tipping point.

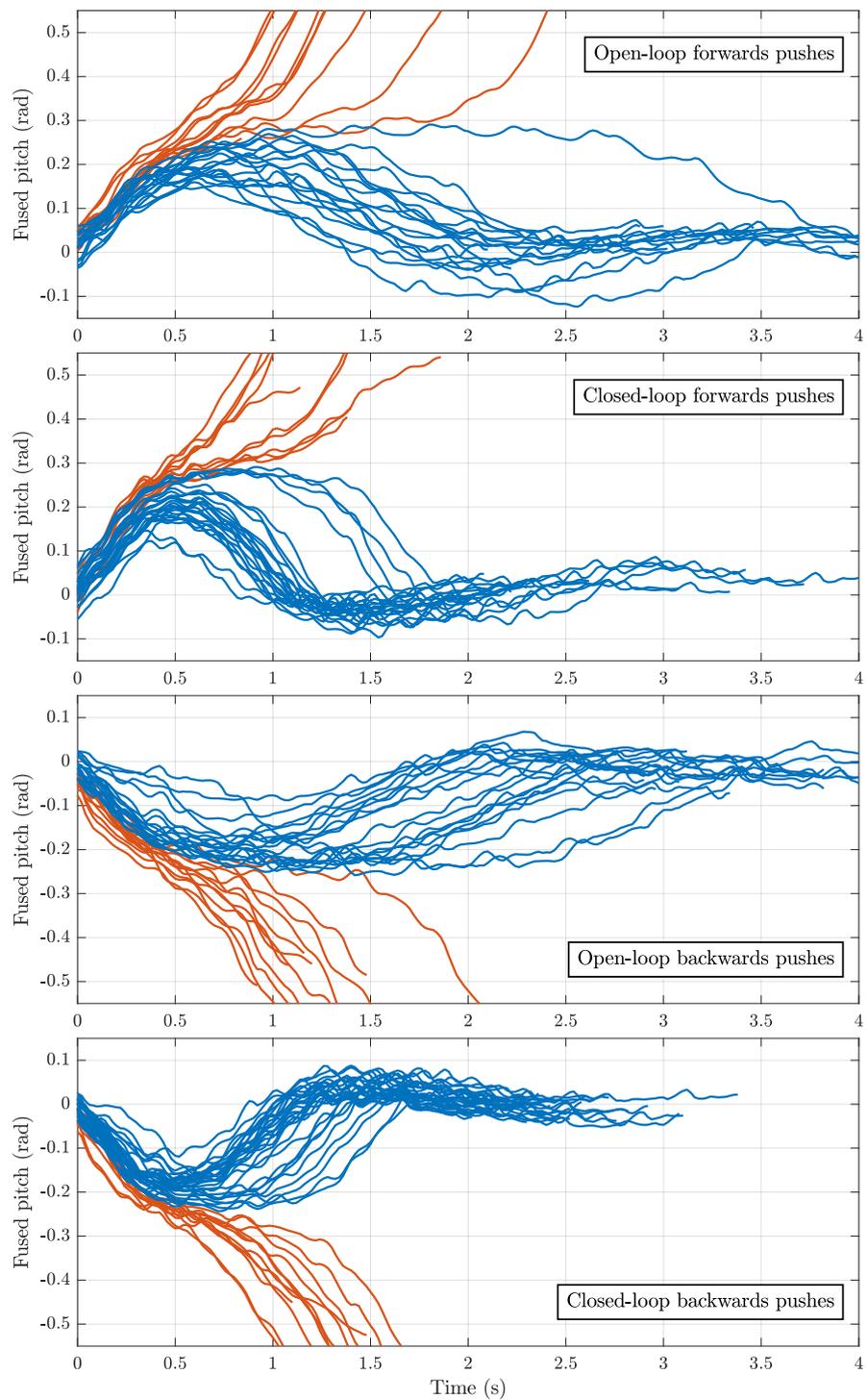

Figure 13.9: Plots of the transient response of a walking igus Humanoid Open Platform robot to sagittal pushes of various strengths. The pushes are grouped by whether they are forwards or backwards, and whether the feedback controller was active at the time or not. From top to bottom, the plots correspond to a) open-loop forwards pushes, b) closed-loop forwards pushes, c) open-loop backwards pushes, and d) closed-loop backwards pushes. The pushes all occurred at time $t = 0$, and each individual curve is coloured blue if the robot survived the push, otherwise red.

The responses of the robot to the pushes shown in Figure 13.9 can also be analysed in the phase space, i.e. in the space of fused pitch velocity vs. fused pitch. Combining the respective forwards and backwards pushes into a single plot yields the curves shown in Figure 13.10. The top plot shows the traces of the push responses for the open-loop gait, while the bottom plot shows the same for the closed-loop gait. Note that the start of each trace ($t = 0$ in Figure 13.9) is marked with a solid dot. It can immediately be observed from Figure 13.10 that the region in the phase space corresponding to stable walking trajectories is larger for the closed-loop gait than for the open-loop gait. This increase in size corresponds predominantly to a wider range of stable fused pitch velocities—an observation that is consistent with the previous observations that the closed-loop gait is more quickly able to dissipate disturbance energies, and can deal with larger initial CoM velocities after a push. These factors together lead to the greater overall push recovery ability that is observed for the closed-loop gait. In reference to Figure 13.10, it should be noted that some of the red traces start from deep within the blue region of stability, and first sharply increase their velocity (a vertical change in the plot) before subsequently starting to diverge in terms of the fused pitch. This is deduced to correspond to situations where the push from the experimenter was still ongoing when the trace was started. As a similar observation, some of the red unstable trajectories start to decrease their velocity towards the end of the trace. This is deduced to correspond to situations where the experimenter caught the falling robot before the trace was stopped.

The plots in Figure 13.10 can alternatively be viewed as a heat map, as shown in Figure 13.11. The states in the phase space are divided into cells (i.e. bins), and the trajectories that pass through any one cell are collected and used to calculate a success rate for the cell. The success rate is a value in $[0, 1]$, and corresponds to the proportion of the trajectories that passed through the cell that did not end in a fall. Extra data of normal balanced walking has been incorporated into the figure to avoid the issue of unvisited states in the middle of the stable region (i.e. holes in the data as can be seen in Figure 13.10). Note that a minimum cell visit count of two was used, explaining why disconnected cell patches are possible. It can be observed from the figure that the robot is 100% stable close to the phase origin, but as the fused pitch and/or velocity increase, there is a sharp transition to 0% stability. This transition is especially sharp in the case of the closed-loop gait, which exhibits a relatively low amount of intermediate transitional cells between the blue and red areas. Overall, like in Figure 13.10, it can be seen that the closed-loop gait has a significantly larger region of stability, and that the main gain in area is due to the greater stability of states with high fused pitch velocities.

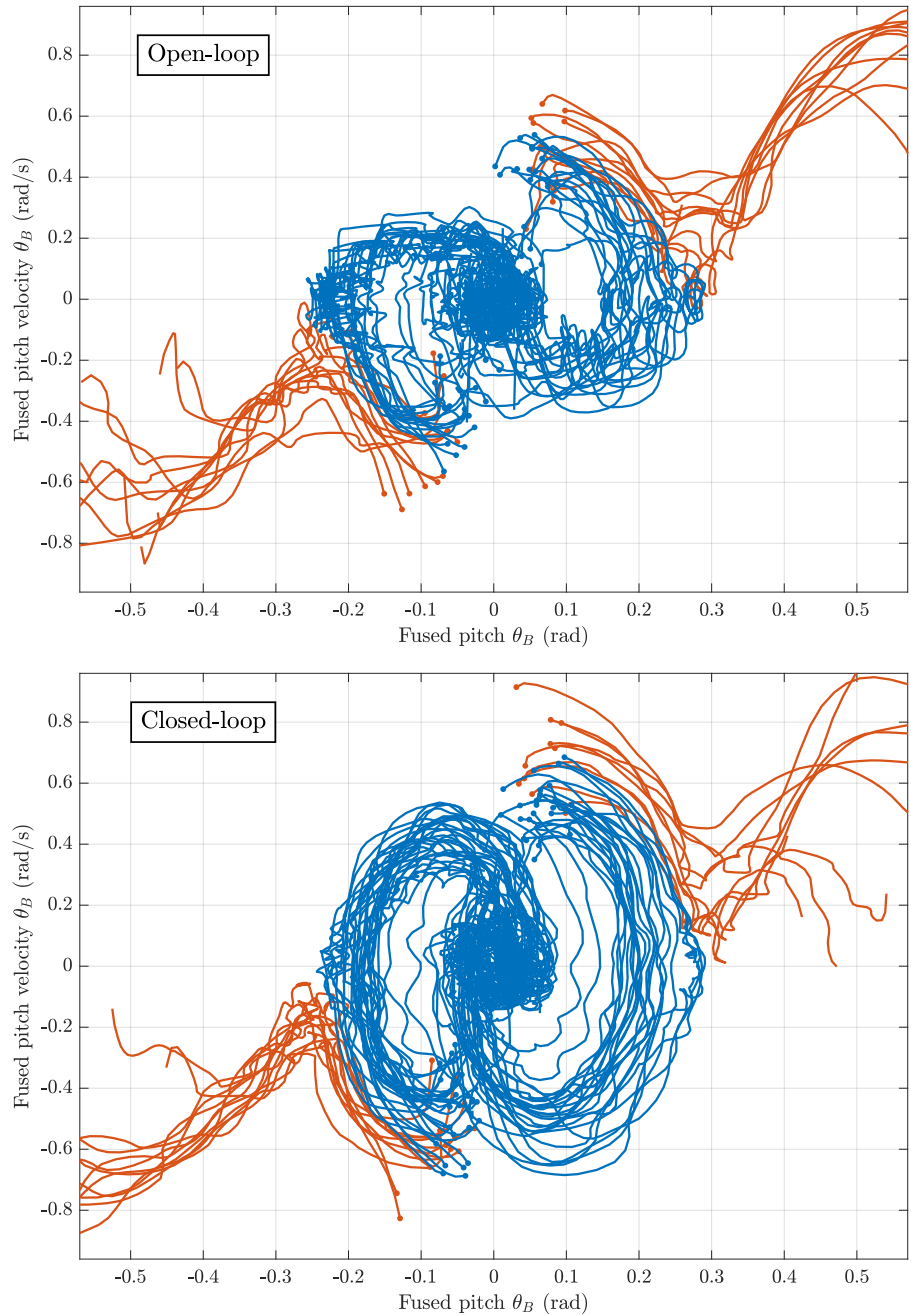

Figure 13.10: Plots of the phase response of a walking igus Humanoid Open Platform robot when sagittal pushes of various strengths are applied. The fused pitch velocity $\dot{\theta}_B$ is plotted against the fused pitch $\theta_B$ for each individual push trajectory, and the resulting curve is coloured according to whether the robot successfully withstood the push (blue) or not (red). The beginning of each trajectory is marked with a solid dot. The top plot corresponds to the performance of the robot with the feedback controller disabled, and the bottom plot shows how this performance improves when the feedback controller is enabled.

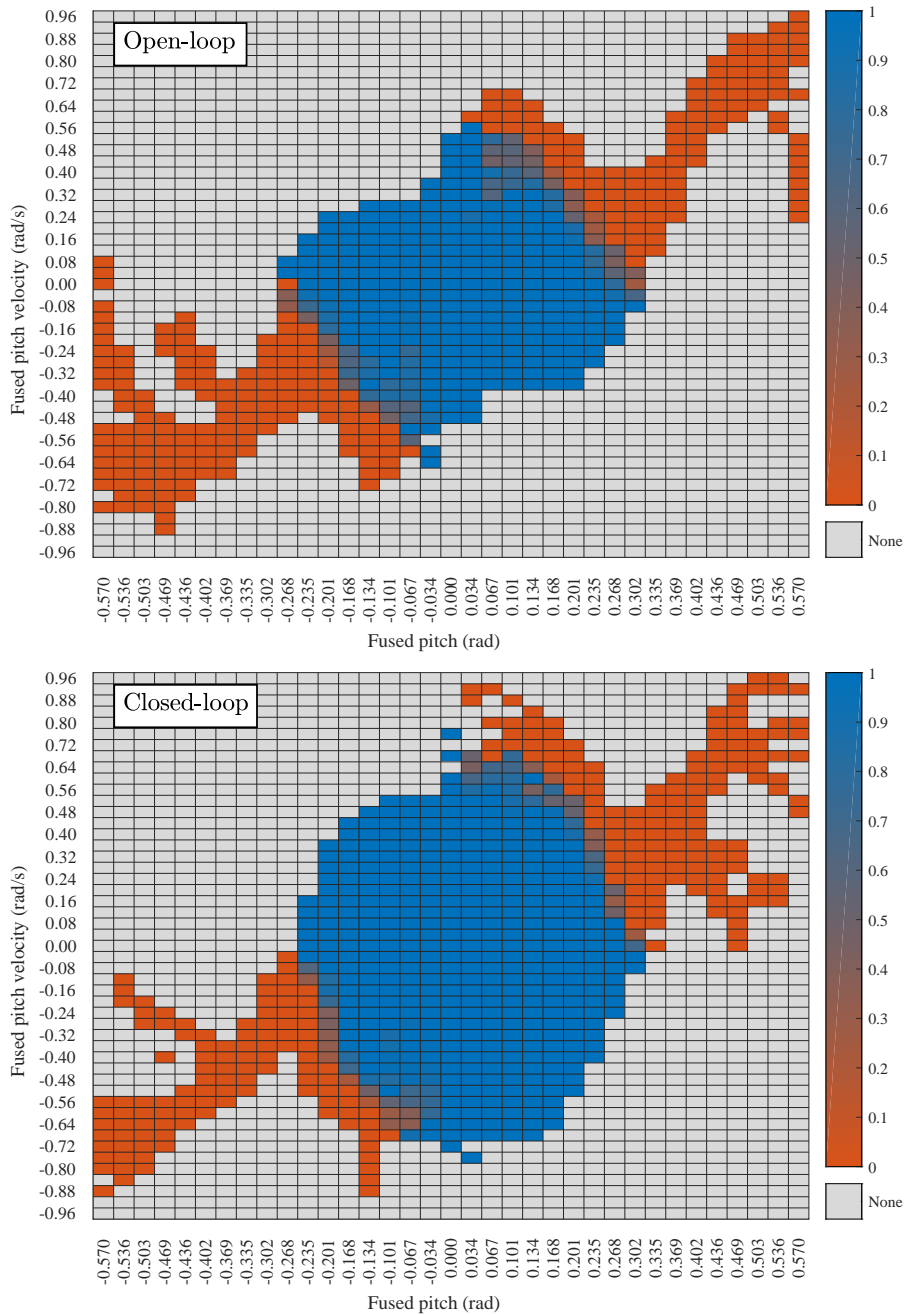

Figure 13.11: Plots of the open-loop vs. closed-loop walking stability of an igus Humanoid Open Platform in the form of a phase space heat map. The blue areas indicate phase states corresponding to stable walking (1.0 = 100% success rate), while the red areas indicate states where the robot tendentially lost balance and fell over (0.0 = 0% success rate). In-between colours indicate that certain binned states were encountered multiple times with different outcomes. All the data from the push tests is shown, including also the states of normal balanced walking between pushes. The grey cells correspond to phase states that were not encountered during the tests.

The task of tuning the PD gains of the direct fused angle feedback controller can usually easily be completed manually, but an alternative learning approach for the tuning of the sagittal gains has been proposed by Rodriguez et al. (2018). They used Bayesian optimisation with Gaussian process regression and multi-fidelity entropy search to share experiments and parameter updates between real-robot experiments and physical simulations, all the while trying to account and correct for systematic errors between the two. A video is available[4] that compares the manually tuned sagittal feedback gains to the final optimised ones for an igus Humanoid Open Platform. Although the difference may be difficult to detect at first, on closer inspection the manually tuned robot can be seen to exhibit slightly greater low frequency oscillations in the pitch direction, which at times can impede clean walking progress. More experiments would be needed to evaluate the true potential of this method in surpassing manual tuning in a time and resource-efficient manner.

The gait presented in this chapter has successfully been run in combination with the capture step controller described in Chapter 12, and produced results notably superior to just using the latter on the igus Humanoid Open Platform. The capture step timing was used in place of the fused angle deviation timing, and the step sizes computed from the capture step algorithm were either used or not used based on the situation (e.g. in testing vs. at competition). The gait was used in this configuration for example at the RoboCup 2016 soccer tournament, and over all games played, none of the five robots ever fell while walking (see Video 1.1 for impressions of the competition), except in cases of strong collisions with other robots. In fact, up to and including 2018 all of the TeenSize and AdultSize robots of the NimbRo RoboCup team always used the direct fused angle feedback controller for gait stabilisation purposes, frequently however with capture step timing enabled. Video 1.4, for example, shows the AdultSize RoboCup performance with capture step timing enabled, while Video 1.2 shows the performance of the same robots with the simpler fused angle deviation timing. The extensibility of the gait to incorporate other methods of step size and timing validates, amongst other things, the use of the presented gait as a stabilising foundation for more complex step size adaptation schemes.

---

## 13.5 CONCLUSION

An inherently robust omnidirectional closed-loop gait has been presented in this chapter that stabilises a central pattern generated open-loop gait using fused angle feedback mechanisms. The gait is simple, model-free, quick to tune, easily transferable between robots, and only requires servo position feedback if feed-forward torque compensation is desired as part of the actuator control scheme (see Section 3.2). The gait is also suitable for larger robots with low-cost sensors and position-controlled actuators. This demonstrates that walking does not always mandate complex stabilisation mechanisms. The gait has been experimentally verified and discussed, and demonstrably made robots walk that were not able to produce even remotely similar results with just a manually tuned open-loop CPG approach. One of the notable merits of the presented gait is that it can combine very well with more complicated model-based approaches that are able to suggest step size and/or alternative timing adjustments. This is what makes the gait so useful and powerful as a building block for more complex and more tailored gait stabilisation schemes.

# KEYPOINT GAIT GENERATOR

The task of bipedal locomotion exposes many facets of the concept of balance—most notably the many and varied methods by which balance can be preserved. While humans, even in early childhood, seem to effortlessly know how to stabilise their gait and best react to pushes while walking, the situation is quite different for humanoid robotic platforms. For robots it must first be delineated by which approaches they can be made to keep their balance, before algorithms can be developed that allow them to reliably execute such strategies.

As discussed in Chapter 8, broadly speaking there are two main paradigms for the implementation of bipedal robotic gaits. A common approach in the state of the art is to use a dynamics model of some kind to capture and predict the physical response of the robot, and calculate or optimise a trajectory to satisfy the required motion and balance criteria. This motion trajectory is then executed on the robot, often with a controller to reject deviations and enforce tracking, and/or under regular recomputation to adapt for differences in the real response of the robot. For imprecise low-cost robots however, where good quality tracking and execution of a trajectory is not given, using such optimised trajectories generated directly from simplified (or even whole-body dynamics) models is often fraught with difficulty. Significant nonlinearities, such as joint backlash, sensor noise, sensor and actuator delays, irregular properties of the contact surface, and unmeasurable external disturbances, are difficult to incorporate into models. This greatly limits the predictive power of such models, and subsequently the applicability of such methods to such robots.

Nevertheless, simpler, cheaper and smaller robots are often easily made to walk despite these nonlinearities using hand-crafted gaits. These gaits are however usually not very flexible or stable on their own. This is the foundation of the second paradigm for the implementation of bipedal gaits—letting the robot find its own natural rhythm and stability with an inspired open-loop gait generator that is entirely self-contained, and extending it with a higher level controller that seeks to preserve and return to that rhythm when there are significant deviations from it. As we did for the Central Pattern Generator (CPG), in this chapter we continue with this second paradigm and present a self-stable omnidirectional gait generator that seeks to strike a balance between the security and simplicity of hand-crafted gaits, and the advantages of analytically computed and optimised gaits. More than just making the robot walk, the so-called Keypoint Gait Generator (KGG) directly embeds a myriad of corrective actions into its generation algorithm, each of which can be commanded and activated by higher level controllers to systematically allow preservation of balance during walking. The diversity of the implemented corrective actions arguably cover the majority of humanlike strategies by which biped robots can balance during walking.

The idea for the keypoint gait generator originated from the CPG, in particular in the light of the CPG's observed shortcomings. The KGG is somewhat like an extension and complete redesign of the CPG from the ground up, with a focus on greater flexibility and general analytical correctness in the way the gait motion waveforms are generated, in particular from a task space perspective. While the CPG strictly only generates the nominal open-loop walking waveforms, and relies on higher level balance controllers like the direct fused angle controller to later change the waveforms to implement corrective actions, the KGG intrinsically incorporates the effects of a diverse array of corrective actions and directly generates the final required joint waveforms. Amongst many improvements of the KGG over the CPG (see Section 14.1.3), the KGG incorporates many new and/or improved corrective actions in its generation algorithm (as compared to the CPG), and allows situations like for example lateral crossing trajectories to be explicitly addressed and dealt with.

An implementation of the Keypoint Gait Generator (KGG) has been released open source as part of the `feed_gait` package of the Humanoid Open Platform ROS Software.[1] In all aspects this can be used as a reference for the algorithms described in this chapter.

## 14.1  MOTIVATION

Before descending into its inner workings, we first try to put the KGG into context by identifying the objectives that it intends to fulfil, the balance strategies it intends to implement, and where it fits into the bigger picture of the gait architecture.

### 14.1.1  Strategies for Balanced Walking

There are many ways in which a humanoid robot is assessed to be able to influence its balance while walking, not just the two that are most commonly addressed in other works—adjustments of step size and timing. Inspired in part by the reactions humans exhibit when disturbed while standing or walking, a list of strategies for preserving or recovering balance in a bipedal robot is as follows:

- Adaptation of step sizes,

- Adaptation of step timing,

---

1 https://github.com/AIS-Bonn/humanoid_op_ros/tree/master/src/nimbro/motion/gait_engines/feed_gait

- Horizontal planar shifts of the torso, and thereby Centre of Mass (CoM) position, relative to the feet,

- Active changes in torso height, and thereby CoM height, above the ground,

- Active leaning of the robot torso from the hips,

- Arm motions to generate reaction moments and/or statically offset the centre of gravity of the robot,

- Swing leg trajectory adjustments, to generate reaction moments and/or bias the weight of the robot in a particular direction by means of modifying how the swing leg moves between lift-off and foot strike,

- Height adjustment of the predicted locations of lift-off and foot strike to adapt to the ground, for example as required in the case of a tilted robot,

- Intentional utilisation of ground normal forces to apply a restoring moment to the robot to make it more upright, for example by pushing more into the ground at foot strike, or less just before lift-off,

- Adjustments of the foot tilt relative to the ground at the instant of foot strike, resulting in large impulses to the balance state of the robot due to the impactful nature of the ground contacts, and,

- Adjustments of the foot tilt relative to the ground during the support and swing phases.

The Keypoint Gait Generator (KGG) presented in this chapter seeks to address all of these strategies by extending its fundamental walking waveforms with a multitude of corrective actions that can be activated simultaneously (as required) to preserve balance.

### 14.1.2 Gait Architecture

The gait presented in this chapter is referred to as a gait generator, as it gives the ability to higher level controllers to execute all of the balance strategies implemented therein, without these controllers needing to know anything about joint motions, how they look like, or how they need to be generated.[2] As such, the gait architecture can logically be partitioned into the higher level controller(s) in use, and the KGG that sits underneath them, with the latter receiving its commands from the controllers via the gait interface described later in Section 14.2.3.1. The complete gait architecture, including the sensorimotor management aspects, is shown in Figure 14.1. Despite the appearance in the figure, it should be noted that no higher level controllers are actually mandatory for walking at all—the gait generator can make the robot walk open-loop by itself as it was designed to do—but walking open-loop does not make use of any of the implemented balancing corrective actions, and is subsequently not as stable as feedback-based walking.[3]

---

2 Recall that this was not the case for the CPG and its corresponding higher level controllers, as for example the direct fused angle controller directly modifies the generated joint waveforms in order to implement its required corrective actions.

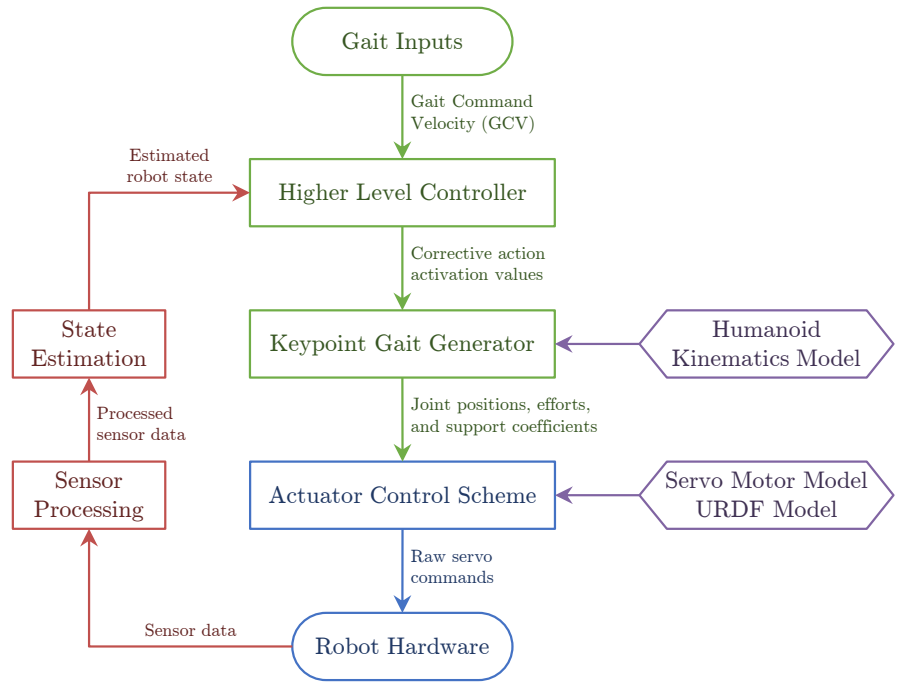

Figure 14.1: Overview of where the KGG fits into the overall keypoint gait architecture. Note that it is possible for there to be multiple higher level controllers running in parallel if each of them calculate activation values for different sets of corrective actions. The actuator control scheme (Chapter 3), sensor processing (Chapter 4), humanoid kinematics model (Chapter 9), and state estimation (Chapter 10) have been covered in previous chapters. A higher level controller that drives all of the corrective actions implemented in the KGG is presented in Chapter 15.

The input to the gait as a whole is the dimensionless Gait Command Velocity (GCV) vector

$$\mathbf{v}_g = (v_{gx}, v_{gy}, v_{gz}) \in [-1, 1]^3. \tag{14.1}$$

This is a vector of three dimensionless values in the range $[-1, 1]$, representing the required x, y and yaw velocities of the robot as a ratio of their respective allowed maximums. The higher level controllers process the input GCV vector $\mathbf{v}_g$, along with the sensory feedback from the robot, and compute commands for the KGG that try to ensure that the robot remains balanced. These commands consist of the required activation values for the KGG corrective actions—including for example the required instantaneous gait frequency $f_g$—and incorporate in particular also the internal GCV vector

$$\mathbf{v}_i = (v_{ix}, v_{iy}, v_{iz}) \in [-1, 1]^3. \tag{14.2}$$

---

3 Note that in the case of open-loop walking without a higher level controller, the input GCV $\mathbf{v}_g$ is passed directly to the keypoint gait generator as the internal GCV $\mathbf{v}_i$, and the gait frequency $f_g$ that is normally set by the higher level controller (see Section 14.2.3.1) is simply kept constant at its nominal value.

The internal GCV vector $\mathbf{v}_i$ tracks the input GCV $\mathbf{v}_g$ in all situations other than if the higher level controller wishes to command step size adjustments for the purpose of balance preservation, and is assumed to be continuous, e.g. through filtering or slope limiting applied by the higher level controller.

Given the gait generator commands calculated by the higher level controllers, the KGG in turn calculates the required instantaneous joint position, joint effort and support coefficient commands, where support coefficients are a feed-forward estimation of the ratios of the weight of the robot that will be carried by each leg when the joint commands are executed. As discussed in Section 3.2, all of these outputs are used by the actuator control scheme to generate raw servo commands that are then sent electrically to the servo motors. The purpose of the actuator control scheme is to ensure good position tracking of the servo motors, by accounting in a feed-forward manner for factors such as battery voltage, joint friction, link inertia, and the relative loadings of the legs.

As discussed on page 380, the KGG and associated gait architecture is somewhat related to the Central Pattern Generator (CPG) presented in Chapter 11. Aside from a fundamental shift in how the walking waveforms are generated, one core difference between the two gait generators is that the corrective actions of the KGG are truly embedded and integrated inside the gait generator, as opposed to being imprecisely superimposed onto individual joints post factum and assuming that this does not adversely affect the gait profile in any way. A further difference and advantage of the KGG over the CPG is that the associated higher level balance controllers are abstracted away from having to work with the raw joint waveforms, leading to a more governed output and better separation of duties.

### 14.1.3 Aims for the Gait Generator

One of the aims of the gait generator is to allow fast walking with large step sizes. In order for this to be possible, the severity of the self-disturbances, impacts and nonessential spikes in acceleration during the gait cycle needs to be minimised as much as possible. This is embodied in the following list of desired properties for the gait generator:

- The final trajectories should be continuous in velocity and acceleration in all dimensions,

- The trunk orientation should remain as constant as possible, up to the moderate lateral oscillations required to assist support exchanges,

- The CoM should not undergo any sudden changes in height, and should generally rise and fall as little as possible during normal walking at a given gait velocity,

- The CoM should avoid start-stopping, i.e. having excessive oscillatory sagittal linear accelerations, as much as possible,

- During ideal walking, the feet should not significantly impact the ground and come down with zero instantaneous normal relative ground velocity, and,

- The feet should always contact the ground with zero instantaneous tangential velocity relative to the ground.

The KGG was designed at every stage with these aims in mind.

In addition to these general desired properties, there are also the following more specific considerations—that in particular were observed to be problems of, or were not addressed by, the CPG presented in Chapter 11. These need to be taken into account in the design of the gait generator:

- The robot torso is not necessarily nominally sagittally upright, so the floor may *nominally* be tilted relative to the torso frame, and the leg lift-swing profile needs to be able to adapt to that.

- The neutral leg pose during walking is not usually vertically downwards, so leg swing rotations in the hip for the purpose of stepping adversely affect the foot height, and in particular do so in a strongly asymmetrical fashion.

- The greater the swing of a leg, the more a shortening of the leg for the purpose of leg lifting also acts to reduce the step size.

- When falling in the direction of walking, if the ankle does not adjust for the extra tilt deviation, the toe may strike the ground prior to the remainder of the foot, preventing a full step from being taken, and possibly destabilising the robot.

- When falling in the direction of walking, the swing leg needs to be lifted higher towards the end of swing to avoid premature contact of the foot with the ground, and vice versa when falling in the direction opposite to walking. Crucially, the continued leg motion should also passively exert a restoring force to counteract the tilt instead of 'giving way' to it.

- For cases of compliant actuation, when a leg is lifted it tends to spring mechanically into a particular direction due to the sudden reduction in joint torque loadings. The leg is only reloaded when it has already reestablished contact with the ground, so leg lifting and placing is inherently asymmetrical and will tend to make the robot drift.

The KGG was designed to address these shortcomings of the CPG.

## 14.2 KEYPOINT GAIT GENERATION

We now look at the internals of the KGG, and how it uses systems of linear equations and cubic spline interpolation between a set of gait phase-dependent keypoints to dynamically generate walking trajectories for the robot in a constraint-based fashion. Before we can do this however, we first need to review some preliminary concepts required by the KGG algorithm, and explicitly define the ten corrective actions that are implemented as part of it. Based on the definitions of these corrective actions, we also explicitly delineate the inputs and outputs of the KGG, before describing the leg and arm trajectory generation methods in detail.

### 14.2.1 Preliminaries

The KGG ties together a wide variety of concepts that were introduced in other parts of this thesis. These concepts are briefly listed and reviewed here for the purpose of understanding:

**Coordinate axis convention:** By convention, the global and body-fixed reference frames used in this chapter are defined such that the x-axis points 'forwards', the y-axis points 'leftwards', and the z-axis points 'upwards'.

**Humanoid kinematics model:** Chapter 9 introduced the humanoid kinematics model, the associated joint, abstract, inverse and leg tip pose spaces, and algorithms with which these spaces can be converted between each other. Algorithms for the corresponding velocity space conversions were also indicated. The model defines various points, lengths and coordinate frames throughout the assumed kinematics of the robot (see Figures 9.1 and 9.3 and Table 9.2), including most notably the body-fixed frame {B} at the so-called *hip centre point*, and the leg tip frame {F} at the so-called *leg tip point* of each leg. As part of the humanoid kinematics model, Section 9.2.4 details an inverse kinematics method for the arms that allows the modelled CoM of the arms to be placed on arbitrary rays through the shoulder points.

**Hip centre point:**   The hip centre point $\mathbf{p}_h^c$ is defined as the midpoint of the left and right hip points $\mathbf{p}_h^{l,r}$, as indicated in Figure 9.1.

**Shoulder point:**   The shoulder points are as visually defined in Figure 9.1, and their respective positions are denoted $\mathbf{p}_s^{l,r}$.

**Inverse leg scale:**   The inverse leg scale parameter $L_i$ (Section 9.1.3) is given by the vertical height between the hip points and ankle points when the robot is in its outstretched zero position (i.e. when all joint angles are zero), and is a measure of the size of the legs with respect to the inverse pose space.

**Leg tip scale:**   The leg tip scale parameter $L_t$ (Section 9.1.4) is given by the vertical height between the hip points and leg tip points when the robot is in its outstretched zero position (i.e. when all joint angles are zero), and is a measure of the size of the legs with respect to the leg tip pose space. Note that the inverse leg scale and leg tip scale parameters are there to scale many parameters of the gait, to make them dimensionless and port easily between robots.

**Soft coercion:**   Soft coercion is a general purpose continuously differentiable transfer function that implements the saturation of a quantity to any particular given range (see Appendix A.1.2.3). It is mainly used to avoid the discontinuities in velocity and acceleration that are caused if standard hard coercion (see Appendix A.1.2.1) is used to limit a motion trajectory in some way.

**Elliptical soft coercion:**   Elliptical soft coercion is a 2D variant of soft coercion that radially limits a 2D vector to the desired axis-aligned elliptical region in space (see Appendix A.1.2.4).

**Tilt angles:**   Tilt angles is a rotation representation $T = (\psi, \gamma, \alpha) \in \mathbb{T}$ that consists of the fused yaw $\psi$, tilt axis angle $\gamma$, and tilt angle $\alpha$, and is rigorously introduced in Section 5.4.3. The $(\gamma, \alpha)$ parameters together define the *tilt rotation component* of the rotation, where $\gamma$ gives the direction of this tilt, and $\alpha$ gives its magnitude. The related *absolute tilt angles* representation is given by $\tilde{T} = (\psi, \tilde{\gamma}, \alpha) \in \tilde{\mathbb{T}}$, where

$$\tilde{\gamma} = \gamma + \psi. \tag{14.3}$$

The $(\tilde{\gamma}, \alpha)$ parameters together define the *absolute tilt rotation component* of the rotation.

**Tilt phase space:**   The tilt phase space is a generic representation of 3D rotations that parameterises them using the phase roll $p_x$, phase pitch $p_y$, and fused yaw $p_z \equiv \psi$ parameters, where

$$P = (p_x, p_y, p_z) = (\alpha \cos \gamma, \ \alpha \sin \gamma, \ \psi) \in \mathbb{P}^3. \tag{14.4}$$

The tilt phase space is detailed in Section 5.4.5.1, and as discussed in Section 5.8.1, is a way of decomposing a rotation into three concurrently acting components in such a way that it gives insight into how rotated a body is in each of the three major planes. The tilt phase space allows for tilt rotations $(p_x, p_y) \in \mathbb{P}^2$ of arbitrary magnitude, and implicitly defines a mathematically correct and commutative form of tilt rotation addition known as *tilt vector addition*. As described in Section 5.4.5.3, this allows expressions such as for example

$$(\gamma_1, \alpha_1) \oplus (\gamma_2, \alpha_2) = (\gamma_3, \alpha_3) \tag{14.5}$$

to be evaluated.

**Absolute tilt phase space:** The absolute tilt phase space (see Section 5.4.5.2) is analogous to the tilt phase space, with the only difference being that it is defined in terms of the absolute tilt axis angle $\tilde{\gamma}$ instead of the (relative) tilt axis angle $\gamma$, i.e.

$$\tilde{P} = (\tilde{p}_x, \tilde{p}_y, \tilde{p}_z) = (\alpha \cos \tilde{\gamma}, \alpha \sin \tilde{\gamma}, \psi) \in \tilde{\mathbb{P}}^3. \tag{14.6}$$

Similar properties and results hold as for the (relative) tilt phase space, like for example the definition of tilt vector addition.

**Referenced rotation:** For any coordinate frames {G}, {A} and {B}, the *referenced rotation*

$$^{GA}_{\ B}R \equiv {}^G_B R\, {}^A_G R \tag{14.7}$$

is the rotation relative to {G} that maps frame {A} onto frame {B}. This is referred to as the rotation from {A} to {B} *referenced by* {G}. As discussed in Section 7.1.1.5, referenced rotations are a direct generalisation of the standard rotation basis notation '$^A_B R$', as

$$^A_B R \equiv {}^{AA}_{\ B}R. \tag{14.8}$$

**Gait phase:** Just like the CPG gait, the entire generation of the KGG gait is a function of the so-called gait phase $\mu_i \in (-\pi, \pi]$. This parameter is continuously incremented based on the required instantaneous gait frequency $f_g$, and wraps around to stay in range. This is embodied by the gait phase update equation

$$\mu_i \leftarrow \mathrm{wrap}(\mu_i + \pi f_g \Delta t), \tag{14.9}$$

which is applied in every execution cycle of the KGG, and indicates how timing feedback (via $f_g$, see Section 14.2.3.1) is incorporated into the gait generator. The motion of each leg is divided into well-defined swing and support phases as a function of the gait phase, with the overlapping support phases of each leg resulting in two brief double support phases (see Table 14.1 for more information). Each double support phase is nominally $D$ radians in length in terms of the gait phase, where $D$ is the double support phase length. By convention, $\mu_i = 0$ corresponds to the instant of foot strike of the right foot (and the beginning of the corresponding double support phase), and $\mu_i = \pi$ corresponds to the instant of foot strike of the left foot. Given the remaining inputs to the KGG as listed in Section 14.2.3.1, the gait phase $\mu_i$ is essentially a sampling parameter that is used to select the right 'snapshot' of the gait cycle at every time instant.

**Limb phase:** When generating the KGG waveforms, it is useful to have an expression of gait phase for each limb independently that always has the swing phase of the limb in the range $(0, \pi)$. As such, the limb phase $\nu_i \in (-\pi, \pi]$ is defined for each arm and leg to be

$$\nu_i = \begin{cases} \mu_i & \text{if swing phase is in } \mu_i \in (0, \pi], \\ \text{wrap}(\mu_i + \pi) & \text{if swing phase is in } \mu_i \in (-\pi, 0]. \end{cases} \quad (14.10)$$

Note that as the arms generally swing opposite to the legs during walking, the swing and support phase definitions for the arms are exactly opposite to those for the legs.

In addition to the concepts reviewed above, we also introduce the notion of the nominal ground plane N, which is a plane in body-fixed coordinates that reflects the nominal orientation of the ground relative to the robot. The nominal ground plane embodies the fact that not all robots nominally walk with a completely upright torso, e.g. as generally assumed in Missura (2015). The N plane is defined by the (fixed) nominal phase pitch $p_{yN}$ of the torso, and the corresponding nominal ground frame {N} is defined by the rotation

$$^B_N R = R_y(-p_{yN}), \quad (14.11)$$

where $R_y(\theta)$ is notation for a pure y-rotation by $\theta$ radians. From Equation (14.11), it can be seen that the nominal ground normal vector $^B\hat{\mathbf{z}}_N \equiv \hat{\mathbf{z}}_N$ is given by

$$\hat{\mathbf{z}}_N = (-\sin(p_{yN}), 0, \cos(p_{yN})). \quad (14.12)$$

The nominal ground plane N is used as a planar reference level for the generation of the arm and leg motion profiles. Given a nominal ground plane and a leg tip point that characterises the centre of a motion profile (i.e. motion centre point), the hip height of that profile is given by the distance perpendicular to the ground plane (i.e. along $\hat{\mathbf{z}}_N$) from the motion centre point to the hip centre point.

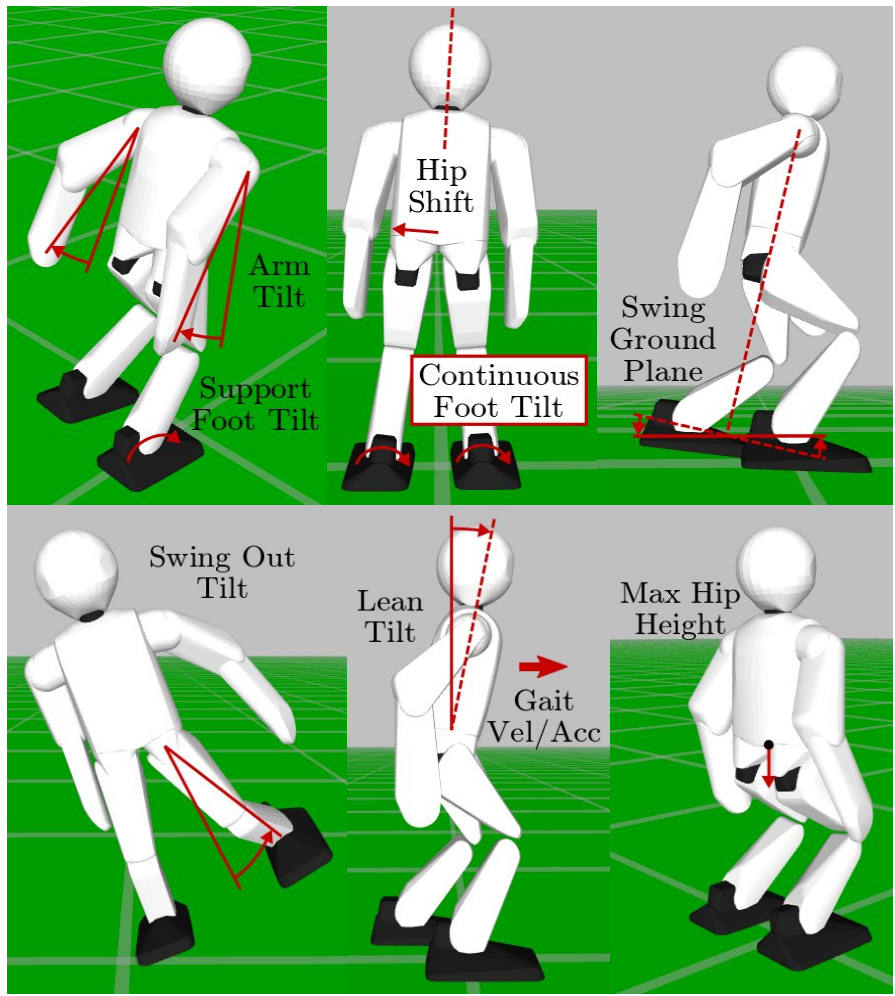

Figure 14.2: Diagrams of the various corrective actions implemented in the KGG. In most cases, the images and annotations are significant 1D simplifications of the true corrective actions for illustrative and explanatory purposes only. For instance, while the arm tilt shown in the top left image seems to be purely sagittal, in truth it is a 3D tilt rotation that is at no point separated into sagittal and/or lateral planes of motion. The arrows in general indicate the effect of the corrective actions in the depicted balance situations. The annotations in the swing ground plane image are trying to show that due to the forwards tilt of the robot, a normal step (lower dashed line) would have collided with the ground, while the adjusted step (solid line) avoids premature contact with the ground and executes the required step size despite the forwards tilt. Note that the swing ground plane is also 3D, and is not restricted in any way to be a pure sagittal tilt. The step size and timing corrective actions are not pictured in this figure as their effects are relatively clear.

### 14.2.2   Corrective Actions

Numerous corrective actions have been implemented in the KGG. All of these corrective actions are illustrated in Figure 14.2—with the exception of the step size and timing actions—and are given as follows:

**Step size:**   The sizes of the commanded steps are adjusted to capture the energy of the robot if it is falling (or predicted to fall) in a particular direction.

**Step timing:**   The rate at which the required stepping trajectories of the robot are executed is adjusted to speed up or slow down the gait as appropriate.

**Arm tilt:**   The modelled CoM positions of the arms are *tilted*[4] to shift their weight and cause corresponding reaction moments.

**Support foot tilt:**   The orientation of the current support foot is tilted to apply a restoring moment to the robot via the thereby altered ground reaction force. Smooth transitions to and from the support foot tilt are used during the double support phases to ensure that the resulting foot orientation trajectories remain continuous and differentiable.

**Continuous foot tilt:**   An equal tilt offset is applied to both feet throughout the entire gait trajectory in order to consistently shift the balance of the robot in a particular 360° direction.

**Hip shift:**   The position of the torso of the robot is adjusted in the **xy** plane to trim the centring of the robot's weight above its feet.

**Maximum hip height:**   The hip height (see page 388) of the generated motion profile is limited to a certain maximum height to temporarily increase the passive stability of the gait.

**Swing ground plane:**   The commanded foot trajectories are adapted to orientation deviations of the torso to avoid premature and/or belated foot strike. Effectively, the swing ground plane S is used as a planar reference level (similar to the nominal ground plane N) for adjusting the relative foot heights and tilts generated by the KGG.

**Swing out tilt:**   The midpoint of the trajectory of the swing leg is tilted around the respective hip point to adjust the path taken by the swing leg to its target footstep location. The modified swing trajectory influences the balance of the robot via the inertial and gravitational effects of the swing leg.

---

4 Note that all uses of the word 'tilted' precisely mean that a pure *tilt rotation* (see Section 5.4.2) is applied to the corresponding entity.

**Lean tilt:** The torso of the robot is tilted at the hips to intentionally make the robot lean in a particular direction.

It should be noted that the corrective actions were not simply chosen arbitrarily on a basis of trial and error, but were the result of an analysis of the conceivable strategies for balanced bipedal walking, the results of which were presented in Section 14.1.1.

### 14.2.3 Gait Generator Interface

As indicated in Figure 14.1, the KGG takes as its inputs the required activation values of the various implemented corrective actions, and outputs the required joint commands for the actuator control scheme. The exact nature of these gait generator inputs and outputs are delineated in this section.

#### 14.2.3.1 *Gait Generator Inputs*

A complete list of the inputs to the gait generator, covering all corrective action activation values, is as follows:

- The dimensionless internal GCV $\mathbf{v}_i = (v_{ix}, v_{iy}, v_{iz})$ to use to set the desired footstep size of the robot,

- The instantaneous gait frequency $f_g$ (in rad/s) to use for updating the gait phase $\mu_i$ in each cycle using Equation (14.9),

- The arm tilt $P_a = (p_{xa}, p_{ya})$ to apply to the arms,

- The support foot tilt $P_s = (p_{xs}, p_{ys})$ to apply to the feet during their respective support phases,

- The continuous foot tilt $P_c = (p_{xc}, p_{yc})$ to apply to the feet as offsets throughout the entire gait trajectory,

- The dimensionless hip shift $\mathbf{s} = (s_x, s_y)$ to apply to the robot (in units of the inverse leg scale $L_i$),

- The maximum hip height $H_{max}$ to allow relative to the feet for the generated motion profile (in units of the leg tip scale $L_t$),

- The 2D tilt phase rotation $P_S = (p_{xS}, p_{yS})$ defining the swing ground plane S relative to the nominal ground plane N,

- The swing out tilt $P_o = (p_{xo}, p_{yo})$ to apply to the midpoint of the leg swing trajectory, and,

- The lean tilt $P_l = (p_{xl}, p_{yl})$ to apply to the robot torso.

Given this list of gait generator inputs, it is important to note that:

(i) All strategies for balanced walking listed in Section 14.1.1 are covered by this palette of gait generator inputs.

(ii) Step size adjustments are effectuated via the internal GCV vector $\mathbf{v}_i$, and timing adjustments are effectuated via the gait frequency parameter $f_g$.

(iii) A small constant bias can be applied to $\mathbf{v}_i$ to negate any minor drifts in the real world walking performance of a particular robot.

(iv) All Cartesian actions are numerically expressed in dimensionless form relative to the nominal ground frame {N}, in units of either the inverse leg scale $L_i$ or leg tip scale $L_t$.

(v) All rotation-based corrective actions are expressed as pure tilt rotations relative to frame {N}, in the 2D tilt phase space rotation representation $(p_x, p_y) \in \mathbb{P}^2$ (see Section 5.4.5.1, and Section 14.2.3.3 for the one slight exception).

The corrective action activation values are all expressed in a dimensionless manner so that near-identical values can be used for robots of different scales. The same approach, for similar reasons, is followed for all configurable constants used throughout the KGG algorithm.

### 14.2.3.2 *Gait Generator Outputs*

As required for the actuator control scheme (Section 3.2), the outputs of the gait generator are as follows:

- The commanded joint positions $\mathbf{q}_o \in \mathbb{R}^N$, specifying the desired angular positions of the $N$ joints,

- The commanded joint efforts $\boldsymbol{\xi} \in [0, 1]^N$, specifying how stiff each joint should be, and,

- The commanded support coefficients $\kappa_l$, $\kappa_r$ for the left and right legs respectively, specifying the proportion of the weight of the robot that is expected to be supported by each.

The vector $\boldsymbol{\xi}$ of commanded joint efforts returned by the KGG is kept constant at the desired manually configured values, and so does not need to be further addressed by the trajectory generation methods presented in the following sections.

As previously mentioned, all of the inputs, outputs and parameters throughout the entire KGG have been chosen and expressed in such a way that they are dimensionless. This is important, as it means that their values are largely independent of size, scale and sample rate, and can be ported directly between different robots (even if, naturally, a renewed fine-tuning would still sometimes be of benefit). It is also worth noting that the entire keypoint gait generator is formulated with relatively few parameters and configuration variables to tune,[5] which offers the benefit of simplicity at least from the user's perspective.

### 14.2.3.3 *Swing Ground Plane Corrective Action*

As explained on page 390, the purpose of the swing ground plane input is to rotationally and positionally adjust the generated foot motion profiles relative to the ground in order to account for any tilt rotations of the robot torso. This can be done in many ways however, and the exact choice of approach is quite important. In the CPG gait (see Chapter 11), the *virtual slope* corrective action performs a similar function to the swing ground plane (at least in the sagittal direction), but is not quite as effective because it induces a moderate form of positive feedback into the balance feedback loop—the more the robot tilts forwards, the more the virtual slope makes the robot lift its legs in front of its body, and therefore the more the robot will as a result end up leaning even further forwards. Despite working as intended to prevent premature foot strike, the problem with the virtual slope corrective action is that it at no point inherently exerts a corrective moment on the robot to counteract the current tilt deviation of the torso. Due to its implementation via foot z-height adjustments, a further problem of the virtual slope corrective action is that even if foot strike occurs exactly when intended, the virtual slope changes the effective commanded step sizes relative to the tilted ground.

The swing ground plane corrective action combats the first of these two problems by ensuring that the foot-ground contacts implicitly generate a restoring force that drives the torso of the robot back towards its nominal orientation. This is done just after foot strike by either intentionally extending the leg into the ground, or by retracting it more than usual. In terms of the second of the two problems, as detailed in Section 14.2.4.4, the swing ground plane leaves the exact desired footstep sizes untouched, and just rotates them into the S plane for further processing. It is assumed that the higher level controller(s) saturate the deviation of the swing ground plane from the nominal ground plane, for example through elliptical soft coercion (see Appendix A.1.2.4), to ensure that the KGG is not commanded to perform unreasonable ground plane adjustments.

Analogously to the N plane, the S plane is defined mathematically with respect to the robot torso by the phase roll and phase pitch variables $(p_{xS}, p_{yS}) \in \mathbb{P}^2$ (see Section 14.2.3.1), albeit not quite as directly. The swing ground normal vector $^B\hat{\mathbf{z}}_S \equiv \hat{\mathbf{z}}_S$ is defined to be the local z-vector (see Section 5.4.2.1) corresponding to the inverse tilt rotation of $(p_{xS}, p_{yS})$. That is,

$$\hat{\mathbf{z}}_S = P_{\hat{z}}^I(-p_{xS}, -p_{yS}, 0), \tag{14.13}$$

---

5 At least, parameters that actually need to be tuned, and whose value are not just a clear analytical design decision from the outset.

Table 14.1: Keypoints used for leg trajectory generation (see also Figure 14.3)

| Key | Limb phase ($\nu_i$) | Characterisation |
|---|---|---|
| **A** | $\pm\pi$ | Swing phase $\rightarrow$ Double support phase |
| **B** | $-\pi + D$ | Double $\rightarrow$ Single support phase |
| **N** | $\frac{1}{2}(-\pi + D)$ | Midpoint of single support phase |
| **C** | $0$ | Single $\rightarrow$ Double support phase |
| **D** | $D$ | Double support phase $\rightarrow$ Swing phase |
| **E** | $D + \nu_{DE}$ | Swing-start tuning keypoint |
| **F** | $\frac{1}{2}(D + \pi)$ | Midpoint of swing phase |
| **G** | $\pi - \nu_{GA}$ | Swing-stop tuning keypoint |

where $P_z^l(\cdot, \cdot, \cdot)$ is the local z-vector corresponding to the enclosed relative tilt phase space parameters. The swing ground frame {S} is then defined to be the frame that results when the pure tilt rotation relative to {N} that rotates $\hat{\mathbf{z}}_N$ onto $\hat{\mathbf{z}}_S$ is applied to {N}. Mathematically, the rotation $_S^N R$ is calculated by evaluating

$$^N\hat{\mathbf{z}}_S = {}_N^B R^T \hat{\mathbf{z}}_S, \tag{14.14}$$

and subsequently applying the methods from Section 7.3.7.2 with $\psi = 0$ to obtain $_S^N R$. The swing ground plane S is given by the **xy** plane of the {S} frame, and is ideally set to correspond to the planar level of the ground relative to the robot at every instant of time.

### 14.2.4  Leg Trajectory Generation

The leg trajectories of the KGG are constructed based on a set of eight keypoints per leg, where each keypoint is labelled by a letter (e.g. **A**, **B**, **C**, …). Each keypoint is fixed to occur at a particular limb phase, where we recall from Equation (14.10) that the limb phase $\nu_i$ is equivalent to the gait phase $\mu_i$, only possibly shifted by $\pi$ radians (for each limb independently) to ensure that the swing phase of each limb occurs within the range $\nu_i \in (0, \pi]$. Table 14.1 and Figure 14.3 show the eight keypoints and their definitions, where $D$ is the configured double support phase length, and $\nu_{DE}$, $\nu_{GA}$ are tuneable phase offset parameters. At its core, each keypoint is just the definition of a particular instant of the gait walking cycle, e.g. **A** corresponds to the instant of foot strike for each leg, and **D** corresponds to foot lift-off. It is the ultimate goal of the KGG to calculate the required abstract space poses at each of these keypoints (based on the gait generator inputs), and use cubic spline interpolation to sample the thereby defined abstract space trajectory at the required limb phases (see Section 14.2.4.14).

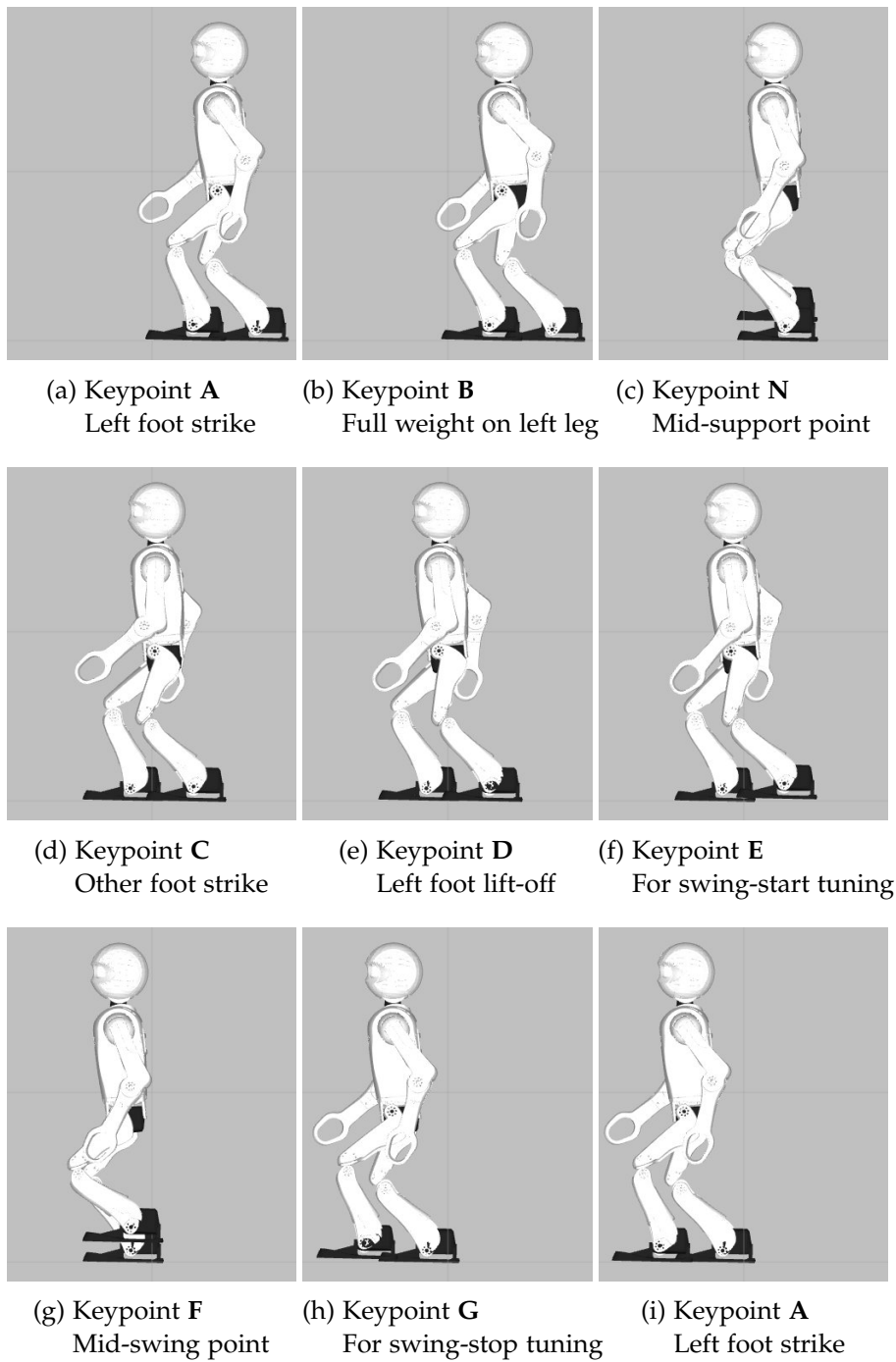

(a) Keypoint **A**
Left foot strike

(b) Keypoint **B**
Full weight on left leg

(c) Keypoint **N**
Mid-support point

(d) Keypoint **C**
Other foot strike

(e) Keypoint **D**
Left foot lift-off

(f) Keypoint **E**
For swing-start tuning

(g) Keypoint **F**
Mid-swing point

(h) Keypoint **G**
For swing-stop tuning

(i) Keypoint **A**
Left foot strike

Figure 14.3: Snapshots of full speed forwards walking of the KGG at the eight keypoints of the left leg. Refer to Table 14.1 for the corresponding keypoint limb phase values $\nu_i$. Note that the equivalent keypoints for the right leg are shifted by exactly $\pi$ radians in terms of the gait phase. Thus, for example, (a) corresponds to keypoint **C** for the right leg, and (b), (c), (d) and (e) correspond to the right leg keypoints **D**, **F**, **A** and **B**, respectively. During the double support phases **A** → **B** and **C** → **D**, it can be seen that the feet remain perfectly still while the torso continues to move forwards.

When the gait generator is executed, the entire trajectories for both legs are first generated and then sampled, as described, at the required limb phases $\nu_i$. This process is broken down into several steps, namely

1. Step size generation,

2. Keypoint rotation,

3. Keypoint reconciliation,

4. Keypoint adjustment,

5. Inclusion of step height,

6. Calculation of foot orientations,

7. Incorporation of leg swing out,

8. Calculation of support phase linear velocities,

9. Calculation of swing phase linear velocities,

10. Calculation of keypoint angular velocities,

11. Keypoint leaning,

12. Keypoint shifting,

13. Keypoint finalisation,

14. Finalisation of leg poses,

15. Calculation of support coefficients.

These steps are addressed in order in the following subsections.

### 14.2.4.1 *Step Size Generator*

To seed the leg tip space positions and orientations of the feet at each of the described keypoints, a step size generator is required. This is simply a function that, given the humanoid kinematics model and internal GCV vector $\mathbf{v}_i$ as inputs, produces the following set of outputs:

- 12 horizontally (i.e. **xy**) coplanar **leg tip position vectors** $\mathbf{x}^{l,r}_{A\text{-}D,F}$,[6] corresponding to all 16 left and right leg keypoints other than $\mathbf{E}$ and $\mathbf{G}$. The returned position vectors should implicitly encode the required translational step sizes of the robot for both legs.

- 12 **foot fused yaws** $\psi^{l,r}_{A\text{-}D,F}$, corresponding to all 16 left and right leg keypoints other than $\mathbf{E}$ and $\mathbf{G}$. The returned fused yaws should implicitly encode the required rotational step sizes for both legs.

---

6 The notation $\mathbf{x}^{l,r}_{A\text{-}D,F}$ is intended as shorthand to collectively refer to the 12 individual leg tip position vectors $\mathbf{x}^l_A$, $\mathbf{x}^r_A$, $\mathbf{x}^l_B$, $\mathbf{x}^r_B$, $\mathbf{x}^l_N$, $\mathbf{x}^r_N$, $\mathbf{x}^l_C$, $\mathbf{x}^r_C$, $\mathbf{x}^l_D$, $\mathbf{x}^r_D$, $\mathbf{x}^l_F$ and $\mathbf{x}^r_F$. Similar 'collective notation' is used throughout the remainder of this chapter.

- The nominal foot yaw $\psi_n$, and nominal foot tilt $(\tilde{\gamma}_n, \alpha_n)$, of each foot relative to the nominal ground plane N.[7]

- A so-called motion centre point $\mathbf{x}_{MC}$ characterising the centre of the motion profile given by the keypoint leg tip positions (see Figure 14.4 for a visual example).

- A so-called height adjustment normal $\hat{\mathbf{h}}$ giving the unit direction for later height adjustments of the keypoint leg tip positions.

- The required scalar step height $h_s$ of the swing leg.

Note that all points and vectors in the above list of step size generator outputs are expressed in body-fixed coordinates {B} relative to the hip centre point. The **E** and **G** keypoints are not required as outputs of the step size generator because they are calculated outright in Section 14.2.4.9 as part of the velocity loop shaping step. The remainder of the KGG algorithm is completely independent of how the listed outputs are computed, so the step size generator is like an interchangeable 'black box' that provides an initialisation for the robot poses at each of the gait keypoints.

As the step size generator produces the required left and right leg step sizes as a function of the internal GCV $\mathbf{v}_i$, ultimately it is the location in the KGG generation pipeline that effectuates the step size adjustment. The step size generator that was developed and implemented for the robots used in this thesis was motivated in part by the CPG gait. The required leg tip positions and foot fused yaws are generated by starting with a neutral halt pose in the abstract space, adding *pushout*, *leg swing* and *hip swing* components, and then projecting the result into the horizontal plane through the required motion centre point. The motion centre point $\mathbf{x}_{MC}$ is taken to be the intersection point between the $\mathbf{x}_B \mathbf{z}_B$ plane and the line joining the two raw calculated **N** points (with just the pushout applied). The pushout component separates the legs laterally and rotationally during walking, while the leg swing component generates the x, y and z hip rotations necessary to realise the commanded GCV. The hip swing component is oscillatory in the lateral direction, and is centred about the double support phases. It reduces the need for the torso to tilt in order to shift the weight of the robot laterally between the left and right legs.

An example figure of the main outputs of the step size generator is shown in Figure 14.4. As a further output, the height adjustment normal $\hat{\mathbf{h}}$ is calculated by spherical linear vector interpolation between $\hat{\mathbf{z}}_N$ and the unit vector pointing from the motion centre point to the hip centre point. The former direction of adjustment keeps the balance of the robot centred above its feet, while the latter keeps the keypoints centred inside the workspace of the robot—both of which are important considerations. The nominal foot yaws and tilts are calculated as the absolute tilt angles representations of the feet in the neutral halt pose relative to the torso frame {B}. As the final output of the step size generator, the step height is calculated as a linear combination of 1, $|v_{ix}|$ and $|v_{iy}|$ (up to a configured maximum step height), to allow step heights to increase with walking velocity.

---

7 Refer to the tilt angles rotation representation $(\psi, \gamma, \alpha) \in \mathbb{T}$ introduced in Section 5.4.3.

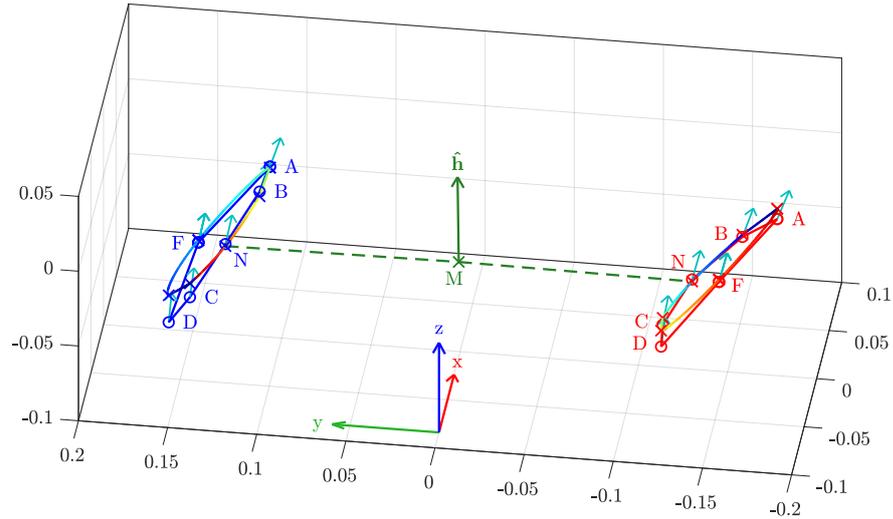

Figure 14.4: Plot of the main outputs of the step size generator for an internal GCV of $\mathbf{v}_i = (1, -0.25, -0.25)$. The smooth internal abstract space trajectory used by the step size generator to sample the raw keypoint leg tip positions '×' is shown by the multicoloured path. These points are projected into the horizontal $\mathbf{xy}$ plane through the motion centre point $M$ to yield the required leg tip positions $\mathbf{x}^{l,r}_{A\text{-}D,F}$, labelled '○'. The height adjustment normal vector $\hat{\mathbf{h}}$ is indicated relative to $M$, and is the direction along which the keypoints are adjusted in height in later steps, if necessary. The cyan arrows indicate the foot fused yaws $\psi^{l,r}_{A\text{-}D,F}$ at the keypoints, where it can be observed that these arrows change their orientation from $\mathbf{A}$ to $\mathbf{D}$, as per the commanded turning velocity of $v_{iz} = -0.25$.

### 14.2.4.2 Rotated Keypoints

As shown in Figure 14.5, the $\mathbf{x}_B\mathbf{y}_B$-planar keypoints produced by the step size generator (i.e. the leg tip positions $\mathbf{x}^{l,r}_{A\text{-}D,F}$) are rotated around the motion centre point $\mathbf{x}_{MC}$ into the N plane (nominal ground plane). This is because the robot, as previously described, is not assumed to walk with a nominally perfectly upright torso, and the step sizes should be relative to the ground (not relative to the torso), which is nominally parallel to the N plane. The update equation for all keypoints $* = \mathbf{A}, \mathbf{B}, \mathbf{N}, \dots$ of both legs is given by

$$\mathbf{x}^{l,r}_* \leftarrow \mathbf{x}_{MC} + {}^B_N R(\mathbf{x}^{l,r}_* - \mathbf{x}_{MC}), \tag{14.15}$$

where ${}^B_N R$ is the tilt rotation from {B} to {N} that was calculated in Equation (14.11). We note that many of the equations in this chapter apply to both the left and right legs of the robot, so for greater readability of the equations, by default we omit all further instances of the superscript '$l, r$'. With this convention, Equation (14.15) becomes

$$\mathbf{x}_* \leftarrow \mathbf{x}_{MC} + {}^B_N R(\mathbf{x}_* - \mathbf{x}_{MC}).  \tag{14.16}$$

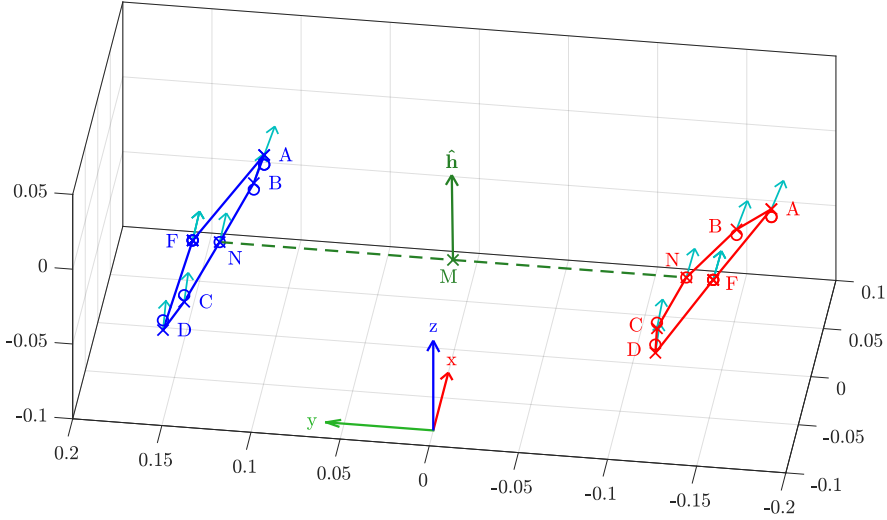

Figure 14.5: Plot of the rotated keypoints '×', obtained by rotating (about $M$) the output leg tip positions '∘' from the step size generator into the nominal ground plane N ($p_{yN} = 0.08$). Note that the foot fused yaws are reinterpreted as being relative to the N plane, and that $\hat{\mathbf{n}}$ and the motion centre point remain unchanged.

In the equations where the individual superscripts $l$ and $r$ are still important however, they are of course still explicitly included.

### 14.2.4.3  *Reconciled Keypoints*

During the double support phases $\mathbf{A} \rightarrow \mathbf{B}$ and $\mathbf{C} \rightarrow \mathbf{D}$, it is intended that both feet are in fixed contact with the ground. As such, even though both feet may move relative to the torso, neither foot should change its position or yaw within the N plane relative to the other foot.[8] To meet this requirement, the positions of the $\mathbf{A}$, $\mathbf{B}$, $\mathbf{C}$ and $\mathbf{D}$ keypoints are first adjusted axially along the diagonal $\mathbf{AC}$ and $\mathbf{BD}$ line segments so that they are of equal length. As indicated by the dashed lines in Figure 14.6, the $\mathbf{AC}$ line segment is the line joining the $\mathbf{A}$ keypoint of one foot to the $\mathbf{C}$ keypoint of the other. The $\mathbf{BD}$ line segment is similarly defined, and the $\mathbf{AC}$ and $\mathbf{BD}$ line segments that belong to the same double support phase are compared for length. Mathematically, we wish to have

$$\|\mathbf{x}_{AC}\| = \|\mathbf{x}_{BD}\|,  \tag{14.17}$$

---

8 Recall from the way that the CPG waveforms were generated that this property was not explicitly addressed by the CPG.

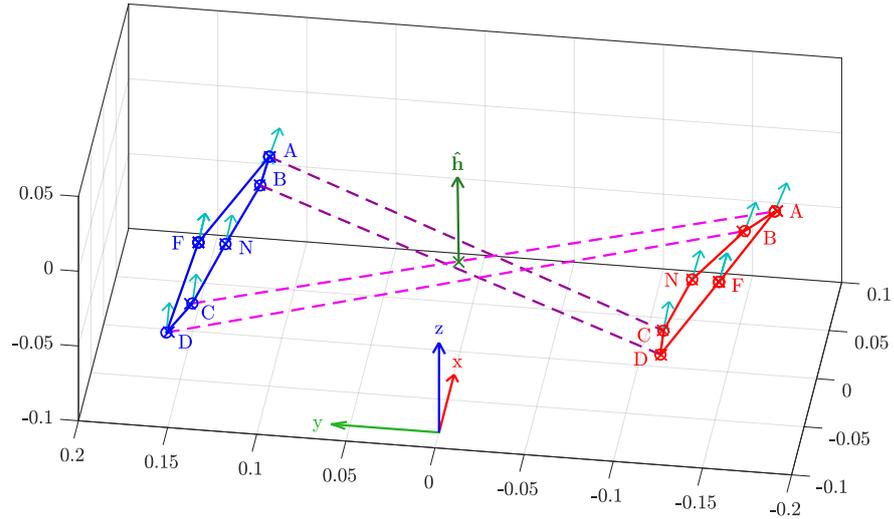

Figure 14.6: Plot of the reconciled keypoints '○', which modify the rotated keypoints '×' in such a way that the relative positions and yaws of the feet stay exactly constant during the double support phases. The effect of reconciliation is slight, but can be observed for example through the shortening of the dashed magenta **AC** line segment, and the equivalent lengthening of the dashed magenta **BD** line segment.

where

$$\mathbf{x}_{AC} = \mathbf{x}_C^o - \mathbf{x}_A, \tag{14.18a}$$

$$\mathbf{x}_{BD} = \mathbf{x}_D^o - \mathbf{x}_B, \tag{14.18b}$$

and where $\mathbf{x}_*$ are the keypoints of one leg, and $\mathbf{x}_*^o$ are the keypoints of the corresponding other leg. Equation (14.17) is made to be satisfied by applying the update equations

$$\mathbf{x}_A \leftarrow \mathbf{x}_A + \Delta\mathbf{x}_{AC}, \qquad \mathbf{x}_B \leftarrow \mathbf{x}_B - \Delta\mathbf{x}_{BD}, \tag{14.19a}$$

$$\mathbf{x}_C^o \leftarrow \mathbf{x}_C^o - \Delta\mathbf{x}_{AC}, \qquad \mathbf{x}_D^o \leftarrow \mathbf{x}_D^o + \Delta\mathbf{x}_{BD}, \tag{14.19b}$$

where

$$\Delta\mathbf{x}_{AC} = \tfrac{1}{4}\big(\|\mathbf{x}_{AC}\| - \|\mathbf{x}_{BD}\|\big)\frac{\mathbf{x}_{AC}}{\|\mathbf{x}_{AC}\|}, \tag{14.20a}$$

$$\Delta\mathbf{x}_{BD} = \tfrac{1}{4}\big(\|\mathbf{x}_{AC}\| - \|\mathbf{x}_{BD}\|\big)\frac{\mathbf{x}_{BD}}{\|\mathbf{x}_{BD}\|}. \tag{14.20b}$$

In addition to an adjustment of length, the **A**, **B**, **C** and **D** keypoints are also adjusted so that the relative **AC** and **BD** foot yaws are equal, i.e. so that

$$\psi_C^o - \psi_A = \psi_D^o - \psi_B, \tag{14.21}$$

where $\psi_*$ are the foot yaws of one leg, and $\psi_*^o$ are the foot yaws of the corresponding other leg (see Section 14.2.4.1). The associated update equations are given by

$$\psi_B \leftarrow \psi_B - \Delta\psi, \tag{14.22a}$$

$$\psi_C^o \leftarrow \psi_C^o - \Delta\psi, \tag{14.22b}$$

where

$$\Delta\psi = \tfrac{1}{2}\big((\psi_C^o - \psi_A) - (\psi_D^o - \psi_B)\big). \tag{14.23}$$

Only the **B** and **C** keypoints are adjusted in terms of foot yaw to ensure that the rotational step sizes are not inadvertently affected. Like the length adjustment, the foot yaw adjustments are generally minor, and do not negatively impact the shape of the foot trajectory.

### 14.2.4.4  *Adjusted Keypoints*

After length and yaw reconciliation, the keypoints are further adjusted to respect the swing ground plane S, referred to in short as just the S plane. To avoid premature foot strike, it is desired for the leg tip points of the two feet to be S-coplanar[9] at the start of each double support phase (i.e. in the moment of **AC**). Furthermore, the feet should push down or lift up from the ground immediately afterwards to exert a restoring moment on the robot to bring it towards the nominal orientation, i.e. towards the nominal ground plane N. The lack of such a restoring moment is a weakness of most classical 'virtual slope' implementations, as the premature foot-ground contact is avoided in such approaches, but the more the robot tilts, the more the foot in the direction of the tilt gives way, and allows the robot to tilt even further. This, despite the reduction of self-disturbances caused by stepping, counterintuitively often leads to a lower resistance to pushes.

As discussed in Section 14.2.3.3, the coordinate frame {S} for the swing ground plane is constructed as the unique frame that has zero fused yaw relative to {N}, and a local z-axis that is consistent with the torso phase roll and pitch $(p_{xS}, p_{yS}) \in \mathbb{P}^2$. For the purposes of keypoint adjustment, two other planes, the intermediate ground plane I, and support ground plane J, are constructed via spherical linear interpolation from {N} to {S}. These two planes determine the strength of the generated swing ground plane corrective action restoring moment, as they trim how much the feet try to return to being N-coplanar after striking the ground in an S-coplanar configuration. It should be noted that the coordinate frames {I} and {J} corresponding to the two new planes are guaranteed to be pure tilt rotations relative to {N}, as both {N} and {S} are, and spherical linear interpolation preserves this property Section 7.3.5.3.

---

9  Two points $\mathbf{x}_1$ and $\mathbf{x}_2$ are *S-coplanar* if the vector between them is parallel to the S plane, i.e. $(\mathbf{x}_2 - \mathbf{x}_1) \cdot \hat{\mathbf{z}}_S = 0$. A similar definition applies, for example, for the notion of two points being *N-coplanar*.

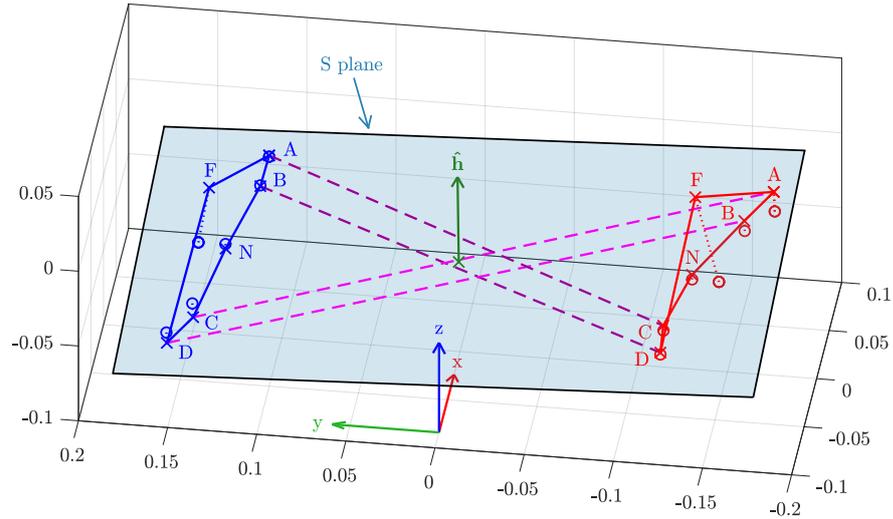

Figure 14.7: Plot of the adjusted keypoints '×' in comparison to the reconciled keypoints '○'. Dotted lines join each of the matching pairs of keypoints, including in particular the **F** keypoints, which incorporate the effect of the desired step height $h_s$. The S plane is illustrated, and passes through all four **A** and **C** keypoints. All four **B** and **D** keypoints lie in the I plane (not pictured for clarity), and the two **N** keypoints lie in the J plane (also not pictured for clarity).

As indicated in Figure 14.7, the keypoints are first rotated around the motion centre point $\mathbf{x}_{MC}$ as follows:

$$\mathbf{x}_{A,C} \leftarrow \mathbf{x}_{MC} + {}^{BN}_{S}R(\mathbf{x}_{A,C} - \mathbf{x}_{MC}), \qquad (14.24a)$$

$$\mathbf{x}_{B,D} \leftarrow \mathbf{x}_{MC} + {}^{BN}_{I}R(\mathbf{x}_{B,D} - \mathbf{x}_{MC}), \qquad (14.24b)$$

$$\mathbf{x}_{N,F} \leftarrow \mathbf{x}_{MC} + {}^{BN}_{J}R(\mathbf{x}_{N,F} - \mathbf{x}_{MC}), \qquad (14.24c)$$

where ${}^{BN}_{S}R$, for example, is the referenced rotation (see Section 7.1.1.5) from {N} to {S} expressed as a rotation relative to {B}. The **A** and **C** keypoints are rotated about the motion centre point into the S plane, the **B** and **D** keypoints are rotated into the I plane, and the **N** and **F** keypoints are rotated into the J plane. This is to account for the fact that the robot is expected to have different tilts at different stages of the restoring step.

After being rotated, the relative heights of the keypoints are adjusted to ensure that the keypoints are centred about the 'neutral' **N** points heightwise, and that the correction from the {S} to the {I} plane during the two double support phases is balanced between corrections of the striking and lifting foot. This, amongst other things, ensures that the final generated support profiles have minimal unnecessary oscillations or accelerations, which would be disruptive to the robot. The four **A** and **C** keypoints are first adjusted in height relative to the **B** and **D** keypoints along the height adjustment normal $\hat{\mathbf{n}}$ (see Section 14.2.4.1). If $h_{AB}$ is how much **AC** would have to be adjusted normal to S in order to make **AB** on average S-neutral,[10] and $h_{CD}$ is how much **AC** would have to be adjusted normal to S in order to make **CD** on average S-neutral, then

$$h_{AB} = -\tfrac{1}{2}\hat{\mathbf{z}}_S \bullet \left(\mathbf{x}_A^l - \mathbf{x}_B^l + \mathbf{x}_A^r - \mathbf{x}_B^r\right), \tag{14.25a}$$

$$h_{CD} = -\tfrac{1}{2}\hat{\mathbf{z}}_S \bullet \left(\mathbf{x}_C^l - \mathbf{x}_D^l + \mathbf{x}_C^r - \mathbf{x}_D^r\right). \tag{14.25b}$$

The required height adjustment of the four **AC** keypoints is then given by the update equation

$$\lambda_{AC} = \tfrac{1}{\hat{\mathbf{z}}_S \bullet \hat{\mathbf{h}}}\left(h_{AB} + \eta_{adj}\left(h_{CD} - h_{AB}\right)\right), \tag{14.26a}$$

$$\mathbf{x}_{A,C} \leftarrow \mathbf{x}_{A,C} + \lambda_{AC}\hat{\mathbf{h}}, \tag{14.26b}$$

where $\eta_{adj} \in [0,1]$ interpolates between making **AB** on average S-neutral, and making **CD** on average S-neutral.

After the height adjustment of the **AC** keypoints relative to the **BD** keypoints, all eight **ABCD** keypoints are then adjusted together in height relative to the two **N** keypoints. If $h_{BC}$ is how much **ABCD** would have to be adjusted normal to the support ground plane J in order to make **BC** on average J-neutral with respect to the **N** keypoints, then

$$h_{BC} = -\tfrac{1}{4}\hat{\mathbf{z}}_J \bullet \left(\mathbf{x}_B^l + \mathbf{x}_C^l + \mathbf{x}_B^r + \mathbf{x}_C^r - 2\mathbf{x}_N^l - 2\mathbf{x}_N^r\right). \tag{14.27}$$

The required update equation of the **ABCD** keypoints is then given by

$$\lambda_{BC} = \tfrac{1}{\hat{\mathbf{z}}_J \bullet \hat{\mathbf{h}}}\,h_{BC}, \tag{14.28a}$$

$$\mathbf{x}_{A,B,C,D} \leftarrow \mathbf{x}_{A,B,C,D} + \lambda_{BC}\hat{\mathbf{h}}. \tag{14.28b}$$

The effect of the combination of both height adjustments can be seen in Figure 14.7. It should be noted that if the swing ground plane S is equal to the nominal ground plane N, then

$$S \equiv I \equiv J \equiv N, \tag{14.29}$$

and all 10 support keypoints are in the N plane through the motion centre point $\mathbf{x}_{MC}$. In this situation, the adjusted keypoints differ from the rotated keypoints (see Section 14.2.4.2) only by the generally minor effects of reconciliation.

---

10 The **AB** double support phases are *S-neutral* if the mean of the changes in height from **A** to **B** relative to the S plane is zero, i.e. $(\mathbf{x}_B^l - \mathbf{x}_A^l + \mathbf{x}_B^r - \mathbf{x}_A^r) \bullet \hat{\mathbf{z}}_S = 0$. A similar definition holds for the **CD** double support phases being S-neutral.

### 14.2.4.5  *Step Height*

So far, the required step height has been calculated by the step size generator in Section 14.2.4.1, but not yet applied to the **F** keypoints. To do this, each **F** keypoint is first interpolated within the J plane towards the corresponding **N** keypoint. This allows the legs to be configured to draw slightly inwards during the swing phase, aiding the lateral shifting of weight. The **F** keypoint is then shifted upwards normal to the S plane, to be the exact step height $h_s$ above the support leg **N** keypoint. The update equations are as follows:

$$\mathbf{p}_F = \mathbf{x}_F + \eta_{FN}(\mathbf{x}_N - \mathbf{x}_F), \tag{14.30a}$$

$$\mathbf{x}_F \leftarrow \mathbf{p}_F + \hat{\mathbf{z}}_S \max\{h_s + \hat{\mathbf{z}}_S \cdot (\mathbf{x}_N^o - \mathbf{p}_F),\, 0\}, \tag{14.30b}$$

where $\mathbf{x}_N^o$ is the **N** keypoint of the other leg, and $\eta_{FN}$ is the required interpolation factor from **F** towards **N**. Figure 14.7 illustrates an example of the resulting **F** keypoints.

### 14.2.4.6  *Foot Orientations*

The required foot fused yaws $\psi_*$ relative to the nominal ground plane N at each keypoint are known from Section 14.2.4.1. The foot tilt rotation components relative to N must still be computed however, as a function of the nominal, continuous and support foot tilts.

There are two different interpretations of foot tilts, torso-fixed and foot-fixed. Torso-fixed foot tilts are applied relative to the frame {N}, and are thereby fixed relative to the torso of the robot, while foot-fixed foot tilts are applied relative to the yaw component of the foot. A parameter $r_t \in [0, 1]$ is used to interpolate between these two interpretations of foot tilt, where $r_t = 0$ corresponds to torso-fixed, and $r_t = 1$ corresponds to foot-fixed.

Taking the nominal foot tilt as an example, non-zero foot tilts are often required due to the local shape and/or symmetry of the foot, or to influence the oscillatory lateral shifting of weight between the two legs. When taking a turning step, in the former case a foot-fixed interpretation would ensure that the foot-ground contact remains constant throughout the entire step, as desired, while in the latter case it would be setting up the next lateral oscillation in the already-turned direction. This is beneficial, but also somewhat of an overcompensation as the torso is not actually in line with the turned direction until the step is over. A torso-fixed interpretation in the same situation however undercompensates the rotation, as the lateral oscillation at the end of the step should actually be in the turned direction. As a result, tuning of the $r_t$ parameter is required to determine which intermediate value works best for each robot. The continuous foot tilt can be seen as a dynamic offset to the nominal foot tilt, and so is also interpreted using the $r_t$ parameter. The support foot tilt however is related to the transient balance of the robot, most notably the torso, and is thus always interpreted as torso-fixed.

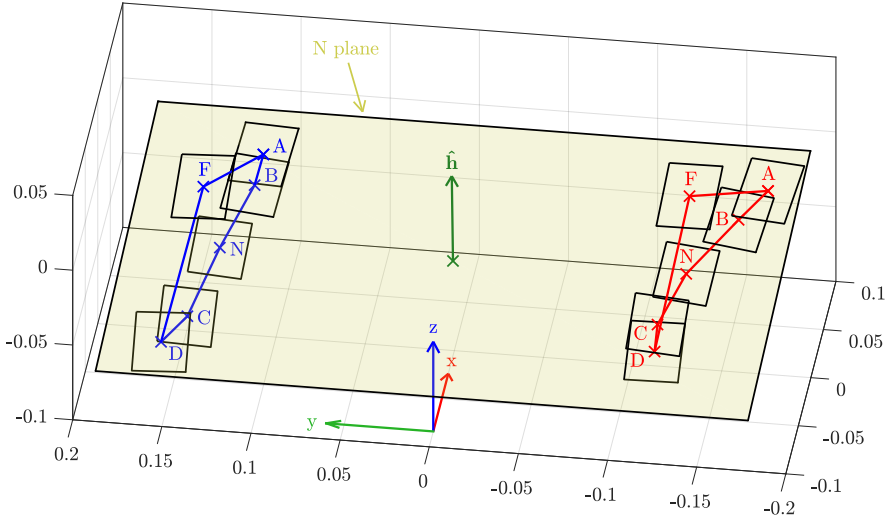

Figure 14.8: Plot of the adjusted keypoints '×' and their respective foot orientations, for a continuous foot tilt of $(\tilde{\gamma}_c, \alpha_c) = (-0.7, 0.02)$ and a support foot tilt of $(\tilde{\gamma}_s, \alpha_s) = (0.5, 0.10)$. Note that the tilts are interpreted relative to the N plane, and that the slight required CW turning velocity of the robot ($v_{iz} = -0.25$) is realised through the smooth changes in foot yaw from keypoints **A** to **D**.

While the continuous and nominal foot tilts are applied throughout the entire trajectory, the support foot tilt is faded in and out during the **AB** and **CD** double support phases respectively. At each instant, and for every keypoint $* = \mathbf{A}, \mathbf{B}, \mathbf{N}, \ldots$, the tilts are first converted into their equivalent torso-fixed representations, and then summed together to give the output torso-fixed foot tilts. If $(\tilde{\gamma}_c, \alpha_c)$ and $(\tilde{\gamma}_s, \alpha_s)$ are the absolute tilt angles rotations corresponding to the continuous foot tilt $P_c = (p_{xc}, p_{yc})$ and support foot tilt $P_s = (p_{xs}, p_{ys})$ respectively, mathematically this process is given by

$$\gamma_{r*} = r_t(\psi_* - \psi_n), \tag{14.31a}$$

$$F_{t*} = (\tilde{\gamma}_n + \gamma_{r*}, \alpha_n) \oplus (\tilde{\gamma}_c + \gamma_{r*}, \alpha_c), \tag{14.31b}$$

$$(\tilde{\gamma}_*, \alpha_*) = \begin{cases} F_{t*} \oplus (\tilde{\gamma}_s, \alpha_s) & \text{for } * = \mathbf{B}, \mathbf{N}, \mathbf{C}, \\ F_{t*} & \text{for } * = \mathbf{A}, \mathbf{D}, \mathbf{F}, \end{cases} \tag{14.31c}$$

where we recall from Section 14.2.4.1 that $\psi_n$ is the nominal foot yaw, and $(\tilde{\gamma}_n, \alpha_n)$ is the nominal foot tilt. The complete foot orientations of the keypoints, illustrated in Figure 14.8, are then given relative to the N plane by the absolute tilt angles representations $(\psi_*, \tilde{\gamma}_*, \alpha_*)$. Note that if $r_t = 0$, i.e. all foot tilts are considered to be torso-fixed, the output absolute foot tilts are independent of the keypoint foot yaws $\psi_*$, as expected. Note also that if a keypoint has the nominal foot yaw, i.e. $\psi_* = \psi_n$, the foot orientation is as a result independent of $r_t$, as expected.

As a final modification, the foot orientations at the **F** keypoints are rotated from the N plane to the S plane, to minimise the likelihood that the toe or heel of a foot prematurely strikes the ground during swing. This is done by applying the rotation $_S^N R$ to the absolute tilt angles rotation $(\psi_F, \tilde{\gamma}_F, \alpha_F)$, and then recomputing these values.

### 14.2.4.7  *Leg Swing Out*

The **F** keypoints are further adjusted to apply the action of <span style="color:red">swing out</span>. The idea of swing out is that if for example the robot is close to falling towards the left during a left support phase, then the swing leg can be made to curve out widely to the right, while it is in the air, to provide a counterbalancing effect. While this may be the typical case, such a modification of the swing leg trajectory is possible continuously in any direction, as given by the required <span style="color:red">swing out tilt</span> $P_o = (p_{xo}, p_{yo})$ relative to the N plane.

The phase roll component $p_{xo}$ of the commanded swing out tilt is first limited on a per-leg basis to avoid swinging either leg too far inwards. For each leg, the resulting tilt rotation is then converted to rotation matrix form, $_O^N R$, where {O} is a leg-dependent placeholder frame representing the resultant frame when swing out is applied to the nominal ground frame {N}. The **F** keypoints $\mathbf{x}_F^{l,r}$ are rotated about their corresponding hip points $\mathbf{p}_h^{l,r}$ (see page 385) by their respective swing out rotations, as follows:

$$\mathbf{x}_F \leftarrow \mathbf{p}_h + {}^{BN}_O R(\mathbf{x}_F - \mathbf{p}_h), \tag{14.32}$$

where

$$^{BN}_O R = {}^B_N R \, {}^N_O R \, {}^B_N R^T \tag{14.33}$$

is the required referenced rotation. The foot tilt parameters $(\psi_F, \tilde{\gamma}_F, \alpha_F)$ are also rotated by $_O^N R$, and recomputed relative to N. The effect of the leg swing out adjustment of the **F** keypoints can be identified in Figure 14.9. Note that after this step in the <span style="color:blue">KGG</span> generation pipeline, no keypoints are adjusted relative to one another ever again, just rotated or shifted as a whole profile.

### 14.2.4.8  *Support Phase Linear Velocities*

In order to join the keypoints together in a cyclic path as a function of the limb/gait phase, keypoint velocities must first be computed that respect the locations of the keypoints and allow for smooth cubic spline path segments to be formed. For the linear velocities this is done in two stages, first for the support phase keypoints **ABNCD**, and then separately for the swing phase keypoints **DEFGA**. The support phase velocities are calculated separately first to make sure that the swing behaviour of the legs, including for example the effects of swing out, cannot possibly affect the highly balance-critical support phases. In addition to this, if the complete cycle were connected in one go, there would be insufficient degrees of freedom (of the connecting cubic splines) to impose the extra constraints on the path that are required. The gained degrees of freedom come from relaxing the constraints of continuous acceleration at the keypoints **A** and **D**. As jumps in acceleration are actually expected at the beginning and end of swing, this weakening of constraints is intentional and advantageous.

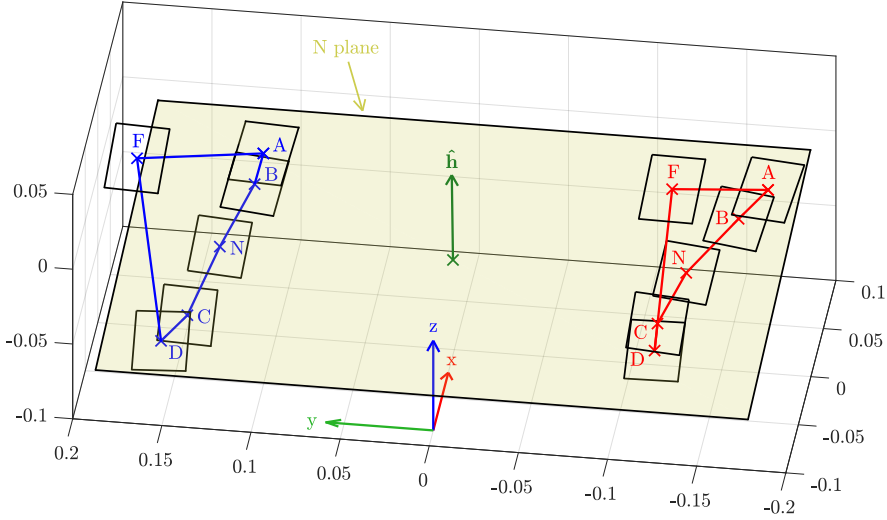

Figure 14.9: Plot of the adjusted keypoints '×' and corresponding foot orientations after the application of a leg swing out tilt of $(\tilde{\gamma}_o, \alpha_o) = (-0.2, 0.1)$. Note that the swing out tilt (equivalent to $P_o$ from Section 14.2.3.1) is interpreted relative to the N plane, and also affects the orientation of the feet at the **F** keypoints.

Care needs to be taken that the calculated linear velocities do not cause unnecessary oscillations in speed or direction, as ultimately the foot, with the inertia of the leg behind it, will need to follow the generated trajectory. To generate keypoint velocities that minimise the need for accelerations, a set of parametric cubic splines as a function of the limb phase $\nu_i$ are fitted through the keypoints, subject to a list of constraints. Given the support keypoint positions $\mathbf{x}_{A\text{-}D}^{l,r}$ and the differences $\nu_{AB}$, $\nu_{BN}$, $\nu_{NC}$ and $\nu_{CD}$ in limb phase between the successive support keypoints,[11] the unknown variables to solve for are the support keypoint velocities $\mathbf{v}_{A\text{-}D}^{l,r}$ (30 scalars). For convenience, we denote the linear interpolant velocities between the keypoints $\tilde{\mathbf{v}}_{**}^{l,r}$, where for example

$$\tilde{\mathbf{v}}_{AB}^{l,r} = \frac{1}{\nu_{AB}}(\mathbf{x}_B^{l,r} - \mathbf{x}_A^{l,r}).$$  (14.34)

Recall that the '$l,r$' superscripts could have been omitted in Equation (14.34), as by convention they are implied when no superscript is present.

The list of constraints for the calculation of the support phase velocities is as follows:

---

11 For example, $\nu_{AB} \equiv \nu_B - \nu_A = (-\pi + D) - (-\pi) = D$ (see Table 14.1).

A) The linear velocity and acceleration should be continuous at all keypoints. A non-cyclic sequence of scalar cubic splines through a set of keypoints with continuous first and second derivatives has exactly two degrees of freedom, regardless of the number of intermediate points. Thus, as there are two legs and three coordinates per keypoint, this results in 18 equations,[12] given by

$$A_p V_p = B_p, \tag{14.35}$$

where once each for the left and right legs ('$l$' and '$r$') we have

$$A_p = \begin{bmatrix} \nu_{BN} & 2(\nu_{AB}+\nu_{BN}) & \nu_{AB} & 0 & 0 \\ 0 & \nu_{NC} & 2(\nu_{BN}+\nu_{NC}) & \nu_{BN} & 0 \\ 0 & 0 & \nu_{CD} & 2(\nu_{NC}+\nu_{CD}) & \nu_{NC} \end{bmatrix},$$

$$V_p = \begin{bmatrix} \leftarrow \mathbf{v}_A \rightarrow \\ \leftarrow \mathbf{v}_B \rightarrow \\ \leftarrow \mathbf{v}_N \rightarrow \\ \leftarrow \mathbf{v}_C \rightarrow \\ \leftarrow \mathbf{v}_D \rightarrow \end{bmatrix}, \quad B_p = 3\begin{bmatrix} \leftarrow \nu_{AB}\bar{\mathbf{v}}_{BN} + \nu_{BN}\bar{\mathbf{v}}_{AB} \rightarrow \\ \leftarrow \nu_{BN}\bar{\mathbf{v}}_{NC} + \nu_{NC}\bar{\mathbf{v}}_{BN} \rightarrow \\ \leftarrow \nu_{NC}\bar{\mathbf{v}}_{CD} + \nu_{CD}\bar{\mathbf{v}}_{NC} \rightarrow \end{bmatrix}. \tag{14.36}$$

B) The component of the acceleration of the cubic spline path within the S plane should be zero at **A**. Similarly, the component of the acceleration of the cubic spline path within the I plane should be zero at **D**. This produces a *relaxed* cubic spline within these planes, and ensures that the spline does not have a turning radius relative to the ground at foot strike and lift-off. The corresponding constraint equations are given by

$$(2\mathbf{v}_A + \mathbf{v}_B) \cdot \hat{\mathbf{x}}_S = 3(\bar{\mathbf{v}}_{AB} \cdot \hat{\mathbf{x}}_S), \tag{14.37a}$$

$$(2\mathbf{v}_A + \mathbf{v}_B) \cdot \hat{\mathbf{y}}_S = 3(\bar{\mathbf{v}}_{AB} \cdot \hat{\mathbf{y}}_S), \tag{14.37b}$$

$$(\mathbf{v}_C + 2\mathbf{v}_D) \cdot \hat{\mathbf{x}}_I = 3(\bar{\mathbf{v}}_{CD} \cdot \hat{\mathbf{x}}_I), \tag{14.37c}$$

$$(\mathbf{v}_C + 2\mathbf{v}_D) \cdot \hat{\mathbf{y}}_I = 3(\bar{\mathbf{v}}_{CD} \cdot \hat{\mathbf{y}}_I). \tag{14.37d}$$

As these equations apply once for the left leg and once for the right leg, this corresponds to a further 8 constraint equations.

C) The component of the velocity normal to the S plane at foot strike should have a mean of zero across the left and right legs, as should the component of the velocity normal to the I plane at foot lift-off. This is to ensure that the robot as a whole does not, on average, have an upward or downward velocity at the beginning and end of stepping (i.e. the robot is not falling into its knees or extending both of its legs at these instants). The corresponding constraint equations are given by

$$(\mathbf{v}_A + \mathbf{v}_C^o) \cdot \hat{\mathbf{z}}_S = 0, \tag{14.38a}$$

$$(\mathbf{v}_B + \mathbf{v}_D^o) \cdot \hat{\mathbf{z}}_I = 0, \tag{14.38b}$$

where as before, the superscript '*o*' refers to the 'other leg' to the one the equation is being applied to. Equation (14.38) corresponds to a total of 4 further constraints (2 constraint equations per leg).

---

12 Two degrees of freedom multiplied by two legs and three coordinates is a total of $2 \cdot 2 \cdot 3 = 12$ degrees of freedom for the whole 30 parameter trajectory, so we can expect $30 - 12 = 18$ constraint equations.

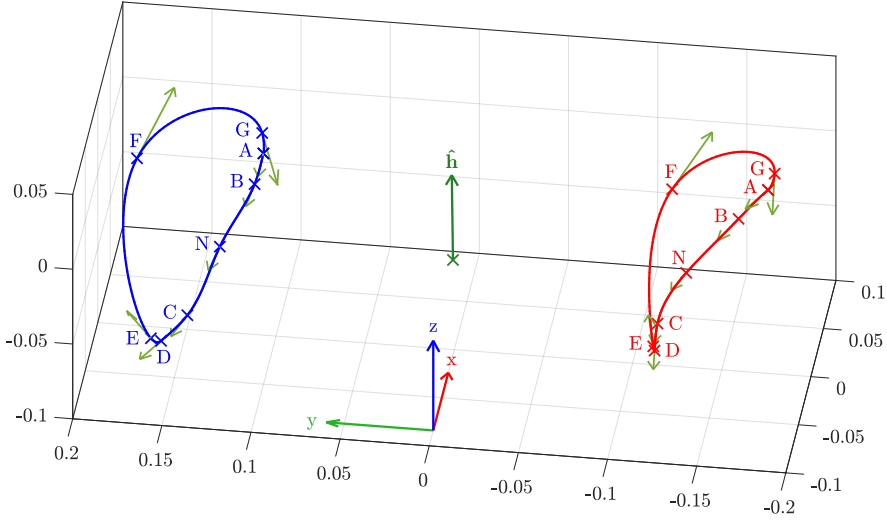

Figure 14.10: Plot of the adjusted keypoints '×', their associated linear velocities (green arrows), and the cubic splines (solid blue/red lines) that were used to calculated them. Note that both support and swing phase velocities are shown, and that the leg tip positions at the **E** and **G** swing keypoints are known from the solution of Equation (14.39).

With a total of 30 constraint equations and 30 scalar unknowns, a single combined matrix equation can be formulated and solved for all 10 support keypoint velocities $\mathbf{v}_{A\text{-}D}^{l,r}$ using QR decomposition. An example of the resulting velocities and cubic spline segments used to calculate them is shown in Figure 14.10. For more details on the whole process of cubic spline interpolation and calculation of the associated intermediate velocities, refer to Appendices A.1.5.3 and A.1.5.4.

It should be noted that this constraint-based method of calculating the support keypoint velocities behaves exactly as expected when S ≡ N. In this case, all 10 support keypoints are N-coplanar, and the resulting velocities and cubic spline path segments all lie completely in the N plane. This observation supports the aim of the KGG to minimise unnecessary CoM height fluctuations.

### 14.2.4.9  *Swing Phase Linear Velocities*

Just like the support phase keypoint velocities, the swing phase keypoint velocities are determined by fitting a set of parametric cubic splines through the keypoints, and solving for the required velocities based on a set of constraints (see Appendix A.1.5.4). Given the keypoint positions $\mathbf{x}_{A,F,D}^{l,r}$, the previously calculated boundary condition velocities $\mathbf{v}_{A,D}^{l,r}$, and the differences $\nu_{DE}$, $\nu_{EF}$, $\nu_{FG}$ and $\nu_{GA}$ in limb phase between the successive keypoints, the unknown variables to solve for are the swing keypoint velocities $\mathbf{v}_{E\text{-}G}^{l,r}$ (18 scalars), and positions $\mathbf{x}_{E,G}^{l,r}$ (12 scalars). The $\mathbf{E}$ and $\mathbf{G}$ keypoints are floating control points that are used for loop shaping of the swing trajectory, and to ensure that leg swing only occurs when the foot is off the ground. The list of constraints is as follows:

A)  The linear velocity and acceleration should be continuous at all keypoints. Similar to Equation (14.35), the resulting 9 constraint equations per leg are given by

$$A_w V_w = B_w, \tag{14.39}$$

where once for each leg ('$l$' and '$r$') we have

$$A_w = \begin{bmatrix} 2(\nu_{DE}+\nu_{EF}) & \nu_{DE} & 0 & 3\nu_{DEF} & 0 \\ \nu_{FG} & 2(\nu_{EF}+\nu_{FG}) & \nu_{EF} & 3\frac{\nu_{FG}}{\nu_{EF}} & -3\frac{\nu_{EF}}{\nu_{FG}} \\ 0 & \nu_{GA} & 2(\nu_{FG}+\nu_{GA}) & 0 & 3\nu_{FGA} \end{bmatrix},$$

$$V_w = \begin{bmatrix} \leftarrow \mathbf{v}_E \rightarrow \\ \leftarrow \mathbf{v}_F \rightarrow \\ \leftarrow \mathbf{v}_G \rightarrow \\ \leftarrow \mathbf{x}_E \rightarrow \\ \leftarrow \mathbf{x}_G \rightarrow \end{bmatrix}, \quad B_w = \begin{bmatrix} \leftarrow 3\left(\frac{\nu_{DE}}{\nu_{EF}}\mathbf{x}_F - \frac{\nu_{EF}}{\nu_{DE}}\mathbf{x}_D\right) - \nu_{EF}\mathbf{v}_D \rightarrow \\ \leftarrow 3\left(\frac{\nu_{FG}}{\nu_{EF}} - \frac{\nu_{EF}}{\nu_{FG}}\right)\mathbf{x}_F \rightarrow \\ \leftarrow 3\left(\frac{\nu_{FG}}{\nu_{GA}}\mathbf{x}_A - \frac{\nu_{GA}}{\nu_{FG}}\mathbf{x}_F\right) - \nu_{FG}\mathbf{v}_A \rightarrow \end{bmatrix},$$

where

$$\nu_{DEF} = \frac{\nu_{DE}}{\nu_{EF}} - \frac{\nu_{EF}}{\nu_{DE}}, \tag{14.40a}$$

$$\nu_{FGA} = \frac{\nu_{FG}}{\nu_{GA}} - \frac{\nu_{GA}}{\nu_{FG}}. \tag{14.40b}$$

B)  Similar to Constraint B of the support phase keypoint velocities, the component of the acceleration of the cubic spline path should be zero at $\mathbf{A}$ within the S plane, and zero at $\mathbf{D}$ within the I plane. The corresponding 4 constraint equations per leg are given by

$$(3\mathbf{x}_G + \nu_{GA}\mathbf{v}_G) \cdot \hat{\mathbf{x}}_S = (3\mathbf{x}_A - 2\nu_{GA}\mathbf{v}_A) \cdot \hat{\mathbf{x}}_S, \tag{14.41a}$$

$$(3\mathbf{x}_G + \nu_{GA}\mathbf{v}_G) \cdot \hat{\mathbf{y}}_S = (3\mathbf{x}_A - 2\nu_{GA}\mathbf{v}_A) \cdot \hat{\mathbf{y}}_S, \tag{14.41b}$$

$$(3\mathbf{x}_E - \nu_{DE}\mathbf{v}_E) \cdot \hat{\mathbf{x}}_I = (3\mathbf{x}_D + 2\nu_{DE}\mathbf{v}_D) \cdot \hat{\mathbf{x}}_I, \tag{14.41c}$$

$$(3\mathbf{x}_E - \nu_{DE}\mathbf{v}_E) \cdot \hat{\mathbf{y}}_I = (3\mathbf{x}_D + 2\nu_{DE}\mathbf{v}_D) \cdot \hat{\mathbf{y}}_I. \tag{14.41d}$$

C) The height of the swing start tuning keypoint **E** above **D** measured normal to the S plane should be exactly $\eta_E \in [0, 1]$ times the corresponding height of **F** above **D**. Similarly, the height of the swing stop tuning keypoint **G** above **A** measured normal to the S plane should be exactly $\eta_G \in [0, 1]$ times the corresponding height of **F** above **A**. For each leg, this leads to the 2 constraint equations

$$\mathbf{x}_E \bullet \hat{\mathbf{z}}_S = (\eta_E \mathbf{x}_F + (1 - \eta_E)\mathbf{x}_D) \bullet \hat{\mathbf{z}}_S, \qquad (14.42a)$$

$$\mathbf{x}_G \bullet \hat{\mathbf{z}}_S = (\eta_G \mathbf{x}_F + (1 - \eta_G)\mathbf{x}_A) \bullet \hat{\mathbf{z}}_S. \qquad (14.42b)$$

By tuning of the $\nu_{DE}$, $\nu_{GA}$ and $\eta_E$, $\eta_G$ parameters, both the limb phase and ground height separations of the **E** and **G** keypoints can be adjusted, allowing both the nature and sharpness of the ground contact transitions to be trimmed.

With 15 constraint equations and 15 scalar unknowns per leg, two independent matrix equations can be formulated and solved using QR decomposition. This yields the required intermediate swing velocities $\mathbf{v}_{E\text{-}G}$ and positions $\mathbf{x}_{E,G}$. An example of the resulting calculated swing phase velocities and leg tip positions $\mathbf{x}_{E,G}$ is shown in Figure 14.10.

### 14.2.4.10  *Keypoint Angular Velocities*

In addition to the linear velocities at each of the keypoints, intermediate angular velocities also need to be computed to allow the foot orientations at each of the keypoints to be connected into a smooth trajectory. As discussed in Section 7.3.3.3, smooth non-linear spline interpolation between multiple orientation keypoints is surprisingly complex, and often relies on the transformation of the orientations into other spaces, such as canonical coordinates or the Cayley-Rodrigues parameters (Kang and Park, 1999). In this work, we transform the keypoint orientations into the absolute tilt phase space $(\tilde{p}_x, \tilde{p}_y, \tilde{p}_z)$, and calculate the required velocities based on cubic spline interpolation within this three-dimensional Cartesian space. This has the distinct advantage, as opposed to other available methods, that the interpolation is partitioned into dimensions that are highly problem-relevant. Most notably, the fused yaw velocities as a result of the interpolation are calculated completely independently from the tilt velocities, which is a desirable trait given the balance and planar contacts nature of the application. The yaw and tilt parameter velocities are easily combined into a single final angular velocity at each keypoint.

While for the calculation of the linear velocities the leg inertias were of significant concern, the rotational inertias of the feet are relatively small. As such, the demand for continuous angular acceleration, at least during the slower moving support phase, is outweighed by the requirement for exact positioning with no overshooting and/or oscillations. As a result, shape-preserving piecewise cubic Hermite interpolating polynomials (Fritsch and Carlson, 1980) were chosen for the calculation of the support keypoint angular velocities (see Appendix A.1.5.5). This method of cubic spline interpolation relaxes the constraint of continuous second derivatives, but in return is direct to evaluate, guaranteed to never locally overshoot data, and is in general more 'local' in that a spline between two keypoints is only affected by the values at those two keypoints and the values at the single next keypoints on either side of them. This means that changes to the foot orientation in one section of the gait cycle has no effect on other sections, increasing independence and robustness.

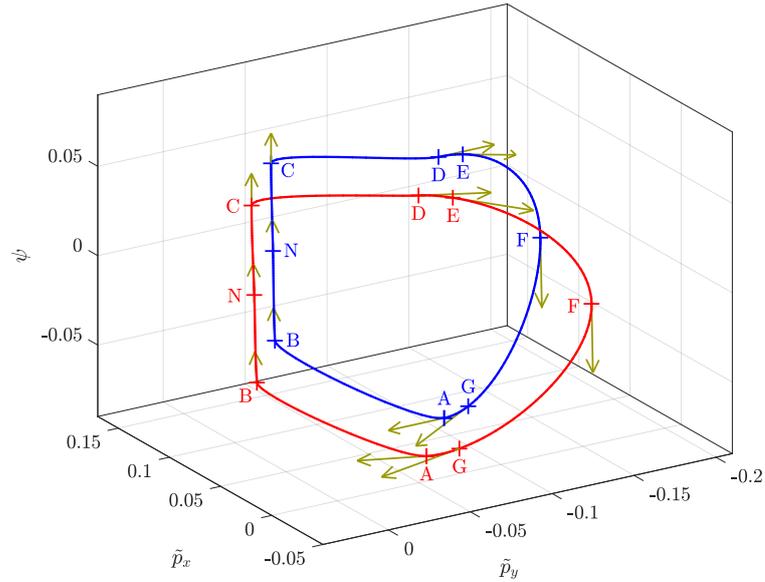

Figure 14.11: Plot of the absolute tilt phase space cubic splines used to calculate the required adjusted keypoint angular velocities. Note that the plotted '+' points correspond to the absolute tilt phase space representations of the orientations of the feet relative to the nominal ground plane N at each of the keypoints. The fitted tilt phase velocities (dark yellow arrows) are converted to angular velocity vectors relative to N, which are then converted to the required body-fixed angular velocities by applying $^B_N R$.

The required support phase keypoint angular velocities $\mathbf{\Omega}^{l,r}_{A\text{-}D}$ (30 scalars) are calculated by applying shape-preserving Hermite cubic interpolation to the periodic **ABNCDF** keypoint loop, as shown in Figure 14.11. The **E** and **G** keypoints are not explicitly considered for the calculation of the angular velocities as they are only there for positional loop shaping. The angular velocities $\mathbf{\Omega}^{l,r}_{E,G}$ at those two keypoints are computed post factum by simply evaluating the fitted splines from **D** to **F** to **A** at the required limb phases. This is equivalent to, but simpler than, the approach of including these two keypoints in the interpolation as free points with the not-a-knot condition.

For notational convenience, we enumerate the cyclic sequence of keypoints **FABNCDF** using the indices $i = 0, 1, \ldots, 6$. Let $h_i$ denote the positive difference in limb phase from keypoint $i$ to $i + 1$, and let $p$ denote any of the absolute tilt phase space components $\bar{p}_x$, $\bar{p}_y$ and $\bar{p}_z$ of the keypoint foot orientations. The slopes of the piecewise linear interpolants between the keypoints are given by

$$\bar{v}_i = \frac{p_{i+1} - p_i}{h_i} \quad \forall\, 0 \le i \le 5. \tag{14.43}$$

The required keypoint phase velocities are given by the weighted harmonic mean of the linear interpolant slopes, with adjustments to ensure monotonicity. This can be expressed for $0 \le i \le 4$ (i.e. for the keypoints **ABNCD**) as

$$\dot{p}_{i+1} = \begin{cases} \frac{\bar{v}_i \bar{v}_{i+1}}{k_1 \bar{v}_i + k_2 \bar{v}_{i+1}} & \text{if } \bar{v}_i \bar{v}_{i+1} > 0, \\ 0 & \text{otherwise,} \end{cases} \tag{14.44}$$

where the constants $k_1$ and $k_2$ are given by

$$k_1 = \frac{2h_i + h_{i+1}}{3(h_i + h_{i+1})}, \tag{14.45a}$$

$$k_2 = \frac{h_i + 2h_{i+1}}{3(h_i + h_{i+1})}. \tag{14.45b}$$

Given the calculated absolute tilt phase velocities at the support keypoints (e.g. $\dot{p}_A \equiv \dot{p}_1$), the corresponding angular velocities $\boldsymbol{\Omega}^{l,r}_{A\text{-}D}$ are calculated using the equations in Section 7.3.6.3.

The only keypoint phase angular velocity that still needs to be calculated is that at **F** (and as a function of that, the angular velocities at **E** and **G**). As the swing phase is generally a fast motion, and as the foot yaw and tilt can both be required to change significantly during this phase, a continuous second derivative constraint is applied at **F**. The resulting trivial one-dimensional matrix equation for the fitting of cubic splines with known endpoint slopes to **DFA** reduces to

$$\dot{p}_F \equiv \dot{p}_0 = \frac{h_5(3\bar{v}_0 - \dot{p}_1) + h_0(3\bar{v}_5 - \dot{p}_5)}{2(h_5 + h_0)}. \tag{14.46}$$

The angular velocities $\boldsymbol{\Omega}^{l,r}_F$ at the **F** keypoints are then calculated by once again applying the conversion equations in Section 7.3.6.3. Recall that Equation (14.46) is applied once for each the $\bar{p}_x$, $\bar{p}_y$ and $\bar{p}_z$ components of the absolute tilt phase space. An example of the final calculated angular velocity vectors $\boldsymbol{\Omega}^{l,r}_*$ for all the keypoints is shown in Figure 14.12.

### 14.2.4.11 *Leaned Keypoints*

Once all of the positions and velocities of the keypoints have been calculated, the entire motion profile is rotated relative to the torso of the robot to effectuate torso leaning. This is referred to as 'perfect leaning', as it is applied analytically and exactly, and leaves the step sizes and relative foot motion profiles untouched. The required rotation to apply to the keypoints is given by the referenced rotation

$$^{BN}_{L}R = ^{B}_{N}R\,^{N}_{L}R\,^{B}_{N}R^{T},\qquad(14.47)$$

where $^{N}_{L}R$ is the rotation matrix corresponding to the inverse of the tilt phase rotation $P_l = (p_{xl}, p_{yl})$ (see Section 14.2.3.1). The frame {L} is implicitly defined by $^{N}_{L}R$, and is referred to as the leaned ground plane L. Using $^{BN}_{L}R$ from Equation (14.47), the required leaning update equations for the keypoint positions and velocities are given by

$$\mathbf{x}_* \leftarrow \mathbf{p}^c_h + {}^{BN}_{L}R(\mathbf{x}_* - \mathbf{p}^c_h)\qquad(14.48a)$$

$$\mathbf{v}_* \leftarrow {}^{BN}_{L}R\,\mathbf{v}_*,\qquad(14.48b)$$

$$\mathbf{\Omega}_* \leftarrow {}^{BN}_{L}R\,\mathbf{\Omega}_*,\qquad(14.48c)$$

where $\mathbf{p}^c_h$, as before, is the hip centre point. The motion centre point $\mathbf{x}_{MC}$ and height adjustment normal $\hat{\mathbf{h}}$ are also affected by the lean tilt, and are updated using the equations

$$\mathbf{x}_{MC} \leftarrow \mathbf{p}^c_h + {}^{BN}_{L}R(\mathbf{x}_{MC} - \mathbf{p}^c_h),\qquad(14.49a)$$

$$\hat{\mathbf{h}} \leftarrow {}^{BN}_{L}R\,\hat{\mathbf{h}}.\qquad(14.49b)$$

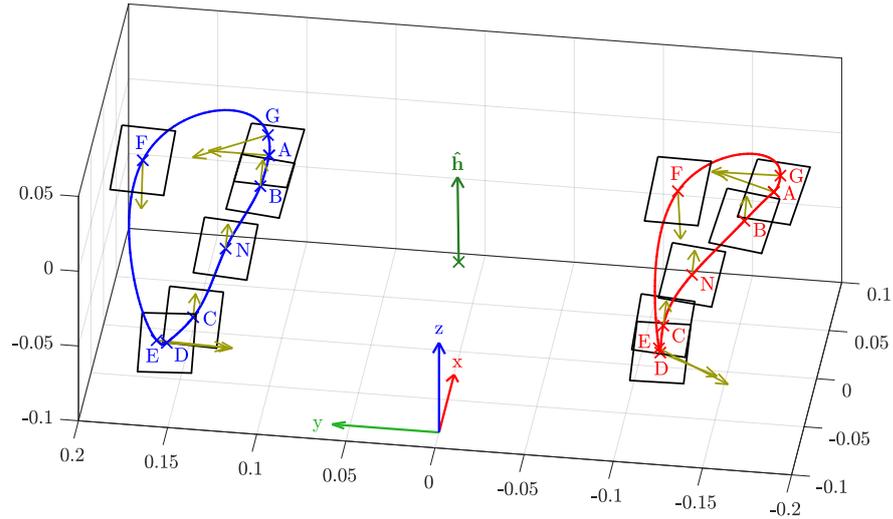

Figure 14.12: Plot of the adjusted keypoints '×', and their corresponding instantaneous angular velocities (dark yellow arrows) and foot orientations (black rectangles). The solid blue/red lines are the same connecting cubic splines as in Figure 14.10. While there are moderately large angular velocities during the quick leg swing phase ($\mathbf{D} \rightarrow \mathbf{F} \rightarrow \mathbf{A}$), one can observe during the support phase ($\mathbf{A} \rightarrow \mathbf{N} \rightarrow \mathbf{D}$) that the main rotational velocity component simply corresponds to the changes in foot yaw required for the rotational GCV component of $v_{iz} = -0.25$.

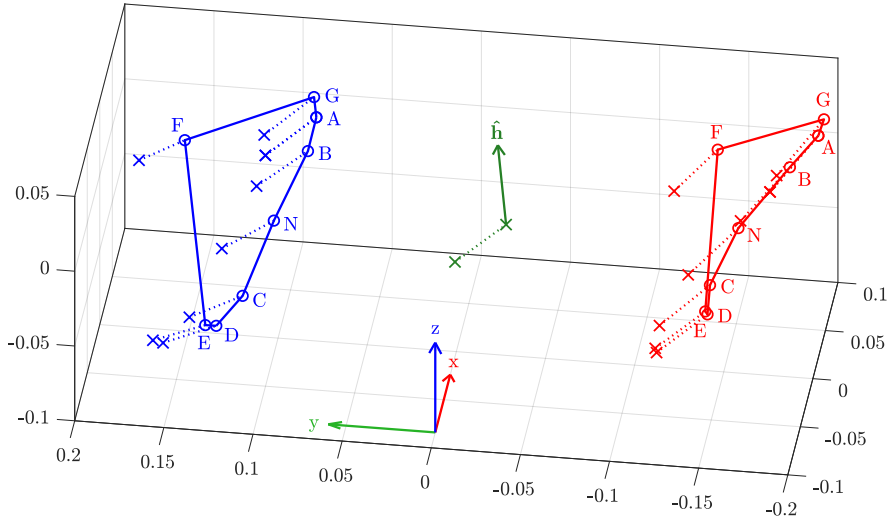

Figure 14.13: Plot of the leaned keypoints '∘', which modify the adjusted keypoints '×' (and the corresponding keypoint velocities) by a rotation of $-P_l$ about the hip centre point. In the pictured case, $P_l$ is equal to a tilt rotation of $(\bar{\gamma}_l, \alpha_l) = (1.0, 0.1)$.

The effect of the lean tilt on the keypoint trajectories is illustrated in Figure 14.13. Note that after leaning, it is assumed that the ground plane is nominally parallel to L during walking, as this is the idealised result of the application of leaning.

#### 14.2.4.12 *Shifted Keypoints*

The leaned keypoints are further adjusted as a whole to apply the hip shift $\mathbf{s} = (s_x, s_y)$. Taking into account that $\mathbf{s}$ is expressed in units of the inverse leg scale $L_i$, this leads to the update equations

$$\mathbf{x}_* \leftarrow \mathbf{x}_* + \Delta\mathbf{x}_s, \qquad (14.50a)$$

$$\mathbf{x}_{MC} \leftarrow \mathbf{x}_{MC} + \Delta\mathbf{x}_s, \qquad (14.50b)$$

where

$$\Delta\mathbf{x}_s = -{}^B_N R \, {}^N_L R (L_i s_x, \, L_i s_y, \, 0). \qquad (14.51)$$

Note that hip shifting is applied relative to the {L} frame, as the shift should always be parallel to the nominal level of the ground. The effect of hip shifting is depicted in Figure 14.14.

#### 14.2.4.13 *Finalised Keypoints*

After the application of shifting to the keypoints, the construction of the leg motion profiles—incorporating almost all of the gait generator inputs and corrective actions—is essentially complete. A final adjustment of the height of the profile is required however, to ensure that it is kinematically reasonable, and that it respects the commanded maximum hip height $H_{max}$. The hip height is given by the upwards distance normal to the leaned ground plane L, in units of the leg tip scale $L_t$, from the motion centre point $\mathbf{x}_{MC}$ to the hip centre point $\mathbf{p}_h^c$. As shown in Figure 14.15, in the final adjustment step, the height of the profile is adjusted linearly along the height adjustment normal $\hat{\mathbf{h}}$. The process whereby this is done ensures that the hip height remains at a nominal value $H_{nom}$, unless that is in breach of $H_{max}$, or causes a leg at one of the keypoints to be too extended. The scalar height adjustment $\lambda_H$ along $\hat{\mathbf{h}}$ that is required to meet the hip height considerations is mathematically given by

$$\lambda_H = \frac{(\mathbf{p}_h^c - \mathbf{x}_{MC}) \cdot \hat{\mathbf{z}}_L - L_t \min\{H_{nom}, H_{max}\}}{\hat{\mathbf{h}} \cdot \hat{\mathbf{z}}_L}, \tag{14.52}$$

where $\hat{\mathbf{z}}_L$ is the z-vector of frame {L}, i.e. the normal vector of the L plane. If $\lambda_K$ is the height adjustment along $\hat{\mathbf{h}}$ that is required to ensure that none of the keypoints have a leg pose that is too extended, the final adjustment of the keypoints can be expressed as

$$\mathbf{x}_* \leftarrow \mathbf{x}_* + \Delta\mathbf{x}_f, \tag{14.53a}$$

$$\mathbf{x}_{MC} \leftarrow \mathbf{x}_{MC} + \Delta\mathbf{x}_f, \tag{14.53b}$$

where

$$\Delta\mathbf{x}_f = \hat{\mathbf{h}} \max\{\lambda_H, \lambda_K\}. \tag{14.54}$$

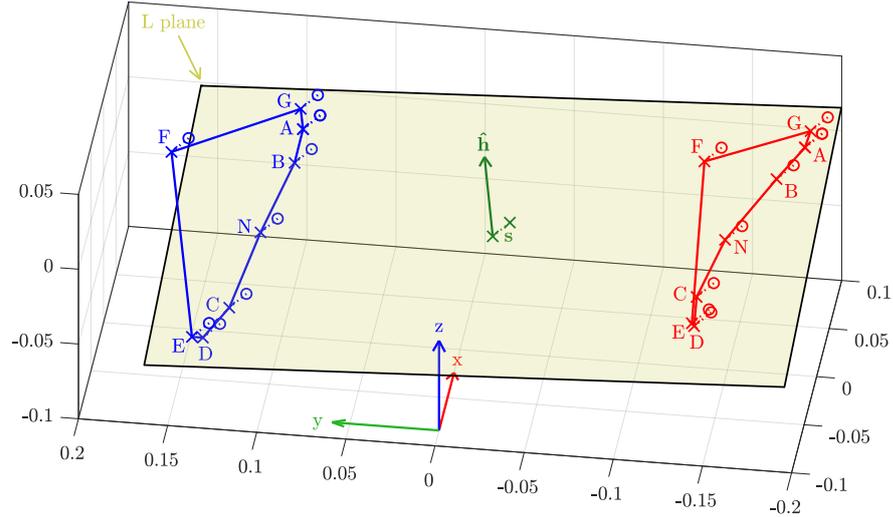

Figure 14.14: Plot of the shifted keypoints '×', which incorporate the hip shift vector $\mathbf{s} = (s_x, s_y)$ into the results of leaning. The hip shift is applied relative to the leaned ground plane L, and in this example is numerically equal to $\mathbf{s} = (0.03, -0.02)$.

An indirect but intentional consequence of the keypoint height adjustment scheme is that the CoM of the robot eventually starts to sink as the commanded GCV increases. This is because the increasingly large step sizes eventually reach the limits of the kinematic workspace at the nominal hip height, and smoothly trigger a decrease in hip height to ensure that they remain feasible. This is a positive trait, as a decreased CoM height for the same support polygon increases the margin of passive stability as per the Zero Moment Point (ZMP) stability criterion.

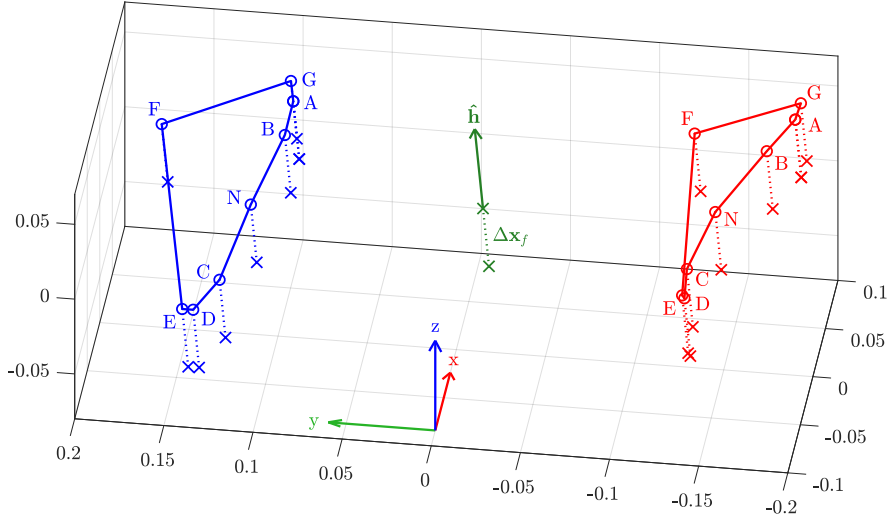

Figure 14.15: Plot of the shifted keypoints '×' and finalised keypoints '○', which incorporate a height adjustment of $\Delta\mathbf{x}_f$ to the entire motion profile. The height adjustment is parallel to $\hat{\mathbf{h}}$, and ensures that the poses have a nominal hip height of $H_{nom} = 0.94$, and that they are safely kinematically feasible ($\epsilon_l \geq 0.025$).

For the serial kinematics model (see Humanoid Kinematics), the calculation of $\lambda_K$ involves an iterative procedure. $\lambda_K$ is defined as the minimum height adjustment for which the leg retraction $\epsilon_l$ of each leg at each keypoint is not less than $\epsilon_{l,min}$. Given a keypoint, leg and current value of $\lambda$, the leg angle $\phi_{lz}$ of

$$\tilde{\mathbf{x}}_* = \mathbf{x}_* + \lambda\hat{\mathbf{h}} \tag{14.55}$$

can be calculated and used to predict a value of $\lambda$ for which the leg retraction of $\tilde{\mathbf{x}}_*$ is exactly $\epsilon_{l,min}$. Recalculating $\tilde{\mathbf{x}}_*$ with this new value of $\lambda$ allows the procedure to be repeated iteratively until (for each leg and keypoint) mutually compatible values of $\lambda$, $\phi_{lz}$ and $\tilde{\mathbf{x}}_*$ result. The final chosen value of $\lambda_K$ is the maximum of all 16 calculated $\lambda$ parameters. This ensures that *every* keypoint has a leg retraction of at least $\epsilon_{l,min}$. The convergence of the iterative procedure is very quick—the changes in the $\lambda$ values are observed to decay at ≈2.7 orders of magnitude per step, and 3 to 4 iterations almost always suffice.

### 14.2.4.14   *Final Leg Poses*

The final height-adjusted keypoints, given by the leg tip positions $\mathbf{x}_{A\text{-}G}^{l,r}$ and foot orientations ($\psi_{A\text{-}G}^{l,r}$, $\tilde{\gamma}_{A\text{-}G}^{l,r}$, $\alpha_{A\text{-}G}^{l,r}$), are converted from the leg tip space to the abstract pose space, for interpolation and final evaluation of the required instantaneous leg poses. Cubic splines (see

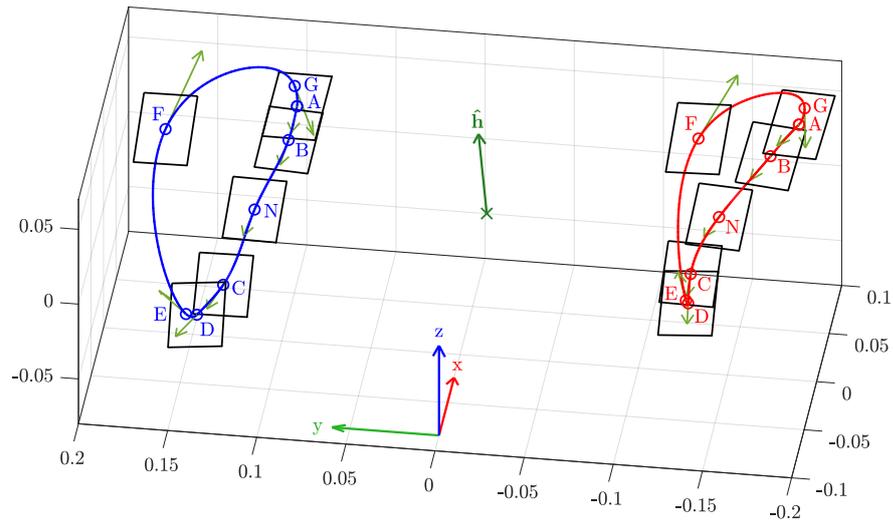

Figure 14.16: Plot of the final leg tip space trajectories of the feet, showing in particular the finalised keypoints '∘', the intermediate keypoint velocities (green arrows), and the keypoint foot orientations (black rectangles). Note that the trajectories of the feet are defined by cubic splines in the abstract space, and have been converted to the leg tip space for visualisation purposes only.

Appendix A.1.5.3) are used to uniquely connect the keypoints within the abstract space, based on the known abstract keypoint velocities, obtained by conversion of $\mathbf{v}_{A\text{-}G}^{l,r}$ and $\boldsymbol{\Omega}_{A\text{-}G}^{l,r}$ using the required Jacobians (see Section 9.2.5). The input limb phase $\nu_i$ for each leg[13] is then used to sample the resulting cubic spline leg trajectories. Asymmetrical tuneable lateral biases are incorporated into the two resulting leg poses, to allow asymmetries in the robot to be corrected, and to achieve regular symmetric walking with the gait. The final abstract leg poses (see Figure 14.17) are then obtained by soft coercion (see Appendix A.1.2.3) of each of the abstract space parameters to their reasonable limits. Converting to the joint space then yields the required commanded joint positions $\mathbf{q}_o$ for the robot. Figure 14.18 shows an example of the final resulting joint waveforms, and Figure 14.16 shows how these joint waveforms look when converted back to the Cartesian leg tip space.

By generating the keypoints and velocities in the leg tip space, yet connecting them in the abstract space, one gets the best of both worlds. The fact that the keypoints are adjusted and processed in a Euclidean space enables the required design goals and constraints to be analytically met, and ensures that the trajectories are approximately physically acceleration minimal. Final interpolation in the abstract space however, ensures that the resulting *joint waveforms* are also approximately acceleration minimal, and guarantees that no aberrant artefacts in the transition from the Cartesian to the abstract space can significantly impact the final joint space profile. Also, during the support phase—where the generated trajectory is most critical—the deviation between the cubic splines that are fit in the leg tip space and the abstract space is generally quite small. The abstract space was chosen for final interpolation in favour of the joint space, predominantly in consideration of the condition number of the corresponding Jacobians required for the velocity conversions.

---

13 Refer to Equation (14.10) for the relation of the limb phase to the gait phase $\mu_i$, and Equation (14.9) for how the gait phase is updated in each cycle, and made to loop through $(-\pi, \pi]$.

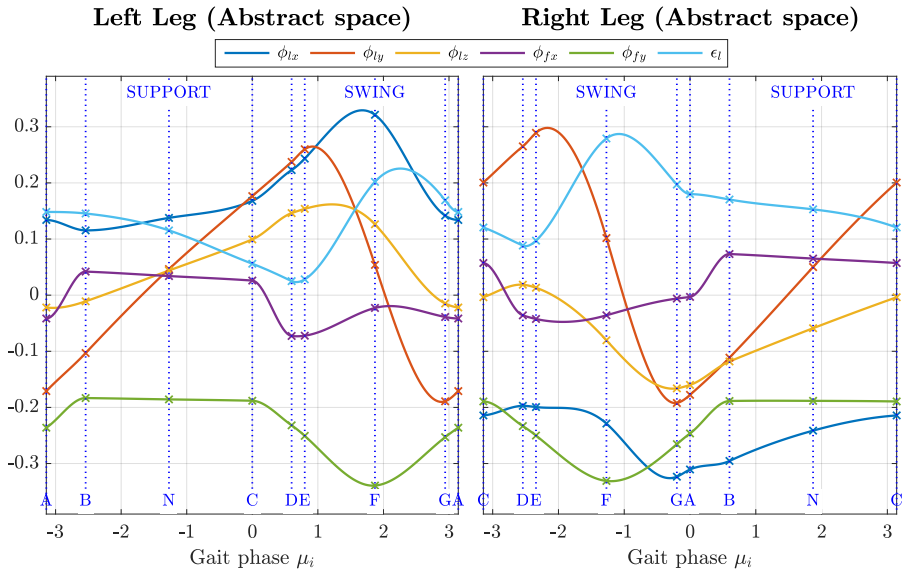

Figure 14.17: Plot of the final abstract space waveforms for the left and right legs respectively, as a function of the input gait phase $\mu_i$. Recall the connection between the gait phase $\mu_i$ and limb phase $\nu_i$ for each leg, as given by Equation (14.10). The keypoints, support and swing phases are marked in blue. Note that due to the large number of significantly activated corrective actions (for illustrative purposes), the pictured waveforms are more dynamic than usual.

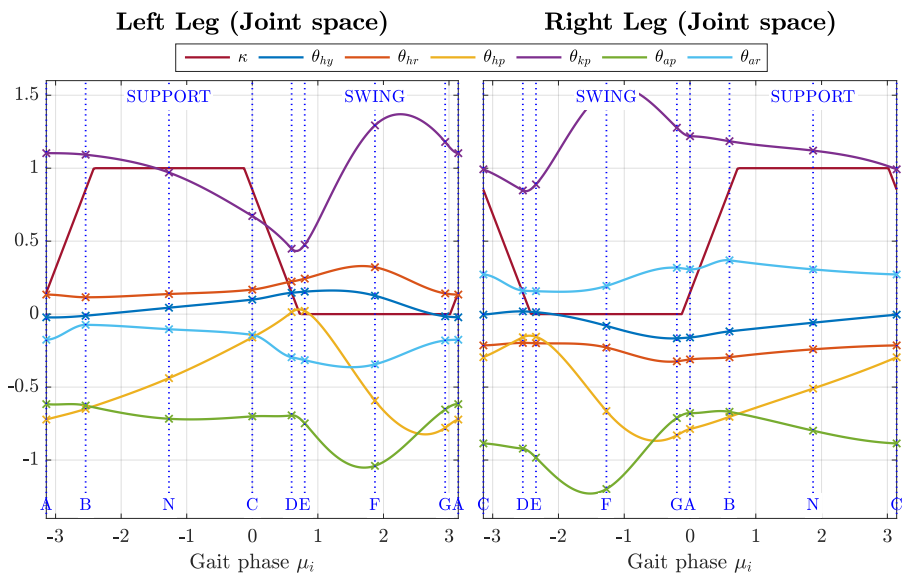

Figure 14.18: Plot of the final joint space and support coefficient waveforms for the left and right legs respectively, as a function of the input gait phase $\mu_i$. Recall the connection between the gait phase $\mu_i$ and limb phase $\nu_i$ for each leg, as given by Equation (14.10). The keypoints, support and swing phases are marked in blue. Note that due to the large number of significantly activated corrective actions (for illustrative purposes), the pictured waveforms are more dynamic than usual.

### 14.2.4.15   *Support Coefficients*

As discussed in Section 14.2.3.2, the joint efforts commanded to the robot are kept constant during the gait profile, and are equivalent to the values used for the configured neutral halt pose. The support coefficients $\kappa_l$, $\kappa_r$ need to change as a function of the limb phase however, as the weight of the robot is constantly being shifted to and from the left and right legs. As shown in Figure 14.18, a trapezoidal waveform is used that oscillates between 0 during the swing phase, and 1 during the support phase. The transitions from 0 to 1 and back are linear and centred about the double support phases. The chosen phase length of the transitions is $D + \nu_{sc}$, where $\nu_{sc}$ is a constant, and $D$ is the double support phase length as before.

Given either leg, the corresponding limb phase $\nu_i$ is first used to calculate two intermediate parameters $\nu_{rise}$ and $\nu_{fall}$, as follows:

$$\nu_{rise} = \text{wrap}\left(\nu_i - \left(\tfrac{1}{2}D - \pi\right)\right) \in (-\pi, \pi], \qquad (14.56a)$$
$$\nu_{fall} = \text{wrap}\left(\nu_i - \tfrac{1}{2}D\right) \qquad \in (-\pi, \pi]. \qquad (14.56b)$$

The required support coefficient $\kappa$ for the chosen leg is then

$$\kappa = \begin{cases} \tfrac{1}{2} + \dfrac{\nu_{rise}}{D + \nu_{sc}} & \text{if } |\nu_{rise}| \leq \tfrac{1}{2}(D + \nu_{sc}), \\ \tfrac{1}{2} - \dfrac{\nu_{fall}}{D + \nu_{sc}} & \text{if } |\nu_{fall}| \leq \tfrac{1}{2}(D + \nu_{sc}), \\ 1 & \text{else if } \nu_{rise} > 0, \\ 0 & \text{otherwise.} \end{cases} \qquad (14.57)$$

Note that by construction, at every instant we have

$$\kappa_l + \kappa_r = 1. \qquad (14.58)$$

This completes the generation of the leg trajectories, and their evaluation at the required gait phase.

### 14.2.5   **Arm Trajectory Generation**

The arm trajectories of the Keypoint Gait Generator (KGG) are constructed in such a way that they are aligned with the keypoints in Table 14.1. Instead of generating a whole cyclic trajectory and sampling it at the required gait phase however, the arm poses are calculated only for the current instantaneous gait phase $\mu_i$. This is because the arm motions produce helpful reactive moments and shifts in balance, but do not have direct balance-critical contacts with the ground like the legs do. As was done for the leg trajectory generation, the arm trajectory generation is also expressed as a function of the arm-specific limb phases $\nu_i$. We recall from page 388 that the limb phase of each arm is equivalent to the limb phase of the corresponding opposite leg. The motivation for this is that in the natural rhythm of walking, opposing arms and legs tend to move together to avoid producing unbalanced yaw moments.

The process of arm trajectory generation is broken down into several steps, namely

1. Base arm motion generation,

2. Arm pose tilting,

3. Arm pose finalisation.

These steps are addressed in order in the following subsections.

### 14.2.5.1  *Base Arm Motion Generator*

To seed the motion of the arms, a base arm motion generator is required that given the internal GCV vector $\mathbf{v}_i$, and input gait phase $\mu_i$—converted for each arm to the corresponding input limb phase $\nu_i$—calculates the required base abstract arm poses $\boldsymbol{\Phi}_a^{l,r}$. The remainder of the arm trajectory generation is completely independent of how this is done. Just like the step size generator, the base arm motion generator is like an interchangeable 'black box' that simply specifies the fundamental walking motions of the arms.

For the robots in this thesis, a base arm motion generator was implemented that generates the required two arm poses by starting with a neutral halt pose in the abstract space, and adding arm swing terms. The arm swing is applied in the sagittal direction only, as a function of the sagittal walking velocity, and has an overall signed magnitude given by a linear combination of 1 and $v_{ix}$. As indicated in Figures 14.19 and 14.21, the motion profile consists of a sinusoidal swing component in the direction of walking during the $\mathbf{D} \rightarrow \mathbf{F} \rightarrow \mathbf{A}$ swing phase, complemented by a linear return swing component during the $\mathbf{A} \rightarrow \mathbf{N} \rightarrow \mathbf{D}$ 'support' phase. As such, the arms are synchronised with the support and swing phases of the corresponding opposite legs.

### 14.2.5.2  *Tilted Arm Poses*

The abstract arm poses produced by the base arm motion generator are first adjusted to incorporate the required arm tilt $P_a = (p_{xa}, p_{ya})$ (see Section 14.2.3.1) relative to the leaned ground plane L. As the purpose of the arm tilt is to produce reaction moments and bias the balance of the robot, it is applied not to the position of the tip of the arm, but instead the position of the modelled CoM of the arm. The position of this modelled CoM relative to the arm is a function of the elbow pitch $\theta_{ep}$, or equivalently the arm retraction $\epsilon_a$, as can be seen in Equations (9.31) and (9.37).

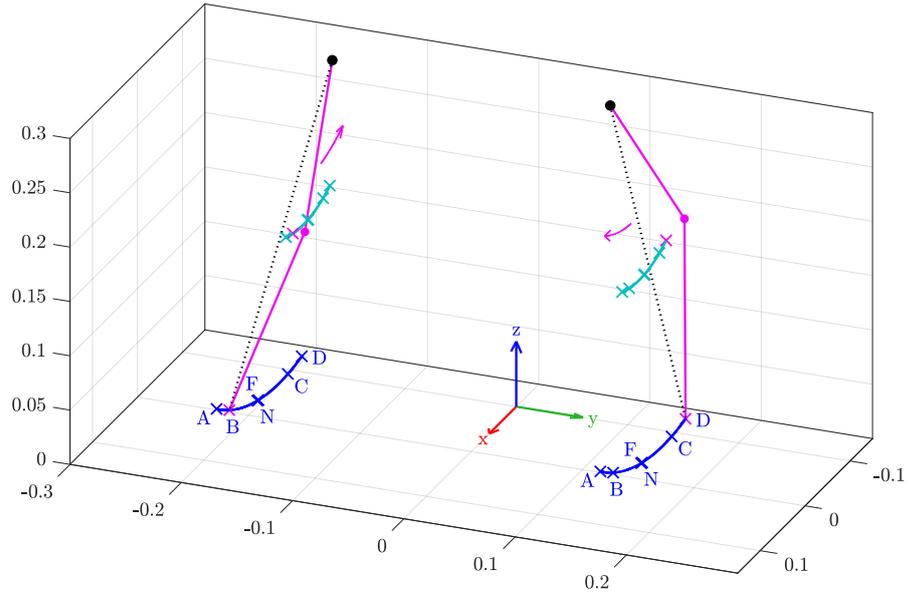

Figure 14.19: Plot of an example output of the base arm motion generator, where we recall that the keypoints of each arm correspond to the keypoints of the opposite leg. The blue curves correspond to the motions of the hand points, and the cyan curves correspond to the motions of the modelled arm CoM points. The full kinematic arm poses are shown in magenta for a gait phase of $\mu_i = -\pi + D$ (see Table 14.1), and the current directions of motion for both arms is indicated by the magenta arrows.

The magnitude $\|P_a\|$ of the arm tilt is first soft-coerced to reasonable limits before it is used, to protect against excessive and/or errant tilt inputs. The resulting 2D tilt phase rotation is then converted to the rotation matrix ${}^L_C R$, where {C} is the frame that is implicitly defined relative to {L} by exactly that rotation. For each arm, if ${}^B\mathbf{c}$ denotes the position of the modelled arm CoM, as calculated from the base abstract poses $\mathbf{\Phi}_a^{l,r}$, then the update equation corresponding to the application of arm tilt is given by

$$
{}^B\mathbf{c} \leftarrow \mathbf{p}_s + {}^{BL}_C R \left( {}^B\mathbf{c} - \mathbf{p}_s \right), \tag{14.59}
$$

where $\mathbf{p}_s^{l,r}$ is the positions of the shoulder points, and

$$
{}^{BL}_C R = {}^B_L R \, {}^L_C R \, {}^B_L R^T \tag{14.60}
$$

is the referenced rotation relative to {B} that rotates {L} onto {C}. Note that the local Cartesian arm coordinate system for each arm is taken to be centred at the corresponding shoulder point, and is aligned with the body-fixed coordinate system {B}.

The CoM position $^B\mathbf{c}$ resulting from Equation (14.59) is normalised and reformulated relative to the nominal ground plane N, to give the unit vector $^N\hat{\mathfrak{c}}$ that defines the required ray through the shoulder point on which the arm CoM should be placed using generalised arm inverse kinematics (see Section 9.2.4). The purpose of reformulating $^B\mathbf{c}$ to be expressed relative to N is to allow the tilt of the CoM ray relative to 'straight down' in the nominal ground plane (i.e. relative to $-\hat{\mathbf{z}}_N$) to be limited. For nominal walking, this effectively symmetrically limits the 'pitch' and 'roll' that the CoM of each arm can have relative to the ground. The limiting is implemented as elliptical soft coercion (see Appendix A.1.2.4), so that different limits for the pitch and roll can be set, with smooth elliptical interpolation between these two values based on the exact tilt direction. The inward y-component of $^N\hat{\mathfrak{c}}$ is also limited for each arm, to ensure that no self-collisions occur with the torso. A corresponding adjustment to the z-component (and x-component if required for more extreme tilts), is made to ensure that $^N\hat{\mathfrak{c}}$ remains a unit vector, and corresponds to the best possible approximation of applying the full required arm tilt $P_a$ to the arm.

The final resulting $^N\hat{\mathfrak{c}}$ vector is reexpressed relative to the local Cartesian arm coordinate system for the purpose of further kinematic calculations. The generalised arm inverse kinematics algorithm from Section 9.2.4 is then applied to generate a joint pose that places the arm CoM on this required ray through the shoulder point. This, when applied once to each arm, generates the required final tilted arm joint poses, as illustrated in Figure 14.20.

### 14.2.5.3 *Final Arm Poses*

The generated arm joint poses are soft-coerced (see Appendix A.1.2.3) in the joint space as a final modification to guarantee the physical safety of the arm motion profile. The output thereof is the final commanded joint positions $\mathbf{q}_o$ for the arms (see Figure 14.21), as required for the gait generator outputs. The corresponding joint efforts $\boldsymbol{\xi}$ are, like for the legs, kept constant at the values used for the configured neutral halt pose. This completes the generation of the arm trajectories, and their evaluation at the required gait phase, and thus completes the evaluation of the entire keypoint gait generator.

### 14.2.6 **Implementation**

As mentioned previously, the Keypoint Gait Generator (KGG) and its surrounding gait architecture have been implemented in both Matlab and C++, and released open source in both cases (Allgeuer, 2018b;

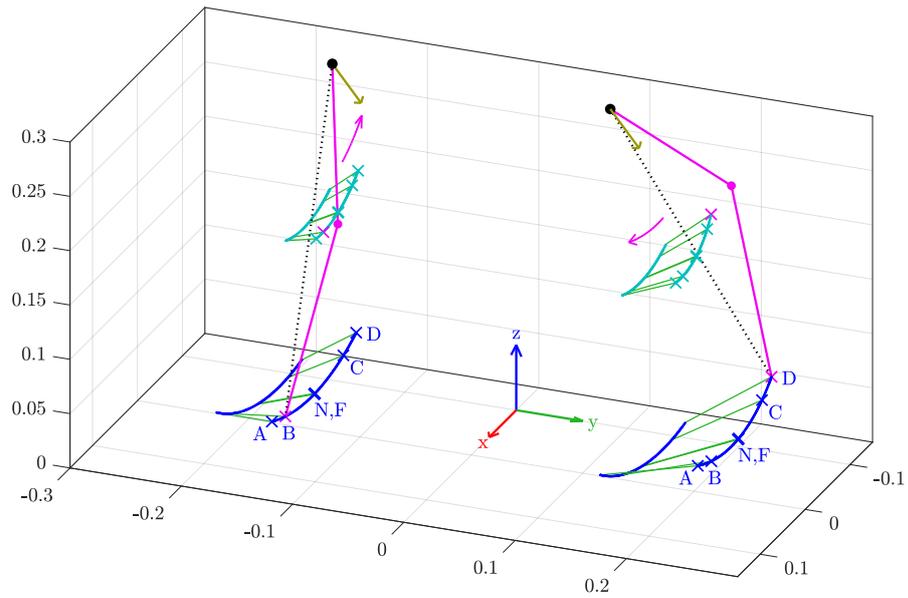

Figure 14.20: Plot of the arm trajectories after the application of an arm tilt of $(\tilde{\gamma}_a, \alpha_a) = (0.6, 0.3)$ relative to the leaned ground plane L. The arm tilt rotations are applied relative to the respective shoulder points, and the axes of rotation are indicated by the dark yellow arrows. The blue curves correspond to the motions of the hand points, and the cyan curves correspond to the motions of the modelled arm CoM points. The full kinematic arm poses are shown in magenta for a gait phase of $\mu_i = -\pi + D$ (see Table 14.1), and the current directions of motion for both arms is indicated by the magenta arrows. The green lines show the effect that the arm tilt had on the base arm motion keypoints.

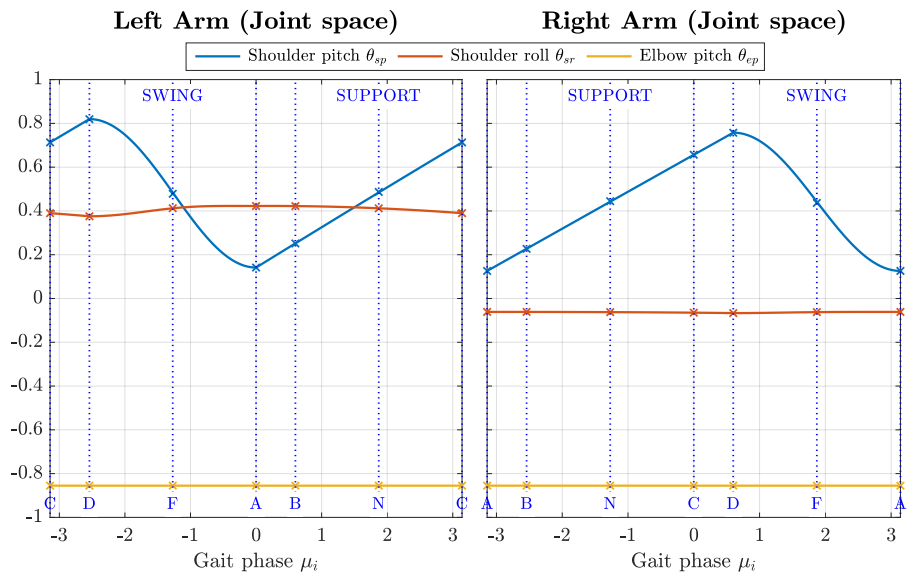

Figure 14.21: Plot of the final joint space waveforms for the left and right arms respectively, as a function of the input gait phase $\mu_i$. The keypoints, support and swing phases are marked in blue.

Team NimbRo, 2018a).[14] The Matlab release serves as a reference implementation and test bed for the algorithms involved, and includes extensive plotting of figures and visualisations, as well as highly comprehensive unit testing to ensure that every component functions exactly as it should. The C++ release serves as the practical performance implementation, that through deep-seated modularity achieves truly flexible and robot-agnostic gait code, with zero code duplication. Virtually every single component of the code is customisable and replaceable, without affecting any of the other components at all (including notably the humanoid kinematics model in use). This is possible due to the well-designed and standardised internal and external interfaces in use. The switching between available implementations of the various components and modules can also be performed freely at runtime. As such, the code is completely flexible and applicable to any robot, any kinematics, any set of higher level controllers, any step size generator, any arm base motion generator, any gait odometry estimator, and even hypothetically any gait generator.

## 14.3 DISCUSSION

There are many positive aspects of the design choices that were made as part of the KGG, the main ones of which are discussed here.

### 14.3.1 Characteristics of the Keypoint Gait Generator

A number of strategies for balanced walking were listed in Section 14.1.1. Every one of these strategies was specifically addressed by one or more of the corrective actions listed in Section 14.2.2. For example,

- The swing leg trajectory adjustment was implemented in the form of swing out tilt,

- The utilisation of ground normal forces to apply a restoring moment to the robot was addressed by the handling of the swing ground plane S, and,

- The adjustment of the foot tilt relative to the ground at the instant of foot strike was realised as part of the continuous foot tilting.

The corrective actions that were built into the KGG are highly independent, and have very clear, specific functions and resulting effects on the motion of the walking robot. This makes the process of tuning relatively simple. As a result of all this, the corrective actions constitute a solid underlying framework on which higher level controllers can be constructed to stabilise the robot.

---

14 Matlab: `https://github.com/AIS-Bonn/keypoint_gait_generator`
   C++:    `https://github.com/AIS-Bonn/humanoid_op_ros/tree/master/src/`
           `nimbro/motion/gait_engines/feed_gait`

The desired properties of the gait generator were listed in Section 14.1.3 as part of the motivation and aims. It can be seen from the details of the KGG that have been presented that each of these points have been addressed, or in some direct manner have influenced the way in which the trajectories are formed. For example, the desire to have continuous velocities and accelerations flowed directly into the constraint equations that were used for the generation of the keypoint velocities, as well as the final joining of the keypoints into abstract space trajectories.

As prescribed by the list of desired gait generator properties, the KGG was designed with the aim of minimal self-disturbances in mind, especially in terms of the desire for a smooth CoM profile that does not 'bounce' up and down during walking (which would require non-negligible oscillations in vertical momentum). The use of a nominal hip height $H_{nom}$ for example, which accounts for the effects of the various ground planes and leaning, ensures that the hip—and therefore the CoM—remains at an essentially constant height off the ground for as long as the walking velocity and kinematic workspace permit it. Another example is the leg support keypoints, which are constructed in a way that ensures that, during ideal walking (i.e. S ≡ N), all of the support keypoints and paths between them are coplanar in the N plane, leading to minimal changes in CoM height during the gait cycle. The achieved coplanarity also leads to a diminution of ground impacts, and produces zero instantaneous normal relative ground velocities at foot strike and lift-off in the ideal case. The corresponding required zero instantaneous tangential relative ground velocities during the support phase—also listed as a desired property of the gait generator in Section 14.1.3—are for example addressed during the reconciliation phase in terms of the adjustment of the **AC** and **BD** line segments, and by optimising all support keypoint velocities at once in Section 14.2.4.8.

Section 14.1.3 also listed other more specific considerations that needed to be, and are indeed taken into account by the KGG. For example, the fact that the robot torso is not always, not even nominally, sagittally upright, is handled in great depth by the definition and use of the N, S, I and J planes. Another example is the mitigation of premature toe strike by adjustment of the ankle tilt during the swing phase to account for torso tilt. This is achieved by the rotation of the foot orientations at the **F** keypoints, from the N to the S plane, as described in Section 14.2.4.6. Numerous decisions were also made in the design of the KGG to ensure that none of the corrective actions alter the step size of the robot, avoiding effects such as some of the ones listed as specific considerations in Section 14.1.3. Self-collisions were also explicitly avoided through soft coercion of the required keypoints and poses at suitable instances along the gait generation pipeline.

In summary, the chosen gait architecture (see Section 14.1.2), and specifically the KGG, are advantageous and well-constructed because the KGG:

- Is analytic, implying that the computational intensity is low, and that there are guarantees about the properties of the generated trajectories,

- Can work with any step size generator and arm base motion generator, allowing the final generated trajectories to be made to have a similar walking style and properties as any purely open-loop gait that is known to essentially work on the robot,

- Can work with essentially any robot kinematics, with appropriate adjustment of the underlying robot-specific humanoid kinematic conversions,

- Has the inherent ability to flexibly perform all corrective actions that it requires, without a higher level controller ever later intervening on a joint trajectory level,

- Utilises virtually all 'normal' ways that a robot may wish to keep its balance while walking, somewhat similar to the reactions a human may have when disturbed while walking, i.e. not just step placement and timing,

- Works in a target-oriented manner that first establishes what exact properties the keypoints and trajectories should have, and then finds a way to calculate them so that they do, as opposed to being a more 'manually intuitively constructed' approach like the CPG, and,

- Is guaranteed to be kinematically feasible and safe for both the robot and human operator(s), through the extensive use of soft coercion and limiting to ensure that all calculations are stable and remain within the allowed workspace.

### 14.3.2 Advantages of the Abstract Space

The advantages of using the abstract space to help generate the keypoint locations and final trajectories are multifold:

- The formulation of the gait generator, at least in part, in the abstract space results in smooth and simple joint trajectories that have few high frequency components or sudden changes, and are more favourable for the servo motors. This comes about due to the close relationship between the joint and abstract spaces.

- Use of the abstract space simplifies the approach of complying with the workspace boundaries, i.e. ensuring that $\epsilon_l \geq 0$ and

that all joints are in range, and provides an easy to work with guarantee of feasibility, along with a known margin thereof. It is significantly more difficult in purely the leg tip space to optimise and maximise a trajectory to 'just fit' inside the robot workspace, and it is especially difficult to make any associated guarantees, especially for varying robots and kinematics.

- Combined use of the abstract and leg tip spaces allows motion profiles to be easily constructed that are simultaneously joint-centred about a neutral halt pose, and inverse-centred in terms of the workspace and balance of the robot.

- The parameters of the abstract space are by design more intuitive and gait-related than both the joint and leg tip spaces, and allow natural concepts like hip swing and leg retraction to be used. This is especially useful for tuning, as using these, the required configuration can be accomplished with fewer and more intuitive parameters. Parameters relating to the abstract space are also more directly portable between robots of different dimensions.

- Planning motions in the abstract space gives relatively direct control over the associated joint velocities, unlike the leg tip space, but still possesses good interpolation performance.

## 14.4 EXPERIMENTAL RESULTS

The Keypoint Gait Generator (KGG) has been implemented on the igus Humanoid Open Platform and NimbRo-OP2X robots, as well as on a virtual igus Humanoid Open Platform in physical simulation. Similar to the open-loop Central Pattern Generator (CPG), the open-loop KGG allows for semi-stable walking—not for indefinite amounts of time, and not in the face of medium to strong external disturbances—but is good enough for basic locomotion. Due to the analytically computed trajectory-sensitive nature of the gait, it has more difficulties than the CPG dealing with backlash and looseness in the servos and robot, but with use of the actuator control scheme (see Section 3.2) these effects are at least for the most part mitigated.

Figure 14.22 shows sample plots of the generated joint space waveforms for the left and right limbs of an igus Humanoid Open Platform robot. The robot started from a standing position, was triggered to commence walking at time $t = 0.65$ s, and accelerated to a constant GCV of $\mathbf{v}_i = (0.7, 0.4, 0.4)$ after first waiting 1.2 s to ensure that it gets into the correct walking rhythm. During this wait time, pose blending was applied (see Step (c) in Section 11.2) to smoothly transition the robot from standing to walking. Note that pre-gait blending (see Step (a) in Section 11.2) was completed instantaneously when the gait was triggered, as the robot was already in its predefined gait halt pose.

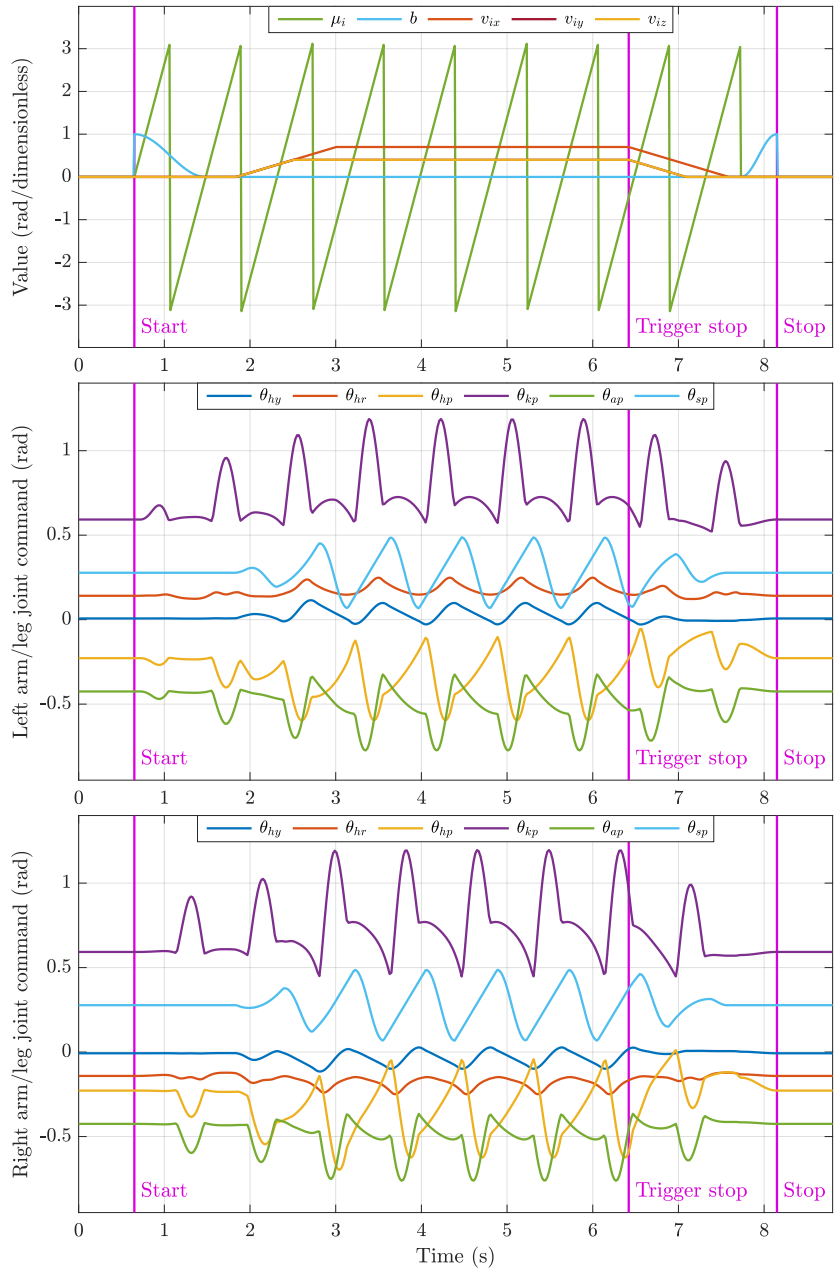

Figure 14.22: Sample output waveforms of the Keypoint Gait Generator (KGG) for an input GCV of $\mathbf{v}_g = (0.7, 0.4, 0.4)$, and a constant input instantaneous gait frequency of $f_g = 2.4\,\text{Hz}$. The top plot shows the values of the gait phase $\mu_i$ and internal GCV vector $\mathbf{v}_i$ (note that $v_{iy}$ is hidden by $v_{iz}$), as well as the dimensionless pose blending factor $b$. The KGG gait is activated at time $t = 0.65\,\text{s}$, and receives the trigger to stop walking at time $t = 6.4\,\text{s}$. After a further two steps, during which the robot slows down to an internal GCV of zero, the robot stops walking and blends back to the halt pose. Note how the waveforms for the left limbs (middle plot) are in general antiphase to the corresponding ones for the right limbs (bottom plot). The effect of the initial GCV zero time (1.2 s), in addition to the effect of the GCV slope limiting, can be seen in the top plot.

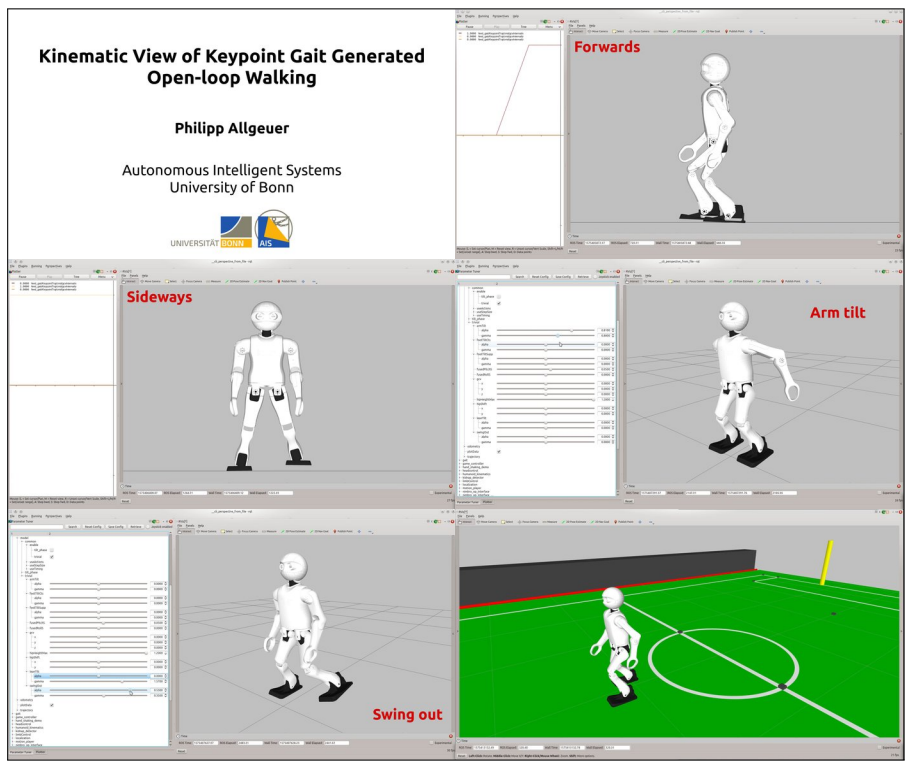

Video 14.1: Kinematic demonstration of the open-loop Keypoint Gait Generator (KGG), including the various corrective actions it is able to actuate. The stepping motion model as applied to the KGG is also demonstrated kinematically in the context of a soccer field.
https://youtu.be/XIfxTRLFbwI
*Kinematic View of Keypoint Gait Generated Open-loop Walking*

After walking for a number of steps with the desired GCV, the robot was triggered to stop walking at time $t = 6.4$ s. After decelerating to a GCV of zero and stopping walking at the next completed step, pose blending (see Step (f) in Section 11.2) was used to return the robot to its halt pose. Note that throughout the entire experiment no corrective actions were used, so the visualised waveforms correspond directly to the open-loop KGG gait. In Figure 14.22, the green curve in the top plot corresponds to the instantaneous gait phase $\mu_i$, which is propagated at a constant gait frequency of $f_g = 2.4$ Hz, and the cyan curve corresponds to the pose blending factor $b$, for which a value of 0 means that the generated gait pose should be used, while a value of 1 means that the gait halt pose should be used. For intermediate values of $b$, linear interpolation is used on a joint level.

Video 14.1 provides a detailed kinematic demonstration of the trajectories generated by the KGG, including in particular sample visualisations of the various implemented corrective actions. Forwards, sidewards and turning motions are shown in isolation before being combined with the stepping motion model (see Section 10.2.2) to achieve full 3D gait odometry. The effects of the individual corrective actions are shown by manually setting activation values live with sliders and visualising the resulting motions of the robot.

Further demonstrations of the KGG and its corrective actions are performed implicitly in the next chapter (see Section 15.3) as part of the evaluation of the tilt phase controller.

## 14.5 CONCLUSION

An analytic method for the generation of bipedal gait trajectories was presented in this chapter. The method incorporates a myriad of 3D corrective actions for the purpose of gait stabilisation into the underlying stepping motions of the robot. The so-called Keypoint Gait Generator (KGG), along with its corresponding overarching gait architecture, which was presented for the purpose of context, is a powerful building block for the construction of complex feedback gaits that require the ability to go beyond just step placement and timing. The role and effect of each corrective action has preliminarily been demonstrated, in anticipation of the tilt phase controller presented in Chapter 15, which properly drives the corrective actions. Two implementations of the KGG and the associated kinematics calculations have been released open source, and can be complemented by higher level controllers, like the tilt phase controller, that utilise the full potential of the 3D corrective actions to preserve the balance of the robot.

## TILT PHASE CONTROLLER

As discussed in detail in Section 14.1.1, many feedback strategies exist by which a robot can seek to maintain its balance while walking bipedally. In related works, the online adjustment of step size and timing is often considered, e.g. by Kryczka et al. (2015). While these are quite effective strategies if done right, numerous other forms of feedback beyond just ankle torque, like for example arm motions and swing leg trajectory adjustments, can also be employed to significantly increase the stability of the robot, especially in a broader spectrum of walking situations. For instance, step size feedback cannot help when a robot is about to tip over the outside of one of its feet, and cannot effectively correct for systematic biases in the robot. It also has little effect until the next step is actually taken, meaning that there is an inherent dead time until disturbances can be counteracted. Furthermore, changing the target step size modifies the footstep locations, and thus directly leads to the non-realisation of footstep plans. As such, step size feedback is envisioned as a valuable tool for gait stabilisation, but one that ideally only activates for large disturbances, when there really is no other option. The corrective actions presented as part of the Keypoint Gait Generator (KGG) in Chapter 14 aim to address all of these issues. The tilt phase controller presented in this chapter (see Figure 15.1) is a higher level controller that suitably drives the corrective actions of the KGG, and solves the more general problem of how to achieve balanced push-resistant walking with minimal changes to the walking intent of the robot.

In the interest of reducing the required tuning effort and making the tilt phase controller applicable to low-cost robots with cheap sensors and actuators, the use of physical models in the feedback path is avoided. Physical models usually require extensive model identification and tuning to sufficiently resemble the behaviour of a robot, and even then, cheap actuators lead to significant nonlinearities that can often cause such models to have poor results or even fail. Physical models are also frequently quite sensitive to small changes in the robot, making frequent retuning necessary. The implementation difficulty and cost of good sensors also limits the type and accuracy of sensors that can be incorporated into a humanoid robot. In order to facilitate the greatest possible portability of the developed gait between robots of different builds and proportions—a design decision that is supported by the nominally model-free nature of the gait—only the presence of a 6-axis Inertial Measurement Unit (IMU) sensor is assumed. Apart from that, no additional sensors, joint positions, robot masses or inertias are assumed at all for the tilt phase controller. The only further 'assumptions' that are made are trivial, like for example that tilting of the support foot in one direction makes the robot tendentially tilt in the other.

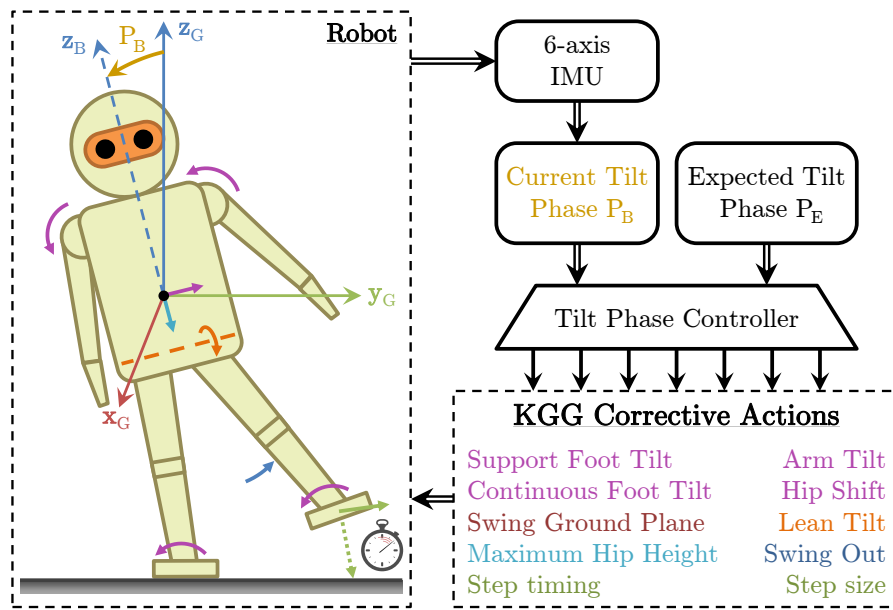

Figure 15.1: Overview of the tilt phase controller approach to walking, and how the controller interacts with the robot, the IMU state estimation, and the Keypoint Gait Generator (KGG). All ten different corrective actions of the KGG are utilised and activated by the tilt phase controller.

When compared to the direct fused angle feedback controller (see Chapter 13), the main advancements of the tilt phase controller lie in the methods of calculation of the various feedback components, which are extensions of the direct fused angle controller only for the Proportional-Derivative (PD), leaning and timing components. Many of the corrective actions are also completely new or substantially revised, with the remaining ones being extended to a full 3D treatment, so as in particular not to treat the sagittal and lateral directions independently. This is aided by the novel use of the tilt phase space as a source of truly axisymmetric orientation feedback.

Ultimately, the tilt phase controller (in combination with the KGG) seeks to demonstrate that not overly complex feedback mechanisms with very limited information of the robot suffice to produce a very stable gait, capable of rejecting significant disturbances. The presented feedback controller has been released open source in C++[1], and works in conjunction with the Humanoid Open Platform ROS Software (Team NimbRo, 2018a).

---

1  https://github.com/AIS-Bonn/humanoid_op_ros/tree/master/src/nimbro/motion/gait_engines/feed_gait/include/feed_gait/model/tilt_phase

## 15.1 GAIT ARCHITECTURE

As illustrated in Figure 14.1, and described in detail in Section 14.1.2, the overall gait architecture consists of three layers, namely

- The **actuator control scheme**, responsible for generating the low-level servo commands that are sent out to the robot hardware,

- The **keypoint gait generator**, responsible for generating the required walking motions in the form of joint angle, joint effort, and support coefficient waveforms, and,

- The **higher level controller**, i.e. the tilt phase controller in this case, responsible for calculating the required activations (see Section 14.2.3.1) of the KGG corrective actions based on the estimated state of the robot and input Gait Command Velocity (GCV). The implemented KGG corrective actions are listed in Figure 15.1 and Section 14.2.2, and are illustrated in detail in Figure 14.2.

### 15.1.1 Gait Command Velocity

The primary input to the tilt phase controller is given by the dimensionless GCV vector

$$\mathbf{v}_g = (v_{gx}, v_{gy}, v_{gz}) \in [-1, 1]^3, \tag{15.1}$$

referred to as the input GCV vector. Based on this, and the effects of any step size feedback calculated from the robot state, the tilt phase controller needs to calculate the required internal GCV vector

$$\mathbf{v}_i = (v_{ix}, v_{iy}, v_{iz}) \in [-1, 1]^3, \tag{15.2}$$

that the KGG should then use for the generation of the desired walking waveforms. As a first step, p-norm limiting as per Equation (11.3) is applied in order to limit the total magnitude of the internal GCV.

As $\mathbf{v}_g$ is an external input, and is thus not in any way guaranteed to be smooth or continuous (for instance, it may come directly from a joystick), care needs to be taken to apply appropriate GCV smoothing, without thereby limiting the ability of the controller to react quickly to large disturbances. The GCV smoothing approach used by the tilt phase controller is to allow each component of the controller to add a contribution to one of three intermediate output GCV vectors, namely $\mathbf{v}_{LF}$, $\mathbf{v}_{HF}$ and $\mathbf{v}_{EOS}$, where

- $\mathbf{v}_{LF}$ is the low frequency GCV, which is significantly slope-limited (yielding $\hat{\mathbf{v}}_{LF}$) to ensure that it is safe for the robot no matter what values are passed to it,

- $\mathbf{v}_{HF}$ is the high frequency GCV, which is only mildly slope-limited (yielding $\hat{\mathbf{v}}_{HF}$) to ensure that more sudden deviations in step size are possible, while still avoiding the possibility of discontinuities, and,

- $\mathbf{v}_{EOS}$ is the end-of-step GCV, which is a step size component that is linearly faded to over the course of the current KGG step.

Thus, if $\hat{\mathbf{v}}_{LF}$, $\hat{\mathbf{v}}_{HF}$ and $\hat{\mathbf{v}}_{EOS}$ are GCV vectors that start at $\mathbf{0}$ at the beginning of the gait, then in each execution cycle of the tilt phase controller, these vectors are updated using the equations

$$\hat{\mathbf{v}}_{LF} \leftarrow \hat{\mathbf{v}}_{LF} + \Delta t\, \text{coerce}\big(\tfrac{1}{\Delta t}(\mathbf{v}_{LF} - \hat{\mathbf{v}}_{LF}),\, -\mathbf{k}_{lfs},\, \mathbf{k}_{lfs}\big), \tag{15.3a}$$

$$\hat{\mathbf{v}}_{HF} \leftarrow \hat{\mathbf{v}}_{HF} + \Delta t\, \text{coerce}\big(\tfrac{1}{\Delta t}(\mathbf{v}_{HF} - \hat{\mathbf{v}}_{HF}),\, -\mathbf{k}_{hfs},\, \mathbf{k}_{hfs}\big), \tag{15.3b}$$

$$\hat{\mathbf{v}}_{EOS} \leftarrow \hat{\mathbf{v}}_{EOS} + \Delta t\, \text{coerce}\big(\tfrac{1}{\hat{t}_s}(\mathbf{v}_{EOS} - \hat{\mathbf{v}}_{EOS}),\, -\mathbf{k}_{eos},\, \mathbf{k}_{eos}\big), \tag{15.3c}$$

where $\Delta t$ is the execution cycle time, $\hat{t}_s$ is the current predicted time to step, and $\mathbf{k}_*$ are suitably configured 3D vector constants specifying the required slope limits. The final GCV vector $\mathbf{v}_i$ that is then passed to the KGG in each cycle is given by

$$\mathbf{v}_i = \hat{\mathbf{v}}_{LF} + \hat{\mathbf{v}}_{HF} + \hat{\mathbf{v}}_{EOS}. \tag{15.4}$$

The low frequency GCV component $\mathbf{v}_{LF}$ works much like the manual GCV slope limiting used for the Central Pattern Generator (CPG) in Section 11.1.1, and is set to the value of $\mathbf{v}_g$ at the beginning of each execution cycle of the tilt phase controller. The $\mathbf{v}_{HF}$ component is a high frequency version of $\mathbf{v}_{LF}$ (albeit not also initialised to $\mathbf{v}_g$ of course), and the end-of-step GCV $\mathbf{v}_{EOS}$ was inspired by the way the CPG makes provisions for step size feedback (see Equation (11.13) in Section 11.1.3.2).

### 15.1.2   The Tilt Phase Space

One significant difference between the tilt phase controller and the direct fused angle controller from Chapter 13 is the full 3D treatment given to the corrective actions, made possible in part by the use of the *tilt phase space* (see Section 5.4.5) instead of *fused angles* (see Section 5.4.4). While fused angles work very well for separate treatments of the sagittal and lateral planes, the tilt phase space has advantages for concurrent treatments, in particular in relation to *magnitude axisymmetry* (see Section 6.2.4.3). This property is important in ensuring that feedback magnitudes are the same scale no matter what continuous direction the robot is tilted in. Furthermore, the tilt phase parameters share all of the critical advantages (see page 146) that fused angles have over lesser options, like in particular Euler angles, mainly due to the tight relationship between the two representations. Further advantages of the tilt phase space include that it can naturally represent and deal with tilt rotations of more than 180°, and that using it, tilt rotations can be unambiguously commutatively added (see Section 5.4.5.3). Both of these are useful features in feedback scenarios where rotation deviation feedback components are scaled by arbitrary gains and need to be combined in a well-defined and logical manner.

If $(\psi, \gamma, \alpha) \in \mathbb{T}$ are the *tilt angles* parameters of a rotation, where $\psi$ is the fused yaw, $\gamma$ is the tilt axis angle and $\alpha$ is the tilt angle (see Section 5.4.3), the corresponding *3D tilt phase* representation is given by

$$P = (p_x, p_y, p_z) = (\alpha \cos \gamma, \alpha \sin \gamma, \psi) \in \mathbb{P}^3. \tag{15.5}$$

Omitting the yaw component, the *2D tilt phase* representation of the resulting *tilt rotation component* is given by

$$P = (p_x, p_y) = (\alpha \cos \gamma, \alpha \sin \gamma) \in \mathbb{P}^2. \tag{15.6}$$

Note that the relative and absolute 2D tilt phase spaces (see Section 5.4.5.2) are identical for pure tilt rotations ($\psi = 0$), so for the majority of the rotation-based corrective actions, just the relative tilt phase notation is used (i.e. no tildes). For the continuous foot tilt corrective action, we recall that the interpretation of the tilt phase rotation varies continuously based on the $r_t$ parameter (see Section 14.2.4.6), so for this we equally just nominally use the relative tilt phase notation.

## 15.2 TILT PHASE CONTROLLER FORMULATION

The aim of the tilt phase controller is to calculate corrective action activation values that will keep the robot balanced and walking in the intended direction. An overview of the feedback pipeline corresponding to the tilt phase controller is shown in Figure 15.2. How the activation values are calculated specifically for each of the individual corrective actions is presented in detail in the following subsections.

### 15.2.1 Preliminaries

We recall from the Keypoint Gait Generator (KGG) described in the previous chapter that

- The nominal ground plane N (along with the corresponding nominal ground frame {N}) is a plane in body-fixed coordinates that reflects the nominal orientation of the ground relative to the robot,

- All Cartesian corrective actions are numerically expressed in dimensionless form relative to the nominal ground frame {N}, in units of either the inverse leg scale $L_i$ or the leg tip scale $L_t$, and,

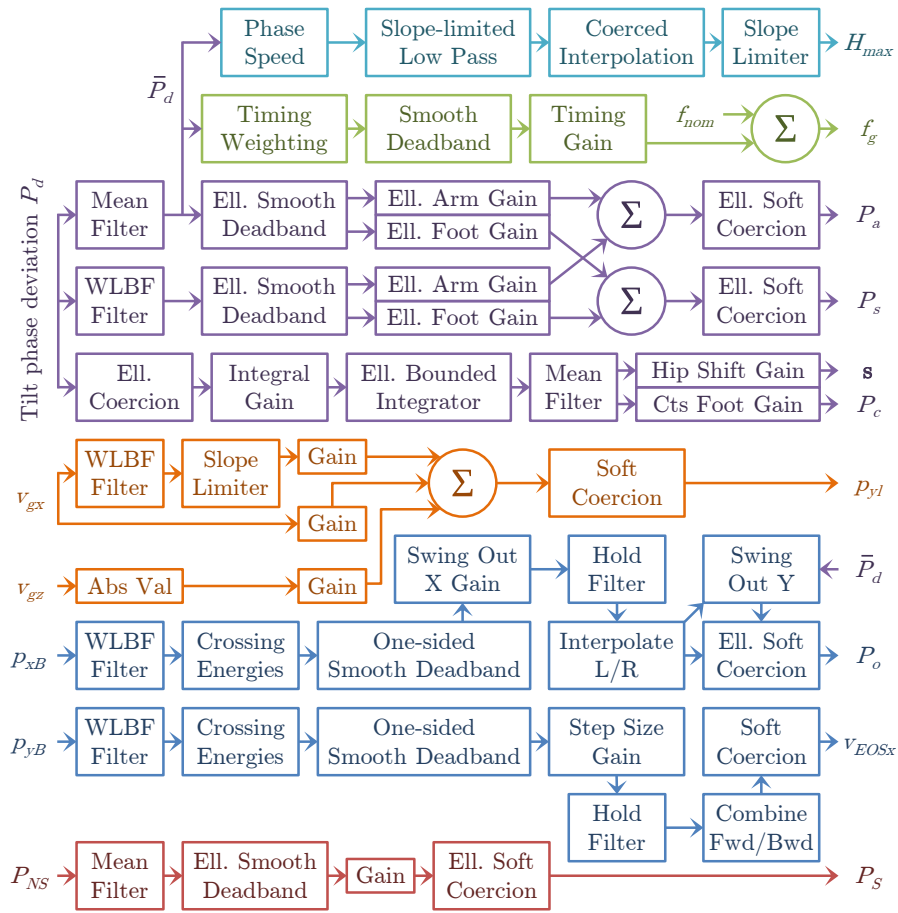

Figure 15.2: Overview of the entire tilt phase controller feedback pipeline, including the calculation of the activations of all ten KGG corrective actions. Refer to Section 15.2.2 for details on how the tilt phase deviation $P_d$ is calculated, and refer to Section 15.2.3 onwards for more details about each individual calculation pipeline.

- All rotation-based corrective actions are expressed as pure tilt rotations in the 2D tilt phase space relative to the {N} frame.

Aside from this, we also note that the tilt phase controller uses a number of recurring filters and mathematical constructs, briefly discussed as follows:

**Filters**    The mean and Weighted Line of Best Fit (WLBF) filters used in Chapter 13 have been taken and generalised to $n$ dimensions (see Appendices A.2.1.1 and A.2.4.1). The former computes the moving average of an n-dimensional vector, while the latter performs weighted time-based linear least squares regression to smooth and estimate the time derivative of n-dimensional data. The advantages of WLBF filters over alternatives for the numerical computation of derivatives are discussed in depth in Section 13.3.2.

**Coerced interpolation**  Standard linear interpolation can lead to extrapolation outside of the interval domain. Coerced interpolation limits the input variable to ensure that the output cannot be outside the range of the two data points (see Appendix A.1.4.1).

**Soft coercion**  The soft coercion function that was used in Chapter 13 has been taken and extended ellipsoidally to $n$ dimensions (see Appendix A.1.2.4). Given an input vector $\mathbf{x}$, the principal semi-axis lengths $\mathbf{a}$ of the required limiting ellipsoid $\mathcal{E}$, and a scalar soft coercion buffer $b$, scalar soft coercion is applied to the magnitude of $\mathbf{x}$ based on $b$ and the radius of $\mathcal{E}$ in that direction. This method of higher dimensional soft coercion is significantly better than applying soft limits along each axis independently, as the latter would result in unexpectedly large radial limits on the diagonals between the principal axes.

**Smooth deadband**  The smooth deadband from Chapter 13 has been taken and extended ellipsoidally to $n$ dimensions (see Appendix A.1.3.4). Given an input vector $\mathbf{x}$ and the principal semi-axis lengths $\mathbf{a}$ of the deadband ellipsoid $\mathcal{E}$, scalar smooth deadband is applied radially along $\mathbf{x}$ with a deadband radius corresponding to the radius of $\mathcal{E}$ in that direction.

### 15.2.2 Deviation Tilt

In the tilt phase controller, most of the calculated corrective action activation values depend directly on how the robot is currently tilted relative to what is expected for the current gait phase $\mu_i$. As such, the first step in the feedback pipeline is to retrieve the current trunk orientation from the attitude estimation (see Section 10.1), and express it as a 2D tilt phase orientation

$$P_B = (p_{xB},\ p_{yB}). \tag{15.7}$$

As only the tilt rotation component (i.e. no heading component) of the trunk orientation is required to construct $P_B$, a single 6-axis IMU suffices for the estimation of $P_B$ (see Section 10.1.6.4).

The two estimated tilt phase parameters from Equation (15.7), $p_{xB}$ and $p_{yB}$, follow an expected periodic pattern as a function of the gait phase during ideal undisturbed walking of the robot. We model this periodic pattern as a function

$$f_{exp} : (-\pi, \pi] \to \mathbb{P}^2,\ \mu_i \mapsto P_E, \tag{15.8}$$

where

$$P_E = (p_{xE},\ p_{yE}) \tag{15.9}$$

is the so-called *expected* 2D tilt phase orientation. For the purposes of the tilt phase controller, a sinusoidal function with an offset is used to model each the expected phase roll $p_{xE}$ and phase pitch $p_{yE}$, as this was seen to sufficiently generalise and describe the observed behaviour of the robot. As such, we define

$$p_{xE} = k_{eox} + k_{emx} \sin(\mu_i - k_{epx}), \tag{15.10a}$$
$$p_{yE} = k_{eoy} + k_{emy} \sin(\mu_i - k_{epy}), \tag{15.10b}$$

where $k_*$ are constants that are tuned based on walking experiments to make $P_E$ closely track the observed $P_B$ data (for as long as the robot is walking undisturbed).

It can be seen by definition that the deviation between $P_B$ and $P_E$ is a measure of the magnitude and direction that the orientation of the trunk is currently disturbed by. We therefore construct a trunk orientation error feedback term that quantifies the unique 3D rotation $q_d$ that rotates $P_E$ onto $P_B$, up to a small possible deviation in fused yaw $\psi_d$. What makes $q_d$ unique is that we constrain it to be a pure tilt rotation relative to the nominal ground frame {N}, as all of the KGG corrective actions that will use $q_d$ as a source of feedback are defined to act relative to the N plane (see Section 15.2.1). Mathematically, if $q_B$ and $q_E$ are the quaternions corresponding to the pure tilt rotations $P_B$ and $P_E$ respectively, then we can express the quaternion $q_d$ as

$$\begin{aligned} q_d \equiv {}^{NE}_B q &= {}^N_B q \, {}^E_B q \, {}^N_B q^* \\ &= {}^N_B q \left( q_E^* \, q_z(\psi_d) \, q_B \right) {}^N_B q^* \\ &= q_y(p_{yN}) \, q_E^* \, q_z(\psi_d) \, q_B \, q_y(-p_{yN}), \end{aligned} \tag{15.11}$$

where $q_y(\cdot)$ and $q_z(\cdot)$ correspond to pure quaternion y and z-rotations respectively, and $p_{yN}$ is the nominal phase pitch of the torso (see page 388). The value of $\psi_d$ is calculated by solving[2]

$$\Psi(q_d) = 0, \tag{15.12}$$

with $\psi_d$ as the only unknown—i.e. by solving the aforementioned constraint that $q_d$ is a pure tilt rotation. The value of $q_d$ can subsequently be evaluated using Equation (15.11), and converted to the 2D tilt phase space representation to give the deviation tilt $P_d = (p_{xd}, p_{yd})$.

To summarise, the deviation tilt $P_d$ is a quantity that indicates in which direction the robot is tilted relative to the N plane away from where it is actually normally expected to be (at that specific moment of walking). Consequently, if the negative (i.e. inverse) of $P_d$ were to be used as the activation of a rotation-based corrective action (such as the arm tilt $P_a$), it would be expected that the resulting effect on the robot would be to tilt it towards its expected orientation $P_E$, as desired. This is the idea behind the proportional feedback in Section 15.2.3, with additional measures being incorporated to address scaling and sensor noise issues.

---

2 The notation $\Psi(q_d)$ refers to the fused yaw of the quaternion $q_d$.

### 15.2.3 PD Feedback: Arm and Support Foot Tilt

The most important thing for the stability of the robot in the short term is to ensure that transient disturbances, such as pushes or steps on larger irregularities in the ground, are swiftly counteracted with little delay. 3D rotational proportional (P) and derivative (D) feedback components, activating the arm tilt and support foot tilt corrective actions, are used for this purpose. The arm tilt rotates the Centre of Mass (CoM) of the arms out in the required direction relative to the N plane, so as to bias the CoM of the entire robot and apply a reactive moment on the torso that helps mitigate the disturbance. At the same time, the support foot tilt applies smooth corrections to the tilt of each foot during its respective support phase, to push the robot back towards its expected orientation.

In order to reduce signal noise, the proportional feedback path of the tilt phase controller first mean filters the deviation tilt using a small filter order, yielding $\bar{P}_d$. This tilt phase is then elliptically smooth deadbanded to ensure that P feedback only takes effect when the robot is non-negligibly away from its expected orientation $P_E$. An elliptically directionally dependent gain is then applied to the resulting tilt phase, once independently for each the arm tilt and the support foot tilt, to get the corresponding P feedback components (see upper purple section in Figure 15.2). The gain in each case is calculated elliptically from specifications of the required gains in the sagittal and lateral directions (see Appendix A.1.1.3). Importantly, the directions of the final proportional feedback components are both unchanged from $\bar{P}_d$—all changes are purely radial.

In the derivative feedback path, a smoothed derivative of $P_d$ is first computed using a 2D WLBF filter. A WLBF filter was chosen for its many advantages, including, amongst other things, its favourable balance between robustness to high frequency noise and responsiveness to input transients (see Section 13.3.2). The computed derivative is elliptically smooth deadbanded to ensure that D feedback only takes effect if the robot torso has a non-negligible angular velocity relative to its expected orientation. Then, as for the P feedback, independent elliptically directionally dependent gains are applied to get the D feedback components for the two nominated PD corrective actions.

Once the separate P and D components have been calculated, they are combined using tilt vector addition (see Section 5.4.5.3) and elliptically soft-coerced to obtain the final required activation values $P_a$ and $P_s$ (see Section 14.2.3.1). Note that although it is not generally acceptable to just add 3D rotations, the special properties of the tilt phase space allow us to do just that in a meaningful, unambiguous, self-consistent and mathematically supported way. In fact, the tilt phase space turns tilt rotations into a well-defined vector space over $\mathbb{R}$ (see Section 7.3.3.2), explaining why the scaling and addition of tilt phases used in this chapter is actually mathematically valid and geometrically meaningful.

The tuning of the PD feedback paths is relatively straightforward, as there are only a few gains, and each gain has a clearly visible effect on the robot. The P feedback is tuned first, and then appropriated with D feedback to add damping to the system and limit oscillatory behaviour.

### 15.2.4   I Feedback: Hip Shift and Continuous Foot Tilt

The implemented PD feedback works well for rejecting the majority of short-term transient disturbances, but if there are continued regular disturbances or a systematic imbalance in the robot, the PD feedback (in combination with other corrective actions) will constantly need to act to oppose them. PD feedback can only act however, if there is a non-zero position and/or velocity error present. Thus, without integral (I) feedback, which in this case is applied in the form of the hip shift and continuous foot tilt corrective actions, the system in such a case would at best settle with a steady state deviation to normal walking, which is undesirable. The continuous foot tilt applies continuous tilt corrections to the generated orientations of the feet, while the hip shift applies an offset to the generated hip positions relative to the feet. Both are applied relative to the N plane, and bias the balance of the robot in the desired direction to overcome systematic errors in the walking of the robot. The implemented I feedback can effectively reduce the need for fine tuning of the robot, and make the gait insensitive to small changes in the hardware or walking surface that would otherwise have been noticeable in the resulting walking quality.

Starting with the deviation tilt $P_d$, standard elliptical coercion (see Appendix A.1.2.2) is first applied, the output of which is scaled by a scalar integral gain (see lower purple section in Figure 15.2). A scalar gain is used instead of a directionally dependent one, so as not to distort the 'aggregated' direction of measured instability once integration is applied. The initial coercion is useful to ensure that the integrated value is determined predominantly by small and consistent deviation tilts, rather than large and brief transients, which have little correlation to the finer balance of the robot. The coerced and scaled deviation tilt is passed to an *elliptically bounded integrator* (see Appendix A.2.3.2). This kind of integrator performs updates of 2D trapezoidal integration and elliptical soft coercion in each step. Note that the two steps are interlinked, as the output of the coercion is used as the starting point for the next integral update. Apart from providing the required integral behaviour to eliminate steady state errors, and ensuring that the integral remains conveniently bounded, this special kind of integrator also inherently combats integral windup in a more effective way than other options. The initial coercion of $P_d$ reduces the extent to which integrator windup is possible, but the elliptically bounded integrator ensures that the integral can move away from the elliptical boundary as quickly as it can approach it, and that it cannot get stuck there due to 'over-integration'. The dynamic response of the corrective actions is on a much quicker time scale than the integration, so this is the main type of windup concern in the integral feedback pipeline.

The integrated tilt phase value is passed through a final mean filter to combat ripple, before being separately scaled to yield the final corrective action activations $P_c$ and $\mathbf{s}$ (see Section 14.2.3.1). The order of the mean filter[3] is chosen to correspond exactly to the duration of an even number of steps at the nominal gait frequency. Due to the periodicity and general regularity of the gait, this leads to almost perfect cancellation of ripple. This would not be achievable with an Infinite Impulse Response (IIR) low-pass filter, which would also have the downside of not as efficiently 'forgetting', i.e. diminishing the influence of, old data.

During tuning, it is attempted to keep at least one of the elliptical integral bounds at 1. This form of normalisation makes the tuning of the integral and corrective action gains relatively simple and intuitive, as the former gain then inversely relates to the parameters of the initial elliptical coercion, and the latter gain then corresponds to the desired maximum magnitude of each respective corrective action.

### 15.2.5   Leaning

The lean tilt corrective action could be activated based on the integral feedback path, but this would promote suboptimal tilted walking postures of the robot, in part because the lean tilt directly changes the measured orientation of the trunk without this necessarily ameliorating the overall balance of the robot. Leaning driven by the PD feedback would also be possible, but although maybe not immediately intuitively obvious, neither attempting to lean forwards nor backwards is particularly useful for dissipating energy when the robot is disturbed and for example falling dynamically forwards. Pure hip rotations are only useful if they are performed quite significantly, early enough so as to precede tipping, and in specifically controlled scenarios, e.g. clean push disturbances, purely in the sagittal direction, with the robot not walking or stopping walking immediately after the disturbance, and so on. In most other situations, reactive leaning has a negative impact on walking robustness. As such, only feed-forward leaning components based on the Gait Command Velocity (GCV) are implemented. These seek to avoid disturbances due to changes in walking velocity before they even occur. The gait acceleration is first estimated using a WLBF filter followed by a slope limiter (see orange section in

---

3 That is, the number of data points that are used in the evaluation of the mean filter.

Figure 15.2). A linear combination of the sagittal velocity $v_{gx}$, absolute turning velocity $|v_{gz}|$ and sagittal gait acceleration is then taken and soft-coerced to give $p_{yl}$ (see Section 14.2.3.1).[4] Feed-forward sagittal leaning in particular helps during strong turns, and when starting or stopping forwards walking.

### 15.2.6 Swing Out

The robot is said to be on a lateral crossing trajectory if it has enough lateral momentum to tip over the outside of its support foot. This is a difficult situation, as no simple reactive stepping or waiting strategy can prevent the fall. Acting alongside the arm tilt and support foot tilt actions, the swing out tilt was designed specifically to allow recovery from lateral crossing trajectories. When significant lateral energy is detected, the current swing leg is rotated outwards to bias the balance of the robot, and to apply a reactive moment that dissipates crossing energy.

The lateral tilt phase $p_{xB}$ is first smoothed and differentiated using a WLBF filter. The line of best fit is evaluated at the mean of the recorded data points so that the resulting estimated phase $\hat{p}_{xB}$ (smoothed) and phase velocity $\dot{\hat{p}}_{xB}$ (differentiated) are synchronised in time. The values of $\hat{p}_{xB}$ and $\dot{\hat{p}}_{xB}$ are used to calculate the left and right crossing angles $\phi_L$ and $\phi_R$ respectively, as well as the corresponding crossing velocities $\dot{\phi}_L$ and $\dot{\phi}_R$, using the equations

$$\phi_L = p_{xL} - \hat{p}_{xB}, \qquad\qquad \dot{\phi}_L = -\dot{\hat{p}}_{xB}, \qquad (15.13a)$$

$$\phi_R = \hat{p}_{xB} - p_{xR}, \qquad\qquad \dot{\phi}_R = \dot{\hat{p}}_{xB}, \qquad (15.13b)$$

where $p_{xL}$ and $p_{xR}$ are tuned constants corresponding to the values of $\hat{p}_{xB}$ at the cusp of crossing for the left and right legs respectively. Both $\phi_L$ and $\phi_R$ are normally negative during normal walking, but become positive in the case of crossing over the respective leg, and more positive as the robot then subsequently falls to the ground.

We model the behaviour of the lateral tilt phase $\hat{p}_{xB}$ as approximately following the nonlinear pendulum model

$$\ddot{\phi}_X = C^2 \sin \phi_X, \qquad (15.14)$$

where $X = L, R$, and $C$ is the pendulum constant. This leads to the core result, and thereby assumption, that the so-called orbital energy

$$E_X(\phi_X, \dot{\phi}_X) = \tfrac{1}{C^2}\dot{\phi}_X^2 + 2(\cos \phi_X - 1) \qquad (15.15)$$

---

4 Actually, to avoid possible discontinuities with $\mathbf{v}_g$ (as it is an external input), $\hat{\mathbf{v}}_{LF}$ is used as the basis of leaning feedback instead (see Section 15.1.1). For all intents and purposes, $\hat{\mathbf{v}}_{LF}$ is essentially just a slope-limited version of $\mathbf{v}_g$ though.

remains constant over any undisturbed trajectory.[5] Note that the orbital energy has intentionally been divided by $\frac{1}{2}C^2$ from its usual form so as to make it dimensionless. This is of significant benefit later on when it comes to the tuning of the swing out feedback mechanism.

One can observe from Equation (15.15) that $E_X(\phi_X, \dot{\phi}_X)$ is the sum of a kinetic energy component (based on $\dot{\phi}_X$) and a potential energy component (based on $\phi_X$). Depending on the signs of $\dot{\phi}_X$ and $\phi_X$, these energy components either help or hinder crossing. As such, we introduce the notion of the crossing energy $CE_X(\phi_X, \dot{\phi}_X)$, which incorporates this fact into the energy calculations, and constructs a measure of how much energy is present in the robot that is going into crossing. For $X = L, R$, we define

$$CE_X(\phi_X, \dot{\phi}_X) = \tfrac{1}{C^2} \dot{\phi}_X^2 \operatorname{sgn}(\dot{\phi}_X) + 2(\cos \phi_X - 1) \operatorname{sgn}(\phi_X). \quad (15.16)$$

$CE_X$ is a $\mathcal{C}^1$ function of $\phi_X$ and $\dot{\phi}_X$, is zero for lateral tilt phase trajectories that come to rest exactly on the verge of crossing, and is more positive the greater the severity of crossing.

In each execution cycle of the tilt phase controller, the crossing energies $CE_L$ and $CE_R$ are evaluated and individually passed through a one-sided smooth deadband function (see upper blue section in Figure 15.2). The result is scaled to give an initial measure of how much swing out is required in the left and right lateral directions. The one-sided deadband ensures that the swing out is zero below a minimum crossing energy of $CE_{min}$, and that it smoothly transitions to a linear relationship beyond that. A pair of hold filters (see Appendix A.2.4.3) is applied to ensure that the greatest activation over the most recent time is kept and used for each side. The filtered lateral swing out values are then linearly interpolated based on the expected support conditions of the robot (a function of the gait phase $\mu_i$). At this point, a sagittal swing out component is added that ensures that the resultant swing out is, within limits, in the direction of $\bar{P}_d$ (see Section 15.2.3). The final resulting swing out tilt $P_o$ is then elliptically soft-coerced to ensure that the swing out stays within reasonable limits.

The tuning of the swing out feedback mechanism is done by examining real crossing trajectories of the robot. The $p_{xL}$ and $p_{xR}$ values are read from the average points of inflection of the observed curves, and $C$ is chosen to give the most constant observed profiles of $E_X$ possible. A suitable value for $CE_{min}$ can be calculated by choosing a value of $\phi_X$ that is just less than zero, and calculating the crossing energy that it would correspond to if the robot were at rest at that tilt.

---

5 Compare this to Equation (12.8) for the LIPM case. Like in Equation (12.9), it is easy to prove that the nonlinear orbital energy stays constant by expanding $\frac{d}{dt}(E_X(\phi_X, \dot{\phi}_X))$.

### 15.2.7    Swing Ground Plane

While the ground nominally coincides with the N plane during walking, if disturbances are present this is no longer the case. This can cause premature or belated foot strike during leg swing, which is both destabilising and prevents the robot from taking the intended step sizes. The swing ground plane S (see Section 14.2.3.3) defines the plane that is used to adjust the stepping trajectories to avoid such issues. This is different to most implementations of virtual slope walking, e.g. Missura (2015), in that it does not just linearly slant the foot motion profile—it analytically computes a smooth trajectory that respects the S plane at foot strike, yet intentionally presses into or eases off the ground immediately after, so as to apply a restoring moment to the robot. Standard virtual slope implementations can often actually *decrease* walking robustness in non-extreme situations, as the more the robot leans forwards for instance, the higher the feet are lifted at the front, and thus the less resistance there is to falling further forwards.

The desired orientation of the S plane is first computed by finding a pure tilt rotation $P_{NS}$ relative to N that makes the N plane coincident with where the N plane would be if the robot had its expected orientation $P_E$. Using the same notation as in Section 15.2.2, the required tilt rotation is mathematically given as

$$P_{NS} = -P_q(q_{SN}),\tag{15.17}$$

where $P_q(\cdot)$ is a function that returns the 2D tilt phase representation of a quaternion and

$$q_{SN} = q_y(p_{yN})\,q_E^*\,q_B\,q_y(-p_{yN}).\tag{15.18}$$

Note that if the robot has its expected orientation, $q_B$ equals $q_E$ and $P_{NS} = (0,0)$, i.e. the identity 2D tilt phase rotation, so S ≡ N. To reduce noise and prevent swing ground plane adjustments from happening when walking is near nominal, a mean filter followed by elliptical smooth deadband is applied to $P_{NS}$ (see red section in Figure 15.2). A nominally unit gain is then applied to allow the strength of the S plane feedback to be tuned if this helps with passive stability. The resulting tilt phase is then passed through elliptical soft coercion to ensure that it always stays within limits. This yields the final required activation $P_S$ of the swing plane corrective action (see Section 14.2.3.1).

### 15.2.8    Maximum Hip Height

As a result of repeated disturbances or self-disturbances, it can occur that the robot enters a semi-persistent limit cycle of sagittal oscillations. In such situations, limiting the height of the hips above the feet can help lower the CoM, and thereby increase the passive stability of the robot, as greater rotations are then required for tipping. As such, by temporarily restricting the maximum hip height of the robot, unwanted oscillations of the robot can be dissipated.

A measure $I$ of the instability of the robot is first computed by applying a slope-limited low pass filter (see Appendix A.2.2.2) to normed speed values $s_d$ of the mean-filtered deviation tilt $\bar{P}_d$, i.e. to

$$s_d = \tfrac{1}{\Delta t} \| \Delta \bar{P}_d \|. \tag{15.19}$$

Note that only *changes* in orientation contribute to $I$, so consistent leaning in a particular direction, for instance, does not contribute to the quantified 'instability' $I$ of the robot. Note also that the slope-limited low pass filter is nominally chosen to have a relatively long settling time, and that $\Delta \bar{P}_d$ can optionally be masked to only include sagittal components, if desired. Given the quantified instability $I$, coerced interpolation is used to map this to a desired range of maximum hip heights, so that greater levels of instability correspond to smaller allowed hip heights. A final slope limiter (see Appendix A.2.4.2) ensures that all changes to the resulting $H_{max}$ activation value occur continuously, and suitably gradually.

The tuning of the maximum hip height feedback mechanism essentially reduces to the choice of a settling time for the low pass filter, usually on the order of a few seconds, and the choice of an instability range to use for interpolation. The former is tuned based on how responsive the maximum hip height is desired to be, and the latter is tuned by artificially disturbing the robot and gauging as of what measured instability hip height feedback would have been suitable.

### 15.2.9  Timing Adjustment

Timing is an important feedback mechanism for the preservation of balance. In addition to its own stabilising effect, it also allows other corrective actions like the swing out mechanism to work most effectively. The approach to timing feedback that was used as part of the direct fused angle feedback controller (see Section 13.3.4) is also used for the tilt phase controller. The only modification is that the calculated feedback is reformulated to be in terms of the lateral deviation tilt $p_{xd}$ instead of the fused roll deviation $d_\phi$. As shown in the green section in Figure 15.2, the final calculated frequency offset $f_{go}$, given by Equation (13.9), is added to the nominal frequency $f_{nom}$, and used to drive the instantaneous gait frequency activation value $f_g$, as required (see Section 14.2.3.1).

### 15.2.10  Step Size Adjustment

Just like swing out is used to combat lateral crossing trajectories, the step size adjustment corrective action is used to combat sagittal crossing trajectories. While other corrective actions like the arm tilt and support foot tilt do indeed dampen any sagittal disturbances that are experienced, sometimes this damping is not enough to stabilise the robot, and reactive steps need to be taken as a last resort of keeping balance. Often, the fact that the robot is on a sagittal crossing trajectory can be inferred from the state (i.e. sagittal energy) of the robot long before the robot actually reaches its tipping point. This means that there is frequently enough time for the robot to take a preemptive step that avoids significant tipping altogether. The quantification of the sagittal crossing energies of the robot, and conversion thereof to suitable step size adjustments, is performed using the so-called tripendulum model. This model is evaluated as a function of the estimated phase pitch $\hat{p}_{yB} \equiv \theta$ and phase pitch velocity $\dot{\hat{p}}_{yB} \equiv \dot{\theta}$ of the trunk, which are calculated by applying a WLBF filter to the raw phase pitch data $p_{yB}$, and by evaluating the resulting line of best fit at the mean of the fitted data points.[6]

When the robot is standing on one foot, the sagittal transient response of the robot has three distinct zones of behaviour. When the robot is leaning far forwards, the contact patch of the robot reduces to just the front edge of the foot, and a forwards-leaning tipping behaviour results. This is referred to as the front zone of the sagittal dynamics. Similarly, if the Centre of Pressure (CoP) moves to the back edge of the foot, a backwards-leaning tipping behaviour results that is referred to as the back zone. In between the front and back zones is a passively stable middle zone that in general returns the robot back to its nominal orientation when small disturbances are applied.

The three zones of behaviour of the sagittal dynamics of the robot are modelled using the nonlinear tripendulum model (see Figure 15.3), which can be summarised by the differential equation

$$\ddot{\theta} = \begin{cases} C_f^2 \sin(\theta - \theta_f) & \text{for } \theta \in [\theta_s, \infty), \\ -C_m^2 \sin(\theta - \theta_m) & \text{for } \theta \in (\theta_t, \theta_s), \\ C_b^2 \sin(\theta - \theta_b) & \text{for } \theta \in (-\infty, \theta_t], \end{cases} \quad (15.20)$$

where

- $\theta$ is the WLBF-filtered phase pitch as before,

- $C_f$, $C_m$ and $C_b$ are three separate pendulum constants,

- $\theta_f$, $\theta_m$ and $\theta_b$ are tuned constants that represent the centres (i.e. the equilibrium points) of the three pendulum zones, and,

- $\theta_s$ and $\theta_t$ are two constants that are calculated such that Equation (15.20) is continuous.

---

6 Recall that this is analogous to the way it was done for the swing out corrective action in Section 15.2.6.

**Tripendulum Model**

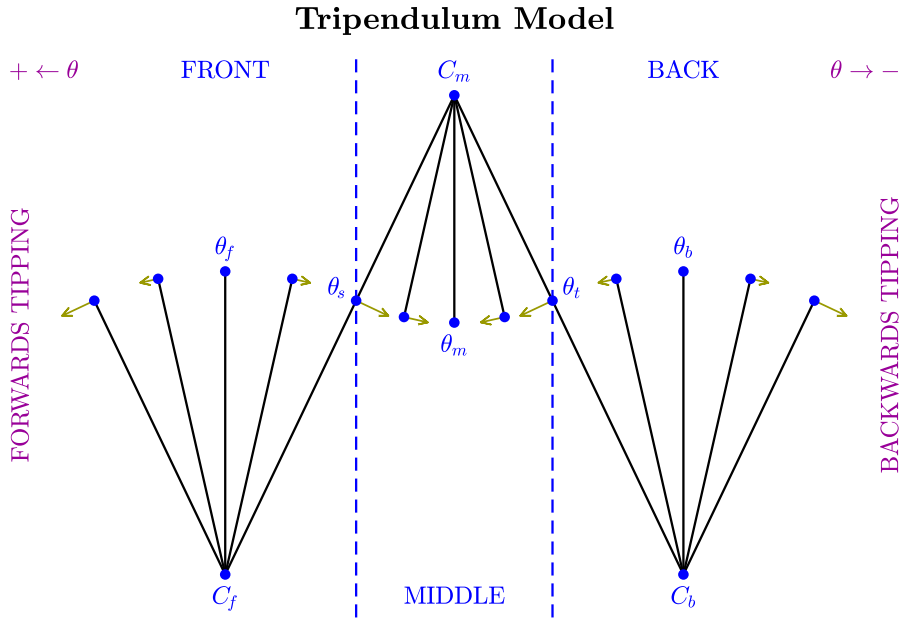

Figure 15.3: An illustration of the tripendulum model, and how it divides up the behaviour of the phase pitch angle $\theta$ of the torso into three distinct zones—the front, middle and back pendulums. Note that these pendulums are figurative—not literal—and are only seen as a mathematical description of how the phase pitch angle behaves for undisturbed trajectories of the robot. The centre angles $\theta_f$, $\theta_m$ and $\theta_b$ of the pendulums (i.e. the equilibrium points) are shown, as well as the corresponding pendulum constants $C_f$, $C_m$ and $C_b$, and the transition angles $\theta_s$ and $\theta_t$. The dark yellow arrows indicate the direction and strength of the angular acceleration $\ddot{\theta}$, as calculated by Equation (15.20).

The phase pitches $\theta_s$ and $\theta_t$ mark the two <span style="color:purple">transition points</span> between the three pendulum zones, and are calculated using

$$\theta_s = \bar{\theta}_{fm} + \text{atan2}\big((C_f^2 - C_m^2)\sin\Delta\theta_{fm},\ (C_f^2 + C_m^2)\cos\Delta\theta_{fm}\big), \quad (15.21a)$$

$$\theta_t = \bar{\theta}_{bm} + \text{atan2}\big((C_b^2 - C_m^2)\sin\Delta\theta_{bm},\ (C_b^2 + C_m^2)\cos\Delta\theta_{bm}\big), \quad (15.21b)$$

where

$$\bar{\theta}_{fm} = \tfrac{1}{2}(\theta_f + \theta_m), \qquad \bar{\theta}_{bm} = \tfrac{1}{2}(\theta_b + \theta_m), \qquad (15.22a)$$

$$\Delta\theta_{fm} = \tfrac{1}{2}(\theta_f - \theta_m), \qquad \Delta\theta_{bm} = \tfrac{1}{2}(\theta_b - \theta_m). \qquad (15.22b)$$

As was the case in Section 15.2.6 for the standard nonlinear pendulum model, instead of using Equation (15.20) to make temporal predictions of the behaviour of the robot and using this as a basis for feedback, we begin by constructing a notion of <span style="color:purple">orbital energy</span>, and use this to define a notion of <span style="color:purple">crossing energy</span> for the forwards and backwards directions. Within the front, middle and back pendulum zones, the following three respective expressions of orbital energy remain constant:

$$E_f(\theta, \dot{\theta}) = \dot{\theta}^2 - 2C_f^2(1 - \cos(\theta - \theta_f)),  \tag{15.23a}$$

$$E_m(\theta, \dot{\theta}) = \dot{\theta}^2 + 2C_m^2(1 - \cos(\theta - \theta_m)),  \tag{15.23b}$$

$$E_b(\theta, \dot{\theta}) = \dot{\theta}^2 - 2C_b^2(1 - \cos(\theta - \theta_b)).  \tag{15.23c}$$

By identifying the kinetic and potential energy components of these orbital energy expressions, and accounting for whether they help or hinder sagittal crossing, the following expressions for the so-called forwards and backwards crossing energies can be developed, similar to as was done in Section 15.2.6:

$$CE_f(\theta, \dot{\theta}) = \dot{\theta}^2 \operatorname{sgn}(\dot{\theta}) - PE_f(\theta),  \tag{15.24a}$$

$$CE_b(\theta, \dot{\theta}) = -\dot{\theta}^2 \operatorname{sgn}(\dot{\theta}) - PE_b(\theta),  \tag{15.24b}$$

where $PE_f(\theta)$ and $PE_b(\theta)$ are expressions of 'required potential energy to crossing' that are given by

$$PE_f(\theta) = \begin{cases} -2C_f^2(1 - \cos(\theta - \theta_f)) & \text{if } \theta \in [\theta_f, \infty), \\ 2C_f^2(1 - \cos(\theta - \theta_f)) & \text{if } \theta \in [\theta_s, \theta_f), \\ 2C_f^2(1 - c_{sf}) + 2C_m^2(\cos(\theta - \theta_m) - c_{sm}) & \text{if } \theta \in [\theta_t, \theta_s), \\ 2C_f^2(1 - c_{sf}) + 2C_m^2(c_{tm} - c_{sm}) \\ \qquad + 2C_b^2(c_{tb} - \cos(\theta - \theta_b)) & \text{if } \theta \in (-\infty, \theta_t), \end{cases}  \tag{15.25a}$$

$$PE_b(\theta) = \begin{cases} -2C_b^2(1 - \cos(\theta - \theta_b)) & \text{if } \theta \in (-\infty, \theta_b], \\ 2C_b^2(1 - \cos(\theta - \theta_b)) & \text{if } \theta \in (\theta_b, \theta_t], \\ 2C_b^2(1 - c_{tb}) + 2C_m^2(\cos(\theta - \theta_m) - c_{tm}) & \text{if } \theta \in (\theta_t, \theta_s], \\ 2C_b^2(1 - c_{tb}) + 2C_m^2(c_{sm} - c_{tm}) \\ \qquad + 2C_f^2(c_{sf} - \cos(\theta - \theta_f)) & \text{if } \theta \in (\theta_s, \infty), \end{cases}  \tag{15.25b}$$

where we use the shorthand $c_{xy} \equiv \cos(\theta_x - \theta_y)$, for $x, y = f, s, m, t, b$. Note that both $PE_f(\theta)$ and $PE_b(\theta)$, and thus $CE_f(\theta, \dot{\theta})$ and $CE_b(\theta, \dot{\theta})$, are $\mathcal{C}^1$ functions of $\theta$ due to the way that $\theta_s$ and $\theta_t$ are calculated in Equation (15.21). The value of the crossing energy $CE_f(\theta, \dot{\theta})$ can be interpreted as the amount of energy that the robot has in the direction of forwards falling ($\theta \to \infty$), while analogously, the value of $CE_b(\theta, \dot{\theta})$ can be interpreted as the amount of energy that the robot has in the direction of backwards falling ($\theta \to -\infty$). Both crossing energy values are respectively zero for trajectories that come to rest exactly on the verge of crossing, i.e. exactly at the front or back pendulum equilibrium points.

In each execution cycle of the tilt phase controller, the front and back crossing energies are calculated and individually passed through a one-sided smooth deadband function (see lower blue section in Figure 15.2). The result is scaled to give a pair of appropriate GCV adjustment values in the sagittal direction. These values are passed through hold filters (see Appendix A.2.4.3) to ensure a more temporally consistent activation of the step size adjustments, after which they are combined through basic addition. Final soft coercion is applied to the resulting combined sagittal GCV to ensure that the final adjustment stays within reasonable ranges. The output of the soft coercion is added to the x-component of the desired end-of-step GCV $\mathbf{v}_{EOS}$, which is then later used to calculate $\mathbf{v}_i$ as described in Section 15.1.1. It is a rare occurrence that both non-zero forwards tipping and non-zero backwards tipping GCV adjustments are produced, but if this happens, the calculated adjustments are in opposite directions, so the process of adding them effectively selects an intermediate adjustment value that balances the strengths of the desired adjustments in each direction.

Tuning of the implemented step size adjustment scheme essentially amounts to tuning of the tripendulum model. While the values of $\theta_f$ and $\theta_b$ are tuned by examining the points of inflection of real sagittal crossing trajectories of the robot, the value of $\theta_m$ is generally kept constant at the nominal phase pitch $p_{yN}$ of the robot (see page 388). The three pendulum constants $C_f$, $C_m$ and $C_b$ are chosen so that the orbital energies calculated in Equation (15.23) remain as constant as possible within the three pendulum zones. The crossing energy thresholds beyond which step size adjustments are invoked (a parameter of the one-sided deadband function) are tuned by choosing a value of $\theta$ that is close to the respective tipping point, and calculating the crossing energy that a robot at rest at that tilt would have.

One important property of the presented approach to step size adjustments is that adjustments are only made as a last resort if they are really needed. The calculated crossing energies will always be significantly negative in normal undisturbed walking situations, leading to zero being emitted by the one-sided smooth deadband functions. Only if significant disturbances are present that risk having the robot tip over sagittally does the deadband function emit non-zero values, and cause reactive steps to be taken.

## 15.3 EXPERIMENTAL RESULTS

The proposed feedback controller has been implemented in C++ in the open-source igus Humanoid Open Platform ROS software (Team NimbRo, 2018a), which also supports the NimbRo-OP2 and NimbRo-OP2X robots. The entire controller takes just 2.1 µs to execute on a single 3.5 GHz core. As such, it is expected that the implementation of this method at 100 Hz on even a modest microcontroller would be possible.

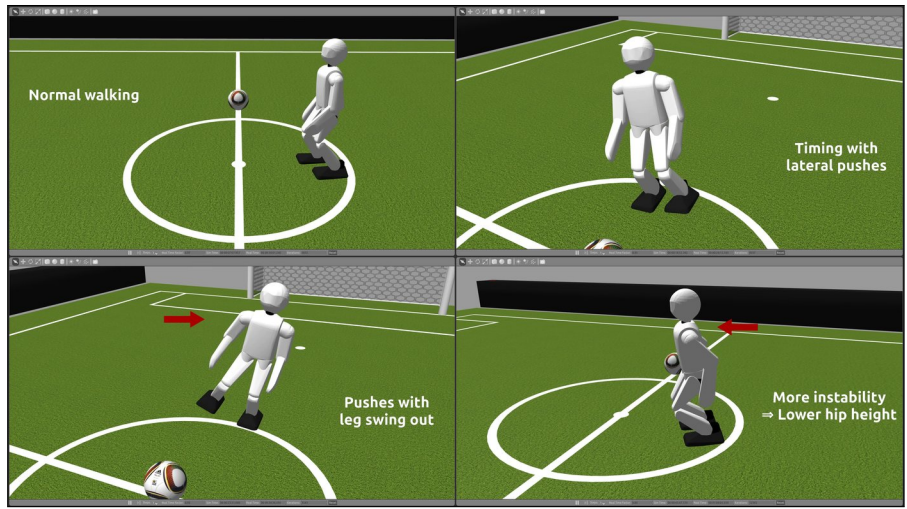

Video 15.1: A short demonstration of bipedal walking in physical simulation using the KGG and tilt phase controller. Not all corrective actions are shown. Note that due to limitations of the version of Gazebo used, uncontrollable fluctuations in the simulation log playback speed resulted in noticeable speed fluctuations in the video.
https://youtu.be/ub0GvZ7AbLc
*Short Demonstration of the Action of the Tilt Phase Controller*

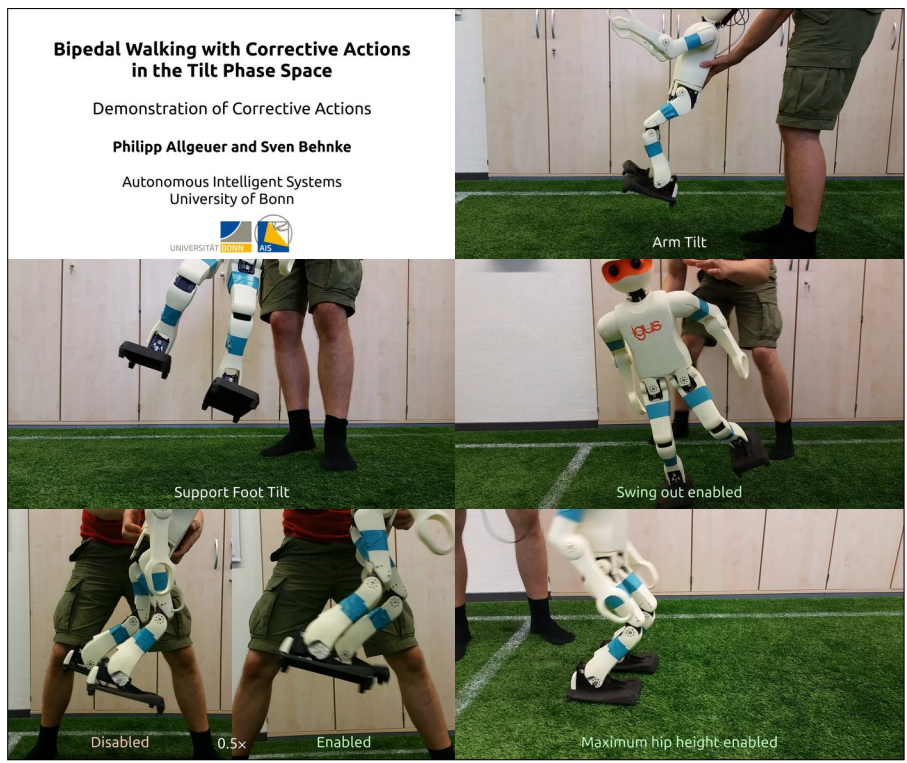

Video 15.2: Individual demonstrations of the effects of the various implemented KGG corrective actions, as activated by the tilt phase controller. Plots of the experiments are provided in Figures 15.4 to 15.6.
https://youtu.be/spFqqktZ1s4
*Demonstration of corrective actions: Bipedal walking with corrective actions in the tilt phase space*

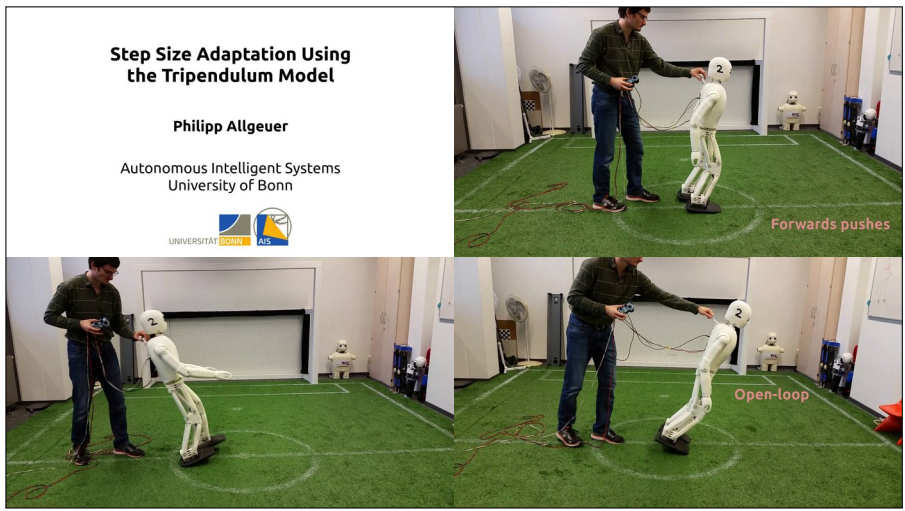

Video 15.3: Demonstration of step size adaptation using the tilt phase controller and tripendulum model. The NimbRo-OP2X is pushed repetitively in the sagittal direction and compared in performance to the same gait without step size adaptation. A plot corresponding to the second-last backwards push of the robot in the video is provided in Figure 15.7.
https://youtu.be/R9gThzV1hTQ
*Step Size Adaptation Using the Tripendulum Model*

Such portability is of great advantage in the area of low-cost robotics. Also, given the relative complexity of the gait and the diverse range of corrective actions, the number of important configuration constants has been kept rather low. The constants are in all cases expressed in a way that they are dimensionless, easy to understand and tune, and more than often just the default values can be used due to these two factors.

A short demonstration of some of the main corrective actions that set the tilt phase controller apart from other controllers is shown in Video 15.1. A more detailed demonstration and analysis of all of the implemented corrective actions is shown in Video 15.2, with step size adaptation being handled separately in Video 15.3. The improvement of the robustness of the gait when the tilt phase controller is enabled is evident across all videos. In particular, in Video 15.3 one can see that the closed-loop gait with step size adjustments enabled requires significantly stronger pushes to make the robot fall than the open-loop gait.[7] Importantly, while the robot reacts quickly with large steps to strong pushes, for small pushes no discernible change in step size occurs. This ensures that the activation of the step size adjustment remains a 'last resort' to combating disturbances, and does not cause a change in walking behaviour for regular lightly-disturbed walking.

---

7 In order to isolate the effect of the step size adaptation, the swing ground plane corrective action was not enabled in the video. It is expected that in particular forwards balance recovery would improve with this additional correction enabled.

The size of the reactive steps taken in Video 15.3 was limited to an internal sagittal GCV of 1.2. More severe reactive steps could have been allowed by increasing this limit, but it is expected that at some point with the increased resulting volatility of the robot, the overall reliability of the gait would degrade and risk damaging the robot. Significant unnecessary self-destabilisations would also be an issue (even if they do not ultimately lead to a fall), e.g. like frequently observed for the capture step controller.

The individual corrective action experiments shown in Video 15.2 have been plotted in detail in Figures 15.4 to 15.6. The experiments were performed on a real igus Humanoid Open Platform, and were specifically designed to isolate and contrast the walking performance of the robot with and without the effect of the various individual corrective actions. For almost all of the experiments, all corrective actions except for the one in focus were turned off to better illustrate the true effect of the feedback.

In Figure 15.4a, it can be observed that the tilt phase corresponds closely to the expected waveforms until a large diagonal push disturbs the robot. The PD activations quickly spike, preventing a forwards fall, and aiding the robot in returning to its expected tilt phase trajectory. Note that the plotted P and D components correspond to the outputs of the elliptical smooth deadband blocks (see Figure 15.2), prior to the application of the individual elliptical arm and foot gains. It can be seen that when the robot starts returning to upright, the sign of the derivative feedback changes quickly to dampen the system and prevent excessive overshoot.

In Figure 15.4b, the robot begins by walking on the spot. At time $t = 1.5\,\text{s}$, a sudden unknown software offset (external to the gait module) is applied to both ankles of the robot, causing the robot to pitch and roll away from its nominal expected tilt phase waveforms. In response, the I components $p_{xI}$ and $p_{yI}$ quickly integrate up the observed 3D deviations in orientation and cause the robot to learn 2D hip shifts and continuous foot tilts that perfectly negate the applied software offsets. Note that the plotted I components correspond to the output of the mean filter in Figure 15.2, i.e. the one that is used right before the hip shift and continuous foot gains in the integral feedback pathway. At approximately 10° of resultant rotation, and despite the relative severity of the applied offsets, the robot returns to completely upright nominal walking within 6.6 s.

In Figure 15.4c, the robot is made to start walking forwards, and then slow down and stop again shortly after. Without the feed-forward effects of the leaning corrective action, the robot falls backwards after taking a few forwards steps. With leaning enabled, the robot tilts slightly in the direction it is accelerating, both at the start and end of walking, and thereby mitigates a fall. In the plot, the magnitude of the sagittal leaning component $p_{yl}$ has been scaled up by a factor of 10 to

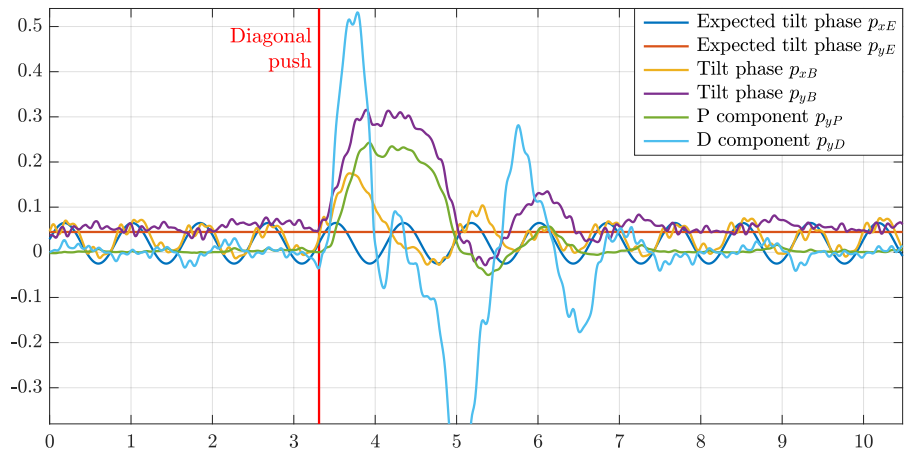

(a) The effect of the arm tilt and support foot tilt PD feedback in recovering balance after a diagonal push with significant destabilising power.

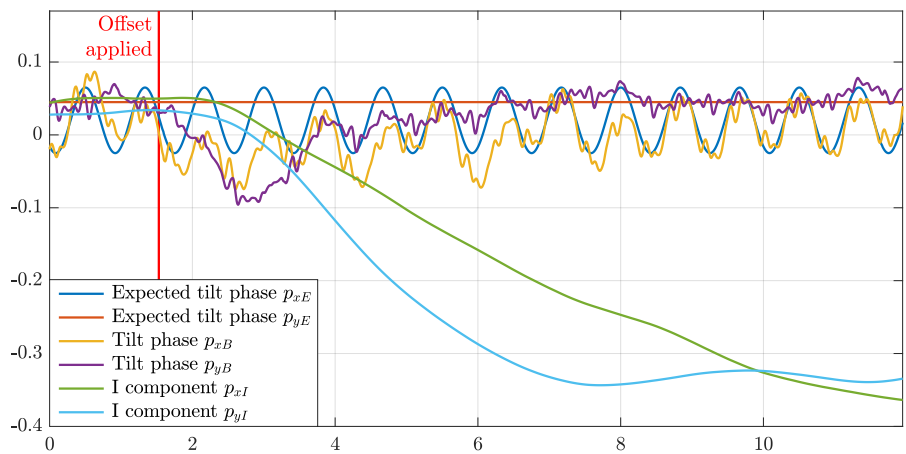

(b) The effect of the hip shift and continuous foot tilt integral feedback after a sudden unknown software offset is applied to the ankles of the robot.

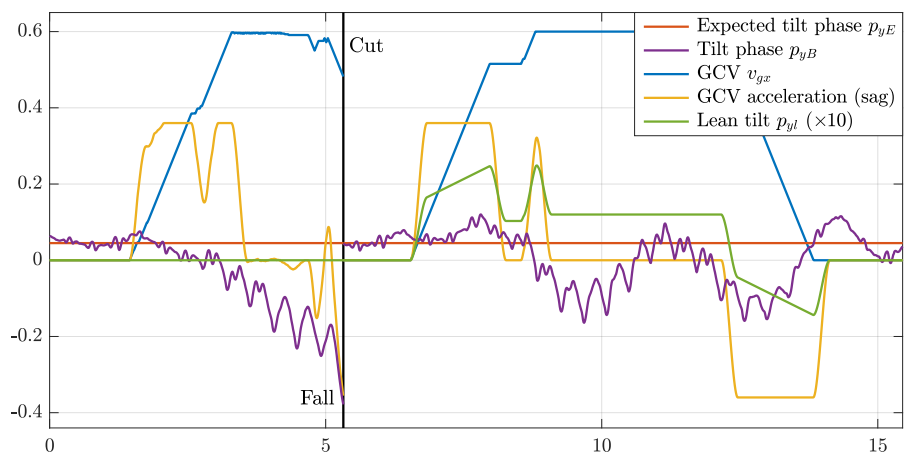

(c) The effect of leaning on sagittal walking. Without leaning (before cut), the forwards GCV acceleration makes the robot tip over backwards. With leaning (after cut), there is no fall.

Figure 15.4: Plots of the PD, I and leaning corrective actions on a real robot. Refer to Video 15.2 for the corresponding footage.

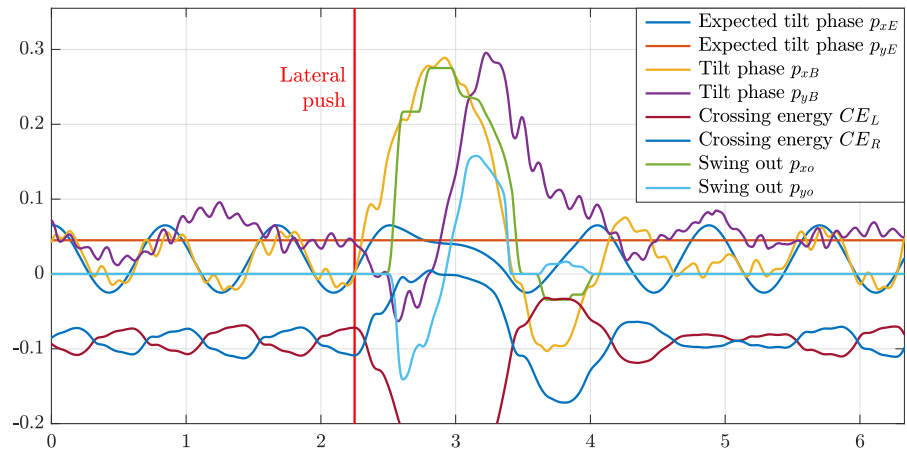

(a) The effect of the swing out tilt in recovering from a lateral push that would otherwise have led to a fall. The full 2D swing out allows the robot to remain balanced sagittally as well in the 1.2 s it takes to return laterally.

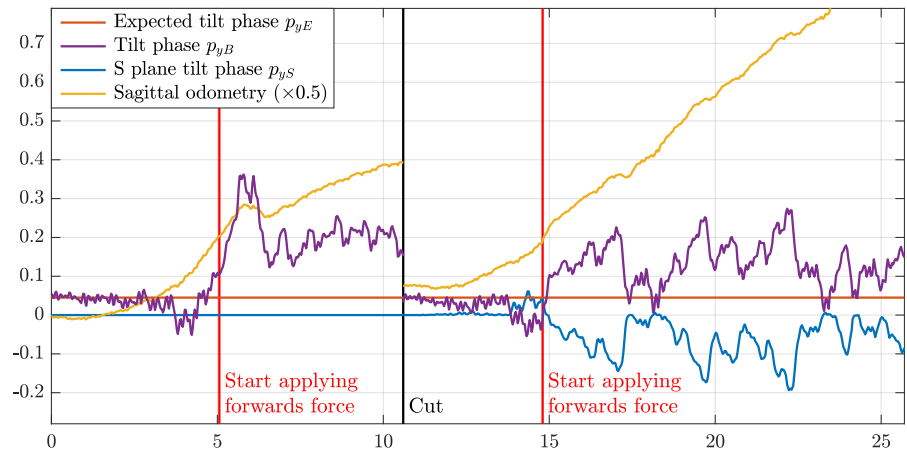

(b) Effect of the swing ground plane in reducing premature foot strike, demonstrated by applying a continuous force to the robot that induces sagittal tilt. Before cut: S plane disabled, After cut: S plane enabled.

Figure 15.5: Plots of the swing out and swing ground plane corrective actions on a real robot. Refer to Video 15.2 for the corresponding footage.

make it more visible in relation to the other plotted quantities. Recall that the plotted sagittal GCV acceleration is calculated by numerically differentiating the sagittal GCV $v_{gx}$ in a smooth manner using a WLBF filter followed by a slope limiter (see Figure 15.2).

In Figure 15.5a, a large lateral push is applied to the robot, putting it on a crossing trajectory that would normally result in a lateral fall over the outside of the support foot. The crossing energies $CE_L$ and $CE_R$, calculated using Equation (15.16), are well below zero prior to the push, but as the excess lateral energy is detected, the right foot crossing energy $CE_R$ quickly exceeds $CE_{min} = -0.044$ and causes swing out of the left leg in the lateral direction. The counterbalancing effect of the left leg dissipates some of the crossing energy, and makes

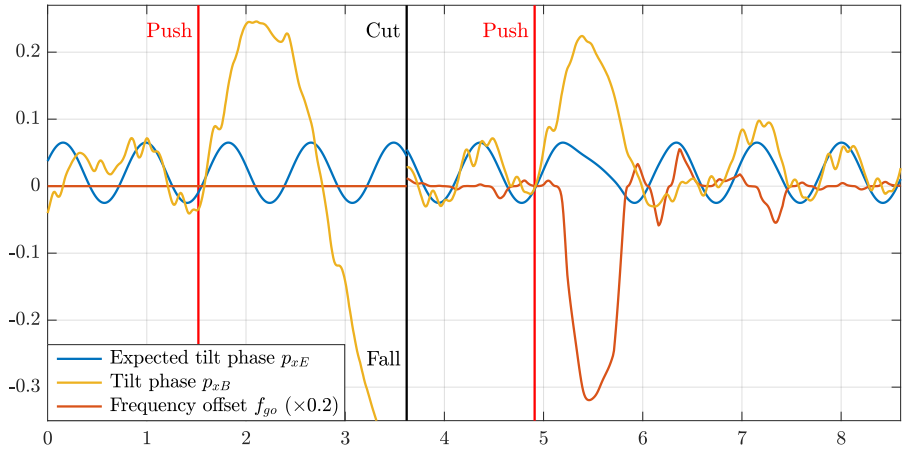

(a) The effect of timing feedback in avoiding self-disturbances that can lead to a fall. Before cut: Timing disabled, After cut: Timing enabled.

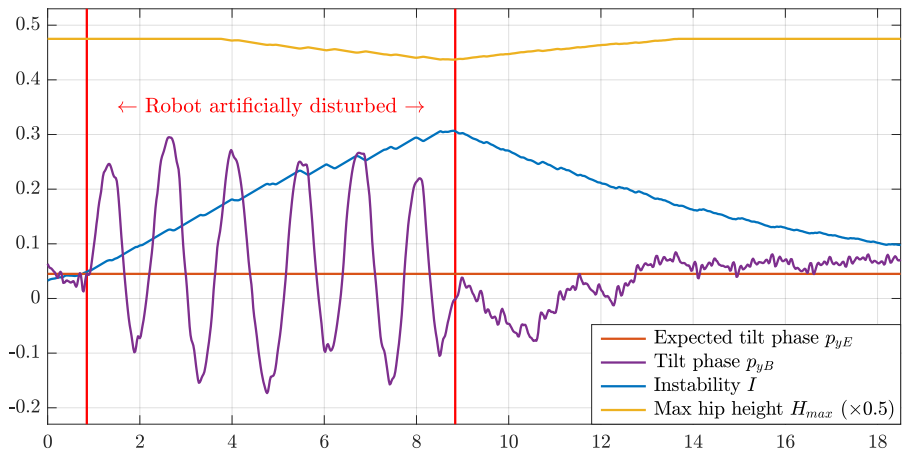

(b) Oscillations in the robot orientation, induced artificially here, cause the calculated instability to increase and limit the allowed hip height.

Figure 15.6: Plots of the timing and maximum hip height corrective actions on a real robot. Refer to Video 15.2 for the corresponding footage.

sure $CE_R$ never significantly surpasses zero, which would indicate a predicted non-returning trajectory. To maintain sagittal balance while the robot waits to return from its lateral trajectory, the sagittal swing out component $p_{yo}$ is used, first in one direction, and then the other.

In Figure 15.5b, a continuous forwards force is applied to the robot during forwards walking, causing the robot to tilt forwards. Without swing ground plane adjustment, the robot tends to walk 'into the ground', and gets more stuck than if the swing ground plane is enabled. This is evidenced by the factor-of-two difference in sagittal gait odometry (see Section 10.2.2), also plotted. The S plane tilt phase $p_{yS}$ is opposite in sign to the torso orientation $p_{yB}$ as, for example, leaning forwards corresponds to a backwards tilt of the ground plane relative to the robot.

In Figure 15.6a, lateral pushes are applied to the robot to disrupt its natural walking rhythm. Without timing adjustments, this leads to a lateral fall, as can be seen in the left half of the plot. With timing adjustments however, the robot slows down its stepping motion as soon as it detects the disturbance, and tries to place its next foot on the ground exactly when the lateral tilt is close to zero again. Importantly, during normal stable walking only negligible contributions of the timing corrective action to the gait frequency can be observed.

In Figure 15.6b, the robot is artificially disturbed over multiple seconds to demonstrate an extreme case of how oscillations lead to a higher quantified level of instability, and subsequent reduction in hip height. During real walking, this is relevant for situations where the robot gets stuck in a limit cycle of oscillations. This can (and does) occur, but is difficult to replicate intentionally. Lowering the hip height increases passive stability and changes the natural frequency of the dynamics of the tilting motions. Both factors generally lead to damping of the oscillations, as can be seen in the plot.

The second-last backwards push in Video 15.3, demonstrating the action of the step size adjustment scheme, has been plotted in detail in Figure 15.7. In the top plot, the instant is shown where the push was applied, causing a significant drop in the phase pitch of the robot. The calculated crossing energies up until that point were decisively negative, but as the robot accelerated as a result of the backwards push, the backwards crossing energy $CE_b$ quickly spiked well above zero, meaning that the robot would have had no chance of avoiding tipping over if it had not been for the reactive steps. The activation of the reactive steps, in the form of an adjustment to the sagittal GCV, is shown in the bottom plot of Figure 15.7. The current support foot at each instant is also plotted in the form of the support leg sign $\delta_t$.[8]

As the robot returned to upright, it had considerable forwards angular momentum, as evidenced by the large rate of change of phase pitch at the time. In order to avoid overshooting and potentially falling forwards, it can be seen that the forwards crossing energy $CE_f$ correctly predicted around time $t = 3.6$–$3.9$ s that this would be an issue, and caused a small forwards step to be taken soon after, as can be seen in the bottom plot. The result is that the robot had very little overshoot in the forwards direction, and was able to return to normal balanced walking very soon after the push. Note that at $t = 3.6$ s the robot was still leaning backwards, far away from upright, yet it was already aware at this time that the greater issue was falling forwards, not falling backwards.

All together, the corrective actions and feedback pipelines implemented by the tilt phase controller make a significant improvement to walking robustness, both on real hardware (see previous exper-iments), and in simulation. The dimensionless construction of the feedback pipeline also allows the controller to be easily tuned and ported between various robot platforms without significant changes. The parameters of the feedback pipeline are generally relatively insensitive to their value, so while tuning the capture step controller for example often requires careful tuning methods and 3–4 digits of precision in order to be successful, for the tilt phase controller it is quite the opposite. Video 15.4 shows the results of the very first walking tests in simulation, after most of the configuration parameters had only been 'guessed', and no previous tests had been done. The controller performs remarkably well, and it takes two minutes of determined pushing to ultimately make the robot fall.

---

8 The support leg sign is the limb sign of the current support leg. See Equation (9.17) for the definition of the limb sign.

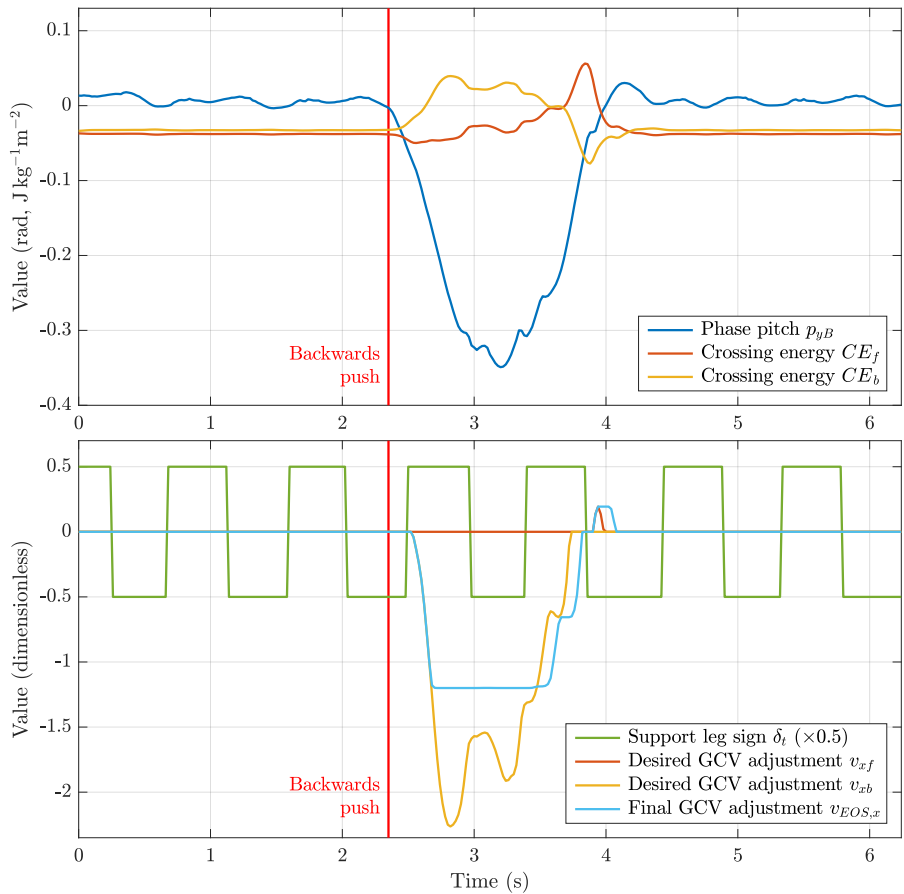

Figure 15.7: Plots corresponding to the second-last backwards push of the NimbRo-OP2X in Video 15.3, demonstrating the effect of step size adjustment in maintaining the balance of the robot. The crossing energies are plotted in units of $\mathrm{J\,kg^{-1}m^{-2}} \equiv \mathrm{s^{-2}}$, as this normalises the values for each robot. This is because the crossing energies then effectively become energy (J) per moment of inertia ($\mathrm{kg\,m^2}$).

The tilt phase controller has also been evaluated quantitatively, as opposed to qualitatively, in Gazebo simulation. Maximum forwards walking speed tests were performed over a 4 m distance with a 1 m run-up. A maximum mean velocity of 45.7 cm/s was achieved, while with the direct fused angle feedback controller (see Chapter 13), 30.5 cm/s was achieved. A video of the associated walking speed tests is provided in Video 15.5. While both gaits provide a good walking performance, it can be seen from the video how the 'smoother' nature of the KGG allows it to achieve higher overall walking speeds than the CPG before self-disturbances become a limiting factor. The minimisation of self-disturbances for the purpose of maximising walking speed was in fact one of the principal aims of the KGG (see Section 14.1.3), and exactly this goal was distilled into a detailed list of desired properties of the KGG that influenced its design at every stage (see page 383).

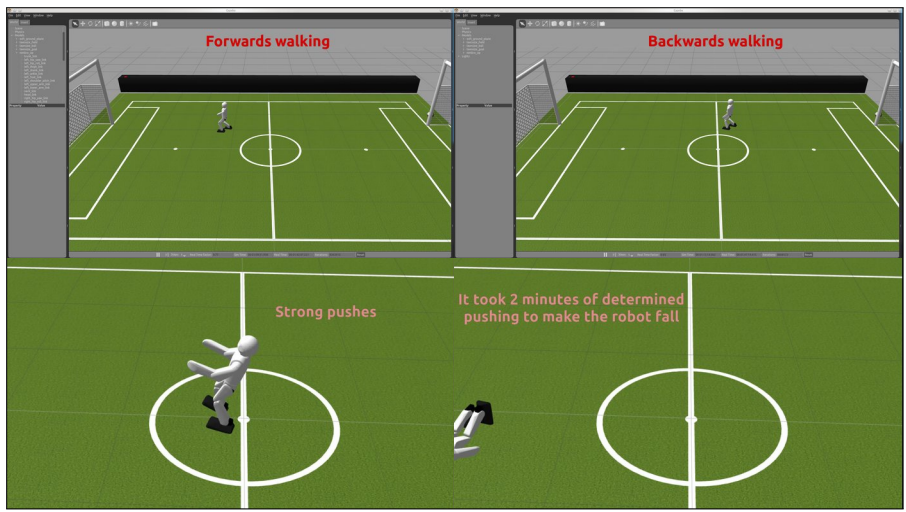

Video 15.4: Summary video of the first walking and balance tests that were
            performed (in simulation) to evaluate the effectiveness of the tilt
            phase controller. The simulated igus Humanoid Open Platform
            withstood two minutes of strong pushes before eventually falling.
            https://youtu.be/A_HQQfCRhDE
            *First Walking and Balance Test of the Tilt Phase Controller*

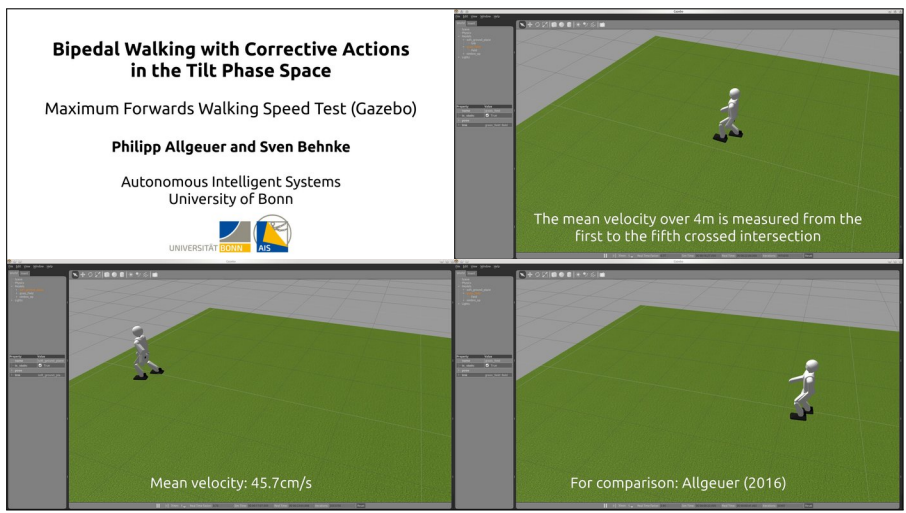

Video 15.5: Maximum forwards walking speed test using the Keypoint Gait
            Generator and tilt phase controller.
            https://youtu.be/yYOkpUZpjO4
            *Maximum forwards walking speed test: Bipedal walking with corrective
            actions in the tilt phase space*

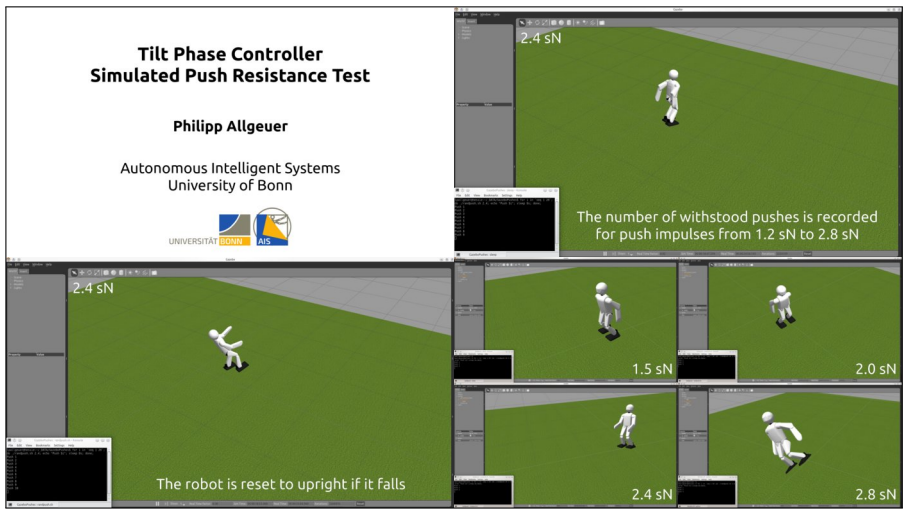

Video 15.6: Push resistance test performed on an igus Humanoid Open Platform in simulation. The robot is walking on the spot with the KGG gait, and is disturbed in random directions by impulses of various magnitudes. The effectiveness of the tilt phase controller with and without step size adaptation enabled is evaluated.
https://youtu.be/OSvHgVIYquc
*Tilt Phase Controller Simulated Push Resistance Test*

The tilt phase controller has also been put to the test in the context of push recovery. As shown in Video 15.6, sets of 20 pushes at a time were applied to an igus Humanoid Open Platform walking on the spot in simulation. Each set of 20 pushes maintained the same impulse strength, in the range from 1.2–2.8 s N, but the pushes were applied in random directions. The aim of the test was for the robot to withstand as many of the 20 pushes as possible. Table 15.1 compares the push recovery performances of the direct fused angle feedback controller, tilt phase controller without step size adaptation, and tilt phase controller with step size adaptation. This data is plotted in Figure 15.8, alongside the corresponding curve for the open-loop CPG gait. It is clear that all controllers provide a great margin of stability beyond the passive stability of open-loop walking. It can be observed however, that the tilt phase controller without step size adjustment (in order to make it a fair comparison) outperforms the direct fused angle controller evenly across all push impulses, with the number of withstood pushes at 2.6 s N, for example, being 7 compared to 3. With step size adaptation enabled, the tilt phase controller makes even further improvements over the direct fused angle controller, with the number of withstood pushes at 2.6 s N, for example, increasing to 12. Note that for reference of scale, a push impulse of 2.8 s N is expected to cause an instantaneous change in CoM velocity of about 42–56 cm/s for the igus Humanoid Open Platform, which is very severe given its CoM height of only 55 cm.

Table 15.1: Number of withstood simulated pushes (out of 20) for various controllers and controller configurations

| Impulse (s N) | 1.2 | 1.5 | 1.8 | 2.0 | 2.2 | 2.4 | 2.6 | 2.8 |
|---|---|---|---|---|---|---|---|---|
| **Direct fused angle** | 19 | 19 | 15 | 12 | 11 | 8 | 3 | 3 |
| **Tilt phase (no steps)** | 20 | 20 | 17 | 16 | 14 | 9 | 7 | 4 |
| **Tilt phase (full)** | 20 | 20 | 18 | 16 | 15 | 15 | 12 | 10 |

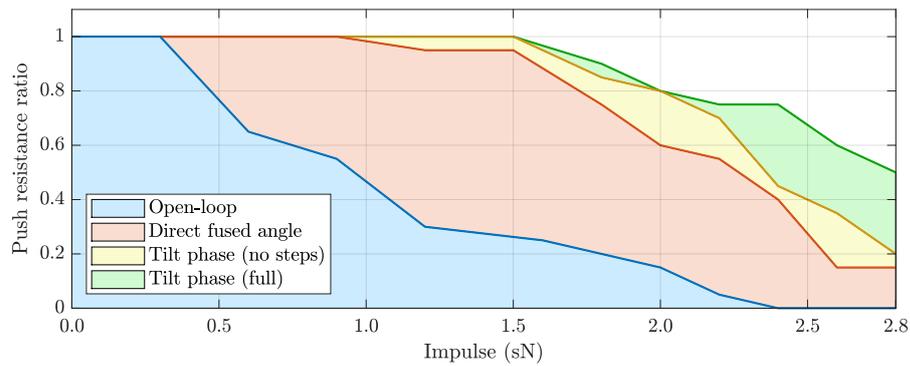

Figure 15.8: Plot of the ratio of withstood pushes against push impulse magnitude for a simulated igus Humanoid Open Platform walking on the spot with the direct fused angle feedback controller, tilt phase controller without step size adaptation enabled, and full tilt phase controller. A push resistance curve for the open-loop Central Pattern Generator is provided as a reference. The raw data corresponding to the plot is given in Tables 13.1 and 15.1.

Although the step size adaptation makes slight improvements for moderate pushes, the true power of the reactive steps is seen for strong pushes, where it makes a decisive difference in the ability of the robot to capture its imbalances. Importantly however, this added stability does not come at the cost of interfering with the reaction of the robot to milder pushes, as it visibly does for the capture step controller.

To further quantify the positive stabilising effects of the tilt phase step size adjustment scheme, in particular in combination with the arm tilt corrective action, push experiments were performed on a real NimbRo-OP2X robot. As can be seen in the excerpts shown in Video 15.3, sagittal pushes were applied to the robot while it was walking on the spot, and the transient responses were recorded. The push recovery performance of the open-loop and tilt phase feedback gaits are compared in Figure 15.9 by means of a phase plot. The blue trajectories are the pushes and subsequent transient responses that were successfully withstood by the robot, while the red trajectories show pushes that led to a fall. It is immediately apparent that the closed-loop gait has much greater push recovery ability than the open-loop gait. The open-loop gait, for example, can only deal with phase pitch velocities from $-0.5$ to $0.45\,\mathrm{rad/s}$, while the closed-loop gait has a stable range closer to $-0.72$ to $0.75\,\mathrm{rad/s}$. The maximum values of phase pitch that occurred during stable blue trajectories also increased from $-0.2$ to $0.2\,\mathrm{rad}$ for open-loop, to $-0.35$ to $0.27\,\mathrm{rad}$ for closed-loop. This increase is mostly attributed to the step size adjustment scheme, as it allows otherwise irrecoverable phase pitches (beyond the sagittal tipping point of the robot) to be recovered through a change of contact point with the ground. Note that the natural tipping point of the NimbRo-OP2X is less than for the igus Humanoid Open Platform due to differences in mass distribution and foot size to height ratio.

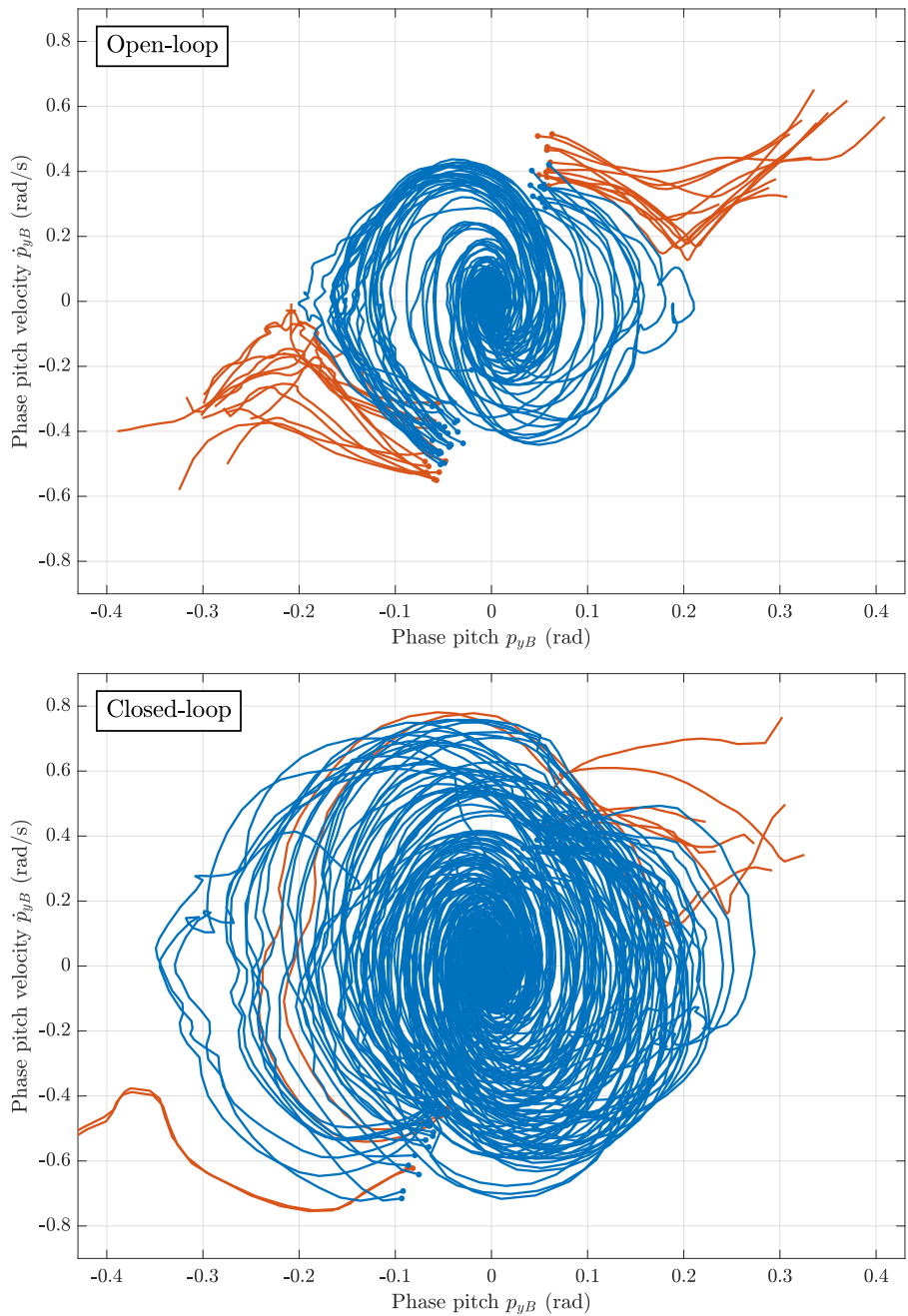

Figure 15.9: Plots of the phase response of a walking NimbRo-OP2X robot when sagittal pushes of various strengths are applied. The phase pitch velocity $\dot{p}_{yB}$ is plotted against the phase pitch $p_{yB}$ for each individual push trajectory, and the resulting curve is coloured according to whether the robot successfully withstood the push (blue) or not (red). The beginning of each trajectory is marked with a solid dot. The top plot corresponds to the performance of the robot with the tilt phase controller disabled (open-loop), and the bottom plot shows how this performance improves with the controller enabled (closed-loop), including in particular with step size adjustments enabled. Refer to Video 15.3 for videos of a subset of the performed pushes.

The difference between stable and unstable trajectories is relatively clear-cut in the open-loop case, as the situation is relatively simple. In the closed-loop case however, the use of reactive steps complicates the response of the robot, and makes it more dependent on the gait phase of the robot at the time of the push. As an effect of this, the exact line of separation between falling and non-falling trajectories becomes somewhat less well-defined. In fact, for two of the failed backwards pushes, the robot actually fell forwards after recovering from the backwards direction.

The data in Figure 15.9 can alternatively be viewed as a heat map, as shown in Figure 15.10. The states in the phase space are divided into cells (i.e. bins), and the trajectories that pass through any one cell are collected and used to calculate a success rate for the cell. The success rate is a value in the range $[0, 1]$, and corresponds to the proportion of trajectories that pass through the cell that did not end in a fall. Overall, Figure 15.10 shows once again the significant improvement in stability that can be attributed to the tilt phase controller, and how it effectively leads to a robust humanoid gait.

## 15.4   CONCLUSION

Walking does not always require overly complex stabilisation mechanisms to achieve high levels of robustness. In this chapter, a feedback controller for robust bipedal walking has been presented that relies solely on measurements from a single 6-axis IMU, and is applicable to low-cost robots with noisy sensors, imperfect actuation and limited computing power. No highly tuned or complex physical models are required, and great portability is ensured through the use of dimensionless parameters and configuration constants. The wide variety of corrective actions that are employed cover many different aspects of balanced walking, including both short-term and long-term stability. In summary, although conceptually relatively simple, the tilt phase controller, in combination with the Keypoint Gait Generator (KGG), achieves genuinely good walking results with minimal assumptions about the robot hardware and performance.

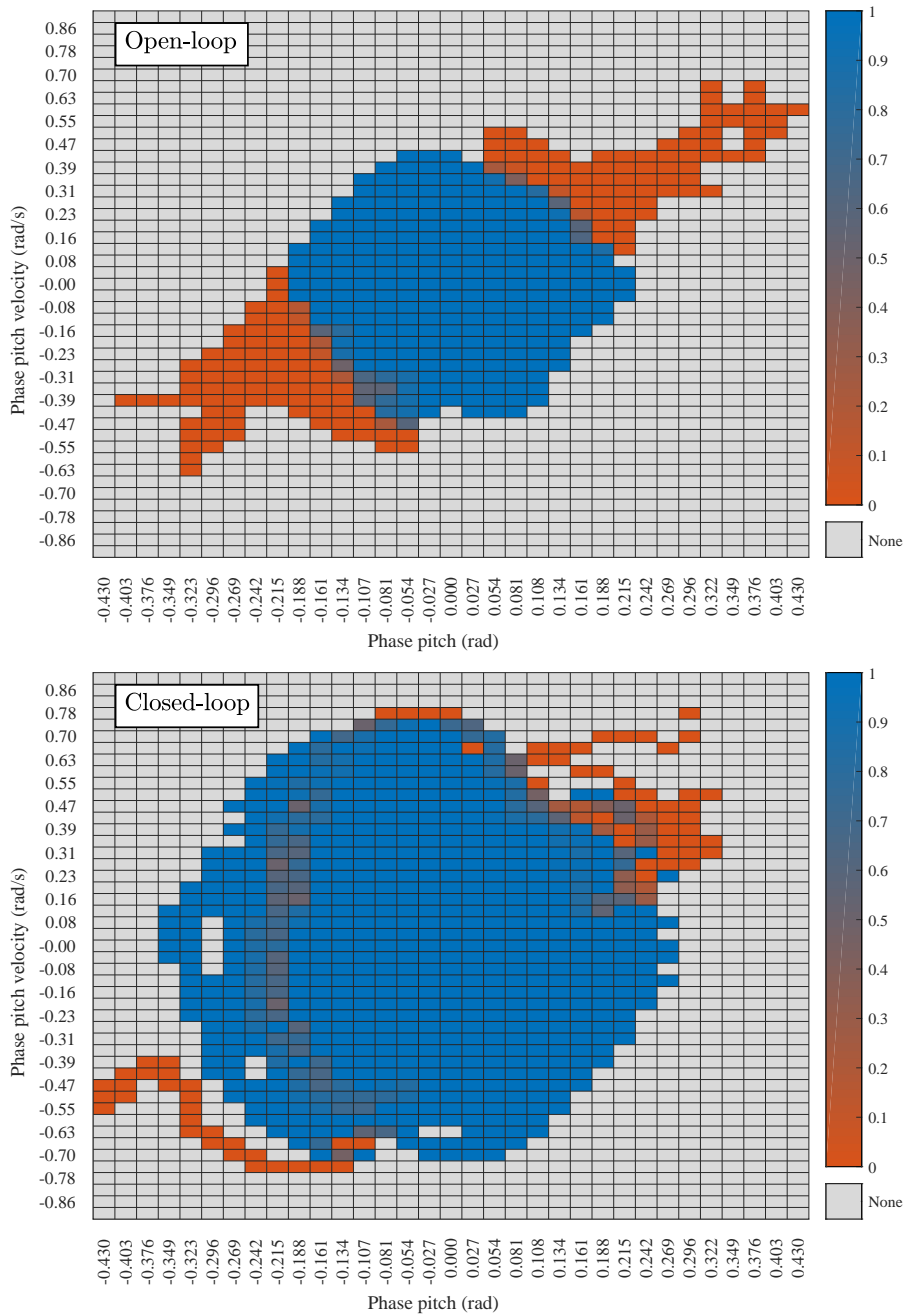

Figure 15.10: Plots of the open-loop vs. closed-loop walking stability of a NimbRo-OP2X robot in the form of a phase space heat map. The blue areas indicate phase states corresponding to stable walking (1.0 = 100% success rate), while the red areas indicate states where the robot tendentially lost balance and fell over (0.0 = 0% success rate). In-between colours indicate that certain binned states were encountered multiple times with different outcomes. The entirety of the data from the push tests has been incorporated into this figure, including also the states of normal balanced walking between pushes. The grey cells correspond to phase states that were not encountered during the tests.

Part V

CLOSING

# 16

CONCLUSION

Bipedal walking does not always require overly complex stabilisation mechanisms to achieve high levels of robustness. The tilt phase and direct fused angle feedback controllers developed in this thesis are a demonstration of this fact. Both controllers extend an underlying open-loop gait with a diverse array of corrective actions that apply effective long and short-term adjustments to the balance of the robot.[1] With just the pitch and roll orientations of the trunk as the sole source of feedback, the controllers apply relatively simple yet well-thought-out feedback pipelines to activate the required corrective actions at the required times. It is a very important feature of the controllers that they do not apply corrective actions at all if the robot is walking within a certain margin of its nominal orientation limit cycle. The application of step size adjustments is crucially also only used as a last resort for keeping balance, i.e. only in cases where the deviation from normal walking is predicted to be too severe for the remaining corrective actions to be able to deal with it. This is important as it minimises the effect that the balance controllers have on disturbing the intended walking direction and footstep placements of the robot.

The use of open-loop gait generators as the underlying principle of walking is advantageous, as it reduces an otherwise high-dimensional control problem into an interface that has just a few very manageable corrective action inputs. Further advantages of the use of gait generators include the flexibility in manual correction and tuning that they provide, and the observation that they allow robots to find their own natural rhythm while walking, instead of imposing one that was calculated from some trajectory optimisation problem. Both the Central Pattern Generator (CPG) and Keypoint Gait Generator (KGG) are analytically formulated, and along with their respective aforementioned balance controllers, are very computationally inexpensive due to this trait.

Most of the methods presented in this thesis were specifically designed with low-cost robots in mind. Low-cost robots often do not have a wide variety of sensors available, and the sensors that they do have are not always of good quality. As a result, the feedback controllers that were designed to stabilise the gaits in this thesis were formulated in such a way that they only assumed the presence of a single 6-axis Inertial Measurement Unit (IMU). This makes the gaits extraordinarily portable between robots, as there is no requirement for force-torque sensors, foot contact sensors, or anything of the like. Extensive thought was also put into calibration and filtering of the sensor data, to make the most of the signal amongst the noise. One particular success in this direction was the introduction of the gyroscope bias autocalibration scheme, which significantly increased the integration accuracy of the attitude estimator presented in Section 10.1.

---

1 Examples of such corrective actions include step size, timing and arm/foot tilt adjustments.

In addition to their general limitations in sensing, low-cost robots are also fundamentally imprecise about the motions they try to execute. Problems such as low power-to-weight ratios, structural non-rigidities, joint backlash and stiction make the tracking of any precise trajectories difficult, or even impossible. The use of tuneable gait generators in combination with a feed-forward actuator control scheme helped deal with these limitations. The sparing use of any physical and/or dynamic models of the robot also contributed to a wider and arguably easier field of applicability of the developed feedback gaits.

Many of the developments in state estimation and bipedal walking that were made in this thesis were founded on other developments that were made in the field of 3D rotation theory. The most significant of these developments were the novel use of fused yaw and tilt to partition 3D rotations into highly problem-relevant rotation subcomponents, and the corresponding development of the tilt angles, fused angles and tilt phase space representations. In addition to providing mathematically and geometrically useful concepts of 'yaw', 'pitch' and 'roll', these developments in 3D rotation theory also led to the introduction of the concepts of referenced rotations and tilt vector addition. Both these concepts proved to be very useful in particular throughout the formulation of the KGG and the corresponding tilt phase controller.

Validated by experiments that were performed in both simulation and on real robot hardware, the core principle of this thesis can be summarised as follows—if the sensor management and feedback chains are carefully constructed, comparatively simple model-free and robot-agnostic feedback mechanisms can successfully and robustly stabilise a generic bipedal gait.

## 16.1 FUTURE DIRECTIONS

Several possible directions of future research exist that would build on this thesis. These directions include:

- Investigating the use of multiple intentionally non-aligned gyroscope and accelerometer sensors for the purposes of increased attitude estimation fidelity and better rejection of noise and outliers.

- Applying the nonlinear tripendulum model to the lateral plane of motion as well, and activating the lateral step size (and possibly swing out) corrective action based on that.

- Using learning methods to learn the interplay between lateral step size and timing and the respective effects on the progression of the lateral tilt of the robot, in order to concurrently predict suitable activation values.

- Using learning methods to learn a higher level controller for the KGG, and all of its corrective actions.[2]

- Reformulating the capture step controller to avoid the complete asynchronicity of the RX and MX models. Gradual total dissociation of the states of these two models is the leading observed cause of failure of tuning of the capture step controller.

- Modelling the changing z-rotations of the feet in a more explicit way in terms of how it affects the balance of the robot throughout the duration of a step, and appropriately modifying the constructed feedback controllers in light of this.

- Extending the capture step controller to a full 2.5D treatment (2D position plus heading), to avoid the segregated treatment of the sagittal and lateral planes of motion, and avoid the lack of consideration of the effects of foot yaw.

---

2 As a step in this direction, for example, Bayesian optimisation for feedback stabilisation of the CPG gait has been performed (Rodriguez et al., 2018).

Part VI

APPENDIX

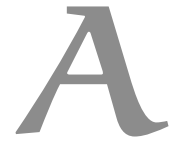

# MATHEMATICAL FUNCTIONS AND FILTERS

Many standard and non-standard mathematical functions and filters are required throughout this thesis. This appendix provides a summary and brief description of each of these, for reference purposes. Complete ready-to-use implementations of all of the presented algorithms are available at:

https://github.com/AIS-Bonn/humanoid_op_ros/tree/master/src/nimbro_robotcontrol/util/rc_utils/include/rc_utils

## CHAPTER CONTENTS

## A.1    MATHEMATICAL FUNCTIONS

The mathematical functions required throughout the remainder of this thesis are detailed in this section.

### A.1.1    Basic Mathematical Functions

We begin with some generic basic mathematical functions.

#### A.1.1.1    *Angle Wrapping*

Angular quantities resulting from numerical calculations often have to be standardised by wrapping them by multiples of $2\pi$ to a particular range of $2\pi$, usually $(-\pi, \pi]$. The standard wrap$(\cdot)$ function does just that:

$$\text{wrap}(\theta) = \theta + 2\pi \left\lfloor \frac{\pi - \theta}{2\pi} \right\rfloor, \tag{A.1}$$

where $\lfloor \cdot \rfloor$ is the floor function. The wrap$_0(\cdot)$ function alternatively wraps an angle by multiples of $2\pi$ to the range $[0, 2\pi)$:

$$\text{wrap}_0(\theta) = \theta - 2\pi \left\lfloor \frac{\theta}{2\pi} \right\rfloor. \tag{A.2}$$

In code, the wrapping functions are referred to as `picut`.

### A.1.1.2 *Sign Function*

The standard mathematical sign function is given by

$$
\text{sgn}(x) = \begin{cases} 1 & \text{if } x > 0, \\ 0 & \text{if } x = 0, \\ -1 & \text{if } x < 0. \end{cases} \tag{A.3}
$$

An alternative custom sign function that avoids an output value of zero is given by

$$
\text{sign}(x) = \begin{cases} 1 & \text{if } x \geq 0, \\ -1 & \text{if } x < 0. \end{cases} \tag{A.4}
$$

### A.1.1.3 *Ellipse Radius*

Given the required ellipse semi-axis lengths $a, b > 0$, the radius of the corresponding ellipse in the direction of the angle $\theta$ (measured counterclockwise (CCW) from the positive x-axis) is given by

$$
R = \sqrt{\frac{1}{\frac{\cos^2 \theta}{a^2} + \frac{\sin^2 \theta}{b^2}}}. \tag{A.5}
$$

The ellipse radius in the direction of any 2D vector $(x, y)$ is given by

$$
R = \sqrt{\frac{x^2 + y^2}{\frac{x^2}{a^2} + \frac{y^2}{b^2}}}. \tag{A.6}
$$

Equation (A.6) generalises easily to the n-dimensional ellipsoid case. If $\mathbf{s} = (s_1, \ldots, s_n)$ is the vector of required semi-axis lengths, and $\mathbf{x} = (x_1, \ldots, x_n)$ is the direction in which the radius of the ellipsoid should be calculated, then the required radius is given by

$$
R = \sqrt{\frac{\|\mathbf{x}\|^2}{\sum_{i=1}^{n} \frac{x_i^2}{s_i^2}}}. \tag{A.7}
$$

### A.1.2 **Coercion**

Coercion, also known as saturation, is the process of limiting a variable to a particular range or region in space.

### A.1.2.1  *Scalar Hard Coercion*

Scalar hard coercion refers to the process of limiting a scalar variable $x \in \mathbb{R}$ to an arbitrary interval $[m, M]$. Mathematically, this is given by

$$\operatorname{coerce}(x, m, M) = \begin{cases} m & \text{if } x \in (\infty, m), \\ x & \text{if } x \in [m, M], \\ M & \text{if } x \in (M, \infty). \end{cases} \tag{A.8}$$

If the coerce(...) function is called with values of $m$ and $M$ such that $m \geq M$, by convention the output is taken to be $\frac{1}{2}(m + M)$. An example of scalar hard coercion can be seen in Figure A.1.

### A.1.2.2  *Elliptical Hard Coercion*

Instead of coercing a scalar variable to an interval on the real number line, it is also possible to coerce a vector $\mathbf{x} \in \mathbb{R}^n$ to an n-dimensional ellipse/ellipsoid defined by the semi-axis lengths $\mathbf{s} \in \mathbb{R}_+^n$. Given Equations (A.7) and (A.8), the required coerced output vector $\mathbf{x}_o \in \mathbb{R}^n$ is given by

$$\mathbf{x}_o = \frac{\mathbf{x}}{\|\mathbf{x}\|} \operatorname{coerce}(\|\mathbf{x}\|, -R, R). \tag{A.9}$$

For the 2D case of an ellipse, note that we have $\mathbf{x} = (x, y)$ and $\mathbf{s} = (a, b)$, and that we can use Equation (A.6) to evaluate $R$.

### A.1.2.3  *Scalar Soft Coercion*

One possible problem with the use of scalar hard coercion in practice is that it is not smooth at the points where it transitions from the linear centre region to the adjoining coerced regions (i.e. at $x = m, M$). This can cause problems, for example, with discontinuous velocities when applied to motion trajectories. The solution to this problem is soft coercion. Scalar soft coercion of the variable $x \in \mathbb{R}$ to the interval $[m, M]$ with a soft coercion buffer of $b \in \mathbb{R}$ is given by

$$x_o = \begin{cases} m + b \exp\big(\frac{S}{b}(x - m - b)\big) & \text{if } x \in (-\infty, m + b), \\ x & \text{if } x \in [m + b, M - b], \\ M - b \exp\big(\frac{S}{b}(M - b - x)\big) & \text{if } x \in (M - b, \infty), \end{cases} \tag{A.10}$$

where nominally $S = 1$, but if $b > b_{max}$ for $b_{max} = \frac{1}{2}(M - m)$, then we instead calculate

$$S = \exp\Big(\frac{b_{max}}{b} - 1\Big), \tag{A.11}$$

after which we set $b = b_{max}$ and evaluate Equation (A.10). It should be noted that soft coercion is a $\mathcal{C}^1$ function of $x$. Examples of scalar soft coercion for various different values of $b$ can be seen in Figure A.1.

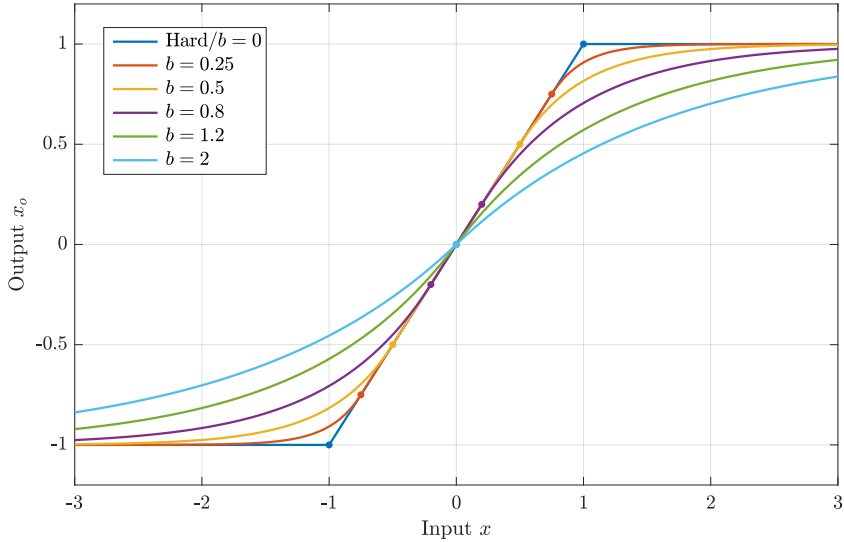

Figure A.1: Illustration of the scalar soft coercion transfer function for $[m, M] = [-1, 1]$ and various different values of the soft coercion buffer $b$. Note that $b = 0$ corresponds to hard coercion. The marked dots show the points of transition where the effects of coercion start.

### A.1.2.4  *Elliptical Soft Coercion*

The same extension of scalar hard coercion to the n-dimensional case can also be applied to scalar soft coercion. Similar to Equation (A.9), this leads to

$$\mathbf{x}_o = \frac{\mathbf{x}}{\|\mathbf{x}\|} \, \text{coerceSoft}\big(\|\mathbf{x}\|, -R, R, b\big), \tag{A.12}$$

where $R$ is calculated using Equation (A.7) as before.

### A.1.3  **Deadbanding**

Deadbanding refers to the concept of applying intentional deadband to a signal, often for the purpose of suppressing noise.

### A.1.3.1  *Sharp Deadband*

Although not necessarily aligned with the traditional signal processing definition of the term 'deadband', in this thesis the notion of deadband (i.e. sharp deadband) refers to the transfer function

$$x_o = \begin{cases} x + r - c & \text{if } x \in (-\infty, c - r), \\ 0 & \text{if } x \in [c - r, c + r], \\ x - r - c & \text{if } x \in (c + r, \infty), \end{cases} \tag{A.13}$$

where $x$ is the input, $x_o$ is the output, $r$ is the deadband radius, and $c$ is the centre of deadband. A visual example of how this looks can be seen in Figure A.2.

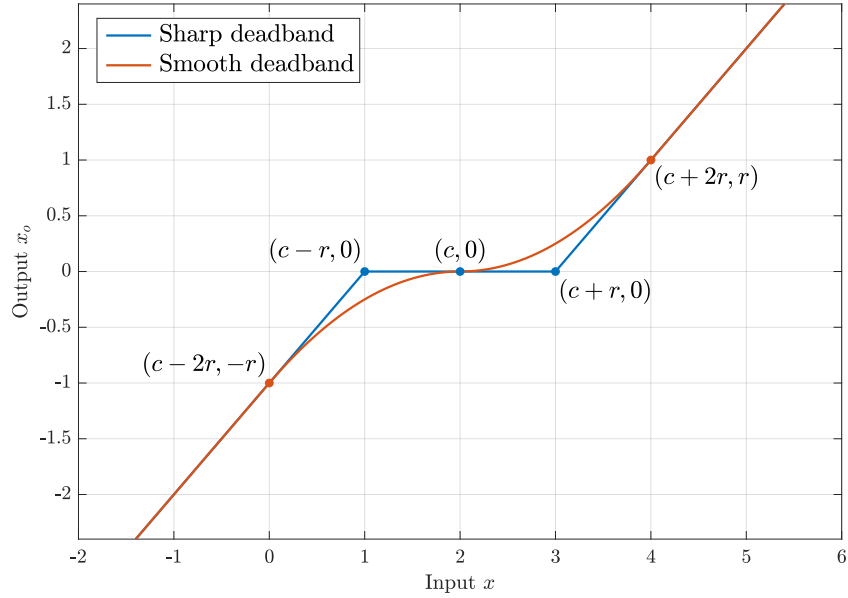

Figure A.2: Illustration of the sharp and smooth deadband transfer functions for $c = 2$ and $r = 1$.

### A.1.3.2  *Elliptical Sharp Deadband*

Sharp deadband can easily be generalised to n-dimensions. The region of deadband then corresponds to an ellipse/ellipsoid, instead of just a line interval. If $\mathbf{x} \in \mathbb{R}^n$ is the input, $\mathbf{s} \in \mathbb{R}^n_+$ is the semi-axis lengths of the required deadband ellipsoid, and $\mathbf{c}$ is the required deadband centre, then

$$\tilde{\mathbf{x}} = \mathbf{x} - \mathbf{c}, \tag{A.14a}$$

$$\mathbf{x}_o = \begin{cases} \mathbf{0} & \text{if } \|\tilde{\mathbf{x}}\| \leq R, \\ \left(\frac{\|\tilde{\mathbf{x}}\| - R}{\|\tilde{\mathbf{x}}\|}\right)\tilde{\mathbf{x}} & \text{otherwise,} \end{cases} \tag{A.14b}$$

where $R$ is calculated from $\tilde{\mathbf{x}}$ and $\mathbf{s}$ using Equation (A.7). For the 2D case of an ellipse, note that we have $\tilde{\mathbf{x}} = (x, y)$ and $\mathbf{s} = (a, b)$, and that we can use Equation (A.6) to evaluate $R$.

### A.1.3.3  *Smooth Deadband*

Sharp deadband is continuous, but not differentiable at the points $x = c \pm r$. This can be a problem if, for example, sharp deadband is applied to a motion trajectory, as it can potentially cause discontinuities in the resulting velocity waveforms. Smooth deadband is a $\mathcal{C}^1$ variant of sharp deadband that avoids this issue. If $x$ is the input, $r$ is the required deadband radius, and $c$ is the centre of deadband, smooth deadband is mathematically given by

$$\tilde{x} = x - c, \tag{A.15a}$$

$$x_o = \begin{cases} \frac{1}{4r}\tilde{x}^2 \operatorname{sgn}(\tilde{x}) & \text{if } |\tilde{x}| \le 2r, \\ \tilde{x} - r \operatorname{sgn}(\tilde{x}) & \text{otherwise.} \end{cases} \tag{A.15b}$$

A visual example of how this looks can be seen in Figure A.2. If only one-sided smooth deadband is desired (keeping only the positive output values), Equation (A.15b) turns into

$$x_o = \begin{cases} 0 & \text{if } \tilde{x} \in (-\infty, 0), \\ \frac{1}{4r}\tilde{x}^2 & \text{if } \tilde{x} \in [0, 2r], \\ \tilde{x} - r & \text{if } \tilde{x} \in (2r, \infty). \end{cases} \tag{A.16}$$

#### A.1.3.4  *Elliptical Smooth Deadband*

The same extension of sharp deadband to the n-dimensional case can also be applied to smooth deadband. Similar to Equation (A.14b), this leads to

$$\mathbf{x}_o = \begin{cases} \frac{\|\tilde{\mathbf{x}}\|}{4R} \tilde{\mathbf{x}} & \text{if } \|\tilde{\mathbf{x}}\| \le 2R, \\ \left(\frac{\|\tilde{\mathbf{x}}\| - R}{\|\tilde{\mathbf{x}}\|}\right) \tilde{\mathbf{x}} & \text{otherwise,} \end{cases} \tag{A.17}$$

where $R$ is calculated using Equation (A.7) as before.

### A.1.4  **Interpolation**

The formula for simple linear interpolation and/or extrapolation is most succinctly given by

$$y = y_1 + \left(\frac{y_2 - y_1}{x_2 - x_1}\right)(x - x_1), \tag{A.18}$$

where $(x_1, y_1)$ and $(x_2, y_2)$ are the two points being interpolated, $x$ is the input x-value, and $y$ is the resulting interpolated y-value. Note that if we define

$$u = \frac{x - x_1}{x_2 - x_1}, \tag{A.19}$$

this can alternatively be written as

$$y = (1 - u)y_1 + uy_2. \tag{A.20}$$

If $x_1 = x_2$, by convention we define $y = \frac{1}{2}(y_1 + y_2)$ irrespective of $x$. In this thesis, we use the notation

$$y = \operatorname{interpolate}\big([x_1, x_2] \to [y_1, y_2], \, x\big) \tag{A.21}$$

for the linear interpolation procedure given by Equations (A.18) and (A.20).

### A.1.4.1  Coerced Interpolation

Coerced interpolation is similar to normal linear interpolation, but ensures that the output y-value is always in the specified y-range, i.e.

$$y \in [y_1, y_2] \text{ for all } x \in \mathbb{R}. \tag{A.22}$$

Essentially, coerced interpolation ensures that extrapolation never takes place, and by convention uses the notation

$$y = \text{interpolateCoerced}\big([x_1, x_2] \rightarrow [y_1, y_2], x\big). \tag{A.23}$$

Mathematically, coerced interpolation is given by

$$u = \text{coerce}\Big(\frac{x - x_1}{x_2 - x_1}, 0, 1\Big), \tag{A.24a}$$

$$y = (1 - u)y_1 + u y_2. \tag{A.24b}$$

### A.1.4.2  Piecewise Linear Interpolation

Linear interpolation can trivially be extended to piecewise linear interpolation via the specification of multiple keypoints. If $(x_i, y_i) \in \mathbb{R}^2$ for $i = 1, \dots, n$ are any collection of keypoints such that

$$x_1 < x_2 < \cdots < x_n, \tag{A.25}$$

then given any input $x \in \mathbb{R}$, we can define

$$j = \begin{cases} 1 & \text{if } x \in (-\infty, x_1], \\ i & \text{if } x \in (x_i, x_{i+1}], \\ n - 1 & \text{if } x \in (x_n, \infty). \end{cases} \tag{A.26}$$

The required piecewise interpolated y-value is then

$$y = \text{interpolate}\big([x_j, x_{j+1}] \rightarrow [y_j, y_{j+1}], x\big). \tag{A.27}$$

### A.1.4.3  Cyclic Piecewise Linear Interpolation

Piecewise linear interpolation can also be applied to the cyclic case, i.e. for angles. Suppose we have $n$ distinct keypoints $(\theta_i, y_i)$ on the cyclic interval $\theta_i \in (-\pi, \pi]$ such that all but exactly one of the following inequalities hold:

$$\theta_1 < \theta_2 < \cdots < \theta_n < \theta_1. \tag{A.28}$$

This is the cyclic equivalent of requiring the values of $\theta_i$ to be strictly monotonically increasing, with the provision of allowing a jump once from $\pi$ back to $-\pi$. Given an input angle $\theta$, the result of cyclic piecewise linear interpolation is given by applying standard piecewise linear interpolation to the data, while being careful to factor in the effects of angle wrapping in the one interval where $\theta_i$ jumps from $\pi$

---

**Algorithm A.1:** Cyclic warping

---

**Input:** $\theta \in (-\pi, \pi]$, and keypoints $(\theta_i, \phi_i)$ for $i = 1, \ldots, n$, where $\{\theta_i\}$ and $\{\phi_i\}$ are both cyclically strictly monotonically increasing

**Output:** Warped output $\phi \in (-\pi, \pi]$

1: **function** CYCLICWARP($\{\theta_i\}, \{\phi_i\}, \theta$)
2:     **for** $i \leftarrow 1$ **to** $n$ **do**
3:         $\hat{\phi}_{i+1} \leftarrow \phi_{i+1}$         $\triangleright$ Note that $\theta_{n+1} \equiv \theta_1$ and $\phi_{n+1} \equiv \phi_1$
4:         **if** $\phi_{i+1} < \phi_i$ **then**
5:             $\hat{\phi}_{i+1} \leftarrow \hat{\phi}_{i+1} + 2\pi$
6:         **end if**
7:         **if** $\theta_i \leq \theta \leq \theta_{i+1}$ **and** $\theta_i \neq \theta_{i+1}$ **then**
8:             $\hat{\phi} \leftarrow$ interpolate$\left([\theta_i, \theta_{i+1}] \rightarrow [\phi_i, \hat{\phi}_{i+1}], \theta\right)$
9:             **break**
10:         **end if**
11:         **if** $\theta_{i+1} \leq \theta_i$ **then**
12:             **if** $\theta \geq \theta_i$ **then**
13:                 $\hat{\phi} \leftarrow$ interpolate$\left([\theta_i, \theta_{i+1}+2\pi] \rightarrow [\phi_i, \hat{\phi}_{i+1}], \theta\right)$
14:                 **break**
15:             **end if**
16:             **if** $\theta \leq \theta_{i+1}$ **then**
17:                 $\hat{\phi} \leftarrow$ interpolate$\left([\theta_i - 2\pi, \theta_{i+1}] \rightarrow [\phi_i, \hat{\phi}_{i+1}], \theta\right)$
18:                 **break**
19:             **end if**
20:         **end if**
21:     **end for**
22:     $\phi \leftarrow$ wrap$(\hat{\phi})$
23:     **return** $\phi$
24: **end function**

---

to $-\pi$. If cyclic interpolation is applied from one set of monotonically increasing angle values to another, it is referred to as cyclic warping or angle warping, and extra care has to be taken at the point where the target y-values also jump from $\pi$ to $-\pi$. Pseudocode for the cyclic warping algorithm is given in Algorithm A.1.

### A.1.4.4 *Sine Function Fillet*

The Leg Lifting component (see page 325) of the Central Pattern Generator (CPG) requires smooth $\mathcal{C}^1$ connections to be calculated between sinusoid hemispheres and the constant zero function. Refer to the difference between the solid and dotted red lines in Figure 11.3 for an illustration of what this means. Suppose we have the generic sinusoid function

$$y(t) = A \sin(Bt), \tag{A.29}$$

for $A, B \in \mathbb{R}$ and $t \in [0, \pi]$. A quadratic sine function fillet

$$q(t) = a(t - T)^2 + b(t - T) + c \tag{A.30}$$

can be calculated, such that for some $T, Z \in \mathbb{R}$ the function

$$\tilde{y}(t) = \begin{cases} 0 & \text{if } t \in [-\pi, -Z), \\ q(t) & \text{if } t \in [-Z, T), \\ y(t) & \text{if } t \in [T, \pi], \end{cases} \tag{A.31}$$

is $\mathcal{C}^1$. Clearly, for this to be the case we must have

$$q(-Z) = 0, \qquad q(T) = y(T), \tag{A.32a}$$

$$q'(-Z) = 0, \qquad q'(T) = y'(T). \tag{A.32b}$$

Knowing $A$, $B$ and the desired fillet time $T$, we can solve these constraints to give

$$c = A \sin(BT), \qquad b = AB \cos(BT), \tag{A.33a}$$

$$a = \frac{b^2}{4c}, \qquad Z = \frac{2c}{b} - T. \tag{A.33b}$$

Based on these parameters, the fillet function $\tilde{y}(t)$ is completely defined, and $\mathcal{C}^1$ as required.

Given this general methodology, sine function fillets can also be calculated for the end of a sinusoid hemisphere by simply mirroring the problem mathematically. If constraints on the magnitude of $Z$ are present, these can be enforced using a simple numerical approach based on Newton's method (refer to the `LinSinFillet` class in the released code).

### A.1.5 Spline Interpolation

Spline interpolation is the task of constructing a waveform that passes through a set of $n$ keypoints $(t_i, x_i)$, with possible velocity boundary conditions $\dot{x}_1$ and $\dot{x}_n$ at the first and last of these keypoints. Note that $t$ in this case is a 'time' variable, and must therefore be strictly monotonically increasing for $i = 1, \ldots, n$.

#### A.1.5.1 *Linear Spline Interpolation*

The simplest approach to spline interpolation is linear spline interpolation, which simply uses the piecewise linear interpolation method from Appendix A.1.4.2 to join the keypoints over time with linear components. Although the individual linear segments are technically smooth functions, i.e. $\mathcal{C}^\infty$, the overall resulting trajectory is not smooth, or even $\mathcal{C}^1$, at the keypoints themselves due to step changes in velocity. It is also not possible to include any velocity boundary conditions into the interpolation.

### A.1.5.2 *Trapezoidal Velocity Spline Interpolation*

While linear spline interpolation can be seen as 'constant velocity' interpolation, trapezoidal velocity spline interpolation can be seen as 'piecewise constant acceleration' interpolation. Each $\mathcal{C}^1$ interpolation segment nominally starts with a burst of constant acceleration in one direction, has a middle phase of the required length with zero acceleration, and finishes with a burst of constant acceleration in the opposite direction. This generally results in a velocity profile that looks distinctly like a trapezoid, especially when a keypoint segment starts and stops from zero velocity. Velocity specifications, i.e. $\dot{x}_i$, are required for all $n$ keypoints, and two further inputs, $v_m$ and $a$, are required to make the solution (essentially) unique. $v_m$ is the maximum absolute velocity to allow during interpolation, and $a$ is the magnitude of acceleration to use in the bursts of acceleration that take place at the beginning and end of each interpolation segment. No $t_i$ values are used as inputs, as the trapezoidal interpolation method automatically calculates the solution that takes the shortest amount of time to reach each successive $x_i$.

The exact algorithm behind the calculation of trapezoidal spline interpolation is involved, due to the presence of multiple possible solutions in general. The details of the algorithm are best documented by the released `TrapVelSpline` code.[1]

### A.1.5.3 *Cubic Spline Interpolation*

Spline interpolation can also be performed using cubic functions. If $\Delta t_i = t - t_i$ for $i = 1, \ldots, n-1$, the cubic spline interpolant corresponding to the $i^{\text{th}}$ segment is given by

$$f_i : [t_i, t_{i+1}] \to \mathbb{R}$$
$$f_i(t) = x_i + \dot{x}_i \Delta t_i + \tfrac{1}{2} a_i \Delta t_i^2 + \tfrac{1}{6} j_i \Delta t_i^3, \quad (A.34)$$

where $(t_i, x_i)$ are the input keypoints, $\dot{x}_i$ are the desired keypoint velocities, and $a_i, j_i$ are the respective calculated keypoint accelerations and jerks. Clearly,

$$f_i(t_i) = x_i, \quad (A.35a)$$
$$f_i'(t_i) = \dot{x}_i. \quad (A.35b)$$

In order to ensure the continuity of the interpolated position and velocity trajectories however, we also need to ensure that

$$f_i(t_{i+1}) = x_{i+1}, \quad (A.36a)$$
$$f_i'(t_{i+1}) = \dot{x}_{i+1}. \quad (A.36b)$$

---

[1] https://github.com/AIS-Bonn/humanoid_op_ros/blob/master/src/nimbro_robotcontrol/util/rc_utils/include/rc_utils/math_spline.h

Defining $h_i = t_{i+1} - t_i$, Equation (A.36) leads to the solution

$$a_i = C - D, \tag{A.37a}$$

$$j_i = \frac{1}{h_i}(3D - 2C), \tag{A.37b}$$

where

$$C = \frac{6}{h_i^2}(x_{i+1} - x_i - \dot{x}_i h_i), \tag{A.38a}$$

$$D = \frac{2}{h_i}(\dot{x}_{i+1} - \dot{x}_i). \tag{A.38b}$$

Using these general solution equations, the complete $\mathcal{C}^1$ cubic spline trajectory passing through all $n$ $(t_i, x_i, \dot{x}_i)$ keypoints can thus be calculated to be

$$f : [t_1, t_n] \to \mathbb{R}$$

$$f(t) = \Big\{ f_i(t) \quad \text{if } t \in [t_i, t_{i+1}], \tag{A.39}$$

for $i = 1, \dots, n-1$.

### A.1.5.4  *Cubic Spline Intermediate Velocities*

Given a sequence of $n$ keypoints $(t_i, x_i)$ and just the velocity boundary conditions $\dot{x}_1$ and $\dot{x}_n$, suitable values for the intermediate velocities $\dot{x}_2, \dots, \dot{x}_{n-1}$ can be calculated by enforcing a continuous acceleration constraint. This makes the total interpolated cubic spline trajectory $\mathcal{C}^2$, and essentially reduces the total process of interpolation to just two 'degrees of freedom'—the choice of the value of $\dot{x}_1$, and the choice of the value of $\dot{x}_n$. Mathematically, the continuity of acceleration constraint corresponds to

$$f_i''(t_{i+1}) = f_{i+1}''(t_{i+1}), \tag{A.40}$$

for $i = 1, \dots, n-2$, which using Equation (A.34) can be expanded to give

$$a_i + j_i h_i = a_{i+1}. \tag{A.41}$$

Using Equation (A.37) and defining

$$\bar{v}_i = \frac{x_{i+1} - x_i}{h_i}, \tag{A.42}$$

Equation (A.41) can be expanded to give

$$h_{i+1}\dot{x}_i + 2(h_{i+1} + h_i)\dot{x}_{i+1} + h_i\dot{x}_{i+2} = 3(h_{i+1}\bar{v}_i + h_i\bar{v}_{i+1}). \tag{A.43}$$

Considering Equation (A.43) once for each $i = 1, \dots, n-2$, we can construct the $(n-2) \times (n-2)$ matrix equation

$$A\dot{\mathbf{x}} = \mathbf{b}, \tag{A.44}$$

where

$$A = \begin{bmatrix} 2(h_1+h_2) & h_1 & & & & \\ h_3 & 2(h_2+h_3) & h_2 & & & \\ & \ddots & \ddots & \ddots & & \\ & & h_{n-2} & 2(h_{n-3}+h_{n-2}) & h_{n-3} & \\ & & & h_{n-1} & 2(h_{n-2}+h_{n-1}) \end{bmatrix},$$

$$\dot{\mathbf{x}} = \begin{bmatrix} \dot{x}_2 \\ \dot{x}_3 \\ \vdots \\ \dot{x}_{n-2} \\ \dot{x}_{n-1} \end{bmatrix}, \qquad \mathbf{b} = \begin{bmatrix} 3(h_2\bar{v}_1 + h_1\bar{v}_2) - h_2\dot{x}_1 \\ 3(h_3\bar{v}_2 + h_2\bar{v}_3) \\ \vdots \\ 3(h_{n-2}\bar{v}_{n-3} + h_{n-3}\bar{v}_{n-2}) \\ 3(h_{n-1}\bar{v}_{n-2} + h_{n-2}\bar{v}_{n-1}) - h_{n-2}\dot{x}_n \end{bmatrix}. \quad (A.45)$$

Note that all omitted entries of $A$ are zero. Equation (A.44) can be solved for the required intermediate velocities $\dot{x}_2, \ldots, \dot{x}_{n-1}$ using standard matrix decomposition methods. Given these intermediate velocities, Equation (A.37) can then be applied, as before, to calculate the required parameters $a_i$ and $j_i$ of the interpolating cubic spline segments. Note that using the $\bar{v}_i$ notation from Equation (A.42), Equation (A.37) can be written more directly as

$$a_i = \frac{2}{h_i}\left(3\bar{v}_i - 2\dot{x}_i - \dot{x}_{i+1}\right), \qquad (A.46a)$$

$$j_i = \frac{6}{h_i^2}\left(\dot{x}_i + \dot{x}_{i+1} - 2\bar{v}_i\right). \qquad (A.46b)$$

This concludes the calculation of the intermediate keypoint velocities (and subsequent cubic spline segments) for the case of a continuous acceleration constraint.

### A.1.5.5  *Hermite Cubic Spline Intermediate Velocities*

The task (described in Appendix A.1.5.4) of calculating suitable intermediate velocities for a cubic spline trajectory through the $n$ keypoints $(t_i, x_i)$ can also be solved using shape-preserving piecewise cubic Hermite interpolating polynomials (Fritsch and Carlson, 1980). If we do not impose a continuous acceleration constraint like in Appendix A.1.5.4, we can instead calculate the required $\dot{x}_2, \ldots, \dot{x}_{n-1}$ velocities using weighted harmonic means. Recalling the definitions

$$h_i = t_{i+1} - t_i, \qquad (A.47a)$$

$$\bar{v}_i = \frac{x_{i+1} - x_i}{h_i}, \qquad (A.47b)$$

the required intermediate velocities are given for $i = 2, \ldots, n-1$ by

$$\dot{x}_i = \begin{cases} \dfrac{\bar{v}_{i-1}\bar{v}_i}{w\bar{v}_{i-1} + (1-w)\bar{v}_i} & \text{if } \bar{v}_{i-1}\bar{v}_i > 0, \\ 0 & \text{otherwise,} \end{cases} \qquad (A.48)$$

where the weight $w$ is given by

$$w = \frac{2h_{i-1} + h_i}{3(h_{i-1} + h_i)}. \qquad (A.49)$$

Equation (A.48) essentially expresses that each intermediate velocity is given by the harmonic mean[2] of the slopes of the linear interpolants of the neighbouring two segments, unless these slopes are in opposite directions, in which case the required intermediate velocity is set to zero. This ensures that all changes in direction of the cubic spline trajectory occur at the keypoints themselves. As a result, no keypoint can ever locally be overshot.

One advantage of cubic Hermite interpolation is that it is direct to evaluate in the sense that it does not require the solution of a matrix equation to be calculated. Cubic Hermite interpolation in the described form is also very local. The velocity at each keypoint is only affected by its position and time, and the position and time of the two keypoints either side of it. As such, the interpolating cubic between any two adjacent keypoints $(t_i, x_i)$ and $(t_{i+1}, x_{i+1})$ is only affected by the adjoining two keypoints $(t_{i-1}, x_{i-1})$ and $(t_{i+2}, x_{i+2})$, and nothing else. This can be an advantageous property in some applications of cubic spline interpolation.

### A.1.5.6  *Cyclic Cubic Spline Intermediate Velocities*

The methods presented in Appendices A.1.5.4 and A.1.5.5 for calculating suitable intermediate keypoint velocities can also be applied to the case of cyclic cubic splines, also known as periodic cubic splines. Cyclic cubic splines are cubic spline trajectories that enforce the additional constraint that the first and last keypoints have to be identical (as the trajectories are generally intended to be traversed in a cyclic loop). This in particular means that we are constrained to have

$$x_1 = x_n, \qquad (A.50a)$$
$$\dot{x}_1 = \dot{x}_n. \qquad (A.50b)$$

If we are given just the $n$ keypoints $(t_i, x_i)$, and no velocities $\dot{x}_i$ at all, in the case of cubic Hermite interpolation we can calculate $\dot{x}_2, \ldots, \dot{x}_{n-1}$ exactly as before, and then proceed to calculate $\dot{x}_1 \equiv \dot{x}_n$ by evaluating

---

2 That is, weighted based on the relative differences between the time intervals of the two adjoining segments. If all segments are equally spaced in time then all $w$ values are equal to exactly $\frac{1}{2}$.

Equation (A.48) for $i = 1$ with the definition $(t_0, x_0) \equiv (t_{n-1}, x_{n-1})$. In the alternative case that we wish to apply a continuous acceleration constraint at all the keypoints (as in Appendix A.1.5.4), we can modify the matrix equation given by Equations (A.44) and (A.45) in light of the new constraint

$$\ddot{x}_1 = \ddot{x}_n. \tag{A.51}$$

The new resulting $(n-1) \times (n-1)$ matrix equation is given by

$$A\dot{\mathbf{x}} = \mathbf{b}, \tag{A.52}$$

where

$$A = \begin{bmatrix} 2(h_{n-1}+h_1) & h_{n-1} & & & & h_1 \\ h_2 & 2(h_1+h_2) & h_1 & & & \\ & \ddots & \ddots & \ddots & & \\ & & & h_{n-2} & 2(h_{n-3}+h_{n-2}) & h_{n-3} \\ h_{n-2} & & & & h_{n-1} & 2(h_{n-2}+h_{n-1}) \end{bmatrix},$$

$$\dot{\mathbf{x}} = \begin{bmatrix} \dot{x}_1 \\ \dot{x}_2 \\ \vdots \\ \dot{x}_{n-2} \\ \dot{x}_{n-1} \end{bmatrix}, \qquad \mathbf{b} = \begin{bmatrix} 3(h_1\bar{v}_{n-1} + h_{n-1}\bar{v}_1) \\ 3(h_2\bar{v}_1 + h_1\bar{v}_2) \\ \vdots \\ 3(h_{n-2}\bar{v}_{n-3} + h_{n-3}\bar{v}_{n-2}) \\ 3(h_{n-1}\bar{v}_{n-2} + h_{n-2}\bar{v}_{n-1}) \end{bmatrix}. \tag{A.53}$$

Once again, all omitted entries of $A$ are by default zero. Recalling that $\dot{x}_1 \equiv \dot{x}_n$, solving Equation (A.52) yields all $n$ required intermediate velocities $\dot{x}_1, \ldots, \dot{x}_n$ as required.

### A.1.6  Function Fitting

Despite its name, linear least squares is a form of linear regression that can be used to fit many different kinds of functions and surfaces to numerical data—not just linear ones. In this section and the following subsections, we examine the ordinary least squares solution to multiple different regression problems.

For the general case, suppose we are given $N$ different data points $(\mathbf{x}_i, \mathbf{y}_i) \in \mathbb{R}^n \times \mathbb{R}^m$, and wish to fit the linear model

$$\mathbf{y} = \mathbf{x}\boldsymbol{\beta}, \tag{A.54}$$

where $\mathbf{x} \in \mathbb{R}^n$ is the row vector of independent variables, $\mathbf{y} \in \mathbb{R}^m$ is the row vector of dependent variables, and $\boldsymbol{\beta} \in \mathbb{R}^{n \times m}$ is an $n \times m$ matrix containing the parameters of the multidimensional linear model. Notationally, we define

$$\mathbf{y} = (y_1, \ldots, y_m) \in \mathbb{R}^m, \tag{A.55a}$$

$$\mathbf{x} = (x_1, \ldots, x_n) \in \mathbb{R}^n, \tag{A.55b}$$

$$\boldsymbol{\beta} = \begin{bmatrix} \beta_{11} & \cdots & \beta_{1m} \\ \vdots & \ddots & \vdots \\ \beta_{n1} & \cdots & \beta_{nm} \end{bmatrix} \in \mathbb{R}^{n \times m}, \tag{A.55c}$$

where, for example, we can see that the linear model models the $y_1$ parameter using the linear equation

$$y_1 = \beta_{11} x_1 + \cdots + \beta_{n1} x_n. \tag{A.56}$$

For any single of the $N$ data points $(\mathbf{x}_i, \mathbf{y}_i)$, we can identify the residual error $\boldsymbol{\epsilon}_i \in \mathbb{R}^m$ of the linear model in describing the data point as

$$\boldsymbol{\epsilon}_i = \mathbf{y}_i - \mathbf{x}_i \boldsymbol{\beta}. \tag{A.57}$$

The task of ordinary linear least squares regression is to find the matrix of coefficients $\boldsymbol{\beta}$ that minimises the total amount of residual error, i.e. that minimises

$$S(\boldsymbol{\beta}) = \sum_{i=1}^{N} \|\boldsymbol{\epsilon}_i\|^2 = \sum_{i=1}^{N} \|\mathbf{y}_i - \mathbf{x}_i \boldsymbol{\beta}\|^2. \tag{A.58}$$

This is equivalent to finding the linear least squares solution to the matrix equation

$$Y = X\boldsymbol{\beta}, \tag{A.59}$$

where

$$Y = \begin{bmatrix} \leftarrow \mathbf{y}_1 \rightarrow \\ \vdots \\ \leftarrow \mathbf{y}_N \rightarrow \end{bmatrix}, \qquad X = \begin{bmatrix} \leftarrow \mathbf{x}_1 \rightarrow \\ \vdots \\ \leftarrow \mathbf{x}_N \rightarrow \end{bmatrix}, \tag{A.60}$$

and the target function to minimise is

$$S(\boldsymbol{\beta}) = \|Y - X\boldsymbol{\beta}\|_F^2. \tag{A.61}$$

The solution to both these formulations of the ordinary least squares regression task is provably given by

$$\hat{\boldsymbol{\beta}} = \underset{\boldsymbol{\beta}}{\operatorname{argmin}} \, S(\boldsymbol{\beta}) = (X^T X)^{-1} X^T Y. \tag{A.62}$$

The final fitted linear model is then given by

$$\mathbf{y} = \mathbf{x}\hat{\boldsymbol{\beta}}. \tag{A.63}$$

### A.1.6.1 *Line of Best Fit*

One simple application of the described linear least squares method is the calculation of a line of best fit through $N$ given keypoints $(x_i, y_i) \in \mathbb{R}^2$. The linear model that is fitted is given by

$$y = \beta_0 + \beta_1 x, \tag{A.64}$$

and the sum of residual errors that is required to be minimised is correspondingly given by

$$S_1^l(\beta_0, \beta_1) = \sum_{i=1}^N \epsilon_i^2 = \sum_{i=1}^N (y_i - \beta_0 - \beta_1 x_i)^2. \tag{A.65}$$

By defining

$$\mathbf{y}_i = (y_i) \in \mathbb{R}^1, \tag{A.66a}$$

$$\mathbf{x}_i = (1, x_i) \in \mathbb{R}^2, \tag{A.66b}$$

$$\boldsymbol{\beta} = \begin{bmatrix} \beta_0 \\ \beta_1 \end{bmatrix} \in \mathbb{R}^{2 \times 1}, \tag{A.66c}$$

we can see that this problem is in the same form as Equation (A.55), and thus is equivalent to the least squares problem

$$Y = X\boldsymbol{\beta}, \tag{A.67}$$

where

$$Y = \begin{bmatrix} y_1 \\ \vdots \\ y_N \end{bmatrix}, \qquad X = \begin{bmatrix} 1 & x_1 \\ \vdots & \vdots \\ 1 & x_N \end{bmatrix}, \qquad \boldsymbol{\beta} = \begin{bmatrix} \beta_0 \\ \beta_1 \end{bmatrix}. \tag{A.68}$$

Recalling Equation (A.62), the required solution can be derived to be

$$\hat{\beta}_1 = \frac{\sum (x_i - \bar{x})(y_i - \bar{y})}{\sum (x_i - \bar{x})^2}, \qquad \hat{\beta}_0 = \bar{y} - \hat{\beta}_1 \bar{x}, \tag{A.69}$$

where the sums go from $i = 1$ to $N$, and the means $\bar{x}$ and $\bar{y}$ are given by

$$\bar{x} = \tfrac{1}{N} \sum_{i=1}^N x_i, \qquad \bar{y} = \tfrac{1}{N} \sum_{i=1}^N y_i. \tag{A.70}$$

While Equation (A.69) is very computationally efficient in computing the line of best fit, it is not necessarily the most numerically robust approach. A more robust method of solving Equation (A.67) involves calculating the QR decomposition of the matrix $X$, and solving the slightly modified equation

$$R\boldsymbol{\beta} = Q^T Y. \tag{A.71}$$

This modified equation has the same solution as Equation (A.67), but is significantly easier and more numerically stable to solve because $R$ is by definition an upper triangular matrix. In fact, in this special case, $R$ only has three non-zero entries, so calculating the required value of $\hat{\boldsymbol{\beta}}$ is trivial.

If a line of best fit is desired for the m-dimensional data points

$$\mathbf{y}_i = (y_{i1}, \ldots, y_{im}) \in \mathbb{R}^m, \tag{A.72}$$

instead of just for the one-dimensional data points $y_i$, the required result can be calculated by simply applying Equation (A.69) once independently for each of the $m$ output y-parameters. Using the notation

$$\boldsymbol{\beta} = \begin{bmatrix} \beta_{01} & \cdots & \beta_{0m} \\ \beta_{11} & \cdots & \beta_{1m} \end{bmatrix} \in \mathbb{R}^{2 \times m}, \tag{A.73}$$

this statement can be seen to be true by examining the required sum of residuals

$$\begin{aligned} S_m^l(\boldsymbol{\beta}) &= \sum_{i=1}^{N} \|\mathbf{y}_i - \mathbf{x}_i \boldsymbol{\beta}\|^2 \\ &= \sum_{i=1}^{N} \sum_{j=1}^{m} (y_i - \beta_{0j} - \beta_{1j} x_i)^2 \\ &= \sum_{j=1}^{m} S_1^l(\beta_{0j}, \beta_{1j}). \end{aligned} \tag{A.74}$$

Clearly, $S_m^l(\boldsymbol{\beta})$ is minimised by minimising each $S_1^l(\beta_{0j}, \beta_{1j})$ individually.[3] We conclude, as stated, that the output y-parameters can thus be treated independently in the calculation of the various entries of $\hat{\boldsymbol{\beta}}$.

### A.1.6.2 *Weighted Line of Best Fit*

If non-negative weights $w_i \geq 0$ are specified for each data point $(x_i, y_i) \in \mathbb{R}^2$, and the sum of residuals minimisation function from Equation (A.65) is modified to be

$$S_1^w(\beta_0, \beta_1) = \sum_{i=1}^{N} w_i \epsilon_i^2 = \sum_{i=1}^{N} w_i (y_i - \beta_0 - \beta_1 x_i)^2, \tag{A.75}$$

then the required equations for $\hat{\beta}_0$ and $\hat{\beta}_1$ become

$$\hat{\beta}_1 = \frac{\sum w_i (x_i - \bar{x})(y_i - \bar{y})}{\sum w_i (x_i - \bar{x})^2}, \qquad \hat{\beta}_0 = \bar{y} - \hat{\beta}_1 \bar{x}, \tag{A.76}$$

where $\bar{x}$ and $\bar{y}$ are now the *weighted* means

$$\bar{x} = \frac{\sum w_i x_i}{\sum w_i}, \qquad \bar{y} = \frac{\sum w_i y_i}{\sum w_i}. \tag{A.77}$$

Clearly, if all weights $w_i$ are equal, Equations (A.76) and (A.77) reduce to Equations (A.69) and (A.70) respectively.

---

3  Due to the independence of the entries of $\boldsymbol{\beta}$, this is always simultaneously possible.

### A.1.6.3 *Orthogonal Line of Best Fit*

It can be seen from Equations (A.65) and (A.75) that the residual errors considered so far (i.e. $\epsilon_i$) have all been discrepancies measured solely along the y-axis. While this is ideal for fitting linear models to temporal data, i.e. applications where $x_i$ corresponds to the timestamps of the data points $y_i$, it is not suitable for cases where, for instance, a line needs to be fit to 2D spatial data. In this case, deviations in both the x and y-axes need to be considered, i.e.

$$\epsilon_i = y_i - \hat{y}_i, \tag{A.78a}$$

$$\eta_i = x_i - \hat{x}_i, \tag{A.78b}$$

where $(\hat{x}_i, \hat{y}_i)$ is the point on the line $y = \beta_0 + \beta_1 x$ that $(x_i, y_i)$ is predicted to most closely correspond to. For this we can formulate the residual error function

$$S_1^d(\beta_0, \beta_1) = \sum_{i=1}^{N} \epsilon_i^2 + \delta\eta_i^2 = \sum_{i=1}^{N} (y_i - \hat{y}_i)^2 + \delta(x_i - \hat{x}_i)^2, \tag{A.79}$$

where $\delta \in [0, \infty)$ is an appropriate fixed scaling factor. The calculation of the parameters $\hat{\beta}_0$ and $\hat{\beta}_1$ that minimise Equation (A.79) is known as Deming regression, and is a special case of total least squares regression. If $\delta = 1$, i.e. if both the x and y-deviations are equally weighted in the residual error function, the problem is referred to as orthogonal regression, and is equivalent to the task of finding the 2D line of best fit through planar data that minimises the perpendicular distances from the data points to the line.[4]

The general solution to the orthogonal minimisation problem described in Equation (A.79) is given by

$$\hat{\beta}_1 = \frac{(s_{yy} - \delta s_{xx}) + \sqrt{(s_{yy} - \delta s_{xx})^2 + 4\delta s_{xy}^2}}{2s_{xy}}, \tag{A.80a}$$

$$\hat{\beta}_0 = \bar{y} - \hat{\beta}_1 \bar{x}, \tag{A.80b}$$

where

$$s_{xx} = \sum_{i=1}^{N} (x_i - \bar{x})^2, \qquad s_{yy} = \sum_{i=1}^{N} (y_i - \bar{y})^2, \tag{A.81}$$

$$s_{xy} = \sum_{i=1}^{N} (x_i - \bar{x})(y_i - \bar{y}). \tag{A.82}$$

If, instead of $y = \beta_0 + \beta_1 x$, the linear model

$$ax + by + c = 0 \tag{A.83}$$

---

4 Or more accurately, minimises the sum of the squared perpendicular distances from the data points to the line.

is used instead, a slightly more numerically robust formulation of the orthogonal regression solution can be expressed as

$$\hat{a} = \begin{cases} (s_{yy} - \delta s_{xx}) + \sqrt{(s_{yy} - \delta s_{xx})^2 + 4\delta s_{xy}^2} & \text{if } s_{yy} \geq \delta s_{xx}, \\ -2s_{xy} & \text{if } s_{yy} < \delta s_{xx}, \end{cases} \quad \text{(A.84a)}$$

$$\hat{b} = \begin{cases} -2s_{xy} & \text{if } s_{yy} \geq \delta s_{xx}, \\ (s_{xx} - \frac{1}{\delta} s_{yy}) + \sqrt{(s_{xx} - \frac{1}{\delta} s_{yy})^2 + \frac{4}{\delta} s_{xy}^2} & \text{if } s_{yy} < \delta s_{xx}, \end{cases} \quad \text{(A.84b)}$$

$$\hat{c} = -\hat{a}\bar{x} - \hat{b}\bar{y}. \quad \text{(A.84c)}$$

Note that these equations are perfectly valid even for $\delta = 0$ and $\delta = \infty$, and that $\delta = \infty$ corresponds to the line of best fit calculated using ordinary least squares in Appendix A.1.6.1. Given $\hat{a}$, $\hat{b}$ and $\hat{c}$, the fitted point $(\hat{x}_i, \hat{y}_i)$ that corresponds to any particular data point $(x_i, y_i)$ is given by

$$(\hat{x}_i, \hat{y}_i) = (x_i - \hat{a}\lambda, \, y_i - \delta\hat{b}\lambda), \quad \text{(A.85)}$$

where

$$\lambda = \frac{\hat{a}x_i + \hat{b}y_i + \hat{c}}{\hat{a}^2 + \delta\hat{b}^2}. \quad \text{(A.86)}$$

For $\delta = 1$, this corresponds to finding the closest point on the fitted line to each data point.

### A.1.6.4  *Ordinary Least Squares for Arbitrary Matrix Equations*

We recall from Equations (A.59) to (A.62) that the ordinary least squares solution to an arbitrary matrix equation

$$X\boldsymbol{\beta} = Y \quad \text{(A.87)}$$

that minimises

$$S(\boldsymbol{\beta}) = \|Y - X\boldsymbol{\beta}\|_F^2 \quad \text{(A.88)}$$

is given by

$$\hat{\boldsymbol{\beta}} = (X^T X)^{-1} X^T Y. \quad \text{(A.89)}$$

Note that the residual vectors $\boldsymbol{\epsilon}_i$ in this case are equal to the row vectors of the matrix $Y - X\boldsymbol{\beta}$, that is,

$$Y - X\boldsymbol{\beta} = \begin{bmatrix} \leftarrow \boldsymbol{\epsilon}_1 \rightarrow \\ \vdots \\ \leftarrow \boldsymbol{\epsilon}_N \rightarrow \end{bmatrix}, \quad \text{(A.90)}$$

and that we can alternatively write

$$S(\boldsymbol{\beta}) = \sum_{i=1}^{N} \|\boldsymbol{\epsilon}_i\|^2, \quad \text{(A.91)}$$

where $N$ is the number of rows in the matrix equation.

A weighted residual error function $S^w(\boldsymbol{\beta})$ can be used instead of $S(\boldsymbol{\beta})$ if desired. If we define

$$S^w(\boldsymbol{\beta}) = \sum_{i=1}^{N} w_i \|\boldsymbol{\epsilon}_i\|^2 = \left\| W^{\frac{1}{2}}(Y - X\boldsymbol{\beta}) \right\|_F^2, \tag{A.92}$$

where $W$ is the diagonal matrix containing the required weights $w_i \geq 0$, the corresponding weighted least squares solution can be derived to be

$$\hat{\boldsymbol{\beta}} = (X^TWX)^{-1}X^TWY. \tag{A.93}$$

This is equivalent to simply scaling each of the rows of the matrix equation $X\boldsymbol{\beta} = Y$ by the corresponding value of $\sqrt{w_i}$. Expressed more formally, this more precisely corresponds to calculating the standard unweighted solution to the slightly modified matrix equation

$$\left(W^{\frac{1}{2}}X\right)\boldsymbol{\beta} = \left(W^{\frac{1}{2}}Y\right). \tag{A.94}$$

### A.1.6.5   *Plane of Best Fit*

If we have a collection of $N$ 3D data points $x_i \in \mathbb{R}^3$, and wish to fit a plane

$$ax + by + cz + d = 0, \tag{A.95}$$

then we cannot do this by calculating the least squares solution to

$$\begin{bmatrix} \mathbf{x}_1 & 1 \\ \vdots & \vdots \\ \mathbf{x}_N & 1 \end{bmatrix} \begin{bmatrix} a \\ b \\ c \\ d \end{bmatrix} = \begin{bmatrix} 0 \\ \vdots \\ 0 \end{bmatrix}, \tag{A.96}$$

as this just results in $a = b = c = d = 0$. As an alternative, we can observe that the normal vector $(\hat{a}, \hat{b}, \hat{c})$ that defines the plane of best fit corresponds to the direction of minimum variance of the data points. This direction is given by the left-singular vector (i.e. column vector of $U$) that corresponds to the smallest singular value of

$$\tilde{X} = \begin{bmatrix} \leftarrow \mathbf{x}_1 - \bar{\mathbf{x}} \rightarrow \\ \vdots \\ \leftarrow \mathbf{x}_N - \bar{\mathbf{x}} \rightarrow \end{bmatrix} \tag{A.97}$$

as given by the Singular Value Decomposition (SVD)

$$\tilde{X} = U\Sigma V^*. \tag{A.98}$$

The remaining value of $\hat{d}$ can then be calculated using

$$\hat{d} = -(\hat{a}, \hat{b}, \hat{c}) \cdot \bar{\mathbf{x}}. \tag{A.99}$$

### A.1.6.6  Circle/Sphere of Best Fit

A general circle

$$(x - c_x)^2 + (y - c_y)^2 = r^2 \tag{A.100}$$

can be fit through the 2D data points $(x_i, y_i) \in \mathbb{R}^2$, by calculating the least squares solution to

$$\underbrace{\begin{bmatrix} x_1 & y_1 & 1 \\ \vdots & \vdots & \vdots \\ x_N & y_N & 1 \end{bmatrix}}_{X} \underbrace{\begin{bmatrix} 2c_x \\ 2c_y \\ r^2 - c_x^2 - c_y^2 \end{bmatrix}}_{\beta} = \underbrace{\begin{bmatrix} x_1^2 + y_1^2 \\ \vdots \\ x_N^2 + y_N^2 \end{bmatrix}}_{Y}. \tag{A.101}$$

This is because substituting $(x_i, y_i)$ into Equation (A.100) yields

$$x_i^2 + y_i^2 = r^2 - c_x^2 - c_y^2 + 2c_x x_i + 2c_y y_i. \tag{A.102}$$

Given the resulting calculated value of $\hat{\boldsymbol{\beta}} = (\hat{\beta}_1, \hat{\beta}_2, \hat{\beta}_3)^T$, the required centre and radius of the circle of best fit can be deduced to be

$$\hat{c}_x = \tfrac{1}{2}\hat{\beta}_1, \qquad\qquad \hat{r} = \sqrt{\hat{\beta}_3 + \hat{c}_x^2 + \hat{c}_y^2}. \tag{A.103}$$
$$\hat{c}_y = \tfrac{1}{2}\hat{\beta}_2,$$

If a weighted circle of best fit is desired, the standard adaptation described in Appendix A.1.6.4 can be applied to achieve this.

Analogous methods can be used to fit spheres and n-dimensional hyperspheres to higher-dimensional data. If we wish to fit the sphere

$$\|\mathbf{x} - \mathbf{c}\|^2 = r^2 \tag{A.104}$$

to the $N$ data points $\mathbf{x}_i \in \mathbb{R}^n$, this requires finding the least squares solution to

$$\begin{bmatrix} \mathbf{x}_1 & 1 \\ \vdots & \vdots \\ \mathbf{x}_N & 1 \end{bmatrix} \begin{bmatrix} \uparrow \\ 2\mathbf{c} \\ \downarrow \\ r^2 - \|\mathbf{c}\|^2 \end{bmatrix} = \begin{bmatrix} \|\mathbf{x}_1\|^2 \\ \vdots \\ \|\mathbf{x}_N\|^2 \end{bmatrix}. \tag{A.105}$$

The required values of $\hat{\mathbf{c}}$ and $\hat{r}$ can then be calculated similarly to before.

### A.1.6.7  Ellipse/Ellipsoid of Best Fit

The methods from Appendix A.1.6.6 can be extended to handle the more general case of ellipse and ellipsoid fitting. Given any set of 2D data points, the mean $(\bar{x}, \bar{y})$ of the points is first computed, and then subtracted from each of them. Suppose that the resulting set of zero-mean data points is given by $(x_i, y_i) \in \mathbb{R}^2$ for $i = 1, \dots, N$. We wish to fit to these data points the general ellipse equation

$$ax^2 + by^2 + 2cxy + 2dx + 2ey = 1. \tag{A.106}$$

This is done by finding the least squares solution to the matrix equation

$$\begin{bmatrix} x_1^2 & y_1^2 & 2x_1y_1 & 2x_1 & 2y_1 \\ \vdots & \vdots & \vdots & \vdots & \vdots \\ x_N^2 & y_N^2 & 2x_Ny_N & 2x_N & 2y_N \end{bmatrix} \begin{bmatrix} a \\ b \\ c \\ d \\ e \end{bmatrix} = \begin{bmatrix} 1 \\ \vdots \\ 1 \end{bmatrix}. \tag{A.107}$$

The centre of the thereby-defined ellipse of best fit (through the original non-zero mean data) can be calculated to be

$$\begin{bmatrix} c_x \\ c_y \end{bmatrix} = \begin{bmatrix} \bar{x} \\ \bar{y} \end{bmatrix} - \begin{bmatrix} \hat{a} & \hat{c} \\ \hat{c} & \hat{b} \end{bmatrix}^{-1} \begin{bmatrix} \hat{d} \\ \hat{e} \end{bmatrix}. \tag{A.108}$$

Although this ellipse of best fit is accurate in most situations, for reasons of total numerical robustness it is recommendable nonetheless to refit the ellipse with a constrained centre to get the best and most consistent results. This is done by calculating the relative offsets $(\tilde{x}_i, \tilde{y}_i) \in \mathbb{R}^2$ of the original data points from the now-constrained centre point $(c_x, c_y)$, and fitting the equation

$$a\tilde{x}^2 + b\tilde{y}^2 + 2c\tilde{x}\tilde{y} = 1. \tag{A.109}$$

This is done by finding the least squares solution to the matrix equation

$$\begin{bmatrix} \tilde{x}_1^2 & \tilde{y}_1^2 & 2\tilde{x}_1\tilde{y}_1 \\ \vdots & \vdots & \vdots \\ \tilde{x}_N^2 & \tilde{y}_N^2 & 2\tilde{x}_N\tilde{y}_N \end{bmatrix} \begin{bmatrix} a \\ b \\ c \end{bmatrix} = \begin{bmatrix} 1 \\ \vdots \\ 1 \end{bmatrix}. \tag{A.110}$$

The final fitted ellipse can then either be expressed as

$$\hat{a}(x - c_x)^2 + \hat{b}(y - c_y)^2 + 2\hat{c}(x - c_x)(y - c_y) = 1, \tag{A.111}$$

or more succinctly as

$$(\mathbf{x} - \mathbf{c})^T A (\mathbf{x} - \mathbf{c}) = 1, \tag{A.112}$$

where

$$\mathbf{x} = \begin{bmatrix} x \\ y \end{bmatrix}, \qquad \mathbf{c} = \begin{bmatrix} c_x \\ c_y \end{bmatrix}, \qquad A = \begin{bmatrix} \hat{a} & \hat{c} \\ \hat{c} & \hat{b} \end{bmatrix}. \tag{A.113}$$

The case of ellipsoid fitting through 3D data is analogous to the case of ellipse fitting through 2D data, only that the fitted equation becomes

$$ax^2 + by^2 + cz^2 + 2dxy + 2exz + 2fyz + 2gx + 2hy + 2iz = 1. \tag{A.114}$$

After calculation of the required centre point

$$\begin{bmatrix} c_x \\ c_y \\ c_z \end{bmatrix} = \begin{bmatrix} \bar{x} \\ \bar{y} \\ \bar{z} \end{bmatrix} - \begin{bmatrix} \hat{a} & \hat{d} & \hat{e} \\ \hat{d} & \hat{b} & \hat{f} \\ \hat{e} & \hat{f} & \hat{c} \end{bmatrix}^{-1} \begin{bmatrix} \hat{g} \\ \hat{h} \\ \hat{i} \end{bmatrix}, \tag{A.115}$$

the simplified equation that is refitted is then

$$a\tilde{x}^2 + b\tilde{y}^2 + c\tilde{z}^2 + 2d\tilde{x}\tilde{y} + 2e\tilde{x}\tilde{z} + 2f\tilde{y}\tilde{z} = 1. \tag{A.116}$$

Solving a least squares matrix equation similar to Equation (A.110) yields the final required values of $\hat{a}$ to $\hat{f}$, and ultimately the equation of the final <span style="color:magenta">ellipsoid of best fit</span>, namely

$$(\mathbf{x} - \mathbf{c})^T A (\mathbf{x} - \mathbf{c}) = 1, \tag{A.117}$$

where

$$\mathbf{x} = \begin{bmatrix} x \\ y \\ z \end{bmatrix}, \qquad \mathbf{c} = \begin{bmatrix} c_x \\ c_y \\ c_z \end{bmatrix}, \qquad A = \begin{bmatrix} \hat{a} & \hat{d} & \hat{e} \\ \hat{d} & \hat{b} & \hat{f} \\ \hat{e} & \hat{f} & \hat{c} \end{bmatrix}. \tag{A.118}$$

## A.2    FILTERS

The standard and non-standard filters that are required throughout the remainder of this thesis are detailed in this section.

### A.2.1    FIR Filters

Discrete time filters that exhibit finite durations of response to any arbitrary finite length input are referred to as <span style="color:magenta">Finite Impulse Response</span> (FIR) filters. The general form of a FIR filter of order $N$ is given by the difference equation

$$y[n] = b_0 x[n] + b_1 x[n-1] + \cdots + b_N x[n-N] \tag{A.119a}$$

$$= \sum_{i=0}^{N} b_i x[n-i], \tag{A.119b}$$

where $x$ is the input variable, $y$ is the output variable, and $b_i$ are fixed constants. If $\mathbf{x}, \mathbf{y} \in \mathbb{R}^m$ are vectors, the generalisation of Equation (A.119) to $m$ dimensions is trivially given by

$$\mathbf{y}[n] = b_0 \mathbf{x}[n] + b_1 \mathbf{x}[n-1] + \cdots + b_N \mathbf{x}[n-N]. \tag{A.120}$$

The name 'Finite Impulse Response' comes from the fact that the response of any such filter to a Dirac delta (i.e. impulse) input is zero after at most $N + 1$ samples.

A.2.1.1 *Mean Filter*

One basic example of a FIR filter is the mean filter, which calculates the moving average of its input data. This corresponds to the filter difference equation

$$\mathbf{y}[n] = \tfrac{1}{N+1}\mathbf{x}[n] + \tfrac{1}{N+1}\mathbf{x}[n-1] + \cdots + \tfrac{1}{N+1}\mathbf{x}[n-N]. \qquad \text{(A.121)}$$

The effect of mean filtering can be seen in Figure A.3.

A.2.1.2 *Savitzky-Golay Filter*

A Savitzky-Golay filter is a kind of FIR filter that implicitly fits low-degree polynomials to incoming data using linear least squares, and uses the result of this fitting to calculate smoothed data and data derivative values. Assuming that the input samples are evenly spaced, tables of FIR coefficients can be calculated for various window sizes and polynomial orders. For instance, if 5 data points are used to fit a quadratic or cubic, the Savitzky-Golay filter equations for smoothing, first and second derivatives are respectively given by

$$\mathbf{y}[n] = -\tfrac{3}{35}\mathbf{x}[n] + \tfrac{12}{35}\mathbf{x}[n-1] + \tfrac{17}{35}\mathbf{x}[n-2] + \tfrac{12}{35}\mathbf{x}[n-3] - \tfrac{3}{35}\mathbf{x}[n-4],$$
$$\dot{\mathbf{y}}[n] = \tfrac{2}{10}\mathbf{x}[n] + \tfrac{1}{10}\mathbf{x}[n-1] - \tfrac{1}{10}\mathbf{x}[n-3] - \tfrac{2}{10}\mathbf{x}[n-4], \qquad \text{(A.122)}$$
$$\ddot{\mathbf{y}}[n] = \tfrac{2}{7}\mathbf{x}[n] - \tfrac{1}{7}\mathbf{x}[n-1] - \tfrac{2}{7}\mathbf{x}[n-2] - \tfrac{1}{7}\mathbf{x}[n-3] + \tfrac{2}{7}\mathbf{x}[n-4].$$

Note that if the samples $\mathbf{x}[\cdot]$ are acquired at regular time intervals of $\Delta t$, we can convert the resulting values of $\dot{\mathbf{y}}[n]$ and $\ddot{\mathbf{y}}[n]$ into derivatives with respect to time by scaling them appropriately:

$$\frac{d\mathbf{x}}{dt}[n] \approx \frac{\dot{\mathbf{y}}[n]}{\Delta t}, \qquad \text{(A.123a)}$$

$$\frac{d^2\mathbf{x}}{dt^2}[n] \approx \frac{\ddot{\mathbf{y}}[n]}{\Delta t^2}. \qquad \text{(A.123b)}$$

A plot of the sample output of a Savitzky-Golay smoothing filter of order 8 (i.e. with a window size of 9) is shown in Figure A.3.

A.2.2 **IIR Filters**

Infinite Impulse Response (IIR) filters are discrete time filters that in general exhibit an infinitely long non-zero response waveform to any impulse input (Dirac delta function). The general form of an IIR filter is given by the difference equation[5]

$$y[n] = b_0 x[n] + b_1 x[n-1] + \cdots + b_N x[n-N]$$
$$\qquad - a_1 y[n-1] - \cdots - a_M y[n-M]. \qquad \text{(A.124)}$$

---

5 Note here that we are simply assuming $a_0 = 1$ to simplify things.

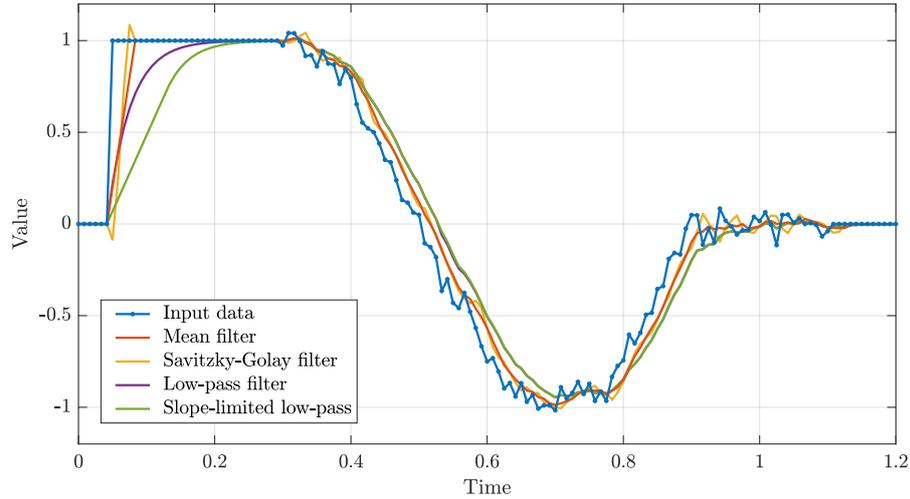

Figure A.3: Comparison of the responses of various FIR and IIR low-pass filters to a generated sample input (blue). Both the mean and Savitzky-Golay filters have a window size of 5—hence why they both exhibit ≈2 samples worth of delay in their output. Both IIR low-pass filters use the same value of $\alpha = 0.22$. One can observe that the IIR filter responses are almost identical, except for the step response rise times in the time period $[0.05, 0.25]$. The Savitzky-Golay filter can be observed to be more sensitive to noise than the mean filter for the same window size, and counterintuitively overshoots both before and after its step response rise time.

In generalised vector form, this can be written as

$$\mathbf{y}[n] = \sum_{i=0}^{N} b_i \mathbf{x}[n-i] - \sum_{j=1}^{M} a_j \mathbf{y}[n-j]. \tag{A.125}$$

Although IIR filters in general never return to zero once disturbed by an impulse input, stable IIR filters approach zero relatively quickly in their limit.

### A.2.2.1 First-order Low-pass Filter

One of the simplest IIR filters is the first-order low-pass filter, given by

$$\mathbf{y}[n] = \alpha \mathbf{x}[n] + (1 - \alpha)\mathbf{y}[n-1], \tag{A.126}$$

for some $\alpha \in [0, 1]$. For extreme cases, where for example $\alpha = 1$, the input value is passed straight through as the output, while for example if $\alpha = 0$, the output value remains constant and is unaffected by the input. For values of $\alpha$ in between, the filter interpolates linearly in each cycle from the current output value $\mathbf{y}[n-1]$ towards the new input value $\mathbf{x}[n]$. This can be thought of as applying (in each cycle) the update equation

$$\mathbf{y} \mathrel{+}= \alpha(\mathbf{x} - \mathbf{y}). \tag{A.127}$$

If $\Delta t$ is the sample time of the filter, the 90% settling time $T_s$ for a step input can be calculated to be

$$T_s = \Delta t \, \frac{\log(0.10)}{\log(1-\alpha)}.$$    (A.128)

Conversely, if the desired 90% settling time is given by $T_s$, the required value of $\alpha$ can be calculated using the formula

$$\alpha = 1 - 0.10^{\frac{\Delta t}{T_s}}.$$    (A.129)

The filter is referred to as a first-order low-pass filter as it uses data from at most one cycle ago, and attenuates the high frequency components of a signal (see Figure A.3).

### A.2.2.2    *Slope-limited First-order Low-pass Filter*

In some applications, the smooth noise-filtering behaviour of a first-order low-pass filter is desired, but the characteristic initial sharp transient that is produced is just too severe for the application. In such situations, slope-limited first-order low-pass filters can be used. These are filters that work exactly like standard first-order low-pass filters, but specify a maximum rate of change $\mathbf{s}$ of the generated output. This can be thought of as modifying Equation (A.127) to become

$$\mathbf{y} \mathrel{+}= \operatorname{coerce}\bigl(\alpha(\mathbf{x} - \mathbf{y}), \, -\mathbf{s}\Delta t, \, \mathbf{s}\Delta t\bigr),$$    (A.130)

where $\operatorname{coerce}(\cdot, \cdot, \cdot)$ is as defined in Appendix A.1.2.1 and is applied elementwise to the given vectors. Note that Equation (A.130) reduces to a standard slope limiter filter (see Appendix A.2.4.2) for $\alpha = 1$. An example of the effect of low-pass slope-limiting is shown in Figure A.3.

### A.2.3    **Integrator Filters**

While the purpose of the filters presented thus far has been to smooth data, we now present discrete time filters that can be used to smoothly integrate data. The standard integrator filter is given by the difference equation

$$\mathbf{y}[n] = \mathbf{x}[n] + \mathbf{y}[n-1].$$    (A.131)

This kind of integrator essentially computes the sum[6]

$$\mathbf{y}[n] = \mathbf{x}[n] + \mathbf{x}[n-1] + \mathbf{x}[n-2] + \cdots$$    (A.132)

and can be observed to never 'forget' any old data, as each $\mathbf{x}[\cdot]$ contributes equally to all future evaluations of the sum.

---

6 Note that the computed $\mathbf{y}[n]$ value still needs to be scaled by the sample time $\Delta t$ to put it into correct temporal units.

### A.2.3.1 *Exponentially Weighted Integrator*

To combat the limitless retention of old data that occurs in the standard integrator filter, weights can be introduced to bias the overall desired sum. If we modify Equation (A.132) to include successive powers of some $\alpha \in [0, 1]$, we arrive at the formulation of the Exponentially Weighted (EW) integrator. This is a kind of 'leaky integrator' that computes the sum

$$\mathbf{y}[n] = \mathbf{x}[n] + \alpha\mathbf{x}[n-1] + \alpha^2\mathbf{x}[n-2] + \cdots \tag{A.133}$$

In terms of incremental updates, this is equivalent to recursively applying the difference equation

$$\mathbf{y}[n] = \mathbf{x}[n] + \alpha\mathbf{y}[n-1]. \tag{A.134}$$

The constant $\alpha$ is referred to as the history time constant. If $\alpha = 0$, the integrator simply returns the last data value, and if $\alpha = 1$ the output is the same as that of a standard integrator. Effectively, the value of $\alpha$ just trims the amount of 'memory' that the integrator has. A suitable value for $\alpha$ is computed from the desired half-life time $T_h$, which is a measure of the decay time of the integrator, using the formula

$$\alpha = 0.5^{\frac{\Delta t}{T_h}}, \tag{A.135}$$

where $\Delta t$ is the sample time. Note that in order for $\mathbf{y}[n]$ to correspond to the true temporal integral of the input data, it first needs to be scaled by $\Delta t$.

### A.2.3.2 *Elliptically Bounded Integrator*

An elliptically bounded integrator is a type of integrator defined specifically for 2D data that inherently soft-coerces the integrated output data to an elliptical region in Cartesian space (see Appendix A.1.2.4). By directly incorporating the coercion step into the filter, problems such as for example integrator windup can be mitigated. In each cycle, the integrated value is first updated using a trapezoidal integration approach:

$$\tilde{\mathbf{y}}[n] = \mathbf{y}[n-1] + \tfrac{1}{2}(\mathbf{x}[n] + \mathbf{x}[n-1]) \tag{A.136}$$

Elliptical soft coercion as per Equation (A.12) is then applied to give the final value of $\mathbf{y}[n]$.

### A.2.4 **Other Filters**

The remaining miscellaneous filters that are required by this thesis are detailed in this section.

### A.2.4.1  *Weighted Line of Best Fit Filter*

The weighted linear regression method presented in Appendix A.1.6.2 can be applied directly to temporal data for the purpose of smoothing and derivative estimation. The result is the so-called Weighted Line of Best Fit (WLBF) filter. The last $N$ data points of the input signal are kept in a cyclic buffer, and used in each cycle to compute the line of best fit, which in turn is evaluated to give the required smoothed position and velocity data. The output position can be calculated by evaluating the line either at the latest data timestamp, or at the mean of the input timestamps.[7] The former approach gives more 'responsive' output position data, but is prone to tracking biases and overshooting. The latter approach, on the other hand, is useful when it is important that the estimated position and velocity correspond closely to each other, and have a matching amount of inherent delay. One advantage of WLBF filters is that they can innately and robustly deal with large irregularities in the timestamps of the acquired data points.

### A.2.4.2  *Slope Limiter*

Slope limiter filters simply apply a limit to the absolute rate of change of a data signal. If $\mathbf{s}$ is the required limit, $\mathbf{x}$ is the input data, $\mathbf{y}$ is the output data, and $\Delta t$ is the time increment of a particular step, the corresponding slope limiter update equation can be expressed as

$$\mathbf{y} \mathrel{+}= \operatorname{coerce}\big(\mathbf{x} - \mathbf{y},\; -\mathbf{s}\Delta t,\; \mathbf{s}\Delta t\big). \tag{A.137}$$

Note that low-pass filtering can easily be added to a slope limiter by adding a multiplicative constant inside the coercion function, as demonstrated in Equation (A.130).

### A.2.4.3  *Time Hold Filter*

There are two types of time hold filters, the maximum time hold filter and the minimum time hold filter. Both variants return the corresponding extremum over the last $N$ cycles of the input data points. These types of filters are useful when short spikes of activation need to be captured and extended into longer activations of a particular output. An example of the effect of both types of time hold filter is shown in Figure A.4.

### A.2.4.4  *Spike Filter*

When signals are not reliable, and are known to contain the possibility of sudden large errant spikes, a spike filter can be used to remove these outliers. Similar to the approach used by slope limiters (see

---

7  Note that evaluating the position at the weighted mean of the buffered timestamps is equivalent to calculating the weighted mean of the position data, resulting in something akin to a weighted mean filter.

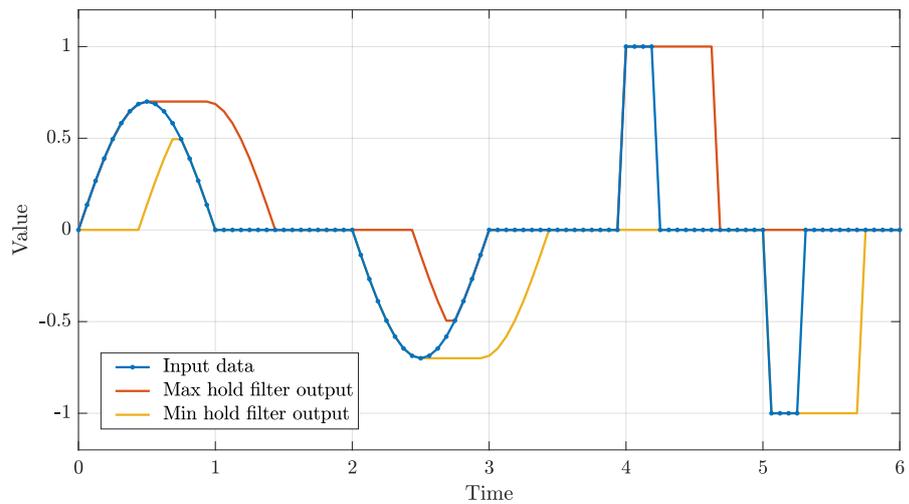

Figure A.4: Plot illustrating the effect of maximum and minimum time hold filters on sample input data. It can be seen that the output of the maximum hold filter (red) prolongs positive peaks and shortens negative troughs, while the output of the minimum hold filter (yellow) does exactly the opposite.

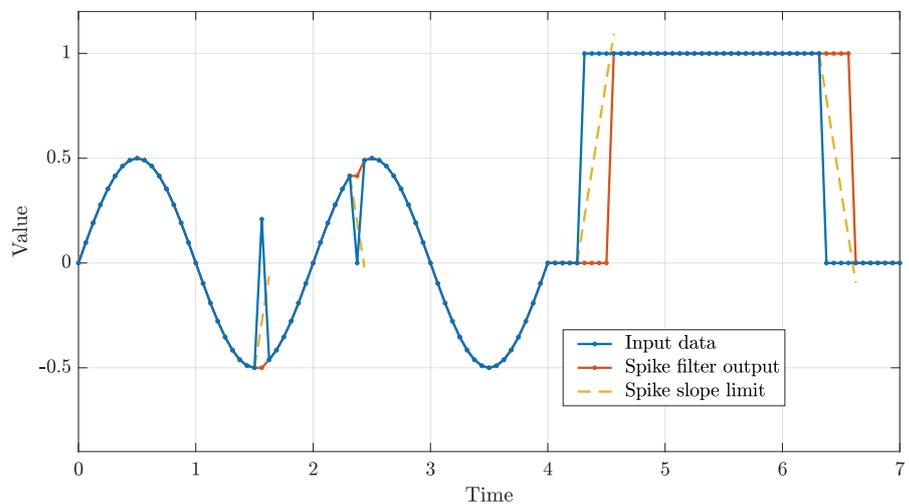

Figure A.5: Plot illustrating the effect of a spike filter on sample input data. Whenever the input (solid blue) jumps further than is allowed by the slope limit (dashed yellow), the filter output (solid red) freezes at its current value and waits until the input is once again in range. Note that at all other times the spike filter just passes the input value through as the output value.

---

**Algorithm A.2:** Spike filter

---

**Input:** Input value $x_n$, last output value $y_{n-1}$, time step $\Delta t$, current hold time $t_h$, absolute slope limit $s$

**Output:** Spike-filtered output value $y_n$, new hold time $t_h$

▷ Note that $y_0$ should be initialised to the value of $x_1$ (the first data point), and $t_h$ should be initialised to zero

1: **function** SPIKEFILTER($x_n$, $y_{n-1}$, $\Delta t$, $t_h$, $s$)
2:     **if** $|x_n - y_{n-1}| > (t_h + \Delta t)s$ **then**
3:         $y_n \leftarrow y_{n-1}$
4:         $t_h \leftarrow t_h + \Delta t$
5:     **else**
6:         $y_n \leftarrow x_n$
7:         $t_h \leftarrow 0$
8:     **end if**
9:     **return** $y_n$, $t_h$
10: **end function**

---

Appendix A.2.4.2), an absolute slope limit $s \in \mathbb{R}^+$ is defined, and used to decide whether an incoming data point is accepted or not. In the case of spike filters, if the slope limit is violated, the output data value is not updated (the value from the previous iteration is retained), and the data point from the following iteration is once again compared to the last accepted data point to see whether it is within the allowed slope limit. The spike filter algorithm is summarised in Algorithm A.2, and plotted in Figure A.5 for sample data.

# B

SUPPORTING PROOFS

---

This appendix contains the proofs of various important non-trivial results that are introduced and used throughout this thesis without proof. The proofs are organised into sections by part and by theme, and are given as follows:

## B.1 ROTATION FORMALISMS IN 3D

This section includes a collection of proofs that are needed for Part III.

### B.1.1 Rotation Representations

#### B.1.1.1 *Proof of Sine Sum Criterion Equivalence*

**See:** Section 5.4.4, Equations (5.60) and (5.61)

**Result:** Given that $\theta, \phi \in [-\frac{\pi}{2}, \frac{\pi}{2}]$, the following two expressions of the sine sum criterion are equivalent:

$$\sin^2 \theta + \sin^2 \phi \leq 1 \iff |\theta| + |\phi| \leq \frac{\pi}{2} \tag{B.1}$$

**Proof:** Keeping in mind that $\theta, \phi \in [-\frac{\pi}{2}, \frac{\pi}{2}]$,

$$\sin^2 \theta + \sin^2 \phi \leq 1 \iff \sin^2 \theta \leq 1 - \sin^2 \phi = \cos^2 \phi = \sin^2(\tfrac{\pi}{2} - \phi)$$

$$\iff |\sin \theta| \leq \left|\sin(\tfrac{\pi}{2} - \phi)\right|$$

$$\iff |\sin \theta| \leq \sin(\tfrac{\pi}{2} - \phi) \text{ as } 0 \leq \tfrac{\pi}{2} - \phi \leq \pi$$

$$\sin(\tfrac{\pi}{2} - \phi) = \sin(\pi - (\tfrac{\pi}{2} - \phi)) = \sin(\tfrac{\pi}{2} + \phi)$$

$$\implies \sin(\tfrac{\pi}{2} - \phi) = \sin(\tfrac{\pi}{2} - |\phi|)$$

$$|\sin \theta| = \begin{cases} \sin \theta & \text{if } \theta \in [0, \tfrac{\pi}{2}] \\ -\sin \theta & \text{if } \theta \in [-\tfrac{\pi}{2}, 0) \end{cases} = \begin{cases} \sin \theta & \text{if } \theta \in [0, \tfrac{\pi}{2}] \\ \sin(-\theta) & \text{if } \theta \in [-\tfrac{\pi}{2}, 0) \end{cases} = \sin|\theta|$$

Thus, the original criterion is equivalent to

$$\sin|\theta| \leq \sin(\tfrac{\pi}{2} - |\phi|) \iff |\theta| \leq \tfrac{\pi}{2} - |\phi| \text{ as } 0 \leq |\theta|, \tfrac{\pi}{2} - |\phi| \leq \tfrac{\pi}{2}$$
$$\iff |\theta| + |\phi| \leq \tfrac{\pi}{2}$$

This completes the proof.    □

### B.1.1.2  *Proof of Smoothness of Tilt Phase Space*

**See:**  Section 5.4.5.1

**Result:**  The tilt phase space is well-defined in terms of the tilt angles parameters $T = (\psi, \gamma, \alpha)$, and is unique for every rotation away from the fused yaw singularity $\alpha = \pi$.

**Proof:**  From Equation (5.62), the definition of the tilt phase space for the tilt rotation $(\gamma, \alpha)$ is given by $P = (\alpha \cos\gamma, \alpha \sin\gamma, \psi)$. Tilt angles have the equivalences $(\gamma, \alpha) \equiv (\gamma + 2\pi k, \alpha) \equiv (\gamma + \pi, -\alpha)$ and $(\gamma, 0) \equiv (0, 0)$ for $k \in \mathbb{Z}$, so in order to be well-defined, the corresponding tilt phase parameters must in each case be equal. Note that only $|\alpha| < \pi$ and tilt rotations are relevant here, as $\alpha = \pm\pi$ corresponds to the fused yaw singularity, and the fused yaw parameter is trivially the same between tilt angles and the tilt phase space.

The tilt rotation $(\gamma + 2\pi k, \alpha)$ corresponds to $(\gamma, \alpha)$ as

$$(\alpha \cos(\gamma + 2\pi k), \alpha \sin(\gamma + 2\pi k), \psi) = (\alpha \cos\gamma, \alpha \sin\gamma, \psi) = P \checkmark$$

The tilt rotation $(\gamma + \pi, -\alpha)$ corresponds to $(\gamma, \alpha)$ as

$$((-\alpha)\cos(\gamma + \pi), (-\alpha)\sin(\gamma + \pi), \psi) = (\alpha\cos\gamma, \alpha\sin\gamma, \psi) = P \checkmark$$

The tilt rotation $(\gamma, 0)$ corresponds to $(0, 0)$ as

$$(0\cos\gamma, 0\sin\gamma, \psi) = (0, 0, \psi) = (0\cos 0, 0\sin 0, \psi) \checkmark$$

Thus, the tilt phase space is well-defined in terms of tilt angles.    □

**Result:**  The tilt phase space parameters are infinitely differentiable, i.e. smooth $\mathcal{C}^\infty$, with respect to the underlying rotation in SO(3), everywhere except at the fused yaw singularity $\alpha = \pi$.

**Proof:**  Note that it is insufficient to say that $\alpha\cos\gamma$ and $\alpha\sin\gamma$ are smooth functions of $\alpha$ and $\gamma$, because these are not in turn smooth functions of the underlying SO(3) rotation! All nine entries $R_{ij}$ of the corresponding rotation matrix *are* smooth functions of the rotation however, as they correspond to the cosines of the angles between each possible pair of rotated and reference coordinate frame axes. We know that $R_{32} = \sin\alpha\cos\gamma$ and $R_{31} = -\sin\alpha\sin\gamma$, so $p_x = \dfrac{R_{32}}{\operatorname{sinc}\alpha}$ and $p_y = \dfrac{-R_{31}}{\operatorname{sinc}\alpha}$, where $\operatorname{sinc}\alpha = \dfrac{\sin\alpha}{\alpha}$ with $\operatorname{sinc}(0) = 1$ is the cardinal sine function. The $\operatorname{sinc}(\cdot)$ function is smooth, even at $\alpha = 0$, where, as stated, by definition of the removable singularity $\operatorname{sinc}(0) = 1$, so it follows that $p_x$ and $p_y$ are smooth for $\operatorname{sinc}\alpha \neq 0$, which is the case for all $|\alpha| < \pi$.

From Equation (5.45), $p_z = \psi = \text{wrap}(2 \, \text{atan2}(z, w))$, where $w$ and $z$ are the associated quaternion parameters. We first note that we consider $p_z$ to be cyclic (circle topology), so jumps between $\pm\pi$ do not constitute discontinuities and the $\text{wrap}(\cdot)$ function can be ignored for the purposes of determining smoothness. The quaternion parameters are smooth functions of the underlying rotation, and $\text{atan2}(z, w)$ is only non-smooth at $w = z = 0$, which corresponds exactly to the fused yaw singularity. Thus, the fused yaw $p_z = \psi$ is also smooth, as required. □

**Corollary:** The tilt phase space parameters are continuous everywhere away from the fused yaw singularity.

**Proof:** Follows trivially from the parameters being infinitely differentiable. □

### B.1.1.3  *Proof of Additive Effect of Z-rotations on Fused Yaw*

**See:** Section 5.7.3, Equations (5.141) and (5.142)

**Result:** Local and global z-rotations are both purely additive to the fused yaw, and in the case of global z-rotations, do not change the tilt rotation component either.

**Proof:** To see why global (i.e. premultiplication) z-rotations are additive in fused yaw, consider the rotation matrix $R$, of fused yaw $\Psi(R) = \psi$. By Equation (5.131a), we can split $R$ into its yaw and tilt rotation components, giving $R = R_f R_t$. Hence,

$$R_z(\psi_z)R = R_z(\psi_z)R_f R_t = R_z(\psi_z)R_z(\psi)R_t = R_z(\psi + \psi_z)R_t$$
$$\implies \Psi(R_z(\psi_z)R) = \text{wrap}(\psi + \psi_z)$$

and the tilt rotation component $R_t$ remains unchanged, so the tilt angles, fused angles and tilt phase space tilt rotation parameters remain the same. This proves Equation (5.142) and the global half of Equation (5.141). Now consider the locally rotated $RR_z(\psi_z)$. Noting that $\Psi(R^{-1}) = -\Psi(R)$ (see Equation (5.148)) and writing $R^{-1} = \tilde{R}_f \tilde{R}_t$, we can see that

$$(RR_z(\psi_z))^{-1} = R_z(\psi_z)^{-1}R^{-1} = R_z(-\psi_z)\tilde{R}_f \tilde{R}_t$$
$$= R_z(-\psi_z)R_z(-\psi)\tilde{R}_t = R_z(-\psi - \psi_z)\tilde{R}_t$$
$$\implies \Psi(RR_z(\psi_z)) = -\Psi\big((RR_z(\psi_z))^{-1}\big) = -\Psi\big(R_z(-\psi - \psi_z)\tilde{R}_t\big)$$
$$= -\text{wrap}(-\psi - \psi_z) = \text{wrap}(\psi + \psi_z)$$

This proves the remaining local half of Equation (5.141). □

## B.1.2  **Spherical Linear Interpolation**

### B.1.2.1  *Proof of Bi-invariance of Slerp*

**See:** Section 7.3.5.1, Equation (7.213)

**Result:**   Given that $q_0, q_1 \in \mathbb{Q}$ and $u \in \mathbb{R}$, for any quaternion $\hat{q} \in \mathbb{Q}$,

$$\text{slerp}(\hat{q}q_0, \hat{q}q_1, u) = \hat{q}\,\text{slerp}(q_0, q_1, u), \tag{B.2a}$$

$$\text{slerp}(q_0\hat{q}, q_1\hat{q}, u) = \text{slerp}(q_0, q_1, u)\hat{q}. \tag{B.2b}$$

**Proof:**   We recall from Equation (7.207) that

$$\text{slerp}(q_0, q_1, u) = q_0(q_0^{-1}q_1)^u$$

Thus, for any $\hat{q} \in \mathbb{Q}$,

$$\begin{aligned}
\text{slerp}(\hat{q}q_0, \hat{q}q_1, u) &= \hat{q}q_0\big((\hat{q}q_0)^{-1}\hat{q}q_1\big)^u \\
&= \hat{q}q_0\big(q_0^{-1}\hat{q}^{-1}\hat{q}q_1\big)^u \\
&= \hat{q}q_0\big(q_0^{-1}q_1\big)^u \\
&= \hat{q}\,\text{slerp}(q_0, q_1, u)
\end{aligned}$$

Similarly, we know from Equation (7.207) that

$$\text{slerp}(q_0, q_1, u) = (q_1 q_0^{-1})^u q_0$$

Thus, for any $\hat{q} \in \mathbb{Q}$,

$$\begin{aligned}
\text{slerp}(q_0\hat{q}, q_1\hat{q}, u) &= \big(q_1\hat{q}(q_0\hat{q})^{-1}\big)^u q_0\hat{q} \\
&= \big(q_1\hat{q}\hat{q}^{-1}q_0^{-1}\big)^u q_0\hat{q} \\
&= \big(q_1 q_0^{-1}\big)^u q_0\hat{q} \\
&= \text{slerp}(q_0, q_1, u)\hat{q}
\end{aligned}$$

Equation (B.2) is thus true, and we conclude that slerp is bi-invariant.

<div align="right">□</div>

### B.1.2.2   *Proof of Slerp Between Tilt Rotations*

**See:**   Section 7.3.5.3, Equation (7.218)

**Result:**   If $q_0, q_1 \in \mathbb{Q}$ are pure tilt rotations, i.e. they have zero fused yaw, then any slerp interpolation between them is also a pure tilt rotation. That is, for $u \in \mathbb{R}$,

$$\Psi(q_0) = \Psi(q_1) = 0 \implies \Psi\big(\text{slerp}(q_0, q_1, u)\big) = 0. \tag{B.3}$$

**Proof:**   As $q_0$ and $q_1$ are pure tilt rotations, from Equation (5.136) the $z$-components of both quaternions must be zero. If we look now at the equation for slerp,

$$\text{slerp}(q_0, q_1, u) = \left(\frac{\sin\big((1-u)\Omega\big)}{\sin\Omega}\right)q_0 + \left(\frac{\sin u\Omega}{\sin\Omega}\right)q_1,$$

we can see that the $z$-component of $\text{slerp}(q_0, q_1, u)$ must as a consequence also be zero, as it is a pure linear combination of $q_0$ and $q_1$. From Equation (5.136) again, we can directly conclude that $\text{slerp}(q_0, q_1, u)$ must be a tilt rotation, and hence has a fused yaw of zero.

<div align="right">□</div>

# INDEX OF TERMS

This index lists the various terms that are defined and used in this thesis, with the appropriate pages where they are explicitly referenced.

## D

## H

## I

## M

### N

### O

### P

## Q

## R

## S

*A parting of ways…*

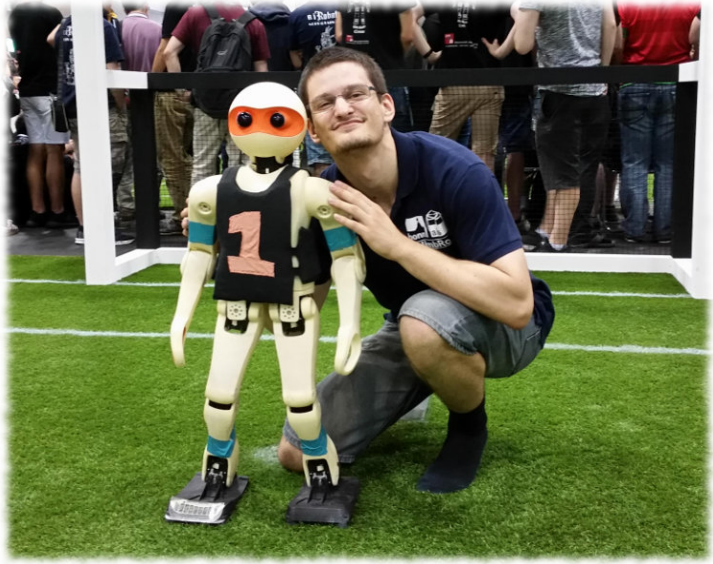

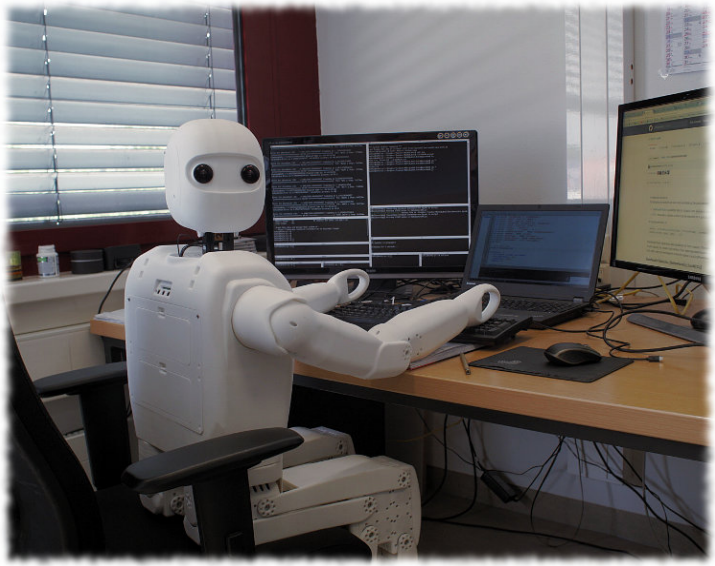

*End of thesis.*